Université Paris Cité

**Frontières de l'Innovation en Recherche et Éducation – ED 474**

***médialab Sciences Po***

# Tu crois que c'est vrai ?

## *Diversité des régimes d'énonciation face aux fake news et mécanismes d'autorégulation conversationnelle*

Par **Manon BERRICHE**

Thèse de doctorat de **Sociologie et sciences de l'éducation**

Dirigée par **Sophie PÈNE**
Et co-dirigée par **Dominique CARDON**

Présentée et soutenue publiquement

le 05/12/2024

Devant un jury composé de :

**Claire BALLEYS**, Professeure associée de sociologie, Université de Genève, *examinatrice*
**Christine BARATS**, Professeure des universités en sciences de l'information et de la communication, Université Paris Cité, *examinatrice*
**Julien BOYADJIAN**, Maître de conférence en sciences politiques, Sciences Po Lille, *examinateur*
**Dominique CARDON**, Professeur associé de sociologie, Sciences Po Paris, *co-directeur de thèse*
**Cyril LEMIEUX**, Directeur d'études en sociologie, EHESS, *examinateur*
**Arnaud MERCIER**, Professeur des universités en sciences de l'information et de la communication, Paris II Panthéon Assas, *rapporteur*
**Dominique PASQUIER**, Directrice de recherche en sociologie au CNRS, Professeure émérite en sociologie, Université Paris Cité, *rapportrice*
**Sophie PÈNE**, Professeure émérite en sciences de l'information et de la communication, Université Paris Cité, *directrice de thèse*

**Titre :** Tu crois que c'est vrai ? Diversité des régimes d'énonciation face aux *fake news* et mécanismes d'autorégulation conversationnelle

**Résumé :** Cette thèse vise à expliquer deux paradoxes : (1) pourquoi la majorité des enquêtes empiriques montre que les *fake news* ne représentent qu'une petite proportion du total d'informations consultées et partagées par les utilisateurs des réseaux sociaux alors que ces derniers ne sont ni soumis à un contrôle éditorial, ni à des règles de déontologie journalistique ? (2) Comment comprendre la montée de la polarisation politique alors que les utilisateurs ne semblent pas si réceptifs aux *fake news* ? Pour répondre à ces questions, deux enquêtes ont été conduites sur Twitter et Facebook. Chacune articule des analyses quantitatives de traces numériques à des observations en ligne et des entretiens. Ce dispositif méthodologique hybride a permis de ne pas réduire les utilisateurs étudiés au fait d'avoir réagi à une *fake news* sur un réseau social particulier et d'examiner la variété de leurs pratiques au sein de différentes situations d'interactions (en ligne comme hors ligne), tout en identifiant certaines de leurs caractéristiques socio-démographiques. La première étude a permis d'identifier l'ensemble des utilisateurs ayant partagé au moins un contenu classé comme une *fake news* par des *fact-checkers* sur la Twittosphère française. En mobilisant un corpus de contenus signalés comme des *fake news* par des utilisateurs de Facebook, la seconde étude a permis de dépasser la seule question de la factualité et d'étudier les réactions des utilisateurs à des énoncés dont la qualité épistémique est incertaine. Trois résultats principaux ressortent de la thèse. Premièrement, le partage de *fake news* est loin d'affecter de façon égale et indifférenciée l'ensemble des utilisateurs des réseaux sociaux, mais n'est en réalité observable que pour un groupe restreint d'internautes dont la particularité n'est pas d'être moins éduqués ou moins dotés en compétences cognitives que les autres, mais d'être davantage politisés et critiques à l'égard des institutions. Bien que minoritaires, ces utilisateurs sont cependant susceptibles de faciliter la mise à l'agenda des opinions défendues par leur camp politique dans le débat public en raison de leur hyperactivité en ligne et des très nombreuses informations d'actualité qu'ils partagent. Deuxièmement, les utilisateurs des réseaux sociaux exposés à des *fake news* sont en mesure de déployer des formes de distance critique de façon plus ou moins importante selon leur position dans l'espace social et les normes d'interactions des situations dans lesquelles ils se trouvent, soit en faisant preuve de prudence énonciative, soit en exprimant des points d'arrêt, c'est-à-dire en intervenant dans le flux d'une conversation pour formuler des désaccords ou des corrections. Troisièmement, ces formes de distance critique permettent rarement l'émergence de véritables débats délibératifs, pas plus que l'expression d'un pluralisme agonistique, mais donnent plutôt lieu à des dialogues de sourds entre une minorité d'utilisateurs particulièrement actifs en ligne. Ces conclusions invitent les futures études académiques, ainsi que le débat public, à se décentrer de la seule question des *fake news* afin de ne pas négliger d'autres troubles de l'information et de la communication comme la manipulation de l'agenda politique ou la brutalisation du débat public par une minorité d'utilisateurs et les mécanismes de spirale du silence qui en découlent.

**Title:** Do you think it's true? Diversity of expression regimes in the face of *fake news* and mechanisms of conversational self-regulation

**Abstract:** This thesis aims to explain two paradoxes: (1) Why do most empirical studies show that *fake news* represents only a small proportion of the total information consulted and shared by social media users, even though they are neither subject to editorial control nor bound by journalistic ethics? (2) How can we understand the rise of political polarization, given that users do not seem to be particularly receptive to *fake news*? To address these questions, two studies were conducted on Twitter and Facebook. Each combines quantitative analyses of digital traces with online observations and interviews. This hybrid methodological approach made it possible not to reduce the users studied to their reaction to a single *fake news* item on a particular social network, but to examine the variety of their practices in different interaction situations (both online and offline), while identifying some of their socio-demographic characteristics. The first study identified all users who shared at least one piece of content classified as *fake news* by fact-checkers in the French Twittersphere. Drawing on a corpus of content flagged as *fake news* by Facebook users, the second study went beyond the issue of factuality to examine users' reactions to statements whose epistemic quality is uncertain. Three main findings emerge from the thesis. First, the sharing of *fake news* does not affect all social media users equally or indiscriminately; it is actually limited to a small group of internet users. These individuals are not less educated or less cognitively skilled than others, but they are more politicized and critical of institutions. Although they are in the minority, these users are likely to facilitate the agenda-setting of their political camp's opinions in the public debate due to their hyperactivity online and the vast amount of news they share. Second, social media users exposed to *fake news* can deploy forms of critical distance to varying degrees, depending on their position in the social space and the interaction norms of the situations they are in, either by exercising "prudence énonciative" (discursive caution) or by expressing "points d'arrêt", i.e., intervening in the flow of a conversation to formulate disagreements or corrections. Third, these forms of critical distance rarely lead to genuine deliberative debates or the expression of agonistic pluralism but instead result in "dialogues of the deaf" among a minority of particularly active online users. These conclusions call for future academic studies, as well as the public debate, to shift their focus away from the sole issue of *fake news*, so as not to overlook other information and communication troubles, such as the manipulation of the political agenda or the brutalization of public debate by a minority of users and the spiral of silence mechanisms that follow.

À la Librairie Presse Papeterie Le Parchemin

Et au Kiosque du métro Jasmin

# Remerciements

Je tiens tout d'abord à remercier du fond du cœur Dominique Cardon et Sophie Pène, mon directeur et ma directrice de thèse. J'ai eu la chance de les avoir tous les deux comme enseignants pendant mes études de master. Je me rends compte que je n'étais pas bien vieille à l'époque – même moi maintenant je me qualifierais de « gamine ». Les rencontrer a été déterminant dans mon parcours académique mais aussi dans ma vie personnelle. Ils m'ont apporté une écoute et un regard très rare leur permettant de voir et d'entendre des intuitions et des émotions que je n'arrivais pas à exprimer à l'époque. Je leur suis très reconnaissante de ne pas s'être arrêtés à mon manque de connaissances, ni à mes propos confus, et d'avoir fait confiance à ma curiosité — bref, d'avoir cru en celle que j'aspirais à devenir plutôt qu'en celle que j'étais. Grâce à leur gentillesse, leur attention, leur disponibilité, leur exigence et leur humilité, je sais aujourd'hui hurler sans faire de bruit.

Je voudrais ensuite remercier chacun et chacune des chercheurs et chercheuses qui ont accepté de faire partie de mon jury — Dominique Pasquier, Arnaud Mercier, Claire Balleys, Christine Barats, Cyril Lemieux et Julien Boyadjian. J'ai beaucoup appris en lisant leurs travaux ou en écoutant leurs présentations à des séminaires, et je suis très honorée qu'ils et elles prennent ce temps pour lire ma thèse, l'évaluer et la commenter. Je me réjouis d'avoir la chance d'échanger avec eux dans le cadre privilégié offert par la soutenance.

J'exprime également toute ma gratitude à tous les chercheurs et à toutes les chercheuses avec lesquel·le·s j'ai eu la chance de travailler ces dernières années. Il y a d'abord évidemment mon ami et co-auteur Sacha Altay — nos discussions, notre correspondance par mail et nos échanges sur Twitter m'ont beaucoup stimulée ; je suis sûre qu'on pourrait les imprimer et en faire un livre pour gorafiser le monde. Il y a aussi, bien entendu, Valentine Crosset avec laquelle j'ai pris beaucoup de plaisir à travailler sur différents projets, ainsi que Pedro Ramaciotti Morales et Aymeric Luneau qui m'ont initiée à écrire en *LaTex* sur *Overleaf*, aidée à mieux comprendre ce que c'était qu'une analyse factorielle, et appris plein d'autres trucs très utiles. Je suis aussi très reconnaissante à Hugo Mercier pour sa « vigilance ouverte », à

Anne-Sophie Hacquin pour son aide technique, ainsi qu'à Alberto Acerbi pour ses recommandations et ses encouragements.

Je tiens à adresser une mention spéciale à Jean-Philippe Cointet. Non seulement j'ai énormément appris à ses côtés — grâce à lui, je n'ai plus peur de manipuler des bases de données de millions de tweets ou de commentaires Facebook sur *Jupyter notebook* ou *Cortext* — mais en plus ses casquettes hautes en couleur, sa bonne humeur, ses farces et ses jeux de mots — sans oublier le prix de consolation qu'on a gagné à Limassol au cours d'une partie de Monopoly grandeur nature — ont contribué à faire de cette thèse une partie de plaisir même dans les moments les plus sombres.

Je suis aussi très reconnaissante à tous les chercheurs et à toutes les chercheuses qui m'ont accordé un peu de leur temps ces dernières années que ce soit en relisant attentivement plusieurs de mes textes, en m'offrant de précieux conseils de lecture, en m'apprenant à utiliser divers outils, en me faisant des retours affutés sur mes travaux ou en me donnant l'opportunité d'enseigner : Celya Gruson-Daniel, Sylvain Parasie, Donato Ricci, Maxime Crépel, Shaden Shabayek, Héloïse Théro, Robin de Mourat, Andrei Mogoutov, Jérémy Ward, Anne Cordier, Bilel Benbouzid, merci à vous !

Parmi tous ces chercheurs et chercheuses, je voudrais adresser une mention spéciale à Thomas Tari qui a gentiment consacré de longues heures à relire et corriger plusieurs de mes textes avec une grande minutie. Je le remercie d'avoir toujours pris le temps, tout au long de cette thèse, de me donner des conseils et de me soutenir. J'ai beaucoup appris grâce à sa finesse, sa rigueur et son humilité.

Mes remerciements vont également à tous les chercheurs et à toutes les chercheuses qui m'ont invitée à présenter mes travaux à des séminaires, des conférences ou des journées d'études : Nathalie Pignard-Cheynel, Vincent Carlino, Julia Cagé, Jérémie Nicey, Marianne Lumeau, Sophie Gebeil, Julien Garric, Mathias Girel, Dominique Boullier, Nicolas Hubé, Christine Balagué, Baptiste Kotras. Ces occasions m'ont permis de confronter mes idées à d'autres perspectives, d'enrichir ma réflexion par des échanges stimulants et de bénéficier de retours constructifs qui ont grandement contribué à nourrir mes recherches.

Je remercie infiniment tou·te·s les ingénieur·e·s de recherche du médialab qui m’ont beaucoup aidée dans les collectes et la gestion de mes données et ont eu la patience de me former à différents outils : Guillaume Plique, Béatrice Mazoyer, Benjamin Ooghe-Tabanou, Audrey Baneyx, Kelly Christensen, Diégo Antolinos-Basso. Bien plus que de simples soutiens techniques, ils et elles m’ont apporté de précieux conseils épistémologiques qui ont grandement contribué à affûter mon regard.

Je tiens aussi à exprimer toute ma gratitude à deux piliers du médialab : Amélie Vairelles et Barbara Bender. Leur disponibilité et leur soutien m’ont permis de bénéficier d’un espace de travail très propice à l’écriture d’une thèse, de participer à quelques conférences à l’étranger, d’organiser des séminaires de recherches dans d’excellentes conditions et de voir mes travaux mis en valeur et diffusés sur le site web du médialab. Je les remercie aussi toutes les deux d’avoir pris le temps de relire certains de mes textes avec beaucoup de finesse et de précision.

Je souhaite également remercier François Taddei, David Tareste, Muriel Mambrini, Élodie Kaslikowski et Camille Gaulon de m'avoir accompagnée tout au long de mon parcours doctoral. Je leur suis particulièrement reconnaissante pour leur engagement à créer un environnement académique stimulant et bienveillant, ainsi que pour leur aide précieuse dans le traitement de toutes mes questions administratives et autres préoccupations.

Je suis également très reconnaissante à la fondation Reboot d’Helen Lee Bouygues de m’avoir accordé une bourse de thèse et d’avoir soutenu mes travaux de recherche.

Je remercie aussi tous les doctorant·e·s du médialab et de son programme doctoral avec lesquel·le·s j’ai pris beaucoup de plaisir à discuter au cours d’ateliers d’écriture, de pauses café, de déjeuner à l’Empire et de conversations WhatsApp : Armin Pournaki, Camille Chanial, Katharina Tittel, Salomé Do, Jules Villa, Jean-Baptiste Garrocq, Tim Faverjon, Valentin Goujon, Benjamin Tainturier, Annina Claesson, Maud Barret Bertelloni, Rubing Shen, Rodrigo Alonso Reyes Cordova, Emma Bonutti D’Agostini, Jean-Marie John-Mathews, Benjamin Gilbert, Axel Meunier, Théophile Lenoir, Héloïse Eloi-Hammer, Benoît Verjat, Florian Cafiero, Duncan Cassells, Ömer Faruk Metin, Sébastien Shulz, Emma Gauthier, Emile Provendier, Valéria Ramirez.

Je remercie plus largement tous les membres actuels du médialab, ainsi que les personnes qui y sont passées de façon temporaire – stagiaire, post-doc, doctorant ou chercheur invité : Émilien Schultz, Pearl Pandya, Hélène Herman, Ana Laura Boeno Malmaceda, Inèse Charlotte Mosgalik, Antoine Sander, Chris Anderson, Michael Bailey, etc.

Je tiens évidement à adresser une pensée à Ysé Vauchez, ma (fausse) jumelle de thèse. J'ai pris beaucoup de plaisir à discuter avec elle et à lire ses travaux. Je la remercie de s'être toujours montrée solidaire et sororale.

Je remercie une seconde fois Amélie Vairelles et Béatrice Mazoyer grâce auxquelles j'ai pu sortir de temps en temps le nez de mes bouquins et de mon ordi pour faire un plouf dans la piscine du Marché Saint Germain.

Je remercie aussi tous les membres et partenaires du projets De Facto avec lesquels j'ai eu la chance d'échanger et de travailler : Ingrid Bertaux, Joyce Zablit, Denis Teyssou, Victor Ebekwumonye, Virginie Sassoon, Elsie Russier, Serge Barbet, Laure Delmoly, Isabelle Feroc-Dumez, Delphine Ramond, Nathan Gallo, Coline Sallois, Christine Buhagiar, Stéphane Laurière, Ludovic Dubost. Cela a été très enrichissant pour moi de discuter à la fois avec des journalistes de l'AFP, des membres du Clemi et d'ExWiki.

Je suis très reconnaissante à plusieurs journalistes, responsables d'associations et personnes engagées dans le débat public d'avoir fait montre d'intérêt pour mes travaux de recherche, de les avoir valorisés et de m'avoir posé des questions pertinentes et stimulantes, me poussant à affiner mes analyses : Xavier de Laporte, Julien Le Bot, Béatrice Kammerer, Julie Charpentrat, Grégoire Lemarchand, Arnaud Muller, Nathalie Puzenat, Alice Bougnères, Paula Forteza, Matti Schneider, Siegrid Henry, Léa Douhard.

Je voudrais aussi remercier plusieurs chercheurs et chercheuses de m'avoir permis de découvrir d'autres disciplines, notamment des travaux de neurosciences ou de psychologie cognitive et sociale : Pauline Martinot, Camille Lakhlifi, Mariam Chammat, Albert Moukheiber, Calliste Scheibling-Sève, Pleen Lejeune, Thibaud Griessinger, Elena Pasquinelli,

Gérald Bronner, Laurent Cordonier, Antoine Marie, Tiffany Morisseau, Melisa Basol, Jon Roozenbeck, Sander Van der Linden, Grégoir Borst, Mathieu Cassotti.

Je suis aussi très reconnaissante à différents membres de l'École d'Affaires Publiques et du LIEPP de Sciences Po, notamment Yann Algan et Bruno Palier, de m'avoir permis de réaliser un mémoire de fin d'études plutôt qu'un stage professionnalisant et d'avoir encouragé ma candidature en thèse.

Je n'oublie pas tous les enquêté·e·s qui ont accepté de répondre à mes messages et qui m'ont accordé leur temps pour réaliser un entretien avec moi.

Un grand merci également à Lise Kiennemann, Jeanne-Louise Rollinger, Justine Nayral, Coralie Degenève, Marine Dupuis, Nancy-wange Moussissa que j'ai eu le privilège d'encadrer avec Dominique Cardon ou Valentine Crosset lors de leur stage d'initiation à la recherche en sociologie – et parfois d'avoir comme étudiante en cours – et qui sont désormais en train de finaliser leur mémoire de fin d'études, de débuter leur thèse ou leur carrière de journaliste ! Échanger avec elles a été très enrichissant pour mon travail de thèse et a renforcé mon goût pour l'enseignement et la transmission.

Je tiens à exprimer ma profonde gratitude à tous les médecins, soignants et aides-soignants qui m'ont soutenue et aidée lorsque j'ai traversé des problèmes de santé et m'ont permis d'être hospitalisée dans de bonnes conditions. Cette thèse leur doit beaucoup.

Je souhaite également remercier plusieurs amis qui ont fortement contribué à aiguiser ma pensée et ma sensibilité ces dernières années — et puis surtout à leur donner des ailes d'alouettes — en me recommandant des films, des chansons et des livres, en partageant avec moi leurs propres travaux et créations, en étant curieux à l'égard de mes projets ou en m'encourageant à écrire : Ada Loueilh, Joëlle Goldberg, Sara Kamidian, Seumboy Vrainom :€, Nicolas Corman, James Altman.

Je voudrais aussi adresser une pensée espiègle à Nicolas Fargues grâce auquel mon écriture a pu dérailler vers des chemins taquins, vagabonder dans des labyrinthes retors et s'amuser à jouer aux poupées russes avec un miroir déformant.

Enfin, je remercie profondément mes parents, mon grand-père et mon grand-frère pour leur amour inconditionnel et leur soutien indéfectible, ainsi que toute ma famille élargie — bref, toute la smala quoi — pour m'avoir transmis un esprit syncrétique et indiscipliné.

Afin de porter un message de paix indélébile, je souhaite signer ces remerciements avec le nom de famille de ma mère et de mon père.

Manon Bessis Berriche

# Table des matières

# Table des figures

# Table des tableaux

# Introduction Générale

## De l’invention d’une scène de la vie quotidienne à la construction d’un objet de recherche

### La réception de l’information à l’ère du numérique : entre réactions en ligne et conversations hors ligne

16 mars 2020

D’ordinaire, Gisèle feuillette le *Télé 7 jours* pendant que son mari Daniel regarde la télévision. Enfin, « regarder » est un bien grand mot. Disons plutôt qu’il passe son temps à zapper d’une chaîne à l’autre, comme l’on picore quelques mets aléatoires tout en passant du coq à l’âne lorsqu’on est invité à un apéro dînatoire et qu’on doit faire la conversation à des inconnus. Ce soir, toutefois, inutile de zapper : l’allocution d’Emmanuel Macron est diffusée en direct sur plusieurs chaînes, même sur *L’Équipe*. Daniel n’a pas tellement d’autre choix que de l’écouter. En plus, pour couronner le tout, sa femme vient de s’emparer de la télécommande.
— Mais chut, lui dit-elle en augmentant le son, faut toujours que tu fasses du bruit au mauvais moment !
Pour une fois, Gisèle tient vraiment à prêter l’oreille au discours du président. La propagation du Covid commence à l’inquiéter. Elle a très peur que son mari ne l’attrape et finisse sous respirateur comme le beau-frère d’une de ses meilleures amies. Celui-ci n’avait pas d’antécédent pourtant, rien, pas même une once de cholestérol.

Quelques jours auparavant, Gisèle avait déjà été mise en garde contre la dangerosité du virus par sa fille Juliette. En stage d’externat dans un service de gériatrie, celle-ci avait entendu plusieurs de ses collègues exprimer de vives craintes sur la situation sanitaire. Certains d’entre eux avaient même préconisé la mise en place d’un confinement. Sans attendre, Juliette avait alors écrit à ses parents et à son grand-frère dans le groupe WhatsApp familial pour les exhorter à rester chez eux. Son frère cependant n’avait pas du tout pris son message au sérieux.

— Tout ça, c’est des rumeurs, avait-il objecté en se contentant d'ajouter un lien renvoyant vers un article des *Décodeurs* en guise d’argument.

De son côté, Gisèle avait préféré discuter de vive voix avec son mari plutôt que d’intervenir directement dans la conversation. Tout comme son fils, celui-ci n’était pas du tout préoccupé par la situation et avait déclaré à sa femme d’un ton très sûr de lui :

— Mais non, voyons, faut pas paniquer ! Tu sais bien comment est ta fille.

Gisèle se souvenait en effet comme si c’était hier de l’épisode de la grippe aviaire. Juliette avait alors dix ou onze ans. Terrifiée à l’idée d’attraper le virus, elle avait refusé de manger du poulet pendant des mois. Amusée par ce souvenir, Gisèle se dit que son mari devait avoir raison, que les inquiétudes de sa fille étaient sans doute exagérées... Elle se ravisa très vite toutefois après le coup de fil reçu par son amie deux jours plus tard.

Dès l’annonce de la mise en place d’un confinement par Emmanuel Macron, Gisèle pressent tout de suite que ça va barder dans le groupe WhatsApp familial. Pour éviter d’être assaillie par les notifications de ses enfants, elle les met en sourdine et décide de ne répondre qu’aux messages de ses amis. Dans l’un de ses groupes, plusieurs de ses copines sont déjà en train de s’organiser pour se voir à distance. Quand l’une d’entre elle suggère de faire un apéro sur Skype, Brigitte, qui n’en rate jamais une pour se faire mousser, intervient en partageant un post Facebook de *Santé + Mag.*

— Figurez-vous que boire du champagne immuniserait contre le Covid !

Étrangement, Claudine est la seule du groupe à ne pas réagir par un émoji mort de rire. Légèrement inquiète, Gisèle décide de lui envoyer un message en privé pour s’assurer qu’elle va bien. Claudine rétorque alors sans ambages qu’elle trouve toute cette histoire de Covid absurde.

— Tout ça, c’est pour nous refourguer un vaccin derrière, s’empresse-t-elle d’ajouter en transférant à Gisèle une vidéo Youtube dans laquelle un homme affirme que le virus a été créé de toute pièce en laboratoire par l’Institut Pasteur en 2004.

Sur le moment, Gisèle ne sait pas quoi répondre. D’habitude, Claudine est plutôt du genre à être bien informée sur les sujets de santé. Mais là, son message lui paraît trop gros pour être vrai. À supposer que l'Institut Pasteur ait réellement créé le Covid en 2004, pourquoi aurait-il attendu seize ans pour le propager ? Rapidement agacée par la musique dramatique de la vidéo, Gisèle n'a pas la patience de la regarder jusqu'au bout, sauf qu'en cherchant à couper le son, elle la *like* sans s'en rendre compte.

## Saisir des « confuses paroles » derrière des « forêts de symboles »[1]

Chers lecteurs, chères lectrices, l'historiette que vous venez de lire est purement fictive. Nous l'avons imaginée après avoir lu une série d'articles[2] publiés par différentes rubriques de *fact-checking*[3] au printemps 2020, alors que la France entamait son premier confinement et nous notre thèse. Ces articles, tout comme ceux parus à la suite de divers événements d'actualité, tels que le *Brexit*, l'élection de Donald Trump ou la guerre en Ukraine, partagent tous la même structure : ils commencent par mettre en exergue les volumes considérables d'engagement (i.e. de *likes*, partages et commentaires) générés par plusieurs contenus sur les réseaux sociaux, puis expliquent pourquoi ceux-ci peuvent être considérés comme des *fake news* ou *infox* — des notions sur lesquelles nous reviendrons dans les pages à venir, mais retenons pour l'heure qu'elles désignent des informations qui ont été évaluées comme fausses, exagérées ou trompeuses par des journalistes spécialisés dans la vérification factuelle.

Si les articles de *fact-checking* mettent fréquemment en avant les millions de *likes*, partages et commentaires suscités par les *fake news* pour illustrer leur circulation virale sur les réseaux sociaux, très peu d'entre eux mentionnent toutefois les profils des personnes à l'origine de ces traces numériques, pas plus que le sens qu'elles confèrent à ces actions ou le contexte dans lequel elles se trouvent au moment de les produire. Tout se passe comme si la réception de *fake news* ne se traduisait que par quelques clics et ne suscitait pas de pensées, de doutes ou de conversations. Autrement dit, alors que les processus de réception d'informations n'ont jamais laissé autant de traces qui soient publiquement accessibles, depuis l'apparition des réseaux sociaux, il ressort paradoxalement que les personnes derrière ces traces – tout comme les contextes sociaux et les situations d'interactions qui les entourent – n'ont jamais été aussi invisibles. C'est précisément cette invisibilité qui nous a poussée à débuter notre thèse par un exercice d'imagination.

---

[1] Ce titre fait écho au quatrain suivant issu du poème « Correspondances » de Baudelaire : « La Nature est un temple où de vivants piliers / Laissent parfois sortir de confuses paroles ; / L'homme y passe à travers des forêts de symboles / Qui l'observent avec des regards familiers. ». Dans ce poème, Baudelaire développe une conception néoplatonicienne de l'Univers et ouvre la voie à l'école symboliste en poésie. Il considère que les poètes doivent jouer un rôle d'intermédiaire entre la Nature et les Hommes en interprétant, à partir d'un jeu de synesthésies (i.e. d'analogies sensorielles et perceptives), « le langage des fleurs et des choses muettes ». L'historiette par laquelle débute cette thèse n'a cependant pas été écrite dans une perspective néoplatonicienne. Son objectif est simplement de rendre compte des potentiels bavardages, commérages, parlementages ou ergotages qu'il peut y avoir derrière les métriques d'engagement qui entourent la circulation de différents énoncés sur les réseaux sociaux.

[2] Maad, A., Audureau, W., Sénécat, A., Vaudano, M., & Dahyot, A. (2020, 13 mars). Coronavirus : notre guide pour distinguer les fausses rumeurs des vrais conseils. *Le Monde.* https://www.lemonde.fr/les-decodeurs/article/2020/03/13/coronavirus-petit-guide-pour-distinguer-les-fausses-rumeurs-des-vrais-conseils_6032938_4355770.html

[3] Il s'agit d'un anglicisme utilisé pour désigner une pratique journalistique qui vise à vérifier la factualité des déclarations énoncées dans les médias et les discours publics.

En offrant la liberté d'imaginer les réactions multiples et variées, parfois vacillantes, qu'il peut y avoir derrière de simples métriques d'engagement, cette invention d'une scène de la vie quotidienne – inspirée par les analyses d'Erving Goffman (1973) et de Michel de Certeau (1980) qui, dans leurs ouvrages *La Mise en scène de la vie quotidienne et L'Invention du quotidien*, invitent à prêter attention aux capacités des individus à ajuster leurs comportements aux attentes des audiences auxquelles ils sont confrontés au cours de leurs interactions sociales, ainsi qu'à déployer des tactiques pour contourner les contraintes imposées par les structures de pouvoir – permet de ne pas appréhender la question de la réception des *fake news* uniquement à partir de leur contenu ou de leurs données d'interactions. Elle donne au contraire l'opportunité de s'interroger non seulement sur « l'espace mystérieux qui s'étend entre texte et lecteur » (Dayan, 1992, p. 143) mais aussi sur celui qui sépare les traces numériques des utilisateurs des réseaux sociaux de leur vie quotidienne.

L'objet de cette thèse ne sera plus d'imaginer cet « espace mystérieux » mais d'enquêter dessus. Nous chercherons à saisir comment les *fake news* sont reçues par les utilisateurs des réseaux sociaux en partant de leurs traces numériques, mais sans les réduire à ces traces.

Afin de transformer ce questionnement initial en problématique de recherche, trois étapes ont été nécessaires. Tout d'abord, nous avons examiné la façon dont les volumes d'engagement générés par les *fake news* sur les réseaux sociaux ont été interprétés par les discours publics dans le but de saisir les représentations véhiculées dans les imaginaires collectifs. Ensuite, nous avons cherché à identifier les éléments ayant amené les discours publics à mettre en avant ces représentations. Enfin, nous avons évalué la validité empirique de ces représentations en les confrontant aux premiers résultats de recherche issus de la littérature académique sur les *fake news*. Au terme de ces trois étapes, le décalage identifié entre les conclusions des études empiriques et les interprétations des discours publics a fait émerger deux paradoxes – tout l'enjeu de cette thèse est de les expliquer.

## Les discours publics à l'épreuve des enquêtes empiriques

### Viralité des *fake news*, infodémie et crédulité

Les millions de *likes*, partages et commentaires suscités par les *fake news* sur les réseaux sociaux sont principalement interprétés de deux manières dans les discours publics, notamment ceux des médias *mainstream* (pour des analyses, voir Farkas et Shou, 2020 ; Vauchez, 2022).

La première interprétation porte sur l'écosystème informationnel contemporain. Les millions d'interactions générées par les *fake news* sur les plateformes numériques sont souvent considérées comme des indicateurs d'un marché dérégulé de l'information, c'est-à-dire d'un espace informationnel sans règle ni hiérarchie, dans lequel des contenus erronés seraient aussi visibles et prévalents que des informations fiables et vérifiées. En établissant régulièrement des listes du Top 10 ou du Top 100 des contenus qui déclenchent le plus d'interactions sur les réseaux sociaux, de nombreux articles concluent que les *fake news* touchent une forte audience et qu'elles ont parfois plus de succès que les informations issues des médias *mainstream*.[4] Par exemple, une analyse conduite par le journaliste Craig Silverman en 2017 montre que les vingt *fake news* les plus partagées au cours de la campagne électorale américaine de 2016 ont suscité 8,7 millions de partages, réactions et commentaires, soit environ 1,3 million de plus que les vingt informations les plus partagées issues des médias traditionnels.[5] L'importante popularité des *fake news* sur les réseaux sociaux a alors conduit de nombreux acteurs à introduire de nouveaux termes dans le débat public pour décrire l'écosystème informationnel contemporain. En 2016, par exemple, d'innombrables titres médiatiques et rapports institutionnels ont proclamé que nous étions entrés dans une ère de la « post-vérité »[6], c'est-à-dire dans une ère au sein de laquelle les individus accorderaient moins d'importance aux faits qu'à leurs propres émotions ou opinions personnelles.[7] En 2020, l'Organisation Mondiale de la Santé a utilisé la notion d'« infodémie » pour souligner qu'en parallèle de la pandémie de Covid-19, les sociétés devaient faire face à un contexte d'abondance informationnel, marqué par la circulation de nombreux contenus peu fiables. Plus récemment, enfin, avec le développement des techniques de *machine learning*, notamment des intelligences artificielles génératives, de nouvelles expressions sont apparues dans le débat public, comme celles d'« infocalypse »[8], pour mettre en garde contre l'émergence d'une nouvelle génération de *fake news*, caractérisée par la production automatisée de contenus artificiels d'une qualité inédite. L'emploi très fréquent de ces nouveaux termes dans le débat public donne ainsi l'impression que les internautes sont submergés par un déluge de fausses informations et naviguent sur le web comme sur un bateau ivre, sans haleur pour les guider.

[4] Pour des exemples, voir les articles suivants :
Silverman, C. (2016, 30 décembre). Here Are 50 Of The Biggest Fake News Hits On Facebook From 2016. *BuzzFeed*. https://www.buzzfeednews.com/article/craigsilverman/top-fake-news-of-2016
Silverman, C., Lytvynenko, J., Pham, S. (2017, 28 décembre). These are 50 of the biggest fake news hits on Facebook in 2017. *Buzzfeed.* https://www.buzzfeednews.com/article/craigsilverman/these-are-50-of-the-biggest-fake-news-hits-on-facebook-in
Silverman, C., Pham, S. (2018, 28 décembre). These are 50 of the biggest fake news hits on facebook in 2018. *Buzzfeed*. https://www.buzzfeednews.com/article/craigsilverman/facebook-fake-news-hits-2018
Sénécat, A. (2017, 19 décembre). Facebook, voyage au cœur de la machine à fausses informations. *Le Monde*. https://www.lemonde.fr/les-decodeurs/article/2017/12/19/facebook-voyage-au-c-ur-de-la-machine-a-fausses-informations_5231640_4355770.html

La deuxième interprétation concerne les capacités de discernement des utilisateurs des réseaux sociaux. Les millions de *likes*, partages et commentaires émis en réaction à des *fake news* sont également souvent perçus comme le signe d'une crédulité de masse. En effet, le fait de *liker* ou de partager une *fake news* est souvent assimilé au fait d'y croire ou d'y adhérer, voire à certaines attitudes et comportements comme le fait de voter pour tel ou tel candidat ou de refuser tel ou tel vaccin. Par exemple, de nombreux articles médiatiques établissent un lien de cause à effet direct entre la circulation virale de *fake news* sur les réseaux sociaux et différents événements d'actualité ou enjeux contemporains, tels que le Brexit (e.g. *Fake news handed Brexiteers the referendum*[9]), l'élection de Donald Trump (e.g. *Click and elect: how fake news helped Donald Trump win a real election*[10]), ou l'hésitation vaccinale (e.g. « Covid-19 : le mouvement antivax porté par la désinformation »[11] ; « Covid-19 : la désinformation sur les vaccins pousse des internautes à rechercher du "sang pur" »[12]). En déclarant ainsi que les *fake news* ont un fort impact sur les croyances, opinions, attitudes et comportements des individus, les discours publics laissent donc entendre que les utilisateurs des réseaux sociaux seraient facilement influençables ou manipulables, et que leur crédulité face aux *fake news* serait la cause principale de divers enjeux contemporains, tels que la montée des populismes, la polarisation politique ou l'hésitation vaccinale.

---

[5] Silverman, C. (2016, 16 novembre). This Analysis Shows how Fake Election News Stories Outperformed Real News on Facebook. *BuzzFeed News*. https://www.buzzfeednews.com/article/craigsilverman/viral-fake-election-news-outperformed-real-news-on-facebook

[6] Fortement employée dans les discours publics, l'expression a été élue mot de l'année 2016 par le dictionnaire Oxford. Flood, A. (2016, 15 novembre). 'Post-Truth' Named Word of the Year by Oxford Dictionaries. *The Guardian.* https://www.theguardian.com/books/2016/nov/15/post-truth-named-word-of-the-year-by-oxford-dictionaries

[7] Oxford English Dictionaries. (2016). Post-truth. https://www.oed.com/dictionary/post-truth_adj?tab=factsheet - 1217123470

[8] Warzel, C. (2018, 11 février). He Predicted the 2016 Fake News Crisis. Now He's Worried About An Information Apocalypse. *BuzzFeed News.* *https://www.buzzfeed.com/fr/charliewarzel/the-terrifying-future-of-fake-news-2?new_user=true&newsletter_modal=true* ;
Naffi, N., Davidson, A.L., Berger, F. (2021, 20 avril), Infocalypse : la propagation des hypertrucages menace la société. *The Conversation.* *https://theconversation.com/infocalypse-la-propagation-des-hypertrucages-menace-la-societe-158335*

En résumé, il ressort donc que les discours publics véhiculent des représentations alarmistes de l'écosystème informationnel contemporain et de ses publics à partir des millions de réactions suscitées par les *fake news* sur les réseaux sociaux. D'un côté, l'écosystème informationnel contemporain est décrit comme un marché de l'information dérégulé. De l'autre, les utilisateurs des réseaux sociaux sont dépeints comme des individus crédules. Le premier chapitre de cette thèse montrera comment ces représentations présentent de nombreuses similarités avec différentes inquiétudes formulées au cours des dernières décennies à la suite de l'apparition de nouveaux médias et de nouvelles technologies. Si ces craintes ont souvent été qualifiées de « panique morale » ou de « panique médiatique » par de nombreux chercheurs en sociologie, nous faisons le choix de ne pas appliquer ce concept dès maintenant aux discours sur les *fake news*. En effet, le concept de « panique morale » fait l'objet de nombreux débats théoriques (Critcher, 2008 ; Chaumont, 2012 ; Mathieu, 2015 ; Mavrot et al., 2022) et est souvent critiqué pour son caractère normatif. Aussi, est-il important d'adopter dans un premier temps une approche compréhensive cherchant à identifier sans *a priori* les raisons ayant conduit les discours publics à proposer ces représentations.

---

[9] Grice, A. (2017, 18 janvier). Fake news handed Brexiteers the referendum - and now they have no idea what they're doing. *The Independent*. https://www.independent.co.uk/voices/michael-gove-boris-johnson-brexit-eurosceptic-press-theresa-may-a7533806.html

[10] Parkinson, J.N. (2016, 14 novembre). Click and elect: how fake news helped Donald Trump win a real election. *The Guardian*. https://www.theguardian.com/commentisfree/2016/nov/14/fake-news-donald-trump-election-alt-right-social-media-tech-companies

[11] Le Point. (2022, 4 juillet). Covid-19 : le mouvement antivax porté par la désinformation. *Le Point*. https://www.lepoint.fr/monde/covid-19-le-mouvement-antivax-porte-par-la-desinformation-04-07-2022-2481974_24.php

[12] Europe 1. (2023, 25 janvier). Covid-19 : la désinformation sur les vaccins pousse des internautes à rechercher du «sang pur». *Europe 1.* https://www.europe1.fr/international/covid-19-la-desinformation-sur-les-vaccins-pousse-des-internautes-a-rechercher-du-sang-pur-4163498

## Numérisation de l'écosystème informationnel et transformations de l'espace public

En diffusant des représentations alarmistes de l'écosystème informationnel contemporain et de ses publics, les discours publics présentent de nombreuses similarités avec les arguments défendus par le sociologue Gérald Bronner. Dans des ouvrages, fréquemment cités dans le débat public, tels que *La Démocratie des Crédules* (Bronner, 2013) ou *Apocalypse cognitive* (Bronner, 2021), celui-ci soutient l'idée qu'Internet a dérégulé le marché de l'information, et favorise la crédulité des individus en les rendant plus susceptibles de se faire piéger par leurs biais cognitifs, notamment par leur biais de confirmation.[13] Par exemple, dans un article intitulé « Ce qu'Internet fait à la diffusion des croyances », il explique, équation à l'appui, que « plus N, le nombre d'informations disponibles augmente, plus la possibilité pour l'individu d'avoir recours au biais de confirmation est forte » (Bronner, 2011). Quelques années plus tard, dans un article intitulé cette fois « Internet peut favoriser la crédulité », il présente un « théorème de la crédulité informationnelle » selon lequel les mécanismes de recherche sélective de l'information sont renforcés par la massification de l'information dans la mesure où « la probabilité de trouver une information qui va dans le sens des attentes idéologiques de l'individu croît à mesure de la diffusion non coordonnée de l'information » (Bronner, 2019).

Afin de comprendre ces arguments, et les représentations qui en découlent, il est nécessaire d'examiner de façon approfondie les transformations provoquées par la numérisation de l'écosystème informationnel — dues à l'essor du web et des réseaux sociaux — sur la circulation et la consommation d'information. En prenant appui sur les travaux de Dominique Cardon (2010), quatre transformations principales peuvent être mises en avant.

[13] Il s'agit d'un mécanismes cognitif défini sur Wikipédia comme la tendance « qui consiste à privilégier les informations confirmant ses idées préconçues ou ses hypothèses, ou à accorder moins de poids aux hypothèses et informations jouant en défaveur de ses conceptions, ce qui se traduit par une réticence à changer d'avis. » https://fr.wikipedia.org/wiki/Biais_de_confirmation

La première transformation tient à l'élargissement de l'accès à la publication d'informations dans l'espace public à des locuteurs qui ne sont pas des professionnels du journalisme. Avant l'apparition du web et des réseaux sociaux, les informations publiées dans l'espace public étaient filtrées par des *gatekeepers,* c'est-à-dire par des professionnels de l'édition et de l'information (e.g. journalistes, éditeurs, experts, institutions, etc.) tenus d'obéir à des normes déontologiques. Ceux-ci déterminaient quels événements d'actualité étaient dignes d'être couverts et d'être rendus publics en fonction de divers critères de *newsworthiness* (Galtung et Ruge, 1965), notamment en évaluant leur degré d'intérêt général. En sélectionnant les énoncés diffusés dans l'espace public, les *gatekeepers* jouaient ainsi un rôle crucial dans la construction de l'agenda public, l'orientation des débats et le façonnement de l'opinion. L'essor du web et des réseaux sociaux a toutefois fait disparaître le monopole que s'étaient attribués les *gatekeepers* sur la mise en circulation d'informations dans l'espace public en donnant à tout un chacun la possibilité de publier des contenus en ligne sans être soumis à un contrôle éditorial, ni à des règles de déontologie journalistique. Les médias traditionnels se sont alors retrouvés en concurrence avec un ensemble disparate de nouveaux producteurs d'information : *pure players*, blogueurs, sites institutionnels, pages personnelles, sans oublier l'ensemble des internautes qui partagent des contenus sur les réseaux sociaux. Si, certains chercheurs ont d'abord vu dans cet élargissement de l'accès à la publication d'informations une opportunité pour favoriser la participation de tous dans le débat public (Shirky, 2008), notamment de « contre publics » qui avaient précédemment un accès très restreint à la visibilité publique (Fraser, 1990), d'autres y ont rapidement perçu des limites (Hindman, 2009), voire des risques pour la démocratie. Selon Gérald Bronner (2021), par exemple, l'ouverture des arènes du débat public à de nouveaux acteurs peut conduire à un relativisme ambiant, dans lequel la parole d'un universitaire ou d'un journaliste se retrouve mise au même niveau que celle d'un youtubeur, d'un blogueur, d'un influenceur (Candel et Gkouskou-Giannakou, 2017) ou d'acteurs malveillants comme des trolls, des robots ou des bots cherchant à introduire des informations trompeuses ou des propos haineux dans l'espace public pour y semer le chaos (Marantz, 2019 ; Pomerantsev, 2019 ; Singer et Brooking, 2018).

La deuxième grande transformation réside dans l'émergence de nouveaux processus de mise en visibilité des informations reposant notamment sur des algorithmes. En effet, si les informations diffusées sur la toile ne sont plus filtrées en étant au préalable passées au crible par des experts et journalistes, elles sont hiérarchisées *a posteriori* par des algorithmes de classement et de référencement, eux-mêmes alimentés par différents signaux, tels que les clics des internautes. En d'autres termes, alors que les journalistes ne sont pas en mesure de contrôler directement quelles informations sont publiées et visibles sur les réseaux sociaux, les algorithmes des plateformes peuvent être paramétrés pour privilégier certains contenus par rapport à d'autres sur les fils d'actualité des utilisateurs. Par ailleurs, n'importe quel utilisateur peut librement *liker*, partager ou commenter n'importe quel contenu, ce qui contribue *in fine* à façonner le fonctionnement des algorithmes. Ainsi, la visibilité d'une information sur les réseaux sociaux dépend à la fois des choix des plateformes, qui cherchent en partie à maximiser leur trafic pour générer des revenus publicitaires, et des préférences des internautes qui utilisent souvent principalement les réseaux sociaux pour se divertir ou pour discuter avec leurs proches plutôt que pour s'informer (Newman et al., 2019). En raison de cet enchevêtrement entre des logiques algorithmiques et des dynamiques de sociabilité, certains contenus peuvent rapidement devenir viraux sur les réseaux sociaux et ainsi bénéficier d'une visibilité qu'ils n'auraient peut-être pas réussi à obtenir sur d'autres espaces de communication. C'est ainsi que Facebook a été critiqué pour avoir favorisé la viralité des *fake news* pendant les élections présidentielles américaines de 2016 en mettant en avant des contenus sensationnalistes et controversés dans le fil d'actualité des utilisateurs (Allcott et Gentzkow, 2017). En outre, alors que la pertinence attribuée à une information ne résulte plus forcément d'une évaluation normative de son contenu mais émane d'une simple « agrégation numérique » et « du seul dénombrement d'actions individuelles inorganisées » (Cardon, 2010, p. 51), il est possible que certains acteurs exploitent de façon méthodique les nouvelles structures de visibilité permises par les réseaux sociaux en ayant recours au *digital labor*, à des fermes à clics ou à des services proposés par des entreprises spécialisées en e-réputation afin d'augmenter de façon artificielle le nombre de *likes* de différents contenus ou le nombre de fans ou de *followers* d'une page Facebook ou d'un compte Twitter (Boullier et Lohard, 2012). Cette importance des métriques de popularité dans la visibilité de l'information permet aussi l'utilisation d'outils numériques pour influencer de manière ciblée les représentations individuelles. En effet, en tirant profit de leurs algorithmes de recommandation, les plateformes ont la possibilité d'offrir à leurs utilisateurs une information personnalisée, adaptée à leurs préférences et à leurs comportements en ligne. Ces algorithmes analysent les interactions passées des utilisateurs avec divers contenus afin de prédire quels articles, vidéos ou publicités sont les plus susceptibles de les intéresser. En utilisant ces données, des acteurs peuvent alors cibler spécifiquement les individus avec des messages conçus pour influencer leurs opinions et leurs comportements que ce soit dans le domaine politique, commercial ou social comme l'a illustré l'affaire *Cambridge Analytica*. Selon le militant Eli Pariser (2011), cette personnalisation peut aussi créer des « bulles de filtres » qui, en exposant principalement des utilisateurs à des informations qui correspondent à leurs intérêts et opinions préalables, sont susceptibles de renforcer leurs croyances existantes plutôt que de les confronter à d'autres points de vue.

La troisième transformation découle des deux premières : l'élargissement de la prise de parole à de nouveaux locuteurs, d'une part, et l'émergence de nouveaux processus de mise en visibilité, d'autre part, ont créé un contexte d'abondance informationnelle fortement concurrentiel, dans lequel l'attention est devenue une denrée rare à capter (Boullier, 2012 ; Citton, 2014), obligeant ainsi les médias à trouver de nouveaux modes de financement et à adopter de nouveaux modèles économiques. Contraints d'être en quête de clics pour monétiser leur audience, de nombreux médias en ligne se sont mis à produire des contenus sensationnalistes ou aguicheurs pour accaparer l'attention des utilisateurs des réseaux sociaux. De cette manière, ils espèrent déclencher des taux d'engagement importants afin de générer du trafic vers leur site Internet et tirer des revenus publicitaires des annonces qu'ils hébergent. Cette poursuite constante du buzz exerce aussi une certaine pression sur les médias traditionnels (Boullier, 2009). En effet, les techniques de mesure d'audience en temps réel de la popularité des articles et de leur taux de partage ont reconfiguré le travail des rédactions (Christin, 2018 ; 2020). Celles-ci n'ont pas toujours le temps et les ressources pour se consacrer à la production d'informations fiables, originales et vérifiées, et cette course effrénée peut conduire à une dégradation de la qualité de l'information. Une étude conduite par Julia Cagé, Nicolas Hervé et Marie-Luce Viaud (2020) a par exemple montré que 64 % des informations diffusées par les médias sur Internet relèvent du pur copié-collé. Par ailleurs, aux États-Unis, certains grands titres médiatiques, contaminés par des logiques commerciales et publicitaires, ont parfois davantage couvert des ragots anecdotiques ou des rumeurs médisantes sur les candidats aux élections présidentielles américaines de 2016 que leurs programmes politiques. Par exemple, en seulement six jours, le *New York Times* a mis autant de fois à la une un scandale lié à la boîte mail d'Hillary Clinton que toutes les questions de politique publiées au cours des soixante-neuf jours qui ont précédé les élections américaines (Watts et Rothschild, 2017 ; Benkler et al., 2018, p. 17).

La quatrième transformation, enfin, se caractérise par l'émergence d'espaces de discussion hybrides où les frontières sont de plus en plus poreuses entre les arènes de débats relatifs à des sujets d'intérêt général, respectant idéalement les principes habermassiens de l'agir communicationnel (Habermas, 1981), et les espaces de conversations dédiés à partager des expériences de la vie quotidienne en suivant des rites de sociabilité (Goffman, 1973). En effet, avec les réseaux sociaux, les conversations interpersonnelles, initialement échangées dans des contextes privés sont venues s'entremêler aux informations relatives au débat public. L'émergence de ces espaces de visibilité en « clair-obscur » (Cardon, 2010) a ainsi favorisé l'apparition de nouveaux registres d'énonciation dans l'espace public, notamment une forme de « parler privé-public » (Cardon, 2012) qui se caractérise sur les réseaux sociaux par la mise en circulation d'énoncés personnels émis dans un registre familier et relâché. La mise en lumière de ces réactions et conversations, qui étaient jusqu'à présent hors des radars des *gatekeepers*, a dès lors conduit ces derniers à craindre que les débats délibératifs, caractéristiques de l'idéal de l'espace public habermassien, soient supplantés par des discussions informelles ou des disputes véhémentes, si ce n'est par une « brutalisation du débat public » (Badouard, 2017). Il est important de garder en tête toutefois que les échanges en ligne reflètent des pratiques sociales qui ont toujours eu lieu, derrière un écran de télévision en famille ou autour d'une table avec des amis. Or, il est possible que, dans ces situations ordinaires de sociabilité, les individus mobilisent des compétences de distance critique, de raisonnement et d'argumentation qui se manifestent sous une autre forme que celles fixées par les attentes normatives des discours publics.

Sous l'essor du web et des réseaux sociaux, un « glissement d'un espace public mass-médiatique vers un espace public en réseau » (Benkler, 2006, p. 39) a ainsi pu être observé. Ce changement de morphologie a conduit différents acteurs du débat public à décrire l'écosystème informationnel contemporain comme un marché dérégulé et les publics comme une collection d'individus atomisés et ciblés par une industrie de la persuasion diffusant des messages taillés sur mesure, selon leurs préférences, via des algorithmes de personnalisation. En effet, l'élargissement de l'accès à la publication d'informations dans l'espace public donne désormais la possibilité à divers acteurs de déployer des stratégies d'influence ou de publier des contenus sans être soumis aux normes de la profession journalistique. Ensuite, le rôle joué par les algorithmes dans la mise en visibilité des informations peut favoriser la viralité de contenus sensationnalistes et surtout le recours à des stratégies de ciblage – ce qui risque de réduire la diversité des points de vue et ainsi de polariser davantage les opinions publiques. Par ailleurs, les nouveaux modèles économiques des médias, axés sur la quête de clics et la maximisation du trafic, peuvent encourager la production de contenus non vérifiés au détriment d'informations fiables. Enfin, le brouillage des frontières entre des espaces de discussions publics et privés peut favoriser le déroulement de conversations informelles, marquées par l'expression d'émotions et d'opinions personnelles, au détriment de débats rationnels, orientés vers l'intérêt général.

Dans ces conditions, l'on pourrait s'attendre à ce que les utilisateurs des réseaux sociaux soient noyés sous un déluge de *fake news* et peu à même de résister à leur influence. Seulement, est-ce vraiment ce qu'on observe empiriquement ? Pour répondre à cette question, la prochaine section rend compte des premiers constats issus de la littérature académique sur les *fake news*.

## Nuances et zones d'ombre dans la littérature académique

Au-delà d'être devenues un sujet de préoccupation majeur dans le débat public, les *fake news* ont aussi fait l'objet d'un nombre important de recherches depuis 2016, avec pas moins de 2 000 articles publiés dans une grande variété de revues académiques, couvrant un large éventail de disciplines, allant de la sociologie à l'informatique en passant par la psychologie, les sciences politiques, l'économie, les sciences de l'information et de la communication ou encore la philosophie (Righetti, 2021). Loin de chercher à offrir une synthèse exhaustive de tous ces travaux, cette partie propose simplement de passer en revue les principaux résultats issus d'enquêtes empiriques qui se sont attachées à : (1) évaluer l'audience des *fake news* ou la place qu'elles occupent dans les habitudes de consommation médiatique des publics, ainsi qu'à (2) mesurer leurs effets sur les croyances, opinions, attitudes et comportements des individus.[14] Se concentrer sur ces deux questions de recherche permet en effet de confronter la façon dont les volumes d'engagement des *fake news* ont été interprétés par les discours publics avec les principaux constats issus de la littérature académique. Avant de rendre compte de ces différents résultats, il est important toutefois de préciser la manière dont les *fake news* sont définies dans le cadre d'enquêtes empiriques car cette définition influence grandement les conclusions des chercheurs (Rogers, 2020).[15]

### ***Définir le terme fake news***

Initialement utilisé par les médias traditionnels pour désigner les informations erronées, exagérées ou trompeuses qui ont circulé sur les réseaux sociaux pendant la campagne électorale américaine de 2016, le terme *fake news* a très vite été critiqué par de nombreux acteurs institutionnels et académiques. Certains d'entre eux ont même préconisé de ne pas l'employer.[16] Tout d'abord, en raison de son caractère vague et imprécis – le mot étant aussi bien utilisé pour désigner des contenus partisans ou sensationnalistes que pour faire référence à des notions plus anciennes telles que celles de rumeur, de propagande ou de théorie du complot (le chapitre 1 questionnera justement la distinction entre le terme *fake news* et ces différents concepts). Ensuite, parce que la notion de *fake news* a été instrumentalisée par de nombreuses personnalités politiques[17] pour décrédibiliser les propos

---

[14] Ces dimensions sont fréquemment distinguées dans les recherches en sciences sociales (pour une discussion sur les différences entre ces concepts, voir : Dargent, 2011 ; Peretti-Watel, et al., 2015).

[15] Une étude menée par Richard Roger (2020) montre que si l'on adopte une définition stricte du terme « misinformation », en ne considérant que les sites étiquetés comme « trompeurs » ou « conspirationnistes » tout en excluant ceux qualifiés d'« hyper-partisans » ou de « sensationnalistes », la proportion de *fake news* qui suscitent davantage d'interactions que les médias traditionnels diminue fortement, passant d'un ratio d'un quart à un neuvième.

[16] Audureau, W. (2017, 31 janvier). Pourquoi il faut arrêter de parler de "fake news". *Le Monde*. https://www.lemonde.fr/les-decodeurs/article/2017/01/31/pourquoi-il-faut-arreter-de-parler-de-fake-news_5072404_4355770.html

[17] Plusieurs articles médiatiques ont par exemple documenté l'emploi fréquent du terme *fake news* par Donald Trump pour critiquer de nombreux médias traditionnels comme *CNN* ou le *Washington Post* ainsi que toutes les informations n'allant pas dans le sens de son point de vue sur certains faits.
Bump, P. (2018, 9 mai). Trump makes it explicit: Negative coverage of his is fake coverage. *The Washington Post.* https://www.washingtonpost.com/news/politics/wp/2018/05/09/trump-makes-it-explicit-negative-coverage-of-him-is-fake-coverage/
Lind, D. (2018, 9 mai). Trump finally admits that "fake news" just means news he doesn't like. *Vox.* https://www.vox.com/policy-and-politics/2018/5/9/17335306/trump-tweet-twitter-latest-fake-news-credentials

de leurs opposants idéologiques (Caplan et al., 2018). Avec cette forme de discours, le sens du mot s'est ainsi déconnecté de ce qu'il était censé qualifier, à savoir l'inexactitude d'une information, pour devenir tributaire de l'interprétation que différents acteurs en ont. Ces tensions entre différentes significations[18] – selon que le terme fasse l'objet d'un usage journalistique, politique, scientifique ou de sens commun (Dauphin, 2019) – ont conduit plusieurs chercheurs comme Johan Farkas et Jannick Shou (2018) à considérer la notion de *fake news* comme un « signifiant flottant ». Plutôt que de constituer une catégorie descriptive permettant d'évaluer la qualité épistémique d'un contenu, le terme *fake news* s'apparente ainsi davantage à une catégorie normative utilisée par différents groupes d'acteurs (journalistes, personnalités publiques, etc.) pour défendre chacun leur définition de la factualité, et souvent, à travers elle, des intérêts idéologiques ou professionnels (Liu et Su, 2020). Toujours est-il qu'en appliquant les propos tenus par Emmanuel Taïeb (2010, p. 267-268) sur le conspirationnisme aux *fake news*, on peut considérer que :

> *Le fait que le terme même de [fake news] puisse être dénoncé comme étiquette, dans des controverses propres au champ intellectuel et scientifique, comme peuvent d'ailleurs l'être les mots « rumeur » ou « propagande » (Taïeb, 2006, 2010), ne doit pas dispenser de l'analyse propre de son objet.*

De nombreux chercheurs ont ainsi entrepris d'importants efforts de conceptualisation ces dernières années pour définir le terme *fake news*. Par exemple, Claire Wardle et Hossein Derakhshan (2017) ont proposé d'utiliser les notions de *misinformation*, *malinformation* et *disinformation* pour distinguer différents types de fausses informations selon que les intentions de leurs producteurs soient malveillantes ou non. D'autres chercheurs ont élaboré des typologies des *fake news* permettant de différencier plusieurs catégories de fausses informations allant de la parodie au contenu piège à clics en passant par les contenus partisans, etc. (Tandoc et al., 2018 ; Kapantai et al., 2021).

[18] Holan A. (2017, October 18). The media's definition of fake news vs. Donald Trump's. *PolitiFact*. https://www.politifact.com/truth-o-meter/article/2017/oct/18/deciding-whats-fake-medias-definition-fake-news-vs/

Malgré ces distinctions conceptuelles et recommandations institutionnelles, le mot *fake news*[19] est toutefois resté le plus souvent utilisé comme un terme parapluie dans la littérature académique pour désigner de façon générique plusieurs types de fausses informations, indépendamment des intentions de leurs producteurs. En pratique, par ailleurs, la majorité des travaux empiriques ont jusqu'à présent délimité leur corpus de *fake news* à partir des *fact-checks* réalisés par des journalistes spécialisés dans la vérification factuelle. Si cette définition opérationnelle a le défaut de réduire l'étiquetage de *fake news* aux choix des *fact-checkers* qui, loin d'être impartiaux, peuvent s'avérer biaisés, arbitraires et surtout très partiels (Uscinski et Butler, 2013 ; Graves, 2016), elle permet de confronter les représentations issues de catégories journalistiques à des données empiriques. En effet, en prenant appui sur les URLs ou noms de domaines[20] évalués comme des *fake news* par des *fact-checkers*, on peut retrouver certains – et certains seulement – des individus qui y ont été exposés, et ainsi mettre en tension la façon dont ceux-ci sont dépeints dans le débat public avec leurs comportements effectifs.

Le travail d'enquête mené dans le cadre de cette thèse repose en partie sur cette définition opérationnelle du mot *fake news*, mais nous proposons de l'élargir, dans le cadre d'une de nos enquêtes, en nous appuyant également sur les signalements des utilisateurs des réseaux sociaux. Cette approche permet de partir d'énoncés perçus comme erronés, exagérés ou mensongers par certains utilisateurs, mais qui n'ont pas été évalués comme des *fake news* par des *fact-checkers*, et ainsi de dépasser la seule question de la factualité afin d'étudier les réactions des utilisateurs des réseaux sociaux à des énoncés dont la qualité épistémique est incertaine au cours de situations qu'ils sont susceptibles de percevoir comme problématiques – au sens de John Dewey[21]. Prendre de la distance avec la question de la factualité des énoncés permet en effet de « se concentrer sur le travail de qualification, donc de labellisation, dont ils font l'objets » et d'examiner « les dynamiques concurrentes d'affirmation et de contestation de l'autorité épistémique, sans préjuger du succès ni de la légitimité de ces entreprises et des groupes qui les mettent en œuvre » (Boullier et al., 2021).

---

[19] Si certaines études anglo-saxonnes privilégient le terme « misinformation » plutôt que celui de *fake news*, l'équivalent français « mésinformation » est peu usité et c'est le terme *fake news* qui reste le plus courant. Dans la même veine, si le néologisme « infox » a été recommandé par la Commission d'enrichissement de la langue française en octobre 2018 pour éviter d'employer l'anglicisme *fake news,* celui-ci reste employé beaucoup plus fréquemment dans les études académiques. Par ailleurs, la définition du terme « infox » se focalise essentiellement sur la dimension idéologique des fausses informations sans tenir compte de leurs logiques économiques et sociales ou de leur absence de teneur informative. En effet, elle désigne « *une information mensongère ou délibérément biaisée, répandue par exemple pour favoriser un parti politique au détriment d'un autre, pour entacher la réputation d'une personnalité ou d'une entreprise, ou encore pour contredire une vérité scientifique* ».
https://www.legifrance.gouv.fr/affichTexte.do?cidTexte=JORFTEXT000037460897&dateTexte=&categorieLien=id

[20] Deux approches principales sont utilisées dans les études académiques pour délimiter des corpus de *fake news*. La première repose sur des listes d'URLs évaluées comme *fake news* par des *fact-checkers*. Cette méthode permet de cibler précisément des articles contenant des informations factuellement erronées, mais de nombreux contenus potentiellement problématiques échappent à cette définition restreinte. La deuxième approche consiste à cibler non pas directement les contenus, mais leurs sources en attribuant aux médias une note de fiabilité à partir d'indicateurs produits par des *fact-checkeurs* (comme le *Décodex* du *Monde*) ou des nouvelles entreprises de notation des médias comme *News Guard*. Cette méthode permet d'englober un nombre plus large de contenus mais risque de qualifier de *fake news* des informations factuelles produites par des médias alternatifs et de ne pas tenir compte des erreurs des médias *mainstream*.

Cette double façon d'opérationnaliser le terme *fake news* ne permettra pas cependant de tenir compte des cas de mésinformation et de désinformation qui se manifestent sous des formes audiovisuelles (à travers des mèmes, des vidéos et des photos) ou reposent sur des techniques d'intelligence artificielle générative (*deepfake*, etc.) et qui sont de plus en plus présentes sur les réseaux sociaux (Yang et al., 2023 ; Peng et al. 2023 ; Brennen et al., 2021).

Tout au long de la thèse, nous continuerons de mener une discussion critique sur la notion de *fake news*. Nous montrerons qu'un élargissement de perspective est nécessaire « s'intéressant moins à des exemples isolés de *fake news* qu'à la manière dont elles et d'autres processus perturbateurs s'inscrivent dans des désordres de désinformation plus vastes » (Bennett et Livingston, 2018, p. 135). Retenons toutefois pour l'instant que la notion de *fake news* désigne l'ensemble des contenus qui ont été qualifiés comme tels par des *fact-checkers*. Cette définition est en effet celle utilisée par la majorité des études que nous allons passer en revue dans les paragraphes suivants.

***Évaluer l'audience des fake news et leur place dans la consommation d'informations des publics***

Si les volumes d'engagement générés par les *fake news* sur les réseaux sociaux paraissent de prime abord impressionnants, ils ne permettent pas d'évaluer de façon rigoureuse la proportion d'internautes exposés à des *fake news* et encore moins la place que celles-ci occupent dans le total d'informations qu'ils consomment. Reprenons par exemple le cas des 8,7 millions de partages, réactions et commentaires suscités par les vingt *fake news* qui ont généré le plus d'interactions sur Facebook au cours de la campagne électorale américaine de 2016 (cf. p. 24-25). Si l'on suppose que, sur la même période, tous les utilisateurs actifs américains ont interagi en moyenne avec un contenu par jour, alors les 8,7 millions d'interactions mentionnées ci-dessus ne représentent plus que 0,006 % du total d'engagement avec des contenus sur Facebook (Watts et Rothschild, 2017).

[21] Dewey (1949) qualifie de « situation problématique » une situation « qui soulève des questions, et qui donc appelle l'investigation, l'examen, la discussion — en bref, l'enquête ».

Cet exemple illustre l'importance de ne pas interpréter trop vite les volumes d'engagement associés aux *fake news* sur les réseaux sociaux, notamment lorsqu'ils proviennent d'une seule plateforme. À la place, il est essentiel de les mettre en perspective avec l'ensemble des pratiques informationnelles et numériques des utilisateurs. En effet, certains médias traditionnels comme la télévision reste le premier support utilisé par les individus pour s'informer (Mitchell, 2018 ; Newman et al., 2019), et chaque plateforme se caractérise par des fonctionnalités spécifiques qui façonnent non seulement le degré de visibilité des informations mais aussi les usages des publics (Newman et al., 2021). Ainsi, ce n'est pas parce qu'un utilisateur a été exposé à une ou plusieurs *fake news* sur un réseau social particulier que celles-ci occupent une large place dans l'ensemble des informations qu'il consomme. Par ailleurs, il est tout à fait possible que ce même utilisateur ne soit pas exposé à des *fake news*, ou n'en partage pas du tout, sur d'autres plateformes.

Les recherches les plus robustes sur l'audience et la consommation de *fake news* sont donc celles qui ont veillé à tenir compte de l'intégralité de l'écosystème informationnel dans lequel baignent les utilisateurs des réseaux sociaux en conduisant des études multiplateformes (Bode et Vraga, 2018). Pour cela, elles ont essayé de mobiliser à la fois des données de consommation de télévision et de navigation sur différents sites web depuis des ordinateurs ou des appareils mobiles (via des entreprises telles que Nielsen et Comscore), ainsi que des données estimant le nombre de vues des contenus sur les réseaux sociaux (Pasquetto et al., 2020). À de rares exceptions près, néanmoins, ces différentes données, notamment le nombre de vues des contenus sur les réseaux sociaux (Guess et al., 2021 ; Allen et al., 2021a ; 2024), sont difficilement accessibles et les chercheurs sont souvent contraints de se contenter de leurs données d'engagement qui constituent un proxy imparfait pour évaluer leur portée et leur audience (Gonzalez-Bailon et al., 2010 ; de Vreese et Tromble, 2023).

Pour estimer de façon rigoureuse la proportion d'internautes exposés à des *fake news*, et la place que celles-ci occupent dans le total d'informations qu'ils consomment, il est donc nécessaire de s'interroger sur les différents types de données mobilisées par les chercheurs, et ainsi de prolonger les questionnements classiques sur les méthodes de mesures d'audience des médias de masse (Méadel, 2010 ; Napoli, 2010), qui, comme l'explique Sabine Chalvon-Demersay (1999, p. 50), « contribuent, par les partis pris qu'elles recèlent, à mettre en forme une certaine idée du public ».

En prenant en compte ces enjeux méthodologiques, nous avons passé en revue une quinzaine d'enquêtes ayant mobilisé différents types de données pour mesurer l'audience des *fake news* et la place qu'elles occupent dans les habitudes de consommation médiatique des individus. Des précisions sur la taille des échantillons, les corpus et les types de données mobilisés dans ces études, ainsi que leurs résultats, ont été listées dans le tableau situé en Annexe 1. Ci-dessous, nous proposons une synthèse des interprétations globales que nous pouvons en tirer. Quatre constats principaux ressortent.

Le premier constat indique qu'une proportion non négligeable d'individus a consulté au moins un site de *fake news*. L'estimation s'élève à 44,3 % dans une étude menée sur un échantillon représentatif de la population américaine pendant la campagne électorale de 2016 (Guess et al., 2020a), mais diminue à 21 % en 2018 (Zhou et al., 2023) et à 26,2 % lors de la campagne présidentielle américaine de 2020 (Moore et al., 2023). En France, l'enquête conduite par Laurent Cordonier et Aurélien Brest (2021) évalue à 39 % la part de la population française qui a consulté au moins un site de *fake news* au cours de l'automne 2020. Sur les réseaux sociaux, des enquêtes ont respectivement évalué à 11 % et 8 % la proportion d'utilisateurs ayant partagé des *fake news* sur Twitter (Osmundsen et al., 2021) et sur Facebook (Guess et al., 2019). Il faut noter cependant que ces proportions ne tiennent pas compte de la part d'individus exposée à des *fake news* sur les réseaux sociaux comme *TikTok* ou *Instagram* et encore moins au sein d'espaces de discussion privés via des applications de messagerie chiffrée telles que Telegram ou WhatsApp (Rossini et al., 2020 ; Rossini, 2023). Aussi, il est possible qu'une plus grosse proportion d'individus ait déjà été exposée à au moins une *fake news*. Cela étant, avoir été exposé à une *fake news* sur une période de six mois ou de 1 an ne veut pas dire grand-chose en soi. Quoi qu'il en soit, cela ne permet pas de conclure que les publics sont davantage exposés à des *fake news* qu'à des informations fiables et vérifiées. Pour évaluer, si l'écosystème informationnel contemporain peut vraiment être décrit par des termes comme ceux de « post-vérité » ou « d'infodémie », il faut comparer l'audience des *fake news* avec celle des *médias mainstream* et se demander le poids qu'elles représentent sur le total d'information consommée. Telles sont les réponses apportées par les deux prochains constats.

Le deuxième constat montre en effet que l'audience des *fake news* reste bien inférieure à celle des médias traditionnels mais varie selon les plateformes. Aux États-Unis, par exemple, les sites de *fake news* ont reçu quarante fois moins de visiteurs uniques que les médias traditionnels au cours de l'année qui a précédé l'investiture de Donald Trump (Nelson et Taneja, 2018). En France et en Italie, chaque site de *fake news* a en moyenne touché seulement 1 % de la population en ligne en 2017, alors que les sites des médias traditionnels, tels que *Le Figaro* (22,3 %), *Le Monde* (19 %), *Repubblica* (50,9 %) et *Il Corriere della Sera* (47,7 %), ont attiré des audiences bien plus larges (Fletcher et al., 2018). Par ailleurs, pendant la crise sanitaire, que cela soit aux États-Unis, en France, au Royaume-Uni ou en Allemagne, les sites de médias *mainstream* ont connu une augmentation de leur trafic beaucoup plus importante que celles des sites de *fake news* (Altay et al., 2022). Il est important de se demander néanmoins dans quelle mesure ces constats obtenus sur des sites Internet persistent lorsque l'analyse est centrée sur les réseaux sociaux. En effet, alors que la consommation d'informations sur le web résulte souvent de recherches intentionnelles, sur les réseaux sociaux l'exposition peut être accidentelle (Valeriani et Vaccari, 2016 ; Matthes et al., 2020). Le fil d'actualité d'un utilisateur est composé des contenus avec lesquels ses contacts ont interagi et dépend du type d'interactions privilégiées par l'algorithme de recommandation de chaque plateforme. Or, cet enchevêtrement entre des logiques algorithmiques et des dynamiques de sociabilité peut favoriser l'exposition des utilisateurs à des contenus qui visent davantage à provoquer des réactions de leur part qu'à les informer. Malgré ces nouveaux processus de mise en visibilité de l'information et les mécanismes d'exposition accidentelle qui en découlent, la majorité des études académiques obtiennent un constat similaire sur les réseaux sociaux et sur les sites web. Sur Facebook, d'une façon générale, les contenus provenant de médias traditionnels sont vus 5,5 fois plus souvent et génèrent 7,5 fois plus d'interactions que ceux provenant de sites de *fake news* (Guess et al., 2021). En revanche, des exceptions sont à noter pour quelques pages comme *La Gauche m'a Tuer* (1.5 million) ou *Le Top de L'Humour* (6 millions) dont les volumes d'engagement sont supérieurs à ceux des médias traditionnels. Le cas de *Santé + Mag*, en France, est particulièrement notable. Identifié comme diffuseurs de prescriptions et conseils médicaux erronés par le *Décodex* du *Monde*[22], ce média suscite autant d'interactions sur Facebook (i.e. 11,3 millions) à lui seul que le regroupement des cinq médias suivants combinés : *Le Monde* ; *Le Figaro* ; *Le Huffington Post* ; *20 Minutes* et *France TV info* (Fletcher et al., 2018). Plus encore, bien que la proportion d'engagement générés par les contenus issus de sites de *fake news* sur Facebook reste faible par rapport à celle des contenus issus de médias traditionnels, il est important de noter que les *fake news* suscitent plus d'interactions sur Facebook que sur des plateformes comme Twitter (Marchal et al., 2019 ; Hopp et al., 2020) et qu'elles y capturent une part d'engagement beaucoup plus élevée (entre 6,8 % et 14 %) par rapport à la proportion qu'elles occupent dans le trafic web global (entre 1,4 % et 2,3 % ; Altay et al., 2022). Ces résultats invitent donc les futures recherches à prêter attention aux divergences des pratiques des publics selon les espaces de communication sur lesquels ils consomment et partagent des informations.

Le troisième constat fait ressortir qu'au global, les *fake news* occupent une faible proportion dans les habitudes de consommation médiatique des individus et ne représentent qu'une infime proportion de tous les contenus avec lesquels ils interagissent. En effet, les enquêtes passées en revue montrent que le temps passé sur les sites de *fake news* est bien inférieur à celui des sites de médias traditionnels (Nelson et Taneja, 2018 ; Fletcher et al., 2018), atteignant au maximum une minute par jour. Même les études indiquant que les *fake news* parviennent à toucher une audience assez large (e.g. 44 %) révèlent qu'elles ne constituent que 5,9 % du total des informations consommées (Guess et al., 2020a). Aux États-Unis, l'enquête la plus complète est celle de Jennifer Allen et ses collègues (2020) qui montre que les *fake news* représentent 1 % de la consommation d'information d'actualité et 0,15 % du total de consommation médiatique. En France, des résultats similaires ont été trouvés par Laurent Cordonier et Aurélien Brest (2021), qui montrent que les *fake news* représentent

[22] Sénécat, A. (2018). Santé+ Magazine, un site emblématique de la « mal-information » sur la santé. *Le Monde*. https://www.lemonde.fr/les-decodeurs/article/2018/05/25/sante-magazine-un-site-emblematique-de-la-mal-information-sur-la-sante_5304505_4355770.html

0,16 % du temps total de connexion à Internet et 5 % du temps total d'information en ligne. Dans la même veine, sur des réseaux sociaux spécifiques comme Twitter, les *fake news* représentent seulement 1,18 % du total d'exposition à des informations politiques (Grinberg, 2019), 3 % de tous les tweets partagés et 4 % du total d'information d'actualité partagées (Osmundsen et al. 2021). Ce constat suggère ainsi de se décentrer des seules *fake news* consommées ou partagées par les individus et de s'intéresser plus largement aux autres contenus auxquels ils sont exposés en ligne ou qu'ils font circuler sur les réseaux sociaux. Bien que la plupart de ces contenus n'aient pas été évalués comme des *fake news*, s'agit-il pour autant d'informations fiables ? Ont-ils un impact sur les croyances, opinions, attitudes et comportements des individus ?

Le quatrième constat montre que la majorité des *fake news* est vue et partagée par une toute petite minorité d'internautes ayant des caractéristiques socio-démographiques spécifiques et des pratiques informationnelles et numériques non représentatives de celles du reste de la population. En effet, différentes études montrent que la majorité des sites de *fake news* a été consultée par une minorité d'internautes. Par exemple, 1 % des utilisateurs a consulté 65,3 % de toutes les pages web de sites de *fake news* (Zhou et al., 2023). De même sur Twitter, la majorité (i.e.~75-80 %) des *fake news* a été vue et partagée par une toute petite minorité d'utilisateurs (i.e. entre 0,1 % et 1 %) d'après les estimations de différentes études (Grinberg et al., 2019 ; Osmundsen et al., 2021 ; Baribi-Bartov et al., 2024). Ces utilisateurs partagent certaines caractéristiques distinctes : ils sont très politisés, plutôt conservateurs et âgés, utilisent les réseaux sociaux de façon intensive et font montre d'un intérêt élevé pour l'actualité et la politique (Nelson et Taneja, 2018 ; Guess et al., 2020a). Cependant, il est important de noter que ces données sont observationnelles ; aussi ne permettent-elles pas d'établir de relation causale entre les caractéristiques socio-politiques des individus et le partage et la consommation de *fake news* en ligne. Une lacune importante subsiste également en considérant que la majorité des études se concentrent sur les États-Unis, où l'espace médiatique est fortement polarisé (Benkler et al., 2018). Il est ainsi essentiel d'explorer si des schémas similaires se manifestent en France, compte tenu des différences potentielles dans le paysage médiatique et politique. Est-ce que les utilisateurs francophones présentent des caractéristiques qui les rendent plus susceptibles de consommer et partager des *fake news* ? Et, inversement, ceux qui ne sont pas touchés ont-ils des caractéristiques qui les rendent plus résistants ?

Ces différents résultats, principalement centrés sur des périodes de campagnes électorales occidentales ou de la pandémie de Covid-19, conduisent donc à nuancer les discours ayant recours aux termes de « post-vérité » et d'« infodémie » pour décrire l'écosystème informationnel contemporain. Toutefois, il est important de rappeler que la définition opérationnelle du mot *fake news,* sur laquelle repose la majorité des études passées en revue, présente différentes limites, notamment celle de restreindre les troubles de l'information et de la communication à la question de la factualité des énoncés. Or, même sans être exposés à des *fake news*, les individus peuvent être ciblés par des mécanismes d'influence. Par exemple, des informations factuelles peuvent être utilisées pour cadrer idéologiquement les débats et produire une représentation déformée de la réalité en mettant à l'agenda certains sujets d'actualité et certaines opinions politiques au détriment d'autres comme l'ont illustré de récents travaux sur des chaînes de télévision telles que *Fox News* (Benkler et al., 2018) et *CNews* (Cagé et al., 2022 ; Labarre, 2024). La prochaine section propose ainsi de passer en revue une série d'enquêtes ayant cherché à évaluer non pas seulement les effets des *fake news*, mais aussi ceux d'un ensemble plus large de contenus, tenant compte des transformations structurelles de l'écosystème informationnel contemporain.

### *Mesurer les effets des fake news*

Pour étayer leur façon d'interpréter les gros volumes d'interactions déclenchées par les *fake news* comme des indicateurs d'une crédulité de la part des utilisateurs des réseaux sociaux, de nombreux articles médiatiques ont souvent tendance à se référer à des résultats d'expériences ou de sondages montrant que les individus peinent à discerner le vrai du faux.[23] Les conclusions de ces études sont néanmoins à interpréter avec prudence pour trois raisons principales.

---

[23] Pour des exemples, voir les articles suivants :
Silverman, C. (2016, 7 Décembre). Most Americans who see fake news believe it, new survey says. *BuzzFeed News.* https://www.buzzfeed.com/craigsilverman/fake-news-survey?utm_term=.kdNMQZ6aD6_-_.obj74yp1zp
Shellenbarger, S. (2016, 21 Novembre). Most students don't know when news is fake, Stanford study finds. *The Wall Street Journal.* https://www.wsj.com/articles/most-students-dont-know-when-news-is-fake-stanford-study-finds-1479752576

Premièrement, il existe souvent une confusion entre les notions de croyances, d'opinions, d'attitudes et de comportements. Alors que celles-ci désignent des phénomènes bien distincts, elles sont fréquemment utilisées de façon interchangeable dans de nombreux articles (médiatiques comme académiques). Une récente recherche ayant passé en revue plus de 500 études expérimentales sur les effets des *fake news* montre que très peu d'entre elles examinent directement les effets des *fake news* sur les comportements (1 %) ou les intentions comportementales (10 %), mais utilisent plutôt des variables comme les croyances (78 %) et les attitudes (18 %) comme proxy (Murphy et al., 2023). Plusieurs de ces enquêtes tirent toutefois des conclusions concernant les conséquences des *fake news* sur des comportements réels comme le vote ou la vaccination. Or, il n'y a pas toujours de liens entre les croyances des individus et leurs comportements (Sperber, 1997 ; Mercier, 2020). Par exemple, des individus en désaccord avec Donald Trump peuvent quand même décider de voter pour lui. D'autres peuvent penser que les vaccins sont nocifs mais malgré tout se faire vacciner. Il faut donc faire attention à ne pas généraliser outre mesure des résultats provenant d'expérience ou de sondages et se demander ce qui est mesuré concrètement dans ces études.

Deuxièmement, les dispositifs méthodologiques des sondages et des expériences présentent des limites importantes, notamment en termes de validité écologique. Les situations d'exposition sont artificielles et éloignées de la réalité. En effet, plutôt que de chercher à reproduire l'écosystème informationnel réel dans lequel baignent les individus, celles-ci sont attachées à neutraliser l'effet d'un stimulus (e.g. la factualité d'une information, la fiabilité de sa source ou son orientation idéologique) en confrontant leurs participants à différentes conditions. Ceux-ci sont alors fréquemment exposés à 50 % d'informations vraies et à 50 % d'informations fausses et/ou à 50 % d'informations pro-démocrates et à 50 % d'informations pro-républicaines. Or, comme l'a montré la section précédente, les *fake news* occupent une faible place dans les habitudes de consommations médiatiques de la majorité des individus – dans tous les cas, elles sont loin de représenter 50 % des contenus auxquels ils sont exposés. Au cours de situations expérimentales ou de sondages, il est ainsi possible que les participants soient exposés de façon forcée à des *fake news* qu'ils n'auraient probablement jamais vues dans leur fil d'actualité. Il est également possible qu'ils se retrouvent contraints de répondre à des questions qu'ils ne se seraient jamais posés dans leur vie de tous les jours ou pour lesquelles ils n'ont pas d'avis tranché. En laissant un choix de réponse très limité à leurs participants, sans leur donner la possibilité de répondre par « je ne sais pas » ou « je ne suis pas sûr », ou en utilisant des formulations les encourageant à deviner, telles que « dans la mesure où vous savez…» ou « diriez-vous que…», les chercheurs risquent de surestimer la proportion d'individus dans l'erreur (Luskin et al., 2018).[24] Enfin, alors qu'ils sont isolés et détachés de tout contexte social, les individus qui participent à des études expérimentales ou répondent à des questionnaires n'ont pas non plus la possibilité d'avoir recours aux stratégies de vérification qu'ils sont susceptibles de mobiliser au cours de leur vie quotidienne pour jauger la qualité d'une information (e.g. solliciter l'avis d'un ami, vérifier la source d'un contenu, faire une recherche en ligne, etc.), d'autant plus que les contenus utilisés comme matériaux expérimentaux ne comportent qu'un titre, un chapeau et une image, mais pas toujours de source, de date ou de paratexte. Ainsi, bien que les protocoles expérimentaux soient plus rigoureux que les études observationnelles pour identifier la cause d'un phénomène, il est possible qu'ils arrivent à des conclusions peu réalistes si leur protocole repose sur une définition du phénomène elle-même peu réaliste (e.g. le fait d'exposer leurs participants à autant de *fake news* que d'informations vraies pour contrôler tous les paramètres alors même que la réalité de l'offre d'informations ne suit pas cet idéal de symétrie).

La troisième limite concerne la fiabilité des réponses déclaratives. Un écart important est fréquemment observé entre les réponses fournies par des individus à des questionnaires fermés et la réalité de leurs opinions ou de leurs comportements. Ce phénomène peut être attribué à divers facteurs. Tout d'abord, les individus peuvent ne pas se souvenir de façon précise de leurs pratiques. Par exemple, au cours d'une enquête, menée sur un échantillon représentatif d'Américains, 15 % des participants ont indiqué se souvenir de *fake news* qui avaient circulé pendant la campagne électorale de 2016 ; mais dans le même temps, 14 % d'entre eux ont également indiqué se souvenir d'avoir vu des *fake news* inventées de toute pièce par les chercheurs pour contrôler la fiabilité de leurs résultats (Allcott et Gentzkow, 2017). On peut ainsi se demander si les participants se souvenaient vraiment des *fake news* qu'ils ont dit avoir vues. Ensuite, les réponses des participants à des questionnaires sont aussi sujettes à des formes d'*expressive responding* ou de *partisan cheerleading*, c'est-à-dire qu'il arrive que des participants déforment sciemment leurs réponses aux questions d'un sondage (e.g. en prétendant croire à des informations qu'ils savent erronées) pour signaler leur soutien à leur camp politique (Bullock et al., 2013 ; Prior et al., 2015). Dans une étude, des chercheurs ont par exemple soumis à leurs participants deux photos d'inauguration présidentielle, celle de Barack Obama en 2009 et celle de Donald Trump en 2017, puis leur ont demandé d'indiquer quelle photo comportait la plus grande foule. Même si la réponse à cette question était plus qu'évidente, les chercheurs ont montré que 26 % des individus républicains sélectionnaient malgré tout la photo représentant l'inauguration de Donald Trump et ce d'autant plus s'ils étaient pourvus d'un diplôme d'études supérieures, ce qui suggère que les individus qui participent à des études scientifiques peuvent très bien déceler l'intention des chercheurs, et ainsi fournir sciemment des réponses erronées pour affirmer leur identité partisane (Schaffner et Luks, 2018). Enfin, les biais de désirabilité sociale peuvent inciter les participants à des études à fournir des réponses socialement acceptables plutôt que des réponses sincères, et donc les conduire à masquer certaines de leurs croyances ou pratiques informationnelles.

[24] Une étude ayant analysé les questionnaires utilisés dans 180 enquêtes montre que plus de 90 % d'entre elles ne donnaient pas la possibilité à leurs participants de répondre par « je ne sais pas » ou « je ne suis pas sûr » et utilisaient des formulations les encourageant à deviner, telles que « dans la mesure où vous savez...» ou « diriez-vous que...». Cette absence d'option « je ne sais pas » et cet encouragement à deviner conduisent à surestimer la proportion de réponses incorrectes de neuf points (25 % au lieu de 16 %), et de 20 points en ne considérant que les personnes qui déclarent être confiantes dans leur réponse (25 % au lieu de 5 %).

Pour dépasser les différentes limites méthodologiques inhérentes aux études expérimentales effectuées en laboratoire, deux approches de recherche peuvent être mobilisées. La première repose sur des méthodes d'économétrie. Celles-ci utilisent des variables instrumentales qui exploitent des sources de variation exogènes influençant la variable indépendante (e.g. l'exposition aux *fake news*) mais n'affectant pas directement la variable dépendante (e.g. les croyances ou comportements des individus) autrement que par cette exposition. Cette approche est particulièrement utile pour inférer des liens de cause à effet dans des contextes où les essais randomisés ne sont pas faisables ou éthiquement acceptables. La seconde approche repose sur la passation d'expériences de terrain directement sur les réseaux sociaux (Marres et Stark, 2020). D'une façon générale, ces expériences de terrain consistent à soumettre de façon randomisée des utilisateurs à différents types de traitements (par exemple, via des messages privés ou des publications publiques), puis à mesurer leurs réactions directement sur les réseaux sociaux ou via des questionnaires, ce qui permet de garantir leur validité interne, externe et écologique et de tirer des inférences causales (Mosleh et al., 2022). Difficiles à mettre en œuvre, ces enquêtes nécessitent souvent des partenariats avec des entreprises et des plateformes privées. Par ailleurs, elles soulèvent de nombreux enjeux éthiques dans la mesure où il est problématique de tester l'effet d'un traitement, tel que l'exposition à des *fake news*, dont on présuppose qu'il aura des effets négatifs sur les individus.

En prenant en compte ces enjeux méthodologiques, nous avons passé en revue près de vingt articles d'économétrie ou d'expériences de terrain ayant cherché à évaluer, en dehors de situations de laboratoire, les effets de différents types de contenus, fréquemment perçus comme problématiques pour le débat public (e.g. informations partisanes ; publicités politiques ; *fake news ;* etc.). De nombreuses informations précisant les tailles des échantillons, les données, les corpus et les résultats de ces enquêtes ont été listées dans le tableau situé en Annexe 2. Ci-dessous, nous rendons compte des conclusions principales avancées par ces études.

Une première série d'études s'est penchée sur les effets de chaînes de télévision partisanes, telles que *Fox News* et *MSNBC*. L'intérêt de ces études réside dans l'attention qu'elles portent aux effets structurels d'une exposition répétée sur le long terme à des discours idéologiques diffusés par des médias à forte audience, plutôt que de se focaliser sur l'effet immédiat d'un contenu unique. Par rapport à cette question, la plupart des études montrent que l'exposition à des chaînes de télévision comme *Fox News* a des effets sur les opinions politiques et les comportements électoraux des individus. Bien que ces effets soient relativement faibles en termes de magnitude (de 0,15 à 0,7 points sur les votes), il est possible qu'ils aient été décisifs dans le cas d'élections présidentielles très serrées comme celle de l'année 2000 aux États-Unis. Par exemple, Stefano DellaVigna et Ethan Kaplan (2007) ont montré que l'entrée de *Fox News* dans différentes villes américaines a augmenté la part des votes républicains de 0,4 à 0,7 points entre 1996 et 2000. Par la suite, Gregory Martin et Ali Yurukoglu (2017) et Elliott Ash et al. (2022) ont obtenu des résultats similaires en utilisant la position de *Fox News* dans la liste des chaînes de télévision par câble comme variable instrumentale. Elliott Ash et al. (2024) ont aussi trouvé des effets sur le respect des mesures de distance sociale. Cependant,

Jens Hainmueller (2012) n'est pas parvenu à répliquer les résultats de Stefano DellaVigna et Ethan Kaplan (2007) en ayant recours à une autre méthode statistique, et Daniel Hopkins et Jonathan Ladd (2013) n'ont trouvé d'effets que sur les intentions de vote des Républicains et des indépendants, mais pas sur l'ensemble de la population. Enfin, David Broockman et Joshua Kalla (2022) ont quant à eux mis en évidence des effets sur les connaissances et les opinions à court terme mais pas sur les attitudes à long terme.

Une deuxième série d'études s'est attachée à mesurer plus spécifiquement les effets de messages de campagnes ou de publicités politiques. Au global, les résultats de ces études indiquent que les effets sont très limités. À partir d'un essai contrôlé randomisé ayant consisté à exposer 61 millions d'utilisateurs de Facebook à un message de mobilisation politique lors des élections législatives américaines de 2010, l'étude de Robert Bond et al. (2016) montre une augmentation de 2,08 % des déclarations de vote, de 0,26 % de la recherche d'informations et de 0,39 % sur la participation électorale. *A contrario*, Jörg Spenkuch et David Toniatti (2018) soulignent que la publicité politique aux États-Unis n'a quasiment aucun impact sur la participation globale des électeurs, mais qu'elle joue un rôle crucial dans l'augmentation des parts de voix des candidats, en modifiant la composition du corps électoral qui se rend aux urnes. Enfin, si l'étude de David Broockman et Donald Green (2014) ne fait apparaître quasiment aucun effet significatif, c'est surtout la méta-analyse conduite par Joshua Kalla et David Broockman (2018), recensant les résultats de quarante expérimentations de terrain et de neuf enquêtes originales, qui permet de conclure à une absence d'effet sur le long terme des messages de campagne ou de publicités politiques.

Une troisième série d'enquêtes a examiné l'effet des filtres exercés par les algorithmes des plateformes (Guess et al., 2023a ; González-Bailón et al., 2023) ou des mesures préconisées dans les discours publics pour améliorer la qualité de l'écosystème informationnel, telles que faciliter l'exposition à des contenus adverses (Bail et al., 2018 ; Levy, 2020) ; réduire l'exposition aux contenus idéologiquement similaires (Nyhan et al., 2023), classer les contenus affichés sur le fil d'actualité d'un utilisateur par ordre chronologique inversé plutôt que selon les paramètres des algorithmes de classement des plateformes (Guess et al., 2023b). La majorité de ces études montre que ces différents paramètres exercent une influence sur les contenus auxquels les utilisateurs des réseaux sociaux sont exposés dans leur fil d’actualité, ainsi que sur les taux d’engagement, mais n’ont pas d’effet durable sur les opinions et attitudes politiques. Par exemple, Brendan Nyhan et al. (2023) montrent qu’une diminution de l’exposition à des contenus venant de sources pro-attitudinales (de 53,7 % à 36,2 %) réduit l’exposition à des contenus incivils (de 3,15 % à 2,81 %), comportant des insultes (de 0,034 % à 0,030 %) et à des *fake news* (de 0,76 % à 0,55 %), mais n’a pas d’effet sur différentes mesures attitudinales, telles que la polarisation affective. De façon intéressante, certaines mesures préconisées dans les discours publics semblent même avoir des effets contre-intuitifs et indésirables. Par exemple, l’étude de Chris Bail et al., (2018) a montré que l’exposition à des contenus adverses pouvait amener les sujets (en particulier les Républicains) à devenir plus extrêmes, plutôt que plus modérés, dans leurs opinions exprimées dans les enquêtes de suivi. En mobilisant un design d’enquête similaire, cependant, l’enquête de Ro’ee Levy (2020) ne fait apparaître aucun changement d’opinion mais observe une atténuation des attitudes émotionnelles à l’égard du camp adverse. Ainsi, il semble que le fait d’être confronté à des vues opposées ne provoque pas toujours une attitude introspective permettant de mieux mettre en balance les idées en compétition et d'éventuellement tempérer ses opinions.

Enfin, une étude très récente a cherché à quantifier l’effet de contenus liés aux vaccins sur les intentions de vaccination des individus en proposant un nouveau cadre méthodologique (Allen et al., 2024). Après avoir mesuré l'effet de 130 contenus portant sur les vaccins, à travers deux expériences, les auteurs ont utilisé le jeu de données *Social Science One* de Facebook pour mesurer l'exposition à 13 206 URLs liées aux vaccins. Ensuite, ils ont développé une méthode intégrant du *crowdsourcing* et du traitement automatique du langage naturel (NLP) pour prédire l'effet de chaque URL Facebook à partir des résultats des deux expériences. L’étude fait ressortir que l'exposition à une seule information erronée sur les vaccins est susceptible de faire diminuer les intentions de vaccination de 1,5 point en moyenne. Mais en tenant compte du nombre de vues des contenus, ceux sceptiques à l'égard des vaccins n’ayant pas été qualifiés de *fake news* par des *fact-checkers* ont réduit les taux de vaccination de −2,28 points comparé à −0,05 point de pourcentage pour les *fake news*. Autrement dit, l'impact des *fake news* était 50 fois moins important que celui de contenus sceptiques à l'égard des vaccins mais non qualifiés de *fake news* par des *fact-checkers.* L’intérêt de cette étude est de ne pas réduire les troubles de l’information et de la communication à la seule question de la factualité des énoncés en montrant comment des informations factuelles venant de média *mainstream* peuvent avoir un impact plus négatif que des *fake news* sur les attitudes vaccinales, notamment parce qu'elles bénéficient d'une plus grande visibilité. Néanmoins, il faut être prudent avant de tirer des implications normatives à partir de ces résultats. En effet, il est problématique de conclure que les contenus sceptiques à l'égard des vaccins venant des médias *mainstream* ne devraient pas être publiés. De nombreuses critiques sur les vaccins sont légitimes et fondées ; les invisibiliser pourraient contribuer à renforcer la méfiance des publics dans les médias *mainstream* et les institutions.

En définitive, la conclusion principale qui ressort des études passées en revue ci-dessus est qu'il est très difficile de mesurer une relation causale entre une exposition à des contenus médiatiques et différentes variables cognitives, attitudinales ou comportementales. Si plusieurs études trouvent des effets, ceux-ci sont dans l'ensemble limités. Plus encore, il est possible que ceux-ci renforcent davantage qu'ils ne causent réellement l'adoption de nouvelles croyances, opinions, attitudes et comportements. En effet, les contenus problématiques touchent souvent des individus qui sont déjà convaincus (Guess et al., 2020b). En fait, bien qu'elles permettent de mesurer de façon robuste une relation de cause à effet, les expériences de terrain et les études d'économétrie ne donnent pas toujours la possibilité d'identifier, et surtout de comprendre, les mécanismes précis qui sous-tendent l'effet observé, notamment lorsque celui-ci est contre-intuitif. Par exemple, c'est grâce aux entretiens menés par son équipe de chercheurs que Chris Bail (2021) est parvenu à expliquer pourquoi l'exposition à des contenus adverses augmentait la polarisation plutôt qu'elle ne l'atténuait. Ainsi, il peut être intéressant de se tourner vers des approches qui ne visent pas à mesurer de lien de cause à effet unidirectionnel à un *instant t* entre un contenu unique et un individu isolé, mais plutôt vers celles qui cherchent à décrire la variété des pratiques des publics en tenant compte de leur contexte social. Autrement dit, il peut être intéressant de dépasser la question des effets des médias et de se tourner vers des approches de réception.

## *Fake news* : les mystères de leur réception

Si les constats empiriques présentés dans la section précédente apportent des nuances importantes aux préoccupations exprimées dans les discours publics concernant la portée et l'impact des *fake news*, ils n'offrent cependant qu'une vue très partielle de la manière dont celles-ci sont véritablement reçues par les publics. Limitées à des mesures de quantification, établies à partir d'indicateurs d'audience (Livingstone, 2019) ou de questionnaires fermés, ces études ne permettent pas de décrire de façon fine et compréhensive comment les individus interprètent les *fake news* ni pourquoi ils y réagissent (ou non). Pour répondre à ces questions, quelques premières enquêtes qualitatives ont cherché à mobiliser les apports théoriques et méthodologiques des travaux de sociologie de la réception. Nous aurons l'occasion de revenir de façon plus approfondie sur les contributions de ce domaine de recherche au moment de construire le cadre théorique et méthodologique de notre thèse (cf. chapitre 2), mais commençons d'ores et déjà par présenter les principaux résultats des premières enquêtes sur les *fake news* qui s'inscrivent dans ce sillon.

### Enquêtes préliminaires

Un premier résultat montre l'intérêt de ne pas réduire les processus de réception de l'information à la question des effets des *fake news*, mais de questionner plus largement les usages sociaux qui les entourent (Duffy et al., 2019). Par exemple, plutôt que de nous arrêter aux gros volumes d'engagement déclenchés par la page Facebook de *Santé + Mag* et de conclure que ceux-ci étaient le signe d'une vulnérabilité importante des utilisateurs à des fausses informations de santé, nous avons analysé, dans le cadre de notre mémoire de fin d'études, environ 500 contenus et 5 000 commentaires publiés sur cette page (Berriche et Altay, 2020). Nos résultats ont montré que les utilisateurs étaient moins réceptifs aux conseils médicaux infondés ou erronés qu'elle publiait qu'à ses « panneaux de citations » (Pasquier, 2018) contenant des formules phatiques (Akrich et Méadel, 2009). Par ailleurs, de nombreux commentaires émis dans un registre humoristique ont été relevés en réaction à des contenus portant sur divers sujets de santé. Concernant les *fake news* politiques, une enquête de Romain Badouard (2021) suggère que les fausses informations constituent moins des outils d'influence que des supports de prise de parole pour des individus déjà convaincus qui partagent des *fake news* pour exprimer des positions politiques. Au-delà de mobiliser des régimes d'énonciation de l'ordre de la conversation informelle et de l'indignation politique, les individus peuvent aussi s'exprimer dans un registre critique face à différents types de contenus, dont certaines *fake news*. Par exemple, des focus groupes menés dans différents pays (e.g. Espagne, Allemagne, Royaume-Uni et États-Unis) font ressortir que les individus adoptent fréquemment une attitude de « scepticisme généralisé » à l'égard des choix éditoriaux des médias ainsi que des informations sélectionnées par des algorithmes de classement sur les réseaux sociaux (Fletcher et Nielsen, 2019). Un scepticisme analogue a également été relevé dans le cadre d'une enquête reposant sur la conduite de 71 entretiens aux États-Unis. Au cours de ces entrevues, les personnes interviewées ont indiqué : (1) avoir une perception globalement négative de la qualité des informations produites par les médias ; (2) se méfier particulièrement des informations qui circulent sur les réseaux sociaux ; et (3) s'inquiéter des effets négatifs que celles-ci sont susceptibles d'exercer sur les autres. En raison de ce scepticisme, les participants ont déclaré avoir recours à différentes stratégies de vérification comme le fait : (1) de se tourner vers des *fact-checks* ou des médias traditionnels ; (2) de croiser leurs sources ; ou (3) de se fier à certains de leurs contacts personnels considérés comme des « leaders d'opinion » (Wagner et Boczkowski, 2019). Enfin, plusieurs enquêtes d'inspiration ethnographique ont montré que les groupes de personnes, souvent disqualifiés dans le discours public comme « irrationnels », « crédules » ou « anti-science », tels que les « complotistes » ou les « anti-vax » (Ward et al., 2019), sont susceptibles de déployer des stratégies de vérification sophistiquées pour « fact-checker » les informations à leur manière (Tripodi, 2018 ; Marwick et Partin, 2022) ou produire une « contre-expertise » objectiviste (Ylä-Anttila, 2018 ; Lee et al., 2021). Par exemple, dans le cadre d'une étude exploratoire menée au tout début de notre thèse, nous avons observé, après avoir analysé de façon approfondie 15 000 commentaires Facebook émis en réaction à des *fake news* et des *fact-checks* sur les vaccins, que les utilisateurs exprimant des positions critiques à l'égard des vaccins citaient beaucoup plus d'études scientifiques dans leurs commentaires pour étayer leurs arguments que les utilisateurs favorables aux vaccins (Berriche, 2021).

Le deuxième résultat montre que les pratiques informationnelles et conversationnelles des publics varient selon des variables socioculturelles et situationnelles. Dans le cadre de notre mémoire de fin d'étude, par exemple, nous avons mis en place un dispositif d'enquête visant à conduire des entretiens semi-directifs à l'aide d'un plateau de mise en situation. Ce plateau permettait à nos enquêtés de désigner les interlocuteurs et les situations sociales dans lesquelles ils étaient susceptibles de partager différentes catégories d'informations (des *hard news* ou des *soft news*, soit vraies, soit fausses). Il est apparu que dans certaines situations nos enquêtés étaient moins concernés par la qualité épistémique d'une information que par sa valeur sociale et sa propension à alimenter des conversations. En effet, au cours des entretiens, plusieurs participants ont décidé de mettre en circulation des informations douteuses ou erronées afin de les utiliser comme des ressources conversationnelles au cours de leurs sociabilités quotidiennes (Berriche, 2019). Le même dispositif a été repris auprès d'élèves d'une classe de 4ème et a donné lieu à des résultats similaires (Berriche, 2023). Enfin, une enquête de Julien Boyadjian (2020), menée auprès de publics socialement et culturellement différenciés, montre que, contrairement aux préjugés sociaux fréquemment associés au partage de *fake news*, les jeunes issus de milieux populaires sont plus enclins à la non-information qu'à la désinformation. À l'inverse, les jeunes de milieux favorisés sont plus politisés et plus susceptibles d'être sujets à des mécanismes d'exposition sélective.

## Un double paradoxe

Les constats empiriques, issus des enquêtes quantitatives et des premières enquêtes de réception, passées en revue dans les sections précédentes, apportent des nuances importantes aux discours publics qui entourent les débats sur les *fake news.* En effet, à cause des transformations provoquées par le numérique sur la circulation et la consommation de l'information, ceux-ci dépeignent l'écosystème informationnel comme un marché dérégulé et les utilisateurs des réseaux sociaux comme des foules d'individus crédules, systématiquement induits en erreur par leurs biais cognitifs. Or, si ces grilles de lecture étaient réalistes, les *fake news* devraient être davantage prévalentes que les vraies informations sur l'ensemble des espaces de communication de l'écosystème informationnel numérique et devraient affecter de façon indifférenciée n'importe quel utilisateur des réseaux sociaux. Pourtant, ce n’est pas ce qui ressort des enquêtes empiriques menées sur le sujet. Si une part non négligeable des populations a pu à un moment donné être exposée à une *fake news*, seule une minorité a réagi publiquement à ces contenus en les likant, partageant ou commentant. Par ailleurs, ces réactions ne signifient pas forcément que les individus adhèrent ou sont influencés par ces contenus. Au contraire, elles peuvent recouvrir une variété de significations.

Ce décalage entre les discours publics sur les *fake news* et les constats empiriques de la littérature académique montre les limites des grilles de lecture qui expliquent différents enjeux contemporains uniquement par des causes technologiques ou cognitives, sans avoir adopté au préalable une approche descriptive et compréhensive (Lemieux, 2014), et conduit à soulever deux paradoxes :

(1) Pourquoi la majorité des enquêtes empiriques montre que les *fake news* ne représentent qu’une petite proportion du total d’informations consultées et partagées par les utilisateurs des réseaux sociaux alors que ces derniers ne sont ni soumis à un contrôle éditorial, ni à des règles de déontologie journalistique ?

(2) Comment comprendre la montée de la polarisation politique alors que les utilisateurs ne semblent pas si réceptifs aux *fake news* ?

Afin d’expliquer ces deux paradoxes, deux enquêtes empiriques ont été conduites dans le cadre de cette thèse. Celles-ci sont centrées sur le cas de la France entre 2016 et 2022 et sont consacrées à étudier les mécanismes sociaux qui modulent les réactions des utilisateurs des réseaux sociaux face aux *fake news*, notamment ceux qui favorisent ou non leur propension à faire preuve de distance critique.

## Plan de la thèse

Après avoir adopté une approche socio-historique et constructiviste, la première partie de la thèse s’attache à délimiter un cadre théorique et méthodologique permettant d’appréhender l’écosystème informationnel contemporain et ses publics d’une manière qui soit conformes aux constats empiriques de la littérature académique.

Le premier chapitre montre que le terme *fake news* ne désigne pas tant un nouveau phénomène social qu'un nouveau « problème public », si ce n'est une nouvelle « panique morale ». Une étude de cas conduite sur la couverture médiatique du documentaire *Hold-up* permet d'illustrer concrètement comment les discours publics contemporains réactivent des grilles de lecture techno-déterministes et psychologisantes, similaires à celles adoptées lors de précédentes « paniques médiatiques et technologiques ». Nous documentons ensuite comment cette manière d'aborder le problème des *fake news* a conduit à une focalisation de l'action publique sur des mesures de régulation ou d'éducation à l'esprit critique, et à orienter les recherches académiques vers des approches de *big data* ou de psychologie cognitive, sans que ne soient vraiment pris en compte les enseignements issus des travaux de sociologie des médias et de la communication. En appliquant le concept de « panique morale » aux discours publics sur les *fake news*, ce chapitre ne vise pas à remettre en cause l'existence d'un problème relatif à la qualité de l'information en circulation dans l'espace public, mais souligne plutôt la nécessité de ne pas s'arrêter à l'unique manière dont celui-ci a été défini par divers acteurs de l'espace politico-médiatique afin de l'aborder sous un angle permettant de ne pas occulter les mécanismes sociaux qui le sous-tendent.

L'enjeu principal du deuxième chapitre est ainsi de construire un cadre théorique et méthodologique permettant de dépasser les limites des grilles de lecture techno-déterministes et psychologisantes issues des discours médiatiques et institutionnels afin de partir de représentations de l'écosystème informationnel contemporain et de ses publics qui soient plus ajustées aux constats empiriques de la littérature académique. Nous montrons l'intérêt de mobiliser un cadre théorique et méthodologique développé par la sociologie de la réception et la sociologie pragmatique afin de tenir compte à la fois du contexte social des récepteurs d'information et du rôle actif que ceux-ci jouent dans la construction du sens d'un message, ainsi que de leur capacité à adapter leurs prises de parole aux contraintes énonciatives des situations d'interactions dans lesquelles ils se trouvent. Nous expliquons également pourquoi ces approches nécessitent certains ajustements conceptuels et méthodologiques, prenant appui notamment sur les apports des travaux de sociologie des usages du numérique, afin d'être plus adaptées au caractère discontinu, nomade et désynchronisé des nouvelles situations de réception engendrées par les transformations numériques de l'écosystème informationnel. Nous exposons ensuite l'intérêt d'articuler des méthodes quantitatives à des méthodes qualitatives, puis présentons les terrains d'enquête que nous avons investis et les matériaux que nous avons recueillis.

La deuxième partie part à la rencontre des utilisateurs qui partagent des *fake news* sur les réseaux sociaux. Elle restitue les résultats d'une enquête menée sur Twitter qui, en articulant des analyses quantitatives de traces numériques à des observations en ligne et des entretiens, a permis d'identifier les caractéristiques spécifiques de la majorité de ces utilisateurs, tout en veillant à ne pas les réduire au fait d'avoir partagé une *fake news* sur un réseau social particulier, ainsi qu'à examiner la variété de leurs pratiques (informationnelles comme conversationnelles) au sein de différentes situations d'interactions (en ligne comme hors ligne).

Le chapitre 3 confirme l'existence d'un écart entre la façon dont les utilisateurs qui partagent des *fake news* sur les réseaux sociaux en France sont dépeints dans les discours publics et leurs caractéristiques réelles. Il montre que la consommation de *fake news* est loin d'affecter de façon aléatoire l'ensemble des utilisateurs des réseaux sociaux, mais n'est en réalité observable que pour un groupe restreint d'internautes dont la particularité n'est pas d'être moins éduqués ou moins dotés en compétences cognitives que les autres, mais d'être davantage politisés et critiques à l'égard des institutions. Bien que minoritaires, ces utilisateurs sont cependant susceptibles de faciliter la mise à l'agenda des opinions défendues par leur camp politique dans le débat public en raison de leur hyperactivité en ligne et des très nombreuses informations d'actualité qu'ils partagent.

Le chapitre 4 approfondit les résultats du chapitre précédent en concentrant ses analyses sur les pratiques informationnelles et conversationnelles de la minorité d'utilisateurs à l'origine du partage de *fake news* sur Twitter. Après avoir décrit comment les comportements de ces utilisateurs, et notamment leurs régimes d'énonciation, sont modulés par leur position dans l'espace social, tout en étant susceptibles de varier d'une situation d'interactions à l'autre, ce quatrième chapitre met en évidence une compétence de distance critique, la prudence énonciative, que parviennent à mobiliser les utilisateurs des réseaux sociaux, même ceux qui relaient des *fake news*, pour préserver leur réputation et maintenir leur intégration dans différents espaces sociaux.

La troisième partie part à la rencontre des utilisateurs des réseaux sociaux qui interviennent dans le flux d'une conversation pour formuler des désaccords ou des corrections. Elle restitue les résultats d'une enquête menée sur Facebook qui, en articulant de nouveau des analyses quantitatives de traces numériques à des observations en ligne et des entretiens, a permis d'identifier et d'analyser un ensemble de commentaires critiques exprimés sur Facebook en réaction à des contenus perçus comme des *fake news*. Cette approche a également permis de prêter attention aux profils des utilisateurs à l'origine de ces commentaires critiques, aux espaces de conversation favorisant ou limitant leur expression, ainsi qu'aux conséquences engendrées par ces interventions dans les discussions qui s'ensuivent.

Le chapitre 5 introduit le concept de « point d'arrêt » pour qualifier des formes de lecture oppositionnelle ou critique qui se traduisent par l'expression publique de désaccords, de signalements, de corrections ou de dénonciations au cours d'échanges en ligne. Après avoir présenté les perspectives théoriques mobilisées pour conceptualiser la notion de « point d'arrêt », le chapitre détaille l'approche méthodologique déployée pour identifier ces formes d'expression au sein de commentaires Facebook, ainsi que pour tenir compte de la multiplicité et de l'hétérogénéité des espaces de communication qui existent sur le réseau social. Les différentes méthodes mobilisées ont ensuite permis d'examiner comment la propension des utilisateurs à exprimer des points d'arrêt, ainsi que leurs régimes d'énonciation, varient selon les divers espaces de communication identifiés sur Facebook. Les résultats montrent que les pratiques consistant à exprimer des points d'arrêt sont plus fréquentes au sein d'espaces de conversation peu politisés et/ou bénéficiant d'une audience importante et/ou rassemblant des personnes avec des points de vue hétérogènes. Ils soulignent également comment, au-delà de ces variables situationnelles, des mécanismes sociaux, tels que le fait d'être membre d'un groupe de modération collective, sont aussi susceptibles d'encourager l'expression de points d'arrêt. En somme, les réseaux sociaux ne constituent donc pas un univers de pratiques homogènes totalement dépourvus de règles. Ils sont plutôt composés d'une pluralité d'espaces, régis par différents contrats de communication et normes d'interactions, favorisant ou entravant des mécanismes d'autorégulation conversationnelle.

Le chapitre 6 approfondit les résultats du chapitre précédent en analysant les profils des utilisateurs des réseaux sociaux qui expriment des points d'arrêt sur Facebook et les réactions que cela déclenche dans la suite des conversations. Tout comme le partage de *fake news*, l'expression de points d'arrêt est l'apanage d'une minorité d'utilisateurs, généralement très actifs sur les réseaux sociaux, plutôt jeunes, politisés et éduqués. Par ailleurs, les points d'arrêt exprimés dans les commentaires Facebook restent dans la plupart du temps sans réponse, quand ils ne déclenchent pas des réactions injurieuses et négatives. Ainsi, bien que les utilisateurs des réseaux sociaux soient en mesure de mobiliser des formes de distance critique au cours d'échanges en ligne, celles-ci permettent rarement l'émergence de véritables débats délibératifs, pas plus que l'expression d'un pluralisme agonistique, mais donnent plutôt lieu à des dialogues de sourds entre une minorité d'utilisateurs particulièrement actifs en ligne. Cela étant, la rareté des réponses suscitées par les points d'arrêt ne signifie pas que les utilisateurs qui y sont exposés y sont totalement indifférents. L'absence de trace numérique est peut-être le signe d'une attention silencieuse ou d'un évitement délibéré de leur part – une hypothèse qui invite les futures recherches à sortir des réseaux sociaux afin de partir à la rencontre de ces utilisateurs invisibles et silencieux.

Les analyses produites à travers chacun de ces six chapitres permettront d'achever notre thèse par une conclusion générale. Après avoir synthétisé les principaux résultats issus des deux enquêtes empiriques réalisées auprès d'utilisateurs de Twitter et Facebook, des explications peuvent être apportées aux deux paradoxes qui ont guidé la rédaction de cette thèse. Ces résultats permettent de formuler deux arguments plus généraux contribuant à enrichir les conceptualisations des notions d'espace public et de rationalité. Des pistes pour de futures recherches sont ensuite proposées pour dépasser les limites de notre dispositif méthodologique. Enfin, des recommandations pour le débat public sont suggérées à partir de bases conceptuelles et empiriques apportées par la thèse, ainsi que d'autres travaux de recherche en sciences sociales, pour reformuler le problème des *fake news* et réorienter l'action publique.

# Première partie. D’un problème public à une enquête de sociologie

Après avoir analysé comment les *fake news* ont été construites en problème public, la première partie de la thèse s’attache à délimiter un cadre théorique et méthodologique permettant d’enquêter sur la réception de *fake news* par les utilisateurs des réseaux sociaux dans l’écosystème informationnel contemporain.

Étant donné l’écart important mis en avant dans l’introduction entre les discours publics sur les *fake news* et les résultats de la littérature académique, le premier chapitre vise à prendre de la distance avec les représentations alarmistes des écosystèmes informationnels et de leurs publics véhiculées dans les médias ces dernières années. En adoptant une approche socio-historique et constructiviste, il montre que le terme *fake news* ne désigne pas tant un nouveau phénomène social qu’un nouveau « problème public »[25], si ce n’est une nouvelle « panique morale ». En effet, réinscrire les débats sur les *fake news* dans une continuité historique permet de faire ressortir que celles-ci ne constituent pas totalement une rupture par rapport à des troubles informationnels passés.[26] Ces débats témoignent plutôt d’une résurgence des préoccupations des élites et des médias face à l’émergence de nouveaux circuits de l’information et de la communication. En expliquant différents enjeux contemporains principalement par la circulation virale de *fake news* sur les réseaux sociaux et par la crédulité de leurs utilisateurs, les discours publics actuels réactivent des grilles de lecture techno-déterministes et psychologisantes, similaires à celles adoptées lors de précédentes « paniques médiatiques et technologiques ». En appliquant le concept de « panique morale » aux discours publics sur les *fake news*, ce chapitre ne vise pas à remettre en cause l'existence d'un problème relatif à la qualité de l'information en circulation dans l'espace public, mais souligne plutôt la nécessité de ne pas s'arrêter à l'unique manière dont celui-ci a été défini par divers acteurs de l'espace politico-médiatique afin de l'aborder sous un angle permettant de ne pas occulter les mécanismes sociaux qui le sous-tendent.

---

[25] Cet argument est traité de façon plus approfondie dans le cadre des recherches doctorales menées par Ysé Vauchez depuis 2018 au CESSP/CRPS. https://theses.fr/s224806

[26] La notion de troubles informationnels fait référence ici à des périodes connues pour avoir été marquées par une circulation importante d'informations contestées et contradictoires, accusées d’avoir engendré des perturbations politiques, économiques ou sanitaires. Comme l’a souligné l’historien Marc Bloch (1921, p. 14) : « les fausses nouvelles, dans toute la multiplicité de leurs formes – simples racontars, impostures, légendes – ont [toujours] rempli la vie de l’humanité ». Pour des exemples, voir :
Le Bras, S. (2018, 24 septembre). Les fausses nouvelles : une histoire vieille de 2 500 ans. *The Conversation*. https://theconversation.com/les-fausses-nouvelles-une-histoire-vieille-de-2-500-ans-101715
Darnton, R. (2017, 20 février). La longue histoire des “fake news”, *Le Monde*. http://www.lemonde.fr/idees/article/2017/02/20/la-longue-histoire-des-fake-news_5082215_3232.html

L'enjeu principal du deuxième chapitre est ainsi de construire un cadre théorique et méthodologique permettant de dépasser les limites des grilles de lecture techno-déterministes et psychologisantes issues des discours médiatiques et institutionnels afin d'appréhender les écosystèmes informationnels contemporains et leurs publics d'une manière qui soit plus en phase avec les constats des études empiriques. Nous montrons l'intérêt de mobiliser un cadre théorique et méthodologique développé par la sociologie de la réception et la sociologie pragmatique afin de tenir compte du contexte social des récepteurs d'information et du rôle actif que ceux-ci jouent dans la construction du sens d'un message, ainsi que de leur capacité à adapter leurs prises de parole aux contraintes énonciatives des situations dans lesquelles ils se trouvent. Nous expliquons également pourquoi ces approches nécessitent certains ajustements conceptuels et méthodologiques, prenant appui sur les apports des travaux de sociologie du numérique, afin d'être plus adaptées au caractère discontinu, nomade et désynchronisé des nouvelles situations de réception engendrées par les transformations numériques des écosystèmes informationnels. Nous exposons ensuite l'intérêt d'articuler des méthodes quantitatives à des méthodes qualitatives, puis présentons les terrains d'enquête que nous avons investis et les matériaux que nous avons recueillis.

# Chapitre 1. La construction d’un problème public

Ce premier chapitre a pour objectif de clarifier le statut des *fake news* en tant qu’objet de recherche en les analysant tout d’abord comme un objet de discours.

L’apparition soudaine de la notion de *fake news* dans le débat public invite en effet à s’interroger sur le caractère inédit de ce qui est désigné. Pourquoi le terme a-t-il bénéficié d’un tel retentissement dans le débat public alors que différents mots existaient déjà pour faire état de troubles informationnels ? Quels acteurs ont contribué à rendre la notion populaire et comment la définissent-ils ?

Ces premières interrogations invitent à puiser dans des travaux d’historiens, d’anthropologues, de sociologues ou de politistes ayant étudié d’anciens troubles de l’information et de la communication, ainsi qu’à mobiliser en toile de fond un cadre d’analyse issu des travaux de sociologie des problèmes publics (Blumer, 1971 ; Spector et Kitsuse, 1977 ; Cefaï, 1996 ; Gusfield, 2009). En effet, tandis que les premiers permettent d’explorer si les inquiétudes associées à divers troubles informationnels ont évolué au fil du temps, face à l’émergence d’innovations médiatiques et technologiques, les seconds sont utiles pour prêter attention à la façon dont le problème des *fake news* a été publicisé et défini par différents acteurs, ainsi que pour analyser dans quelle mesure ce cadrage a pu façonner l'agenda des politiques publiques et des recherches académiques.

Les analyses de ce chapitre ont également été nourries par une enquête exploratoire menée en deux temps.

Tout d’abord, une phase d’exploration a été amorcée pendant nos études de master et poursuivie au début de notre thèse. En prenant part à divers événements, rencontres et projets portant sur le thème des *fake news*, nous avons pu prêter attention aux discours de différents acteurs, échanger avec eux et conduire des entretiens informels. Par exemple, nous avons (1) assisté à des événements organisés par des associations d’étudiants et de doctorants en psychologie cognitive et neurosciences impliquées dans la promotion de l'esprit critique telles que le *Critical Thinking Lab*, *Chiasma*, *Décrypte*, *Imhotep* ; (2) réalisé un stage de recherche et collaboré à des études expérimentales sur les *fake news* avec des chercheurs du laboratoire *La PsyDée* de l'Université Paris Cité et du *Social Decision-Making Lab* de l'Université de Cambridge ; (3) participé à des réunions du groupe de travail Esprit critique du Conseil Scientifique de l'Éducation Nationale piloté par Elena Pasquinelli et Gérald Bronner ; (4) discuté au cours de colloques, conférences et tables rondes de l'impact des *fake news* ou des mesures prises pour lutter contre avec des journalistes, *fact-checkers*, enseignants du secondaire, chercheurs de disciplines variées et responsables institutionnels. Cette démarche de participation observante nous a permis de partir à la rencontre de différents acteurs, de nous confronter à des points de vue divergents et ainsi de nous imprégner des discours et débats entourant les *fake news*.

Après cette phase d'imprégnation, nous avons souhaité cartographier les acteurs mobilisés dans les débats sur les *fake news* et analyser leurs discours d'une façon plus systématique. Pour cela, nous avons, d'une part, réalisé une étude de cas sur la couverture médiatique du documentaire *Hold-up* et, d'autre part, piloté la constitution d'une « Galaxie de la Raison » dans le cadre d'un projet de recherche réalisé pour l'Anses (Luneau et al., 2023). Nous proposons de renommer cette cartographie « Galaxie du Vrai/Faux » car elle ne cartographie pas exclusivement des acteurs appartenant à des mouvements rationalistes et défendant la rationalité mais plutôt une variété de communautés ayant pour particularité de s'exprimer de façon réflexive – et parfois normative – sur la démarche scientifique et la production de connaissances. Ces communautés mobilisent différents « régimes épistémiques » reposant chacun sur des conceptions divergentes du valide et des méthodes distinctes pour délimiter les frontières entre le vrai et le faux (Carbou et Gilles, 2019). Ce travail de cartographie a permis d'identifier les multiples catégories d'acteurs impliqués dans les débats sur les *fake news*, d'examiner leurs façons de poser le problème et d'analyser les solutions qu'ils préconisent pour y remédier.

Après avoir questionné la nouveauté désignée par le terme *fake news* par rapport à des notions plus anciennes comme celles de rumeurs, de propagande, de désinformation ou de théories du complot, ce chapitre montre comment les discours publics actuels réactivent des grilles de lecture techno-déterministes et psychologisantes, similaires à celles adoptées lors de précédentes « paniques médiatiques et technologiques ». Nous documentons ensuite comment cette façon de cadrer les débats et de définir la problématique des *fake news* a conduit à une focalisation de l'action publique sur des mesures de régulation ou d'éducation à l'esprit critique, ainsi qu'à orienter les recherches académiques vers des approches de *big data* ou de psychologie cognitive, sans que ne soient vraiment pris en compte les enseignements issus des travaux de sociologie des médias et de la communication.

## 1.1. L'irruption des *fake news* dans les discours publics

Jusqu'alors peu usité dans les discours publics[27], le terme *fake news* a connu une explosion massive de son utilisation au lendemain de l'élection présidentielle américaine de novembre 2016 ; ce qui lui a valu d'être élu mot de l'année 2017 par le dictionnaire Collins.[28] L'outil *Gallicagram*, développé par Benjamin Azoulay et Benoît De Courson (2021), permet en effet d'observer une forte augmentation de sa fréquence dans les quotidiens *Le Monde* et *The New York Times* à partir de l'année 2016 (cf. figures 1.1 et 1.2). De façon intéressante, le même pic se retrouve du côté des recherches effectuées sur Google par les internautes du monde entier.[29] Cette irruption soudaine du mot *fake news* dans les discours publics invite à s'interroger sur le caractère inédit de ce qui est désigné. Le terme est-il le signe d'un nouveau phénomène social ou renferme-t-il d'autres significations ?

Pour répondre à cette question, cette section commence par mettre en perspective le terme *fake news* avec des notions plus anciennes auxquelles il est fréquemment assimilé ou comparé dans les discours publics contemporains. Cette confrontation fait ressortir que le terme *fake news* ne désigne pas à proprement parler un nouveau phénomène social, mais atteste plutôt d'une résurgence des préoccupations des élites et des médias, quant à la question de la qualité de l'information, dues à l'apparition de nouveaux circuits de l'information et de la communication. Partant de ce constat, la deuxième section interroge la valeur heuristique du concept de « panique morale » et de ses sous-catégories « panique

[27] Le terme *fake news* a commencé à être utilisé de façon occasionnelle dans les années 1990, notamment dans l'émission satirique américaine *The Daily Show*. Il désignait alors des contenus parodiques ou des canulars (Harsin, 2018).
[28] Flood, A. (2017, 2 novembre). Fake news is 'very real' word of the year for 2017. *The Guardian*. https://www.theguardian.com/books/2017/nov/02/fake-news-is-very-real-word-of-the-year-for-2017
[29] Google Trends. « fake news ». https://trends.google.fr/trends/explore?date=all&q=fake news

médiatique » et « panique technologique » pour analyser les discours publics sur les *fake news*. Enfin, dans un troisième temps, une étude de cas conduite sur la couverture médiatique du documentaire *Hold-up* permet d'illustrer concrètement comment les débats contemporains sur les *fake news* s'inscrivent dans un nouveau cycle de « paniques médiatiques et technologiques ».

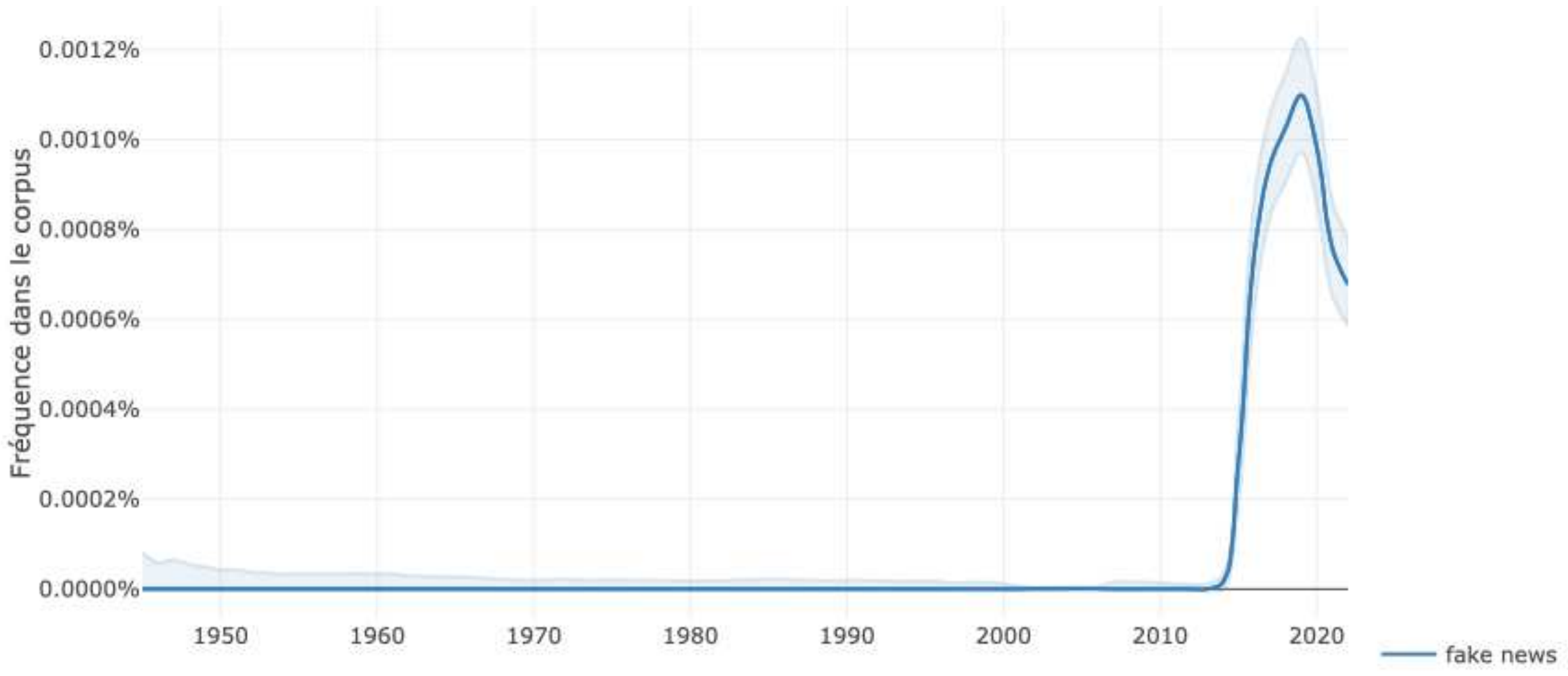

*Figure 1.1. Évolution de l'utilisation du terme fake news dans le journal Le Monde entre 1945 et 2022*

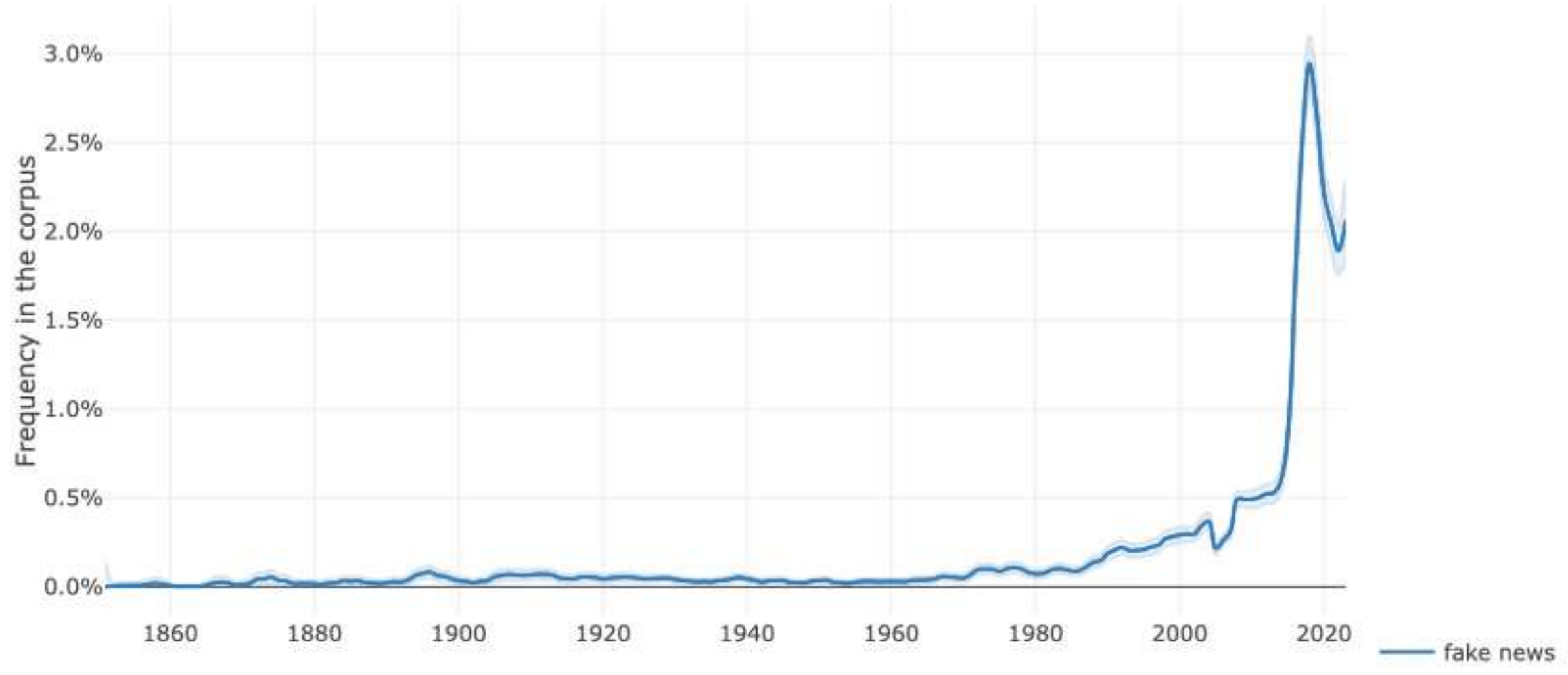

*Figure 1.2. Évolution de l'utilisation du terme fake news dans le New York Times entre 1851 et 2022*

### 1.1.1. *Fake news* : un phénomène pas si nouveau

Aujourd'hui, de nombreux exemples historiques, autrefois désignés par des notions telles que celles de « rumeur », de « propagande », de « désinformation » ou de « complot », sont rebaptisés sous le nom de *fake news*. Par exemple, le numéro d'octobre 2022 de la revue *Historia*[30], intitulé « Les *fake news* qui ont changé l'histoire », revient sur divers troubles informationnels associés à des événements majeurs comme la Révolution française, l'Affaire Dreyfus ou la Guerre du Golfe. Ces troubles observés dans le passé présentent-ils des similitudes avec ceux rencontrés dans le contexte contemporain ? La notion de *fake news* est-elle appropriée pour les décrire correctement ? Inversement, les concepts utilisés dans le passé sont-ils applicables aux enjeux contemporains ?

Répondre à ces questions nécessite de prendre en compte les travaux de sciences sociales ayant déjà développé différents cadres conceptuels et théoriques pour étudier divers troubles informationnels (Dauphin, 2019). En prenant appui sur un article de Julien Giry (2021), nous avons choisi de nous concentrer sur les travaux relatifs à quatre concepts : ceux de « rumeur », de « propagande », de « désinformation » et de « théorie du complot ». L'objectif de cette section n'est pas de dresser un panorama exhaustif de toutes les recherches menées sur chacun de ces concepts, mais plutôt d'examiner l'évolution de leur sens et de leurs conceptualisations, ainsi que de mettre en lumière ce qui les rapproche et ce qui les éloigne du terme *fake news*. Ainsi, après avoir offert une synthèse des recherches relatives à chacun de ces quatre concepts et dégagé les principaux éléments permettant de les caractériser — notamment au niveau de (1) leur statut épistémique, (2) leur sujet, (3) leur intention, (4) leur source et (5) leur mode de diffusion — nous les confrontons à la notion de *fake news* afin d'identifier leurs convergences et divergences.

Le premier concept avec lequel la notion de *fake news* est souvent associée est celui de « rumeur ». Du latin *rumor*, le terme apparaît en français dès le Moyen Âge ; il désigne à l'époque des bruits vagues et confus ou des bruits qui courent (Fargette, 2007). Toutefois, comme le signale Pascal Froissart (2011) dans un entretien pour la revue *Sens-Dessous* : « le concept de rumeur a une histoire qui ne recoupe pas l'histoire du mot ». En effet, bien que la rumeur soit souvent considérée comme le plus vieux média du monde (Kapferer, 1987), ce n'est qu'à partir du XXe siècle, parallèlement au développement des médias de masse et aux deux guerres mondiales, que les premières conceptualisations et théorisations scientifiques du terme ont émergé (Froissart, 2002). En montrant les processus de transformation des messages (réduction, accentuation, assimilation), qui arrivent au cours d'expériences réalisées en laboratoire, les travaux des psychologues Stern (1902), puis Allport et Postman (1945), ont contribué à appréhender la circulation de rumeurs comme un symptôme d'un dérèglement de la communication. Ces études de psychologie sociale ont cependant fait l'objet de nombreuses critiques (Peterson et Gist, 1951 ; Boon et Davis, 1987 ; Treadway et McCloskey, 1987 ; 1989 ; Rosnow, 1991), et des approches moins normatives, notamment de sociologie compréhensive (Lagrange, 1996 ; Aldrin, 2005) ou d'anthropologie interactionniste et pragmatique (Bonhomme, 2009) se sont développées ultérieurement pour interroger la fonction sociale des rumeurs sans les considérer comme un dysfonctionnement communicationnel. Aujourd'hui, bien que le concept de rumeur reste controversé, cinq caractéristiques principales peuvent être soulignées : il s'agit (1) d'énoncés non vérifiés mais pas forcément faux[31], (2) relatifs à un sujet circonstancié, (3) visant à compenser un défaut d'information, (4) dont la source est anonyme (e.g. « un ami d'un ami » ; voir Campion-Vincent et Renard, 1992) ; (5) et qui se diffusent par bouche à oreille.

[30] Revue Historia (2022, octobre). Les fake news qui ont changé l'histoire n°910. https://www.historia.fr/parution/mensuel-910

Le deuxième concept avec lequel la notion de *fake news* est parfois rapproché est celui de « propagande ». Apparu en 1622, au moment de la fondation de la *Congregatio de Propaganda Fide* (Congrégation pour la propagation de la foi), le terme désigne à l'époque les efforts déployés par l'Église catholique pour évangéliser les populations. À partir de la Révolution française, le mot prend une signification plus large et caractérise toute forme de communication organisée pour répandre une opinion, une idéologie ou une doctrine. Ce n'est qu'à la suite des deux guerres mondiales – période au cours desquelles des bureaux de propagande et des ministères sont créés pour monopoliser la production des représentations

[31] De nombreuses rumeurs s'avèrent vraies finalement (pour des exemples, voir : Pound et Zeckhauser, 1990 ; DiFonzo et Bordia, 2006).

collectives – que le terme acquiert une connotation péjorative (d'Almeida, 2002). Si la propagande est souvent associée à des régimes autoritaires ou totalitaires, des auteurs considèrent qu'elle est également utilisée dans les régimes démocratiques par des entreprises ou des organisations non gouvernementales, et peut prendre diverses formes, telles que la diffusion de discours politiques, la publication de documents officiels, la création de campagnes publicitaires ou la manipulation des médias de masse (Taïeb, 2010). Pour certains chercheurs, néanmoins, il importe de différencier une campagne de propagande, qui s'affirme comme détenant le monopole de la vérité, et une publicité, qui présente positivement une offre parmi d'autres (Chomsky, 2004, p. 39). Là encore, bien que la définition de la notion de propagande fasse l'objet de débats, on peut dégager quelques critères pour la définir : il s'agit (1) d'informations partiellement vraies, déformées ou tronquées, (2) imposant une vision du monde (3) pour influencer les opinions, les émotions, les attitudes ou les comportements d'un public ou d'une population (4) produites par des acteurs politiques, des gouvernements, ou des partis politiques et (5) diffusées de façon organisée et contrôlée (Ellul, 1967 ; Jowett et O'Donnell, 2015 ; Augé, 2007 ; Taïeb, 2010).

Le troisième concept avec lequel la notion de *fake news* est fréquemment identifiée est celui de « désinformation ». Introduit dans les années 1920 en Union Soviétique, le terme *dezinformatsia* désignait initialement les campagnes de propagande attribuées aux pays capitalistes contre l'URSS. L'usage du mot a culminé pendant la Guerre froide, restant principalement associé au monde soviétique pour dénoncer les actions, réelles ou supposées, des puissances occidentales contre le bloc communiste (Godson et Schultz, 1978). Si la notion de désinformation était utilisée historiquement pour désigner les fausses informations produites par les élites au pouvoir et diffusées via les mêmes canaux que les informations officielles, aujourd'hui, le terme est davantage employé pour décrire les fausses informations qui émanent de groupes situés en dehors des sphères de pouvoir et exploitent surtout les canaux alternatifs qu'offrent Internet et les réseaux sociaux pour les diffuser (Durandin, 1993 ; Giry, 2021). Autrement dit, à la différence de la propagande, la désinformation constitue un contre-discours visant non pas à servir le pouvoir en place mais à le dénoncer. Le rapport de Claire Wardle et Hossein Derakhshan (2017) a contribué à rapprocher la notion de *fake news* de celle de désinformation en précisant toutefois que cette dernière désigne des informations erronées produites dans l'intention de manipuler délibérément l'opinion publique, alors que les *fake news* peuvent être diffusées par inadvertance sans intention de tromper ou de causer du tort. De nouveau, bien que le terme de désinformation ne bénéficie pas d'une définition vraiment stabilisée, plusieurs éléments permettent de le caractériser : il s'agit de (1) fausses informations ; (2) relatives à des sujets d'intérêt public mais circonstanciées à un événement ou un groupe de personnes ; (3) diffusées avec l'intention de tromper l'opinion publique à des fins politiques, militaires ou économiques (4) par des groupes situés en dehors des sphères du pouvoir ; (5) via des circuits de communication alternatifs.

Le quatrième concept avec lequel la notion de *fake news* est souvent comparée est celui de « théorie du complot ». Si les travaux d'historiens font remonter les premières traces d'accusations de complot au Moyen Âge, et plus particulièrement à partir de la Révolution française[32], ce n'est qu'après la Seconde Guerre mondiale que le terme devient un concept scientifique, notamment à la suite de la publication des ouvrages *La Société ouverte et ses ennemis* en 1945 par le philosophe Karl Popper[33] et *The Paranoid Style in American Politics* en 1964 par l'historien Richard Hofstadter (Dieguez et Delouvée, 2021). Michael Butter et Peter Knight (2018, p. 34) qualifient les recherches dérivées de cette approche de « paradigme pathologisant » dans la mesure où elles envisagent les théoriciens du complot comme des groupes de « radicaux » ou de « paranoïaques » opérant en marge de la société. L'expression « théorie du complot » devient populaire à partir des années 1980 (Reichstadt, 2015 ; France, 2019), en parallèle de l'essor d'Internet, et bondit après les attentats du 11 septembre 2001, où diverses théories sont avancées pour contester la version officielle des événements. En France, la question est devenue hautement médiatisée depuis les attentats qui se sont déroulés à Paris en janvier 2015 (Kreis, 2015). Le traitement médiatique des « théories du complot » met en lumière un ensemble de thématiques associées, telles que la jeunesse, les quartiers populaires, l'islam, les problèmes éducatifs, l'influence d'Internet et des réseaux sociaux, ainsi que l'extrémisme politique et l'antisémitisme. Alors que la notion de théorie du complot est utilisée de façon stigmatisante depuis les années 1950 (Thalmann, 2019), des approches plus compréhensives se sont développées récemment (France et Motta, 2017 ; Giry, 2017 ; Harambam et Aupers, 2015). Ainsi, de nouveau bien que le terme de théorie du complot soit contesté (Butter et Knight, 2015, p. 30-31 ; Taguieff, 2013 ; 2017), quelques éléments de définition peuvent être proposés : il s'agit (1) d'énoncés non officiels et souvent non vérifiables (2) contenant une explication alternative d'événements historiques ou contemporains, (3) attribuant leur origine à une organisation clandestine regroupant des personnes puissantes (e.g. Illuminatis, Francs-Maçons, le FBI) agissant dans l'ombre pour atteindre des objectifs cachés et malveillants (Keeley, 1999) ; (4) produits par des individus ou groupes non officiels ; (5) et dont la diffusion est surtout associé à l'essor d'Internet.

[32] Un exemple notable est la publication des *Mémoires pour servir à l'histoire du jacobinisme* de l'abbé Augustin Barruel, où il accuse les philosophes et les francs-maçons d'avoir ourdi la Révolution française dans le dessein de détruire la chrétienté
[33] Dans ce livre, il introduit l'expression de « conspiracy theory of Society » qu'il définit comme « *l'opinion selon laquelle l'explication d'un phénomène social consiste en la découverte des hommes ou des groupes qui ont intérêt à ce qu'un phénomène se produise (parfois il s'agit d'un intérêt caché qui doit être révélé au préalable) et qui ont planifié et conspiré pour qu'il se produise* ».

Maintenant que nous avons retracé l'histoire de différents concepts fréquemment employés en sciences sociales, ainsi que dans le débat public, pour désigner différents troubles informationnels, confrontons chacune de ces notions à celle de *fake news* afin de comprendre les spécificités du phénomène désigné par ces dernières. D'un point de vue épistémique, tout d'abord, alors que les notions de rumeurs, de propagande et de théorie du complot ne désignent pas nécessairement des contenus faux, le terme *fake news* met l'accent, comme celui de désinformation, sur le caractère erroné d'un contenu. Au sens propre, il désigne d'ailleurs des énoncés falsifiés. En pratique, par ailleurs, il est utilisé dans les discours publics comme dans le langage courant pour qualifier des énoncés perçus comme faux. Au niveau de leur contenu, ensuite, les *fake news* ne sont pas, contrairement à la propagande et aux théories du complot, hégémoniques et globales mais circonstanciées et événementielles comme les rumeurs et la désinformation. Par ailleurs, alors que la propagande et la désinformation cherchent délibérément à induire en erreur leur audience, les rumeurs et théories du complot sont souvent diffusées pour compenser un défaut d'informations ou dévoiler une vérité cachée. Autrement dit, même si ces dernières s'avèrent erronées, leur intention n'est pas toujours malveillante. Dans le cas des *fake news*, si celles-ci sont souvent produites pour influencer l'opinion publique ou générer des revenus publicitaires, elles peuvent être partagées sur les réseaux sociaux par erreur ou pour une variété de raisons (e.g. provoquer, faire rire, etc.) qui restent à ce jour peu étudiées. Enfin, alors que la propagande est souvent associée à des acteurs institutionnels ou gouvernementaux, les *fake news*, comme les rumeurs, la désinformation et les théories du complot, peuvent être produites et diffusées par des individus ou des groupes non officiels. Si le terme *fake news* présente ainsi des spécificités par rapport à ceux de rumeur, de propagande, de désinformation et de théorie du complot (cf. Tableau 1.1), l'on peut rapprocher les cinq concepts par leur caractère normatif et péjoratif, ainsi que par leur contexte d'apparition. En effet, tous ces termes sont majoritairement utilisés (sauf par certaines approches compréhensives) pour désigner des dysfonctionnements communicationnels ou déviances informationnelles et sont devenus populaires lors de troubles sociopolitiques (guerre, attentat, etc.) attribués à des innovations médiatiques et technologiques (e.g. média de masse, Internet, réseaux sociaux).

| | **Rumeur** | **Propagande** | **Désinformation** | **Théorie du complot** | ***Fake news*** |
|---|---|---|---|---|---|
| **Statut épistémique** | Non vérifié | Biaisé, partial | Faux, trompeur | Non officiel | Falsifié, faux |
| **Intention** | Compenser un défaut d'information | Promouvoir une idéologie | Nuire, tromper | Expliquer un événement extraordinaire | Influencer l'opinion publique ou générer des revenus publicitaires |
| **Sujet** | Circonstancié | Global, totalisant | Circonstancié | Global, totalisant | Circonstancié |
| **Source** | Anonyme | État, institution, gouvernement | Média alternatif | Groupes marginaux, Lanceur d'alerte | Médias, sites web, blog |
| **Mode de diffusion** | Bouche-à-oreille | Média de masse | Média alternatif Internet | Internet | Réseaux sociaux, Amplification algorithmique |

*Tableau 1.1. Synthèse des éléments de définition des notions de rumeur, propagande, désinformation, théories du complot et fake news*

Au terme de cette analyse, il ressort ainsi que les *fake news* s'inscrivent dans une continuité historique de troubles informationnels et de manipulation de l'information. Autrement dit, bien que le terme soit récent, la réalité qu'il décrit n'est pas totalement inédite. En fait, ce qui justifie l'augmentation de l'utilisation du terme *fake news* en novembre 2016, alors que la problématique des troubles de l'information n'est finalement pas si nouvelle, ce sont les nouveaux modes de l'information et de la communication engendrés par le développement des plateformes numériques et des réseaux sociaux. Si les *fake news* ne désignent pas un phénomène totalement nouveau, leur émergence dans le discours contemporain reflète une évolution dans la manière dont l'information est produite, diffusée et consommée. En définitive, il apparaît dès lors que l'irruption soudaine du terme *fake news* dans le débat public est moins le signe de l'apparition d'un nouveau phénomène social que le signalement d'une brutale inquiétude des pouvoirs publics, des journalistes et de nombreux experts devant les risques que les nouveaux circuits de l'information font peser sur la qualité du débat public à l'ère du numérique.

### 1.1.2. L'éternel retour des paniques médiatiques et technologiques

La brève analyse historique effectuée dans la section précédente montre à quel point les inquiétudes relatives à divers types de troubles informationnels sont étroitement associées à l'arrivée d'un nouveau format médiatique ou d'une nouvelle technologie dans les pratiques du grand public. En sociologie des médias et du numérique, ces préoccupations sont fréquemment qualifiées sous le nom de « paniques médiatiques » (Drotner, 1999 ; Critcher, 2008)[34] ou de « techno-paniques » (Marwick, 2008). Ces deux notions découlent du concept de « panique morale » initialement utilisé en sociologie de la déviance et des problèmes sociaux par Stanley Cohen (1972) pour décrire l'anxiété publique ressentie quand :

> *une condition, un incident, une personne ou un groupe de personnes sont brusquement définis comme une menace pour la société, ses valeurs et ses intérêts ; ils sont décrits de façon stylisée et stéréotypée par les médias ; des rédacteurs en chef, des évêques, des politiciens et d'autres personnes bien pensantes montent au créneau pour défendre les valeurs morales ; des experts reconnus émettent un diagnostic et proposent des solutions ; les autorités développent de nouvelles mesures ou — plus fréquemment — se rabattent sur des mesures existantes ; ensuite la vague se résorbe et disparaît, ou au contraire prend de l'ampleur* (*Ibid*, p. 9).

Différents travaux ont ensuite approfondi et révisé le concept, notamment pour nuancer le rôle prédominant des médias et mettre davantage l'accent sur celui des groupes de pression ou des élites (Hall et al., 1979), ainsi que pour délimiter des indicateurs empiriques permettant de l'objectiver (Goode et Ben Yehuda, 1994)[35]. Dans les décennies suivantes, des analyses de nature socio-historique ont surtout contribué à mettre en lumière la récurrence des paniques suscitées par différents formats médiatiques et outils technologiques (Springhall, 1998) — qu'il s'agisse de romans-feuilletons (Dumasy, 1999), d'émissions télévisées (Buckingham, 1996), de bandes dessinées (Critcher, 2009), de films (Rapp, 2002) ou de jeux vidéo (Fergueson, 2007 ; 2008) — et ont documenté pour chacun d'eux comment « plusieurs catégories d'acteurs (magistrats, médecins, journalistes, experts) incriminent l'objet nouveau [...] en utilisant les mêmes arguments » (Wibrin, 2012, p. 90).

[34] Le terme média est souvent utilisé dans un sens élargi pour désigner aussi bien les moyens de communication de masse traditionnels, tels que la presse écrite, la radio et la télévision, que les plateformes numériques et les réseaux sociaux.

Dans son article *The Sisyphean Cycle of Technology Panics,* la chercheuse Amy Orben (2020) a développé une grille de lecture permettant de décomposer ces cycles de panique en quatre étapes. Elles sont d'abord initiées par (1) la mise à jour de troubles sociaux, psychologiques ou politiques directement imputés à la pratique d'une nouvelle technologie ; (2) elles suscitent ensuite la mobilisation des institutions et des acteurs politiques qui encouragent et attisent la panique afin de tirer des bénéfices politiques de la prise en charge du problème et de la promesse de le juguler ; (3) elles sollicitent alors les chercheurs qui réorientent leur agenda de recherche vers l'étude du nouveau problème public sans cependant jamais parvenir à fournir des conclusions faisant consensus ; (4) enfin, les résultats scientifiques, aussi incertains que contradictoires, sont oubliés parce qu'une nouvelle technologie a relancé un nouveau cycle de paniques et de recherches sur une autre question.

Les exemples de cas où des innovations médiatiques et technologiques ont suscité des inquiétudes sont si nombreux[36] qu'il est impossible de tous les recenser et d'illustrer pour chacun d'eux comment se sont manifestées les quatre étapes identifiées par Amy Orben. Contentons-nous ici de citer simplement l'exemple de la diffusion de l'émission *La Guerre des Mondes* d'Orson Wells à la radio sur le réseau CBS aux États-Unis. Cet événement est en effet souvent considéré comme un cas emblématique de « panique médiatique » (Lagrange, 2005 ; Pooley et Socolow, 2013a ; Schwartz, 2015) et est fréquemment utilisé aujourd'hui pour introduire les débat sur les *fake news*.[37] Comme dans le « cycle sisyphéen des paniques technologiques » décrit par Amy Orben (2020), de nombreux articles de presse ont tout d'abord multiplié les éditoriaux indignés et les reportages chocs attestant de la frayeur des Américains.[38] Par exemple, dans un éditorial pour le *New York Herald Tribune*, la journaliste Dorothy Thompson a écrit que :

> *Sans le vouloir, M. Orson Welles et le Mercury Theater on the Air ont réalisé l'une des démonstrations les plus fascinantes et importantes de tous les temps. Ils ont prouvé que quelques voix efficaces, accompagnées d'effets sonores, peuvent convaincre des masses de personnes d'une proposition totalement déraisonnable et complètement fantastique au point de créer une panique à l'échelle nationale* (notre traduction).[39]

---

[35] Dans leur livre *Moral Panics:The Social Construction of Deviance*, Erich Goode and Nachman Ben-Yehuda (1994) ont proposé cinq indicateurs : (1) préoccupation, (2) hostilité, (3) consensus, (4) disproportion et (5) volatilité.

[36] Un compte Twitter nommé « Pessimist Archive » documente ces inquiétudes en publiant fréquemment des extraits de journaux décriant les dangers de différentes innovations technologiques et médiatiques, comme les vélos, le télégraphe, la radio, les bandes-dessinées.

Dans un second temps, des campagnes de régulation ont été menées pour imposer des règles aux programmes radiophoniques. Le président de la Federal Communications Commission (FCC), Franck McNinch, a par exemple été sollicité par des hommes politiques, des éditorialistes et des ligues de vertus pour interdire la radiodiffusion de fausses informations. En parallèle, des recherches ont aussi été entreprises pour étudier les réactions du public à la radiodiffusion de l'émission. Par exemple, le psychologue Hadley Cantril a obtenu un financement de la Rockefeller Foundation pour une « étude sur l'hystérie de masse ».[40] La publication en 1940 de son ouvrage *The Invasion from Mars: A study in the Psychology of Panic* a ainsi apporté une légitimation scientifique aux préoccupations des médias et des régulateurs concernant la vulnérabilité des publics. Enfin, presque tous les journaux ont rapidement délaissé l'affaire ; ce qui suggère que l'hystérie des populations n'a pas été aussi généralisée que l'avaient laissé entendre les médias (Campbell, 2016).

---

[37] Voir : Cardon, D. (2018, 27 février) Fake news Panic. Circuits et chemin de l'information dans l'espace public numérique, *Communication pour le colloque : « La démocratie à l'âge de la post-vérité »,* Collège de France.)
Cardon, D. (2019, 20 juin) « Pourquoi avons-nous si peur des fake news ? », *AOC*
https://aoc.media/analyse/2019/06/20/pourquoi-avons-nous-si-peur-des-fake-news-1-2/
https://aoc.media/analyse/2019/06/21/pourquoi-avons-nous-si-peur-des-fake-news-2-2/
[38] New York Times. (1938, 31 octobre). Radio Listeners in Panic, Taking War Drama as Fact. *New York Times*.
https://www.nytimes.com/1938/10/31/archives/radio-listeners-in-panic-taking-war-drama-as-fact-many-flee-homes.html
[39] Thompson, D. (1938). Mr. Welles and mass delusion. *New York Herald Tribune*.
[40] Tracy, J. F. (2012, 29 avril). Early "Psychological Warfare" Research and the Rockefeller Foundation. *Global Research*.
https://www.globalresearch.ca/early-psychological-warfare-research-and-the-rockefeller-foundation/30594

Aujourd'hui, de nombreux chercheurs en sciences sociales proposent d'utiliser le concept de « panique morale », ou l'une de ses sous-catégories (i.e. « panique médiatique » et « technopanique »), pour désigner les inquiétudes formulées par les discours publics à propos des *fake news* (Anderson, 2021 ; Carlson, 2020 ; Mitchelstein et al., 2020). Néanmoins, rares sont ceux qui s'appuient sur un travail empirique rigoureux pour vérifier la pertinence du concept et pour décrire de façon fine et précise les différents types d'acteurs qui sont à l'origine de la mise à l'agenda du problème des *fake news* dans le débat public, les causes qu'ils lui attribuent et les catégories de la population qu'ils considèrent comme les plus vulnérables. En France, l'enquête menée par Ysé Vauchez (2022) sur un corpus de 1 403 émissions de télévision, diffusées entre le 13 décembre 2016 et le 10 mai 2019, offre des éléments de réponse concrets pour illustrer ces différents aspects. En effet, son étude montre comment les débats sur les *fake news* s'inscrivent pleinement dans la continuité des « paniques médiatiques et technologiques » en étant principalement associés à l'émergence d'Internet et des réseaux sociaux et en désignant principalement les jeunes et les classes populaires comme les groupes sociaux les plus vulnérables.

Pour prolonger ces travaux et acquérir une connaissance approfondie des acteurs contemporains, de leurs arguments et de leur manière de cadrer le problème des *fake news*, il est utile, dans le cadre de cette thèse, de conduire une analyse des discours médiatiques sur une période postérieure à celle étudiée par Ysé Vauchez. L'attention s'est portée spécifiquement sur la couverture médiatique du documentaire *Hold-Up*, mis en ligne le 11 novembre 2020. Ce film se présente comme une enquête révélant les « mensonges » et la « corruption » ayant marqué la gestion de la pandémie de Covid-19, tout en suggérant que le Forum économique mondial utiliserait la crise pour mettre en œuvre un « plan global » (ou *Great Reset*) visant à asservir l'humanité. Le choix de se concentrer sur ce documentaire s'explique par l'opportunité qu'il offre d'examiner la réaction des médias face à un cas particulièrement médiatisé de troubles de l'information, comparable à l'épisode de *La Guerre des Mondes*.

### 1.1.3. Du bruit à la fureur : le cas du documentaire Hold-Up

Le corpus étudié comporte un ensemble d'articles ayant couvert le documentaire *Hold-Up* entre le 11 août 2020 (jour du lancement de sa campagne de *Crowdfunding*) et le 11 mars 2021 (un mois après la sortie de sa version augmentée).

Pour constituer ce corpus, tous les tweets contenant les mots clés suivants ont été récoltés : « Holdup » ; « Holdup_ledoc » ; « Hold_Up » ; « HoldUpStopLaPeur » ; « Hold-Up ». Au total, environ 400 000 tweets, dont près de 90 000 contenant une URL, ont été obtenus. Après avoir exclu les URLs dirigeant vers des plateformes comme Facebook ou Instagram (et donc vers des posts de réseaux sociaux et non directement vers des articles médiatiques), environ 4 000 URLs uniques ont été conservées. Un nettoyage manuel de cette liste a ensuite été opéré pour ne garder que les URLs relatives au documentaire *Hold-Up* et retirer celles portant sur un autre sujet (e.g. un *hold-up* dans une banque). Enfin, les contenus textuels et les métadonnées des pages web correspondant aux 1 134 URLs de notre liste nettoyée ont été extraits à l'aide de l'outil *Minet* (Plique et al., 2019).

Une fois cette collecte réalisée, nous avons cherché à capturer la position adoptée par chacun des articles du corpus par rapport au documentaire. Une première lecture exploratoire d'une centaine de titres et de chapeaux introductifs a permis d'identifier quatre types de prises de position (cf. Tableau 1.2).

Toutes les URLs du corpus ont ensuite été classées selon ces quatre modalités. Puis, nous avons conduit une analyse séquentielle pour suivre l'évolution temporelle des différentes prises de position des médias sur le documentaire (cf. Figure 1.3).

| Catégorie | Description | Exemple |
|---|---|---|
| **Soutien** | l'article parle du documentaire sur un ton globalement positif et/ ou cherche à faire sa promotion | « Hold-up : le film-choc sur le scandale sanitaire imposé aux Français »[41] |
| **Debunk** | l'article dénonce la circulation virale du documentaire et remet en cause la crédibilité de certains passages | « "Hold-up" : on a décrypté le docu qui alimente les fantasmes sur le Covid-19 »[42] |
| **Critique des médias *mainstream*** | l'article ne porte pas directement sur le contenu du documentaire mais critique les réactions des médias *mainstream* suite à sa diffusion | « Panique à bord ? Le formidable documentaire "Hold-Up" fait scandale dans les médias officiels et chez les macronistes »[43] |
| **Nuance** | l'article s'apparente à une analyse réflexive qui cherche davantage à comprendre le succès du documentaire qu'à le dénoncer | « Critique constructive d'une journaliste réputée modérée sur le film Hold-Up »[44] |

*Tableau 1.2. Grille d'annotations utilisée pour identifier la position des articles sur le documentaire Hold-Up*

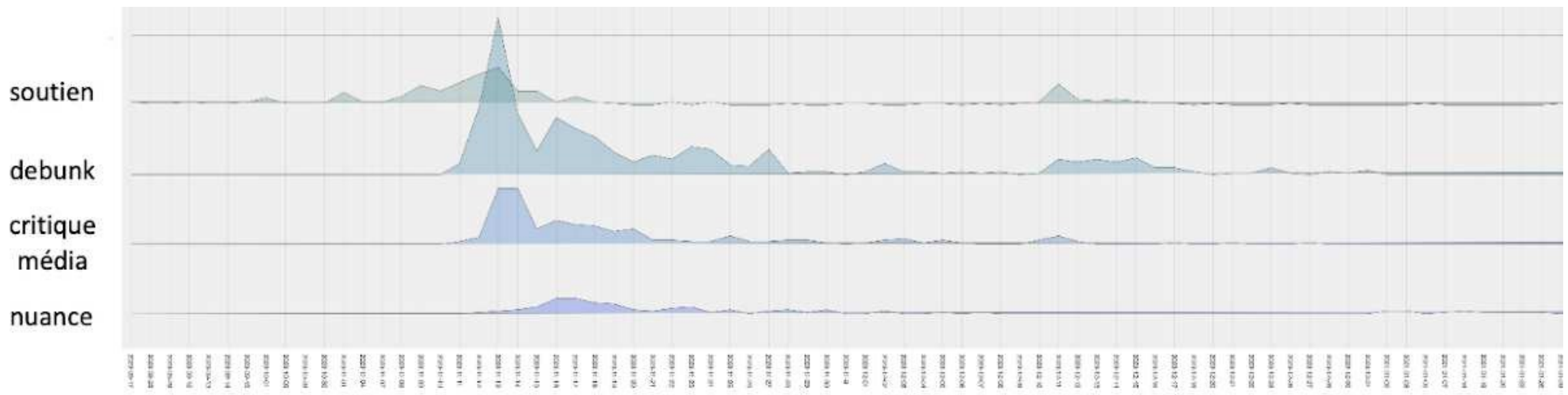

*Figure 1.3. Évolution temporelle des différentes prises de position des médias sur le documentaire Hold-up*

[41] Tournesol, G. (2020, 14 septembre). Hold-up : le film-choc sur le scandale sanitaire imposé aux Français. *Riposte Laïque*. ripostelaique.com/hold-up-le-film-choc-qui-denonce-le-scandale-sanitaire-impose-aux-francais.html

[42] Berrod, N., Gallet, L., Rougerie, P. (2020, 12 novembre). «Hold-up» : on a décrypté le docu qui alimente les fantasmes sur le Covid-19. *Le Parisien*. leparisien.fr/amp/societe/covid-19-labos-masques-et-domination-du-monde-on-decrypte-hold-up-le-docu-qui-agite-les-complotistes-12-11-2020-8408046.php

[43] Anonyme. (2020, 12 novembre). Panique à bord ? Le formidable documentaire « Hold-Up » fait scandale dans les médias officiels et chez les macronistes. *Sott.net.* http://fr.sott.net/article/36278-Panique-a-bord-Le-formidable-documentaire-Hold-Up-fait-scandale-dans-les-medias-officiels-et-chez-les-macronistes

[44] Polony, N. (2020, 18 novembre). Critique constructive d'une journaliste réputée modérée sur le film Hold-Up. *Agoravox*. agoravox.tv/tribune-libre/article/critique-constructive-d-une-87644

Le graphique ci-dessus rend compte du volume d'articles publiés entre le 11 août et le 11 mars 2020 pour les quatre types de positions adoptées par les médias par rapport au documentaire. Six phases peuvent être dégagées :

(1) Entre la fin de l'été et le début de l'automne 2020, seule une poignée d'articles de soutien (n=18) ont été mis en ligne par légers à-coups pour annoncer la production du documentaire par la société *T prod*, encourager à participer à sa campagne de *crowdfunding* sur la plateforme Ulule et inviter à suivre son actualité sur les réseaux sociaux, par exemple par les sites *Riposte Laïque*[45] et *Résistance Républicaine*[46].

(2) À partir du début du mois de novembre, on observe une augmentation importante du nombre d'articles de soutien publiés (n=44). Par exemple, quelques jours avant la sortie du documentaire, Christophe Cossé a rédigé une longue tribune pour le média *France Soir*[47] afin d'expliquer les raisons qui l'ont conduit à le produire.

(3) Le jour de la sortie du documentaire, le 11 novembre, puis dans les quelques jours qui ont suivi, le nombre d'articles faisant la promotion du documentaire a continué à augmenter (n=75), mais ce qui est particulièrement notable, c'est l'apparition rapide et volumineuse des premiers articles de debunk sur la même période (n=213). Ceux-ci ont d'abord été réalisés par des sites web comme *Conspiracy Watch*[48] ou *Fact and Furious*[49], puis par des médias traditionnels comme *France Inter*[50], *Le Parisien*[51] ou *Le Figaro*[52].

(4) Cette mobilisation des médias *mainstream*, couplée aux décisions de certaines plateformes comme *Vimeo* de retirer le documentaire, a été perçue par certains médias alternatifs comme une forme de censure ou de musellement, ce qui a pu les pousser à publier davantage d'articles visant à critiquer les médias dominants qu'à soutenir le documentaire *Hold-Up*.

---

45 Tournesol, G. (2020, 14 septembre). Hold-up : le film-choc sur le scandale sanitaire imposé aux Français. *Riposte Laïque*. https://ripostelaique.com/hold-up-le-film-choc-qui-denonce-le-scandale-sanitaire-impose-aux-francais.html

46 Tassin, C. (2020, 23 août). "Hold-up" ! "LE" film sur le Covid en préparation, avec les Résistants Perronne, Raoult et compagnie !. *Résistance Républicaine*. https://resistancerepublicaine.com/2020/08/23/hold-up-le-film-sur-le-covid-en-preparation-avec-les-resistants-perronne-raoult-et-compagnie/

47 Cossé, C. (2020, 7 novembre). Hold-Up, film en sortie nationale 11 novembre. Pourquoi j'ai produit ce film par Christophe Cossé. *France Soir*. https://www.francesoir.fr/opinions-tribunes-culture-cinema/hold-film-en-sortie-nationale-11-novembre-pourquoi-jai-produit-ce

48 Conspiracy Watch. (2020, 12 novembre). « Hold-up ». *Conspiracy Watch*. https://www.conspiracywatch.info/notice/hold-up

49 Daoust, A. (2020, 11 novembre). HOLD-UP, la comédie fiction de Pierre Barnérias. *Fact and Furious*. http://factandfurious.com/archives-instant-critique/hold-up-la-comedie-fiction-de-pierre-barnerias

50 France Inter. (2020, 12 novembre). "Hold-Up" : "Pour démonter ce documentaire, il faudrait des heures, des jours de travail". *France Inter*. franceinter.fr/societe/hold-up-pour-demonter-ce-documentaire-il-faudrait-des-heures-des-jours-de-travail

51 Berrod, N., Gallet, L., Rougerie, P. (2020, 12 novembre). «Hold-up» : on a décrypté le docu qui alimente les fantasmes sur le Covid-19. *Le Parisien*. https://www.leparisien.fr/societe/covid-19-labos-masques-et-domination-du-monde-on-decrypte-hold-up-le-docu-qui-agite-les-complotistes-12-11-2020-8408046.php

52 Renault, C. (2020, 12 novembre). Covid-19 : Hold-up, le documentaire sur un complot mondial qui fait polémique. *Le Figaro*. https://www.lefigaro.fr/actualite-france/covid-19-hold-up-le-documentaire-sur-un-complot-mondial-qui-fait-polemique-20201112

(5) Ensuite, dans les deux-trois semaines qui ont suivi la sortie de *Hold-Up,* quelques longs articles ont été publiés, parfois par des universitaires, sur des magazines en ligne, des blogs ou des sites web indépendants afin d'analyser de façon distante et réflexive aussi bien le succès du documentaire que les discours de vérification et de dénonciation qu'il a suscités de la part des médias *mainstream*. Par exemple, le chercheur Olivier Ertzscheid[53] a publié un article sur son blog *Affordance info*, ainsi que François Provenzano sur *Diacritik*[54].

(6) Enfin, malgré la sortie d'une version augmentée puis de deux autres opus, il n'y a quasiment plus eu d'articles publiés sur le documentaire à partir de mi-décembre 2020. Seule une petite dizaine d'articles de soutien et de debunk ont été publiés vers la mi-février au moment de la sortie de la version augmentée.

Si l'analyse séquentielle qui vient d'être conduite permet de distinguer six phases différentes dans la manière dont le documentaire a été couvert par les médias, l'une d'entre elles se démarque particulièrement par son volume de publications : celle de debunk. En effet, nous avons pu observer l'important nombre d'articles publiés par des médias traditionnels dès le lendemain de la sortie du documentaire pour alerter sur son caractère complotiste. Nous allons donc à présent concentrer la suite de notre analyse sur les 500 articles qui se sont attachés à debunker le documentaire. Tous ces articles ont été lus et annotés afin d'identifier (1) la façon dont le documentaire *Hold-Up* a été présenté dans les médias ; (2) les différents acteurs qui se sont mobilisés ou ont été sollicités pour le commenter ; (3) les causes et (4) les conséquences qui ont été attribuées à son succès.

Voyons d'abord comment le documentaire a été présenté dans les médias et dans quelle mesure il a été associé au problème des *fake news*. Initialement décrit comme un documentaire « complotiste », « [inspiré] – à la limite du plagiat – du film conspirationniste américain

[53] Ertzscheid, O. (2020, 15 novembre). De la "compol" au complot : le hold-up des eschatologies narratives. *Affordance Info.* https://affordance.framasoft.org/2020/11/compol-et-complot-hold-up/

[54] Provenzano, F. (2020, 7 décembre). Je suis le spectateur de Hold-up. *Diacritik*. https://diacritik.com/2020/12/07/je-suis-le-spectateur-de-hold-up/

*Plandemic*, le documentaire *Hold-Up* est très vite devenu un objet de débats cristallisant des inquiétudes plus larges sur les « fake news », « l'infodémie » ou la « post-vérité ». Par exemple, le programme de *fact-checking* de *20 Minute*s, *Fake off*, l'a intégré « aux intox de la semaine », et de nombreux médias l'ont présenté comme un « monument de désinformation », « truffé de fausses informations », « d'affirmations mensongères », « trompeuses » ou « fallacieuses » ou de « contre-vérités » sur le Covid-19, « allant parfois totalement – et grossièrement — à l'encontre des données scientifiques ». Au-delà de pointer les erreurs factuelles diffusées dans le documentaire, les articles de debunk ont aussi, dans la lignée de la majorité des articles de *fact-checking* évoqués dans l'introduction, mis en avant sa circulation virale en parlant d'une vidéo « largement partagé[e] sur les réseaux sociaux », « visionné[e] plusieurs milliers de fois » qui a « agité la Toile » et « a eu un retentissement inégalé en France ».

Penchons-nous, à présent, sur les différents types d'acteurs qui se sont mobilisés pour commenter le documentaire. En première ligne, se trouvent les médias traditionnels et journalistes spécialisés dans la vérification factuelle qui, comme le confirme le nombre important de debunks publiés, « sont immédiatement montés au front » et « ont levé une multitude de boucliers incluant mises en garde et fact checking » pour « apporter un éclairage objectif et constructif ». Par exemple, *Oh My fake,* le programme de *20 Minutes* qui se décrit comme un « antidote contre les rumeurs » qui « rend [les publics] fort contre les fake news » a passé en revue une dizaine d'affirmations contenues dans Hold-Up, et *France TV* a produit, de son côté, une émission « Vrai ou fake » spéciale *Hold-Up*, présentée par Adrien Rohard et Julien Pain.

La lecture des articles issus des médias traditionnels montre qu'ils ont donné la parole à différents experts ou universitaires, connus pour être des spécialistes des théories du complot, tels que Thomas Huchon, journaliste et réalisateur de documentaires pour *Spicee* ; Tristan Mendès France, maître de conférences, associé à l'université de Paris Diderot spécialisé en cultures numériques et collaborateur à l'Observatoire du conspirationnisme ; et Rudy Reichstadt de *Conspiracy Watch.* On retrouve aussi : Marie Peltier, historienne, chercheuse et enseignante à l'institut supérieur de pédagogie Galilée de Bruxelles (ISPG) ; Laurence Kaufmann, professeure à l'université de Lausanne ; Cédric Terzi, enseignant à l'Ecole des hautes études en sciences sociales (EHESS) et Divina Frau-Meigs, sociologue des médias, professeure à l'université Sorbonne Nouvelle. La parole a également été donnée à des médecins comme le docteur Jérôme

Marty médecin généraliste, président du syndicat de l'Union française pour une médecine libre (UFML).

Différentes institutions et personnalités politiques se sont également mobilisés pour dénoncer le documentaire. Par exemple, « de nombreux députés ou marcheurs ont invité à ne pas regarder ce documentaire, comme Mounir Mahjoubi, Laetitia Avia ou Coralie Dubost ».[55] Les Académies nationales de médecine, pharmacie, sciences, technologies ont aussi alerté l'opinion sur les contre-vérités véhiculées par le film en publiant un communiqué dénonçant un « bric-à-brac d'inepties ». Enfin, l'institut Pasteur a de son côté décidé de se pourvoir en justice et a porté plainte en diffamation contre le réalisateur du documentaire.

Par ailleurs, on peut aussi remarquer l'intervention d'entrepreneurs ou d'entreprises dans les débats. En effet, alors que certains ont dénoncé le rôle des plateformes dans la diffusion virale du documentaire, celles-ci ont été obligées de se justifier. Selon le CEO d'Ulule, Alexandre Boucherot[56], si le projet *Hold-Up* a pu passer entre les mailles des filets des modérateurs, c'est parce que son pitch initial était « euphémisé ». Au début, précise-t-il, il s'est « positionné sur le mode "d'autres voix sont possibles", mais très vite on s'est rendu compte qu'il débordait du cadre initial supposé (le pluralisme des voix) pour devenir un étendard de thèses complotistes très éloignées de ce que l'on défend sur Ulule ». Dans un thread posté le 12 novembre sur Twitter[57], Alexandre Boucherot explique alors avoir fait le choix de ne pas couper le projet sur la plateforme, mais que l'intégralité de la commission Ulule perçue sur le projet allait être reversée à une association de défense des droits ou du journalisme comme le stipule la politique de modération de la plateforme si un projet contrevient à sa mission de diversité et d'ouverture.

Enfin, plusieurs associations, collectifs et vidéastes spécialisés dans la vulgarisation scientifique et la promotion de l'esprit critique, tels que le collectif Covid 19 Fédération, les

---

[55] La Rédaction. (2020, 13 novembre). Covid-19. Hold-Up : le documentaire qui fait polémique. *Actu-Mag*. https://www.actu-mag.fr/2020/11/13/covid-19-hold-up-le-documentaire-qui-fait-polemique/
[56] https://twitter.com/cemonsieur/status/1326873749531074560
[57] https://twitter.com/cemonsieur/status/1326873749531074560?ref_src=twsrc%5Etfw

Éclaireurs du Numérique, Esprit Zététique, le Youtubeur Vincent Verzat de la chaîne « Partagez c'est sympa », sont également intervenus pour faire un long travail d'analyse afin de réfuter les thèses défendues par le documentaire. Par exemple, le collectif Covid 19 Fédération a indiqué avoir passé une demi-journée de travail rien que pour analyser les quinze premières minutes du documentaire.[58]

Pour expliquer le succès du documentaire, les articles de debunk des médias et les différents types d'acteurs auxquels ils ont donné la parole mettent en avant une cause principale : les nouveaux modes de circulation de l'information permis par Internet, le web et les réseaux sociaux. Dans un article intitulé « L'arbre du complotisme et la forêt de la désinformation »[59], l'Afis explique que « la diffusion virale du documentaire a exploité ce que le sociologue Gérald Bronner décrit sous l'expression de « marché dérégulé de l'information » où « la désinformation, qu'elle soit intentionnelle ou inconsciente, a trouvé avec Internet un puissant moyen d'amplification ». De son côté, Tristan Mendès France relève que les réseaux sociaux « permettent à des productions marginales [...] de dépasser le cercle naturel de leur audience, qui devrait se limiter à quelques dizaines de milliers d'énervés [...], mais qui, en réalité, obtiennent une visibilité allant bien au-delà de la réalité de ce que représente l'audience naturelle de ce documentaire ».[60] Cette forte visibilité est souvent attribuée aux algorithmes des plateformes :

> *Ce succès révèle en effet que les algorithmes des réseaux sociaux suivent toujours la logique de marché et continuent d'afficher ce qui est susceptible de plaire à l'utilisateur (et donc d'être cliqué). Dès lors, en créant des « bulles de filtre », Internet et les réseaux sociaux continuent de jouer un rôle décisif dans le succès des théories du complot.*[61]

---

[58] Pour accéder à l'analyse : https://express.adobe.com/page/xiBTzlo8ML8Hv/

[59] Afis (2020, 2 Décembre). L'arbre du complotisme et la forêt de la désinformation. *Association Française pour l'Information Scientifique*. https://www.afis.org/L-arbre-du-complotisme-et-la-foret-de-la-desinformation

[60] Beliaeff, L., Cadeau, L., & Quioc, M. (2020, 14 novembre). Covid-19 : "Hold-Up", ou comment un film aux théories complotistes est devenu viral. *France Info.* https://france3-regions.francetvinfo.fr/provence-alpes-cote-d-azur/covid-19-hold-up-comment-film-accuse-complotisme-est-devenu-viral-1894768.html

[61] Alemany Oliver, M. (2020, novembre 18). «Hold-up», ou comment faire une bonne théorie du complot. *Slate*. https://www.slate.fr/story/197214/hold-up-film-documentaire-covid-19-theorie-complot-big-pharma-gafam

Les gros volumes d'engagement déclenchés par le documentaire sont aussi fréquemment attribués au manque d'esprit critique et de rationalité des utilisateurs des réseaux sociaux. Par exemple, pour le chercheur Antoine Bristielle ou le youtubeur Vincent Verzat de la chaîne « Partager C'est sympa », « il y a une perméabilité d'un nombre assez important de la population à ces théories » [62]; « Hold-Up rend les spectateurs vulnérables, confus et terrifiés ».[63] Certains acteurs évoquent également des causes socio-politiques mais ces perspectives sont assez rares ou très étroitement associées aux biais cognitifs des individus comme l'illustrent ces propos de Rudy Reichstadt :

> *Si les gens sont prêts à donner de l'argent pour entendre ce genre de propos, c'est qu'il y a une demande sociale. Je pense qu'il faut considérer qu'il y a toute une partie du public qui est demandeuse, qui est cliente de ce genre de contenu. Il y a des raisons psychologiques, des biais cognitifs : on sait par exemple que l'anxiété est un facteur d'adhésion qui facilite l'adhésion à ces théories. Mais il y a aussi, évidemment, la variable des sympathies politiques, idéologiques qui joue à plein.*[64]

En mettant en avant le pouvoir des réseaux sociaux et en présentant les publics comme vulnérables, la plupart des articles de debunk attribuent des effets forts au documentaire. Ils le décrivent comme « une véritable entreprise paranoïaque qui a pour seul but de prendre le contrôle de la personne qui la regarde, sans lui laisser la moindre possibilité de la réflexion et de la critique »[65], ou encore comme une « technique parfaitement mise au point par les sectes ou des groupements intégristes qui, ainsi, harponnent les personnes fragilisées, crédules ou peu cultivées »[66]. Pour différents acteurs, ces effets forts se traduisent par une influence directe sur les croyances et comportements des individus. Par exemple, Sylvain Delouvé relève que « l'adhésion à ces croyances a un impact direct sur les comportements, qui peut pousser à ne pas porter le masque, à refuser un vaccin ».[67]

---

[62] *Ibid.*

[63] Lemoine, T. (2020, 16 novembre). Coronavirus : Face à « Hold-Up », certains ont choisi l'humour. *20 Minutes.* https://www.20minutes.fr/societe/2909771-20201116-coronavirus-face-hold-ups-certains-choisi-rire

[64] Lopez, L.-V., & Demagny, X. (2020, 17 novembre). Pourquoi la France est "extrêmement anti-vaccin" : interview croisée de deux experts. *Radio France.* https://www.radiofrance.fr/franceinter/pourquoi-la-france-est-extremement-anti-vaccin-interview-croisee-de-deux-experts-6175288

[65] Guelff, P. (2020, 19 novembre). « Hold-up » est au complotisme, ce que fut « Forces Occultes » au nazisme. *Fréquence Terre.* https://www.frequenceterre.com/2020/11/19/hold-up-est-au-complotisme-ce-que-fut-forces-occultes-au-nazisme/

[66] *Ibid.*

[67] 24 heures. (2020, 13 novembre). «Hold-Up»: un documentaire au service des thèses complotistes. *24heures.* https://www.24heures.ch/hold-up-documentaire-au-service-des-theses-complotistes-445791295675

En établissant ainsi, à partir d'indicateurs d'audience, un lien de cause à effet entre la circulation virale du documentaire et des croyances ou comportements, les discours publics contribuent à réactiver des paradigmes d'effets forts. Aucun article ne questionne toutefois de manière critique les indicateurs d'audience et métriques d'engagement cités pour illustrer le succès d'Hold-up : par exemple, une vue signifie-t-elle vraiment que chaque utilisateur a visionné l'intégralité des 2h45 du documentaire ? En effet, il existe peu de transparence quant à la définition de ce qui constitue un « visionnage » sur YouTube. Plusieurs sources indiquent que « le visionnage d'une annonce payante est comptabilisé comme une vue lorsque le spectateur regarde au moins 30 secondes d'une vidéo de plus de 30 secondes ». [68] Il est donc probable qu'une part importante de ces vues ne reflète pas une exposition complète au contenu du documentaire. Et si comme Gisèle, le personnage principal de l'historiette racontée dans l'introduction de cette thèse, plusieurs utilisateurs avaient liké la vidéo par erreur en voulant couper le son ou la mettre sur pause ? Cette absence de prise en compte des nuances relatives aux modes de consommation des contenus en ligne interroge la validité des conclusions qui prêtent des croyances ou comportements aux individus juste en reposant sur des données d'engagement, et met en évidence la nécessité d'adopter une approche plus critique et nuancée des traces numériques laissées par les utilisateurs sur les réseaux sociaux.

En mettant en garde contre les risques soulevés par l'émergence de nouveaux modes d'information et de communication (en l'occurrence les réseaux sociaux), et en dénonçant la crédulité des publics, les discours contemporains sur le documentaire *Hold-up*, et plus largement sur les *fake news* présentent en cela de nombreuses similarités avec d'anciens cas de « paniques médiatiques et technologiques ». En effet, on retrouve le même cadrage techno-déterministe et psychologisant mis en avant par le même type d'acteurs (e.g. journalistes, décideurs publics, experts). Mais ce qui permet surtout d'appréhender les débats contemporains sous l'angle d'une panique morale, c'est leur décalage avec les constats empiriques de la littérature académique. Mobiliser le concept de « panique morale » pour qualifier les débats contemporains sur les *fake news* ne revient pas pour autant à nier l'existence d'un problème (Hunt, 1997). L'intérêt est plutôt de questionner les répercussions qu'a cette unique façon de cadrer le problème sur l'action publique et les recherches académiques.

[68] Pour des éléments d'explications sur les méthodes de calcul des métriques d'engagement sur Youtube : https://support.google.com/youtube/answer/2991785?hl=fr#:~:text=Consulter%20le%20nombre%20de%20vues,du%20nombre%20potentiel%20de%20vues.

## 1.2. Des entrepreneurs de cause et leur problème

L'étude effectuée dans la section précédente sur la couverture médiatique du documentaire *Hold-Up* a permis d'amorcer une cartographie des acteurs qui ont participé ces dernières années, par leurs interventions dans le débat public, à définir les *fake news* comme un problème public. Ce travail de cartographie et d'analyse de discours a pu être complété grâce à la « Galaxie du Vrai/Faux » mentionnée ci-dessus. Cette section vise à décrire plus précisément qui sont les différents types d'acteurs mobilisés en France dans les débats sur les *fake news*, ainsi qu'à présenter leurs actions et à questionner leur efficacité ou leurs limites. Nous proposons de nous concentrer sur quatre types d'acteurs et d'actions : (1) le *fact-checking*, (2) la régulation, (3) l'éducation aux médias et à l'esprit critique et (4) la vulgarisation scientifique.

### 1.2.1. L'essor du journalisme de vérification

Parallèlement à l'augmentation du mot *fake news* dans les discours publics, un autre terme a également gagné en popularité à partir de 2016 : celui de *fact-checking*. Tout comme celle de *fake news*, cette notion n'est pas complètement nouvelle, mais a pris un autre sens aujourd'hui. Par exemple, en 1923, le rôle des *fact-checker*s du magazine *Time* était de vérifier scrupuleusement l'exactitude de toutes les informations avant leur publication. *A contrario*, les *fact-checkers* contemporains exercent un contrôle *a posteriori* sur la qualité de l'information, c'est-à-dire qu'ils vérifient la véracité des chiffres et des propos énoncés publiquement après leur publication (Bigot, 2017).

Si des initiatives de vérification *a posteriori* comme la clinique des rumeurs existaient déjà au début de la Seconde Guerre mondiale (Froissart, 2022), les rubriques de *fact-checkin*g ont largement augmenté depuis une dizaine d'années (Graves, 2016 ; Graves et Cherubini, 2016).

En Europe, par exemple, plus de 90 % des sites de *fact-checking* ont été établis depuis 2010, avec une intensification accrue ces dernières années. En France, *Libération*, via sa rubrique *Désintox* (2008), fait figure de pionnier. D'autres médias ont suivi avec notamment *Les Décodeurs* (2014) pour le journal *Le Monde*, l'émission de radio *Le Vrai du Faux* (2012) de *France Info* ou encore *AFP Factuel* (2017) pour l'*Agence France Presse*.

Initialement dédié à contrôler ponctuellement la factualité des propos tenus par des responsables politiques ou des personnalités publiques, le travail des *fact-checkers* est aujourd'hui de plus en plus animé par une volonté pédagogique : celle d'aider les individus à évaluer la fiabilité des sources et des contenus auxquels ils sont confrontés au cours de leur navigation en ligne (Bigot, 2019 ; Graves et Mantzarlis, 2020). Afin de toucher une audience dépassant le simple cercle de leurs lecteurs, plusieurs médias traditionnels ont rejoint le programme de *fact-checking* lancé par Facebook en décembre 2016. En France, quatre médias — *20 Minutes*, *l'AFP*, *Les Décodeurs du Monde* et *Les Observateurs de France 24* — collaborent actuellement avec la plateforme pour vérifier la qualité des contenus qui y circulent.[69] Lorsque l'un d'eux classe une information comme fausse, Facebook envoie un message d'alerte aux utilisateurs souhaitant la partager ou affiche un lien redirigeant vers un article de *fact-checking* juste en-dessous pour indiquer que le contenu est contesté.[70] Les classifications des *fact-checkers* peuvent aussi être utilisées par Facebook pour réduire la visibilité des publications dans les fils d'actualité des utilisateurs (Théro et Vincent, 2022).

L'augmentation des rubriques de *fact-checking* dans de nombreux pays, ainsi que l'élargissement de leurs contenus, dédiés non plus seulement à vérifier les propos de personnalités politiques mais plus largement à corriger les erreurs attribuées au grand public, peut s'expliquer par deux dynamiques. D'une part, la numérisation de l'écosystème informationnel. Après avoir perdu leur monopole sur la publication d'information dans l'espace public face à l'émergence des réseaux sociaux, les médias *mainstream* ont cherché à réaffirmer l'autorité de la profession journalistique. La multiplication des rubriques de *fact-checking* est ainsi devenue l'un des principaux outils de cette relégitimation (Vauchez, 2019).

---

[69] https://www.facebook.com/journalismproject/programs/third-party-fact-checking/partner-map

[70] Les Décodeurs. (2018, 9 janvier). Fausses nouvelles : comment fonctionne le partenariat entre « Le Monde » et Facebook. *Le Monde*. https://www.lemonde.fr/les-decodeurs/article/2018/01/09/comment-fonctionne-le-partenariat-entre-le-monde-et-facebook-sur-les-fausses-nouvelles_5239464_4355770.html

D’autre part, ce renouvellement des pratiques de vérification relève, entre autres, d’un mouvement plus général de technicisation et de rationalisation des métiers de l’information et d’une transformation du système de formation des journalistes. À l'ère de l'*open data* et du *big data*, un nombre croissant de données est mis à disposition par les gouvernements, les institutions publiques, les entreprises ou les citoyens. Ces ressources offrent de nouvelles opportunités aux journalistes pour produire des enquêtes approfondies et rigoureuses, fondées sur des données chiffrées, et donnent ainsi lieu à des pratiques émergentes comme le *data journalism* (Antheaume, 2016 ; Parasie, 2022).

Comment les publics réagissent-ils face à ces différents types de corrections factuelles ? Dans quelle mesure parviennent-elles à réduire le partage de *fake news* ou l’adhésion à des énoncés erronés ? Ces dernières années, de nombreuses enquêtes reposant sur des méthodes expérimentales ont cherché à évaluer l’efficacité du *fact-checking* pour lutter contre les *fake news*. Si quelques études ont noté une absence d’effet (Garrett et Weeks, 2013), ou ont observé un effet rebond (i.e. *backfire effect*), c’est-à-dire un renforcement des croyances ou attitudes préexistantes des individus (Nyhan et Reifler, 2010), plusieurs méta-analyses ont permis de faire ressortir un effet globalement positif du *fact-checking* (Chan et al., 2017 ; Walter et al., 2020 ; 2021). Cependant, ces études présentent certaines limites importantes. Tout d’abord, la majorité d’entre elles ont été menées dans des conditions expérimentales artificielles, ce qui soulève des questions quant à la validité écologique de leurs résultats. Les conclusions tirées de ces recherches ne sont pas nécessairement représentatives des réactions réelles des publics dans des situations de la vie quotidienne. De plus, les recherches existantes se concentrent principalement sur les effets correctifs à court terme de *fact-checks* isolés, en négligeant souvent l’impact à plus long terme ou les changements culturels et systémiques que les initiatives de *fact-checking* cherchent à induire (Dias et Sippitt, 2020 ; Roozenbeek et al., 2024). Un manque d'études longitudinales empêche ainsi de mesurer l’effet durable de *fact-checks* sur les croyances et comportements des individus, ce qui est crucial pour évaluer leur réelle efficacité.

Le *fact-checking* fait aussi l’objet de nombreuses critiques dans le débat public. Yarif Tsfati et al. (2020) interrogent notamment la responsabilité des médias *mainstream* dans la dissémination de *fake news*. Alors que ces médias visent à corriger certaines informations erronées, ils contribuent par là même à leur conférer une certaine visibilité. Autrement dit, même si la majorité du public n'est pas directement exposée aux *fake news,* elle est constamment informée par les médias *mainstream* de leur existence, souvent sans avoir eu connaissance de ces informations préalablement. Par ailleurs, les classements produits par les *fact-checkers* tendent à reproduire les critères dominants de légitimité symbolique au sein de l'espace professionnel. Cela s'explique en partie par le fait que les *fact-checkers* occupent eux-mêmes une position privilégiée dans les hiérarchies de la profession, bénéficiant d'une proximité avec les grandes écoles de journalisme et d'équipes de journalistes associés aux journaux dominants. Parfois accusées de maintenir un regard dominant sur l'espace médiatique français (Lordon, 2017), les rubriques de *fact-checking* recensent peu de *fake news* provenant des médias *mainstream*. Ces choix éditoriaux peuvent contribuer à augmenter la défiance des publics envers les médias traditionnels. Par exemple, le Pew Research Center a montré que 70 % des Républicains aux États-Unis estiment que les *fact-checkers* sont biaisés (Walker et Gottfried, 2019).

### 1.2.2. Régulation institutionnelle et responsabilisation des plateformes

À travers le monde, de nombreuses institutions étatiques ou supranationales ont mis en place des lois ou des réglementations pour lutter contre les *fake news* et encadrer la circulation de contenus sur les plateformes numériques.[71] Cette section se concentre sur les principales mesures de régulation adoptées en France et au sein de l'Union européenne, tout en présentant les différents enjeux et questions qu'elles soulèvent (pour une synthèse, voir Badouard, 2020).

En France, la diffusion de fausses nouvelles est encadrée par le droit français depuis la loi de 1881 sur la liberté de la presse.[72] De nouvelles mesures ont cependant été prises ces dernières années pour faire face aux nouveaux mécanismes de viralité et de propagation associés au

---

[71] Par exemple, la loi *NetsDG* a été adoptée en Allemagne en juin 2017 par le Bundestag, et le *Protection from Online Falsehoods and Manipulation Act* par le Parlement de Singapour en mai 2019. Pour découvrir plus d'initiatives mises en place à travers le monde, cette carte interactive réalisée par Poynter peut être consultée : https://www.poynter.org/ifcn/anti-misinformation-actions/

[72] « Art. 27 – La publication ou reproduction de nouvelles fausses, de pièces fabriquées, falsifiées ou mensongèrement attribuées à des tiers, sera punie d'un emprisonnement […]. »

numérique. À la suite des vœux à la presse exprimés par Emmanuel Macron, le 4 janvier 2018, plusieurs députés ont déposé une proposition de loi « relative à la lutte contre les fausses informations ». Jugée « inefficace, voire dangereuse »[73], celle-ci a été rejetée par le Sénat en juillet 2018. Après la convocation d'une commission mixte paritaire, une proposition de loi « relative à la lutte contre la manipulation de l'information en période électorale » a finalement été adoptée en novembre 2018.[74] Celle-ci vise à renforcer l'obligation de transparence et de coopération des plateformes, à confier de nouvelles compétences au Conseil supérieur de l'audiovisuel (CSA, désormais Arcom) et à encourager l'éducation aux médias. Le texte permet également, pendant les trois mois précédents une élection, à un juge des référés d'ordonner le retrait sous quarante-huit heures « d'allégations ou imputations inexactes ou trompeuses d'un fait de nature à altérer la sincérité du scrutin à venir [...] diffusées de manière délibérée, artificielle ou automatisée et massive par le biais d'un service de communication au public en ligne. ». Cette dernière mesure a toutefois été largement critiquée en raison du risque à limiter la liberté d'expression par une censure rapide, ainsi que de la difficulté à définir clairement ce que constitue une « allégation trompeuse » (Hochmann, 2018). D'autres initiatives législatives ont également été mises en place, telles que la loi Avia de mai 2020, qui cible les contenus haineux en ligne et inclut des dispositions pour lutter contre la diffusion de fausses informations. Cette loi impose aux plateformes de retirer les contenus manifestement illicites dans un délai de vingt-quatre heures, sous peine de sanctions. Enfin, la création du Service de vigilance et de protection contre les ingérences numériques étrangères (Viginum) en juillet 2021 a renforcé la lutte contre la désinformation, en ciblant spécifiquement les tentatives d'ingérence étrangère, en particulier lors des périodes électorales.

Au niveau européen, la Commission européenne a créé en 2017 un Groupe d'experts de haut niveau sur les *fake news* et la désinformation (HLEG) et a lancé une consultation publique. En janvier 2018, la remise d'un rapport intitulé : « Vers une approche multi-dimensionnelle de la désinformation » initie un ensemble de politiques européennes : la création de l'East Stratcom Task Force destinée à surveiller la désinformation étrangère, le programme

[73] Bardo, A. (2018, 5 novembre). Loi sur les fake news : le Sénat rejette à nouveau le texte. *Public Sénat*. https://www.publicsenat.fr/actualites/non-classe/loi-sur-les-fake-news-le-senat-rejette-a-nouveau-le-texte-135105
[74] http://www.legifrance.gouv.fr/affichTexte.do?cidTexte=JORFTEXT000037847559&categorieLien=id

European Digital Media Observatory (EDMO) qui vise à constituer dans chaque pays européen des hubs rassemblant journalistes, chercheurs et éducateurs aux médias comme le projet DE FACTO en France[75]. En 2018, la Commission européenne a élaboré un code de bonne pratique sur la désinformation (*Code of Practice on Disinformation*) signé par des plateformes en ligne telles que Facebook, Google et Twitter. Révisé en 2022, ce code encourage les plateformes à prendre des mesures pour lutter contre la désinformation en ligne, notamment en favorisant la transparence et en limitant les revenus des sites diffusant de fausses informations. Avec le *Digital Service Act* (DSA), adopté en octobre 2022, puis entré en vigueur progressivement jusqu'en février 2024, la Commission européenne a engagé un grand chantier destiné à mettre en place un cadre réglementaire permettant de définir un cahier des charges beaucoup plus précis et contraignant à l'égard des plateformes concernant la désinformation et les discours de haine. L'objectif, notamment, est de renforcer les obligations des plateformes en matière de transparence, de modération des contenus et de protection des utilisateurs, tout en affirmant leur responsabilité quant aux contenus hébergés en ligne.

Les enjeux soulevés par ces différents textes de lois et mesures de régulation sont à la fois nombreux et complexes. Il est difficile d'en faire une synthèse exhaustive et précise. L'une des principales limites qui peut être soulevée est le manque d'accès pour les chercheurs aux données des plateformes afin qu'ils puissent évaluer de façon rigoureuse l'efficacité des régulations mises en œuvre et vérifier que celles-ci soient adaptées aux constats des recherches académiques pour éviter de restreindre la liberté d'expression et de créer des obstacles à la diffusion légitime d'informations (Jungherr, 2024).

### 1.2.3. Éducation aux médias et à l'esprit critique

En parallèle des actions entreprises par des journalistes et des régulateurs, des institutions et des organismes publics, ainsi que des acteurs du secteur privé et associatif ont soutenu diverses initiatives destinées à développer l'éducation aux médias et à l'information (EMI), la littératie numérique, le raisonnement ou l'esprit critique.[76] Ces initiatives se distinguent des précédentes par leur approche proactive et préventive. Plutôt que d'agir *a posteriori*, elles visent à équiper les citoyens de mécanismes de défense et de résistance face aux *fake news* afin d'éviter qu'ils n'y adhèrent ou ne les propagent.

[75] Le projet européen DE FACTO réunit Sciences Po, l'Agence France Presse (AFP), le Centre pour l'Éducation au média et à l'information (CLEMI) et XWiki. Le site web qui rassemble les productions du projet est : https://defacto-observatoire.fr/

[76] Frau-Meigs, D. (2017, 15 octobre). Piloter et coordonner le développement de la « littératie numérique ». *The Conversation*. https://theconversation.com/piloter-et-coordonner-le-developpement-de-la-litteratie-numerique-85434

En France, l'éducation aux médias et à l'information a commencé à connaître des développements institutionnels importants à partir des années 1980, notamment en 1982-1983 avec la remise du rapport d'orientation « Gonnet/Vandervoorde » et la création du CLEMI (i.e. Centre de Liaison de l'Enseignement et des Moyens d'Information) dont la mission est de « promouvoir, [...] par des actions de formation, l'utilisation pluraliste des moyens d'information dans l'enseignement afin de favoriser une meilleure compréhension par les élèves du monde qui les entoure tout en développant leur sens critique ».[77] Toutefois, c'est surtout depuis 2015, à la suite des attentats commis contre le journal *Charlie Hebdo*, que l'éducation aux médias et à l'information a connu un regain d'activités en France — un regain renforcé par les événements associés aux *fake news* et par l'assassinat de Samuel Paty en 2020. Par exemple, le 9 février 2016 a été organisée une Journée d'étude intitulée « Réagir face aux théories du complot ». [78] Par ailleurs, à partir de 2020, différents rapports publics et tribunes ont été publiés, tels que le rapport « sur le renforcement de l'éducation aux médias et à l'information et de la citoyenneté numérique », dirigé par Serge Barbet, délégué général du CLEMI, remis le 1er juillet 2021 ; le rapport sur le « développement de l'esprit critique chez les élèves », dirigé par Alain Abécassis et Paul Mathias, rendu public le 28 avril 2022 ; ou encore le rapport « les lumières à l'ère numérique », dirigé par Gérald Bronner, remis le 11 janvier 2022 à Emmanuel Macron.

Initialement liée à l'Éducation Morale et Civique (EMC) et à la promotion des valeurs républicaines, l'EMI est de plus en plus perçue comme une éducation à l'esprit critique pouvant être intégrée dans toutes les disciplines scolaires. Cette extension permet d'impliquer davantage d'acteurs éducatifs et d'éviter la stigmatisation de certains élèves. Ce

---

Pierre, S. (2018, 8 février). Former à l'esprit critique : une arme efficace contre les fake news. *The Conversation*. https://theconversation.com/former-a-lesprit-critique-une-arme-efficace-contre-les-fake-news-91438

[77] Décret n° 93-718 du 25 mars 1993 relatif au centre de liaison de l'enseignement et des moyens d'information. https://www.clemi.fr/sites/default/files/clemi/Reperes/Decret-définition des missions du CLEMI.pdf

[78] https://www.education.gouv.fr/cid98418/journee-d-etude-reagir-face-aux-theories-du-complot.html

glissement d'un cadrage politique à un cadrage éducatif (Haderbache, 2018) s'accompagne de l'émergence d'une approche axée sur les sciences cognitives (Hupé et al., 2021), qui met l'accent sur les « biais cognitifs » et failles de raisonnement des individus pour expliquer leur porosité à certaines théories du complot ou *fake news*. L'objectif central devient alors d'accompagner les jeunes dans le développement d'un esprit critique, perçu comme un pilier essentiel pour la démocratie. Certaines activités pédagogiques proposent de travailler sur l'évaluation des informations en initiant à la découverte de « mécanismes » de la pensée, en s'appuyant sur la vulgarisation des connaissances sur le cerveau et la cognition.

### 1.2.4. Debunkage sceptique et vulgarisation scientifique

En marge des actions menées contre les *fake news* par des institutions ou des professionnels des mondes du journalisme, du droit et de l'éducation, coexistent également des initiatives de vulgarisation scientifique (Letierce et al., 2010 ; Debove et al., 2021) ou de débunkage portées par des individus, des associations, des collectifs ou des mouvements qui se disent engagés pour la défense de la science, de la raison ou de l'esprit critique. Il s'agit en particulier d'hommes jeunes et assez éduqués (e.g. ingénieurs, médecins, jeunes chercheurs, enseignants) qui se qualifient souvent de « sceptiques », de « zététiciens », de « debunkers », de « rationalistes » ou encore de « gatekeepers » (Dauphin, 2022 ; 2023).

Ces acteurs sont très actifs sur des réseaux sociaux comme Twitter ou Youtube pour démystifier des croyances qu'ils jugent « fausses » ou pour vulgariser différents enjeux et résultats de recherche (Luneau et al., 2023). Comme l'explique Sylvain Cavalier, créateur de la chaîne Youtube Le DeBunker des étoiles : « Jusqu'en 2011, YouTube était quasi exclusivement une tribune pour les complotistes. Désormais, on voit fleurir des vidéos de contre-argumentation. Il est essentiel que les 'débunkers' et les sceptiques occupent le terrain ».[79]

---

[79] Bilem, V. (2018, 14 janvier). Qui sont les zététiciens, ces chasseurs de fake news sur YouTube ? *Les Inrockuptibles*. https://www.lesinrocks.com/actu/qui-sont-les-zeteticiens-ces-chasseurs-de-fake-news-sur-youtube-129789-14-01-2018/

Initialement focalisés sur la démystification des phénomènes paranormaux ou sur la réfutation des théories complotistes, les vidéos et tweets des membres des cercles zététiciens et rationalistes ont progressivement élargi leur champ d'intervention en se tournant vers des sujets scientifiques et techniques, comme le glyphosate, les OGMs ou le nucléaire, faisant l'objet d'importantes controverses dans le débat public. Par exemple, l'étude d'Antoine Segault (2021) sur Twitter a mis en évidence comment la diffusion rapide de rumeurs portant sur la contamination radioactive de l'eau potable par du tritium durant l'été 2019 a vite été contrecarrée par des contre-arguments visant à réaffirmer la place de la rationalité scientifique dans le débat public.

À ce jour, peu d'enquêtes empiriques ont étudié l'impact de ces interventions sur le grand public. Quelques critiques ont néanmoins été formulées à l'égard de ce mode d'action par différents acteurs. Par exemple, dans leur livre *Les Gardiens de la Raison*, Stéphane Foucart, Stéphane Horel et Sylvain Laurens (2020) avancent que les individus qui font appel à l'autorité scientifique dans l'espace public seraient animés par des intérêts économiques ou idéologiques. Selon les trois auteurs, la défense de la science et de la raison est « un projet politique volontiers financé par l'argent des industriels libertariens, [...] qui porte la marque d'une anti-environnementaliste et antiféministe ». D'après une analyse historique de Sylvain Laurens (2019) retraçant l'évolution des valeurs des mouvements rationalistes, la défense de la science et de la raison s'inscrirait depuis les années 1980 dans « une épistémologie de marché » après être passés par une « épistémologie engagée » dans les années 1930 et une « épistémologie expérimentale » dans les années 1960. Pour le chercheur, le rationalisme, initialement porteur d'un engagement social et humaniste, s'est transformé en un mouvement davantage axé sur la défense du progrès technologique et industriel. D'autres approches critiques voient également dans ces mouvements rationalistes et zététiques des tentatives de dérégulation des normes de véridiction savante, évacuant toute la dimension collective du travail de véridiction scientifique et transposant la méthode scientifique à l'échelle du raisonnement individuel (Andreotti et Noûs, 2020).

Si l'enquête réalisée sur la « Galaxie du Vrai/Faux » dans le cadre du projet de recherche mené pour l'Anses (Luneau et al., 2023) apporte des nuances aux arguments défendus par Stéphane

Foucart et ses co-auteurs, elle confirme l'augmentation des discours positivistes et rationalistes dans le débat public et la place occupée par les communautés qui défendent des méthodes expérimentales et reposent sur des approches de sciences cognitives. Un questionnement peut alors être soulevé sur les conséquences de l'augmentation d'un tel régime discursif sur les échanges qui se déroulent dans les espaces publics numériques. En mettant en avant l'irrationalité supposée des individus, leurs biais cognitifs et leur manque de connaissances scientifiques (Ward et al., 2019), ces acteurs contribuent-ils à imposer des contraintes de validité scientifique de plus en plus rigides sur les prises de parole en ligne ? Cette pression ne risque-t-elle pas alors de pousser les personnes les plus méfiantes à l'égard des autorités scientifiques à mener leurs propres recherches, non seulement en allant puiser auprès de références n'ayant pas encore fait preuve d'autorité dans les arènes de débats scientifiques, mais aussi directement dans l'expertise scientifique ? (Berriche, 2021). Par exemple, si l'on reprend le cas du documentaire *Hold-Up*, il est intéressant de souligner deux éléments qui ont peu été mis en avant par les discours médiatiques. D'une part, le fait que les internautes ne sont plus seulement exposés aux résultats scientifiques qui font autorité dans les arènes de débats scientifiques par le biais des instances servant d'intermédiaires entre la science et le grand public, tels que les médias traditionnels et les institutions sanitaires, mais sont de plus en plus confrontés à des disputes épistémiques sur les réseaux sociaux. D'autre part, la manière dont ce déconfinement de l'expertise et l'omniprésence des termes scientifiques dans l'espace public numérique peut conduire une partie des utilisateurs des réseaux sociaux à douter de la science institutionnelle et à la contrecarrer par une rhétorique statistique et positiviste afin d'établir leurs propres « vérités alternatives ». L'un des enjeux de cette thèse sera ainsi de questionner si la constitution des *fake news* en problème public n'aurait pas paradoxalement concouru à renforcer les pratiques de vérification des publics en imposant de nouvelles contraintes sur leurs prises de parole.

L'analyse effectuée dans cette section a permis de montrer comment une pluralité d'acteurs — journalistes, juristes, hauts fonctionnaires, enseignants, chercheurs, etc. — ainsi que divers types de structure — médias, institutions, associations, collectifs de citoyens, *think tank*, etc. — se sont mobilisés ces dernières années pour lutter contre le problème des *fake news*. Ces acteurs et structures peuvent être considérés comme des entrepreneurs de cause dans la mesure où ils ont contribué à définir les *fake news* comme un enjeu de société majeur nécessitant des réponses politiques, législatives, technologiques, etc. En expliquant principalement le problème des *fake news* par des causes technologiques et cognitives, ces acteurs ont orienté l'action publique vers des initiatives visant à réguler la circulation d'informations et à éduquer les citoyens, rappelant les réactions face à des paniques morales passées, telles que celle déclenchée par *La Guerre des Mondes*. Si ces mesures et actions ne sont pas dénuées d'intérêt et d'utilité publique, elles sont parfois en décalage avec les constats des enquêtes empiriques et risquent d'occulter d'autres manières d'appréhender les *fake news*, tenant compte notamment de dimensions sociales et structurelles plus profondes. Dans une perspective critique, il peut être pertinent de se demander si cette mobilisation contre les *fake news* ne contribue pas à renforcer une forme d'idéologie dominante, soutenue par les élites politiques, les grands médias et les institutions traditionnelles (Bourdieu et Boltanski, 1976). En effet, en désignant les *fake news* comme une menace commune, ces acteurs consolident leur place dans l'espace public et réaffirment l'autorité des structures déjà établies.

## 1.3. La production d'un paradigme dominant

Selon le « cycle sisyphéen des paniques technologiques » décrit par Amy Orben (2020), une fois que des troubles sociaux ont été directement imputés à l'émergence d'innovations médiatiques et technologiques, et que divers acteurs se sont mobilisés dans le débat public pour les contrer, des équipes de chercheurs sont à leur tour sollicitées pour étudier l'ampleur du problème, mesurer son impact et évaluer l'efficacité des interventions mises en place pour le juguler. L'enjeu de cette section est ainsi de rendre compte des principaux soutiens institutionnels, politiques et financiers accordés aux études sur les *fake news* et d'identifier les questions de recherche et les approches disciplinaires favorisées jusqu'à présent. Nous examinons ensuite si les cadres théoriques et méthodologiques actuellement dominants dans la littérature académique sur les *fake news* renforcent les représentations des publics et de l'écosystème informationnel véhiculées par les discours dominants en réactivant des paradigmes d'effets forts.

### 1.3.1. Appels à projets, financements et orientation des questions de recherche.

Les préoccupations soulevées par les *fake news* ont eu pour conséquence d'offrir un volume considérable de financements venant d'institutions publiques, d'organismes de recherche, d'entreprises ou de fondations privées créées pour soutenir les initiatives de *fact-checking*, de régulation de contenus, d'éducation aux médias ou à l'esprit critique visant à préserver la démocratie et la qualité de l'écosystème informationnel contemporain (Anderson, 2021).

Par exemple, ces dernières années, l'Union européenne a financé plusieurs projets de recherche sur les *fake news* à travers son programme Horizon 2020.[80] Des initiatives comme SocialTruth, WeVerify ou Co-Inform ont développé des solutions technologiques pour identifier et contrer les *fake news*. Le Conseil Européen de la Recherche (ERC) a également soutenu des travaux de recherche plus théoriques tels que le projet COMPROP, dirigé par Philip Howard de l'Oxford Internet Institute, qui étudie l'impact des algorithmes et des bots sur le discours politique en Europe. Le projet FARE, quant à lui, explore les mécanismes de propagation des *fake news* en combinant des méthodes expérimentales et informatiques, tandis que le projet GoodNews utilise le *deep learning* pour détecter les *fake news*.

Parallèlement, des partenariats entre universitaires et plateformes ont aussi vu le jour. Par exemple, en 2019, le Conseil de recherche en sciences sociales (SSRC) a accordé des financements à 12 projets de recherche, dont un porté par le médialab de Sciences Po, visant à mieux comprendre l'impact des réseaux sociaux sur la société et la démocratie en leur permettant d'accéder à des données de Facebook via l'organisme *Social Sciences One*.[81] En août 2020, Facebook a par ailleurs lancé une nouvelle initiative de recherche pour étudier si les réseaux sociaux amplifient la polarisation ou influencent la participation démocratique, comme le vote. [82]

---

[80] https://commission.europa.eu/strategy-and-policy/coronavirus-response/fighting-disinformation/funded-projects-fight-against-disinformation_en

[81] Social Media and Democracy Research Grants, 2019
https://items.ssrc.org/from-our-programs/social-media-and-democracy-research-grants-grantees/
https://socialscience.one/blog/first-grants-announced-independent-research-social-media%E2%80%99s-impact-democracy?admin_panel=1

Enfin, plusieurs acteurs privés participent également activement à la promotion de l'esprit critique et du *fact-checking*. Google par exemple finance via son programme Google News Initiative des projets d'éducation aux médias et propose des formations en partenariat avec l'AFP et First Draft pour sensibiliser à la désinformation. De son côté Facebook, par le biais de son initiative Meta Journalism Project, soutient le travail des médias de vérification. [83] En France, des fondations et organismes ont aussi été créés pour promouvoir l'esprit critique et soutenir des recherches liées à ce sujet. [84]

Ces différents exemples de financement et de partenariats montrent une priorité accordée aux projets qui visent à évaluer les effets des *fake* ou à développer des solutions techniques ou éducatives pour les combattre. En plaçant les réseaux sociaux et l'esprit critique au cœur de la problématique des *fake news*, ces soutiens institutionnels ou privés favorisent dès lors des disciplines scientifiques centrées sur l'étude des technologies numériques ou des mécanismes cognitifs des individus au détriment de perspectives davantage orientées vers l'analyse de dynamiques sociales et structurelles.

Deux grandes approches dominent ainsi actuellement le champ des études contemporaines sur les *fake news*. La première, issue des sciences computationnelles (Lazer et al., 2009), utilise des données massives extraites des plateformes pour analyser les comportements numériques, définissant l'influence principalement comme une conséquence de l'exposition à certains contenus, sans toujours approfondir les méthodes de mesure de cette influence. La seconde, venant de la psychologie et des sciences cognitives, appréhende l'influence à travers des expériences ou questionnaires sur les opinions, en l'interprétant comme une croyance motivant des attitudes ou des comportements.

[82] Meta. (2020, August 31). *New Facebook and Instagram research initiative to look at US 2020 presidential election*. https://about.fb.com/news/2020/08/research-impact-of-facebook-and-instagram-on-us-election/
[83] Meta. (2022, 1 avril). *Meta's investments in fact-checking*. PROGRAMME IMPACT. https://www.facebook.com/formedia/blog/third-party-fact-checking-industry-investments
[84] On peut citer par exemple la fondation Reboot : https://reboot-foundation.org/

Si cette thèse a bénéficié de certains de ces soutiens, les analyses présentées dans ce chapitre ont permis de prendre du recul vis-à-vis de ce cadrage dominant afin de développer un questionnement autonome et indépendant (Ferron et Le Bourhis, 2020). Par ailleurs, cet intérêt de recherche pour les médias s'inscrit dans un parcours personnel et professionnel de longue date.

### 1.3.2. *Big data*, sciences cognitives et modèle de l'homo numericus irrationnel.

Cette section explore dans quelle mesure les travaux de recherche reposant sur des méthodes de *big data* ou de psychologie cognitive contribuent à renforcer les représentations des publics et de l'écosystème informationnel véhiculées par les discours publics en réactivant des paradigmes anciens de communication tirés de la « psychologie des foules » (Le Bon, 1895) ou du modèle de la « seringue hypodermique » (Laswell, 1927).

La plupart des articles publiés dans des revues d'informatique ou de psychologie cognitive sont presque systématiquement introduits par une dramatisation des désordres de l'écosystème informationnel numérique mettant en évidence le volume, la variété et la virulence de la production d'énoncés faux, douteux ou manipulés sur les réseaux sociaux (Lewandowsky et al., 2017). Cette toile de fond d'un immense capharnaüm conduit les chercheurs à mobiliser de façon plus ou moins explicite des approches théoriques représentant les individus comme irrationnels. En effet, les modèles classiques de rationalité, postulant l'existence de citoyens rationnels et informés sont perçus comme inadaptés par les chercheurs en sciences cognitives pour appréhender les comportements des individus, notamment dans l'écosystème informationnel numérique, dans la mesure où ces modèles reposent sur l'idée d'un marché de l'information optimal et ne tiennent pas compte de la surcharge cognitive associée à un contexte d'abondance informationnel (Bronner, 2021).

De nombreux chercheurs considèrent alors que les théories à processus duels (Evans, Stanovich, 2013 ; Pennycook et al., 2015 ; De Neys, 2017) – également connues sous le nom de « système 1/système 2 » (Kahneman, 2011) – offrent un cadre explicatif plus pertinent pour étudier les réactions des individus face aux *fake news.* Ces approchent invitent à distinguer deux modes de pensée : d'une part, le Système 1, intuitif, rapide et émotionnel ; de l'autre le Système 2, lent, logique et rationnel. À la gourmandise impulsive du Système 1 pour tout ce qui est rapide et intuitif, s'oppose la paresse du Système 2 pour tout ce qui demande effort et réflexion. Dans cette perspective, c'est donc la capacité à laisser le système 2 prendre le relais du système 1 qui distingue les bons des mauvais raisonneurs.

Afin de mesurer les compétences de raisonnement des individus, les études reposant sur ce cadre théorique administrent fréquemment à leurs participants un test de réflexion cognitive (CRT) composé d'une série d'énigmes pour lesquelles la première réponse qui vient à l'esprit est souvent fausse. Par exemple, un problème de mathématique en apparence élémentaire est fréquemment mobilisé : « Une batte de baseball et une balle coûtent 1,10 $. La batte coûte 1 $ de plus que la balle. Combien coûte la balle ? ». À cette question, plus de 80 % des personnes interrogées répondent : 0,10 $. Une réponse intuitive qui semble couler de source. Néanmoins, celle-ci est erronée. La bonne réponse est en réalité 0,05 $ car 0.05 + 1.05 = 1,10. Ce simple exercice de logique montre, d'après les théories à processus duels, comment le système 1 peut induire les individus en erreur, et comment le fait d'inhiber ses intuitions premières par le système 2 permet d'aboutir à une réponse correcte. En recourant au CRT, de nombreuses études ont ainsi montré que les individus qui étaient les plus réflexifs (i.e. ceux obtenant les meilleurs scores au CRT) étaient aussi les plus aptes à discerner les informations vraies des informations fausses (Bronstein et al, 2019 ; Pennycook et Rand 2019 a ; 2020 ; Stanley et al., 2020), et cela que les contenus médiatiques soient idéologiquement concordants ou discordants avec leurs opinions politiques. Ces travaux concluent ainsi que la principale cause de l'adhésion à de fausses informations ou du partage de *fake news* se trouve dans l'esprit humain : si les individus ont tant de mal à discerner le vrai du faux, c'est parce que leurs capacités de raisonnement sont défaillantes. Au lieu de procéder par inférence bayésienne ou par déduction, les individus suivent leurs intuitions et leurs émotions avant de faire des choix ou de prendre des décisions.

Les études dominantes sur les *fake news* tendent ainsi à réactiver des paradigmes d'effets forts en présentant les individus comme peu réflexifs et facilement manipulables. Ces études reposent cependant sur des unités d'analyse différentes que celles des modèles de la psychologie des foules ou de la seringue hypodermique tout en passant d'une approche béhavioriste (i.e. consigner les comportements humains sans chercher leurs raisons) à une perspective cognitiviste. Les émetteurs et récepteurs d'informations ne sont plus considérés comme de grands agrégats, tels que les industries culturelles, les médias de masse ou les foules anomiques souvent appréhendées sous l'angle d'une psychologie collective d'inspiration freudienne (Lasswell, 1930). Ils sont désormais transformés en une collection d'individus singularisés, ciblés par une industrie de la persuasion qui ne diffuse plus un message unique, mais des contenus personnalisés et sur mesure.

En réactivant des paradigmes d'effets forts à travers des modèles de recherche centrés sur les biais cognitifs, ces travaux exercent un effet performatif (MacKenzie, 2008) sur la façon dont nous nous représentons la réception de l'information à l'ère du numérique. Ils contribuent à renforcer l'idée que les internautes, submergés et désorientés dans un chaos informationnel, sont devenus particulièrement vulnérables aux manipulations.

### 1.3.3. Cécité par rapport aux débats de sociologie sur les effets des médias

Alors que la grande majorité des études de *big data* et de psychologie expérimentale conduites sur les *fake news* réactivent des paradigmes d'effets forts, elles font rarement référence à la longue tradition des études de communication portant sur les effets des médias qui les ont pourtant contestés (Marwick, 2018 ; Anderson, 2021). Cette section est l'occasion de mettre en perspective les études contemporaines sur les *fake news* avec des débats qui animent les recherches menées en sociologie des médias et de la communication sur les effets des médias depuis les années 1940 (Katz, 2001 ; Bryant et Thompson, 2002). Elle permet d'explorer si d'autres paradigmes de sciences sociales que celui des effets forts sont plus adaptés pour étudier les *fake news* et ainsi d'amorcer la construction du cadre théorique de la thèse.

Commençons par revenir sur l'enquête pilotée par Hadley Cantril après la diffusion de l'émission *La Guerre des Mondes* d'Orson Wells. À l'époque, ses collaboratrices Herta Herzog et Hazel Gaudet étaient déjà parvenues à des conclusions beaucoup plus nuancées après avoir mené des entretiens avec des auditeurs. Toutefois, celles-ci n'ont pas pu les faire entendre et leurs contributions à l'étude ont été peu valorisées (Pooley et Socolow, 2013b). Cette situation a généré des tensions personnelles et méthodologiques entre Hadley Cantril et Paul Lazarsfeld, qui ont finalement mené à leur séparation.

Le Princeton Radio Research Project a alors été transféré à l'université de Columbia, où il est devenu en 1944 le Bureau of Applied Social Research (BASR), l'un des centres de recherche les plus dynamiques de l'Amérique d'après-guerre, sous la direction de Paul Lazarsfeld (Cefaï, 2008). Dans une succession d'ouvrages fondateurs, *The People's Choice* (Lazarsfeld et al., 1948), *Voting* (Berelson et al., 1954) et, surtout, *Personal Influence* (Katz et Lazarsfeld, 1955), Paul Lazarsfeld et l'École de Columbia vont remettre en question le modèle traditionnel de la communication de masse, qui supposait une influence directe et puissante des médias sur l'audience, et introduire le modèle de la communication à double étage. Par exemple, dans l'enquête pionnière réalisée en 1946 dans la ville de Decatur (Illinois), Katz et Lazarsfeld ont mené des entretiens auprès de 800 femmes pour étudier l'influence des médias sur leurs opinions et choix de consommation. Leurs résultats ont montré que les décisions des individus étaient davantage façonnées par leurs relations personnelles que par leur exposition directe aux médias. Ils ont ainsi établi que les effets médiatiques étaient « faibles » parce que les messages étaient filtrés par des processus sociaux collectifs. Les opinions et les comportements des individus étaient médiés par l'influence d'autres personnes dans leur entourage — les « leaders d'opinion » — et non directement par le contenu médiatique.

La grande originalité de l'enquête d'Elihu Katz et Paul Lazarsfeld est d'avoir prêté attention aux groupes d'appartenance (familiaux, amicaux, professionnels, territoriaux) des individus afin d'opposer à l'influence directe du média, une autre force de persuasion, indirecte, issue de la structure des liens interpersonnels au sein de laquelle s'isole la figure du leader d'opinion. Devenu central en science politique, ce paradigme conçoit la société comme une structure stable dans laquelle les grandes variables de la socialisation des individus jouent un rôle prépondérant dans la formation des croyances, l'interprétation des informations ou les attitudes comme le vote et les décisions d'achat. L'identification de ces variables assure le passage du niveau individuel au niveau collectif sans que des perturbations très fortes des choix individuels ne puissent être attribuées aux stratégies de persuasion des médias.

À partir des années 1970, le paradigme des effets minimaux est de plus en plus critiqué (Gitlin, 1978). Le financement des recherches du Bureau of Applied Social Research (BASR) de Columbia par des entreprises et des institutions, dans un but de marketing et d'évaluation de l'efficacité des messages, est souvent mis en avant pour souligner la dépendance de ce paradigme aux demandes des élites dominantes et à l'essor de l'industrie publicitaire (Pollak, 1979). Les critiques se concentrent également sur l'approche libérale de Paul Lazarsfeld, qui cherche à concilier l'autonomie des choix citoyens face à une offre politique plurielle, tout en ne mesurant que des effets à court terme et facilement quantifiables. L'approche empirique d'une mesure individuelle, comportementale et de court terme de l'influence minimise le pouvoir des médias (Mills, 1953). Elle place les messages médiatiques sur le même plan que les réseaux d'interconnaissance et ne porte pas attention à des effets plus structuraux, profonds et de long terme de l'exposition continue aux discours dominants relayés par le système médiatique. Cette vision minimaliste est perçue comme de plus en plus inadaptée face à l'influence croissante de la télévision qui fidélise de larges audiences, structure le débat politique et alimente activement la conversation publique. La société de consommation de masse, les effets uniformisant des industries culturelles, ainsi que les techniques de persuasion de la publicité et du marketing, suggèrent un pouvoir médiatique plus fort et direct sur les publics.

Dans les années 1980, les critiques du modèle dominant des effets faibles fragmentent le champ des recherches sur les médias et la communication et le divisent entre deux approches contrastées.

La première radicalise l'hypothèse des effets faibles mais opère un renversement dans l'appréciation des compétences du public par rapport à la théorie de la communication à double étage. Dans le modèle d'Elihu Katz et Paul Lazarsfeld, les médias ont des effets faibles parce que les individus ne disposent que d'une faible autonomie de jugement et que leurs appartenances sociales et leurs réseaux interpersonnels structurent si fortement leurs choix qu’il est difficile de les faire changer d’opinion. À mesure que les individus s’émancipent de ces appartenances, une plus grande autonomie leur est conférée, conduisant à l’émergence des recherches sur les audiences « actives ».

L'approche des *uses and gratifications*, déjà dans la lignée de Lazarsfeld, reconnaissait une certaine capacité des publics à interpréter les messages médiatiques, mais cette agentivité s'est intensifiée des années 1970 aux années 1990. Les individus gagnent en liberté d'action face aux déterminations sociales, ouvrant la voie à une multitude de lectures et interprétations médiatiques. Cette réorientation des *media studies* des années 1980 reflète les nouvelles formes d’individuation sociale et l'accroissement des capacités critiques des individus. Ceux-ci, désormais moins contraints par les variables sociales lourdes, s’identifient à des communautés interprétatives diverses et dynamiques. Les effets des médias deviennent alors « très faibles », tant les individus s’affranchissent du poids structurant des déterminations de la société industrielle et inventent un nombre considérable de variations interprétatives qui témoignent simultanément de leur dépendance et de leur distance à l’égard des industries culturelles.

Cette valorisation des audiences actives des années 1990 inspire les premières recherches sur les pratiques numériques. Les études sur la culture participative du web, apparues dans les années 2000, s’inscrivent en effet dans cette continuité (Jenkins, 2006 ; Marwick, 2018). Les internautes sont perçus comme créateurs autonomes, produisant du contenu original, plutôt que soumis à l'influence des médias (Castells, 2009 ; boyd, 2016 ; Allard et Vandenberghe, 2003 ; Cardon, 2008). Ces recherches soulignent les capacités expressives des communautés en ligne (Jenkins et al., 2015), laissant de côté la question de leur conformisme ou aliénation. Il faudra cependant attendre la massification des pratiques numériques dans les années 2010 pour que l'internaute, perçu comme actif dans les années 2000, soit à nouveau envisagé sous un angle plus passif, réintroduisant le débat sur les effets et la suggestibilité des publics.

Dans les années 1980, parallèlement au courant des audiences actives, émerge une autre approche critique, soulignant le pouvoir hégémonique des médias et leur rôle dans la fermeture de l’espace public. La thèse des effets forts a toujours coexisté au sein du paradigme lazarsfeldien, notamment à travers le financement par le BASR des recherches de certains exilés de l'école de Francfort. S'inscrivant dans la lignée d'Adorno et Horkheimer (1974), cette critique se concentre sur le rôle des industries culturelles et médiatiques dans la normalisation du conformisme et l'affaiblissement des appartenances traditionnelles, au profit d'une société uniformisée (Ang, 1991). La crise des identifications collectives et des engagements civiques est en partie absorbée par l'influence conformiste des grandes audiences télévisuelles, comme le souligne Robert Putnam (2000).

De nombreux modèles théoriques vont alors se développer pour mettre en évidence des effets plutôt importants des médias comme ceux de la spirale du silence d'Elisabeth Noelle-Neumann (1974), de la structuration asymétrique des débats (Lang et Lang, 1983) ou de l'influence directe, impersonnelle et dirigée des grands médias vers des individus atomisés (Mutz, 1998 ; Zaller, 1992). Ces travaux reprochent au paradigme des effets limités de négliger ce pouvoir vertical, asymétrique et dominant qu'exercent les grands médias pour définir la réalité, rendre légitime certains types d'enjeux et donner de l'importance aux discours des puissants tout en minorant ceux des minorités (Gitlin, 1978).

L'ouvrage *Manufacturing Consent* d'Edward Herman et Noam Chomsky (1988) approfondit cette critique, dénonçant la concentration de la production médiatique par des conglomérats et l'inégale répartition des flux d'information (Bourdieu, 1996 ; McChesney et Nichols, 2002). Leur analyse se concentre sur les contraintes économiques et idéologiques des médias, soulignant leur influence structurante sur le public, bien que la réception des messages soit peu étudiée (Cardon et Granjon, 2013). Le développement d'Internet, en diversifiant l'offre d'information, affaiblit cet argument, et la critique de l'hégémonie médiatique se tourne alors vers les plateformes de réseaux sociaux (Napoli, 2019).

Au sein de cet ensemble de travaux renouant avec les effets forts apparaît également la théorie de l'effet d'agenda (*agenda-setting*) selon laquelle les médias ne disent pas aux gens ce qu'il faut penser mais à quoi ils doivent penser et comment y penser. Cette approche montre comment la couverture médiatique, par des mécanismes comme l'amorçage (*priming*) et le cadrage (*framing*), façonne la perception des événements (Iyengar, 1991). Contrairement aux approches critiques des médias, cette théorie se focalise sur le lien entre discours médiatique et opinion publique à travers des mesures empiriques, sans toutefois s'attarder sur le pouvoir idéologique des médias (Wojcieszak, 2008).

Ce récit standard de la succession des paradigmes de la théorie des médias fait l'objet d'un examen critique ininterrompu au sein de la discipline (Katz, 2001 ; Neuman et Guggenheim, 2011 ; Simonson, 2013), mais il constitue cependant une base suffisamment reconnue pour se demander quel devrait être le cadre théorique des recherches contemporaines lorsque les médias ne dispersent plus leurs messages vers des *foules* ni des *masses*, mais vers des individus en *réseau* (Bennett et Iyengar, 2008 ; Boltanski et Esquerre, 2022). Ce que nous allons faire dans le prochain chapitre.

## Conclusion du premier chapitre

En analysant les *fake news* comme un objet de discours, plutôt que de les considérer d'emblée comme un nouveau phénomène social, ce premier chapitre a pu documenter comment les *fake news* ont été définies comme un problème public. En expliquant différents enjeux contemporains principalement par la circulation virale de *fake news* et par la crédulité des utilisateurs des réseaux sociaux, les discours publics actuels réactivent des grilles de lecture techno-déterministes et psychologisantes, similaires à celles adoptées lors de précédentes « paniques médiatiques et technologiques ». Ce cadrage du problème a favorisé la mise en place de mesures destinées à réguler l'écosystème informationnel et à éduquer les publics, ainsi que la conduite de recherches reposant sur des méthodes de *big data* ou de psychologie expérimentale. Sans être totalement dénuées d'intérêt, ces approches sont limitées car elles partent de représentations des écosystèmes informationnels contemporains et de leurs publics qui sont en décalage avec les constats empiriques de la littérature académiques et négligent les enseignements issus des travaux de sociologie des médias et de la communication. L'objectif du prochain chapitre est donc de définir un autre cadre d'analyse, plus en phase avec les constats empiriques des études académiques, tout en intégrant les acquis des travaux de sociologie des médias et de la communication.

# Chapitre 2. Enquêter sur la réception de *fake news* : décisions théoriques et méthodologiques

Le chapitre précédent a montré comment la construction des *fake news* en problème public a orienté les recherches vers des approches privilégiant des méthodes de *big data* ou de psychologie expérimentale. Sans y faire explicitement référence, la plupart de ces travaux réactivent d'anciens modèles de communication réduisant l'activité de réception d'une information à de simples mécanismes de stimulus et réponses, tout en introduisant de nouvelles représentations de l'écosystème informationnel et de ses publics : l'information n'est plus injectée par des médias de masse dans les cerveaux d'une foule d'individus crédules, mais personnalisée par des algorithmes de recommandations qui isolent dans des bulles de filtre des utilisateurs des réseaux sociaux en proie à des milliers de biais cognitifs.

Ces grilles de lecture sont cependant peu conformes aux constats issus des enquêtes empiriques sur les *fake news* et ne tiennent pas compte des apports des travaux de sociologie des médias et de la communication, en particulier de ceux qui ont remis en cause les théories des effets forts des médias. L'objectif de ce deuxième chapitre est de spécifier notre positionnement au sein de ce champ de recherche et de présenter le cadre théorique et méthodologique adopté dans cette thèse, ainsi que les terrains d'enquête investis, afin d'étudier la réception de *fake news* par les utilisateurs des réseaux sociaux dans l'écosystème informationnel contemporain.

## 2.1. Cadre théorique : une approche pragmatique de la réception en contexte numérique.

Étant donné que les *fake news* n'occupent qu'une faible place dans les habitudes de consommation médiatique des individus et ne sont partagées que par une minorité d'entre eux, il est difficile de concevoir l'écosystème informationnel comme un marché dérégulé et les utilisateurs des réseaux sociaux comme des individus crédules, systématiquement induits en erreur par leurs biais cognitifs. Afin d'expliquer pourquoi la majorité des *fake news* ne sont consultées et partagées que par une petite minorité d'utilisateurs tout en veillant à comprendre pourquoi différents événements dénotant une augmentation des phénomènes de polarisation ont malgré tout été observés dans la société ces dernières années, cette thèse propose de mobiliser les apports théoriques et conceptuels issus de trois grands domaines de recherche : la sociologie de la réception, la sociologie pragmatique et la sociologie des usages du numérique.

### 2.1.1. Participation active des publics et rôles des contextes sociaux dans la construction du sens des énoncés

À rebours des modèles de communication cherchant à évaluer l'influence directe des médias de masse sur les individus, les travaux de sociologie de la réception se sont attachés à ne pas réduire les processus de réception d'informations à la question de leurs effets en interrogeant « non pas ce que les médias font aux gens mais ce que les gens font des médias » (Molénat, 2016, p. 225). Deux apports principaux résultent de ce renversement de perspective.

Le premier est de ne pas considérer les récepteurs d'informations comme des individus passifs, dépourvus de sens critique, mais comme des personnes qui participent activement à la construction du sens des énoncés. Dans l'ouvrage *La culture du pauvre*, publié en 1957 et traduit en Français en 1970, Richard Hoggart a décrit avec une grande finesse les capacités de résistance des classes populaires d'Angleterre face aux produits culturels de masse. Celles-ci « savent en prendre et en laisser » en adoptant une « consommation nonchalante » et en faisant montre d'une « attention oblique » (Passeron, 1994, p. 288). Cet ouvrage a ouvert la voie à l'émergence de l'école de Birmingham et aux *cultural studies*. Dans les années 1970, Stuart Hall, désormais à la tête du Center for Contemporary Cultural Studies, fondé par Richard Hoggart, construit l'un des modèles théoriques les plus influents au sein des travaux de sociologie de la réception : le modèle « codage/décodage ». Selon ce modèle, les médias sont fortement influencés par les structures économiques et institutionnelles qui caractérisent leur environnement. Ils produisent alors une lecture préférentielle qui oriente la présentation des faits. Si le décodage des énoncés médiatiques est largement conditionné par l'ordre culturel dominant, Stuart Hall fait l'hypothèse que les récepteurs peuvent adopter trois modes de lectures différents : (1) une lecture de conformité qui s'aligne sans résistance sur la culture dominante ; (2) une lecture négociée où le récepteur prélève certains segments et les adapte à ses propres codes et valeurs ; enfin, (3) une lecture oppositionnelle à la lecture prescrite (Hall, 1973). Ainsi le modèle de Stuart Hall souligne que bien qu'un sens préféré soit encodé dans les messages médiatiques par des producteurs d'informations, leur interprétation et leur réception sont sujettes à des processus de décodage divergents par les récepteurs. La signification des énoncés n'est pas fixe mais émerge dans la relation entre le message et son destinataire, ouvrant la voie à des interprétations multiples et souvent conflictuelles.

Le second apport met en lumière l'importance des contextes socioculturels dans la réception des messages. Les hypothèses de lecture proposées par le modèle de Stuart Hall ont été vérifiées par des enquêtes empiriques portant sur la réception d'émissions de télévision. David Morley (1980), dans son étude sur un magazine d'actualité télévisé de la BBC dans les années 1970, a utilisé le modèle de Stuart Hall pour examiner comment les récepteurs décodent les contenus médiatiques en fonction de leurs appartenances socio-économiques et culturelles. Ses analyses montrent que les interprétations des messages télévisuels varient en fonction des contextes sociaux des récepteurs et de leur position sociale, mettant en évidence une pluralité de lectures et une diversité de significations attribuées par les publics. Dans la même veine, d'autres recherches ont également montré l'importance des ressources sociales et culturelles des individus dans l'intérêt, l'exposition et la lecture des informations (Le Grignou et Neveu, 1988 ; Katz et Liebes, 1990).

Malgré leurs contributions importantes, les études de réception inscrites dans le sillon des *Cultural Studies* de l'École de Birmingham ont fait l'objet d'importantes critiques au point que certains sociologues comme Louis Quéré (1996) se sont demandés si elles ne seraient pas dans une impasse et s'il ne faudrait pas les abandonner. Selon Louis Quéré, la réception est trop souvent perçue comme un « processus individuel [...] privé ou interne [...] momentané, limité dans le temps et l'espace ». Face à ces limites, il se demande comment reconceptualiser la notion de réception pour l'appréhender comme un phénomène temporel, social et pratique. Pour cela, il propose entre autres de mobiliser les outils issus de la microsociologie goffmanienne ou de l'ethnométhodologie. Cette démarche a été appliquée par Dominique Boullier (2004) dans son livre *La télévision telle qu'on la parle. Trois études ethnométhodologiques.* En adoptant comme terrain d'observation des groupes de travail dans des entreprises, il montre que l'usage de la télévision se prolonge en dehors du laps de temps et du lieu même où le téléspectateur entre en interaction directe avec le téléviseur. Autrement dit, ce n'est donc pas seulement la situation de présence devant la télévision, ou la lecture effective du journal, qui permettent de comprendre la réception : il faut aussi prendre au sérieux les conversations qu'ont les individus entre eux et les traiter comme des prolongements discursifs externes de la réception.

Cette reconnaissance de la multiplicité des situations de réception invite à compléter les approches classiques de sociologie de la réception par une approche pragmatique et interactionniste.

### 2.1.2. Analyse des situations et pluralité des régimes d'action

Alors que les études de réception, ancrées dans la tradition des *cultural studies*, ont montré comment l'interprétation d'un contenu médiatique est largement influencée par les contextes socio-culturels des individus, une approche pragmatique et interactionniste invite à dépasser les modèles qui analysent les processus de réception d'informations uniquement à travers des structures sociales globales ou des schémas culturels figés. Comment ?

Tout d'abord, en restaurant l'analyse de la situation, trop souvent négligée selon Goffman (1964). Or, pour ce dernier, celle-ci possède « une structure propre », « ses propres règles » et « des processus propres » (*Ibid*, p. 149). Il la définit comme « un environnement fait de possibilités mutuelles de contrôle, au sein duquel un individu se trouvera partout accessible aux perceptions directes de tous ceux qui sont “présents” et lui sont similairement accessibles. » (*Ibid*, p. 146). Cette perspective met en avant le rôle central du contexte immédiat de l'interaction, où la coprésence physique crée une dynamique d'influence réciproque entre les participants. Pour William Thomas (1938), il s'agit d'analyser comment les individus agissent selon leur propre définition et compréhension de la situation. D'après son fameux théorème : « si les hommes définissent une situation comme réelle, elle est réelle dans ses conséquences ». Cette formule invite les chercheurs à ne pas définir les situations depuis des cadres analytiques extérieurs mais plutôt à partir des conséquences pratiques qui émanent de la façon dont les acteurs définissent eux-mêmes les situations. En observant et en décrivant des situations concrètes, c'est à dire « le présent de l'action dans son déroulement » (Barthe et al., 2013, p. 180), les chercheurs peuvent ainsi dégager les « règles de grammaire propres à chaque situation », c'est-à-dire « l'ensemble des règles à suivre pour agir d'une façon suffisamment correcte aux yeux des partenaires de l'action » (Lemieux, 2000, p. 110). Par exemple, l'analyse de Luc Boltanski et ses collègues (1984) conduite sur un corpus de lettres adressées au quotidien *Le Monde* a permis de montrer comment les individus doivent désingulariser leur énonciation de leur personne et généraliser leurs propos afin d'émettre des dénonciations d'injustice jugées recevables dans l'espace public.

Ensuite, en adoptant un principe de pluralisme et en partant du principe que les personnes sont dotées de capacités critiques qu'ils déploient de manière dynamique et évolutive en situation (voir Lemieux, 2018, notamment p. 16-18 et 30-33). « Parmi ces capacités, celle à évaluer les capacités des autres et les siennes propres n'est pas la moindre. Ainsi la sociologie pragmatique s'intéresse-t-elle non seulement à la façon dont les individus démontrent aux autres et à eux-mêmes ce dont ils sont capables ou incapables, mais encore à la manière dont de telles capacités et incapacités sont évaluées socialement, et avec quelles conséquences pratiques » (*Ibid*, p. 16). Par exemple, de nombreux individus préfèrent éviter de parler de politique en public de peur d'apparaître incompétents aux yeux des autres, mais se sentent plus légitimes pour formuler des réflexions politiques dans des contextes intimes et privés (Eliasoph, 1998). L'apport des notions d'épreuve et de régime (Boltanski et Thévenot, 1991) est particulièrement pertinent ici. Une situation sociale devient une épreuve lorsque les participants sont confrontés à des défis qui mettent en jeu leur capacité à justifier leurs actions ou à prouver leur compétence. Cela se manifeste notamment lorsque les individus doivent rendre légitimes leurs opinions dans des contextes publics ou semi-publics. Chaque interaction se déroule selon des régimes de justification ou d'action (Boltanski, 1990 ; Thévenot, 2006), à travers lesquels les individus ajustent leurs actions et discours en fonction des attentes sociales, des normes implicites et explicites, et des enjeux spécifiques à chaque situation.

Mobiliser une approche pragmatique et interactionniste offre ainsi des outils complémentaires à ceux des travaux de sociologie de la réception en intégrant une attention fine aux situations concrètes dans lesquelles les individus évoluent et à la pluralité des régimes d'action qu'ils déploient. En reposant sur des notions comme celles d'épreuve ou de de régime, cette approche permet de révéler les capacités critiques des individus et leurs ajustements en fonction des situations d'interactions. De cette manière, elle enrichit les perspectives des sociologies de la réception en offrant un cadre plus dynamique et interactif pour analyser les pratiques qui entourent la réception de contenus médiatiques.

### 2.1.3. Culture participative, interactions asynchrones et visibilité en clair-obscur

Écouter la radio dans sa cuisine en prenant son petit déjeuner. Lire le journal dans le métro en allant au travail. Discuter de l'actualité le midi avec ses collègues. S'indigner des propos tenus par une personnalité politique pendant un dîner avec des amis. Regarder la télé en famille le soir dans son salon. Si les concepts et théories développés par les travaux de sociologie de la réception et de sociologie pragmatique permettent d'étudier ces situations traditionnelles de réception d'informations et d'interactions sociales, la place croissante occupée par les technologies numériques dans les pratiques du grand public invite à mettre à jour certaines catégories d'analyse pour étudier la façon dont les publics s'informent et discutent de l'actualité. En effet, les pratiques informationnelles et conversationnelles des publics ont grandement évolué avec le développement des technologies numériques. Si la télévision reste le premier support d'informations de la majorité des populations, les réseaux sociaux occupent une place de plus en plus importante dans les pratiques informationnelles des individus, notamment chez les plus jeunes. Par exemple, selon des enquêtes du Pew Research Center et du Reuters Institute, 62 % des Américains et 46 % des Européens accèdent aux informations à travers les réseaux sociaux (Gottfried et Shearer, 2016 ; Newman et al., 2016). Toutefois, il ne s'agit pas d'une simple substitution des anciens médias par les nouveaux, mais plutôt d'une accumulation et d'un entrelacement des pratiques. Les technologies numériques ont introduit de nouvelles modalités de réception et d'interaction qui coexistent avec les formes plus traditionnelles. Désormais, l'information peut être consultée de manière fragmentée et sporadique : par exemple, en parcourant des vidéos humoristiques sur TikTok en faisant la queue à la boulangerie ou en répondant à des notifications pendant une réunion. Face à ces transformations, trois changements majeurs méritent d'être soulignés. Nous indiquons parallèlement les contributions des travaux en sociologie des usages du numérique pour appréhender ces nouvelles réalités.

***De la réception à la participation***

Le premier changement concerne l'effacement progressif des frontières entre émetteurs et récepteurs, auteurs et lecteurs de textes (Chartier, 2000 ; Beaudoin, 2002). Alors, certes, bien avant l'existence du web et des réseaux sociaux, les récepteurs et lecteurs d'informations étaient déjà des producteurs de sens à part entière, que ce soit en interprétant librement un énoncé ou en en discutant avec d'autres personnes. Néanmoins, dans la majorité des cas, ces activités d'interprétation et de discussion se déroulaient dans des contextes privés, sans interaction possible avec les émetteurs et producteurs initiaux des messages. Depuis l'émergence des réseaux sociaux, les individus peuvent désormais réagir publiquement et instantanément aux contenus des médias, en les likant, commentant ou remixant. Autant d'actions qui permettent aux utilisateurs de superposer au contrat de lecture fixé par le média (Véron, 1985) un contrat de conversation (Granier, 2011) à travers lequel ils co-produisent du sens en interaction avec d'autres utilisateurs.

Pour mieux appréhender ces dynamiques, il est utile de recourir à la notion de « contrat de communication » proposée par Patrick Charaudeau (2017). Ce concept définit un socle commun qui rend possible l'intercompréhension entre émetteurs et récepteurs, en posant des bases partagées qui stabilisent partiellement le sens tout en laissant place à des variations et des réinterprétations (Charaudeau, 2004, p. 120). En reconnaissant que le sens est en constante construction et négociation, cette notion permet d'articuler à la fois le contrat de lecture et le contrat de conversation qui émerge entre les utilisateurs dans un espace de communication en perpétuelle reconfiguration.

À l'heure actuelle, l'étude de la communication médiatique se pense ainsi de plus en plus dans le cadre du paradigme de la participation (Jenkins, 2006 ; Ségur, 2017). Comme l'ont montré les travaux sur les publics *fans* (Bourdaa, 2016), la réception médiatique devient une expérience sociale qui dépasse le simple moment d'exposition et les récepteurs ne se contentent plus de réagir aux messages, mais participent parfois dès la production (Maigret, 2013). Pour certains chercheurs, l'évolution de ces pratiques remet en cause la notion même de public et il conviendrait désormais de penser la question de la réception plutôt en termes d'usages (Donnat, 2017). Pourtant, cette participation reste inégalitaire. Il existe une large part d'utilisateurs, souvent appelés *lurkers*, qui se contentent d'observer sans interagir (Falgas, 2016). Il apparaît dès lors important d'appréhender la réception de l'information en contexte numérique à l'aune des travaux sur la culture participative sans perdre de vue les apports des approches classiques de sociologie de la réception.

***Interactions asynchrones et co-présence écranique***

Le deuxième changement important est l'émergence de nouvelles configurations spatio-temporelles qui bouleversent les cadres traditionnels d'analyse des interactions. Les études classiques de sociologie de la réception ou de sociologie interactionniste étaient surtout centrées sur la coprésence temporelle et spatiale et les échanges se déroulant en face à face Dans les environnements en ligne, cette coprésence n'est plus nécessairement requise. Dans les interactions dites synchrones comme les messageries instantanées et les chats, les participants sont en situation de coprésence temporelle, à défaut d'être en coprésence spatiale. Dans les interactions dites asynchrones, tels que les forums ou les listes de discussion, les participants ne sont présents ni au même moment, ni dans le même lieu. Valérie Beaudoin et Julia Velkovska (1999) ont proposé de parler de « cadre de participation asynchrone médiatisé par un support électronique » pour désigner ces formes d'interaction.

Toutefois, malgré l'absence de coprésence physique, il est possible d'évoquer une forme de « coprésence écranique ». Cette notion désigne le fait que les participants partagent un même contexte communicationnel défini par des textes visibles à l'écran. Bien que ces textes ne soient pas lus simultanément ni depuis le même endroit, ils constituent une ressource commune (Beaudoin, 2002, p. 219). Dans ce cadre, la coprésence se manifeste à travers l'inscription des participants à l'écran, qui les rend virtuellement présents aux yeux des autres. Les marques de déictiques et d'indexicaux, présentes dans les échanges textuels, suggèrent même souvent une continuité dans la situation d'énonciation, malgré la désynchronisation des interactions. Ainsi, même dans les situations de communication différée, certains principes de l'analyse conversationnelle peuvent s'appliquer.

***Espace de visibilité en clair-obscur et context collapse***

Enfin, le troisième changement important est l'émergence d'espaces de visibilité en « clair-obscur » (Cardon, 2010) et la possibilité croissante de *context collapse* ou d'effondrement de contexte (boyd, 2008)[85]. Sur les réseaux sociaux, les différentes sphères de la vie d'une personne sont plus fréquemment mélangées en un seul espace qu'hors ligne. Une personne peut publier un message qui sera visible par ses amis, sa famille, ses collègues de travail, voire par des inconnus, et cela même si le message initial n'était pas destiné à toutes ces audiences mais seulement à quelques contacts. Ainsi, les réseaux sociaux ont grandement contribué à rendre visible et à entrecroiser ce que la pratique antérieure pouvait garder soigneusement cloisonnée. Cette collision entre des groupes très divers, ne partageant pas du tout les mêmes normes sociales, peut alors entraîner des erreurs d'interprétation, voire des conflits. Par exemple, lorsqu'un utilisateur publie un tweet, il peut être retweeté par n'importe qui, ce qui introduit le contenu à un nouveau public. Ainsi, un message ironique ou humoristique - une fameuse « *private joke* » - initialement destiné à un cercle restreint, peut rapidement être pris au premier degré lorsque diffusé à une large audience, ce qui peut entraîner de nombreux malentendus, voire donner lieu à des rumeurs[86]. Dès lors, les troubles potentiellement entraînés par les situations de *context collapse* n'incombent pas seulement aux personnes à l'origine de publications problématiques mais également aux récepteurs qui les sortent de leur contexte en les partageant auprès d'une audience plus large.

[85] Popularisée par la chercheuse danah boyd (2008), la notion de *context collapse* trouve son origine dans les travaux d'Erving Goffman et de Joshua Meyrowitz. Dans son livre *The Presentation of Self in Everyday Life*, Goffman (1956) a développé la notion de « ségrégation d'audience » pour décrire la façon dont les individus gèrent et adaptent leurs comportements en fonction des différents publics auxquels ils sont confrontés dans leur vie quotidienne. Dans son livre *No Sense of Place*, Meyrowitz (1985) a, quant à lui, été le premier à montrer comment de nouveaux médias comme la télévision et la radio ont créé un « effondrement des frontières sociales » en estompant les barrières entre différents types de publics. Par exemple, des individus peuvent être témoins d'événements internationaux en direct dans le confort de leur foyer, ce qui abolit la distance physique et sociale entre les individus et les événements mondiaux.

Pour éviter ces troubles, les utilisateurs des réseaux sociaux peuvent cependant jouer avec les divers paramètres de visibilité proposés par les plateformes. Cela leur permet ainsi de créer des représentations de leur identité adaptées aux différents publics rassemblés autour de leurs comptes (Cardon, 2008). Les utilisateurs savent également détourner les fonctionnalités proposées par les plateformes pour les adapter aux logiques propres à leur sociabilité. Par exemple, il y a une dizaine d'années, les adolescents distinguaient leurs amis les plus proches de leurs simples contacts Facebook en les indiquant comme des membres de leur famille ou en leur demandant de passer en « d.i » (i.e. discussion instantanée) dans un fil de commentaires publics afin de « rendre visible au public le plus large possible l'intimité la plus privée possible » (Balleys, 2014, p. 79). Ainsi, malgré le brouillage des frontières entre les sphères de la vie privée et publique, les utilisateurs ne subissent pas passivement ce phénomène. Ils redessinent activement ces frontières, jonglant avec les outils numériques pour gérer les audiences multiples et les enjeux d'interprétation. Ces transformations induites par le numérique invitent à examiner comment les notions de règles de grammaire et de compétence, concepts centraux en sociologie pragmatique, s'appliquent dans un environnement numérique.

## 2.2. Cadre méthodologique

Avant de présenter les choix méthodologiques adoptés dans cette thèse, reprenons brièvement l'historiette racontée en début d'introduction. Si l'écriture de fiction se nourrit volontiers d'une fascination curieuse pour la vie quotidienne, elle offre une liberté que l'écriture académique ne permet pas : celle d'adopter un point de vue omniscient, permettant de prétendre tout voir et tout entendre – et même de lire dans les pensées des gens. Mais supposons que Gisèle existe pour de vrai et qu'elle fasse partie de l'échantillon d'utilisateurs des réseaux sociaux étudiés dans cette thèse. Pour analyser les réactions de Gisèle face à différents contenus médiatiques, tout en prenant en compte son contexte social et ses interactions quotidiennes, une approche ethnographique aurait pu être envisagée. Cependant, cette approche peut se heurter à des difficultés pratiques importantes, notamment à l'ère du numérique. Observer une famille dans son salon, par exemple, alors qu'elle regarde la télévision, peut rapidement devenir intrusif. De plus, si chaque membre de la famille consulte en parallèle son téléphone pour lire des messages ou naviguer sur Internet, il est impossible pour un ethnographe de suivre attentivement toutes ces conversations simultanément. Et puis, comment envisager de conduire des observations *in situ* en se plaçant directement « du côté du public » (Le Grignou, 2003) alors que des mesures de distanciation sociale viennent d'être mises en place ?

[86] Slate (2020, 23 Septembre). De la «private joke» à la fake news, itinéraire d'une bien étrange histoire. https://korii.slate.fr/et-caetera/nagui-chevaux-course-au-clic-transforme-blague-rumeur-twitter-medias

Cette section vise à construire un cadre méthodologique permettant de répondre à ces défis. Après avoir discuté des apports et des limites des méthodes classiques d'ethnographie de la réception et des méthodes numériques, nous défendons l'intérêt de développer une approche hybride et combinatoire, articulant des méthodes quantitatives et qualitatives.

### 2.2.1. Apports et limites des méthodes ethnographiques traditionnelles

Afin d'étudier les usages des médias dans les formes ordinaires de la vie sociale des publics et d'éviter d'imposer des catégories déjà constituées à leurs pratiques, les études de réception reposent depuis les années 1970 en grande partie sur des méthodes ethnographiques. Les réactions des publics ne sont plus supposées par des analyses de contenus, ni construites par des indicateurs d'audience, mais observées, écoutées et entendues à partir d'observations *in situ*. Cette approche méthodologique a permis aux chercheurs de donner une voix aux récepteurs d'informations afin de saisir de façon précise et nuancée comment ils interagissent avec les médias, comment ils attribuent du sens aux messages médiatiques et comment ces significations sont influencées par leur milieu social. Dans *Inside Family Viewing*, James Lull (1990) a par exemple réalisé des observations *in situ* dans des foyers afin de comprendre comment la télévision est regardée, quelles discussions elle génère et comment elle façonne les relations entre les membres de la famille. En prêtant attention aux conversations qui surgissent à propos de ce qui est diffusé à l'écran, ainsi qu'aux comportements non verbaux des spectateurs, Lull montre que la télévision joue un rôle central dans l'organisation du temps domestique et dans la régulation des interactions familiales. Il souligne que la télévision ne se contente pas d'être un simple divertissement, mais qu'elle participe à la structuration des routines familiales, à la création d'espaces de socialisation et parfois à la gestion des conflits. Par exemple, il note que la télévision peut être utilisée pour réunir la famille, mais aussi pour marquer des moments de séparation ou d'isolement, comme lorsque chaque membre regarde des émissions différentes ou s'isole dans une autre pièce.

Malgré l'immense apport de ces travaux, quelques chercheurs ont souligné leurs limites. En effet, l'observation prend parfois la forme d'une intrusion qui ne permet pas d'accéder à la situation ordinaire et spontanée de réception. Comme le dit Daniel Dayan (1992, p. 150), le risque est alors de « donner la parole à une fiction de public à propos d'une fiction de texte, en inventant une relation fictive entre les deux ». Une solution pour dépasser ces limites peut être de conduire des études de réception multi-méthodes. Par exemple, Dominique Pasquier (1995) a montré la pertinence de coupler des observations de la réception du feuilleton *Hélène et les garçons* au sein de familles à une lecture attentive du courrier des adolescents et à la passation d'un questionnaire. Cette approche « multi-méthode » lui a permis de développer une compréhension fine des usages que les adolescents font du feuilleton dans les relations qu'ils nouent entre eux et avec les adultes tout en mettant en relation certains types de situation sociale et certains types d'interprétations de la série.

Au-delà des limites inhérentes aux approches ethnographiques traditionnelles, la numérisation de la réception de l'information soulève de nouveaux défis méthodologiques pour les approches ethnographiques (Markham, 2013). Avec l'essor des réseaux sociaux, des plateformes de *streaming* en ligne et des technologies de communication virtuelle, les pratiques de réception ont évolué de manière significative Ces pratiques se déploient désormais dans des contextes éloignés des cadres traditionnels de réception médiatique, rendant l'observation *in situ* plus complexe. En effet, les individus ne consomment plus seulement des informations de manière linéaire dans des contextes spatio-temporels bien circonscrits, mais naviguent à travers des écosystèmes informationnels hybrides (Chadwick, 2017). Ils consultent leurs fils d'actualité de manière sporadique, souvent via leur smartphone, entre deux stations de métro ou en attendant l'arrivée de leur bus. Comme le soulignent Amandine Kervella et Marlène Loicq (2015), cette consommation est « nomade, individuelle, désynchronisée ». Comment enquêter dès lors sur ces pratiques informationnelles multi-situationnelles, fragmentées et discontinues ? Les approches multi-méthodes offrent de nouveau des pistes intéressantes pour appréhender cette complexité. Claire Balleys, par exemple, a récemment mené une ethnographie multi-site sur les cultures juvéniles urbaines en Suisse, qui pourrait inspirer une enquête sur les pratiques informationnelles contemporaines. [87]

La troisième raison tient à la pandémie de Covid-19 et aux différentes mesures de distanciation sociale instaurées pour enrayer la propagation du virus. Au tout début de notre thèse, nous avons été en contact avec une muséographe, chargée du développement du projet de l'Exposition itinérante « Esprit Critique » coproduite par Universcience à Paris, Cap sciences à Bordeaux et le Quai des savoirs à Toulouse.[88] Intéressée par le dispositif d'enquête élaboré dans le cadre de notre mémoire de fin d'études, cette muséographe nous a proposé de l'adapter pour qu'il devienne l'une des activités proposées aux visiteurs de l'exposition. Le dispositif d'enquête que nous avions conçu à l'époque reposait sur un plateau de mise en situation permettant de mener des entretiens semi-directifs à travers une expérience se déroulant sous la forme d'un jeu et visant à enquêter sur les situations d'interactions (en ligne comme hors ligne) favorisant (ou non) le partage de différents types d'informations. Pour l'exposition, les membres d'Univers Science ont souhaité transposer notre dispositif sous la forme d'une table de jeu équipée d'un écran et complétée par un bracelet connecté permettant de mesurer certaines interactions en temps réel.

Après de nombreux échanges avec la muséographe, nous avions prévu de mener des observations participantes auprès des visiteurs de l'exposition, puis de compléter ces observations par des entretiens afin de mettre en perspective leurs impressions et réflexions sur l'expérience avec leurs pratiques quotidiennes de partage d'information. Le bracelet connecté aurait par ailleurs permis d'accéder aux données des visiteurs et d'analyser leurs pratiques de partage d'informations pendant le jeu. En raison de la pandémie, néanmoins, l’ouverture des expositions a pris beaucoup de retard. En novembre 2021, nous avons pu tester un prototype du jeu à Bordeaux, mais celui-ci n’était pas encore accessible au public en raison de divers bugs. En parallèle, nous avons eu l’occasion d'expérimenter une variante du dispositif avec des élèves de 4ème. Une visite de l’exposition a également été organisée à Toulouse en juin 2022, mais l’activité restait encore indisponible pour le grand public. Étant donné que nos enquêtes principales étaient déjà bien avancées, nous avons finalement décidé de ne pas inclure ce projet dans notre thèse, tout en maintenant un contact ponctuel avec la muséographe pour d’éventuelles futures enquêtes.

[87] Balleys, C. (2023, 7 novembre). « C'est gênant ». Les normes de la pudeur juvénile dans les espaces urbains et numériques [Présentation de séminaire]. Séminaire de recherche, Sciences Po, Paris, France.
[88] https://www.cap-sciences.net/au-programme/exposition/esprit-critique.html

### 2.2.2. Apports et limites des méthodes numériques

Alors que les méthodes ethnographiques traditionnelles et les entretiens risquent de donner la parole à une « fiction de public à propos d’une fiction de texte, en inventant une relation fictive entre les deux » (Dayan, 1992, p. 150), les traces numériques laissées par les utilisateurs sur les réseaux sociaux permettent de prime abord aux chercheurs d’accéder à des réactions spontanées non sollicitées par leur dispositif d’enquête. Toutefois, bien que les méthodes de *big data* offrent de nouvelles opportunités pour tirer parti de leur volume massif (Lazer, 2009), elles n’en font pas moins l’objet d’importants débats épistémologiques et soulèvent de nombreux enjeux méthodologiques (boyd et Crawford, 2012 ; Kitchin, 2014 ; Tufekci, 2014 ; Ollion et Boelaert, 2015 ; Boyadjian, 2017).

Un premier écueil à éviter est de croire que les données parlent d'elles-mêmes et que les catégories créées par les plateformes permettent de saisir les usages et les pratiques numériques des internautes sans prendre le temps de les rencontrer et de mener un minutieux travail d’interprétation (Bastard et al., 2013). En réalité, comme le note Alice Marwick (2014, p. 109) : « les inférences faites à partir des propriétés de larges ensembles de données sont limitées dans ce qu'elles peuvent expliquer [...]. Les méthodes qualitatives, telles que les entretiens, les observations ethnographiques et l'analyse de contenu, fournissent une source riche de données qui permet d'aller au-delà de la simple description » (notre traduction). Par exemple, ce n’est qu’après avoir réalisé des entretiens avec des utilisateurs de Twitter que Chris Bail et ses collègues (2021) ont pu comprendre et surtout expliquer les résultats contre-intuitifs issus de leur expérience de terrain (Bail et al., 2018). Ils ont mis en évidence que l’exposition à des contenus adverses ne conduit pas tant à une modération des opinions et attitudes qu’à leur radicalisation. Ce phénomène s’explique par un sentiment de menace pesant sur l’identité des individus, les incitant alors à se replier sur leurs positions initiales. Les chercheurs doivent donc se demander si les données collectées à partir des plateformes de médias sociaux mesurent réellement ce qu'ils cherchent à étudier. Les interactions en ligne (comme les likes ou les retweets) sont souvent prises comme des indicateurs de comportements ou d'attitudes, mais ces signaux peuvent être trompeurs. Par exemple, un retweet ne signifie pas nécessairement que l'utilisateur approuve le contenu partagé. La validité de telles interprétations nécessite donc des précautions méthodologiques.

Deuxièmement, les données scrapées sur les réseaux sociaux sont souvent « désincarnées » et « hors-sol » (Boyadjian, 2017), c’est-à-dire dépourvues de références socio-démographiques (âge, sexe, profession et catégorie socio-professionnelle, niveau de diplôme, origine sociale, etc.) permettant d’identifier avec précision la position des enquêtés dans l’espace social. Or, de nombreuses enquêtes contemporaines, réalisées à partir des méthodes classiques de sciences sociales - questionnaires, entretiens, etc. - continuent de montrer le poids du capital éducatif dans la distribution des pratiques numériques, et notamment les modalités de consommation des biens culturels ou informationnels en ligne (Selwyn, 2004 ; Boyadjian, 2016 ; Pasquier, 2018 ; Seux, 2018).

Enfin, il existe des biais dans la représentativité de l’échantillon. Si l’on prend Twitter, par exemple, il ne représente pas « le grand public ». Nous ne pouvons pas non plus supposer une équivalence systématique entre compte et utilisateur. Certains utilisateurs ont plusieurs comptes, tandis que certains comptes sont utilisés par plusieurs personnes. Certaines personnes ne vont jamais créer de compte, et simplement accéder à Twitter via le Web. Certains comptes sont des « robots » qui produisent du contenu automatisé sans impliquer directement une personne. De plus, la notion de compte « actif » est problématique. Alors que certains utilisateurs vont publier du contenu fréquemment sur Twitter, d'autres participent comme « auditeurs » (Crawford 2009, p. 532). Une autre dimension importante à prendre en compte est que les ensembles de données massifs tendent à exclure les individus disposant de moins de ressources, qu'elles soient financières, techniques ou d'un autre ordre. Ainsi, face à chaque grand ensemble de données, il est essentiel de se demander : quelles personnes en sont exclues ? Quels groupes sont sous-représentés ? Il est donc primordial de comprendre les limites et les spécificités d'un ensemble de données, quelle que soit sa taille. Un volume de plusieurs millions de données ne garantit ni une sélection aléatoire ni une représentativité. Toute analyse statistique fondée sur un ensemble de données nécessite non seulement de connaître son origine, mais aussi d'en appréhender les biais et les limites.

### 2.2.3. L'intérêt d'une approche hybride et combinatoire

Afin d'ajuster les méthodes classiques de sociologie de la réception aux nouvelles situations de réception d'informations engendrées par l'essor des réseaux sociaux, tout en tirant parti des possibilités offertes par les approches de *big data* sans tomber dans leurs écueils, nous avons fait le choix de développer un dispositif d'enquête hybride et combinatoire, inspiré d'une série de travaux ayant démontré la fécondité des enquêtes empiriques multi-méthodes et à entrées multiples (Balleys, 2011 ; Boyadjian, 2014 ; 2016 ; Pasquier, 2018 ; Le Caroff et Foulot, 2019 ; Bail, 2021).

Plutôt que de parler de méthodes mixtes – ou plus spécifiquement d'une approche « quali-quanti » (Venturini et al., 2014) – il nous semble plus pertinent de parler de dispositif d'enquête hybride et combinatoire. En effet, de la même manière que l'apprentissage de la lecture ne se fait pas en saisissant un mot dans sa globalité mais syllabe par syllabe, chaque méthode séparément ne permet pas de saisir la réalité dans son ensemble. C'est en combinant chacune de ces méthodes que l'on obtient une vision plus complète et plus juste. Ainsi, tout comme la méthode combinatoire en lecture permet de déchiffrer et de comprendre un texte en assemblant différentes unités, notre approche méthodologique vise à saisir la complexité des comportements et des pratiques informationnelles en combinant différentes techniques de collecte et d'analyse des données.

Le cadre méthodologique mis en place a ainsi impliqué des allers et retours nombreux entre (1) des traitements massifs de données (*distant reading*), (2) des observations manuelles de corpus (*close reading*) et (3) des entretiens recueillis auprès des acteurs. Des précisions et réflexions sur la mise en œuvre de chacune de ces approches seront données dans la prochaine section, ainsi qu'au fil des chapitres de cette thèse. L'enjeu pour l'instant est simplement d'offrir une vue d'ensemble des méthodes mobilisées.

Une fois les terrains d'enquête délimités et les données collectées sur les réseaux sociaux (*infra*), des méthodes numériques ont dans un premier temps été mobilisées pour explorer, visualiser et analyser les données. Par exemple, des analyses de réseaux et de co-occurrences ont été réalisées avec des outils comme *Gephi* ou *Cortext* (Breucker et al., 2016) afin de repérer les mots les plus présents dans un corpus ou d'identifier des liens entre différents types de données. Des premières analyses statistiques descriptives ont également été conduites à partir des métadonnées extraites des plateformes (e.g. nombre de *followers* ; nombre de tweets ; nombre de *likes*) offrant un aperçu des pratiques des utilisateurs et de leur visibilité en ligne ou du succès de certains contenus.

Ensuite, des observations en ligne ont été menées sur des profils d'utilisateur et des espaces de discussion publiquement accessibles sur les réseaux sociaux en mobilisant des approches d'ethnographie virtuelle (Hine, 2000) ou de netographie (Kozinets, 2019), reposant sur une immersion prolongée du chercheur dans les espaces étudiés (Jouët et Le Caroff, 2013). Une approche passive (Loveluck, 2016), parfois qualifiée de « deep lurking » (Lee et al., 2021), c'est-à-dire « sans implication dans le cours des interactions » a été privilégiée. En effet, à la différence d'étude comme celle conduite par la chercheuse Gabriella Coleman (2016) au sein du collectif *Anonymous*, l'objectif de cette thèse n'était pas de suivre au jour le jour les pratiques d'un groupe social spécifique, mais d'analyser de façon rétrospective des réactions spontanées et individuelles eues par des utilisateurs *a priori* non coordonnés les uns avec les autres.

Si les réseaux sociaux offrent un terrain d'enquête privilégié pour accéder facilement aux pratiques des utilisateurs, les méthodes d'observation en ligne comportent de nombreuses limites et font l'objet d'importantes critiques (Pastinelli, 2011). Contrairement au long processus d'intégration traditionnellement nécessaire pour être accepté dans un groupe social, l'ethnographie en ligne permet une accessibilité immédiate aux données. Cependant, cette transparence apparente est souvent trompeuse : elle réduit les individus à leurs traces numériques sans révéler le contexte ni les motivations qui sous-tendent la production de ces traces ; ce qui risque d'offrir une vision partielle et déformée de leur comportement (Bail, 2021). De plus, les caractéristiques sociodémographiques des utilisateurs restent souvent inaccessibles aux chercheurs, d'autant plus que ces informations sont souvent masquées sous des pseudonymes. Ainsi, le champ d'observation des ethnographies en ligne est souvent beaucoup plus limité que celui des ethnographies classiques ; celles-ci se restreignant aux usages visibles sans permettre de saisir les dimensions cachées et les intentions des pratiques sociales dans les espaces de communication virtuelle.

Afin de dépasser les limites inhérentes aux analyses quantitatives de traces numériques et aux observations en ligne, nous avons cherché à entrer en contact avec des utilisateurs des réseaux sociaux faisant partie de nos échantillons d'enquête (*infra*) pour réaliser des entretiens avec eux. L'enjeu principal était d'éviter de tirer des conclusions sur leurs pratiques informationnelles et conversationnelles uniquement à partir de leur comportement en ligne, ainsi que de les mettre en perspective avec leur parcours biographique et leur contexte social. Cette approche permet de confronter des données déclaratives à des données comportementales et d'identifier d'éventuels écarts ou contradictions entre les deux (Marwick, 2014). Par exemple, l'étude de Coralie Le Caroff et Mathieu Foulot (2019) montre que les propos des utilisateurs qui adhèrent à des théories conspirationnistes sont bien plus nuancés hors ligne que sur Facebook. De même, Chris Bail (2021) a montré que les dynamiques d'expression peuvent varier considérablement selon les contextes en ligne et hors ligne. Si de prime abord ces écarts et contradictions peuvent donner l'impression de résultats incohérents et difficilement interprétables, le fait d'adopter un cadre théorique pragmatique invite à voir dans ces contradictions une capacité de la part des personnes à adopter des identités multiples et à ajuster constamment leurs pratiques aux contextes d'interactions.

Enfin, un travail d'annotation important a été réalisé à partir des données récoltées (e.g. les textes des bios des profils Twitter ; les noms d'utilisateurs ; les noms des pages et des groupes Facebook ; les textes de commentaires Facebook ; etc.). Les observations réalisées sur différents comptes et espaces de discussion, ainsi que les lectures exploratoires des différents corpus récoltés, ont permis de faire émerger des variables qualitatives et d'élaborer des grilles d'annotations. Plusieurs types de catégories ont ensuite été attribués à divers éléments constitutifs de nos bases de données. Quand les corpus étudiés étaient relativement modestes ou que les modalités à attribuer étaient multiples et complexes, l'ensemble des données a été annoté, souvent grâce à l'aide précieuse d'étudiantes effectuant un stage d'initiation à la recherche dans le cadre de leur troisième année d'étude.[89] Ces annotations ont fait l'objet de discussions collectives pour assurer un bon accord inter-annotateurs. Pour les corpus plus volumineux, 10% des données ont été annotées. Ce travail d'annotation a ensuite servi de base pour entraîner un algorithme. Nous avons notamment utilisé le modèle de langage pré-entraîné BERT (Delvin et al., 2019) dans sa version francophone CamemBERT (Martin et al., 2019).

Ces dernières années, de nombreux chercheurs et chercheuses en sciences sociales se sont tournés vers l'apprentissage automatique supervisé pour analyser de vastes corpus de textes. Cette procédure implique d'étiqueter un échantillon relativement modeste de données pour « entraîner » un algorithme. Avant d'être entraîné à classifier les traces textuelles issues des corpus annotés, le réseau de neurones a été entraîné sur une tâche de texte à trous, la *Cloze Task*, qui consiste à masquer aléatoirement une faible proportion de mots au sein d'une phrase, puis à faire inférer ces mots au modèle à partir du contexte connu du modèle. Cette tâche de pré-entraînement permet au modèle d'acquérir à la fois des « connaissances » syntaxiques et grammaticales (le modèle doit inférer la fonction, conjugaison éventuelle, etc. du mot), sémantiques (le modèle apprend à reconnaître que deux mots synonymes peuvent s’employer dans le même contexte), mais aussi « culturelles », lorsque les mots masqués sont des personnes, des dates ou des lieux, par exemple. En apprenant à réaliser cette tâche sur des corpus très volumineux, comme le *Common Crawl* (i.e. un ensemble des pages web indexées), le modèle pré-entraîné obtient des performances au niveau de l’état de l’art lorsqu’il est ensuite adapté à d’autres tâches, comme la classification de texte, la traduction automatique, la réponse automatique à des questions, etc.

[89] En raison de la pandémie de Covid-19, de nombreux étudiants du Collège Universitaire de Sciences Po n'ont pas pu partir à l'étranger pour leur troisième année. À la place, des stages d'initiation à la recherche, d'une durée de 10 semaines à raison de 12 heures par semaine, leur ont été proposés. Quatre étudiantes ont ainsi été accueillies au médialab pendant les semestres d'automne 2021 et de printemps 2022 pour travailler sur un projet de modération de contenus, financé par le *Project Liberty's Institute*. Ces stages ont été crédités de 10 ECTS et co-encadrés avec Valentine Crosset et Dominique Cardon.

Une fois que le classifieur est jugé suffisamment performant (« phase de test »), il est utilisé pour annoter un ensemble de données bien plus vaste (« phase de prédiction »). Des recherches récentes, telles que celles de Barberá et al. (2021) et de King, Pan et Roberts (2013), ont démontré l’efficacité de cette approche. Laura Nelson et ses collègues (2021) ont également apporté des avancées significatives dans l’étiquetage automatique de textes appliqué à la sociologie, en montrant que les annotations manuelles pouvaient, dans certains cas, être remplacées par un modèle d’apprentissage supervisé. Cette approche fonctionne particulièrement bien pour des tâches relativement simples. Toutefois, de nombreuses tâches restent difficiles pour les machines, notamment l’identification de motifs subtils ou implicites, ainsi que la reconnaissance précise des sujets et objets au sein des phrases. Une récente étude a cependant montré qu'un modèle d'apprentissage automatique pouvait obtenir des résultats presque équivalents à ceux d’assistants de recherche et meilleurs que ceux de micro-travailleurs (Do et al., 2022). Dans le cadre de cette thèse, des méthodes d’apprentissage supervisé ont ainsi été mobilisées dans certains cas pour faciliter le travail d’annotation des données.

Une fois le modèle entraîné et les tâches de prédictions réalisées, un score F1 est calculé pour évaluer la performance d'un modèle de classification, et vérifier qu’il y un équilibre entre précision et rappel. Pour expliquer le sens de ces notions, illustrons par un cas concret. Imaginons une tâche d’annotation consistant à classer des noms d’utilisateurs comme féminin ou masculin. La notion de précision désigne alors la proportion de véritables noms masculins parmi ceux identifiés par le modèle. Si le modèle identifie 100 noms masculins et que 80 sont corrects (d’après les annotations des chercheurs), la précision est de 80 %. La notion de rappel désigne la proportion de noms masculin correctement identifiés parmi tous les noms masculins existant dans le corpus. Si, parmi 100 noms masculins présents dans le corpus, le modèle en identifie 80, le rappel est de 80 %. Le score F1 est la moyenne harmonique de ces deux mesures et permet de vérifier si le modèle est très précis mais manque beaucoup de noms masculins (faible rappel), ou s'il identifie beaucoup de noms masculins (haut rappel) mais avec de nombreux faux positifs (faible précision). Cela permet de juger de manière équilibrée la qualité d'un modèle, en s'assurant qu'il ne se concentre pas uniquement sur la précision ou le rappel.

En résumé, cette approche hybride et combinatoire, articulant statistiques descriptives, observations en ligne et entretiens compréhensifs permet d'examiner des pratiques informationnelles, des prises de paroles et des usages des réseaux sociaux non générées ni altérées par nos dispositifs d'enquête, tout en parvenant à les associer à des propriétés socio-démographiques.

## 2.3. Terrains d'enquête

Afin de délimiter des terrains d'enquête adaptés à notre objet d'étude, ainsi qu'à notre cadre théorique et méthodologique, trois étapes ont été nécessaires. Tout d'abord, nous avons eu besoin de constituer des corpus de *fake news* – et donc de retenir une définition opérationnelle du terme. Ensuite, nous avons cherché à identifier les utilisateurs des réseaux sociaux ayant été exposés à ces contenus sans « [inventer] une relation fictive entre les deux » (Dayan, 1992, p. 150). Enfin, nous avons essayé de négliger le moins possible les contextes sociaux de ces utilisateurs, tout en prêtant attention à la diversité des situations d'interactions (en ligne comme hors ligne) dans lesquelles ils ont été amenés à recevoir ou à partager des informations.

### 2.3.1. Constituer des corpus de *fake news*

Comme cela a été expliqué en introduction (cf. p. 35-36), nous avons fait le choix de définir le terme *fake news* en reposant sur deux types d'évaluation de la qualité d'une information : d'une part, sur les vérifications factuelles produites par des journalistes français ; de l'autre, sur les signalements réalisés par des utilisateurs de Facebook. Cette double définition permet d'inscrire notre thèse dans la continuité de la majorité des études académiques sur les *fake news* tout en proposant une nouvelle approche pour dépasser certaines de leurs limites.

***Des fact-checks des journalistes…***

Dans un premier temps, nous sommes partie de deux bases de données contenant chacune une liste d'URLs ayant été classées comme des *fake news* par des journalistes spécialisés dans la vérification factuelle *:* (1) la base de données des *Décodeurs* en accès libre et mise à jour en temps réel sur le site du *Monde*[90] ; (2) la base de données *Condor* de Facebook[91] mise à disposition de plusieurs équipes de recherche par le consortium *Social Science One* (Messing et al., 2020) donnant accès entre autres aux évaluations produites par les *third-party fact-checkers* francophones collaborant avec Facebook, à savoir : l'*AFP Factuel* ; les *Décodeurs* du *Monde* ; les *Observateurs* de *France 24* ; et *Fake off* de *20 Minutes*. Afin de pouvoir situer chaque source dans l'espace médiatique français, nous avons filtré ces listes d'URLs pour ne garder que celles issues de 420 médias identifiés par une étude du médialab de Sciences Po (voir Cardon et al., 2019 et Cointet et al., 2021 pour la méthodologie détaillée). Au moment de lancer notre collecte de données en automne 2020 la base de données des *Décodeurs* contenait 1 080 URLs étiquetées comme fausses et celle de Facebook 468, portant le total de notre liste d'URLs fact-checkées à 1 548.

***… aux signalements des utilisateurs des réseaux sociaux***

Dans un second temps, nous avons de nouveau mobilisé la base de données *Condor* de Facebook. Cette fois, nous avons restreint notre sélection aux 1 163 URLs qui ont été signalées au moins 30 fois[92] comme des fausses informations par des utilisateurs du réseau social. Nous

---

[90] Les Décodeurs. (2017, 19 décembre). Fausses informations : les données du Décodex en 2017. *Le Monde*. https://www.lemonde.fr/le-blog-du-decodex/article/2017/12/19/fausses-informations-les-donnees-du-decodex-en-2017_5231605_5095029.html

[91] Cette base de données *Condor* contient 38 millions d'URLs qui ont été partagées au moins 100 fois publiquement sur Facebook entre le 1er janvier 2017 et le 31 juillet 2019.

[92] Nous avons choisi ce seuil de 30 car les données de Facebook sont bruitées pour éviter que l'on puisse retrouver l'identité des utilisateurs. Ce bruit est un ajout ou une soustraction aléatoire au nombre réel de signalements obtenus par une URL ; il varie entre +10 et -10. Nous avons choisi de ne conserver que les URLs au-dessus d'un seuil de 30 signalements afin d'être

avons passé en revue cette liste afin d'écarter les URLs incomplètes, indisponibles, non francophones ou ne renvoyant pas vers une information mais vers un autre type de contenu (publicité, jeux, concours, etc.). Après cette phase de nettoyage, nous avons conservé 904 URLs signalées. Pour rappel, ces URLs ne renvoient pas forcément vers des contenus factuellement faux (au sens des *fact-checkers*) mais vers des contenus signalés comme « faux » par des utilisateurs de Facebook. Or, les boutons de signalement sur les réseaux sociaux peuvent être utilisés à des fins stratégiques plutôt que pour produire des évaluations épistémiques sincères (Crawford et Gillespie, 2016 ; Andersen et Søe, 2020). Il faut donc considérer les contenus des URLs signalées comme des énoncés dont la qualité épistémique est incertaine, susceptibles de faire l'objet de labellisations contradictoires par différentes personnes, et permettant ainsi aux chercheurs d'accéder à des situations qui sont potentiellement perçues comme problématiques par les utilisateurs des réseaux-sociaux eux-mêmes.

Avant de détailler la procédure utilisée pour identifier les utilisateurs ayant été exposés à ces contenus sur les réseaux sociaux, prenons un moment pour examiner concrètement ce que les *fact-checkers* et les utilisateurs de Facebook mettent derrière le terme *fake news*. Quels sont les médias à l'origine de ces contenus ? Sur quels sujets et thématiques portent-ils ? Observe-t-on des différences entre les choix des *fact-checkers* et ceux des utilisateurs de Facebook ? Répondre à ces questions permettra de cerner plus pratiquement les énoncés que recouvrent l'imprécise notion de *fake news* et d'explorer si les jugements des utilisateurs des réseaux sociaux convergent ou divergent avec ceux des *fact-checkers*.

Pour identifier les sources médiatiques à l'origine de la diffusion de *fake news*, nous avons utilisé la classification réalisée par une étude du médialab de Sciences Po sur la structure de l'écosystème médiatique français (Cardon et al., 2019 ; Cointet et al., 2021). L'étude a mesuré l'autorité de 420 sites médias français en examinant le réseau d'hyperliens qui les relient les uns aux autres. Sur la base du crédit que les sites de médias s'accordent entre eux, quatre blocs ont été distingués : (1) les médias grand public regroupant les principaux médias français de presse, radio et télévision, qui ont la plus grande audience dans l'espace médiatique sûre qu'elles aient été signalées au moins une fois et de minimiser la taille de notre corpus (avec un seuil de 20 par exemple, on se retrouvait avec des millions d'URLs).

français (e.g. *Le Monde*, *Le Figaro*, *TF1*, *BFM TV*) ; (2) les médias engagés ou partisans comportant des sites qui défendent une ligne politique de droite ou de gauche (e.g. *Valeurs Actuelles*, *Reporterre*) ; (3) les médias de contre-information qui ne reçoivent pratiquement aucun lien du reste de l'écosystème médiatique français alors qu'ils envoient beaucoup de citations à l'ensemble de l'écosystème (e.g. *FdeSouche*, *Stop Mensonges*, *Alternative Santé*) ; et enfin (4) les médias périphériques contenant différents types de médias spécialisés allant de la presse locale à la presse technologique en passant par les magazines de loisir et de divertissement (e.g. *Nice Matin*, *Télé Loisirs*).

Cette classification montre que la très grande majorité (72,5 %) des contenus évalués comme des *fake news* par des *fact-checkers* a été diffusée par des médias de contre-information (cf. Figure 2.1). En tête de liste, on retrouve notamment le média *La Gauche m'a tuer* (135 URLs), suivi par *Les moutons rebelles* (83 URLs), *Wikistrike* (55 URLs) et *Stopmensonge* (54 URLs). Bien connus des organes de vérification, la majorité des médias de contre-information se caractérisent par une ligne éditoriale très critique à l'égard des médias *mainstream* comme l'attestent les bannières de certains sites web : « Les moutons en ont marre, ils s'informent » (*Les moutons enragés*), « Votre bouffée d'air d'information du matin pour supporter les mensonges des médias » (*Dreuz.info*), « Articles à contre-courant de la doxa médiatique et politique qui impose une information « prête-à-penser » (*Adoxa*), « La publication anti bourrage de crâne » (*Les4Vérités*). Un volume non négligeable a aussi été diffusé par des magazines de loisir situés dans la catégorie des médias périphériques, notamment par *Santé + Mag* (53 URLs) ou *Gentside* (39 URLs).

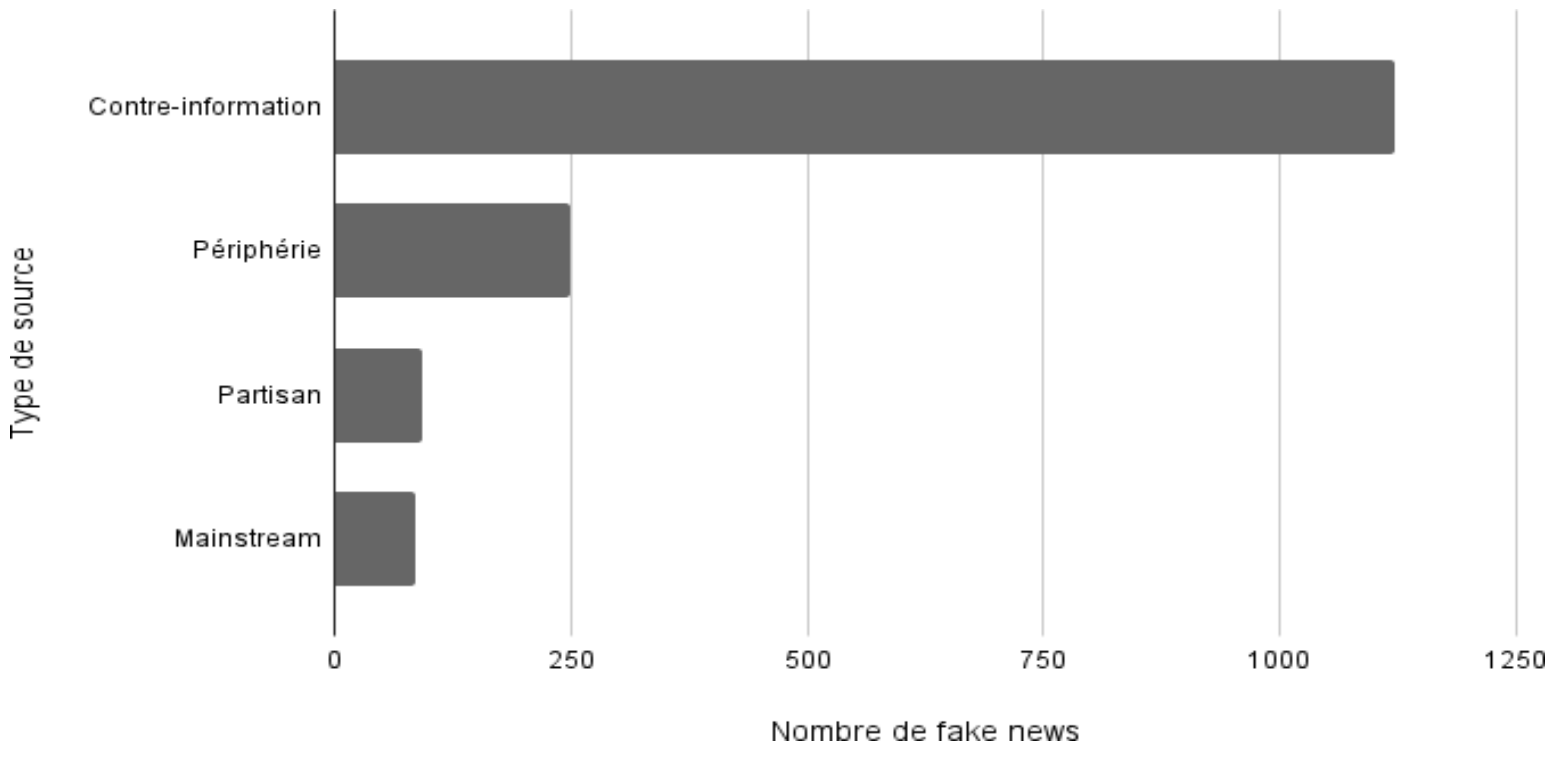

*Figure 2.1. Types de médias dont sont issus les contenus évalués comme des fake news par des fact-checkers*

L'analyse des sources des URLs signalées fait quant à elle apparaître une distribution plus homogène parmi les différentes catégories de médias (cf. Figure 2.2). À la grande différence du classement issu des évaluations des *fact-checkers*, les médias *mainstream* sont en tête des signalements (36,5 %). Viennent ensuite les médias périphériques (29,5 %), notamment spécialisés dans les loisirs ou la santé, et les médias de contre-information (26,2 %) ; et enfin, loin derrière, les médias partisans (7,7 %).

Ces différences entre les *fact-checks* et les signalements ne signifient pas que la hiérarchie des sources issue de la structure de l'espace médiatique n'a pas été intériorisée par les publics. De nombreuses études ont d'ailleurs mis en évidence une corrélation importante entre les évaluations de sources du grand public et des *fact-checkers* (Pennycook et Rand, 2019b ; Allen et al., 2021b ; Roitero et al., 2020). En fait, les nombreux signalements reçus par les médias *mainstream* peuvent s'expliquer par leur audience plus large que celle des médias de contre-information, ce qui les rend plus susceptibles d'être vus et donc potentiellement signalés par des utilisateurs sur les réseaux sociaux. En effet, à partir des constats résumés dans l'introduction, il est possible de faire l'hypothèse que c'est précisément parce que les internautes sont exposés à peu de sites de *fake news* mais davantage aux médias *mainstream* qu'ils sont plus susceptibles de signaler ces derniers. Cela ne signifie pas qu'ils les jugent moins fiables mais que leur propension à émettre des signalements concerne davantage ces médias car ceux-ci occupent une place plus grande dans leurs habitudes de consommation médiatique. Afin d'estimer plus précisément quels types de sources sont considérés comme fiables ou non fiables par les utilisateurs des réseaux sociaux, le chapitre 5 s'attachera à évaluer la proportion de commentaires critiques suscités par ces différents médias sur leur volume total de commentaires.

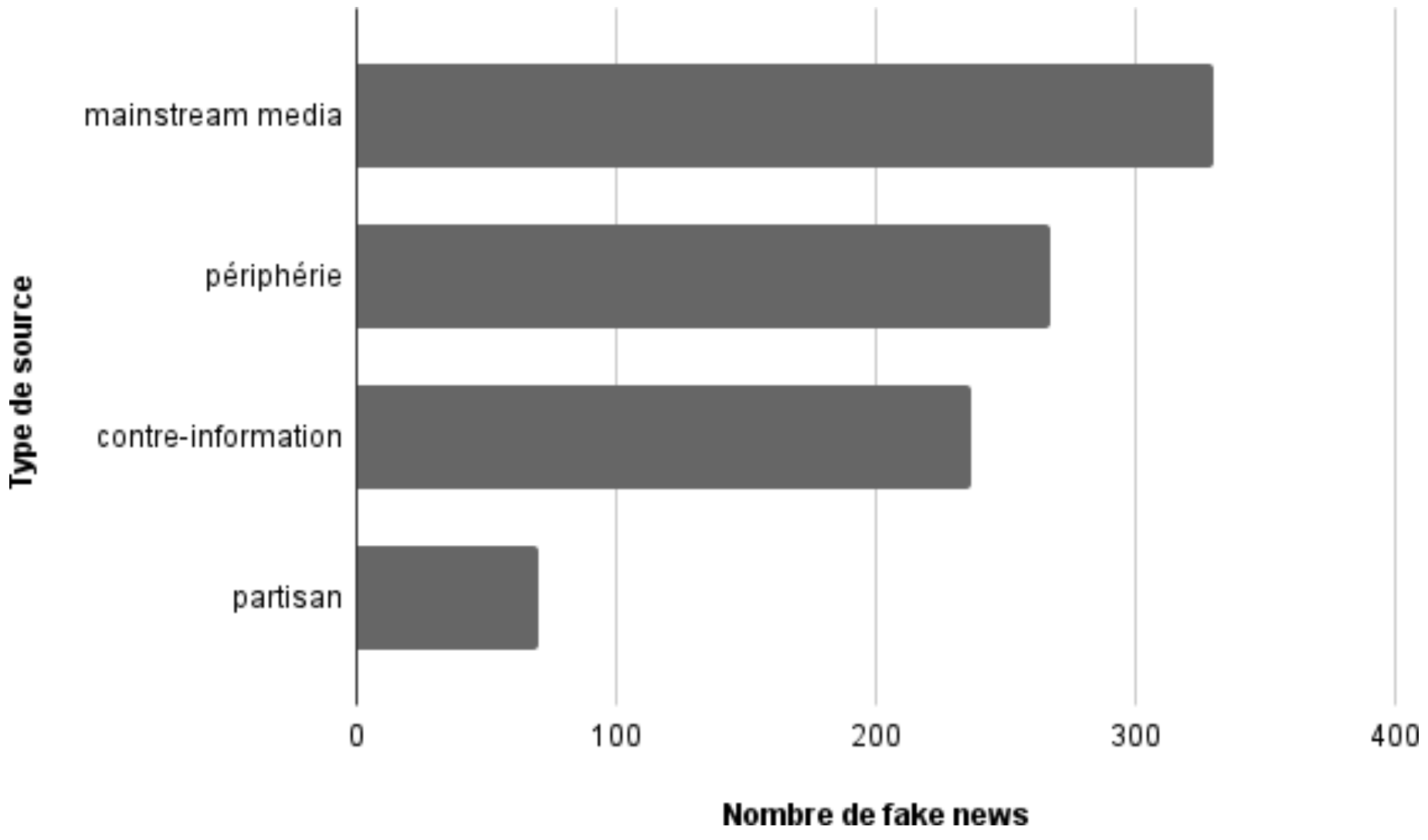

*Figure 2.2. Types de médias dont sont issus les contenus signalés comme des fake news par des utilisateurs*

Alors que les *fact-checks* des journalistes et les signalements des utilisateurs ne portent pas tout à fait sur les mêmes types de sources médiatiques, l'on peut se demander dans quelle mesure les évaluations de ces deux groupes concernent les mêmes sujets et thématiques. Afin de répondre à cette question nous avons réalisé un réseau de co-occurences, permettant de cartographier la fréquence avec laquelle des paires de mots apparaissent ensemble dans un même article et ainsi de faire émerger des communautés de mots révélant des thèmes ou des sujets récurrents dans le corpus. Pour produire ce réseau, nous avons collecté les contenus de l'ensemble des articles signalés et fact-checkés avec l'outil de *web mining* Minet (Plique et al., 2019). Au total, 1 915 articles ont été obtenus et 1 830 articles ont été conservés après nettoyage. Ensuite, nous avons extrait les 3 000 *n-grams* (séquences de mots) et les 3 000 *monograms* (mots individuels) les plus fréquents grâce à l'outil *Cortext* (Breucker et al., 2016). Après nettoyage et suppression des doublons, nous avons construit un réseau de co-occurrences comprenant 1 500 termes.

La carte ci-dessous (cf. Figure 2.3) présente les co-occurrences de termes dans les articles fact-checkés et signalés de notre corpus. La proximité entre deux termes indique qu'ils apparaissent ensemble dans les mêmes articles. Après l'analyse de cette carte, nous avons ajouté manuellement des étiquettes permettant de rendre compte du thème principal de chaque cluster. Des analyses plus fines de contenus seront réalisées dans les chapitres 3 et 5 pour préciser les sujets abordés dans les contenus des *fake news* de notre corpus.

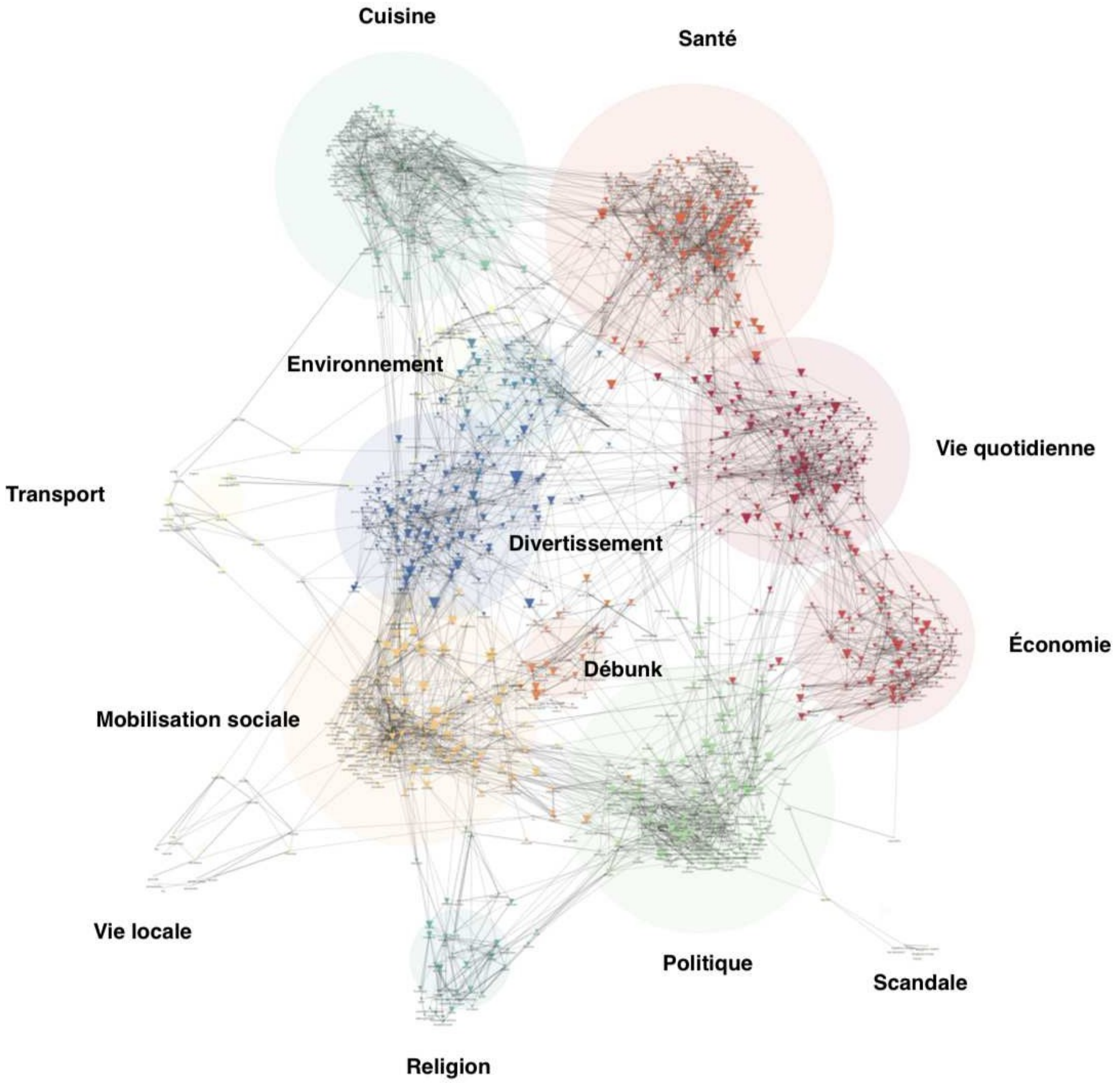

*Figure 2.3. Cartographie des thématiques des contenus identifiés comme des fake news par des fact-checkers ou signalés comme faux par des utilisateurs de Facebook*

Afin de comparer les signalements des utilisateurs aux *fact-checks* des journalistes, des cartes de chaleur ont été réalisées grâce à l'outil *Cortext* (cf. Figure 2.4). Celles-ci montrent une concentration des *fact-checks* au sein des clusters relatifs à des sujets portant sur la religion, la politique, l'environnement et la santé. En ce qui les concerne, les signalements des utilisateurs sont un peu plus dispersés. Bien qu'ils soient également présents de manière significative dans les clusters portant sur la religion et la politique, ils couvrent également des thématiques plus variées comme les divertissements, les mobilisations sociales, la vie locale et l'économie. Ces constats suggèrent que, tandis que les journalistes se concentrent sur des sujets fréquemment discutés dans le débat public, les utilisateurs interviennent sur une plus grande diversité de problématiques, parfois plus locales, personnelles ou anodines, reflétant un éventail plus large de préoccupations.

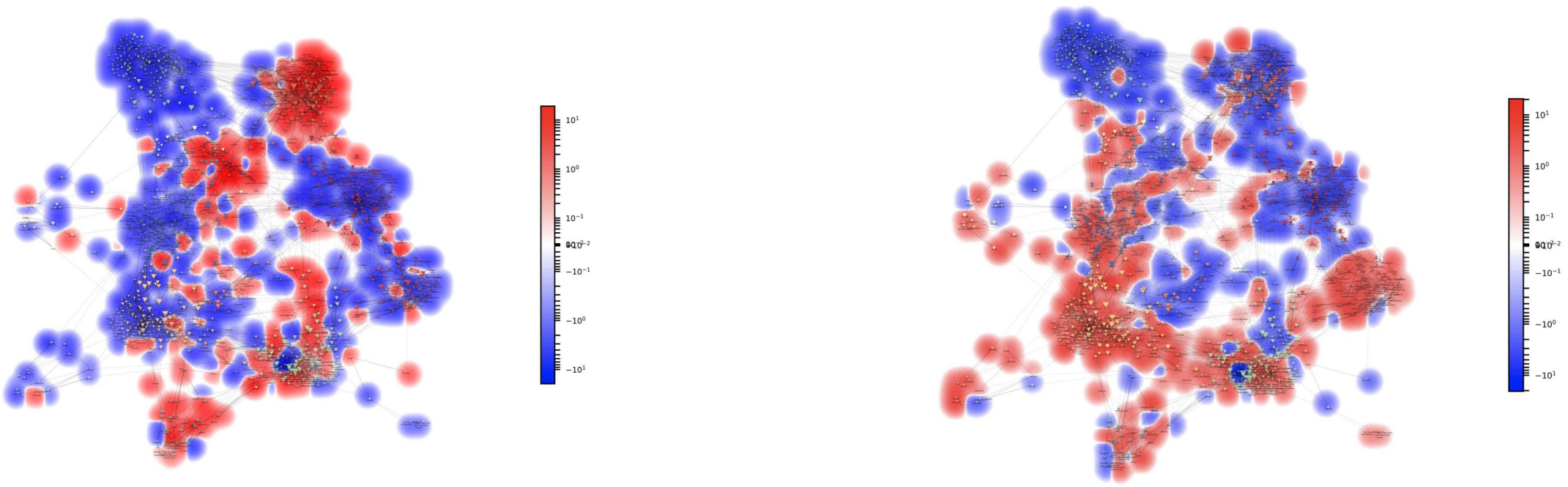

*Figure 2.4. Cartographie des zones qui ont reçu le plus de fact-checks (à gauche) ou de signalements (à droite)*

Il est important de souligner que les analyses réalisées ci-dessus visent moins à préciser la définition de la notion de *fake news* qu'à comprendre comment celle-ci est définie par différents acteurs. Autrement dit, elles ne permettent pas de conclure de façon rigoureuse que les contenus faux ont plutôt tendance à porter sur tel ou tel sujet ou à provenir de telle ou telle source, mais offrent l'opportunité d'identifier vers quels types de médias et de contenus les *fact-checkers* et les utilisateurs dirigent leur attention. Alors que l'identification de *fake news* par les *fact-checkers* dépend de normes professionnelles, la question se pose de savoir si les utilisateurs qui signalent des contenus sur les réseaux sociaux adoptent ces mêmes normes ou s'ils s'appuient sur d'autres critères et suivent d'autres principes – et dans ce cas lesquels ? Cette question sera abordée dans les chapitres 5 et 6.

### 2.3.2. Identifier des récepteurs de *fake news* et leurs réactions spontanées

Une fois nos corpus de *fake news* constitués, nous avons cherché à retrouver les utilisateurs des réseaux sociaux ayant réagi spontanément à ces contenus, c'est-à-dire sans que cela leur soit imposé par notre dispositif d'enquête.

À partir de la liste de 1 548 URLs fact-checkées, nous avons identifié 87 364 tweets contenant au moins une *fake news*, émis par 31 760 comptes, parmi un ensemble de données permettant de reconstituer la Twittosphère française. Le chapitre 3 détaillera la méthode utilisée pour reconstituer cette Twittosphère et évaluer la proportion d'utilisateurs qui relaie des *fake news* parmi cet ensemble. Il analysera également les types de *fake news* qui ont été les plus partagées afin de comprendre les thématiques et rhétoriques que cherchent à mettre en avant les utilisateurs qui les relaient.

À partir de la liste de 904 URLs signalées, nous avons collecté via l'outil de web mining *Minet* (Plique et al., 2019) l'ensemble des posts Facebook les ayant partagées (n = 30 157) ainsi que leurs commentaires associés (n = 441 149). Ceux-ci ont été émis par 294 988 utilisateurs. Les chapitres 5 et 6 présenteront le dispositif méthodologique mis en place pour évaluer la proportion d'utilisateurs ayant formulé des commentaires critiques face à ces contenus.

Si les traces numériques émises en réaction à des *fake news* sur les réseaux sociaux constituent le point d'entrée de nos deux terrains d'enquête, un ensemble plus vaste de données a été récolté au fil de nos analyses grâce aux différentes méthodes mobilisées afin de ne pas réduire les utilisateurs de nos échantillon au fait d'avoir réagi à une *fake news* sur un réseau social particulier et d'examiner plus largement la variété de leurs pratiques (informationnelles comme conversationnelles) au sein de différentes situations d'interactions (en ligne comme hors ligne), tout en identifiant certaines de leurs caractéristiques socio-démographiques.

### 2.3.3. Négliger le moins possible le contexte social et les situations d'énonciation

Cette section documente comment le dispositif méthodologique hybride et combinatoire adopté dans cette thèse, ainsi que les différentes données collectées, ont permis de négliger le moins possible les contextes sociaux des enquêtés et la variété des situations d'interactions au sein desquelles ils évoluent malgré le fait de ne pas avoir reposé sur une approche ethnographique classique.

Tout d'abord, les positions politiques d'un peu plus de 30 % des relayeurs de *fake news* identifiés sur Twitter ont pu être estimées en mobilisant des travaux de sciences sociales computationnelles réalisés au médialab de Sciences Po (Cointet et al., 2021 ; Ramaciotti Morales et al. 2021). La propension des comptes à être plus ou moins politisés a également pu être évaluée à partir d'analyses mobilisant des méthodes d'apprentissage supervisé réalisées sur les bios de 1 907 comptes de relayeurs de *fake news* par contraste avec celle d'un groupe de contrôle composé de 958 non-relayeurs de *fake news* ayant les mêmes positions politiques. Plus de précisions seront données dans le chapitre 3 pour expliquer la construction de ces différents échantillons et les méthodes mobilisées.

Ces analyses quantitatives ont ensuite été complétées par des observations en ligne en sélectionnant aléatoirement 150 comptes au sein de notre échantillon de 1 907 relayeurs de *fake news* et 150 au sein de notre échantillon de 958 non-relayeurs de *fake news.* Parmi les comptes sélectionnés de relayeurs de *fake news*, 11 ont été suspendus, 4 sont passés en privé et 8 étaient introuvables (probablement car leurs utilisateurs ont décidé de les supprimer). Parmi les comptes sélectionnés de non-relayeurs de *fake news*, 4 ont été suspendus, 5 sont passés en privé et 5 étaient introuvables. Le plus grand nombre de suspendus chez les relayeurs de *fake news* est intéressant à remarquer car il suggère que ces derniers sont plus susceptibles d'enfreindre les règles de modération de Twitter. Par ailleurs, seulement 3 comptes de relayeurs de *fake news* appartiennent à une organisation (e.g. média, collectif, association, entreprise) tandis que c'est le cas de 21 des comptes des autres utilisateurs. Ce constat sera approfondi dans le chapitre 4 mais d'ores et déjà il est possible de suggérer que les comptes qui s'expriment au nom d'une organisation sont moins susceptibles de partager des *fake news* que ceux qui s'expriment de façon individuelle. Au total, nos observations ont donc concerné 124 relayeurs de *fake news* et 115 non-relayeurs de *fake news*. Concrètement, nous nous sommes rendue sur les comptes Twitter de nos enquêtés et nous avons prêté attention à de nombreux détails (e.g. niveau de langage, diversité des sujets, éléments autobiographiques auto-déclarés, affiliation partisane, pratiques militantes). Nous nous sommes aidée de l'outil de recherche avancé de Twitter (pour plus d'informations sur l'utilisation de cet outil, voir l'encadré ci-dessous) afin de capturer plus facilement les caractéristiques socio-démographiques de nos enquêtés et pour analyser si leurs opinions politiques, leurs usages de Twitter et leurs modes d'expression avaient évolué depuis leur inscription sur le réseau social. Ces observations ont permis de réaliser des portraits détaillés des enquêtés sur un journal de terrain.

## Ethnographie en ligne : capturer les caractéristiques socio-démographiques d’utilisateurs grâce à l’outil de recherche avancée de Twitter

Accessible à tous les internautes, l’outil de recherche avancé de Twitter offre une gamme étendue d’options pour effectuer des recherches ciblées. Il permet de retrouver des tweets contenant des termes ou des hashtags spécifiques, émis par des comptes précis ou ayant suscité un minimum d’engagement, et même de limiter les recherches à des plages temporelles ou à des zones géographiques particulières.

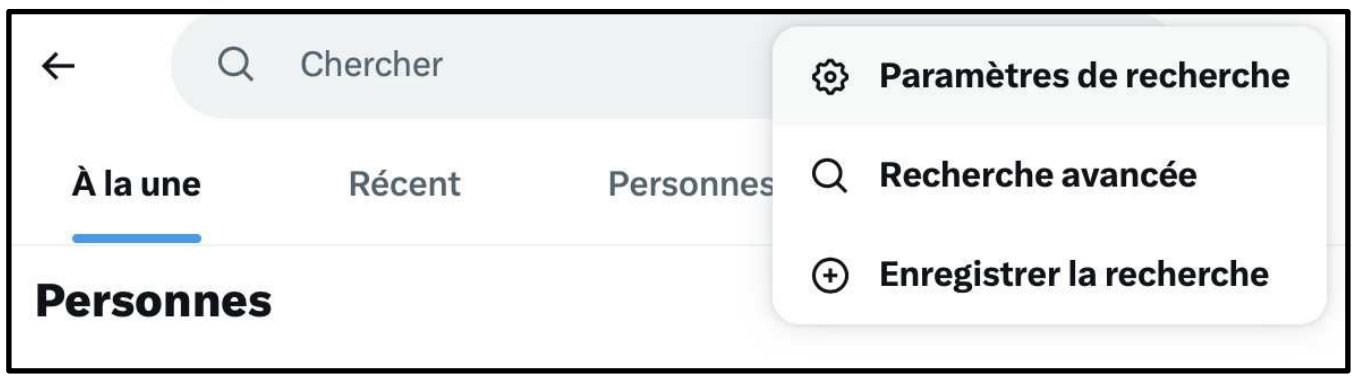

Pour capturer les caractéristiques socio-démographiques de comptes Twitter, les sections « Mots » et « Comptes » sont les plus utiles. Elles permettent de rechercher la présence de mots ou groupes de mots au sein des tweets publiés par chacun des comptes étudiés.

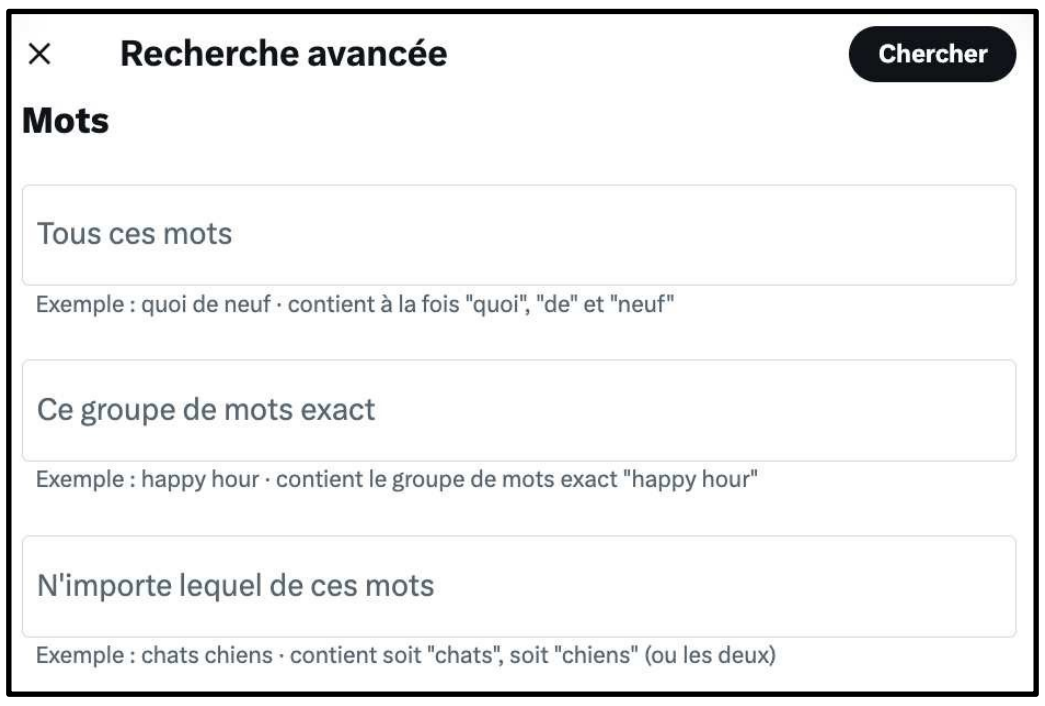

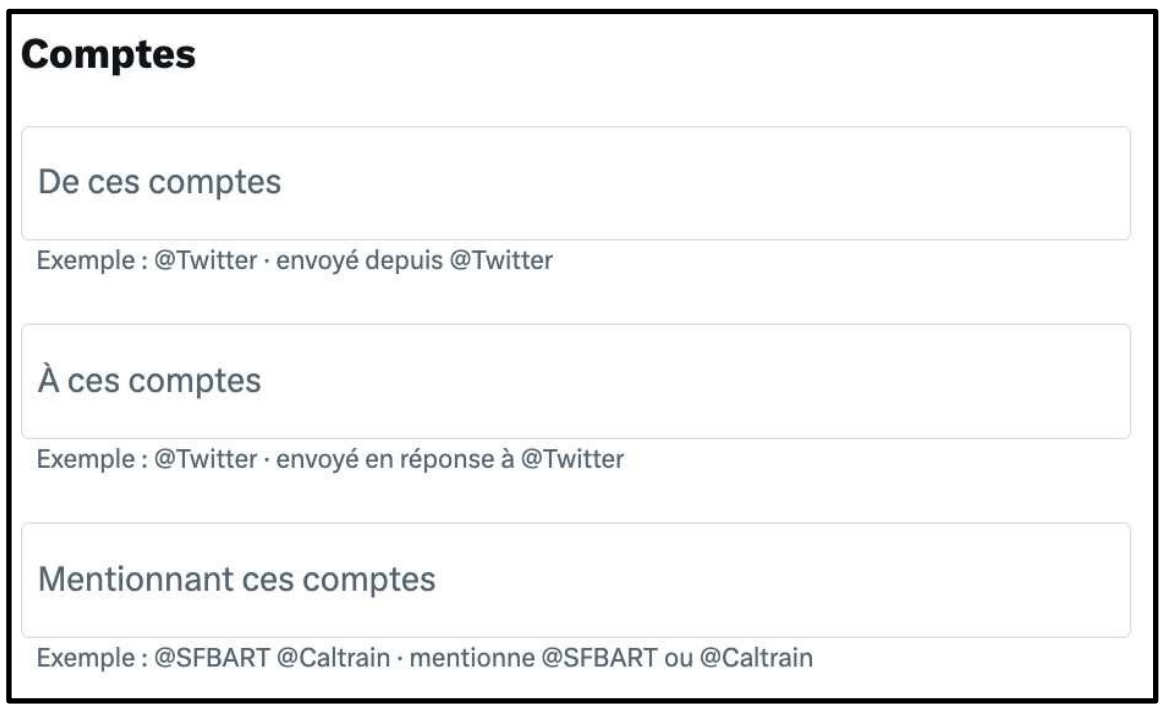

Après de longues observations sur plusieurs comptes, nous avons eu l’idée de répertorier les différents mots et groupes de mots à travers lesquels des internautes sont susceptibles de dévoiler leur âge, leur genre, leur profession ou leur engagement politique. Ces mots peuvent être classés en quatre grandes catégories :

**1. Les verbes « être » et « avoir » conjugués à la première personne du singulier :** parmi les plus couramment utilisés de la langue française, les verbes « être » et « avoir » permettent aux individus d’exprimer diverses informations relatives à leur identité, leur état, leurs émotions ou leurs actions. Par exemple, lorsqu’elle est suivie d'un adjectif qualificatif ou d’un participe passé accordé au féminin ou masculin, l'expression « je suis » permet dans de nombreux cas d’identifier le genre d'une personne, et parfois, quoique beaucoup plus rarement, sa profession ou sa date de naissance. L'expression « j'ai » quant à elle donne parfois l'opportunité d'identifier l'âge d’une personne, ses opinions ou ses pratiques. Plus largement, les verbes être et avoir, conjugués à la première personne du singulier, sont d'une grande utilité pour capturer des détails qui n'ont pas pu être observés directement au cours d’observations en ligne.

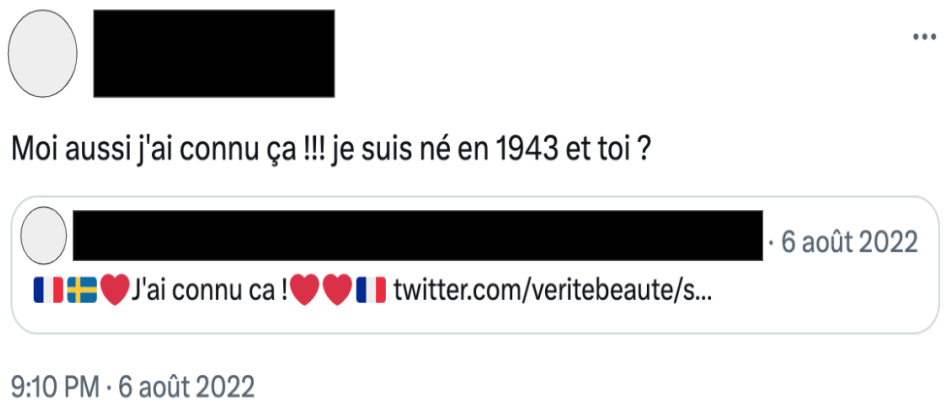

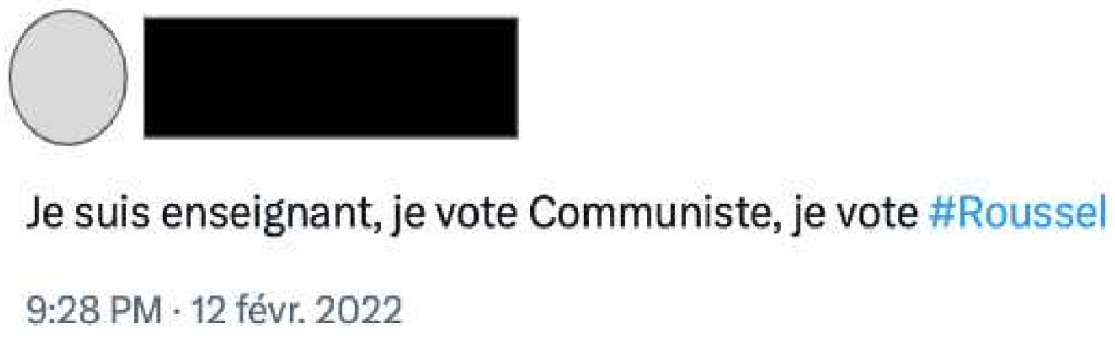

**2. Les mots relatifs à l'âge ou à la date de naissance d'une personne :** cette catégorie comprend des termes et expression tels que « âge », « ans », « années », « anniversaire » ou « naissance ».

**3. Les mots liés à l'exercice d'un travail** : ce groupe englobe des termes tels que « travail », « métier », « profession », « collègue », « patron », « salarié », « entreprise », « boulot », « taf », « job », « retraite », etc. Ils nous ont aidée à identifier les discussions liées à l'emploi ou à la profession, et ainsi à saisir les caractéristiques socio-professionnelles de certains utilisateurs.

4. **Les mots associés à l'affiliation ou au militantisme politique** : cette dernière catégorie est composée de termes tels que « carte » ; « encarté » ; « adhérent » ; « adhésion » ; « cotisation » ; « vote » ; « élection » ; « soutien » ; « manifestation » ; « pétition » ; « abstention » ;

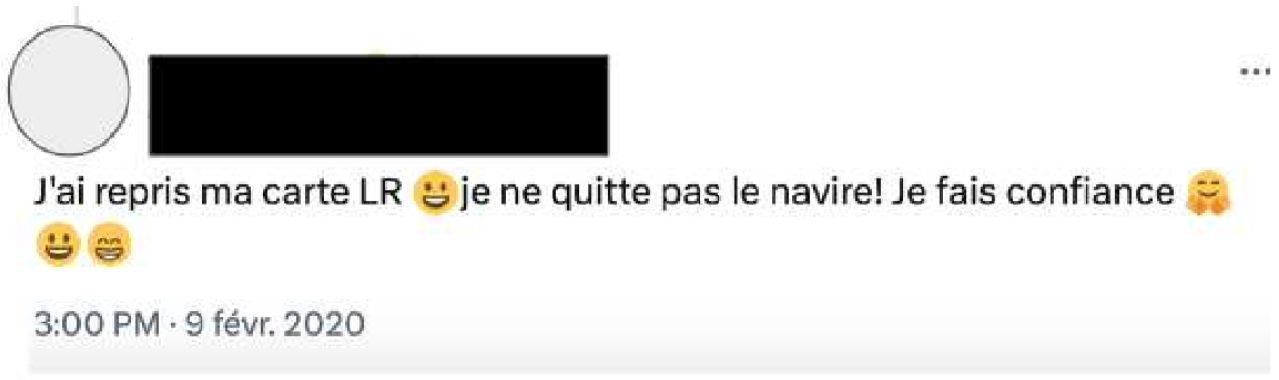

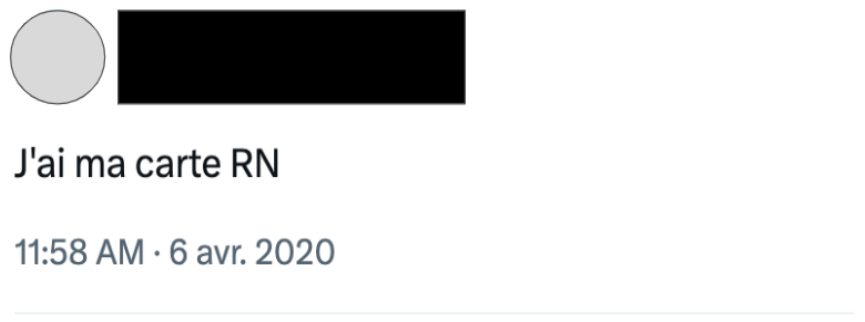

Enfin, nous avons essayé de contacter entre février et juin 2022 l'ensemble des 124 comptes de relayeurs de *fake news* analysés dans le cadre de notre ethnographie en ligne afin de réaliser des entretiens avec quelques-uns d'entre eux. Suivant l'approche développée par Alice Marwick (2014), nous avons créé un compte Twitter distinct de notre compte personnel, mais ne masquant pas notre identité civile, puis nous avons envoyé un message[93] spécifiant notre affiliation académique et nos objectifs de recherche, sans pour autant préciser aux comptes qu'ils étaient contactés pour avoir partagé un ou plusieurs contenu(s) qualifié(s) de *fake news* pour éviter qu'ils ne se sentent disqualifiés. La majorité de nos messages sont restés non lus ou sans réponse : seulement 10 internautes ont accepté notre demande d'entretien et 2 l'ont déclinée. À la fin d'un entretien, une enquêtée nous a proposé de nous mettre en contact avec son mari. Après avoir vérifié que le compte Twitter de ce dernier faisait bien partie de l'échantillon de 31 760 relayeurs de *fake news*, nous avons accepté sa proposition. Nous avons essayé de lancer une deuxième campagne d'entretiens en novembre 2022 en contactant 50 nouveaux comptes sélectionnés de façon aléatoire dans notre échantillon de relayeurs de *fake news* : 3 ont répondu de façon positive et 2 de façon négative. Au total, nous avons donc réalisé 14 entretiens semi-directifs (cf. Tableau 2.1) : 7 se sont déroulés en visio ; 5 au téléphone et 2 en face à face. Chacun d'eux a duré entre 45mn et 2h30.

Ces entretiens et observations ont permis d'avoir accès à des informations sur les parcours biographiques des enquêtés, de saisir leur position dans l'espace social et d'analyser la variété de leurs pratiques selon différentes situations d'interactions. Bien que nous n'ayons pas eu l'opportunité d'observer directement ces individus lors de conversations en face à face, nous avons pu accéder à certaines scènes de leur quotidien qu'ils relataient sur Twitter, notamment à travers des récits d'expériences ou des moments de vie partagés en ligne. Par ailleurs, les entretiens menés avec celles et ceux qui ont accepté notre invitation nous ont permis de questionner leurs interactions hors ligne.

---

[93] Au moment de lancer notre campagne d'entretiens, les utilisateurs de Twitter avaient le choix d'empêcher les personnes qu'ils ne suivaient pas de leur envoyer des messages privés. Ainsi, nous n'avons pu contacter que 56 comptes par message privé. Nous avons écrit aux 68 comptes restants en les mentionnant dans un tweet.

| Participants | Date | Pseudonyme | Âge | Genre | Niveau d'étude | Profession | Situation familiale |
|---|---|---|---|---|---|---|---|
| n°1 | 28 février 2022 | Éric | 47 | H | Bac +2 | inactif, en situation de handicap | divorcé, 2 enfants |
| n°2 | 2 mars 2022 | Jean-Yves | 60 | H | Bac +6 | pharmacien | célibataire, sans enfant |
| n°3 | 6 mars 2022 | Christian | 55 | H | Bac +8 | chercheur puis à son compte | célibataire, sans enfant |
| n°4 | 15 mars 2022 | Viviane | 70 | F | Bac +8 | retraitée, ex chercheuse | mariée, 2 enfants |
| n°5 | 23 mars 2022 | Jean-Pierre | 64 | H | Bac +6 | structureur financier | marié, 2 enfants |
| n°6 | 18 mai 2022 | Moïse | 45 | H | Bac +4 | fonctionnaire | séparé, 1 enfant |
| n°7 | 18 mai 2022 | William | 53 | H | Bac +8 | chercheur | marié, 3 enfants |
| n°8 | 19 mai 2022 | Jocelyne | 67 | F | Certificat d'étude | retraitée | mariée, 5 enfants |
| n°9 | 20 mai 2022 | José | 39 | H | Bac +3 | infirmier | marié, 2 enfant |
| n°10 | 20 mai 2022 | Richard | 57 | H | Bac +5 | chef d'entreprise | veuf, 2 enfants |
| n°11 | 9 juin 2022 | Sylvia | 58 | F | Bac +3 | inactive, ex enseignante | célibataire, sans enfant |
| n°12 | 19 novembre 2022 | Serge | 56 | H | Bac +2 | consultant | séparée, 2 enfants |
| n°13 | 22 novembre 2022 | Patrick | 60 | H | Bac technologique | employé | marié, 2 enfants |
| n°14 | 5 décembre 2022 | Gauthier | 23 | H | Bac +4 | étudiant | célibataire, sans enfant |

*Tableau 2.1. Liste des entretiens réalisés auprès de comptes ayant relayé au moins une fake news sur Twitter*

Sur Facebook, le travail de caractérisation des utilisateurs s’est concentré uniquement sur ceux ayant émis des commentaires critiques en réaction à des *fake news* (n=20 944). Le chapitre 5 précisera comment ces commentaires critiques ont été définis et identifiés. Le genre de l’ensemble de ces utilisateurs a pu être inféré à partir de leurs noms d’utilisateurs en ayant recours à des méthodes d’apprentissage supervisé. Autrement, les données récoltées sur les utilisateurs étaient beaucoup plus limitées que sur Twitter et n’ont pas permis d’identifier d'autres caractéristiques sociodémographiques à grande échelle (en mobilisant par exemple des méthodes d’inférence idéologique).

Afin d'explorer les caractéristiques socio-démographiques des profils d’utilisateurs à l’origine de commentaires critiques, nous avons sélectionné un échantillon aléatoire de 200 comptes Facebook et conduit des observations sur leur profil. Parmi tous ces comptes, environ 50 % étaient fermés ; 35 % partiellement ouverts et 15 % très ouverts. En effet, contrairement à Twitter, sur Facebook la majorité des utilisateurs ont des profils privés. Néanmoins, il est d’usage d’indiquer sur son profil Facebook son vrai nom et prénom, son niveau d’étude, son métier ou son lieu d’habitat. Par ailleurs, les utilisateurs étudiés sont beaucoup plus actifs que la moyenne, ce qui pourrait les inciter à configurer leur compte de manière à être plus visibles. En effet, certains contenus partagés par les comptes partiellement ouverts et très ouverts ont été délibérément paramétrés pour être visibles en dehors de leurs cercles d’amis. Cela nous a permis d’effectuer diverses observations, notamment sur la fréquence de partage d’informations, le niveau d'éducation, la profession ou la position politique des utilisateurs. Ensuite, des messages privés ont été envoyés à chacun de ces comptes. Sur l’ensemble des messages envoyés, 15 réponses positives ont été obtenues contre 3 négatives. Nous avons par ailleurs eu l’opportunité d’entrer en contact avec une personne responsable d’une association spécialisée dans la lutte contre les discours de haine et la désinformation. Tous les entretiens ont duré entre 40 minutes et 2h45. La majorité se sont déroulés par téléphone, 4 en visio et 1 en face à face.

| Participants | Date | Pseudonyme | Âge | Genre | Niveau d'étude | Profession | Situation familiale |
|---|---|---|---|---|---|---|---|
| **n°1** | 1er août 2022 | Julien | 34 | H | Bac | accompagnant éducatif et social | marié, deux enfants |
| **n°2** | 1er août 2022 | Nathan | 32 | H | Bac +8 | enseignant | célibataire, sans enfant |
| **n°3** | 2 août 2022 | Esteban | 44 | H | Bac +8 | médecin | marié, deux enfants |
| **n°4** | 31 octobre 2022 | Hamza | 20 | H | Bac pro | en recherche d'emploi | célibataire, sans enfant |
| **n°5** | 8 novembre 2022 | Paul | 41 | H | Bac | manager | divorcé, deux enfants |
| **n°6** | 16 janvier 2023 | Bastien | 45 | H | Bac +4 | ingénieur du son | célibataire, sans enfant |
| **n°7** | 18 janvier 2023 | Arthur | 26 | H | Bac +6 | étudiant | célibataire, sans enfant |
| **n°8** | 26 janvier 2022 | Youssef | 30 | H | Bac +5 | entrepreneur | célibataire, sans enfant |
| **n°9** | 30 janvier 2022 | Amadou | 30 | H | Brevet des collèges | conducteur de bus | marié, un enfant |
| **n°10** | 13 février 2023 | Oualid | 34 | H | Bac +3 | styliste-modéliste | marié, un enfant |
| **n°11** | 16 février 2023 | Damien | 36 | H | Bac | superviseur des bagages dans un aéroport | célibataire, un enfant |
| **n°12** | 2 mars 2023 | Maxence | 28 | H | Bac +5 | développeur | célibataire, sans enfant |
| **n°13** | 9 mars 2023 | Lucas | 33 | H | Bac +5 | data scientist | marié, sans enfant |
| **n°14** | 21 mars 2023 | Thibaud | 38 | H | Bac +5 | consultant | célibataire, sans enfant |
| **n°15** | 29 août 2023 | Théo | 24 | H | Bac +5 | étudiant | célibataire, sans enfant |
| **n°16** | 12 août 2024 | André | 47 | H | Bac +5 | ingénieur | célibataire, sans enfant |

*Tableau 2.2. Liste des entretiens réalisés auprès de comptes ayant exprimé un point d'arrêt sur Facebook*

Enfin, un travail important a été réalisé pour tenir compte de la pluralité des espaces de communication sur Facebook, décrire leurs caractéristiques et saisir comment les situations d'énonciation peuvent être perçues par les utilisateurs en couplant analyse de réseaux, observations en ligne, annotations manuelles, analyse factorielle et statistiques descriptives. Ce travail sera documenté de façon approfondie dans le chapitre 5, notamment dans la section 5.2.

En résumé, deux terrains d'enquêtes ont été investis au cours de cette thèse. Le point d'entrée sur chacun d'eux a été un corpus de contenus définis comme des *fake news,* soit par des *fact-checkers*, soit par des utilisateurs de Facebook. Cette façon d'opérationnaliser la définition du terme *fake news* a permis d'identifier un large ensemble d'utilisateurs ayant réagi de façon spontanée à au moins une *fake news* sur Facebook ou Twitter par la production d'une trace numérique. Des analyses quantitatives de leurs profils et de leurs pratiques numériques ont ensuite pu être réalisées. Cependant, un effort a été fait pour ne pas limiter ces analyses aux seules traces numériques ou à un comportement unique observé sur un réseau social en particulier. Le contexte et la situation de chaque utilisateur ont été pris en compte grâce à des observations en ligne et des entretiens. Le schéma ci-dessous offre une synthèse de l'ensemble des données analysées (cf. Figure 2.5). Des précisions seront apportées au fil des prochains chapitres sur les méthodes d'échantillonnage et d'analyse.

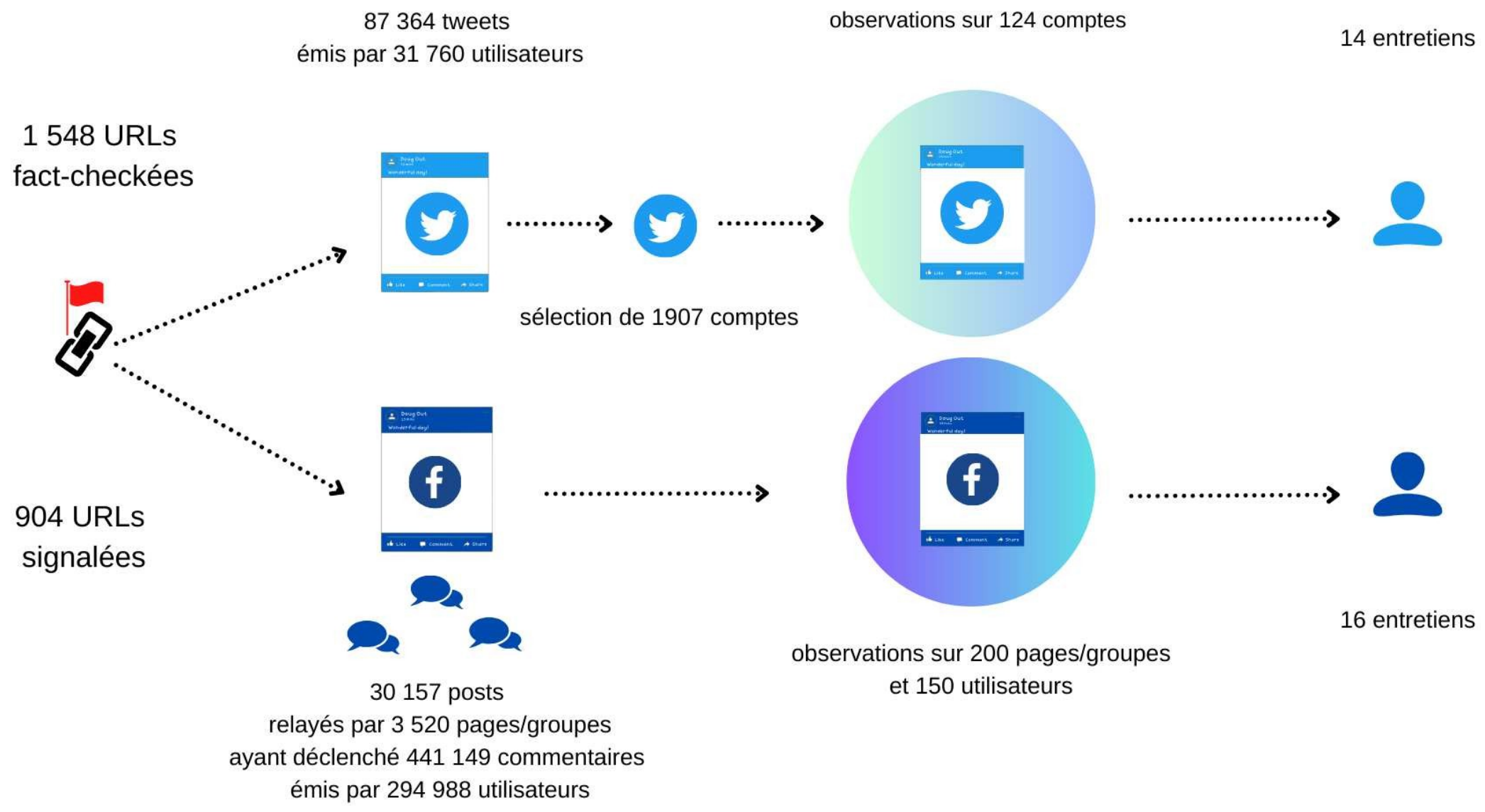

*Figure 2.5. Schéma récapitulatif des données analysées*

Il est important de noter que toutes les données recueillies sur les réseaux sociaux étaient publiquement accessibles à condition de disposer d'un compte Twitter ou Facebook. Cependant, compte tenu de l'incertitude concernant leur statut public ou privé (y compris pour les utilisateurs eux-mêmes), toutes les données ont été anonymisées, à l'exception des noms des personnalités publiques régulièrement cités dans les médias. Par ailleurs, le consentement des personnes interrogées a été systématiquement demandé avant chaque entretien afin de vérifier leur accord pour l'enregistrement de la conversation. Les informations recueillies lors de ces entretiens ont également été anonymisées par la suppression ou le remplacement des détails susceptibles de permettre l'identification des individus. Tous les noms ont été remplacés par des pseudonymes. Ceux-ci ont été attribués à l'aide du site de Baptiste Coulmont[94] afin de conserver une cohérence en termes d'âge, de position sociale et d'origine géographique. Enfin, à la demande de quelques participants, des modifications ont été apportées à certaines informations personnelles, comme l'âge ou la profession, tout en veillant à maintenir également une cohérence avec leurs caractéristiques réelles.

---

[94] https://coulmont.com/bac/results.php

La principale limite de nos terrains d'enquête réside dans le fait qu'ils sont concentrés uniquement sur les utilisateurs ayant laissé des traces numériques en réaction à des *fake news* sur les réseaux sociaux. Ils laissent donc de côté les utilisateurs ayant été exposés à des *fake news* mais qui n'y ont pas réagi publiquement, ainsi que ceux qui n'y ont pas été exposés du tout. Plutôt que de partir d'une liste de contenus définis comme des *fake news* et d'identifier seulement les utilisateurs qui y réagissent de façon publique sur les réseaux sociaux, il serait important pour de futures recherches d'adopter une approche inversée : à savoir de partir des pratiques hors ligne de différentes groupes sociaux, notamment des utilisateurs qui ne laissent aucune trace publique sur les réseaux sociaux, et de remonter à leurs usages de différentes applications ou plateformes, par exemple en leur demandant leur accord pour réaliser des observations sur leurs profils privés. Cependant, adopter une telle approche aurait présupposé de rejeter d'emblée les *fake news* en tant que catégorie construite par les discours publics. Or, il nous a semblé important dans un premier temps de prendre appui sur les contenus désignés par cette catégorie et d'étudier empiriquement les pratiques des utilisateurs qui y réagissent.

## Conclusion du deuxième chapitre

Au terme de ce chapitre, nous voici désormais équipée d'outils conceptuels et méthodologiques permettant d'aborder les pratiques informationnelles des publics en tenant compte de leurs capacités interprétatives, de leurs contextes sociaux, et de la diversité des situations d'énonciation dans lesquelles ces pratiques s'inscrivent.

En articulant des approches de sociologie de la réception et de sociologie pragmatique tout en tenant compte des apports des travaux de sociologie du numérique, le cadre théorique et méthodologique de cette thèse donne l'opportunité de : (1) penser les récepteurs d'informations comme des producteurs actifs de sens, imprégnés par leurs environnements sociaux ; (2) ne pas considérer leurs pratiques comme figées et de tenir compte de leurs oscillations ; (3) repenser la question de réception de l'information face à la numérisation de l'écosystème informationnel.

Contrairement aux discours publics et aux approches dominantes dans le champ des *fake news*, cette thèse s'ancre ainsi dans une représentation des publics numériques fondamentalement relationnelle, et une vision de l'écosystème informationnel plus ouverte aux logiques d'interaction sociale et moins réduite aux influences technologiques. L'enjeu n'est pas de confronter les utilisateurs à un contenu isolé dans un cadre expérimental, mais de comprendre comment leurs pratiques s'inscrivent dans la trame sociale de leur quotidien, au croisement de multiples situations d'interactions, qui influencent autant leurs prises de position que leurs manières de s'exprimer.

Munie de ces outils conceptuels et méthodologiques, nous voici en mesure de regarder autrement les traces numériques laissées par les utilisateurs sur les réseaux sociaux en réaction aux *fake news*.

# Deuxième partie. Du partage de *fake news* à son évitement

Cette deuxième partie propose de partir à la rencontre des utilisateurs qui partagent des *fake news* sur les réseaux sociaux en France. Sont-ils aussi nombreux que le prétendent certains acteurs du débat public ? Leurs caractéristiques correspondent-elles à celles des populations auxquelles les discours publics attribuent un plus grand manque d'esprit critique ? Constituent-ils un bloc homogène ou bien existe-t-il des disparités d'un compte à l'autre ? Observe-t-on par ailleurs des situations, notamment hors ligne, où certains utilisateurs évitent de partager des *fake news* ? Le cas échéant, comment expliquer cette variabilité ?

Pour répondre à ces questions, cette partie restitue les résultats d'une enquête menée sur la Twittosphère française. En articulant des analyses quantitatives de traces numériques à des observations en ligne et des entretiens, celle-ci a permis d'identifier les caractéristiques qui distinguent les utilisateurs qui relaient des *fake news* des autres, et d'examiner la variété de leurs pratiques (informationnelles comme conversationnelles), au sein de différentes situations d'interactions (en ligne comme hors ligne), afin de ne pas les réduire au fait d'avoir partagé une *fake news* sur un réseau social particulier.

Le chapitre 3 confirme l'existence d'un écart entre la façon dont les utilisateurs qui partagent des *fake news* sur les réseaux sociaux en France sont dépeints dans les discours publics et leurs caractéristiques réelles. Il montre que la consommation de *fake news* est loin d'affecter de façon aléatoire l'ensemble des utilisateurs des réseaux sociaux, mais n'est en réalité observable que pour un groupe restreint d'internautes dont la particularité n'est pas d'être moins éduqués ou moins dotés en compétences cognitives que les autres, mais davantage politisés et défiants à l'égard des institutions. Bien que minoritaires, ces utilisateurs sont cependant susceptibles de faciliter la mise à l'agenda des opinions défendues par leur camp politique dans le débat public en raison de leur hyperactivité en ligne et des très nombreuses informations d'actualité qu'ils partagent.

Le chapitre 4 approfondit les résultats du chapitre précédent en concentrant ses analyses sur les pratiques informationnelles et conversationnelles de la minorité d'utilisateurs à l'origine du partage de *fake news* sur Twitter. Après avoir décrit comment les comportements de ces utilisateurs, et notamment leurs régimes d'énonciation, sont modulés par leur position dans l'espace social, tout en étant susceptibles de varier d'une situation d'interaction à l'autre, ce quatrième chapitre met en évidence une compétence de distance critique, la prudence énonciative, que parviennent à mobiliser les utilisateurs des réseaux sociaux, même ceux qui relaient des *fake news*, pour préserver leur réputation et maintenir leur intégration dans différents espaces sociaux.

# Chapitre 3. Manque de raisonnement analytique ou contestation politique ?

À rebours des discours publics dépeignant les utilisateurs qui partagent des *fake news* sur les réseaux sociaux comme des foules crédules, et des études de psychologie expérimentale soulignant leur manque de raisonnement analytique (Bronstein et al, 2019 ; Pennycook et Rand 2019a ; 2020 ; Stanley et al., 2020), les enquêtes empiriques reposant sur des données observationnelles montrent que le partage de *fake news* ne concerne qu'une minorité d'internautes ayant pour particularité d'être très politisés, plutôt conservateurs et âgés (Guess et al., 2019 ; Grinberg et al., 2019 ; Osmundsen et al., 2021). Cependant, la majorité de ces constats sont centrés sur les États-Unis, un pays caractérisé par un système politique bipartisan et un espace médiatique très polarisé entre la droite et la gauche (Benkler et al., 2018). À ce jour, quasiment aucune étude à grande échelle n'a été menée auprès d'utilisateurs francophones, permettant de vérifier si les constats empiriques des enquêtes américaines s'appliquent également au cas de la France. De plus, alors que la plupart des études de psychologie expérimentale présentent d'importantes limites en termes de validité écologique pour étudier le partage de *fake news* tel qu'il se manifeste naturellement sur les réseaux sociaux, les approches observationnelles ne permettent pas de conclure que le partage de *fake news* s'explique principalement par des variables socio-politiques et non par des variables cognitives.

Pour éclairer ces différentes zones d'ombre, ce chapitre poursuit trois objectifs principaux. Sur un plan empirique, tout d'abord, il vise à vérifier si les constats des études américaines s'appliquent également au cas de la France afin de confirmer l'existence d'un écart entre la façon dont les utilisateurs qui partagent des *fake news* sur les réseaux sociaux sont dépeints dans les discours publics et leurs caractéristiques réelles – du moins telles qu'elles apparaissent lorsqu'on les étudie de manière empirique. Dans une optique méthodologique, ensuite, il cherche à dépasser les limites inhérentes aux approches expérimentales ou purement observationnelles en construisant un dispositif d'enquête permettant de déterminer de façon rigoureuse les caractéristiques qui distinguent les utilisateurs qui partagent des *fake news* des autres, tout en reposant sur des données comportementales. D'un point de vue théorique, enfin, il s'attache à valider la pertinence des choix théoriques de cette thèse, à savoir d'écarter les modèles d'explication attribuant le partage de *fake news* uniquement à des variables cognitives, notamment à un manque de raisonnement analytique, au profit d'une approche pragmatique prenant en considération les compétences critiques des acteurs. Plus largement, ce chapitre s'inscrit dans un débat académique très resserré (et surtout divisé) entre deux grandes hypothèses : celle du « raisonnement analytique » et celle de la « polarisation politique » (Pennycook et Rand, 2021 ; Osmundsen et al., 2021 ; van der Linden, 2022). En adoptant une approche descriptive, et en dialoguant avec des travaux de psychologie cognitives et de sciences politiques, il permet d'explorer si d'autres variables que le raisonnement analytique ou la position politique distinguent les utilisateurs qui relaient de *fake news* des autres et ainsi d'affiner les modèles d'explication sur le partage de *fake news*.

Pour remplir ces trois objectifs, ce chapitre se propose de mener une enquête sur la Twittosphère française. Après avoir évalué la proportion d'utilisateurs francophones à l'origine du partage de *fake news* sur Twitter, une estimation de leur position politique est ensuite proposée en mobilisant des méthodes de sciences sociales computationnelles. Étant donné que les relayeurs de *fake news* sont fortement concentrés à des endroits bien précis de l'échiquier politique, une étude comparative reposant sur une procédure de *matching* est ensuite réalisée. Celle-ci permet de comparer un échantillon de relayeurs de *fake news* à un groupe de contrôle composé de comptes Twitter présentant exactement les mêmes caractéristiques politiques afin d'identifier les spécificités propres aux utilisateurs qui partagent des *fake news* sur Twitter.

## 3.1. Une minorité d'utilisateurs concentrés aux extrêmes de l'échiquier politique et majoritairement opposés aux institutions

Cette section commence par présenter la méthodologie utilisée pour estimer la proportion d'utilisateurs qui partagent des *fake news* sur la Twittosphère française. Elle montre ensuite que ces utilisateurs ont tendance à partager principalement des *fake news* portant sur des thématiques politiques souvent accompagnées d'une rhétorique très critique envers les élites, c'est-à-dire envers un groupe restreint de personnes ou d'institutions qui exercent un pouvoir politique, économique, social, ou culturel susceptible d'influencer les décisions politiques et d'orienter les politiques publiques (Mudde, 2004). Ce constat invite alors à s'interroger sur la position politique des relayeurs de *fake news* : sont-ils répartis de façon aléatoire sur l'ensemble de l'échiquier politique français ou sont-ils au contraire fortement concentrés au sein de certaines tendances politiques employant notamment une rhétorique anti-élites ?

### 3.1.1. Identifier les relayeurs de *fake news* sur la Twittosphère française

Afin d'identifier les utilisateurs ayant partagé au moins une *fake news* sur la Twittosphère française, deux étapes ont été nécessaires (cf. Figure 3.1). Tout d'abord, nous avons sélectionné un ensemble de données permettant de représenter la Twittosphère française, c'est-à-dire l'ensemble des utilisateurs émettant des tweets en France. Pour cela, nous sommes partie d'une liste de 3 661 690 comptes d'utilisateurs ayant partagé au moins une information issue d'un corpus composé de 420 sources médiatiques françaises sur Twitter entre mai 2018 et mai 2020. Il est important de noter qu'il s'agit d'une définition très restreinte de la Twittosphère française car sur Twitter, comme sur d'autres réseaux sociaux, le nombre de personnes qui partagent des informations médiatiques est très faible (Odabaş, 2022). Sur cette période, 184 092 449 tweets ont été extraits via l'API Twitter (pour plus de détails sur la méthodologie, voir Cardon et al., 2019 et Cointet et al., 2021). Ensuite, nous avons utilisé deux bases de données contenant chacune une liste d'URLs ayant été évaluées comme des *fake news* par des journalistes spécialisés dans la vérification factuelle : (1) la base de données des *Décodeurs* du *Monde*[95] ; et (2) la base de données Condor de Facebook[96] mise à disposition de plusieurs équipes de recherche par le consortium *Social Science One* (Messing et al., 2020). Afin de pouvoir situer chaque source dans l'espace médiatique français, nous nous sommes limitée aux 1 548 URLs issues des 420 médias utilisés plus haut. Nous avons alors pu identifier 87 364 tweets (soit 0,05 %) contenant au moins une *fake news* et 31 760 comptes les ayant partagés (soit 0,8 %). À l'instar des recherches réalisées aux États-Unis, il apparaît ainsi que la mise en circulation de *fake news* sur la Twittosphère française est un phénomène marginal qui ne touche qu'une petite minorité d'utilisateurs.

Il est également important de noter que le nombre de *fake news* partagées varie grandement d'un utilisateur à l'autre (de 1 à 163). Alors que 90,7 % des utilisateurs ont partagé moins de 5 *fake news*, seulement 0,1 % en ont partagé plus de 50. Ainsi, parmi l'infime minorité de relayeurs de *fake news* identifiés sur la Twittosphère française, une très large majorité n'en a partagé qu'un petit nombre. Par ailleurs, ce n'est pas parce que tous ont partagé au moins une *fake news* sur Twitter que cela a également été le cas dans d'autres contextes. Par exemple, il est possible qu'hors ligne ou sur d'autres réseaux sociaux, certains fassent preuve d'une plus grande réserve à l'égard des informations qu'ils partagent (ou l'inverse). Le prochain chapitre s'attachera précisément à explorer pourquoi le nombre de *fake news* partagées varie autant d'un compte à l'autre et si les modes d'énonciation des utilisateurs à l'origine du partage de *fake news* sur Twitter fluctuent d'un contexte à l'autre.

---

[95] Les Décodeurs. (2017, 19 décembre). Fausses informations : les données du Décodex en 2017. *Le Monde*. https://www.lemonde.fr/le-blog-du-decodex/article/2017/12/19/fausses-informations-les-donnees-du-decodex-en-2017_5231605_5095029.html

[96] Cette base de données *Condor* contient 38 millions d'URLs qui ont été partagées au moins 100 fois publiquement sur Facebook entre le 1er janvier 2017 et le 31 juillet 2019.

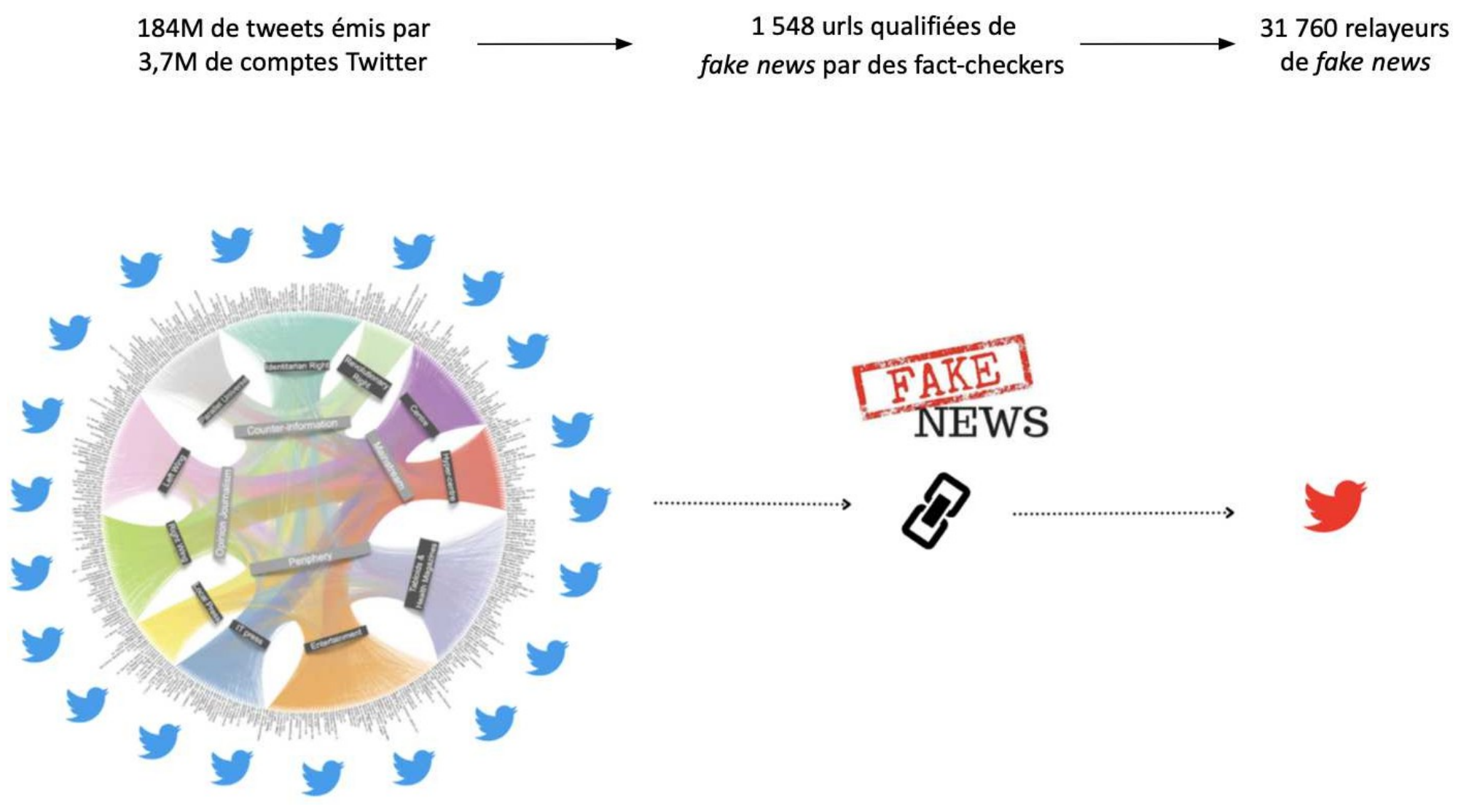

*Figure 3.1. Bases de données utilisées pour identifier les relayeurs de fake news sur la Twittosphère française*

Si les résultats obtenus mettent en avant la rareté du partage de *fake news* sur la Twittosphère française, il est toutefois important de souligner que la définition opérationnelle utilisée du terme *fake news* donne probablement lieu à une sous-estimation de leur prévalence. En effet, les *fact-checkers* n'ont pas les ressources nécessaires pour évaluer tous les tweets contenant des informations douteuses ou erronées. Il est donc très probable qu'un nombre plus élevé d'utilisateurs ait déjà partagé des *fake news* sur Twitter. Cela étant, il est possible que cette sous-estimation soit contrebalancée par notre définition restreinte de la Twittosphère française. En effet, nous avons identifié un groupe de relayeurs de *fake news* parmi des utilisateurs de Twitter ayant partagé au moins une information issue d’un média français. Or, sur Twitter, comme sur de nombreux autres réseaux sociaux, la majorité des utilisateurs sont des *lurkers*, c’est-à-dire des internautes qui consultent les publications des autres mais ne publient que très rarement des contenus. Par exemple, d'après une étude du Pew Research Center, 49 % des adultes américains qui utilisent Twitter publient moins de cinq tweets par mois (Odabaş, 2022). Si ces *lurkers* avaient été pris en compte dans notre reconstitution de la Twittosphère française, le pourcentage de relayeurs de *fake news* aurait sûrement été beaucoup plus faible. Sans pouvoir affirmer avec certitude que la prévalence exacte de relayeurs de *fake news* sur la Twittosphère française est de 0,8 % d’utilisateurs, nos résultats contribuent donc à nuancer l'ampleur réelle de la propagation des *fake news* sur la Twittosphère française.

Notons cependant que nos résultats sont circonscrits à la Twittosphère française et non généralisables à l'ensemble des réseaux sociaux. En effet, pour évaluer plus largement si le partage de *fake news* sur les réseaux sociaux en France est un phénomène aussi répandu que ce que laissent entendre certains acteurs du débat public, il aurait été nécessaire de mener une étude multi-plateforme (Bode et Vraga, 2018), comprenant des réseaux sociaux comme Facebook ou Instagram, dont les utilisateurs reflètent davantage la diversité de la population française que ceux de Twitter (Barberá et Rivero, 2015). Une telle approche nous aurait également permis de déterminer si la prévalence et les caractéristiques des internautes qui partagent des *fake news* varient selon les réseaux sociaux. Néanmoins, au moment de lancer notre collecte de données, l'API de Twitter était beaucoup plus accessible aux chercheurs que celle d'autres réseaux sociaux. De plus, sur Twitter, la majorité des comptes sont publics, contrairement à d'autres plateformes où la plupart sont privés. Par conséquent, il nous était impossible d'identifier à grande échelle des comptes d'utilisateurs impliqués dans le partage de *fake news* sur d'autres réseaux sociaux que Twitter.

### 3.1.2. Analyser les sujets des *fake news* et leur rhétorique

Avant de s'intéresser aux caractéristiques des utilisateurs qui relaient des *fake news* sur la Twittosphère française, prenons un moment pour examiner concrètement quels types de *fake news* ils ont partagé sur la période étudiée. L'objectif de cette analyse est d'identifier si certains types de *fake news* sont plus fréquemment propagées que d'autres sur Twitter.

Si le réseau de co-occurences présenté dans le chapitre 2 (cf. Figure 2.3) a permis de faire ressortir les thématiques principales des contenus identifiés comme des *fake news* par des *fact-checkers*, ou signalés comme tels par des utilisateurs de Facebook, nous proposons ici d'affiner ces analyses par des annotations manuelles afin de saisir plus précisément les thématiques spécifiques à chaque *fake news*, ainsi que la rhétorique employée par ces contenus. Une première lecture exploratoire de la liste de 1 548 URLs a permis de faire émerger 5 thématiques principales et 8 types de rhétorique (cf. Tableaux 3.1 et 3.2).

| Thématique | Description | Exemple |
|---|---|---|
| **Politique** | Le contenu mentionne des personnalités ou des partis politiques ou évoque des événements liés à l'actualité politique (e.g. adoption de loi, manifestation, scandale) | « Macron compte diminuer de 1,2 milliard d'euros la masse salariale des hôpitaux, soit la suppression de 30 000 infirmières » |
| **Société** | Le contenu porte sur un sujet d'intérêt général, en abordant des problèmes sociaux ou des enjeux sociétaux, sans faire explicitement référence à des personnalités ou à des partis politiques. | « France : c'est confirmé, la langue arabe sera enseignée dès le CP » |
| **Santé** | Le contenu fait référence à des maladies, des médicaments ou des vaccins. | « Le vapotage augmenterait le risque de cancer et de maladies cardiaques » |
| **Écologie** | Le contenu concerne des questions environnementales, le changement climatique, la conservation de la nature. | « C'est de l'arrogance de croire qu'en 150 ans d'industrialisation nous avons changé le climat » |
| **Faits divers** | Le contenu rapporte un incident inhabituel ou anecdotique (e.g. accident, disparition, phénomène étrange) et n'est pas lié à un sujet d'intérêt public. | « Un événement cosmique qui n'arrive que tous les 35 000 ans : le 27 juillet 2018, la planète Mars sera aussi grosse que la Lune » |

*Tableau 3.1. Grille d'annotations utilisée pour identifier les thématiques de chaque fake news partagée sur Twitter*

Les analyses réalisées font ressortir que la majorité des *fake news* partagées sur Twitter portent sur des sujets politiques (34 %) et de santé (28 %), et emploient fréquemment une rhétorique anti-élite (29 %) ou vaccino-sceptique (18 %). Étant donné le contexte de la période étudiée, il est important de noter que les *fake news* relatives à des sujets de santé peuvent également comporter une dimension politique importante et exprimer une certaine défiance envers les élites et les institutions. En effet, la crise sanitaire a mis en évidence une politisation croissante des enjeux de santé, comme l'a par exemple illustré la controverse sur le traitement par hydroxychloroquine, ou encore les réticences vaccinales de la population (Stroebe et al., 2021 ; Ward et al., 2021). Les individus ont été nombreux à appréhender ces enjeux au prisme de leurs propres convictions idéologiques ou partisanes (Gauchat, 2012 ; Gostin, 2018) et à contester la gestion de la crise par le gouvernement (Berriche et al., 2020).

| Rhétorique | Description | Exemple |
|---|---|---|
| **anti-élite** | Le contenu s'oppose aux discours et actions portés par des personnes ou des institutions exerçant un pouvoir politique, économique, social, ou culturel significatif au sein de la société. | « 120 000€ détournés par Macron : Internet expose ses casseroles puisque les médias ne le font pas » |
| **identitaire** | Le contenu défend les valeurs, la culture ou la religion de la population française ou européenne mais rejette celles d'autres groupes. | « La voilée Yasmine ne peut pas recevoir le prix du jeune européen, elle ne peut pas être européenne » |
| **vaccino-sceptique** | Le contenu exprime une ou plusieurs critiques à l'égard de différents sujets liés à la vaccination. | « Vaccin hépatite B: La Cour européenne reconnaît le lien avec la sclérose en plaques » |
| **chimiophobe et/technophobe** | Le contenu dénonce l'utilisation de produits chimiques ou de nouvelles technologies. | « Le Dr Henri Joyeux, cancérologue a déclaré "85 % des chimiothérapies sont contestables, voire inutiles" » |
| **climatosceptique** | Le contenu remet en cause l'existence du réchauffement climatique ou nie la responsabilité humaine dans le phénomène. | « Une étude scientifique finlandaise conclut que le GIEC se trompe : il n'y a pas de preuves que le changement climatique est dû à l'homme » |
| **sensationnaliste** | Le contenu ne cherche pas à influencer l'opinion des internautes mais à les surprendre ou à les choquer. | « Cette femme morte sort de sa tombe et marche 3 ans après son décès » |
| **sécuritaire** | Le contenu met l'accent sur le maintien de l'ordre et dénonce des actes de violence ou de délinquance. | « Notre-Dame : la piste criminelle doit être privilégiée, ce qui est arrivé était impossible » |
| **autre** | Le contenu ne rentre dans aucune catégorie. | « Le journal de 18h : hausse des accidents sur les routes à 80km/h » |

*Tableau 3.2. Grille d'annotations utilisée pour identifier la rhétorique employée par chaque fake news partagée sur Twitter*

Il est important de souligner également qu'une proportion non négligeable de *fake news* (15 %) cible spécifiquement des groupes perçus comme « assistés » ou « étrangers » plutôt que des élites. Ce type de discours résonne avec la notion de « conscience sociale triangulaire » développée par Olivier Schwartz (1990), qui souligne que la perception de la société par les individus n'est pas uniquement hiérarchisée entre « eux » (les élites) et « nous » (le peuple), mais inclut aussi une dynamique d'opposition vis-à-vis des groupes perçus comme appartenant aux échelons les plus bas de l'échelle sociale.

En résumé, ces analyses suggèrent que les utilisateurs de Twitter sont particulièrement sensibles aux *fake news* liées à des enjeux politiques ou sanitaires, surtout lorsqu'elles s'accompagnent d'un discours critique envers les élites ou les vaccins, ou encore à l'encontre des immigrés ou des étrangers. Cela invite à s'interroger sur les positions politiques des utilisateurs à l'origine de la diffusion de ces contenus, notamment en explorant si leur comportement est associé à une opposition aux élites ou aux institutions sanitaires, ainsi qu'à des discours identitaires et xénophobes.

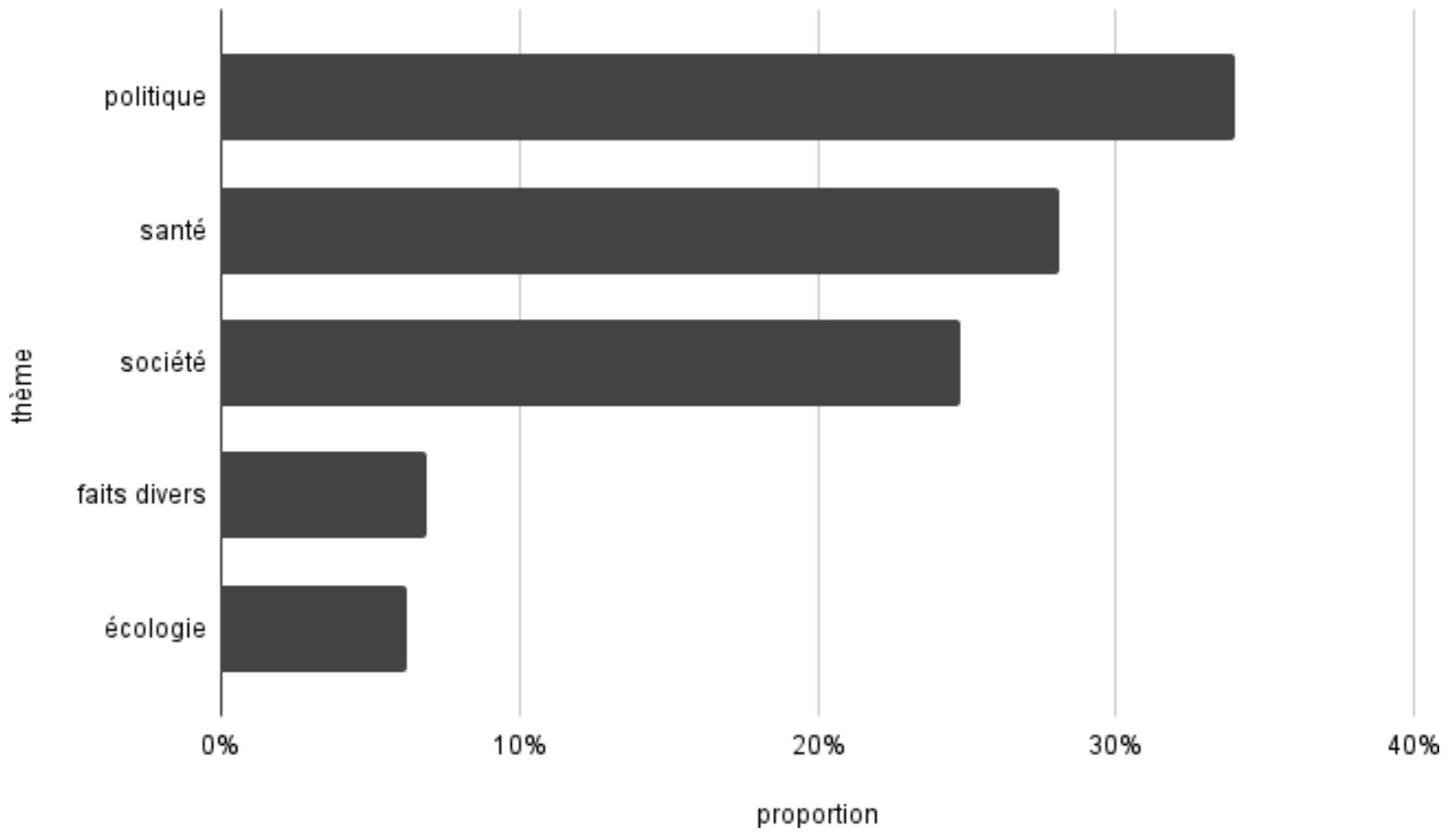

*Figure 3.2. Proportion de chaque thématique parmi l'ensemble des fake news partagées sur Twitter*

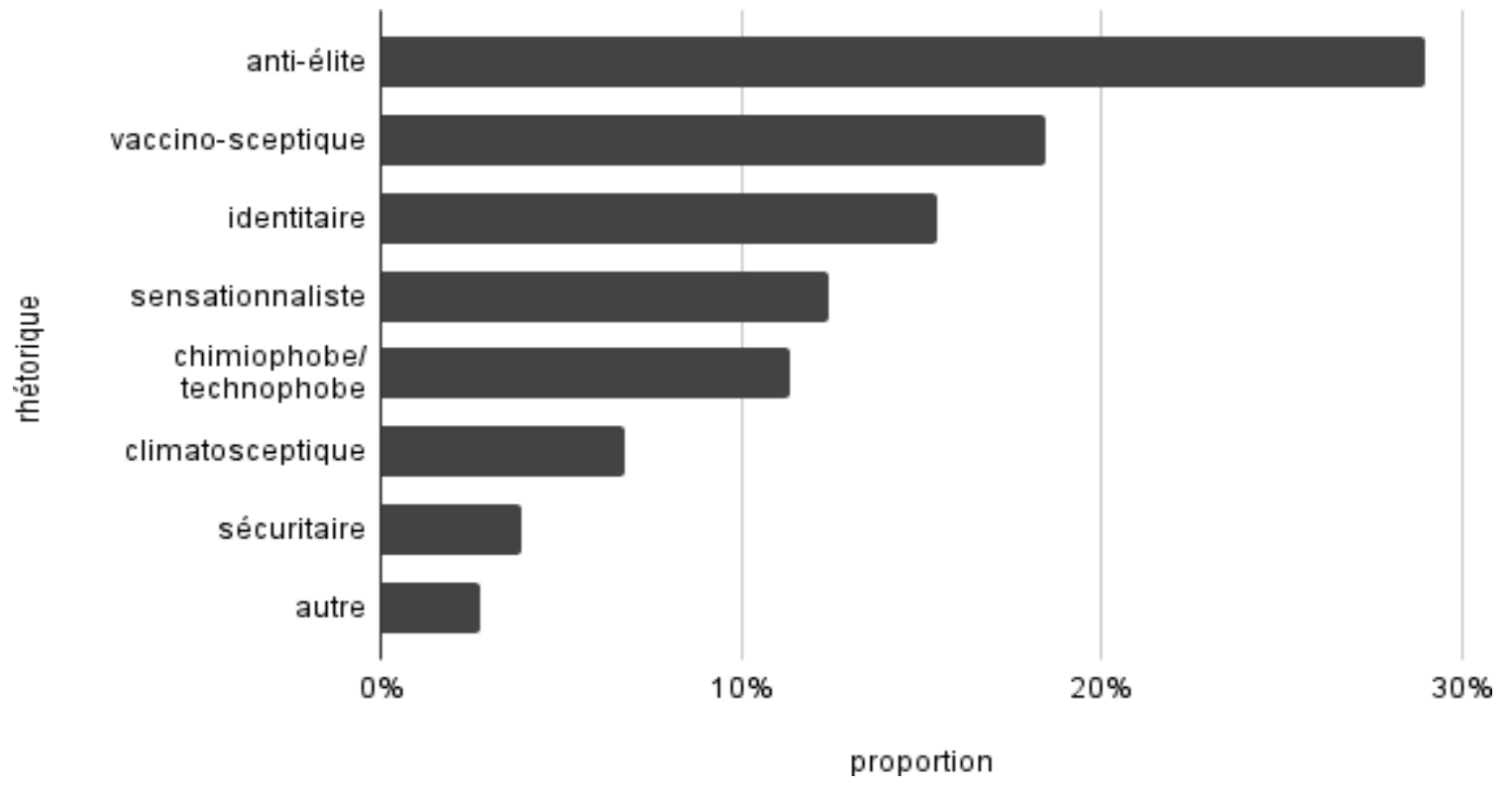

*Figure 3.3. Proportion de chaque type de rhétorique parmi l'ensemble des fake news partagées sur Twitter*

### 3.1.3. Déterminer la position politique des relayeurs de *fake news*

Dans la mesure où la majorité des *fake news* partagées sur Twitter portent sur des sujets politiques, nous avons cherché à estimer la position politique des internautes qui les ont relayées. Pour ce faire, nous avons tiré parti de l'espace idéologique latent construit par des travaux de sciences sociales computationnelles réalisés au médialab de Sciences Po (Cointet et al., 2021 ; Ramaciotti Morales et al. 2021). Ces travaux prennent appui sur l'approche méthodologique développée par Pablo Barberá (2015), elle-même inspirée d'une étude de Robert Bond et Solomon Messing (2015) sur Facebook, et reposent sur une hypothèse d'homophilie classique : deux individus ayant des opinions politiques similaires devraient avoir tendance à suivre les mêmes comptes d'élus sur Twitter. Après avoir listé tous les comptes des députés français actifs sur Twitter (soit 812 sur 925), puis collecté tous leurs *followers*, et enlevé ceux suivant moins de 3 députés et/ou ayant moins de 25 abonnés, Jean-Philippe Cointet, Pedro Ramaciotti Morales et leurs collègues ont pu inférer la position politique de 368 831 utilisateurs de Twitter en France. L'originalité de leur approche, par rapport aux études pionnières de Robert Bond et Solomon Messing (2015) ou de Pablo Barberá (2015), est d'avoir dégagé deux axes principaux pour interpréter l'espace idéologique latent : le premier peut être considéré comme une représentation du clivage droite-gauche, tandis que le second permet de capturer les sentiments et les attitudes des comptes envers les élites (Ramaciotti Morales et al., 2021). Si la polarisation des individus sur un axe gauche/droite constitue une dimension particulièrement pertinente pour étudier des pays bipartites comme les États-Unis, elle s'avère limitée pour comprendre différents phénomènes politiques concernant des pays plus divers idéologiquement, notamment en Europe (Benoit and Laver, 2012 ; Bakker et al., 2012). Plusieurs études ont souligné l'importance des positions des individus sur des questions liées à l'intégration européenne, l'immigration et la mondialisation pour expliquer différents comportements politiques (Hix et al., 2006 ; Ramaciotti Morales et al., 2021). Par ailleurs, quelques études ont montré que l'opposition des publics au gouvernement et aux partis politiques (Uscinski et al., 2021), ou la méfiance envers les médias, les institutions et les experts étaient décisives pour expliquer la propagation des rumeurs politiques (Petersen et al., 2023) ou l'adhésion à des discours conspirationnistes (Miller et al., 2016).

Pour estimer les positions politiques des utilisateurs partageant des *fake news* sur la Twittosphère française, il a alors suffi de retrouver lesquels parmi eux figuraient dans la liste de 368 831 utilisateurs de Twitter établie par Jean-Philippe Cointet, Pedro Ramaciotti Morales et leurs collègues, puis d'observer leur distribution dans l'espace idéologique latent. Sur les 31 760 relayeurs de *fake news* de départ, 10 910 comptes ont pu être retrouvés, soit un peu plus de 30 %. Cette proportion est importante puisque seulement 8,6 % des comptes d'utilisateurs de la base de données initiales sont positionnés dans l'espace idéologique latent ; ce qui suggère que les relayeurs de *fake news* sont beaucoup plus politisés que la moyenne dans la mesure où le fait de suivre des députés sur Twitter est un indicateur d'un intérêt marqué pour la politique. Bien entendu, ce résultat ne signifie pas que les 20 850 relayeurs de *fake news* restants n'ont pas d'opinion politique, mais seulement qu'il n'a pas été possible de l'inférer à partir de leurs traces comportementales sur Twitter car ils ne suivent pas suffisamment de comptes de parlementaires ou n'ont pas suffisamment d'abonnés. Autrement dit, nous reposons ici sur une définition très restrictive de la notion politique, mais il ne faut pas perdre de vue qu'en marge du « politique spécialisé » il existe des formes de politisation plus ordinaires issues d'expériences quotidiennes et d'interactions sociales informelles (Duchesne et Haegel, 2001 ; Kokoreff et Lapeyronnie, 2013). La suite des analyses va porter sur les relayeurs de *fake news* rattachés à l'espace idéologique latent, mais il serait très important pour de futures recherches de tenir compte des utilisateurs qui se situent en dehors de cet espace.

Afin d'étudier la distribution des relayeurs de *fake news* dans l'espace idéologique latent, nous avons repris le découpage en 16 rectangles effectué par Jean-Philippe Cointet et ses collègues (2021) dans leur étude sur les Gilets Jaunes. Ce découpage correspond à un partitionnement de l'espace idéologique en parcelles rectangulaires de taille variable mais regroupant le même nombre de comptes. Le calcul des frontières pour ce partitionnement résulte de la construction d'un arbre kd (*k-dimensional tree*). Cette procédure de partitionnement de l'espace consiste à opérer des bi-sections successives sur un ensemble de données de sorte que le nombre de points dans chaque cellule reste constant. À chaque étape, l'arbre kd divise l'espace en deux parties selon une dimension choisie en cherchant à équilibrer le nombre de points (en l'occurrence de comptes) dans chaque cellule. Ce processus se répète jusqu'à ce que l'ensemble des données soit découpé en cellules. Ce découpage en 16 cellules a été retenu car il permet d'obtenir des segments relativement bien définis sur le plan politique tout en garantissant une taille suffisante des populations pour permettre des calculs de proportions statistiquement significatives. Cette approche est avantageuse car elle facilite la visualisation des différences de densité des relayeurs de *fake news* au sein d'un espace initialement hétérogène, où certaines zones peuvent être très peuplées alors que d'autres le sont beaucoup moins.

Alors que l’ensemble de l’espace idéologique latent contient 368 831 comptes, chaque rectangle représente environ 22 000 individus. Si le partage de *fake news* concernait de façon égale et indifférenciée toutes les communautés politiques, les 10 910 comptes de notre échantillon devraient être répartis équitablement dans les 16 rectangles. Autrement dit, on devrait s’attendre à ce qu’il y ait environ 682 relayeurs de *fake news* dans chaque rectangle, ce qui représenterait 3,2 % de l’ensemble des comptes de chaque communauté politique. Or, comme le montre la Figure 3.4, les internautes qui relaient des *fake news* sont répartis de manière très inégale dans l'ensemble de l'espace idéologique latent : ils sont très concentrés dans sa partie inférieure, avec une densité encore plus élevée dans les rectangles inférieurs de droite et, dans une moindre mesure, dans les rectangles inférieurs de gauche. En effet, la densité est de 8,6 % à l’extrême gauche et de 10 % à l’extrême droite. À l’inverse, c’est dans les rectangles de centre-droit qu’elle est la plus faible (entre 0,7 et 0,8 %). Il est important de souligner ici que les notions d’« extrême droite » et d'« extrême gauche » sont utilisées dans un sens géométrique et spatial, sans jugement de valeur, pour qualifier les comptes qui se trouvent aux extrémités de l’axe droite-gauche de l’espace idéologique latent. Il est possible que ces termes soient contestés par les acteurs eux-mêmes et ne reflètent pas la classification réalisée par l’instruction relative à l’attribution des nuances des candidats des élections sénatoriales, qui en août 2023 a par exemple classé La France Insoumise à gauche et non à l’extrême gauche.[97]

[97] https://www.legifrance.gouv.fr/download/pdf/circ?id=45472

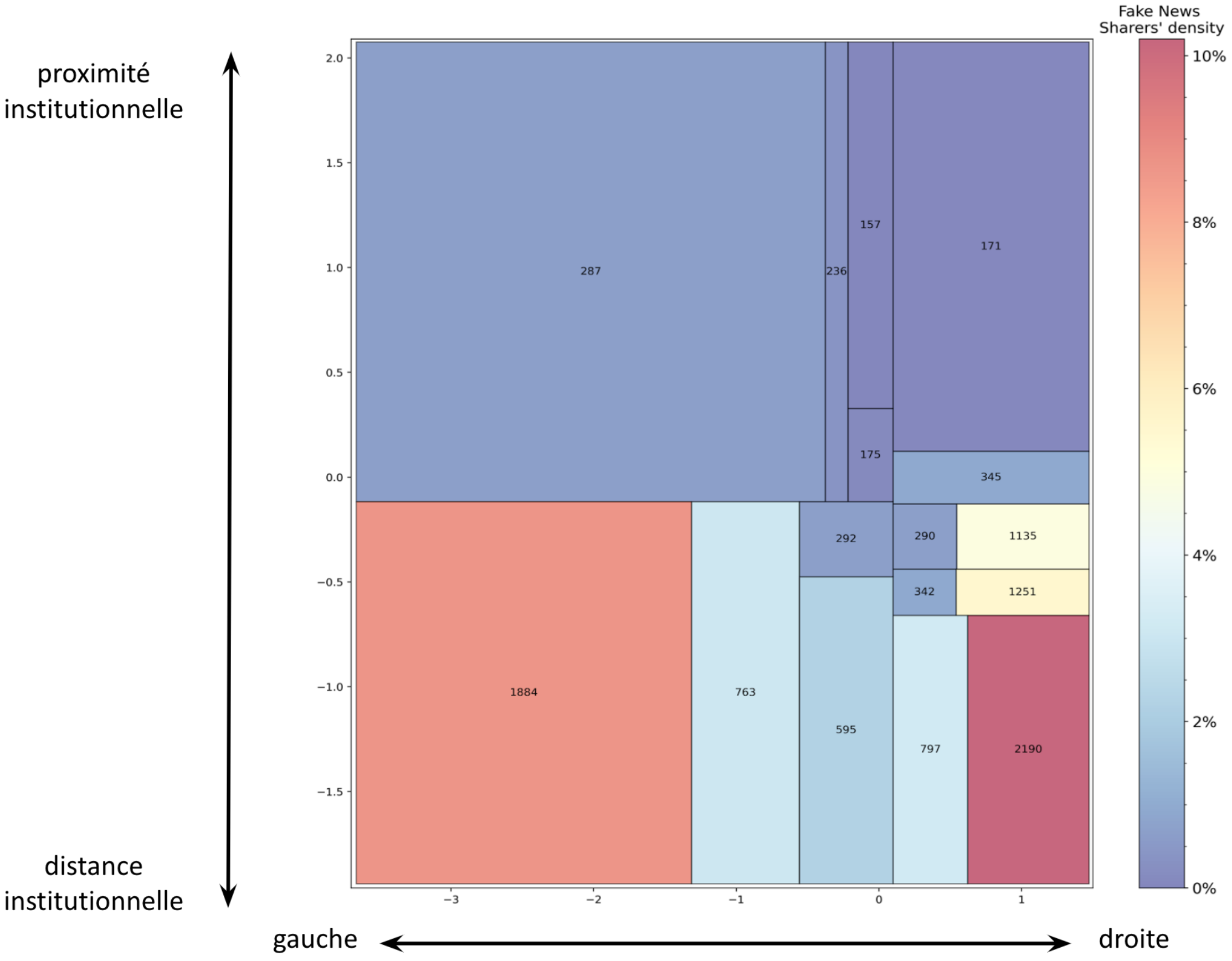

*Figure 3.4. Distribution des relayeurs de fake news dans un espace idéologique latent*

Note de lecture : chaque rectangle, indépendamment de son périmètre, représente environ 22 000 comptes Twitter. Le nombre affiché dans chaque rectangle indique le nombre de comptes ayant partagé au moins une *fake news*. Plus la couleur d'un rectangle tend vers des couleurs chaudes (orange et rouge), plus la présence de relayeurs de *fake news* est élevée. Inversement, plus la couleur d'un rectangle tend vers des couleurs froides (bleu et violet), plus la présence de relayeurs de *fake news* est faible.

En France, donc, si le partage de *fake news* concerne principalement des internautes positionnés à l'extrême droite, comme aux États-Unis, il touche également une part non négligeable d'internautes situés à l'extrême gauche. La plus forte densité des relayeurs de *fake news* au sein de l'extrême droite par rapport à l'extrême gauche est cependant particulièrement notable dans la mesure où l'électorat de gauche est surreprésenté sur Twitter, alors que celui d'extrême-droite est largement sous-représenté (Boyadjian, 2016).[98] En fait, l'originalité de nos résultats est surtout d'apporter une dimension supplémentaire au cadre d’analyse mobilisé jusqu’à présent par les études américaines en montrant que les relayeurs de *fake news* d’extrême droite comme d'extrême gauche se caractérisent par un fort sentiment d’hostilité envers les élites. Il peut dès lors s’avérer pertinent de s’interroger sur les dynamiques socio-politiques qui alimentent le plus fortement la mise en circulation de *fake news* au sein de la Twittosphère française. Ce phénomène résulte-t-il davantage d’un clivage traditionnel entre la gauche et la droite ou plus largement d’une hostilité des internautes envers les élites ?

[98] Ces chiffres ont cependant pu évoluer depuis 2016, notamment depuis le rachat de Twitter par Elon Musk en 2022 et les changements qui en ont découlé.

En calculant la somme de chaque rectangle situé de part et d’autre de l’axe horizontal et vertical, il ressort que les utilisateurs situés à droite de l’échiquier politique (n=6 521) sont 1,5 fois plus susceptibles que ceux situés à gauche (n=4 389) de relayer des *fake news*, tandis que ceux qui ont les positions anti-élites les plus marquées (n=9 539) sont 7 fois plus susceptibles que ceux étant le moins opposés aux élites (n=1 371) de partager des *fake news*. Il ressort ainsi que l’axe mesurant les attitudes des utilisateurs envers les élites est plus prédictif de la probabilité de partager des *fake news*. Néanmoins, cette seule dimension n’est pas suffisante pour expliquer complètement les positions des propagateurs de *fake news*. En effet, il faut noter que les taux les plus élevés d'utilisateurs partageant des *fake news* se trouvent parmi ceux qui ont de forts sentiments anti-élites, mais seulement dans la mesure où ils sont également polarisés dans l'échelle gauche-droite.

Dans l'ensemble, ces résultats indiquent que le partage de *fake news* sur Twitter est un symptôme d'une crise de la représentation politique, voire d’une crise de la démocratie représentative car comme l’exprime Didier Mineur (2010) dans son livre *Archéologie de la représentation politique. Structure et fondement d’une crise* :

> *Quand l’insatisfaction des citoyens s’exprime en faveur de partis qui se veulent « antisystème », ou qui prônent le rejet ou la rupture vis-à-vis des institutions représentatives, on change de registre : il ne s’agit plus d’une crise « classique » de la représentation, mais d’une crise de la démocratie représentative en tant que telle.*

En partageant des *fake news* (notamment à caractère politique et anti-élite), les utilisateurs de Twitter cherchent sans doute à exprimer un sentiment de déconnexion entre leurs préoccupations, leurs aspirations et les décisions prises par les élites au pouvoir. Cette hostilité envers les élites les pousse à rechercher des informations alternatives qui correspondent à leurs points de vue et défendent leurs intérêts, même si ces informations s'avèrent être fausses. La propagation de *fake news* peut donc être vue comme une manifestation de la crise de la représentation, où les citoyens cherchent à perturber l'ordre établi en raison de leur mécontentement envers le système politique traditionnel. Cette interprétation peut être mise en perspective avec la notion de *need for chaos* (i.e. besoin de chaos) utilisée par les politologues Michael Bang Petersen, Mathias Osmundsen et Kevin Arceneaux (2023). Selon les auteurs, le besoin de chaos — défini comme « le désir d'un nouveau départ par la destruction de l'ordre et des structures établies » — est étroitement lié à un sentiment de marginalisation dans la société. La perturbation politique apparaît alors comme une manière de renverser le pouvoir au sein de la société afin de conquérir un statut social. Le besoin de chaos pousse ainsi les individus à cibler l'ordre politique dans son ensemble plutôt qu'un camp particulier du spectre politique. Il est ainsi possible que les relayeurs de *fake news* soient moins animés par des motivations partisanes, cherchant à promouvoir les idées de leur camp politique, que par des attitudes cyniques et contestataires à l'égard des institutions et des organes gouvernementaux. Cette hypothèse invite à s'interroger sur le degré d'affiliation à un parti politique des utilisateurs qui partagent des *fake news*. Bien qu'ils soient plus proches de certains courants politiques, notamment d'extrême droite et d'extrême gauche, sont-ils encartés dans leurs partis ? Se sentent-ils représentés par leurs élus ? Adhèrent-ils aux idées défendues par leur programme ? Ou bien les soutiennent-ils principalement pour leur discours contestataire à l'égard de l'ordre politique établi ? Répondre à ces questions permettra d'affiner l'identification des mécanismes socio-politiques qui favorisent ou inhibent le partage de *fake news*.

## 3.2. Saisir les spécificités des relayeurs de *fake news*

Nous venons de voir que le partage de *fake news* ne concerne qu'une minorité de comptes Twitter en France dont la particularité est d'être très concentrés aux extrêmes de l'échiquier politique et fortement opposés aux élites et aux institutions. Si ce résultat, obtenu à partir d'une approche observationnelle, suggère que la mise en circulation de *fake news* sur les réseaux sociaux n'est pas le fait d'une foule crédule mais plutôt d'une minorité d'utilisateurs très défiants à l'égard des élites, il ne permet pas de conclure que le partage de *fake news* ne s'explique que par les positions politiques des utilisateurs et non par d'autres facteurs tels qu'un défaut de raisonnement analytique. En effet, il est possible que les relayeurs de *fake news* aient des motivations politiques tout en manquant d'esprit critique — et même que le manque d'esprit critique soit le facteur le plus prédictif du partage de *fake news*. Cet argument a été formulé par les chercheurs en sciences cognitives Gordon Pennycook et David Rand (2021) : après avoir passé en revue les données de 14 études expérimentales dans lesquelles des participants américains ont dû évaluer la crédibilité de différentes informations d'actualité (vraies ou fausses et pro-démocrates ou pro-républicaines), ceux-ci ont montré que l'effet de la concordance politique (entre l'opinion des individus et celle véhiculée par le contenu informationnel) était plus faible que celui de la véracité des informations dans la mesure où les informations vraies mais discordantes avec les opinions politiques des enquêtés étaient considérées comme plus crédibles que les informations fausses mais politiquement alignées avec leur opinion politique. Nous avons cependant montré au début de notre thèse que les méthodes expérimentales pouvaient donner lieu à des inférences causales factices, notamment en raison de leur manque de validité écologique et des limites des données déclaratives pour mesurer des opinions ou des comportements. À l'inverse, les approches observationnelles permettent de décrire avec plus de justesse la réalité en examinant une population ou un phénomène sans intervenir sur le cours naturel des choses, mais ne permettent pas de démontrer que les liens (corrélations) observés sont de nature causale car elles ne peuvent déterminer quelles variables agissent les unes sur les autres. Dès lors, la question s'est posée de savoir comment mettre en place un dispositif d'enquête permettant de dépasser les limites inhérentes aux approches expérimentales ou purement observationnelles afin d'identifier de façon rigoureuse non pas directement la cause du partage de *fake news*, mais les caractéristiques qui distinguent les utilisateurs qui en partagent des autres, tout en reposant sur des données comportementales.

Pour répondre à cette question, une procédure de *matching* a été utilisée (Rubin, 1973). Étant donné que les relayeurs de *fake news* sont très concentrés à des endroits bien précis de l'échiquier politique, cette approche permet de neutraliser l'effet de leur position politique dans l'étude du partage de *fake news*. En comparant un échantillon de relayeurs de *fake news* à un groupe de contrôle composé de comptes Twitter présentant exactement les mêmes caractéristiques politiques, il devient possible de distinguer les spécificités propres aux utilisateurs qui partagent des *fake news* sur Twitter. Concrètement, nous avons sélectionné un échantillon de 1 907 partageurs de *fake news* et de 958 utilisateurs n'en ayant pas partagé mais occupant la même position politique (i.e. appartenant au même rectangle de l'espace idéologique présenté ci-dessus). Nous avons ensuite collecté les 3 200 derniers tweets de chaque compte en octobre 2020. Nous avons obtenu 3 644 421 tweets publiés par des relayeurs de *fake news* et 459 422 tweets publiés par des utilisateurs n'en ayant pas partagé. Après avoir vérifié que les comptes de notre échantillon n'étaient pas des bots et que l'hyperactivité des partageurs de *fake news* n'expliquait pas à elle seule leur partage de *fake news* (cf. Annexes 3 et 4), nous avons construit un ensemble de variables permettant d'analyser l'identité en ligne, les pratiques informationnelles, les usages de Twitter et le style de langage des comptes (cf. Figure 3.5). Pour chaque variable, nous avons effectué soit des tests du $\chi^2$ (pour vérifier si les fréquences observées dans chacun de nos deux groupes correspondaient aux fréquences attendues), soit des tests de Wilcoxon-Mann-Whitney[99] (pour comparer les distributions de nos deux échantillons).

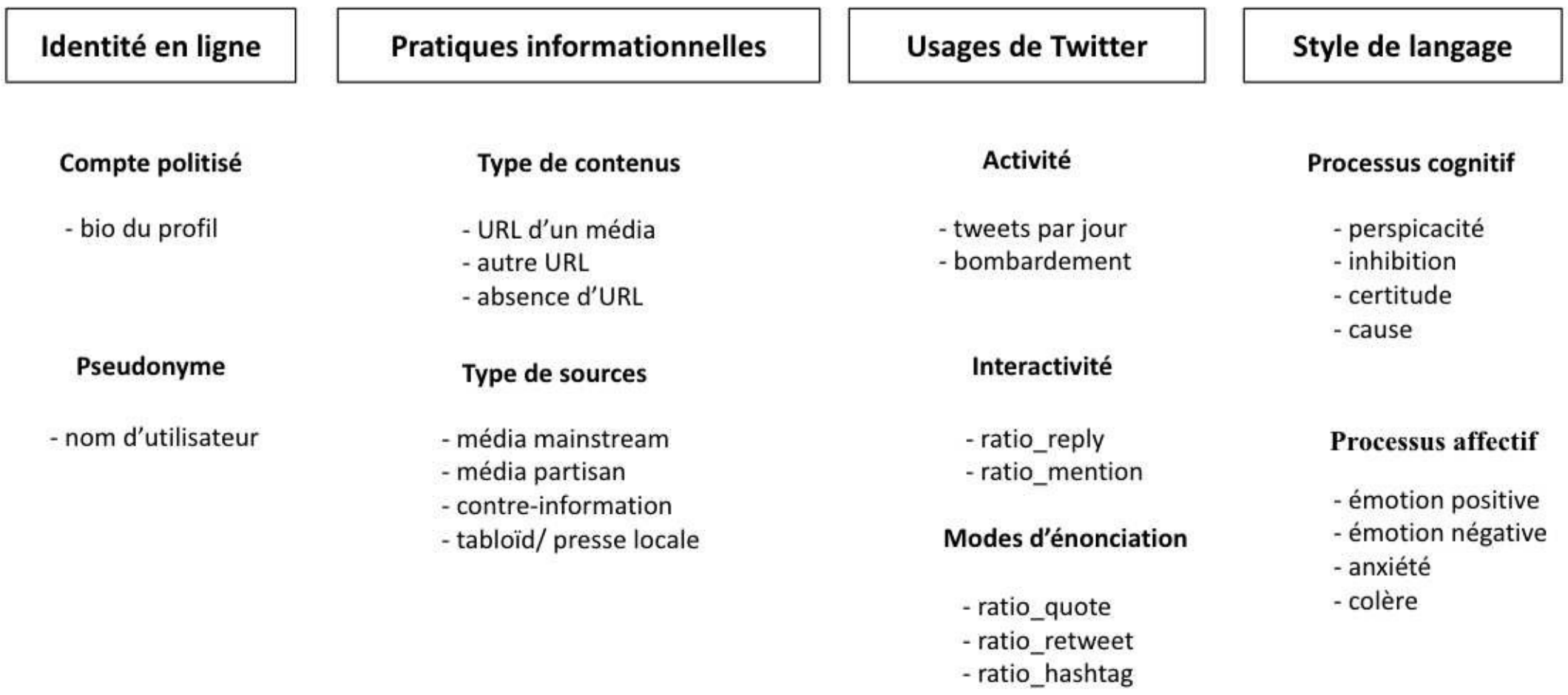

*Figure 3.5. Liste des variables étudiées par l'analyse quantitative*

[99] Nous avons eu recours à ce test statistique non paramétrique plutôt qu'à un test de *Student* car les données des deux échantillons étudiés ne suivaient pas une loi normale.

Nous avons ensuite complété ces analyses statistiques par des observations en ligne réalisées sur 124 comptes de relayeurs de *fake news* et 115 non-relayeurs de *fake news*, ainsi que par 14 entretiens menés auprès d'utilisateurs ayant partagé au moins une *fake news* sur Twitter (pour plus d'informations sur la constitution de ces échantillons et la méthodologie déployée voir Chapitre 2, section 2.2.3 et 2.3.3).

L'articulation de ces différentes approches méthodologiques a permis d'identifier les caractéristiques distinguant les relayeurs de *fake news* des utilisateurs du groupe de contrôle indépendamment de leur position politique. Les résultats de l'étude comparative sont détaillés ci-dessous. Il est très important de souligner de nouveau que les analyses conduites dans le cadre de cette enquête ne concernent qu'un type particulier de relayeurs de *fake news*, ceux qui suivent au moins trois députés sur Twitter. Par conséquent, elles ne peuvent pas être généralisées à l'ensemble des utilisateurs qui relaient des *fake news* sur les réseaux sociaux.

### 3.2.1. Des comptes très politisés

Une première variable a été utilisée pour examiner si les utilisateurs à l'origine du partage de *fake news* sur Twitter sont davantage politisés que les autres. Traditionnellement, la notion de politisation renvoie à « l'attention accordée au fonctionnement du champ politique » (Gaxie, 1978, p. 240). Ici, l'analyse a plutôt porté sur la place occupée par la politique dans l'identité numérique présentée par les utilisateurs (Greene, 2004).

Afin d'explorer si l'identité politique des utilisateurs prime sur leurs autres appartenances sociales (Duchesne et Scherrer, 2003), du moins sur Twitter, les bios des comptes ont été passées en revue. Lors de l'ouverture d'un compte Twitter, chaque utilisateur est invité à se présenter aux autres en écrivant une courte biographie de 160 caractères. En général, les utilisateurs profitent de cet espace pour partager leurs centres d'intérêt, évoquer leur situation familiale, indiquer leur métier ou partager une citation qui les inspire. Ce texte d'auto-description est ensuite publiquement accessible pour les autres utilisateurs. Il permet d’indiquer comment les utilisateurs souhaitent être perçus par leur audience et quelles facettes de leur identité ils veulent mettre en avant (Rogers et Jones, 2021).

Après une lecture exploratoire d’une centaine de bios, nous avons décidé de nous focaliser sur la présence de marqueurs politiques. Nous avons considéré qu'une bio contenait des marqueurs politiques lorsqu'un utilisateur exprimait son appartenance à un parti politique ou son soutien à une personnalité politique ; lorsqu'il mettait en avant ses valeurs idéologiques ou ses convictions en général ; ou encore lorsqu’il critiquait les opinions ou personnalités de camps adverses (cf. Figure 3.6 pour des exemples de bio avec et sans marqueur politique). Nous avons annoté manuellement 500 bios. Pour généraliser ces annotations à l’ensemble des bios des comptes de l’échantillon, un classifieur basé sur le modèle CamemBERT a ensuite été entraîné. Les performances du classifieur se sont avérées très satisfaisantes avec un score F1 de 0,87 (pour plus d’informations sur l’utilisation de modèle de langage pré-entraîné en sciences sociales, voir Chapitre 2, section 2.2.3).

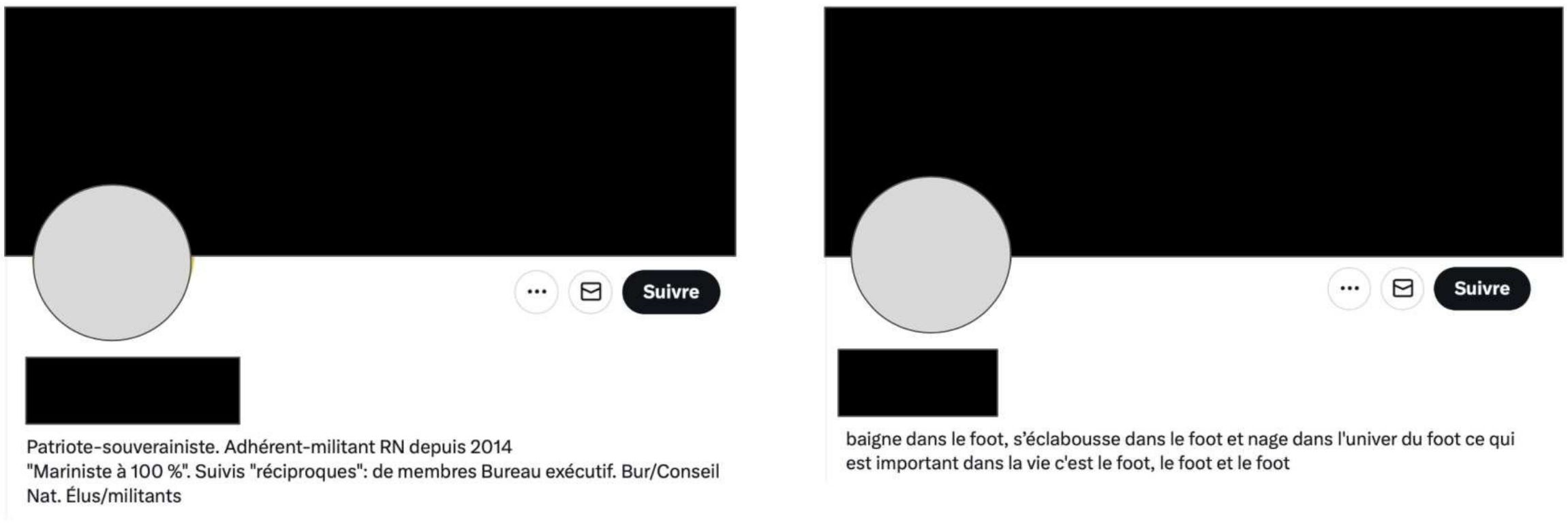

*Figure 3.6. Exemples de bio Twitter avec (à gauche) et sans (à droite) marqueur politique*

Les résultats montrent que les utilisateurs qui relaient des *fake news* sont deux fois plus susceptibles que les autres (e.g. 42,31 % contre 19,15 % ; $\chi^2$ (1, N=2,865)=73.709, $p<.001$, $\phi=0.16$) d’afficher leurs opinions politiques ou leur affiliation partisane dans leur bio (cf. Figure 3.7). Par exemple, des hashtags et mots-clés désignant des candidats et des partis politiques du Rassemblement National, de Reconquête ou de La France Insoumise (e.g, #RN2022, #Marine2022, #Zemmour, #JLM2022, #LFI2022) étaient très fréquents dans les bios des relayeurs de *fake news*. Un certain nombre d’entre eux semblaient néanmoins préférer indiquer leur soutien à des courants de pensée généraux, tels que le patriotisme ou le féminisme, ou à des mouvements sociaux (e.g. les Gilets Jaunes ou la communauté LGBTQ+), plutôt que leur appartenance à des partis politiques précis. Plusieurs comptes arboraient également le hashtag #TeamPatriote dans leur bio, faisant ainsi référence au groupe Télégram du même nom, créé pour permettre aux utilisateurs dont les comptes sont fréquemment suspendus de discuter en privé à l'abri de toute modération de contenu.[100]

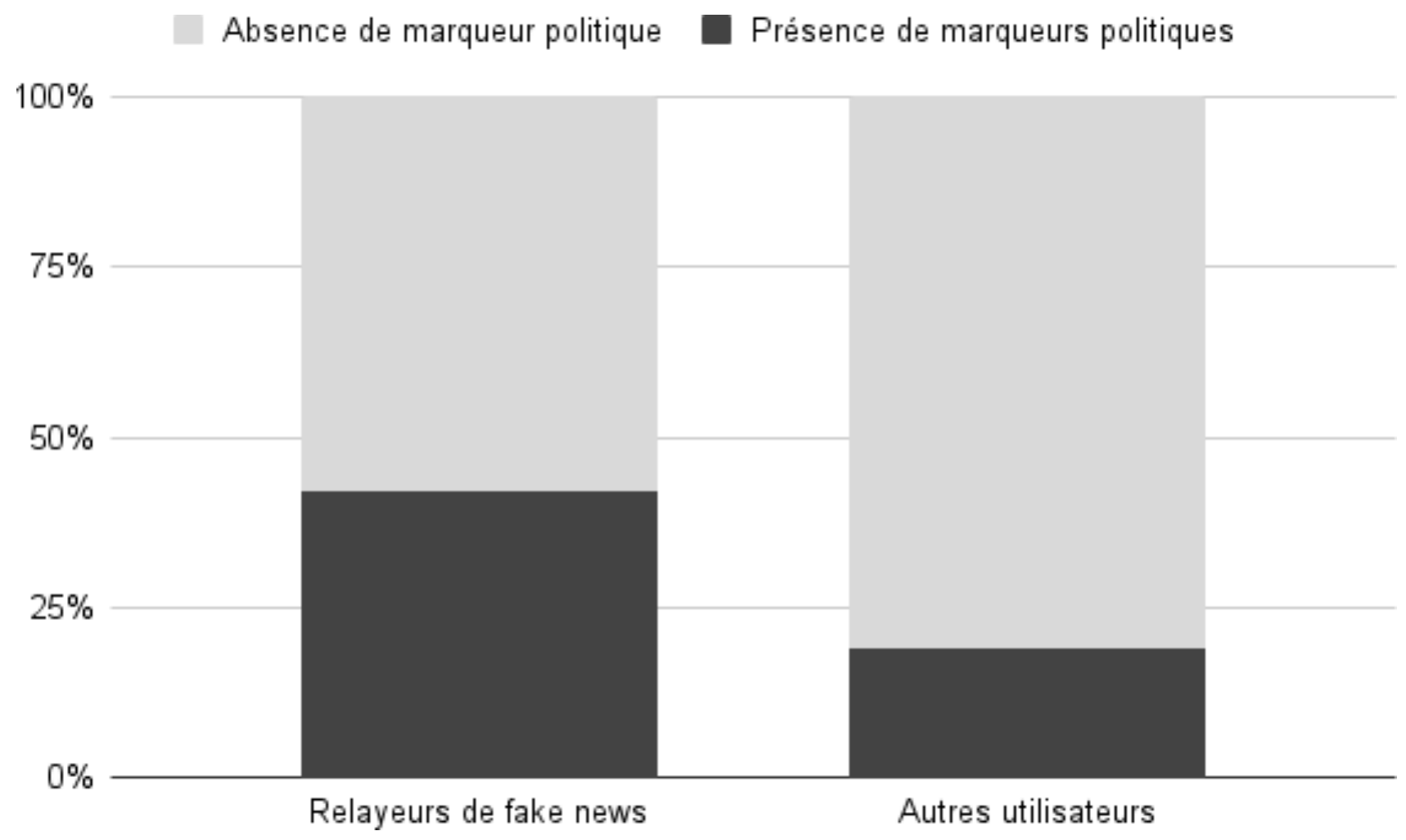

*Figure 3.7. Proportion de bio Twitter avec et sans marqueur politique selon que les comptes aient relayé ou non au moins une fake news*

La plus grande propension des relayeurs de *fake news* à mettre en avant leur affiliation partisane ou leurs opinions politiques dans la présentation qu'ils font d'eux sur Twitter ressort de façon encore plus frappante lorsqu'on la met en perspective avec celle d'un échantillon représentatif de l'ensemble des utilisateurs américains de Twitter. En effet, une étude du Pew Research Center montre que seulement 6 % des adultes américains présents sur Twitter incluent dans leur bio des mots suggérant une opposition ou un soutien à un parti politique, une idéologie, une personnalité, une organisation ou un mouvement politique (Widjaya, 2022).

[100] Laurent, S. (2020, 21 juillet). Plongée dans la haine en ligne avec « Team patriote », groupe privé de militants nationalistes. *Le Monde*. https://www.lemonde.fr/societe/article/2020/07/21/team-patriote-plongee-dans-la-haine-en-ligne_6046790_3224.html

Les observations en ligne ont par ailleurs permis d'examiner si les pratiques politiques des relayeurs de *fake news* étaient plus intenses que celles des autres utilisateurs. Twitter offre en effet un terrain d'enquête numérique privilégié pour analyser les comportements politiques des individus et la façon dont ils s'approprient de nouveaux outils de communication pour exprimer leurs opinions politiques (Jungherr, 2014). L'analyse des tweets postés par les relayeurs de *fake news* révèle que la majorité d'entre eux se concentrent principalement sur l'actualité politique, avec une focalisation sur des thèmes récurrents tels que l'immigration, la délinquance ou la fraude fiscale. Certains comptes montrent une insistance marquée sur ces sujets, allant jusqu'à relayer plusieurs fois d'affilée des titres médiatiques très similaires ou des commentaires portant sur un même événement. En comparaison, les publications des utilisateurs n'ayant pas relayé de *fake news* sont plus variées : elles ne concernent pas seulement des sujets politiques, mais sont également liées au divertissement, au sport ou à d'autres centres d'intérêt.

Les observations ont également permis de capturer certaines pratiques militantes des utilisateurs de Twitter. En effet, extrêmement actifs sur le réseau social, de nombreux relayeurs de *fake news* ont tendance à poster de multiples photos ou captures d'écran permettant de saisir des formes classiques de participation politique comme la signature de pétitions, le collage d'affiches, la participation à des manifestations, l'adhésion à un parti ou à une association (Boyadjian, 2016 ; Greffet et al., 2014 ; Oser et al., 2013).

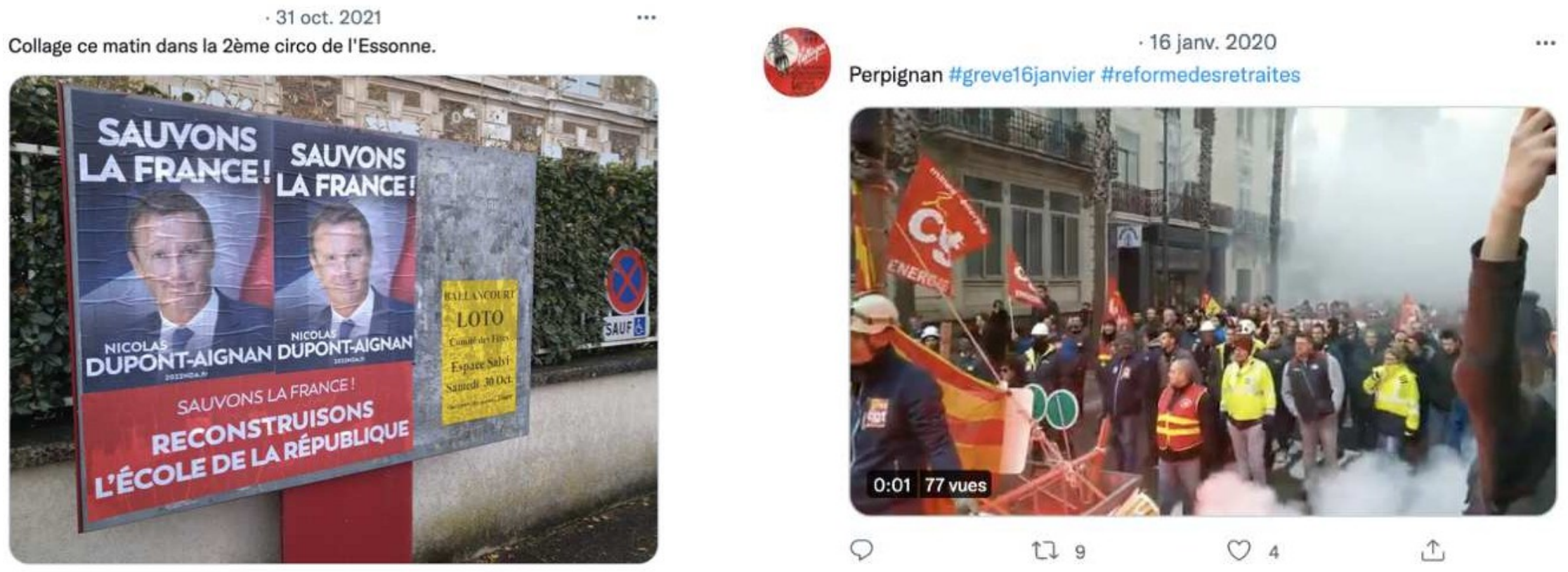

*Figure 3.8. Exemples de posts Twitter ayant permis d'observer certaines pratiques militantes*

L’analyse des bios des comptes, couplée aux observations réalisées sur les profils Twitter des utilisateurs, a fait émerger plusieurs modes de participation politique et d’affiliation partisane. Par exemple, tandis que certains se positionnent comme des électeurs votant par défaut, en attente d’un représentant convaincant, d'autres expriment une forme de désillusion totale à l’égard du système politique. Quelques utilisateurs indiquent également s’ils sont membres d’un parti ou s’ils exercent une fonction d’élu. En prenant appui sur différents modèles d’affiliation partisane (Duverger, 1951 ; Scarrow, 2014 ; Gibson et al., 2018), nous avons ainsi pu distinguer 5 types de profils en fonction de leur degré d’adhésion et d’identification envers un parti politique : (1) élu ; (2) adhérent ; (3) militant ; (4) indépendant ; (5) peu politisé. Cette classification a ensuite été étendue de façon systématique à l’ensemble des 124 comptes analysés dans le cadre de l’ethnographie en ligne à l’aide de l’outil de recherche avancé de Twitter (pour plus d’informations sur l’utilisation de cet outil, voir l’encadré de la section 2.2.3 du Chapitre 2). Plusieurs mots clés comme « cotisation » ; « membre » ; « encarté » ; « adhésion » ; « élection » ; « abstention » ; « vote » ont en effet permis d’estimer si un compte était affilié à un parti politique particulier ou au contraire s’il ne se reconnaissait dans aucun parti (cf. Figure 3.9).

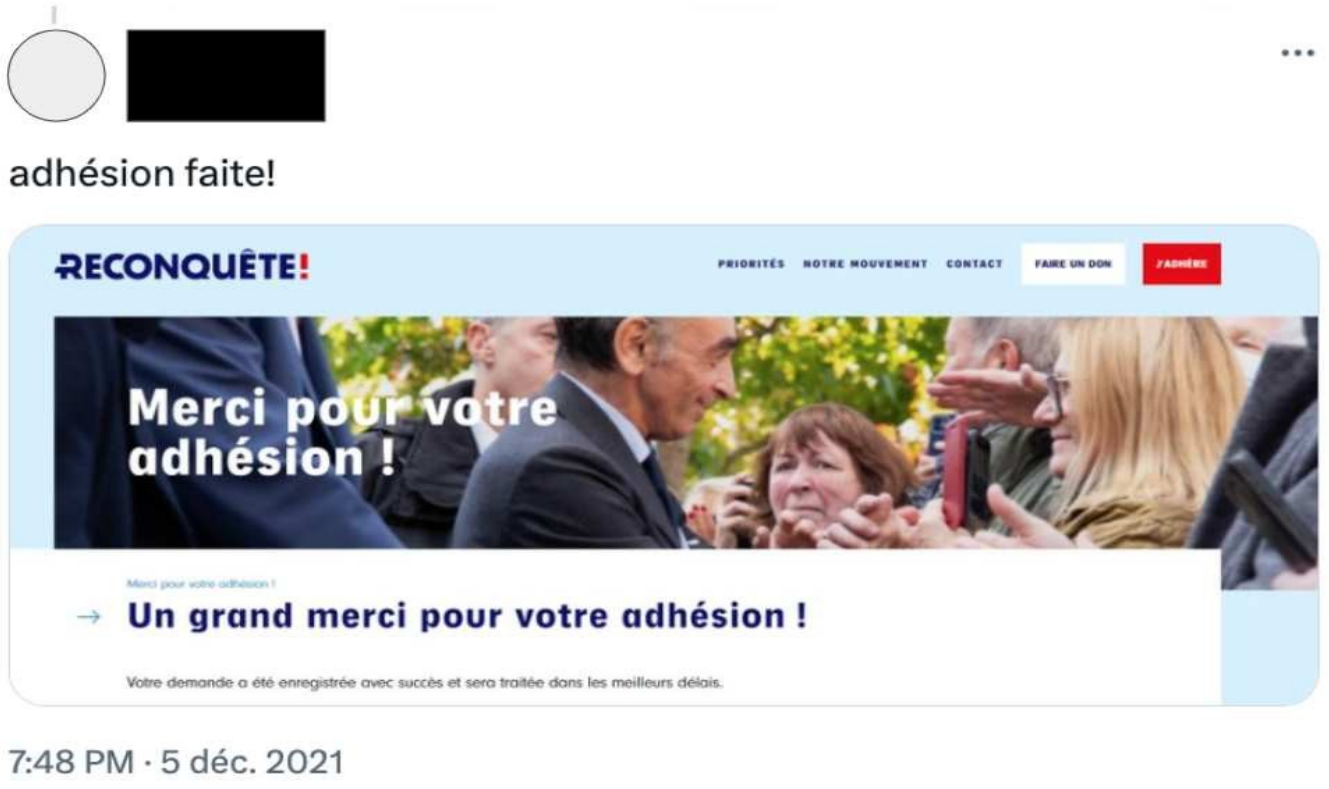

*Figure 3.9. Exemple de post Twitter ayant permis d’estimer le mode d’affiliation partisane d’un compte*

Cette diversité de profils met en lumière l'importance de prendre en compte les multiples nuances dans l'engagement politique des utilisateurs au-delà de leur position sur différents axes (droite/gauche ou pro/anti-élite). Examiner ces différents niveaux d'implication pourrait permettre de mieux comprendre les dynamiques qui sous-tendent le partage de *fake news*. Par exemple, dans la mesure où nous avons vu dans la section précédente que le partage de *fake news* était beaucoup plus fortement associé au fait d'avoir des positions anti-élites qu'à celui de se situer aux extrêmes de l'échiquier politiques, nous pouvons faire l'hypothèse que les relayeurs de *fake news* dépourvus d'attache partisane partageront plus de *fake news* que ceux qui endossent un rôle d'élu ou qui sont affiliés à un parti politique. Cette question sera approfondie dans le chapitre suivant afin d'enrichir notre compréhension des dynamiques sociales et politiques qui contribuent à favoriser (ou à inhiber) le partage de *fake news*.

Les entretiens ont aussi confirmé que les usages de Twitter des relayeurs de *fake news* étaient étroitement liés à leur engagement politique. Plusieurs enquêtés par exemple ont indiqué s'être inscrits sur Twitter pour des raisons politiques :

> *Je me suis inscrit vers 2014-2015 ou 2016. C'est un twitter politique. En fait, mon twitter est uniquement politique. Mon usage est revendicatif et surtout pour m'informer. Moi je ne suis pas sur Twitter pour parler de truc de machin, moi c'est vraiment un instrument politique.*[101]

Au cours des entretiens, il est également apparu très clairement que plusieurs participants avaient été exposés à des influences politiques très jeunes. Par exemple, parmi les personnes interrogées, un participant a raconté avoir été profondément marqué par la guerre du Golfe alors qu'il n'avait que 12-13 ans[102] ; une enquêtée a précisé avoir été impliquée avec la Jeunesse Ouvrière Chrétienne à l'âge de 14-15 ans[103] ; et un troisième répondant a partagé des souvenirs de ses premières manifestations étudiantes :

> *Oui je me souviens de mes premières manifs étudiantes contre le CEP ou un truc comme ça c'était à l'époque. C'était un contrat éducation je ne sais plus trop quoi. [...] Nous on s'était battu à l'époque contre ça, les emplois jeunes, les choses comme ça. Les emplois au rabais. À l'époque on appelait ça comme ça. Et ouais, j'étais mineur encore. C'était avant mon bac. Ça devait être mes années de seconde, première. Oui j'étais… À l'époque, j'étais délégué de classe, j'étais délégué d'établissement. Et ouais bah pareil on avait des réunions avec des trucs étudiants à l'époque.*[104]

---

[101] Homme, 39 ans, Bac +3, infirmier, entretien réalisé le 20 mai 2022.
[102] Homme, 45 ans, Bac +4, fonctionnaire, entretien réalisé le 18 mai 2022.
[103] Femme, 67 ans, certificat d'étude, retraitée, entretien réalisé le 19 mai 2022.
[104] Homme, 47 ans, Bac +2, inactif, en situation de handicap, entretien réalisé le 28 février 2022.

Cette politisation précoce est souvent attribuable à la socialisation familiale. Par exemple, un enquêté a mentionné avoir grandi dans une famille d'extrême droite « depuis le début de [sa] vie », ajoutant que son père était « l'un des premiers militants du Front National en France ».[105] Pour d'autres participants, cependant, c'est la socialisation professionnelle qui semble avoir joué un rôle plus déterminant. Ainsi, un autre enquêté, fils d'instituteur, a partagé un souvenir de jeunesse, racontant qu'il avait « collé des affiches avec [son] père avant [ses] 18 ans pour Mitterrand ». Bien qu'il ait apprécié cette implication à l'époque, il nous a confié que cela ne l'avait pas empêché de développer par la suite des opinions politiques différentes de celles de son père. À la question de savoir ce qui l'avait fait basculer à droite, il a répondu sans détour : « parce que je ne suis pas devenu fonctionnaire ».[106]

L'importante politisation des utilisateurs laisse ainsi penser que les opinions politiques qu'ils détiennent sont ancrées en eux depuis longtemps. Dans cette perspective, il est plausible que les *fake news* renforcent davantage des opinions préexistantes qu'elles n'en provoquent de nouvelles. Les individus semblent aborder l'information avec des convictions antérieures bien établies, et les *fake news* servent probablement à consolider ces croyances plutôt qu'à les modifier ou à les infléchir.

En somme, les analyses conduites dans cette section montrent que les utilisateurs qui relaient des *fake news* sont plus politisés que les autres utilisateurs de Twitter. Il est important de rappeler qu'ils ont été comparés avec un groupe de contrôle composé d'utilisateurs ayant une distribution similaire dans l'espace idéologique latent et ayant également partagé au moins une information issue d'un média français. Autrement dit, cela signifie que les utilisateurs du groupe de contrôle sont déjà plus politisés et intéressés par l'actualité que la moyenne. Ainsi, le partage de *fake news* n'est pas la conséquence d'un manque de compétence ou d'intérêt pour la politique, mais au contraire d'un fort intérêt pour la politique et d'un surinvestissement de l'espace politique.

[105] Homme, 23 ans, Bac +4, étudiant, entretien réalisé le 5 décembre 2022.
[106] Homme, 56 ans, Bac +2, consultant, entretien réalisé le 19 novembre 2022.

### 3.2.2. Des pratiques de partage d'informations très élevées malgré une défiance importante dans les médias *mainstream*

Un deuxième ensemble de variables a servi à examiner les pratiques de partage d'informations des relayeurs de *fake news* par rapport à celles des autres utilisateurs de Twitter. Tout d'abord, nous avons mesuré le volume de tweets contenant au moins un lien vers un site externe pour étudier dans quelle mesure les utilisateurs ont posté des contenus web non reliés à des informations d'actualité. Puis, nous avons calculé le volume de tweets contenant au moins un lien redirigeant vers l'un des 420 sites de médias répertoriés dans notre base de données pour examiner la propension des utilisateurs à publier des contenus issus de l'écosystème médiatique français. Les résultats suggèrent que les relayeurs de *fake news* sont de plus gros consommateurs d'informations que les autres utilisateurs. En effet, 23,3 % de leurs tweets contiennent une URL redirigeant vers un média français alors que ce n'est le cas que pour 13,7 % des autres utilisateurs. En revanche, aucune différence notable n'a été relevée au niveau des autres URLs partagées, correspondant à des sites non reliés à l'information d'actualité (e.g. plateforme de jeux, sites de *e-commerce*, etc.). Ces résultats suggèrent que le partage de *fake news* n'est pas forcément associé à un manque d'accès à l'information d'actualité mais au contraire à un intérêt marqué pour celle-ci.

Pour aller plus loin, nous avons ensuite analysé les types de sources médiatiques partagées par les utilisateurs de notre échantillon. Pour ce faire, nous avons de nouveau utilisé la classification des sources médiatiques réalisée par une étude du médialab de Sciences Po sur la structure de l'écosystème médiatique français (Cardon et al., 2019 ; Cointet et al., 2021). Cela nous a permis d'évaluer la part de médias *mainstream*, de médias partisans, de médias de contre-information et de médias de presse locale ou de tabloïd relayés par chacun de nos deux groupes (cf. Tableau 3.3).

| Measures | Variables | Relayeurs de *fake news* | Autres utilisateurs | $\chi^2$ | p-value |
|---|---|---|---|---|---|
| Type de contenu | URL issue d'un média | 23,33 % | 13,68 % | 24087.989 | < .0001 |
| | autre URL | 26,55 % | 28,42 % | | |
| | aucune URL | 50,12 % | 57, 90 % | | |
| Type de source | média mainstream | 52,14 % | 62,44 % | 3898.738 | < .0001 |
| | média partisan | 19,47 % | 15,63 % | | |
| | média de contre-information | 14,79 % | 7,80 % | | |
| | tabloïd/ presse locale | 13, 60% | 14,13% | | |

*Tableau 3.3. Répartition des types de contenus et de sources partagés par les relayeurs de fake news et les autres utilisateurs*

Les résultats montrent que les relayeurs de *fake news* partagent une proportion de médias de contre-information environ deux fois plus élevée que le groupe de contrôle (14,79 % contre 7,80 %) ainsi qu'un peu plus de médias partisans (19,47 % contre 15,63 %). En revanche, ils partagent moins de médias *mainstream* (52,14 % contre 62,44 %) et on observe peu de différence en ce qui concerne la presse locale et les tabloïds (cf. Tableau 3.3). Ces constats suggèrent que les utilisateurs qui relaient des *fake news* sont plus méfiants que les autres à l'égard des médias *mainstream* et cherchent à les court-circuiter en partageant des sources alternatives. Toutefois, ils ne semblent pas pour autant déconnectés des médias *mainstream* dans la mesure où ceux-ci représentent plus de la moitié des médias qu'ils partagent. Ce constat nuance l'idée d'un enfermement complet dans des bulles de filtres se traduisant par une coupure et un isolement : bien que les relayeurs de *fake news* privilégient des sources alternatives, ils maintiennent une exposition importante aux sources d'information centrales.

Ces résultats sont cohérents avec ceux d'autres travaux ayant montré que les individus polarisés consomment souvent de grandes quantités de médias *mainstream* (Garrett, 2009 ; Weeks et al., 2016) et ne se désabonnent pas systématiquement des sources offrant des perspectives contradictoires à leurs propres opinions (Ross Arguedas et al., 2022). Cela montre que même les publics polarisés restent exposés au débat public central. En cela, ils n'illustrent pas un phénomène d'évitement sélectif strict.

Ce constat d'une défiance, couplée à une forte consommation d'informations médiatiques, a été confirmé lors des entretiens. Par exemple, plusieurs enquêtés ont déclaré qu'ils s'étaient inscrits sur Twitter pour partir à la recherche de sources d'information alternative.

> *Alors c'était en 2009. C'était… En fait, je suivais beaucoup à cette époque le combat qu'il y avait à Notre Dame des Landes pour la construction de l'aéroport. Et donc voilà, je cherchais un peu des médias un peu plus indépendants que ce que je voyais à la télévision qui était assez uniforme dans l'idée que voilà c'était des méchants zadistes qui occupaient un terrain, qu'il y avait une décision de justice. J'ai appris que c'était pas vraiment ça. Y avait vraiment un écart énorme entre ce que je voyais à la télé et ce qui se passait sur le terrain. Je me suis dit : j'ai besoin d'une autre source d'information en fait, différente de ce que je voyais à la télévision, de ce que je pouvais lire dans les journaux classiques. C'était surtout ça. À l'époque j'étais encore sur Facebook et donc il y avait des gens qui partageaient des liens Twitter et donc je me suis dit je vais aller voir. Voilà quoi. [...] Aujourd'hui avec Internet, on a quand même accès à beaucoup beaucoup de choses quoi, donc euh… je trouve que c'est vraiment dommage de se contenter de ce qu'on nous sert, de pas aller chercher plus loin. Et je trouve injuste le traitement de l'information comparativement à ceux qui se battent sur le terrain et qui voient la réalité des choses et comment on les expose, comment on les montre ou comment on les décrit. Et voilà, pour moi, c'est une injustice. Et donc de ce côté-là les médias indépendants ont un rôle important à jouer.*[107]

> *En fait, moi j'utilise essentiellement Twitter et Telegram pour m'informer parce que je ne crois pas un mot [en mettant de l'emphase] de ce qui est écrit dans les journaux et pas un mot de ce qui est dit à la télévision. C'est une désinformation invraisemblable. Je n'ai pas la télévision. Ça doit faire… Déjà je n'avais pas la télévision quand j'ai quitté mes parents à votre âge à peu*

[107] Homme, 47 ans, Bac +2, inactif, en situation de handicap, entretien réalisé le 28 février 2022.

*près. J'avais pas la télévision dans mon studio. Et puis je n'ai pas senti la nécessité d'avoir la télévision.*[108]

De façon intéressante, lorsque le sujet des *fake news* a été abordé avec les enquêtés plusieurs ont répondu que leurs principaux instigateurs étaient les médias *mainstream*.

*Tous les gens qui accusent de fake news pour moi c'est des gens suspects de mensonge.*[109]

*Dans le régime UMP-PS les gens qui parlent de fake news sont eux-mêmes des sources de désinformation et de mise en biais des discours …*[110]

Ce contre-emploi de la notion de *fake news* montre que celle-ci est considérée comme une étiquette disqualifiante et stigmatisante. L'accusation de croire ou de propager des *fake news* apparaît alors comme une forme de marqueur spécifique de la compétition politique. Les tensions sur la définition du mot *fake news* reflètent une lutte pour détenir le monopole de la vérité. Cette bataille pour déterminer ce qui est vrai ou faux sert avant tout à renforcer la légitimité de son propre camp, soulignant les enjeux de pouvoir sous-jacents à ce discours.

Malgré leur importante défiance à l'égard des médias *mainstream*, les relayeurs de *fake news* n'en sont pas pour autant déconnectés. En effet, la majorité des contenus médiatiques qu'ils partagent sont issus de ces médias. Certains sont même abonnés au *Monde* et au *Figaro* et à *The Economist*.

*À une époque j'étais abonné au Figaro, au Monde, à The Economist et j'achetais quelques magazines papiers, et en fait depuis 3 ans je ne suis plus abonné parce que le Monde et le Figaro ont merdé avec les reconductions d'abonnement par carte bleue et donc j'ai laissé tomber. Et The Economist, je vais peut-être me réabonner, mais enfin ils sont sur une ligne éditoriale qui m'énerve. Ils ont des avantages mais ils ont aussi des inconvénients. Ils ont l'avantage d'apporter des faits, des graphiques et des données mais ils sont sur une ligne éditoriale très néolibérale qui est assez énervante.*[111]

*Disons qu'en règle générale je les regarde quand même, je les suis quand même. Je ne vais pas dire qu'ils disent des mensonges. Des fois c'est dans le traitement de l'information. L'information*

[108] Femme, 70 ans, Bac +8, retraitée, ex chercheuse, entretien réalisé le 15 mars 2022.
[109] Homme, 64 ans, Bac +6, structureur financier, entretien réalisé le 23 mars 2022.
[110] Homme, 55 ans, Bac +8, à son compte, ex chercheur, entretien réalisé le 6 mars 2022.
[111] Homme, 55 ans, Bac +8, à son compte, ex chercheur, entretien réalisé le 6 mars 2022.

*qu'ils donnent n'est pas forcément mauvaise, après c'est dans le traitement, la façon de traiter le sujet, des choses comme ça. Mais j'ai quand même besoin… On a quand même besoin des médias mainstream quoi. Après si on veut aller plus loin dans la réflexion ou dans la recherche des causes et tout ça, il vaut mieux faire ses recherches tout seul et avoir des points de vue différents qu'un seul en fait. Eux ils traitent les sujets que d'un angle. En fait c'est ça le problème des médias mainstream c'est qu'ils n'ont qu'un angle de traitement sur un sujet, ils n'en auront pas deux, ils ne vont pas confronter deux idées. C'est assez rare on va dire qu'ils opposent deux idées et qu'ils ouvrent un débat parce que généralement ils ont une ligne éditoriale et ils la tiennent quoi. C'est ça le problème. Mais après dans ce qu'ils racontent au départ, ils sont quand même sources des actualités.*[112]

En résumé, ces résultats montrent que le partage de *fake news* n'est pas associé à un manque d'accès à l'information d'actualité mais au contraire à des pratiques actives de partage d'informations. Loin d'être confinés dans des « chambres d'échos », les relayeurs de *fake news* s'exposent à des médias également consommés par la majorité des publics. Non seulement ils n'ignorent pas les informations qui contredisent leurs opinions, mais certains les recherchent explicitement pour mieux les critiquer. Leur recours à des sources alternatives n'est donc pas motivé par un déficit d'information, mais par une défiance vis-à-vis des récits promus par les médias dominants. Autrement dit, le partage de *fake news* n'est pas symptomatique d'un manque de connaissances, mais d'une crise de confiance à l'égard de l'information institutionnalisée, en particulier celle des médias *mainstream*. Au fond, la véritable question à se poser pour les chercheurs et chercheuses n'est pas tant : « Pourquoi adhèrent-ils aux *fake news* ? », mais plutôt « Pourquoi sont-ils si méfiants envers les médias traditionnels ? » Ce glissement de perspective permet de réorienter l'analyse vers les dynamiques de méfiance et de contestation, en interrogeant la relation complexe entre ces utilisateurs et les institutions médiatiques, au-delà d'une simple explication par un déficit de discernement ou de compétences informationnelles.

[112] Homme, 47 ans, Bac +2, inactif, en situation de handicap, entretien réalisé le 28 février 2022.

### 3.2.3. Des comptes hyperactifs mais peu expressifs et interactifs

Un troisième ensemble de variables a été construit pour comparer les usages de Twitter des relayeurs de *fake news* à ceux des autres utilisateurs. La propension des comptes à publier ou à partager des tweets sur une période de temps donnée a tout d'abord été évaluée en calculant leur nombre médian de tweets par jour. Les résultats montrent que les relayeurs de *fake news* sont nettement plus actifs sur Twitter que les utilisateurs du groupe de contrôle, avec un nombre médian de 22 tweets par jour contre moins d'un tweet par jour pour les autres (cf. Tableau 3.4). Cette activité est vraiment hors-norme dans la mesure où la moitié des utilisateurs de Twitter publie moins de 5 tweets par mois (Odabaş, 2022).

Ensuite, le nombre de comptes ayant publié ou partagé au moins 10 tweets consécutifs en moins d'une heure a été mesuré afin de construire un indicateur de « bombardement ». Ici encore, les relayeurs de *fake news* se distinguent très fortement, avec 108 comptes concernés contre seulement 15 parmi les autres utilisateurs. Il ressort donc que le rythme de publication des relayeurs de *fake news* est beaucoup plus soutenu que celui des autres utilisateurs. Ces constats rejoignent les conclusions de nombreuses études qui montrent que les utilisateurs les plus actifs en termes de volume de publications sont souvent les plus militants et les plus radicaux. Par exemple, le Pew Research Center a souligné qu'un petit groupe d'utilisateurs très politisés est responsable d'une part disproportionnée du contenu politique sur Twitter : alors que ceux qui se déclarent « très démocrates » ou « très républicains » ne représentent que 6 % des comptes, ils génèrent à eux seuls 20 % de l'ensemble des tweets et 73 % des tweets portant sur la politique nationale (Wojcik et Hughes, 2019).

| Measures | Variables | Relayeurs de *fake news* | Autres utilisateurs | W | p value |
|---|---|---|---|---|---|
| Activité | tweets_per_day | 22.2 | 0.9 | 573354 | <.0001 |
| Mode d'énonciation | ratio_quote | 0.01 | 0.02 | 2332206 | .302 |
| | ratio_retweet | 0.78 | 0.58 | 2822832 | <.0001 |
| | ratio_hashtag | 0.36 | 0.33 | 2735082 | .001 |
| Interactivité | ratio_reply | 0.08 | 0.14 | 2590114.5 | <.0001 |
| | ratio_mention | 0.13 | 0.23 | 2675266.5 | <.0001 |

*Tableau 3.4. Comparaison des usages de Twitter des deux échantillons et résultats des tests de Wilcoxon-Mann-Whitney*

Dans un deuxième temps, un indicateur d'interactivité a été construit afin d'évaluer la propension des utilisateurs étudiés à échanger avec d'autres comptes soit en répondant à leurs tweets (*reply*), soit en les mentionnant lors de la rédaction d'un tweet (*mention*), notamment en ayant recours au sigle @ qui est un marqueur d'interaction (Honeycutt et Herring, 2009). Les résultats montrent que les relayeurs de *fake news* sont presque deux fois moins enclins à interagir avec d'autres utilisateurs que ceux qui n'ont partagé aucune *fake news*. En effet, seulement 8 % des tweets des relayeurs de *fake news* sont des réponses adressées à d'autres comptes et seulement 13 % mentionnent d'autres utilisateurs, tandis que ces proportions sont respectivement de 15 % et 23 % pour les tweets des utilisateurs n'ayant pas partagé de *fake news*. En somme, les relayeurs de *fake news* semblent préférer publier des tweets à la cantonade à un public indéterminé plutôt que de s'adresser à des destinataires précis pour engager une conversation. Cela suggère qu'ils cherchent avant tout à faire entendre leur voix et à se positionner dans l'espace public sans nécessairement vouloir s'engager dans des interactions directes et interpersonnelles. Cette posture rappelle celle des « égo-visibles » sur Facebook (Bastard et al., 2017) qui n'utilisent pas les réseaux sociaux pour créer un espace de conversation large et ouvert, mais plutôt pour afficher leur identité et partager des contenus sans s'investir pleinement dans des échanges interactifs qu'ils ne contrôlent pas. Cette faible interactivité souligne une dynamique expressive particulière chez les relayeurs de *fake news* : ils semblent moins intéressés par le dialogue que par la diffusion de leurs opinions, ce qui traduit une conception unilatérale de la communication en ligne, plus proche de la proclamation que de la conversation.

Enfin, différentes variables ont permis d'analyser les modes d'énonciation des utilisateurs étudiés. En effet, trois possibilités s'offrent aux utilisateurs qui souhaitent mettre en circulation des énoncés sur Twitter : ils peuvent soit produire leurs propres tweets ; soit citer ceux des autres en y ajoutant un nouveau message ; soit les retweeter simplement sans exprimer de nouveau message (cf. Figure 3.10). Dans une certaine mesure, ces trois modes d'énonciation que constituent les *tweets*, les *quotes* et les *retweets* permettent d'analyser la propension des utilisateurs à assumer la responsabilité des énoncés qu'ils mettent en circulation. En effet, en retweetant un contenu, ils ne se positionnent ni comme des sujets de l'énonciation, ni comme des locuteurs, mais comme des simples relais d'une information ou d'une opinion, tandis qu'en écrivant leurs propres tweets ils sont responsables des énoncés mis en circulation. L'analyse de ces différents modes d'énonciation sera approfondie dans le prochain chapitre, notamment en mobilisant les distinctions conceptuelles apportées par Oswald Ducrot (1984) entre sujet parlant, locuteur et énonciateur.

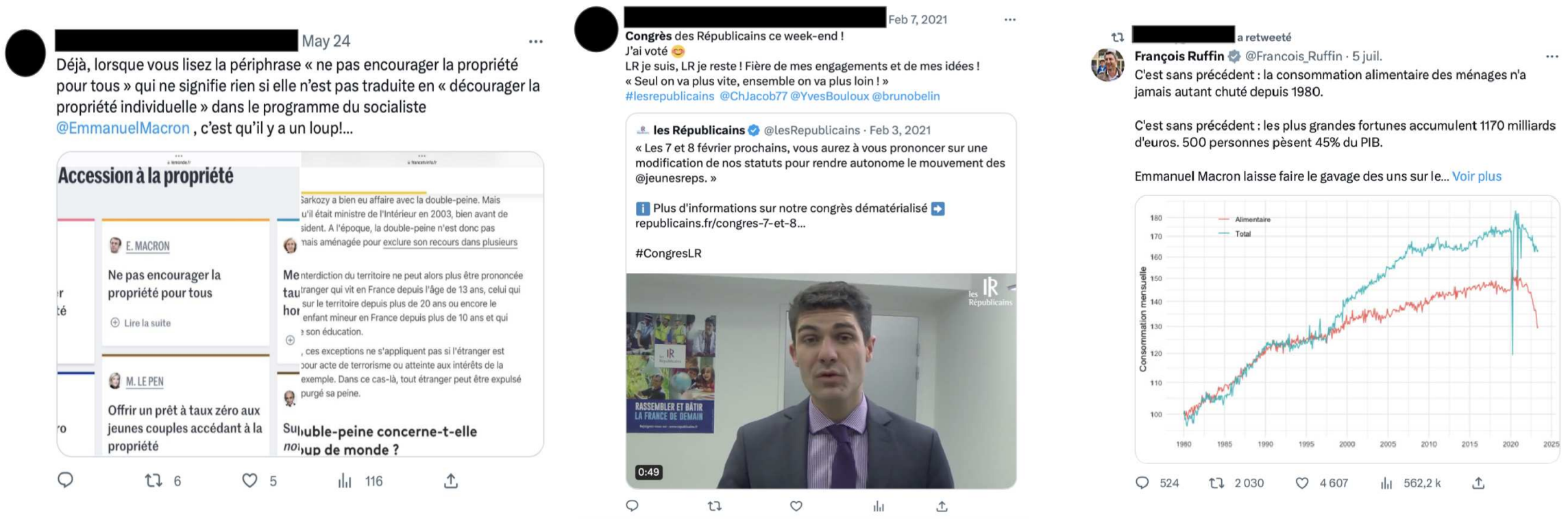

*Figure 3.10. Les différents modes d'énonciation possibles sur Twitter : tweet ; quote ; retweet*

En calculant la proportion de *tweets*, *quotes* et *retweets* émis par chaque utilisateur, on observe que les tweets originaux ne représentent que 13 % des publications des relayeurs de *fake news* contre 27 % de celles des autres utilisateurs, et qu'environ 79 % des publications des relayeurs de *fake news* sont de simples retweets contre 57 % de celles des autres utilisateurs. Très peu de différences ont cependant été observées au niveau des *quotes* qui représentent pour les relayeurs de *fake news* comme pour les autres utilisateurs un très faible pourcentage de l'ensemble de leurs publications (1 % contre 2 %). Par ailleurs, en regardant de plus près sous quel mode d'énonciation les *fake news* ont été mises en circulation, nous pouvons remarquer que la majorité ont été simplement retweetées (85 %), mais que seulement 15 % ont été initiées par un tweet. Il ressort ainsi que plutôt que de produire leurs propres énoncés pour exprimer leur opinion, les utilisateurs qui partagent des *fake news* ont surtout la particularité d'être des relayeurs d'informations très actifs et prolifiques. D'un côté, ce mode d'énonciation suggère une forme de diffusion rapide et peu réflexive, privilégiant la circulation d'informations déjà existantes. De l'autre, cela indique que les relayeurs de *fake news* ne prennent pas directement en charge les contenus qu'ils mettent en circulation sur Twitter (notamment les *fake news*). Dans une certaine mesure, le mode de diffusion par *retweet* peut alors être perçu comme une manière de se distancer des propos relayés, évitant ainsi toute responsabilité directe vis-à-vis de l'énoncé, en particulier lorsqu'il s'agit de *fake news*. Par ailleurs, bien que le *retweet* soit souvent interprété comme un acte d'adhésion à l'information partagée, cette équivalence n'est pas toujours évidente. Quelques utilisateurs prennent même le soin d'indiquer dans leur bio des expressions comme « RT is not endorsement ». Ces pratiques soulèvent des questions quant à la conscience des utilisateurs concernant le caractère contestable des informations qu'ils relaient et suggèrent que le recours au retweet pourrait être une tactique délibérée pour éviter d'être tenus directement responsables des propos partagés.

Une caractéristique majeure de relayeurs de *fake news* est donc de retweeter sans relâche. Ils retweetent à la chaîne – parfois jusqu'à 100 tweets par jour – des comptes de personnalités politiques, de polémistes réputés, de médias alternatifs, d'éditorialistes, etc. Un listing systématique des comptes retweetés par l'échantillon de relayeurs de *fake news* étudiés montre à quel point les discours de personnalités comme Jean Messiha, Eric Zemmour ou Nicolas Dupont-Aignan sont relayés par les internautes qui propagent des *fake news* sur Twitter (cf. Figure 3.11). Ces pratiques révèlent ainsi l'engagement de ces comptes à défendre l'agenda politique des *leaders* de leur camp et invitent à élargir les études sur les *fake news* par des analyses examinant plus largement comment des informations, même factuelles, peuvent être utilisées pour cadrer idéologiquement les débats et produire une représentation déformée de la réalité en mettant à l'agenda certains sujets d'actualité au détriment d'autres.

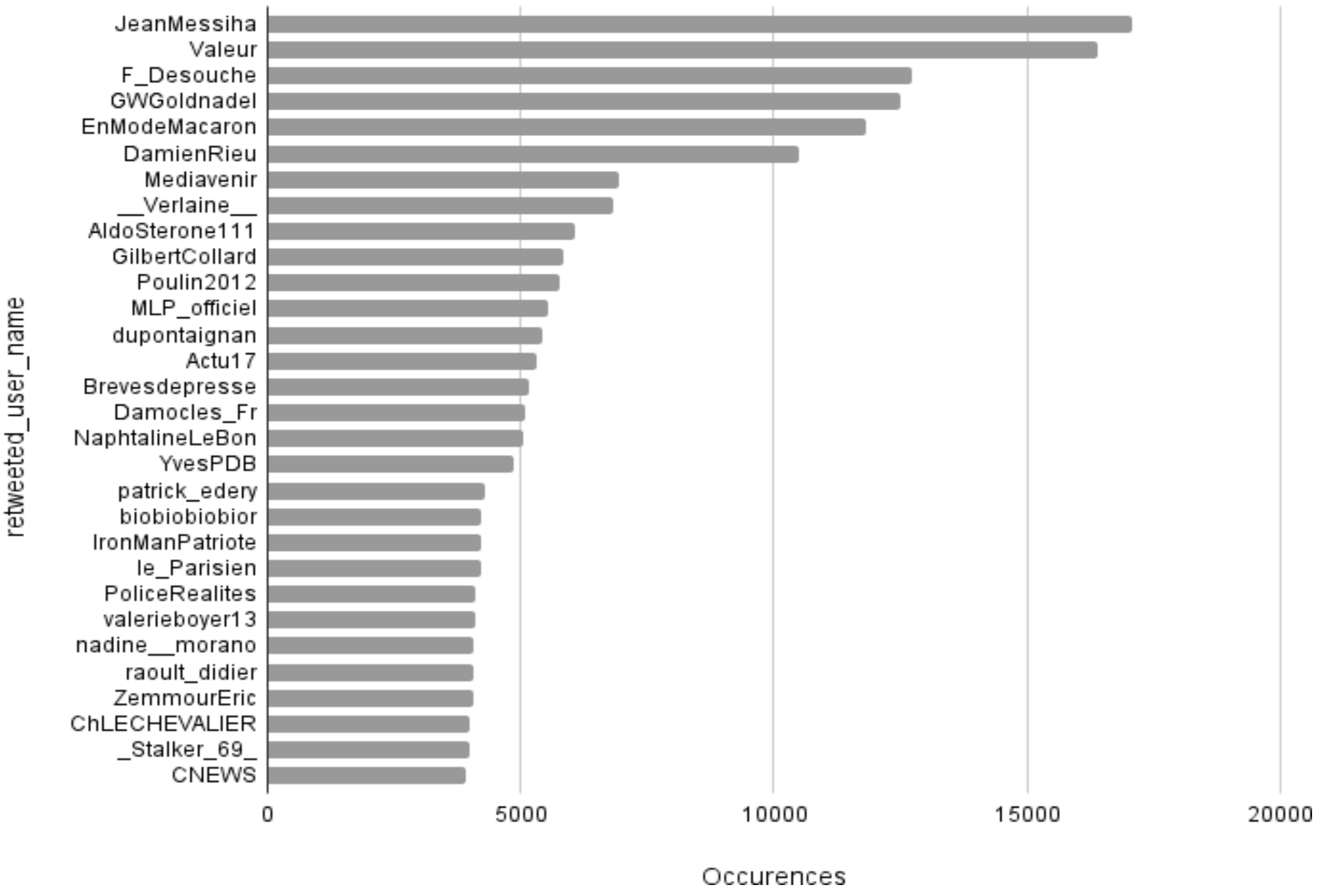

*Figure 3.11. Liste des 30 comptes les plus retweetés par les relayeurs de fake news*

En résumé, les résultats issus de cette section montrent que les utilisateurs qui relaient des *fake news* sur Twitter sont beaucoup plus actifs que les autres. Derrière leur hyperactivité, toutefois, ils semblent moins chercher à interagir avec d'autres utilisateurs qu'à lancer des tweets à la cantonade. Par ailleurs, ils ont recours à des modes d'énonciation leur permettant de ne pas être tenus pour responsables des énoncés qu'ils mettent en circulation, d'autant plus lorsqu'il s'agit de *fake news.*

### 3.2.4. Des comptes sous pseudonyme

Une quatrième variable a été introduite pour comparer la propension des relayeurs de *fake news* à masquer leur identité sous un pseudonyme par rapport aux autres utilisateurs de Twitter. 250 comptes de chacun des deux groupes ont été annotés de façon manuelle, soit 500 au total. Nous avons considéré qu'un utilisateur indiquait son nom complet lorsque figuraient sur son profil à la fois un prénom et un nom de famille. En revanche, lorsqu'un compte avait recours à un simple prénom, à un surnom, à un diminutif, à un nom d'objet, de fleur ou d'animal, ou encore à un adjectif qualificatif, nous avons considéré qu'il utilisait un pseudonyme. Par exemple, des noms d'utilisateur tels que « Pierre Durand » ou « Latifa Mansouri » auraient été classés comme des noms complets, tandis que « Révolutionnaire415 » ou « MusicJunkie87 » auraient été catégorisés comme des pseudonymes. Il est important de noter que certains comptes peuvent avoir choisi délibérément d'utiliser des prénoms et des noms de famille différents de leur identité civile, ce qui introduit une certaine marge d'erreur dans nos annotations. Pour étendre notre travail d'annotation manuelle à l'ensemble des comptes de notre échantillon, nous avons par la suite entraîné un classifieur basé sur le modèle CamemBERT. Les performances du classifieur se sont révélées très satisfaisantes avec un score F1 de 0,95 (pour plus d'informations sur l'utilisation de modèle de langage pré-entraîné en sciences sociales, voir Chapitre 2, section 2.2.3).

Alors que les deux-tiers (66 %) des internautes à l'origine du partage d'une *fake news* ont recours à un pseudonyme, ce n'est le cas que pour la moitié (52 %) des autres utilisateurs ($\chi^2$ (1, N=2,865)=38.195, $p$ <.001, $\phi$=0.12). Il ressort ainsi que les comptes Twitter qui sont à l'origine du partage de *fake news* ont plus tendance que les autres utilisateurs à dissimuler leur identité derrière un pseudonyme. Cet attachement des comptes à préserver leur anonymat s'est retrouvé au cours de plusieurs entretiens. Par exemple, trois comptes ont souhaité nous appeler sous un numéro masqué. Pour se justifier, l'un de nos enquêtés[113] nous a dit : « Je ne cède pas mon anonymat aussi facilement. Comme j'ai un pseudo, j'essaie de garder mon pseudo ». L'utilisation d'un pseudonyme peut être interprétée comme un moyen de diminuer la responsabilité liée à l'énonciation, tout en prenant des précautions vis-à-vis des répercussions potentielles associées à la diffusion de contenus en ligne. En dissociant les énoncés qu'ils partagent sur Twitter de leur identité civile, les relayeurs de *fake news* montrent qu'ils sont conscients que les contenus qu'ils mettent en circulation sont susceptibles d'être contestés par d'autres personnes et d'entacher leur réputation. Le recours au pseudonyme peut être vu comme une stratégie pour minimiser ces risques tout en continuant à participer activement au débat en ligne. En ce sens, ils font preuve d'une certaine « prudence énonciative », une pratique qui sera explorée plus en détail dans le chapitre

[113] Homme, 55 ans, Bac +8, à son compte, ex chercheur, entretien réalisé le 6 mars 2022.

suivant.

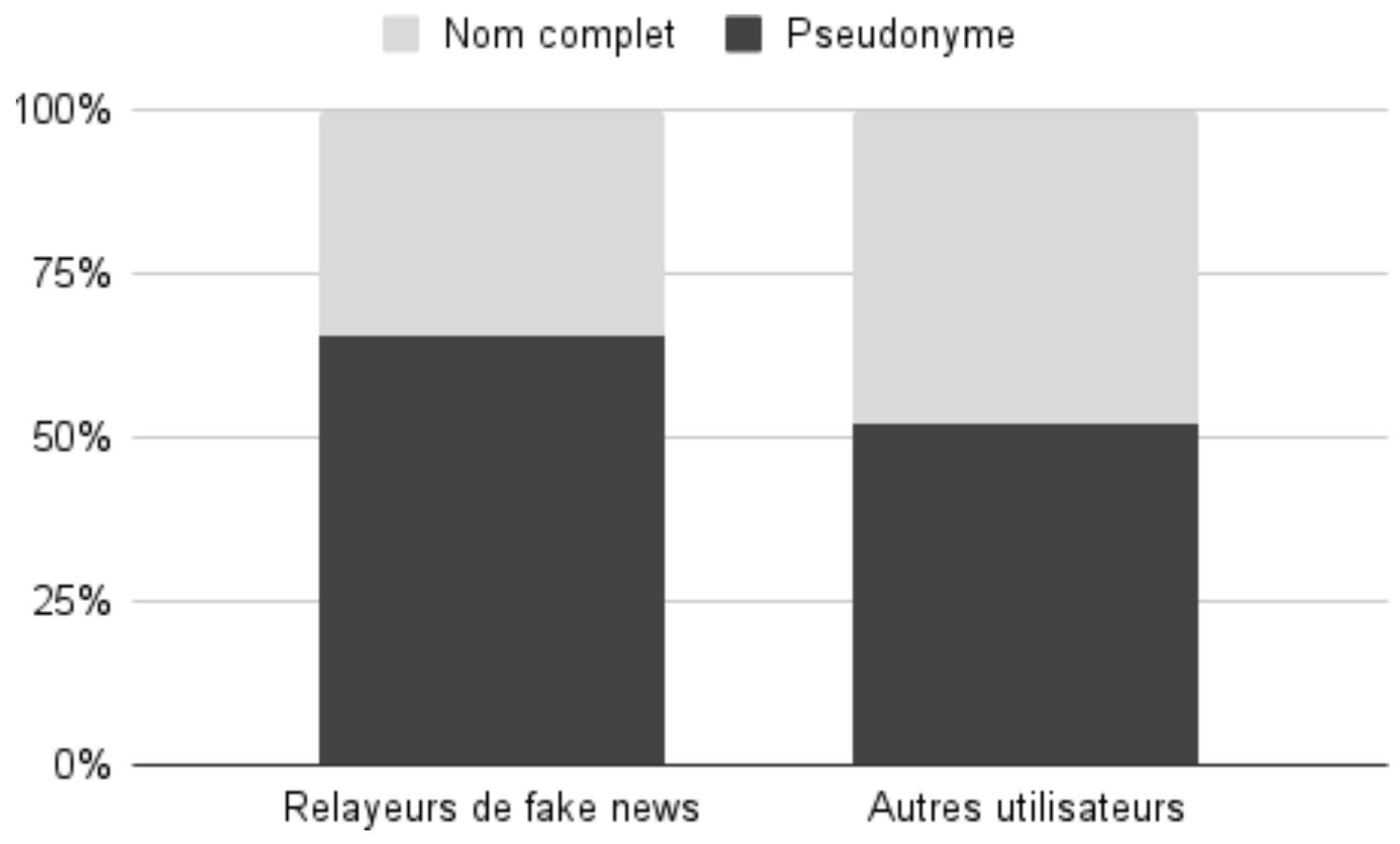

*Figure 3.12. Proportion de comptes Twitter indiquant un nom complet ou recourant à un pseudonyme selon que les comptes aient relayé ou non au moins une fake news*

### 3.2.5. Un langage davantage teinté d'émotions négatives mais pas moins réflexif

Un cinquième ensemble de variables a été défini pour comparer le langage employé par les relayeurs de *fake news* à celui des utilisateurs du groupe de contrôle. À cette fin, le logiciel LIWC (Linguistic Inquiry and Word Count) a été mobilisé. Développé par des psycholinguistes, celui-ci vise à identifier les modes de pensées et les états mentaux des individus à partir des mots qu'ils utilisent (Pennebaker et al., 2001 ; Pennebaker et al., 2015). Le logiciel compare les mots présents dans les corpus de textes étudiés avec des dictionnaires préétablis contenant des mots regroupés en différentes catégories couvrant divers aspects du langage, tels que les émotions ou les processus cognitifs. Le LIWC calcule ensuite le pourcentage de mots classés dans chaque catégorie issue des dictionnaires préétablis.

Bien que le LIWC soit largement utilisé dans la littérature académique (pour des exemples voir Bail et al., 2017 ; Vosoughi et al., 2018), il ne s'agit pas du logiciel le plus adapté au raisonnement sociologique (Cointet et Parasie, 2018) dans la mesure où il ne permet pas aux chercheurs de faire émerger leurs catégories d'analyse à partir des textes produits par les utilisateurs. Dans le cadre de cette enquête, le LIWC a moins été utilisé pour analyser les modes d'expression des utilisateurs étudiés que pour prendre position dans les débats académiques qui entourent les *fake news*. Recourir au logiciel a en effet permis de répliquer ou de réfuter les résultats obtenus par différentes études de psychologie ayant étudié le langage d'individus conspirationnistes (Samory et Mitra, 2018 ; Fong et al., 2021), de comptes Twitter qui relaient des sites peu fiables (Mosleh et al., 2021a), ou encore d'utilisateurs qui commentent des *fake news* (Jiang et Wilson, 2018 ; Metzger et al., 2021). Comme la majorité de ces études, deux catégories principales du LIWC ont été retenues : celles regroupant des variables permettant d'analyser les styles cognitifs et émotionnels des individus (pour des exemples de mots associés à chacune des catégories dans la version française du LIWC, voir tableau 3.5).

Afin de minimiser les biais d'interprétation, seuls les tweets originaux ont été pris en compte, tandis que les retweets ont été exclus, réduisant ainsi le corpus de 4 millions à 1,3 million de tweets. Les modes d'expression et registre d'énonciation des utilisateurs seront explorés plus finement dans le prochain chapitre grâce aux analyses qualitatives réalisées.

| Catégorie LIWC | Variables | Exemples |
|---|---|---|
| **Processus Cognitifs** | Perspicacité | penser, savoir, considérer |
| | Cause | parce que, car, d'où |
| | Certitude | toujours, jamais |
| | Inhibition | bloquer, contraindre |
| **Processus Affectifs** | Émotions positives | joyeux, joli, bon |
| | Émotions négatives | haine, bon à rien, ennemi |
| | Anxiété | nerveux, tendu, craintif |
| | Colère | détester, tuer, furieux |

*Tableau 3.5. Exemples de mots pour chacune des catégories du LIWC utilisées*

Les résultats montrent que les utilisateurs qui relaient des *fake news* sur Twitter utilisent plus fréquemment des mots de vocabulaire ayant une connotation émotionnelle négative, exprimant notamment de la colère ou de l'anxiété, par rapport à ceux du groupe de contrôle. En revanche, ils ont moins fréquemment recours à des mots dénotant des émotions positives. Sur le plan cognitif, cependant peu de différences ont été observées, mis à part que les relayeurs de *fake news* semblent établir davantage de liens de cause à effet. Ces résultats sont en partie alignés avec ceux de l'étude de Amos Fong et ses collègues (2021) et suggèrent que les relayeurs de *fake news* n'ont pas un langage moins réflexif que celui des utilisateurs du groupe de contrôle et n'ont pas moins d'esprit critique (ou de capacité de raisonnement).

| **Mesures** | **Variables** | **Relayeurs de *fake news*** | **Autres utilisateurs** | **W** | **p value** |
|---|---|---|---|---|---|
| Processus Cognitifs | Perspicacité | 1.35 | 1.34 | 1469415 | .301 |
| | Cause | 0.65 | 0.56 | 3462612 | <.0001 |
| | Certitude | 1.35 | 1.40 | 1498026 | .045 |
| | Inhibition | 1.46 | 1.49 | 1157448 | .146 |
| Processus Affectifs | Émotion positive | 2.04 | 2.36 | 552043.5 | <.0001 |
| | Émotion négative | 1.88 | 1.56 | 897930 | <.0001 |
| | Anxiété | 0.23 | 0.15 | 2773479 | <.0001 |
| | Colère | 0.64 | 0.46 | 3460650 | <.0001 |

*Tableau 3.6. Comparaison des styles de langage des deux échantillons et résultats des tests de Wilcoxon-Mann-Whitney*

Les constats issus des observations en ligne et des entretiens vont dans le même sens que les résultats obtenus avec le LIWC et permettent d'affiner nos interprétations. Plusieurs enquêtés ont reconnu être dans un état de colère ou de mécontentement lorsqu'ils consultent leur fil d'actualité Twitter et ont indiqué adopter parfois un style véhément ou pouvoir vite s'emporter face à certains contenus :

> *Même si j'essaie de rester courtoise, je ne suis pas tendre [rire].*[114]

> *Ah [ricanement] oui je sais pourquoi je me suis inscrit sur Twitter. Alors c'était en 2014-2015 et c'était pour taper sur François Hollande, pour le massacrer.*[115]

> *Le Covid. La guerre en Ukraine. L'augmentation du pétrole. J'avoue que je ne comprends plus trop vraiment ce qu'il se passe sur cette planète. Et après je suis un peu débordé c'est pour ça que des fois j'ai des réactions épidermiques. Mais je réfléchis bien quoi avant de marquer quelque chose. À part pour Lola où là c'était plus fort que moi. La peine de mort. Faut remettre la peine de mort pour ce genre de personnes.*[116]

Bien que certains relayeurs de *fake news* admettent eux-mêmes se montrer parfois impulsifs ou agressifs sur Twitter, il est essentiel de souligner que ce comportement n'est pas systématique. Leur façon d'agir et de parler ne peut être réduite à leur langage en ligne ni à leur usage de Twitter. En effet, plusieurs participants ont mentionné qu'ils s'autorisaient à dire des choses sur Twitter qu'ils ne diraient pas dans un contexte hors ligne.

> *Si vous dites votre opinion de but en blanc dans la vie réelle, vous choquez les gens qui n'ont pas forcément les mêmes éléments que vous sur le dossier, ou les mêmes convictions et vous n'avez pas envie de choquer pour des raisons familiales ou autre. Donc en fait, dans la vie réelle, la dynamique est différente que sur Twitter où vous pouvez mettre ce que vous pensez ou ce que vous avez constaté. Ce que vous ne pouvez pas faire forcément dans la réalité, que ce soit dans une bulle amicale, une bulle professionnelle ou une bulle familiale.*[117]

Par ailleurs, les pratiques de vérification des relayeurs de *fake news* s'inscrivent souvent dans un cadre réflexif. Plusieurs considèrent que l'accès à l'information en ligne leur permet de contourner les sources officielles et de trouver eux-mêmes les faits qu'ils jugent pertinents, parfois en s'appuyant sur des données disponibles sur Internet ou des études scientifiques.

---

[114] Femme, 58 ans, Bac +3, inactive, ex enseignante, entretien réalisé le 9 juin 2022.
[115] Homme, 56 ans, Bac +2, consultant, entretien réalisé le 19 novembre 2022.
[116] Homme, 60 ans, Bac technologique, employé, entretien réalisé le 22 novembre 2022.
[117] Homme, 55 ans, Bac +8, à son compte, ex chercheur, entretien réalisé le 6 mars 2022.

> *Mais très vite quand on commençait à chercher, on avait toutes les données. Puisque tout est sur la toile. On trouve tout. Les gens qui disent : « ha non, on savait pas ». Si on savait. On savait que le vaccin avait des effets secondaires redoutables ; on savait qu'il n'y avait pas de vaccin qui avait tué autant que celui-là. Tout, tout, est sur la toile. Ça c'est une chose extraordinaire qui n'existait pas moi quand j'étais jeune : pour avoir l'information, c'était très dur ; maintenant vous vous installez dans votre fauteuil, vous regardez, vous trouvez tout, et pour peu que vous lisiez les langues étrangères, surtout l'anglais, vous avez accès à tout.*[118]

Ces constats renforcent l'idée que le partage de *fake news* sur Twitter n'est pas le résultat d'une absence de réflexion critique. Les relayeurs de *fake news* étudiés sont engagés dans une démarche active de recherche de vérité, souvent en opposition à ce qu'ils perçoivent comme une information biaisée des médias traditionnels. Ils se considèrent comme des acteurs investis dans une quête de contre-expertise, avec une volonté affirmée de remettre en question les récits officiels. Ils illustrent ainsi un rapport complexe à l'information, où la défiance envers les sources institutionnelles coexiste avec un certain souci d'objectivité dans la recherche et la vérification des faits (voir aussi Tripodi, 2018 ; Marwick et Partin, 2022 ; Berriche, 2021).

### 3.2.6. Une population plutôt masculine, éduquée et âgée

Avant de conclure ce chapitre, présentons les données socio-démographiques qui ont pu être recueillies au cours des observations en ligne et des entretiens afin de dresser des portraits plus incarnés des utilisateurs qui relaient des *fake news* sur la Twittosphère française et d'explorer si ceux-ci se distinguent des autres par leur âge, leur profession, leur genre ou leur niveau d'étude.

Bien qu'il soit difficile, voire impossible, de déterminer la position des utilisateurs de Twitter dans l'espace social uniquement à partir de leurs traces numériques, il peut arriver que leur nom d'utilisateur, leur bio ou leur photo de profil fournissent des indications suffisantes pour identifier leur âge, leur genre ou leur profession. Parfois, ces éléments permettent également de les retrouver sur des réseaux sociaux comme *Linkedin* ou *Copains d'avant*, où il est d'usage de détailler son parcours académique et professionnel, ce qui donne l'opportunité de vérifier ou de compléter les informations déclarées par un compte sur Twitter. Dans la plupart des cas, cependant, ces informations ne suffisent pas à saisir pleinement les caractéristiques socio-démographiques des comptes, et un travail de fouille plus approfondi s'avère nécessaire. En effet, il peut arriver qu'un compte divulgue spontanément des informations sur son âge, sa profession ou son genre à travers ses tweets, même lorsque son nom est masqué par un pseudonyme. Mais comment capturer de telles informations lorsque les comptes observés sont très actifs et/ou inscrits sur Twitter depuis de nombreuses années et que leur profil contient des milliers de tweets ? Si de longues immersions sur les comptes des utilisateurs sont indispensables pour s'imprégner de leurs façons de s'exprimer sur Twitter, il n'est pas concevable pour une chercheuse ou un chercheur de passer au peigne fin tous les tweets publiés par ses enquêtés, alors même que les chances de tomber sur des éléments socio-démographiques sont très faibles. Pour faire face à ces défis méthodologiques, très fréquents lors de la conduite d'observations en ligne, l'outil de recherche avancé de Twitter a été mobilisé (pour plus d'informations sur l'utilisation de cet outil, voir l'encadré de la section 2.2.3 du Chapitre 2).

[118] Femme, 70 ans, Bac +8, retraitée, ex chercheuse, entretien réalisé le 15 mars 2022.

Recourir à cet outil a permis de déterminer le genre de 94,1 % des comptes de l'échantillon mais l'âge et la profession de seulement 52,2 % et 61,4 % des comptes. Ces proportions sont à peu près équivalentes pour les relayeurs de *fake news* et les autres utilisateurs (cf. Tableau 3.7). Il est important de rappeler toutefois que les comptes étudiés – qu'il s'agisse de ceux de relayeurs de *fake news* ou non – ne sont pas représentatifs de l'ensemble des utilisateurs de Twitter mais seulement de ceux qui partagent des informations issues de média français. Or, en raison de leur activité plus élevée sur Twitter par rapport aux utilisateurs moyens qui sont souvent des *lurkers*, il est possible que ceux-ci divulguent davantage d'informations sur eux à travers leurs tweets. Il est également important de noter qu'un compte d'utilisateurs n'est pas forcément associé à un individu, mais peut appartenir à un collectif, une organisation ou une association. Par ailleurs, il arrive qu'un même individu ait plusieurs comptes.

| | **Relayeurs de *fake news*** (n=124) | **Autres utilisateurs** (n=115) |
|---|---|---|
| Âge | 58 % | 62 % |
| Genre | 97 % | 95 % |
| Profession | 72 % | 69 % |

*Tableau 3.7. Proportion de comptes pour lesquels l'âge, le genre ou la profession ont pu être déterminés*

Les données socio-démographiques collectées révèlent une population de relayeurs de *fake news* plutôt masculine, éduquée et âgée. En effet, parmi les comptes de relayeurs de *fake news* pour lesquels des données socio-démographiques ont pu être obtenues, une légère prédominance masculine peut être constatée (64 %), ainsi qu'une proportion importante d'individus inactifs (33 %) ou appartenant à des catégories socio-professionnelles supérieures (30 %). Par exemple, 22 % ont indiqué être à la retraite et nous avons identifié 3 médecins, 7 enseignants ou chercheurs et 5 chefs d'entreprise. Enfin, l'âge moyen des relayeurs de *fake news* était de 61,7 ans.

Les entretiens réalisés ont également confirmé ces constats. En effet, l'âge moyen des relayeurs de *fake news* interrogés était relativement élevé : 55 ans. Le participant le plus jeune avait 23 ans tandis que le plus âgé avait 70 ans, et seulement deux enquêtés étaient âgés de moins de 40 ans (cf. Chapitre 2, Tableau 2.1). De plus, parmi tous les utilisateurs interrogés, une importante prédominance masculine a aussi été constatée (avec 11 hommes sur 14 enquêtés au total, soit près de 80 %). Par ailleurs, la grande majorité des personnes interrogées, à l'exception de deux, ont suivi des études supérieures. Trois d'entre eux sont même titulaires d'un doctorat. Plusieurs enquêtés exercent ou ont exercé des métiers définis par l'INSEE comme des professions intellectuelles supérieures (e.g. universitaire, pharmacien, ingénieur, professeur dans le secondaire, chef d'entreprise). Il est toutefois possible que les individus appartenant à des catégories socio-professionnelles supérieures ou ayant suivi de longues études soient surreprésentés parmi nos enquêtés car ils sont moins susceptibles de ressentir un sentiment d'illégitimité lorsqu'ils sont sollicités pour participer à une étude académique contrairement à ceux issus de milieux populaires (Mauger et Pouly, 2019).

Si les analyses descriptives présentées ci-dessus révèlent des profils de relayeurs de *fake news* plutôt âgés, masculins et éduqués, notre analyse comparative montre toutefois que seul l'âge différencie vraiment les relayeurs de *fake news* des autres utilisateurs. En effet, parmi notre échantillon de comptes n'ayant pas partagé de *fake news*, nous avons également observé une prédominance masculine (67 %) et de CSP+ (43%), mais une population plus jeune avec une moyenne d'âge de 44,6 ans. En fait, il est possible que la surreprésentation des hommes et des cadres et professions intellectuelles supérieures, aussi bien parmi les relayeurs de *fake news* que les autres utilisateurs, s'explique par la particularité sociologique des utilisateurs de Twitter, caractérisées par une prédominance masculine et une surreprésentation des catégories socio-professionnelles élevées (Boyadjian, 2014). En revanche, alors que les utilisateurs de Twitter sont plus jeunes que le reste de la population française[119], il est surprenant de constater que les relayeurs de *fake news* sont nettement plus âgés que les autres utilisateurs sur cette plateforme.

Ce résultat n'est pas surprenant lorsqu'on le met en perspective avec les conclusions issues de nombreuses enquêtes reposant sur des données comportementales (Munger et al. 2018 ; Grinberg et al., 2019). À travers deux études, Andy Guess et ses collègues (2019 ; 2021) ont par exemple montré que les utilisateurs âgés de plus de 65 ans diffusaient 7 fois plus de *fake news* que les jeunes âgés de 18 à 29 ans et que 20 % des URLs partagées par les plus de 65 ans étaient issues de sources peu crédibles, alors que ce n'était le cas que pour 11 % des URLs partagées par les 24 à 35 ans. Ce constat d'un lien entre un âge élevé et le partage de *fake news* entre cependant en contradiction avec les résultats de nombreuses études expérimentales qui trouvent que la capacité des individus à discerner le vrai du faux augmente avec l'âge (Allcott and Gentzkow, 2017 ; Pennycook and Rand, 2018 ; Roozenbeek et al., 2020). Comment expliquer ce décalage entre les résultats issus des données comportementales et ceux issus des études basées sur des questionnaires déclaratifs ?

[119] D'après l'enquête réalisée par Julien Boyadjian (2014), l'âge moyen des utilisateurs est de 31 ans et l'âge médian de 26 ans.

Une recherche conduite par Ben Lyons (2023) offre des éléments de réponse à cette question. Dans un premier temps, le chercheur a adressé trois questionnaires à des panels représentatifs de la population américaine. Au cours de chaque sondage, les participants ont dû évaluer la crédibilité de trois types de contenus médiatiques (1) vrais ou faux ; (2) orientés à droite ou à gauche ; (3) hyper partisans (i.e. factuels mais idéologiquement orientés à droite ou à gauche). L'exposition médiatique des répondants aux sondages a ensuite été analysée à partir de leurs traces de navigation en ligne. La comparaison entre ces deux dispositifs d'enquête fait une nouvelle fois apparaître une discordance entre données comportementales et données déclaratives. Alors que les traces de navigation des enquêtés montrent que les personnes âgées de plus de 60 ans consomment plus d'informations issues de sources médiatiques peu fiables que les 18-34 ans, les résultats issus des questionnaires montrent que les personnes âgées ne sont pas moins capables de discerner le vrai du faux que les plus jeunes.

Pour expliquer cette différence, Bens Lyons fait émerger d'autres variables que celles prises en compte dans les études expérimentales de psychologie. Tout d'abord, les traces de navigation des enquêtés révèlent que les personnes âgées consomment globalement beaucoup plus d'informations que les plus jeunes — un constat corroboré par de nombreuses études (Moretto et al., 2022).[120] Cette surexposition peut expliquer pourquoi les personnes âgées sont plus susceptibles de relayer des informations douteuses ou peu fiables. Ensuite, les réponses des enquêtés aux questionnaires indiquent que les personnes âgées de plus de 60 ans sont plus sujettes à un biais idéologique lorsqu'on leur demande d'évaluer des contenus hyper partisans. Autrement dit, en dépit de capacités de discernement égales, voire supérieures, à celles des plus jeunes, les Américains plus âgés semblent plus motivés à partager des informations correctes mais présentant des faits sous un angle biaisé afin de soutenir leur camp politique ou de critiquer leurs opposants.

---

[120] Grâce à un accès à la base de données Condor de Facebook comportant des informations démographiques sur les internautes qui voient et partagent des URLs sur le réseau social, cette étude a montré que l'âge moyen des personnes qui partagent des URLs est considérablement plus élevé que l'âge de ceux qui les voient simplement, et que ce phénomène s'observe dans 43 pays différents. Après avoir procédé à une classification manuelle des URLs partagées en Amérique du Sud, les chercheurs ont également constaté que l'âge moyen des utilisateurs qui partagent les URLs augmente lorsque le contenu est politique et issu de médias partisans, et augmente encore lorsqu'il est de droite.

En somme, il est possible de conclure que la corrélation entre un âge élevé et un partage de *fake news* plus important n'est probablement qu'une conséquence d'un intérêt plus marqué pour l'actualité et la politique (Glenn et Grimes 1968 ; Moretto et al., 2021 ; Thurman et Fletcher, 2019) et d'un niveau de politisation plus important. En effet, les personnes âgées ont des identités partisanes plus établies et éprouvent davantage que les plus jeunes des sentiments négatifs envers ceux qui ne partagent pas leur point de vue politique (Phillips, 2022).

Il ressort ainsi que les personnes qui dans les enquêtes partagent le plus fréquemment des *fake news* ne correspondent pas aux caractéristiques des populations auxquelles les discours publics attribuent un plus grand manque d'esprit critique – à savoir les jeunes et les moins éduqués. Ce résultat contribue à renforcer l'idée que les variables cognitives jouent un rôle sans doute moins important que les valeurs idéologiques dans les troubles informationnels contemporains.

## Conclusion du troisième chapitre

Les analyses conduites tout au long de ce chapitre confirment l'existence d'un écart entre la façon dont les utilisateurs qui partagent des *fake news* sur les réseaux sociaux sont dépeints dans les discours publics et leurs caractéristiques réelles. Loin d'être le fruit d'une foule d'individus crédules, le partage de *fake news* est plutôt l'apanage d'une petite minorité d'utilisateurs ayant pour particularités d'être très fortement concentrés aux extrêmes de l'échiquier politique et surtout particulièrement opposés aux élites et aux institutions. Ils se démarquent des autres utilisateurs par une identité politique plus marquée ; des pratiques informationnelles plus importantes ; une activité en ligne beaucoup plus intense ; une utilisation plus fréquente d'émotions négatives ; ainsi qu'un âge plus avancé. Néanmoins, ils ne semblent pas moins dotés de capacités de raisonnement et ne sont pas moins éduqués que les autres utilisateurs.

Ces constats empiriques permettent d'apporter des premiers éléments de réponse aux deux paradoxes que cette thèse cherche à élucider. Si le partage de *fake news* touche seulement une minorité d'utilisateurs sur les réseaux sociaux, c'est parce que ce phénomène est davantage associé à des facteurs socio-politiques que cognitifs. Comme l'ont montré de nombreuses études, la plupart des utilisateurs consacre très peu de temps à consulter des informations politiques sur les réseaux sociaux (Allen et al., 2020 ; Nyhan et al., 2023). Ce détachement relatif vis-à-vis de la politique les préserve donc en quelque sorte contre le partage de *fake news*, bien qu'ils puissent toujours être exposés à ces contenus. *A contrario*, les utilisateurs des réseaux sociaux qui sont les plus engagés politiquement peuvent percevoir les *fake news* comme un levier pour défendre leur identité politique et critiquer les institutions ou leurs opposants idéologiques. Par ailleurs, leur forte activité en ligne peut aboutir à une surreprésentation de leurs opinions dans les espaces de visibilité numérique et renforcer la polarisation politique sur les réseaux sociaux, ou du moins donner l'impression d'une polarisation accrue dans le débat public.

Ces constats empiriques permettent également d'apporter des contributions méthodologiques et théoriques importantes dans le champ des études contemporaines sur les *fake news*.

Sur le plan méthodologique, ils montrent comment il est possible d'identifier de façon rigoureuse les caractéristiques spécifiques d'un groupe d'utilisateurs de réseaux sociaux en s'appuyant sur leurs traces numériques. En mobilisant des travaux de sciences sociales computationnelles reposant sur des méthodes d'inférences idéologiques (Cointet et al., 2021 ; Ramaciotti Morales et al. 2021), il a été possible d'estimer la position politique de plus d'un tiers des relayeurs de *fake news* identifiés sur la Twittosphère française. Une procédure de *matching* a ensuite permis de comparer un groupe de relayeurs de *fake news* à un groupe d'utilisateurs n'en ayant pas partagé en neutralisant l'effet de leur position politique. Enfin, une ethnographie en ligne et des entretiens ont permis de dresser des portraits plus détaillés et incarnés des relayeurs de *fake news* et d'avoir une compréhension plus fine et nuancée de leurs pratiques. Bien entendu, malgré ces contributions importantes, ce dispositif méthodologique n'est pas sans limite. Tout d'abord, il est restreint à Twitter, un réseau social qui n'est pas représentatif de l'ensemble des plateformes. Ensuite, la méthodologie utilisée se concentre sur un type particulier de relayeurs de *fake news* : ceux qui suivent des députés. Il est donc possible que les utilisateurs qui ne suivent pas de députés sur Twitter partagent des *fake news* pour d'autres raisons que politiques, par exemple en raison d'un intérêt important pour des sujets liés à la santé, ce qui pourrait éventuellement les amener, à terme, à s'engager sur des thématiques plus explicitement politiques. Enfin, il convient de souligner la distinction entre le partage de *fake news* et l'exposition à celles-ci. Si cette étude se focalise sur les individus qui partagent activement et publiquement des *fake news*, il est tout à fait possible que d'autres utilisateurs, bien qu'exposés à ce type de contenu, ne les relaient pas (ou seulement en privé) mais s'en trouvent tout de même influencés.

Sur le plan théorique, ces constats permettent d'écarter les modèles explicatifs qui attribuent le partage de *fake news* exclusivement à des variables cognitives, comme un déficit de raisonnement analytique. Ils confirment l'intérêt d'adopter une approche pragmatique, qui prend en compte les compétences critiques des acteurs dans l'étude de la réception des *fake news* au sein de l'écosystème informationnel contemporain. De plus, ces résultats affinent les modèles d'explications du partage de *fake news* en suggérant que les motivations politiques des utilisateurs ne sont pas nécessairement partisanes mais émanent davantage d'une opposition aux élites au pouvoir.

Enfin, il convient de noter que les observations et les entretiens ont révélé quelques nuances et divergences entre les relayeurs de *fake news*. Si la plupart d'entre eux utilisent des pseudonymes, sont très actifs en ligne et ont plutôt tendance à retweeter qu'à écrire leurs propres messages, quelques-uns se distinguent par l'utilisation de leur vrai nom et une activité plus modérée, privilégiant des publications originales. L'enjeu du prochain chapitre consistera à explorer ces variations de manière plus approfondie afin d'identifier plus finement les mécanismes qui favorisent (ou limitent) la mise en circulation de *fake news*.

# Chapitre 4. Intégration dans l'espace social, normes d'interactions et prudence énonciative

Le chapitre précédent s'est attaché à comparer un groupe d'utilisateurs ayant relayé des *fake news* sur Twitter avec un groupe d'utilisateurs n'en ayant pas partagé, tout en contrôlant leur position politique. Cette approche comparative, reposant principalement sur des analyses quantitatives, a permis de montrer que les utilisateurs qui partagent des *fake news* sur Twitter ne se caractérisent pas principalement par un manque de raisonnement analytique mais plutôt par une forte politisation, un âge élevé, une activité en ligne très importante et un intérêt marqué pour le partage d'informations.

Derrière cette vue d'ensemble, toutefois, les observations en ligne et les entretiens ont fait ressortir d'importantes nuances. D'un côté, plusieurs divergences sont apparues entre certains comptes de relayeurs de *fake news*, notamment au niveau de leurs pratiques de partages d'informations et de leurs façons de s'exprimer sur les réseaux sociaux. De l'autre, quelques écarts ont été relevés entre leurs prises de parole en ligne et hors ligne. Par exemple, alors que la majorité des comptes utilise un pseudonyme et relaie de nombreux contenus issus de médias de contre-information sous la forme de retweets « secs », certains utilisateurs écrivent leurs propres tweets sans masquer leur identité civile. D'autres, quant à eux, ont indiqué s'exprimer sur un ton beaucoup plus modéré hors ligne que sur Twitter.

L'objectif de ce chapitre est précisément d'approfondir l'analyse de ces variations (tant inter-individuelles qu'intra-individuelles). Pourquoi certains comptes partagent-ils un très grand nombre de *fake news* tandis que d'autres en partagent très peu ? Certains font-ils plus attention que d'autres aux propos qu'ils tiennent sur les réseaux sociaux ? Et puis, qu'en est-il hors ligne ?

Pour répondre à ces différentes questions, ce chapitre s'appuie sur les matériaux d'enquête recueillis en menant des observations sur 124 comptes de relayeurs de *fake news* et en conduisant des entretiens avec 14 d'entre eux. Comme cela a été documenté dans le chapitre 2, ce dispositif méthodologique hybride, visant à compléter des analyses quantitatives de traces numériques par des observations en ligne et des entretiens, a permis de ne pas réduire les enquêtés étudiés au fait d'avoir partagé une *fake news* sur un réseau social particulier, mais de prêter attention à la diversité des situations d'interactions (en ligne comme hors ligne) dans lesquelles ils ont l'occasion de recevoir ou de partager des informations, tout en prenant en compte leur position dans l'espace social.

Les données ont été analysées en combinant une approche microsociologique avec une perspective macrosociologique. Cette posture épistémologique a permis d'examiner si les comportements des relayeurs de *fake news* varient selon les audiences auxquelles ils sont confrontés dans des situations d'interactions spécifiques tout en tenant compte de variables plus structurelles. Cette articulation entre un niveau micro et macro est importante car selon leur position sociale les individus n'évoluent pas dans les mêmes configurations interactionnelles et ne sont pas soumis aux mêmes prérogatives et obligations (Giddens, 1979).

Ce chapitre commence par présenter les différentes positions sociales occupées par les relayeurs de *fake news* étudiés, ainsi que les attentes et obligations associées aux rôles qu'ils doivent endosser. Il décrit ensuite les diverses situations d'interactions dans lesquelles ces utilisateurs ont l'occasion de s'exprimer et les contraintes énonciatives qui pèsent sur leurs prises de parole. La troisième section examine si ces attentes et contraintes façonnent leurs pratiques informationnelles et numériques, ainsi que leurs régimes d'énonciation sur Twitter. Enfin, la dernière section revient sur plusieurs pratiques, observées au cours des analyses précédentes, à travers lesquelles les utilisateurs intègrent dans leur énonciation une anticipation des potentielles réactions de leur audience réelle ou imaginée, et plus précisément des sanctions qu'ils sont susceptibles d'encourir. La notion de « prudence énonciative » est alors proposée pour désigner l'ensemble de ces pratiques.

En résumé, les analyses réalisées tout au long de ce chapitre permettent de mettre en évidence une compétence de distance critique, la prudence énonciative, que peuvent mobiliser les utilisateurs des réseaux sociaux, même ceux qui relaient des *fake news*, pour préserver leur réputation et maintenir leur intégration dans différents espaces sociaux.

## 4.1. Position sociale, attentes normatives et risques de sanctions

Selon Anthony Giddens (1979, p. 117), la position sociale correspond à « une identité sociale qui s'accompagne d'un ensemble de prérogatives et d'obligations que peut mettre en œuvre ou remplir un acteur à qui cette identité est accordée ». L'objectif de cette section est donc double : identifier, d'une part, les différentes positions occupées dans l'espace social par les utilisateurs de Twitter ayant partagé au moins une *fake news* sur la plateforme, et décrire, d'autre part, l'ensemble des attentes associées à ces positions, ainsi que les sanctions qu'ils peuvent encourir en cas de faute ou de transgression.

Les analyses ont fait ressortir une diversité de niveaux d'engagement et d'appartenance à des sphères politiques, professionnelles, familiales ou amicales, et ont permis de positionner les utilisateurs étudiés sur trois axes principaux, selon (1) leur mode d'affiliation partisane ; (2) leur degré d'insertion professionnelle ; (3) l'intensité de leurs relations sociales ou leur proximité affective.

### 4.1.1. Affiliation partisane

Dans l'ensemble les utilisateurs qui relaient des *fake news* sur Twitter se démarquent des autres par un niveau de politisation beaucoup plus élevé. Cette forte politisation se traduit cependant par des formes variées d'affiliation partisane et de participation politique (Duverger, 1951a ; Scarrow, 2014 ; Gibson et al., 2018). Pour rappel, les observations et les entretiens réalisés ont permis de faire émerger cinq types de profils en fonction de leur degré d'adhésion et d'identification envers un parti politique. Parmi l'échantillon de relayeurs de *fake news* analysé de manière qualitative, 12 comptes appartiennent à des élus ; 23 à des adhérents ; 50 à des militants ; 36 à des utilisateurs dépourvus d'attache partisane spécifique ; et 3 à des utilisateurs peu politisés. Une description détaillée de chacun de ces profils est proposée ci-dessous, à l'exception des 3 comptes peu politisés – ceux-ci étant trop peu nombreux et trop peu actifs sur Twitter pour être analysés.

***Les élus***

Parmi les 12 comptes d'élus identifiés, 9 jouent actuellement – ou ont joué par le passé – un rôle de représentant politique à un échelon local, tandis que 3 seulement exercent – ou ont exercé – leurs fonctions à un niveau national ou européen. Les trois quarts de ces élus sont rattachés à des partis d'extrême droite, tandis que 2 appartiennent à des formations de centre-droit et 1 seul au parti Europe Écologie Les Verts (Les Écologistes depuis 2023). Cette prédominance d'élus d'extrême droite pourrait s'expliquer par un contrôle interne moins strict et une plus grande porosité entre les discours de ces partis et certains contenus controversés ou haineux. Par exemple, une enquête conduite par des journalistes de *BuzzFeed News*[121] a montré que sur les 573 candidats présentés par le FN (Rassemblement National depuis 2018) aux législatives en 2017, une centaine ont posté, aimé ou partagé des contenus homophobes, antisémites, islamophobes ou racistes. Plus récemment, une enquête de *Libération*[122] a identifié 28 élus RN, représentant environ 20 % des 142 sièges remportés par l'alliance d'extrême droite, impliqués dans des propos discriminants ou violents. Ces dérives ne sont toutefois pas exclusives à l'extrême droite. Par exemple, au sein de La France Insoumise, des élus ont également été critiqués par des médias comme *Le Point*[123] ou par le gouvernement[124] pour avoir relayé des informations erronées ou tenu des propos controversés. De tels incidents ont néanmoins été moins fréquemment observés qu'à l'extrême droite.

Bien que les codes de conduite varient d'un parti à l'autre et que les responsabilités des élus divergent selon leur rôle (e.g. maire, député, conseiller municipal, etc.), tous sont tenus de servir l'intérêt général, de respecter la discipline de leur parti (Leconte et al., 2017 ; Castagnez,

[121] Perrotin, D., Maad, A., Aveline, P., Darmanin, J., Kirschen, M., & Léchenet, A. (2017, 6 juin). «Lobby Juif», «Banania», «connards de Français» : on a scruté les comptes Facebook et Twitter des 573 candidats FN. *Buzzfeed*. https://www.buzzfeed.com/fr/davidperrotin/enquete-fn-legislatives.
[122] Macé, M., et Plottu, P. (2024, 9 juillet). Froid dans le dos : Racistes, complotistes, homophobes, pro-Poutine : la galerie des horreurs des députés RN qui siègeront à l'Assemblée. *Libération*. https://www.liberation.fr/politique/racistes-complotistes-homophobes-pro-poutine-la-galerie-des-horreurs-des-deputes-rn-qui-siegeront-a-lassemblee-20240709_FFQ3KX6AAZALHCPQREEH7DOQJQ/
[123] Michelet, H. (2024, 5 août). Quand la députée LFI Ersilia Soudais diffuse des fake news sur les JO. *Le Point*. https://www.lepoint.fr/politique/quand-la-deputee-lfi-ersilia-soudais-diffuse-des-fake-news-sur-les-jo-05-08-2024-2567249_20.php - 11
[124] Le Monde avec AFP. (2024, 29 avril). Le gouvernement va porter plainte contre Jean-Luc Mélenchon pour « injure publique » à la suite de ses propos sur le nazi Eichmann. *Le Monde*. https://www.lemonde.fr/politique/article/2024/04/29/le-gouvernement-va-porter-plainte-contre-jean-luc-melenchon-pour-injure-publique-apres-ses-propos-sur-le-nazi-eichmann_6230458_823448.html

2006), et de répondre aux attentes de leurs électeurs. Le non-respect de ces règles expose les élus à des risques élevés de sanctions, d'exclusion ou de critiques publiques. Par exemple, lorsque les élus partagent des informations non vérifiées ou tiennent des propos polémiques, ils risquent de décrédibiliser l'ensemble de leur parti. Afin de modérer l'image de son parti, Marine Le Pen a ainsi entrepris des purges et a exclu certains membres du RN avant les législatives de 2017[125]. Récemment des membres du RN[126] et de LFI[127] ont également été écartés des listes électorales de leur parti en raison d'accusations d'antisémitisme.

Ces observations suggèrent que les élus appartenant à des partis moins intégrés dans l'espace politique ont moins tendance à contrôler leur énonciation que ceux plus proches du gouvernement. Toujours est-il que ces élus font tout de même l'objet d'une surveillance médiatique et partisane importante. Pour éviter d'être sanctionnés ou exclus de leur parti, il est ainsi possible qu'ils se sentent contraints d'avoir recours à des stratégies pour policer leurs discours, une hypothèse qui sera explorée dans les sections suivantes de ce chapitre.

***Les adhérents***

Le deuxième ensemble de profils identifiés (n=23) est composé de relayeurs de *fake news* ayant officiellement obtenu une carte de membre d'un parti à un moment donné. En effet, bien que les partis politiques contemporains offrent des formules d'engagement de plus en plus flexibles et à intensité variable (Scarrow, 2014 ; Gibson et al., 2018), l'adhésion à un parti reste un acte formel qui se matérialise par la signature d'un bulletin d'adhésion, le versement d'une cotisation et l'obtention d'une carte officielle de membre (Duverger, 1951b).

Le fait d'adhérer à un parti politique ne se limite pas à un simple alignement idéologique, mais implique également des engagements et des responsabilités. Souvent, les nouveaux adhérents sont tenus de signer des chartes ou des codes de conduite pour certifier qu'ils approuvent les principes et les valeurs du parti. Certains partis peuvent également assigner des tâches spécifiques aux adhérents, telles que la participation à des actions de terrain, la collecte de fonds ou la diffusion d'informations sur les réseaux sociaux. En période électorale, la discipline partisane devient particulièrement importante, avec des attentes accrues en termes de contribution à la campagne électorale. Les adhérents sont tenus de respecter la ligne politique définie par le parti et de soutenir ses positions publiques.

---

[125] de Bonni, M. (2016, 13 septembre). Jean-Lin Lacapelle, le « nettoyeur » du FN. *Le Figaro*. https://www.lefigaro.fr/politique/le-scan/2016/09/13/25001-20160913ARTFIG00207-jean-lin-lacapelle-le-nettoyeur-du-fn.php
[126] France info avec AFP. (2024, 20 juin). Législatives 2024 : le candidat suspendu du RN pour un tweet réfute tout antisémitisme. *France Télévision*. https://www.francetvinfo.fr/elections/legislatives/le-candidat-suspendu-du-rn-pour-un-tweet-refute-tout-antisemitisme_6616041.html
[127] Bruandet, L. (2024, 27 juin). Tweets antisémites : LFI retire son investiture au candidat aux législatives Reda Belkadi. *Le Journal du Dimanche*. https://www.lejdd.fr/politique/tweets-antisemites-lfi-retire-son-investiture-au-candidat-aux-legislatives-reda-belkadi-146845

Si jamais les adhérents s'écartent du discours du parti, ils encourent différents risques de sanctions, tels que des avertissements, des suspensions temporaires de leurs activités au sein du parti, voire une exclusion totale comme en attestent les propos de cet enquêté concernant un de ses amis membre du parti La France Insoumise :

> *À la LFI, j'ai un copain qui s'est fait exclure du parti de gauche parce que justement il tenait des positions anti-injection.*[128]

***Les militants***

Le troisième ensemble de profils identifiés (n=50) est composé d'individus qui expriment leur soutien à un parti politique par leurs actions et leurs comportements, notamment en ligne, sans pour autant en être officiellement membres (Scarrow, 2014 ; Gibson et al., 2018) :

> *J'suis à la gauche de la gauche mais encarté dans aucun parti.*[129]

> *Alors je n'adhère et je n'ai adhéré à aucun parti. En revanche, je suis assez militante sur Twitter pour relayer des propos, des invitations à voter pour tel ou tel candidat. Je me reconnais plutôt à l'heure actuelle dans Reconquête mais ça ne m'empêche pas de m'intéresser à d'autres personnes que ce soit Bardella pour le RN, que ce soit Nicolas Dupont Aignan, François Asselineau, Florian Philippot. Voilà... Même si je ne voterai pas forcément pour eux.*[130]

---

[128] Femme, 67 ans, certificat d'étude, retraitée, entretien réalisé le 19 mai 2022.
[129] Homme, 45 ans, Bac +4, fonctionnaire, entretien réalisé le 18 mai 2022.
[130] Femme, 58 ans, Bac +3, inactive, ex enseignante, entretien réalisé le 9 juin 2022.

Contrairement aux adhérents, les militants ne sont pas soumis aux obligations institutionnelles d'une organisation partisane et bénéficient d'une plus grande latitude pour définir les modalités de leur engagement en fonction de leurs préférences individuelles et de leurs convictions personnelles. Dans une certaine mesure, ce mode d'engagement, moins institutionnalisé, plus individualisé (Duchesne, 1997), fait écho au modèle du « militantisme distancié » identifié par Jacques Ion (1997), dans son livre *La fin des militants ?* Ce modèle suggère une forme d'engagement « post-it », où la participation aux activités partisanes traditionnelles est remplacée par des actions ponctuelles et autonomes. D'une certaine manière, ces possibilités d'engagement à la carte ont été renforcées au cours des dernières décennies par le développement des outils de communication numérique (Blondeau et Allard, 2007 ; Greffet et al., 2014). Par exemple, les militants ne sont pas astreints à aller aux réunions d'un parti ou à participer activement à ses actions de terrain, et peuvent se contenter de tweeter lors d'un *meeting* pour soutenir un candidat, de commenter un blog politique ou de partager l'actualité d'un parti sur Facebook quand bon leur semble.

***Les indépendants***

Le quatrième groupe de relayeurs de *fake news* identifiés (n=36) est composé d'utilisateurs qui ne se sentent représentés par aucun parti, ni par aucune personnalité politique. Ils se caractérisent par une forte propension à critiquer le gouvernement et les partis traditionnels — qu'ils n'hésitent pas d'ailleurs à regrouper sous une même étiquette en fusionnant leurs acronymes et en parlant de « régime UMPS » ou de « système LR-LREM ». Toutefois, ce n'est pas parce que ces utilisateurs sont dépourvus d'attache partisane qu'ils sont dénués d'opinion ou de compétence politique. En réalité, aucun n'est véritablement totalement « hors du jeu politique » (Jaffré et Muxel, 2000). Tous manifestent un intérêt marqué pour l'actualité politique et possèdent une bonne connaissance du fonctionnement des institutions politiques et du système électoral français, même ceux qui s'abstiennent de voter depuis plusieurs années. Sur certains sujets de débats, tels que l'immigration, le terrorisme, le réchauffement climatique, l'obligation vaccinale, l'euthanasie ou la PMA, leurs positions sont très tranchées et rejoignent souvent celles défendues par des partis politiques de droite ou de gauche radicale. Certains d'entre eux ont également soutenu une personnalité ou un parti politique par le passé, mais ont fini par en être déçus ou par se sentir trahis, et indiquent aujourd'hui voter « par défaut » pour le candidat qu'ils considèrent comme le « moins pire » plutôt que de s'abstenir :

> *Je ne me sens rattaché à aucun parti, aucun candidat. Mais alors vraiment aucun. À l'époque, je me reconnaissais dans Nicolas Sarkozy et Sarkozy m'a déçu. Et puis là je ne me reconnais plus dans aucun candidat.*[131]

> *Alors mon engagement en politique, ça se limite à mes posts sur Twitter et j'aimerais qu'on quitte ce régime UMPS qui a détruit le pays dans tous les domaines depuis 35 ans. Donc en l'occurrence, comme il y a une élection qui se rapproche, je soutiens Zemmour qui malheureusement sort quelques conneries de temps en temps. C'est le problème de l'engagement en politique, c'est que quand y a un choix limité, on est obligé de faire avec les faiblesses du candidat. Mais c'est le moins pire pour moi.*[132]

Malgré leurs désillusions, ces utilisateurs restent politiquement engagés. Néanmoins, leurs activités se limitent pour la plupart à leurs publications sur les réseaux sociaux. Contrairement aux élus et adhérents, qui doivent souvent aligner leurs propos sur les discours officiels de leur formation politique, les indépendants peuvent exprimer leurs opinions de manière plus spontanée et directe sans avoir de compte à rendre à aucun groupe politique spécifique, et sans craindre les répercussions qu'un adhérent ou un élu pourrait redouter de la part de son parti ou de son électorat.

Si les utilisateurs étudiés dans le cadre de cette enquête ont été classés selon leur degré d'implication et d'engagement envers un parti politique, il est important de noter que ces catégories n'ont rien de figées et d'exclusives. En réalité, les frontières entre différents modes d'affiliation partisane et de participation politique sont souvent mouvantes et poreuses. Par exemple, il arrive que des membres d'un parti finissent par se désengager et par continuer à militer simplement en ligne (Théviot, 2020), ou que des citoyens ordinaires en viennent à adhérer à un parti et à y exercer des responsabilités. Sur les 23 comptes identifiés comme des adhérents,

[131] Homme, 39 ans, Bac +3, infirmier, entretien réalisé le 20 mai 2022.
[132] Homme, 55 ans, Bac +8, à son compte, ex-chercheur, entretien réalisé le 6 mars 2022.

6 comptes, notamment issus du parti Les Républicains, ont alterné plusieurs fois entre des phases d'adhésion et de non-adhésion entre 2017 et 2022. Aussi, dans la suite des analyses, une attention particulière sera portée aux éventuels changements de statut des enquêtés dans leur trajectoire.

Ces différents degrés d'affiliation à un parti politique sont également fréquemment associés à des propriétés socio-professionnelles et démographiques particulières. Par exemple, les élus occupent souvent des positions élevées dans l'espace social (Rouban, 2011). La plupart des élus de notre échantillon exercent en parallèle de leurs fonctions politiques – ou ont exercé antérieurement – des professions ayant nécessité de longues études ou impliquant d'importantes responsabilités (e.g. pharmacien, directeur de banque, juriste, cadre, etc.). *A contrario*, on retrouve davantage d'individus plus isolés ou appartenant à des groupes sociaux moins favorisés chez les utilisateurs indépendants. L'objectif des deux prochaines sections est de présenter les contraintes et les risques de sanctions associés (ou non) à ces différents degrés d'insertion professionnelle et d'intensité relationnelle.

### 4.1.2. Insertion professionnelle

Alors que la majorité des relayeurs de *fake news* étudiés dans le cadre de cette enquête ne sont pas des professionnels de la politique (Gaxie, 1973 ; Lagroye, 1994), il est important de ne pas analyser leurs pratiques uniquement à travers le prisme de leur affiliation à un parti politique. Pour mieux comprendre leur comportement, il est nécessaire d'adopter une perspective plus large et d'explorer leurs liens avec d'autres types d'organisations, notamment professionnelles.

Comme cela a été montré dans le chapitre précédent, une proportion assez importante de relayeurs de *fake news* appartient à des catégories socio-professionnelles supérieures, mais peu d'entre eux proviennent des catégories sociales les plus défavorisées. En revanche, une proportion importante est constituée d'inactifs, notamment de retraités, mais aussi de personnes en arrêt maladie, en situation de handicap ou reconnues comme souffrant d'une affection de longue durée.

Si les analyses conduites n'ont pas permis de comparer finement les pratiques des relayeurs de *fake news* selon leur catégorie socio-professionnelle, elles ont fait ressortir des contrastes importants entre ceux qui exercent une activité professionnelle et ceux qui n'en ont pas.

### *Responsabilités professionnelles et devoir de réserve*

Les individus qui exercent une activité professionnelle sont beaucoup plus susceptibles d'être tenus de respecter un devoir de réserve que les inactifs. Par exemple, trois personnes issues du monde académique (dont l'une reconvertie dans un autre secteur et l'autre à la retraite) ont rapporté s'être déjà senties contraintes de faire profil bas et de s'auto-censurer de peur que leurs prises de position n'aient de répercussions sur leur carrière.

> *On est pas du tout dans un régime de liberté. Surtout pas de liberté de paroles. Moi je connais beaucoup de gens qui sont obligés d'être extrêmement opaques. J'ai des amis qui sont. Je pense à un en particulier qui est un universitaire, qui est un grand professeur d'économie, qui a longtemps été sur Twitter. Il n'y est plus maintenant. Il était en danger. Il se mettait en danger. J'en connais un autre qui n'y est plus aussi, qui était un philosophe, qui est très connu aussi, et bon pareil. Enfin c'est… Les gens ont… Vous savez, l'institution académique c'est… Moi je faisais profil bas. De tout façon il y a un tel conformisme, on va dire de gauche, où toutes les évidences de gauche sont présentées avec la force de l'évidence incontestable qu'on va pas rentrer à batailler, à argumenter, c'est pas possible. Donc bon on dit : « oh oui » ou on dit rien.*[133]

De façon intéressante, Gauthier, un enquêté encore étudiant, soutenant le Rassemblement National, a également indiqué se sentir dans l'obligation de faire profil bas face à ses professeurs et à ses camarades de peur que ses opinions politiques n'aient de conséquences négatives sur ses évaluations académiques, sa future carrière ou ses relations avec les autres étudiants.

> *À l'université, je ne le dis surtout pas ! J'ai envie d'avoir mon année ma chère. Par contre, ça c'est pas un mythe, quand t'es d'extrême droite, t'es au chômage. Je ne sais pas, t'es responsable commercial dans une entreprise et tu fais comprendre à tes collègues que voilà… Putain, crois-*

[133] Femme, 70 ans, Bac +8, retraitée, ex-chercheuse, entretien réalisé le 15 mars 2022.

*moi que le chômage il va être très rapide. L'université c'est quand même majoritairement à gauche, je suis obligé de faire profil bas, même pas profil bas, de me faire tout petit. Je ne ramène jamais ma fraise. Moi à la fac, personne ne le sait, je suis un caméléon.*[134]

Dans la même veine que les élus ou que les membres d'un parti, les personnes rattachées à une organisation professionnelle risquent d'être sanctionnées en cas d'impair ou de manquement à une règle. Ces sanctions peuvent aller d'un simple rappel à l'ordre à l'exclusion.[135] Par exemple, un enquêté, William, a raconté s'être fait reprocher d'avoir défendu certaines prises de positions politiques à l'extérieur de son travail.

*J'ai eu une mauvaise expérience professionnelle pour des gens qui ne comprenaient rien à ça et qui voulaient contrôler très étroitement le message à la fois de l'organisation mais aussi de tous ceux qui étaient dans l'organisation. Ils comprenaient pas forcément que ce n'est pas parce qu'on travaille dans une organisation que ce qu'on dit doit refléter… Ça n'a pas trop plu, donc je me suis dit : « bon ok ».*[136]

Les risques de réprimande ont été accrus pendant la pandémie, notamment pour les personnes travaillant dans le milieu de la santé, où les attentes de conformité aux discours officiels ont été particulièrement élevées. Dans ce contexte, prendre position contre le vaccin ou le pass sanitaire pouvait être perçu comme un acte de défiance, non seulement vis-à-vis des autorités, mais également envers les collègues et l'institution hospitalière elle-même. Un enquêté travaillant comme infirmier dans un hôpital a ainsi indiqué rester neutre lorsque ces sujets étaient abordés avec ses collègues ou avec des patients. Malgré ses réticences à l'égard du vaccin contre le Covid-19 et son opposition au pass sanitaire, il a préféré respecter ces mesures plutôt que de perdre son emploi.[137]

---

[134] Homme, 23 ans, Bac +4, étudiant, entretien réalisé le 5 décembre 2022.
[135] The Economist. (2020, February 27th). The case for free speech at work. *The Economist*. https://www.economist.com/leaders/2020/02/27/the-case-for-free-speech-at-work
[136] Homme, 53 ans, Bac +8, chercheur, entretien réalisé le 18 mai 2022.
[137] Homme, 39 ans, Bac +3, infirmier, entretien réalisé le 20 mai 2022.

***Inactivité professionnelle et disponibilité de temps***

Contrairement aux relayeurs de *fake news* qui exercent une profession, ceux qui sont inactifs ne risquent pas de sanction directe liée à une prise de parole publique qui pourrait nuire à leur carrière ou à leur réputation au sein d'une organisation. Par ailleurs, ils bénéficient d'une disponibilité de temps plus importante qui leur permet de s'investir dans la lecture de l'actualité ou dans différentes activités, selon leurs centres d'intérêts.

> *J'utilise mes journées à étudier ce qui m'intéresse… Les sujets d'actu… À regarder sur les chaînes Youtube des reportages, des débats ; à m'enrichir personnellement. Voilà c'est ce que je fais de mes journées. Plutôt que d'enrichir un patron, je m'enrichis personnellement.*[138]

> *Sinon, je fais partie d'une association où je suis beaucoup investie parce j'ai plus de temps en tant que retraité, c'est un magasin coopératif. Il est dans ma commune [...]. Ça demande pas mal de temps quand on est référente produit et tout. Mais c'est intéressant, on a du bio pour 30 à 40 % moins cher que dans le commerce. On est bénévole.*[139]

> *Quand je me suis retrouvée à la retraite, j'ai commencé une autre formation pour faire de l'hypnose. En fait, plus précisément, pour devenir hypnothérapeute. Je n'ai pas exercé parce que le confinement est arrivé. Mais bon j'ai quand même fait une formation et très vite j'ai vu arriver des techniques de manipulation qui procédaient tout à fait de techniques comparables.*[140]

Il apparaît ainsi que les personnes inactives sur le plan professionnel sont davantage susceptibles d'évoluer dans des espaces sociaux peu exposés à des risques de sanctions formelles, où les interactions sont relativement détendues et assez homogènes du point de vue des opinions partagées. Cela ne veut pas dire pour autant qu'aucune attente sociale ne pèse sur eux.

---

[138] Homme, 47 ans, Bac +2, inactif, en situation de handicap, entretien réalisé le 28 février 2022.
[139] Femme, 67 ans, certificat d'étude, retraitée, entretien réalisé le 19 mai 2022.
[140] Femme, 70 ans, Bac +8, retraitée, ex-chercheuse, entretien réalisé le 15 mars 2022.

### 4.1.3. Intensité relationnelle et proximité affective

En dehors de leurs responsabilités politiques et professionnelles, les utilisateurs qui relaient des *fake news* sur Twitter peuvent aussi être conduits à jouer différents rôles au cours de leur vie quotidienne, selon leur statut marital, leur situation familiale, leurs cercles d'amis et leur appartenance à différents groupes sociaux, qu'il s'agisse d'associations locales ou de clubs de loisirs. Selon le degré d'intensité des relations qu'ils entretiennent et la proximité affective qui en découle, les utilisateurs peuvent être répartis en trois groupes distincts.

Le premier groupe est composé d'individus dont la vie sociale est surtout centrée autour de quelques liens forts (e.g. conjoint, enfants, amis proches). Ils sont davantage attachés à ne pas mettre leurs relations en péril plutôt qu'à emporter un débat. La politique est ainsi régulièrement un sujet évité, notamment en famille. Autrement, les enquêtés risquent de se retrouver isolés :

> *Moi je suis un peu en rupture avec ma famille. Je les vois de temps en temps, mais depuis longtemps il n'y a plus de dîner de Noël. J'ai l'impression d'être comme le vilain petit canard.*[141]

Un deuxième groupe se caractérise par une vie sociale plutôt centrée autour de nombreux liens faibles, mais avec peu de temps ou d'investissement émotionnel pour des relations plus proches. Ces utilisateurs, souvent des élus ou des personnes exerçant un poste avec d'importantes responsabilités, jonglent entre plusieurs obligations professionnelles, sociales et politiques, qui les éloignent des cercles intimes.

Enfin, un troisième groupe d'enquêtés se caractérise par un isolement social important. Bien que les utilisateurs étudiés n'appartiennent pas aux couches de la population les plus marginalisées, une part non négligeable d'entre eux indiquent souffrir d'isolement social. Ce ressenti est souvent associé à des situations de célibat, de maladie invalidante, de vieillesse ou fait suite à divers événements de la vie, tels qu'un divorce, un licenciement, une rupture ou un deuil. Dans de nombreux cas, ce sentiment d'isolement s'est accentué avec les périodes

---

[141] Homme, 64 ans, Bac +6, structureur financier, entretien réalisé le 23 mars 2022.

de confinement, la mise en place de mesures de distanciation sociale et l'instauration du pass sanitaire :

> *— Vous disiez que depuis le Covid vous étiez un peu isolé, est-ce que c'était le cas aussi avant ?*
> *— Euh bah oui avant le Covid oui mais là… Là j'ai coupé les ponts. Bon déjà j'étais assez isolé avant pour des raisons professionnelles, pour des raisons de complications familiales et amicales et là le Covid a empiré les choses, surtout que j'ai refusé de me faire vacciner, donc ça venait limiter encore plus. Donc le Covid a empiré les choses oui.*[142]

Dans ce contexte, Twitter devient un substitut social pour ces utilisateurs et leur offre une forme de compagnie virtuelle à différents moments de la journée, leur permettant d'éviter une solitude trop pesante, notamment au moment des repas ou de voir que d'autres sont d'accord avec eux comme l'illustrent ces deux témoignages :

> *Bah disons qu'en règle générale, j'ouvre le matin pendant le petit déjeuner comme je vis tout seul.*[143]

> *Ça fait du bien de trouver des gens, notamment pour moi qui suis célibataire et même si j'ai des amis y en a qui sont assez loin et puis ma santé ne me permet pas toujours d'être en contact et de téléphoner ou d'écrire donc oui ça fait du bien de trouver des gens, y en a plein avec lesquels je ne suis pas d'accord, qui ne partagent pas mes idées sur Twitter mais il y en a qui les partagent et ça fait du bien, on se sent moins seule.*[144]

À la différence des individus fortement intégrés à divers groupes sociaux, ces utilisateurs ne risquent pas d'être exclus par les autres puisqu'ils sont déjà marginalisés. Au contraire, ils sont souvent en quête de reconnaissance et d'un statut social (Bail, 2021). Il est ainsi possible que les individus les plus isolés encourent moins de risques que les autres à s'exprimer de façon plus radicale sur les réseaux sociaux ou à soutenir des contre-discours. En effet, l'adhésion à des théories alternatives peut permettre aux individus de rejoindre certains groupes sociaux en brûlant les ponts qui les relient à d'autres groupes. Comme l'explique Hugo Mercier (2020, p. 193), brûler les ponts avec le plus grand nombre possible de groupes concurrents peut permettre aux utilisateurs de signaler de manière crédible aux groupes restants qu'ils leur seront fidèles car ils n'ont plus d'autre option.

---

[142] Homme, 55 ans, Bac +8, à son compte, ex-chercheur, entretien réalisé le 6 mars 2022.
[143] Homme, 47 ans, Bac +2, inactif, en situation de handicap, entretien réalisé le 28 février 2022.
[144] Femme, 58 ans, Bac +3, inactive, ex enseignante, entretien réalisé le 9 juin 2022.

## 4.2. Situations d'interactions et contraintes énonciatives

Selon leur position dans l'espace social, les utilisateurs qui relaient des *fake news* ne sont pas amenés à naviguer dans les mêmes situations d'interactions. Le dispositif méthodologique mis en place dans le cadre de cette thèse n'a pas permis de suivre chaque enquêté dans toutes ses interactions quotidiennes à la manière de Nina Eliasoph (1998). Néanmoins, les observations réalisées sur Twitter ont permis d'accéder à quelques scènes de la vie quotidienne rapportées par des utilisateurs sur leur profil, et les entretiens ont permis d'interroger quelques utilisateurs sur leurs interactions hors ligne ou sur d'autres réseaux sociaux que Twitter. Trois types de situations principales ont ainsi pu être relevées. Celles-ci se caractérisent par un degré (1) de publicité, (2) de visibilité en ligne ou de (3) proximité affective, plus ou moins fort. À chacun de ces types de situations correspondent des opportunités discursives et des contraintes énonciatives divergentes.

### 4.2.1. Prise de parole en public et enjeux réputationnels

Plus les enquêtés occupent une place élevée dans la hiérarchie d'une organisation, plus ils sont appelés à prendre la parole dans des situations publiques requérant d'importantes ressources langagières, ainsi qu'une maîtrise fine de différentes compétences énonciatives.

Les agendas des élus sont particulièrement chargés et font peser sur eux une certaine « injonction à l'ubiquité » (Lefebvre, 2014 ; Ollion, 2019). Au quotidien, ils doivent assister à diverses réunions, participer à des émissions de télévision ou de radio, accepter de multiples *interviews*, se rendre à différentes rencontres locales sur des marchés, des brocantes ou des braderies, etc. En raison de leurs multiples responsabilités, les élus sont ainsi appelés à prendre part à de nombreuses situations de prise de parole en public au sein desquelles il est attendu qu'ils désingularisent leur énonciation tout en restant engagés et concernés (Cardon et al., 1995). Autrement dit, leurs discours ne doivent être ni trop personnels ni trop distants afin de fédérer le plus grand nombre possible de personnes.

Dans une certaine mesure, les contraintes de publicité qui pèsent sur les prises de parole et les comportements des individus au sein d'espaces publics sont renforcées par les réseaux sociaux. Désormais, chaque prise de parole ou comportement peut être enregistré, photographié, ou capturé sous forme de vidéo, parfois à l'insu des personnes. Comme l'a souligné un ancien conseiller municipal, au cours d'un entretien, cette omniprésence des technologies numériques dans l'espace public pousse les acteurs politiques à surveiller tous leurs faits et gestes avec une vigilance accrue :

> *Je ne me poserais pas la question de boire un coup, d'être bourré en public, de faire des blagues à la con ou que sais-je encore s'il n'y avait pas les réseaux sociaux. Je ne me poserais même pas la question. Par exemple, un mec qui va me voir avec un verre à une inauguration, s'il prend une photo, il peut dire que j'en ai bu dix.*[145]

Bien qu'ils n'aient pas les mêmes responsabilités publiques que les élus, les membres d'un parti politique sont généralement appelés à participer activement à la vie de l'organisation. Cela peut inclure une participation régulière à des réunions, des assemblées générales ou à d'autres événements organisés par le parti. Ceux qui ne maîtrisent pas suffisamment les codes langagiers et corporels nécessaires à ce type de situations sont souvent marginalisés et mis à l'écart. Les travaux de Raphaël Challier (2015 ; 2021) montrent bien comment les individus issus de classes populaires peinent à accéder à des responsabilités au sein de leur parti et sont souvent cantonnés à des tâches moins valorisées comme le collage d'affiches ou la distribution de tracts. Cette marginalisation peut aller jusqu'à l'interdiction explicite de prendre la parole, par exemple lors de distributions de tracts, pour éviter tout dérapage.

En ce qui les concerne, les utilisateurs qui exercent une activité professionnelle sont régulièrement amenés à prendre la parole dans des contextes formels comme des réunions, des séminaires ou des conférences. Dans ces situations, ils sont particulièrement contraints de contrôler leur image et de maîtriser leur discours pour éviter des maladresses susceptibles de nuire à leur crédibilité professionnelle. Le témoignage de Jean-Pierre, qui travaille dans le monde de la finance, illustre bien comment cette vigilance varie en fonction du degré de publicité des situations d'énonciation. Dans son milieu, explique-t-il, l'opinion politique de quelqu'un peut facilement être décelée, et il est courant d'adopter un discours neutre ou apolitique. Cette norme professionnelle vise à maintenir une image crédible et fiable mais n'a rien de systématique : lorsque la relation est plus informelle et marquée par une « connivence », Jean-Pierre indique qu'il peut se permettre davantage de liberté dans son discours.

[145] Homme, 60 ans, Bac +6, pharmacien, ex-conseiller municipal, entretien réalisé le 2 mars 2022.

> *— Bon principalement on tourne quand même autour de la chose non pas politique mais économique, mais cela dit c'est presque pareil sur beaucoup de sujets. Mais quand même on peut tenir un discours neutre, en tout cas apolitique. Je regarde beaucoup d'analystes le faire.*
> *— Et vous aussi, vous tenez plutôt un discours neutre ?*
> *— Ben ça dépend avec qui. Si par exemple je suis avec un collègue avec qui j'ai une certaine connivence, bon ben là ça ne pose pas de problème. Si c'est un client, je vais rester strictement professionnel, je vais faire très attention.*[146]

*A contrario*, les utilisateurs qui sont à la retraite ou ne sont pas encartés dans un parti politique spécifique ont peu d'occasions de s'exprimer dans des arènes de débat public. Cela ne signifie pas pour autant qu'ils sont invisibles dans l'espace politique ou médiatique. Les réseaux sociaux leur offrent la possibilité d'obtenir une forte visibilité en ligne et de participer activement aux discussions politiques. Cette participation peut s'exercer de manières diverses : certains choisissent d'utiliser un pseudonyme pour préserver leur anonymat, tandis que d'autres préfèrent s'exprimer dans des espaces de communication plus privés ou restreints. La prochaine section vise ainsi à présenter les opportunités et les risques associés à la prise de parole en ligne selon différents degrés de visibilité.

---

[146] Homme, 64 ans, Bac +6, structureur financier, entretien réalisé le 23 mars 2022.

### 4.2.2. Visibilité en ligne : gagner en popularité sans perdre en réputation

D'une façon générale, plus les utilisateurs bénéficient d'une notoriété publique importante, plus ils sont exposés à une forte visibilité sur le web et les réseaux sociaux. Par exemple, dans le cas de l'échantillon étudié, le nombre de *followers* des élus est assez élevé sur Twitter, atteignant en moyenne 12 120 abonnés, soit 2 fois plus que celui des relayeurs de *fake news* qui adhèrent à un parti sans exercer de responsabilité et près de 10 fois plus que celui des relayeurs de *fake news* qui ne sont affiliés à aucun parti. Les élus sont également très présents sur de nombreuses plateformes et disposent souvent de profils publics. Il n'est pas rare non plus qu'ils possèdent un site web ou une fiche biographique sur Wikipédia. Ainsi, au-delà d'être soumis à de nombreuses contraintes de temps et de publicité hors ligne, les élus sont aussi exposés à une forte visibilité sur le web et les réseaux sociaux.

Cette forte visibilité s'accompagne d'un risque accru de critiques. Par exemple, au-delà des possibles sanctions de leur parti ou des médias, les élus font aussi l'objet d'une forte surveillance de la part des utilisateurs des réseaux sociaux. En raison de leur audience importante, les contenus qu'ils partagent suscitent un grand nombre de réactions, et deviennent ainsi rapidement virales. Le revers de la médaille de cette forte visibilité, toutefois, est que les propos des élus sont davantage susceptibles de susciter des critiques de la part de leur audience. Ainsi, de nombreux internautes interviennent régulièrement en réponse aux tweets des élus pour qualifier leurs propos de *fake news*. Dans de nombreux cas, ces énoncés ne constituent pas à proprement parler des *fake news*, dans la mesure où il ne s'agit pas de contenus médiatique relatifs à un événement d'actualité ayant fait l'objet d'une vérification factuelle par des journalistes, mais plutôt d'opinions ou d'observations personnelles (cf. Figure 4.1). Par exemple, le tweet ci-dessous d'Yves Pozzo di Borgo s'apparente à une hypotypose : en ayant recours à une énumération, ainsi qu'au présent de narration, il offre une description réaliste, imagée et animée d'une scène donnant l'impression qu'elle est vécue à l'instant de son expression. Si cette stratégie rhétorique peut produire un effet de réel auprès de son audience, cela n'empêche pas de nombreux utilisateurs de faire montre d'un scepticisme important. De façon intéressante, les critiques exprimées sont souvent énoncées sur un ton plus familier et moins argumenté que les discours contredits. Dans la majorité des cas, le ton est même plutôt moqueur et railleur. Par exemple, certains utilisateurs interpellent les élus par leur prénom ou encore par le mot « mec » et les tutoient. Ces observations invitent à se demander si, en permettant des formes d'énonciation plus souples et relâchées, les réseaux sociaux ne faciliteraient pas l'expression de critiques par le grand public. Cette question sera approfondie dans le chapitre suivant.

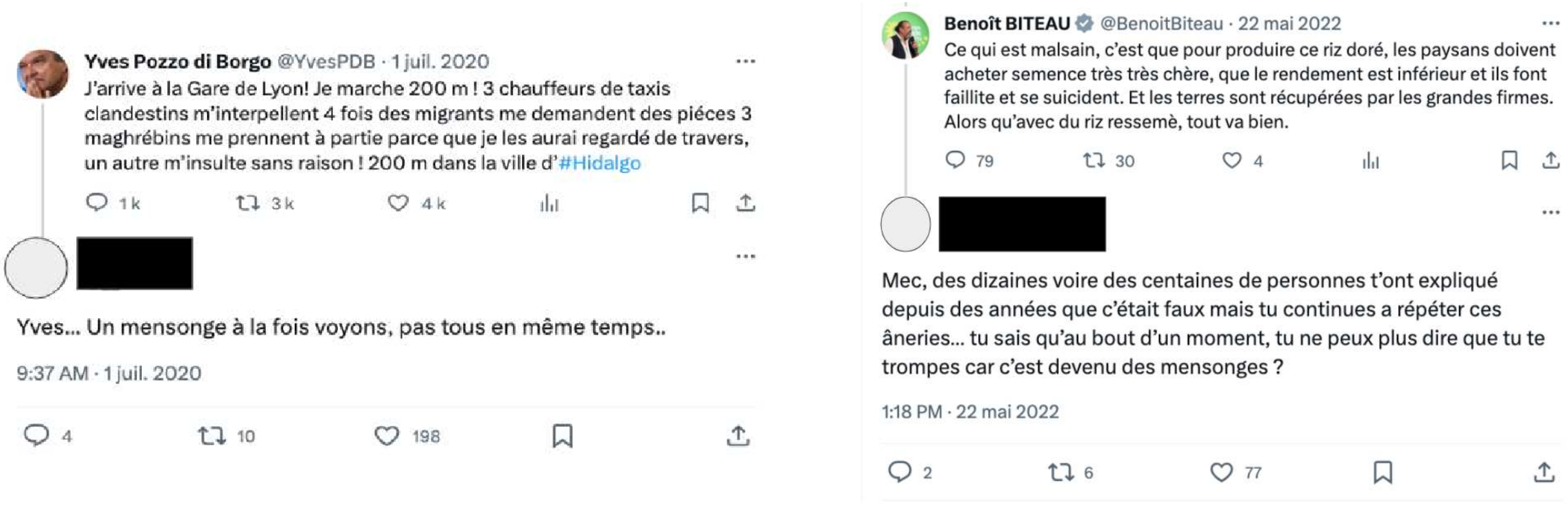

*Figure 4.1. Exemples de réponses qualifiant des tweets d'élus de fake news*

Si la forte visibilité des élus les expose à un risque accru de critiques par rapport aux utilisateurs moins visibles, ces derniers ne sont pas pour autant à l'abri de sanctions ou de rappels à l'ordre. Par exemple, bien que les militants soient plutôt amenés à prendre la parole dans des espaces de discussion informelle leur permettant de s'exprimer dans un registre familier et relâché, voire injurieux et ordurier, sans craindre de sanctions officielles de la part d'un organisation politique, puisqu'ils n'en sont pas officiellement membres, cela n'empêche pas d'observer un certain contrôle de l'expression partisane sur la toile. En effet, même en dehors des cadres formels des partis politiques, de nombreuses règles sont régulièrement rappelées aux militants en ligne que ce soit par des modérateurs (Matuszak, 2007) ou par leurs amis adhérents comme l'illustre le témoignage ci-dessous d'une enquêtée, faisant écho aux résultats d'une étude d'Anaïs Théviot (2020) :

> *On a bien vu que c'était l'extrême droite qui s'est emparée du sujet, qui a été contre cette thérapie génique, et pas tellement la gauche en fin de compte. D'ailleurs je fais partie d'un collectif où on est tous de gauche et puis ouais tous LFI, et on a interpellé Mélenchon, on a envoyé des lettres et il n'y a eu aucune réponse à ce niveau-là. Et moi j'ai des amis à moi qui m'ont reproché de relayer la propagande d'extrême droite. [...] J'ai mis sur Facebook certains articles* *justement qui expliquaient le danger de cette injection et du coup ça ne leur a pas plu. Enfin voilà, ils m'ont dit que je relayais des infos d'extrême droite alors que ben non ce n'était pas de Philippot. Enfin bon ok c'était de Martine Wonner. Je pense qu'elle est de droite. Mais elle est médecin. Pour moi on faisait de la politique mais de la politique sanitaire pas de la politique politicienne pour parler d'un parti politique. Ça a été vraiment désolant de se retrouver sur des positions défendues seulement par l'extrême droite.*[147]

Par ailleurs, si dans l'ensemble, une forte visibilité en ligne expose les utilisateurs à un risque de sanctions plus élevé, il ne faut pas oublier les multiples paramètres de visibilité proposés par les plateformes pour permettre aux utilisateurs de créer des représentations de leur identité adaptées aux différentes audiences qu'ils imaginent derrière leurs comptes (Cardon, 2008). Des fonctionnalités, telles que le recours à l'anonymat, permettent aux utilisateurs de contourner les contraintes traditionnelles de l'énonciation publique tout en bénéficiant d'une forte visibilité. Cela donne l'opportunité aux utilisateurs de s'exprimer de façon plus relâchée sans avoir à engager directement leur responsabilité énonciative. Par exemple, l'ancien conseiller municipal déjà mentionné plus haut a expliqué avoir créé deux comptes Twitter. Tandis que sur le premier, il veille à ne pas froisser son audience, il se permet sur le second d'avoir un ton plus virulent, si bien qu'il a fini par être suspendu pendant plusieurs mois :

*J'ai deux comptes Twitter : un compte institutionnel et un compte défouloir. Sur le premier, je reste assez consensuel, mais sur le second, je peux me lâcher un peu plus. Ça permet d'exprimer un avis plus tranché.*[148]

Ces pratiques de dédoublement ou de dissociation énonciative permettent ainsi aux utilisateurs de faire varier leurs régimes d'énonciation — par exemple, de passer d'un régime consensuel et lisse à un régime polémique. Il est intéressant de relever que si les discours virulents tenus par des élus via un pseudonyme peuvent toujours entraîner une suspension de leur compte ou des réprobations de leur audience, ils ne leur font pas encourir cependant de risques vis-à-vis de leur parti dans la mesure où ils sont détachées de leur identité civile. Au contraire, ces prises de parole peuvent même leur faire gagner des points de popularité auprès de leurs électeurs et militants. Par exemple, le conseiller municipal mentionné ci-dessus a indiqué que son second compte était connu par des personnes de son entourage et qu'il avait « même plus de *followers* sur ce compte ».

[147] Femme, 67 ans, certificat d'étude, retraitée, entretien réalisé le 19 mai 2022.
[148] Homme, 60 ans, Bac +6, pharmacien, ex-conseiller municipal, entretien réalisé le 2 mars 2022.

De son côté, Christian, qui ne se sent représenté par aucun parti et se montre très critique envers les formations politiques traditionnelles, a expliqué publier régulièrement des contenus traitant de sujets d'actualité politique sur Twitter mais éviter de le faire sur Facebook, où l'absence d'anonymat rend ces prises de position plus risquées socialement :

> *Sur Facebook, je ne faisais pas de commentaires politiques parce que pour le coup ce n'était pas anonyme. Donc sur Facebook si vous faites des commentaires politiques vous perdez des amis ; au contraire de Twitter où vous en gagnez peut-être parfois.*[149]

D'autres enquêtés, comme Richard, chef d'entreprise et adhérant au parti Les Républicains, dont le compte Twitter bénéficie d'une visibilité relativement importante avec un nombre d'abonnés avoisinant les 5 000, préfèrent se tourner vers des espaces de communication moins publics pour exprimer leurs opinions plus librement. Des plateformes comme Facebook leur permettent de contrôler plus facilement la visibilité de leurs publications en filtrant les demandes d'amis ou en paramétrant leur compte pour restreindre certains posts à un cercle précis de contacts :

> *J'essaie de faire attention à ce que je relaie sur Twitter. Facebook, c'est plus restreint. Les usages ne sont pas du tout les mêmes puisque Facebook on limite, on accepte les gens si on veut, etc. Twitter non. C'est pour ça que je suis plus méfiant, j'essaie de faire attention à ce que je relaie sur Twitter parce que l'incidence n'est pas du tout la même et là c'est ouvert à tous. C'est vraiment deux réseaux totalement différents.*[150]

Bien que les réseaux sociaux permettent aux utilisateurs d'utiliser des stratégies pour contourner les contraintes de publicité traditionnellement associées à une forte visibilité, notamment via le recours à des pseudonymes, ces plateformes ne sont pas exemptes de règles et engendrent parfois de nouveaux risques. D'une part, il existe un risque d'exhumation. Si l'anonymat offre une certaine protection aux utilisateurs qui s'expriment en ligne, la permanence des traces numériques rend toujours possible la récupération d'informations passées. Ainsi, les utilisateurs peuvent faire l'objet d'enquêtes minutieuses à travers des recoupements d'informations disponibles en ligne. Par exemple, Jordan Bardella a récemment été soupçonné par plusieurs médias d'avoir tenu un compte anonyme « sulfureux » sur Twitter.[151] Le RN a également retiré son soutien à l'un de ses candidats aux dernières législatives après un tweet datant de 2018 « déterré » par des journalistes de *Libération.*[152] D'autre part, les plateformes elles-mêmes imposent des régulations sur la circulation de contenus. Si l'utilisation de pseudonymes permet aux utilisateurs de s'exprimer sans risque immédiat pour leur réputation personnelle, les contenus qu'ils publient ou leurs comptes peuvent à tout moment être signalés, voire supprimés ou suspendus. Cependant, ces sanctions n'entraînent pas toujours une modération accrue des discours. Au contraire, elles peuvent exacerber la colère des utilisateurs, en renforçant leur sentiment d'être censurés, et les pousser à se replier vers des plateformes alternatives moins modérées, comme Telegram, Parler, Gettr ou VK, où les prises de parole deviennent encore plus virulentes. [153]

[149] Homme, 55 ans, Bac +8, à son compte, ex-chercheur, entretien réalisé le 6 mars 2022.
[150] Homme, 57 ans, Bac +5, chef d'entreprise, entretien réalisé le 20 mai 2022.

> *J'ai basculé sur un autre réseau social du fait d'une censure. Je me suis inscrit sur Gettr.*[154]

> *C'était des gens plein d'esprit, il y avait beaucoup de dérision. Intellectuellement, je les trouvais très brillants [...]. Ils avaient du style quoi. [...] Ils ne sont plus sur Twitter maintenant. Ils ont estimé que le combat était ailleurs. Ils sont sur Telegram.*[155]

> *Télégram j'ai utilisé un tout petit peu parce que c'est des fachos. Quand je dis facho, je veux dire que moi je suis patriote au sens de non-facho. Moi j'aime la France. Par contre, vous allez sur Telegram c'est : « mort aux noirs, mort aux arabes ».*[156]

---

[151] Guillou, C. (2024, 18 janvier). Jordan Bardella associé par « Complément d'enquête » à un compte Twitter sulfureux. *Le Monde*. https://www.lemonde.fr/politique/article/2024/01/18/jordan-bardella-associe-par-complement-d-enquete-a-un-compte-twitter-sulfureux_6211610_823448.html

[152] Libération avec AFP. (2024, 19 juin). Antisémitisme : le RN retire son soutien à l'un de ses candidats après un tweet déterré par « Libé ». *Libération*. https://www.liberation.fr/politique/elections/antisemitisme-le-rn-retire-son-soutien-a-lun-de-ses-candidats-apres-un-tweet-deterre-par-libe-20240619_VFK2I3WDG5CUDEMEZJLMALP6LM/

[153] Laurent, S. (2020, 21 juillet). Chassée de Twitter, l'extrême droite en ligne migre vers des réseaux sociaux alternatifs. *Le Monde*. https://www.lemonde.fr/societe/article/2020/07/21/chassee-de-twitter-l-extreme-droite-en-ligne-migre-vers-des-reseaux-sociaux-alternatifs_6046809_3224.html

[154] Homme, 55 ans, Bac +8, à son compte, ex-chercheur, entretien réalisé le 6 mars 2022.

[155] Femme, 70 ans, Bac +8, retraitée, ex-chercheuse, entretien réalisé le 15 mars 2022.

[156] Homme, 39 ans, Bac +3, infirmier, entretien réalisé le 20 mai 2022.

Il ressort ainsi de ces analyses que les utilisateurs qui s'expriment en ligne doivent maîtriser un subtil jeu d'équilibriste afin de maximiser la popularité de leurs énoncés tout en protégeant leur réputation personnelle. Pour cela, ils peuvent soit se conformer aux contraintes traditionnelles de l'énonciation publique en polissant leurs discours, soit tirer parti des outils proposés par les réseaux sociaux, comme le blocage de comptes, le passage en mode privé ou l'usage de pseudonymes afin de mieux contrôler leur visibilité en ligne.

### 4.2.3. Conversations interpersonnelles, risques de dispute et enjeux affectifs

Selon le degré d'intensité des relations qu'ils entretiennent et la proximité affective qui en découle, les individus ont plus ou moins l'occasion d'échanger avec leurs pairs dans des contextes conversationnels privés et informels. Dans ces situations d'interactions, les discussions sont davantage orientées vers le renforcement des liens sociaux que vers la transmission d'informations d'intérêt général et prennent ainsi souvent la forme de bavardages familiers.

Il n'est pas rare cependant que des discussions autour de sujets d'actualité émergent dans ces moments de socialisation quotidienne (Boczkowski et al., 2022). Alors que de nombreuses personnes évitent de parler de politique en public (Eliasoph, 1998), elles se sentent souvent plus libres de le faire dans des milieux d'interconnaissance et des espaces privés, par exemple autour d'un repas de famille ou d'un café entre amis (Wyatt et al., 2000). En effet, les discussions en famille ou entre amis, notamment hors des réseaux sociaux, sont souvent marquées par une faible diversité d'opinions, doublées d'une forte proximité affective. Cette homophilie favorise une énonciation souple et relâchée, voire permissive, car bien souvent ce n'est pas la factualité d'une information qui prime, ni son intérêt public, mais le renforcement des liens sociaux. Il est ainsi possible que les situations d'interactions privées et informelles favorisent le partage de *fake news* (Berriche, 2019).

Ces situations d'interactions n'empêchent pas néanmoins que des désaccords surgissent au cours d'une conversation. Par exemple, dans la famille de Sylvia, tous sont de droite et tous partagent les mêmes réticences par rapport aux vaccins contre le Covid-19. Si cette homophilie a favorisé le partage d'informations vaccino-sceptiques au cours de leurs échanges, cela n'a pas empêché certaines mises en garde et légères divergences d'être exprimées. Ces désaccords n'ont cependant pas été frontaux. Les choix des uns et des autres ont tous été respectés. En réalité, si les membres de la famille de Sylvia ont échangé des informations sur les vaccins – qu'il s'agisse de *fake news* ou de *fact-check* – c'était moins pour argumenter et contre-argumenter les uns avec les autres que pour manifester une attention bienveillante à l'égard de leur santé.

> *Dans ma famille proche on a un bon socle commun on va dire [rire]. Et avec mes amis aussi en général. On est d'accord dans les grandes lignes. Après certains se sont quand même fait vacciner parce que sinon ils ne pouvaient plus continuer à bosser même s'ils ne sont pas du tout dans le milieu médical. Ils sont dans le milieu enseignant. Mon frère est au conservatoire et ma belle-sœur à l'université. Et les enfants de mon frère se sont je crois fait tous vacciner malgré certaines mises en garde que leurs parents et mon père ont pu leur faire. Mais y a pas eu d'hostilité sur ces sujets-là.*[157]

Bien que les discussions en famille ou entre amis soient souvent marquées par une faible diversité d'opinions, il arrive parfois que des désaccords plus prononcés émergent. En cas de d'oppositions importantes, discuter de politique peut alors s'avérer être une « pratique sociale risquée » (Duchesne et Haegel, 2004, p. 884). Par exemple, dans la famille de Jocelyne, bien que tous soient de gauche, le vaccin a été un sujet de division entre elle et son mari d'un côté, et leurs enfants de l'autre.

> *Alors il y a aussi la famille. Ça a été chaud en famille. Moi j'ai 5 enfants. [...]. Au tout début je leur balançais un tas d'info. Ben j'ai dû arrêter. Parce que je me suis faite… Enfin voilà ça s'est mal passé avec eux. C'est passé, on est en bon terme et tout, mais ça a été une période difficile [...]. On commence à en parler et il y a toujours quelqu'un pour dire : « non, stop, allez machin » [...] Mon fils avait institué une amende sur WhatsApp en disant : « le premier qui parle du Covid, il a une amende ». Il avait fait une caisse, une tirelire à la maison, donc j'ai récupéré les sous. Ça aurait pu être pire mais à un moment donné, c'était vraiment chaud.*[158]

---

[157] Femme, 58 ans, Bac +3, inactive, ex enseignante, entretien réalisé le 9 juin 2022.
[158] Femme, 67 ans, certificat d'étude, retraitée, entretien réalisé le 19 mai 2022.

Au même titre que la politique ou la religion, le Covid est ainsi devenu un sujet de conversation à éviter dans de nombreuses familles pendant la pandémie.[159] L'idée du fils de Jocelyne de mettre en place une amende dans le groupe WhatsApp familial témoigne d'un effort pour détourner la conversation de conflit par un stratagème subtil visant à préserver les liens affectifs.

Il est intéressant de noter que, bien que Jocelyne et son mari fassent souvent front commun contre leurs enfants dans leur groupe WhatsApp familial, son mari n'hésite pas à lui signaler de temps à autre qu'il trouve qu'elle va trop loin lorsqu'ils sont seul à seul. Par exemple, Jocelyne a expliqué, avec un brin d'amusement dans la voix, que dès qu'elle mentionne au cours d'une conversation que Bill Gates finance les campagnes de vaccination contre le Covid-19 pour réduire la population mondiale, son mari la tempère immédiatement : « même mon mari qui n'est pas vacciné, qui entend des choses, il va me dire : "non là t'exagères" ». Le sourire aux lèvres de Jocelyne au moment de raconter cette anecdote laisse percer une forme de tendresse et d'indulgence dans le « t'exagères » de son mari, suggérant que la formulation d'une critique n'est pas forcément vécue comme une forme de sanction.

Ainsi, bien que les groupes familiaux et amicaux soient souvent perçus comme des « chambres d'échos », ils peuvent aussi parfois servir d'instances de régulation de la parole. Si, au sein de ces espaces de conversation informels, les sujets de désaccord sont généralement écartés pour éviter les disputes, cela n'empêche pas l'émergence occasionnelle d'interventions critiques. D'une part, il est possible qu'il soit plus facile et moins risqué d'exprimer des critiques dans ces situations informelles car les propos des individus n'ont pas nécessairement besoin d'être argumentés ou sourcés pour être jugés recevables par leurs pairs ; d'autre part, ces désaccords n'entraînent pas nécessairement une rupture du lien social puisqu'ils peuvent être exprimés avec tendresse et indulgence ou animés par des sentiments d'affection et de prévenance. Toujours est-il que ces interventions critiques peuvent inciter les participants à des discussions informelles à modérer leurs discours. Ces processus d'autorégulation seront approfondis dans les deux prochains chapitres.

---

[159] Cordier, S. (2021, 18 Septembre). Au sein des familles, le Covid, c'est pire que la religion et la politique. *Le Monde*. https://www.lemonde.fr/societe/article/2021/09/18/au-sein-des-familles-le-covid-c-est-pire-que-la-religion-et-la-politique_6095160_3224.html

## 4.3. Mise en scène de soi et variations énonciatives sur Twitter

Les analyses réalisées dans les deux sections précédentes ont permis de positionner les utilisateurs étudiés sur trois axes principaux permettant de cerner leur position dans l'espace social selon (1) leur mode d'affiliation partisane ; (2) leur degré d'insertion professionnelle ; (3) l'intensité de leurs relations sociales. En fonction de ces positions, les utilisateurs sont amenés à recevoir ou à partager des informations dans des situations d'interactions variées caractérisées par un degré (1) de publicité, (2) de visibilité ou de (3) proximité affective, plus ou moins fort. Par exemple, les élus se trouvent régulièrement dans des situations de prise de parole en public et sont exposés à une forte visibilité en ligne, mais disposent généralement de peu de temps à consacrer aux relations sociales. À l'inverse, les utilisateurs sans attache politique ou inactifs sur le plan professionnel disposent de plus de temps pour converser avec leurs proches dans des cadres informels, mais sont rarement amenés à prendre la parole au sein d'espaces publics.

Chaque position dans l'espace social est associée à une série d'attentes normatives, auxquelles viennent s'ajouter des contraintes énonciatives spécifiques à chaque situation d'interactions. Cette section propose d'analyser si ces différentes attentes normatives et contraintes énonciatives façonnent les comportements et les pratiques des utilisateurs sur Twitter. Plus précisément, il s'agit d'examiner si les règles intériorisées par les individus à travers leurs interactions sociales, notamment hors ligne, se reflètent dans leurs usages de Twitter et modulent (1) la manière dont ils se présentent sur leur profil, (2) leurs pratiques informationnelles et numériques (y compris leur propension à partager des *fake news*), ainsi que (3) leurs régimes d'énonciation.

### 4.3.1. Des stratégies de présentation de soi différenciées

D'emblée, un fort contraste apparaît entre la façon dont les élus se présentent sur Twitter et l'image générale des relayeurs de *fake news* vue au chapitre précédent. Parmi les douze comptes d'élus identifiés, un seul utilise un pseudonyme dans son nom d'utilisateur, ce qui représente 8 % de l'échantillon contre 66 % pour l'ensemble des relayeurs de *fake news*. Par ailleurs, presque tous affichent leur visage en photo de profil ou de couverture. La majorité des élus s'expriment donc en leur nom propre et engagent ainsi leur responsabilité énonciative en publiant des tweets. Dans la plupart des cas, leurs photos de profil et de couverture semblent avoir été prises par des professionnels de la communication, plutôt que de façon spontanée par des amateurs. En effet, sur ces clichés la majorité porte des tenues formelles et pose d'un air sérieux dans un cadre mettant en avant leurs responsabilités politiques et professionnelles. À ce titre, le cas de Jonas est très illustratif. Alors que dans sa bio Twitter, celui-ci indique être à la fois pharmacien hospitalier, conseiller municipal de l'opposition dans la mairie de sa commune et remplaçant au conseil départemental de son département, ses photos de profil et de couverture reflètent ses différentes casquettes : en blouse blanche sur sa photo de profil, on voit qu'il pose puisqu'il fait mine de signer un document de travail protégé par une pochette plastique transparente tout en levant les yeux vers la personne qui le photographie ; en blazer bleu marine sur sa photo de couverture, il se tient aux côtés de trois autres personnes pour afficher sa candidature aux élections départementales sur une liste d'un groupe Divers Droite et Centre (cf. Figure 4.2). Ainsi, bien que Twitter donne l'opportunité aux élus de se présenter sous un jour personnel, de manière conviviale et informelle, en favorisant une apparente proximité et spontanéité avec leurs *followers*, ils utilisent surtout la plateforme pour se présenter sous un jour professionnel (Roginsky et De Cock, 2015). On retrouve ainsi sur Twitter comme sur d'autres espaces des stratégies identitaires mises en place par les élus pour contrôler leur image publique (Collovald, 1988 ; Planche et al., 2022).

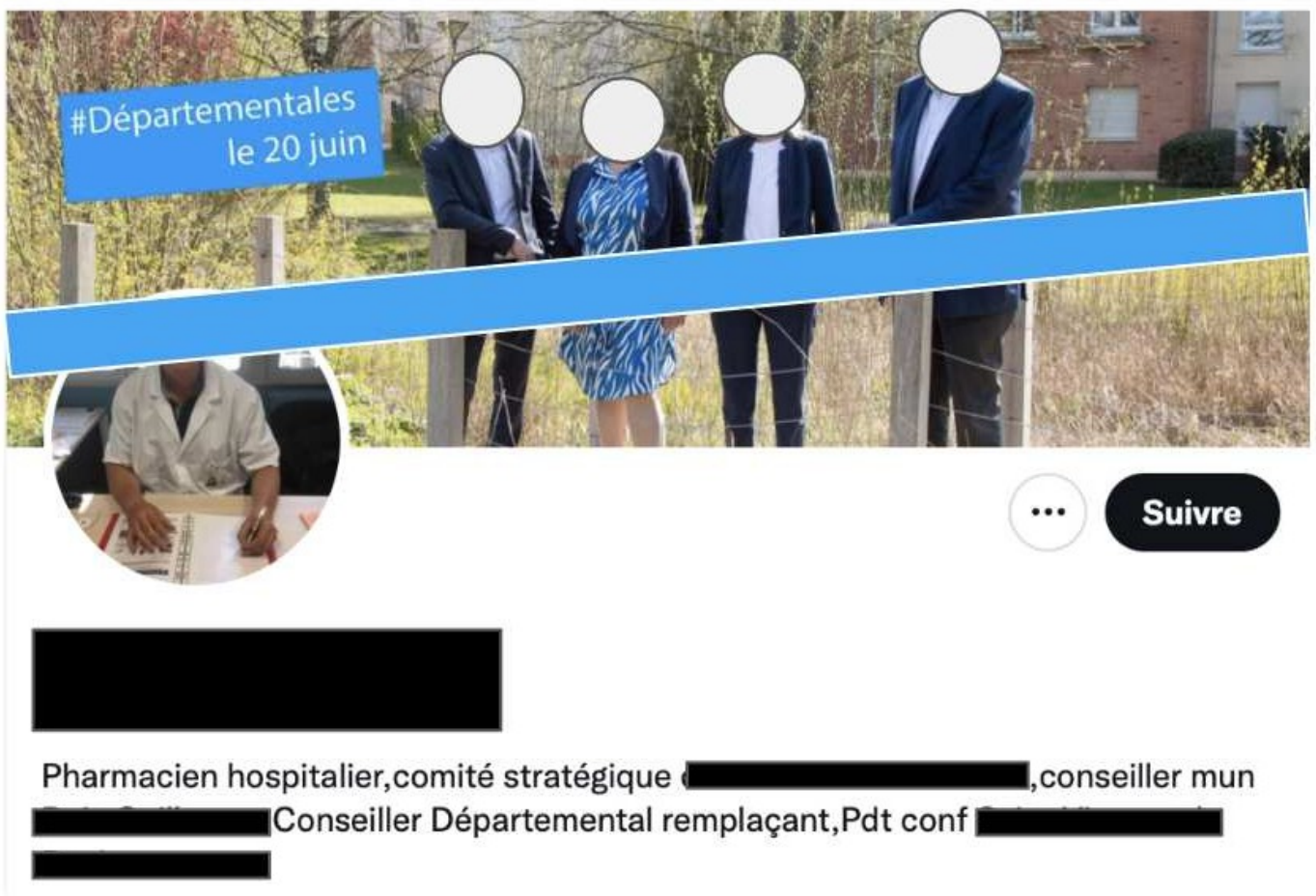

*Figure 4.2. Exemple de profil Twitter d'un élu politique*

Par rapport aux élus, les adhérents des partis politiques ont beaucoup plus fréquemment recours à des pseudonymes, bien que cette pratique soit moins répandue chez eux que chez l'ensemble des autres relayeurs de *fake news*. En effet, environ 50 % des adhérents utilisent un pseudonyme contre 65,7 % pour les autres relayeurs de *fake news*. En ce qui concerne leurs photos de profil et de couverture, on observe un style moins formel et professionnel que celui des élus. Leurs photos de profil affichent souvent des bannières ou logos de leur parti politique (cf. Figure 4.3), tandis que ceux qui choisissent de montrer leur visage optent généralement pour des photos amateurs, prises dans des contextes plus personnels. Ces choix de présentation révèlent une stratégie identitaire moins encadrée par les codes de la communication politique institutionnelle, tout en maintenant un lien visible avec leur affiliation partisane.

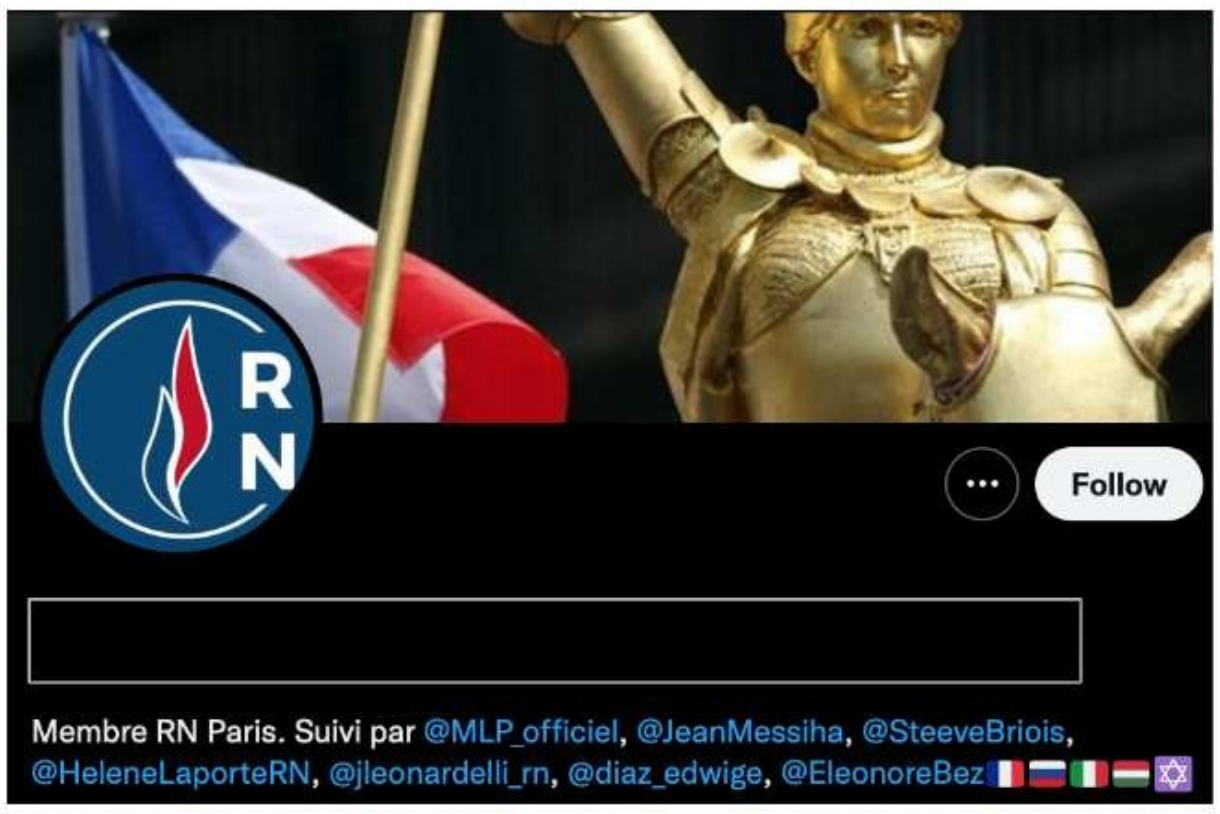

*Figure 4.3. Exemple de profil Twitter d'un adhérent au RN*

La première image renvoyée par les militants via leur profil Twitter contraste fortement avec celle des élus et des adhérents. Plutôt que d'afficher le parti politique qu'ils soutiennent, ils cherchent surtout à mettre en avant leurs préoccupations personnelles. Leur communication est purement individuelle et ne fait pas appel à des ressources collectives partisanes (Enli et Skogerbo, 2013). Ils sont 76 % à masquer leur identité civile par un pseudonyme et très peu utilisent des photos de profils ou de couverture montrant leur visage ou le logo d'un parti politique. Plusieurs d'entre eux ont plutôt opté pour des photos représentant symboliquement des causes qu'ils défendent. Par exemple, le compte ci-dessous, tenu par une militante écologiste, expose sur son profil une photo de coquelicot, un emblème des revendications contre l'utilisation des pesticides, et différents badges ou autocollants indiquant qu'elle est concernée par plusieurs problèmes environnementaux (cf. Figure 4.4).

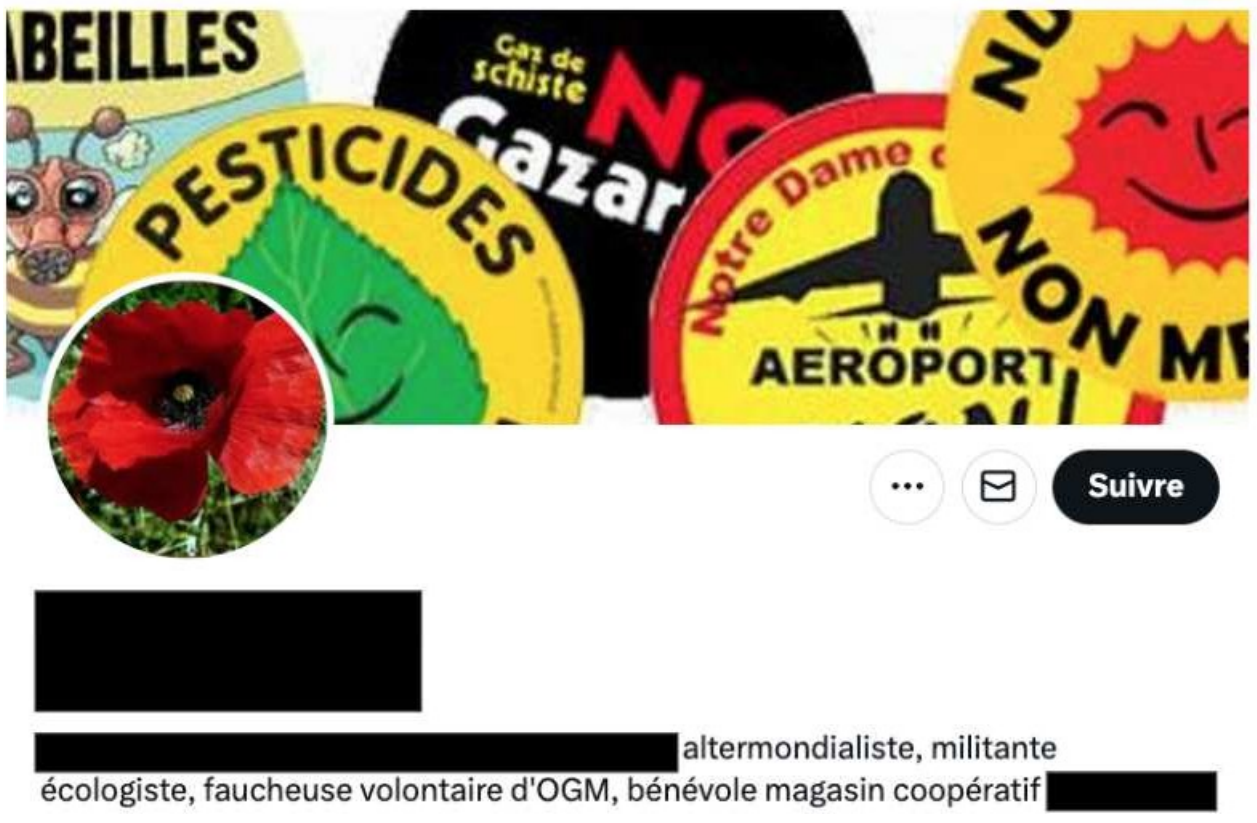

*Figure 4.4. Exemple de profil Twitter d'une militante*

Enfin, de leur côté, les profils des relayeurs de *fake news* dépourvus d'attache partisane spécifique ne cherchent pas à mettre en avant leur appartenance à des groupes sociaux, ni les rôles qu'ils jouent dans la société, mais leur opposition au système politique. Bon nombre d'entre eux indiquent dans leur bio Twitter qu'ils sont « non encarté[s] », « sans obédience », « sans étiquette », voire « inclassable[s] » ou affichent leur rejet des institutions via leur photo de profil et de couverture (cf. Figure 4.5). Par ailleurs, près de 92% ont recours à un pseudonyme, soit 27 points de plus que l'ensemble des relayeurs de *fake news,* et très peu d'entre eux dévoilent des informations sur leur profession ou leur âge à travers leurs tweets. Parmi les quelques internautes qui donnent des informations sur leur statut, près de la moitié sont inactifs, un quart sont des CSP+ et l'autre quart sont des CSP-. Il ressort ainsi que les relayeurs de *fake news* dépourvus d'attache partisane sont moins intégrés socialement que les élus ou les adhérents, mais n'appartiennent pas non plus à des groupes sociaux aussi défavorisés que les abstentionnistes (Lancelot, 1968 ; Gaxie, 1993 ; Héran, 1997 ; Chiche et Dupoirier, 1998). Concernant les utilisateurs qui ne révèlent aucune information sur leur statut, il est difficile de savoir si c'est parce qu'ils craignent pour leur réputation ou si c'est parce qu'ils exercent des professions qui ne sont pas valorisées socialement. Quoi qu'il en soit, on peut faire l'hypothèse que l'utilisation de pseudonymes, ainsi que l'absence de marqueurs sociaux, leur permettent d'exprimer leurs opinions sans risquer de compromettre le « pacte de stabilité sociale » qui régule d'autres relations, par exemple les relations de voisinage ou les relations amicales (Dutton, 1996).

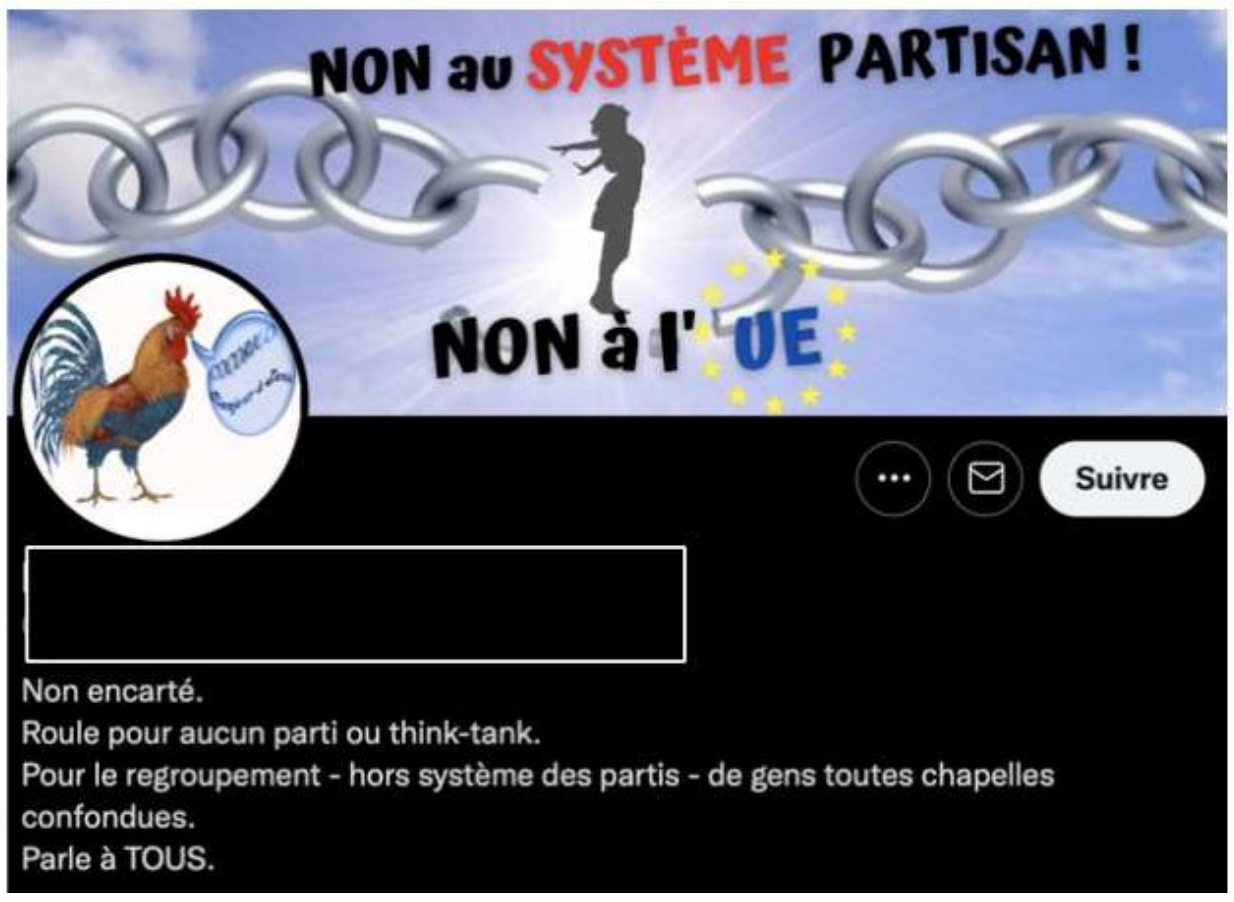

*Figure 4.5. Exemple de profil Twitter d'un utilisateur sans attache partisane*

L'analyse des stratégies de présentation de soi des différents groupes de relayeurs de *fake news* menée dans cette section a mis en lumière des différences significatives selon leur niveau d'affiliation à un parti politique et leur degré d'exposition publique. En revanche, il n'a pas été possible de réaliser des comparaisons aussi précises en fonction de leur profession ou de leur type de sociabilité. Il est néanmoins ressorti que les utilisateurs les plus isolés étaient plus enclins à utiliser un pseudonyme ou à adopter des photos de profil affichant un message contestataire ou militant.

### 4.3.2. Des pratiques informationnelles et numériques divergentes

Après avoir mis en avant des différences dans les stratégies de présentation de soi sur Twitter des relayeurs de *fake news* en fonction de leur position sociale et des contraintes énonciatives qu'ils ont pu intérioriser au fil de leurs expériences, voyons maintenant si des divergences se retrouvent également au niveau de leurs pratiques informationnelles et numériques. Plus précisément, il s'agit d'analyser leur activité sur Twitter et la fréquence à laquelle ils partagent des *fake news*.

Sur Twitter, l'activité des élus est quatre fois moins importante que celle des utilisateurs qui ne sont pas directement affiliés à un parti politique ou qui déclarent être sans attache partisane. En effet, alors que les élus émettent en moyenne 26 (re)tweets par jour, les utilisateurs militants et dépourvus d'attache partisane en font respectivement 103 et 107. De leur côté, les adhérents sont légèrement plus actifs que les élus mais nettement moins que les militants et les sans attache avec un nombre moyen de (re)tweets par jour de 38.

L'intensité de l'activité des utilisateurs semble également fortement liée à leur degré d'intégration à une organisation professionnelle et aux sanctions ou à la disponibilité de temps qui peuvent en découler ou non. Par exemple, après avoir connu quelques mauvaises expériences professionnelles à la suite de prises de positions politiques, Willam a décidé de ne plus publier de contenus en lien avec la politique sur les réseaux sociaux.

> *Je n'écris pratiquement plus parce qu'à chaque fois j'ai l'impression d'avoir une mauvaise expérience. Je m'autocensure complètement. Dès qu'on tweete sur la politique ou sur n'importe quel type d'actualité, ça peut toujours déraper. Donc je m'abstiens…* [160]

*A contrario*, deux militants, l'un pour La France Insoumise, l'autre pour Reconquête, tous deux inactifs professionnellement à cause d'une maladie invalidante, ainsi que célibataires, ont confié se sentir « addicts » à Twitter et ont souligné qu'ils avaient du mal à limiter leur utilisation du réseau social.

> *Ça devient presque addictif au bout d'un moment. Donc si on ne s'impose pas un minimum de pause et de distance par rapport à ça, ça peut être très chronophage, trop chronophage on va dire.* [161]

> *Alors je ne m'abonne pas à tous ceux qui s'abonnent à mon compte mais quand même il y en a beaucoup que je trouve intéressants et finalement bon ben j'aurais envie de tout voir. C'est ça qui peut m'emmener au bout de la nuit.* [162]

L'écart observé entre Viviane et son mari Jean-Pierre est également très parlant. À la retraite depuis 2 ans, Viviane, est très prolixe sur Twitter avec une moyenne de plus de 100 (re)tweets par jour. À l'inverse son mari, toujours en activité, a indiqué avoir uniquement un Twitter de retweets ou de consultation :

> *Je ne mets jamais de commentaire. Je ne fais que lire. Je n'ai pas tellement le temps d'écrire. Ni l'intérêt. Ma femme, elle, veut faire changer d'avis l'autre. Moi sur de nombreux sujets j'ai renoncé depuis longtemps. En tout cas je pense que ça ne passe pas du tout par ce genre de canal.*[163]

[160] Homme, 53 ans, Bac +8, chercheur, entretien réalisé le 18 mai 2022.
[161] Homme, 47 ans, Bac +2, inactif, en situation de handicap, entretien réalisé le 28 février 2022.
[162] Femme, 58 ans, Bac +3, inactive, ex enseignante, entretien réalisé le 9 juin 2022.

Si ces écarts importants en termes d'activité suggèrent que les élus, les adhérents ou les personnes exerçant une activité professionnelle ont un usage de Twitter plus modéré que les autres relayeurs de *fake news*, il ne signifie pas non plus qu'ils fassent preuve d'une grande retenue dans la mesure où leur activité reste très importante par rapport à celle d'un utilisateur ordinaire.

Le nombre de *fake news* partagées sur Twitter varie également selon les positions sociales des utilisateurs. Le nombre de *fake news* partagé par les comptes d'élus sur Twitter est 2 à 3 fois moins élevé que celui des militants et des utilisateurs sans attache partisane. Ils en ont partagé 5 en moyenne contre 10 et 15 pour les militants et les comptes sans attache partisane. Ces *fake news* sont souvent issues de médias partisans plutôt que de médias de contre-information, et sont moins dirigées contre les élites que contre les émigrés et délinquants, comme l'illustre le contenu ci-dessous (à gauche), publié par le média *Valeurs Actuelles* et relayé par l'un des élus de notre échantillon, ainsi que par d'autres personnalités politiques comme Nicolas Dupont-Aignan, Valérie Boyer ou Nicolas Bay. Si ce contenu a été qualifié de *fake news* par des *fact-checkers* de l'AFP[164], de nombreux autres tweets analogues (mettant également en avant un cas de violence sexuelle où un étranger est accusé) ont été publiés sur les comptes des élus de notre échantillon, mais n'ont pas fait l'objet de *fact-check*. Par exemple, le tweet ci-dessous (à droite) de Valérie Boyer a été relayé par le même élu que celui ayant partagé le contenu de *Valeurs Actuelles.* On voit ainsi comment les *fake news* partagées par les élus s'inscrivent dans leurs pratiques discursives globales. Plutôt que de diffuser délibérément des informations fausses pour promouvoir leurs idées, ils semblent surtout enclins à recourir à des stratégies de communication, telles que le fait de tirer des généralités à partir de faits divers, leur permettant de mettre à l'agenda certains sujets qui défendent les valeurs de leurs partis. Ces pratiques soulignent l'importance pour les travaux de sciences sociales de ne pas se focaliser sur la seule question de la factualité d'une information et de tenir compte plus largement du rôle que jouent les usages médiatiques, en particulier des élites, dans le cadrage idéologique de certains débats.

[163] Homme, 64 ans, Bac +6, structureur financier, entretien réalisé le 23 mars 2022.
[164] Daudin, G., & Coupeau, C. (2018, 28 novembre). Non, un homme n'a pas été acquitté de viol parce qu'il n'"avait pas les codes culturels". *AFP*. https://factuel.afp.com/non-un-homme-na-pas-ete-acquitte-de-viol-parce-quil-navait-pas-les-codes-culturels

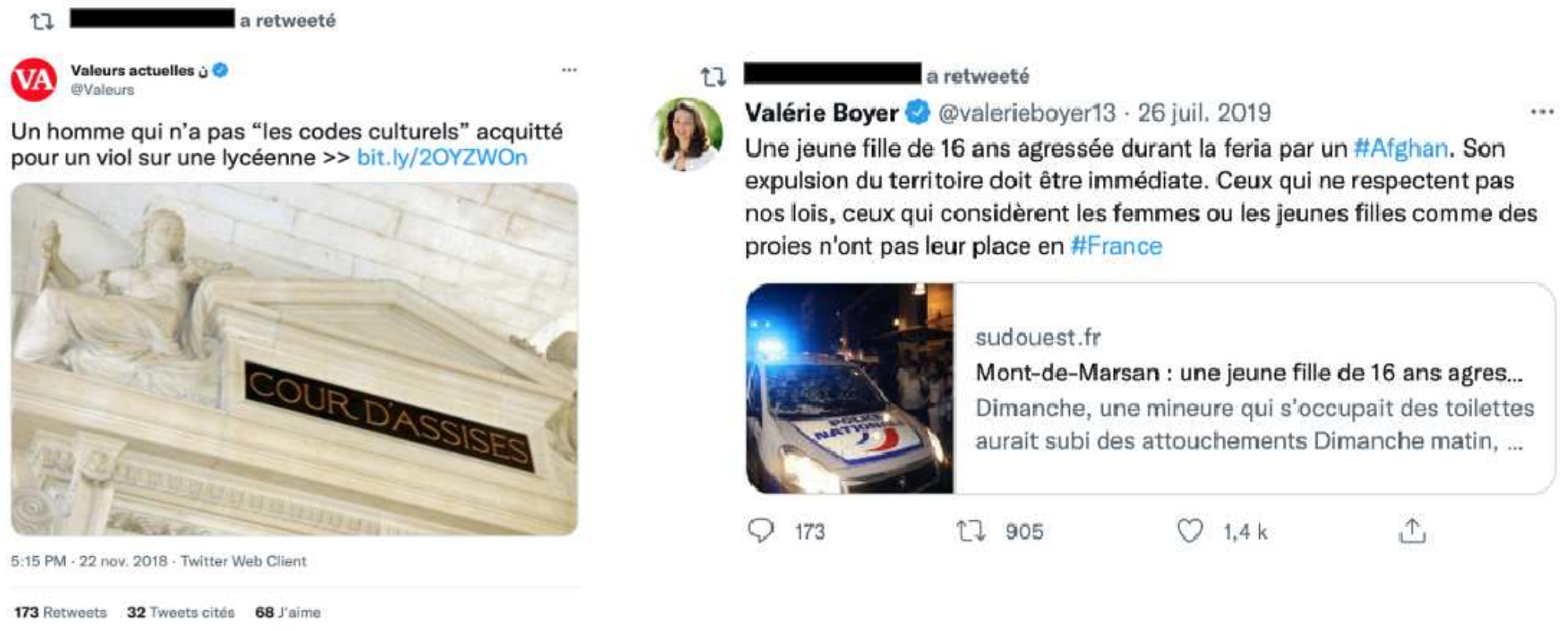

*Figure 4.6. Exemples d'une fake news (à gauche) et d'un énoncé analogue mais non qualifié de fake news (à droite) partagés par un élu*

Les *fake news* relayées par les utilisateurs militants ou sans attache partisane proviennent surtout de médias de contre-information et ont plus tendance à cibler les élites au pouvoir, à remettre en cause des résultats de sondages ou à dénoncer des scandales financiers en adoptant parfois une rhétorique complotiste. Quelques utilisateurs ont par ailleurs surtout partagé des *fake news* liées à des sujets de santé critiquant particulièrement les vaccins. Parmi ces utilisateurs, certains sont devenus très actifs sur Twitter seulement à partir de 2020, suggérant que la période de la pandémie a été pour eux un moment de politisation important. Néanmoins, des recherches plus approfondies, reposant notamment sur des entretiens, seraient nécessaires pour étudier si certains utilisateurs sont devenus politisés après avoir été exposés à des *fake news* ne concernant pas directement des sujets politiques et pour comprendre de façon plus fine et nuancée si d'autres caractéristiques que celles vues aux chapitre 3 sont associées au partage de *fake news* relatives à des sujets de santé.

Enfin, il convient également de souligner que le nombre moyen de *fake news* partagées par les utilisateurs exerçant une activité professionnelle est inférieur (~9) à celui des utilisateurs sans activité (~12). Cela suggère que des mécanismes liés à la réputation ou à la disponibilité de temps pourraient soit favoriser, soit inhiber le partage de *fake news*, ou du moins moduler la prise de risque associée aux pratiques de partage d'informations en ligne.

### 4.3.3. Des modes d'énonciation contrastés

Au-delà de l'intensité de l'activité des utilisateurs sur Twitter et du nombre de *fake news* qu'ils partagent, il est essentiel d'examiner les types d'usages qu'ils ont du réseau social.

En regardant de plus près le fil Twitter des élus, il ressort que l'essentiel de leur activité consiste à (1) retweeter des contenus dans lesquels ils sont identifiés ou mentionnés par les comptes de leur parti, ou de leurs adhérents et militants, pour mettre en avant leurs actions et leurs discours — une pratique que l'on peut qualifier par le terme d'« ego retweet » (boyd et al., 2010) — et (2) à publier de nombreuses photos d'eux en train de faire campagne ou les représentant dans l'exercice de leurs fonctions (cf. Figure 4.7), ce qui leur permet de signifier leur « conformité à un ensemble de normes qui définissent [leur] rôle et qui confortent l'image que [les] publics se font de ce rôle. [En effet, les] réunions politiques, les meetings électoraux, les présentations de bilans ou autres inaugurations, sont autant de moments où, à travers ses discours et ses actes, un élu mobilise un ensemble de signes et de symboles qui accréditent l'image qu'il veut donner de lui-même et celle qu'il anticipe être attendue par ses interlocuteurs » (Briquet, 1994, p. 19). Sur ces clichés, les élus taguent ou mentionnent fréquemment le compte Twitter du parti politique auquel ils sont rattachés, ainsi que ceux de leurs collègues. Autrement dit, leurs stratégies de promotion et de mise en scène ne se limitent pas à leur propre personne, mais s'étendent également à leur parti. En partageant des contenus sur Twitter, ils n'engagent donc pas seulement leur propre responsabilité en tant que locuteur, mais impliquent également celle de leur parti.

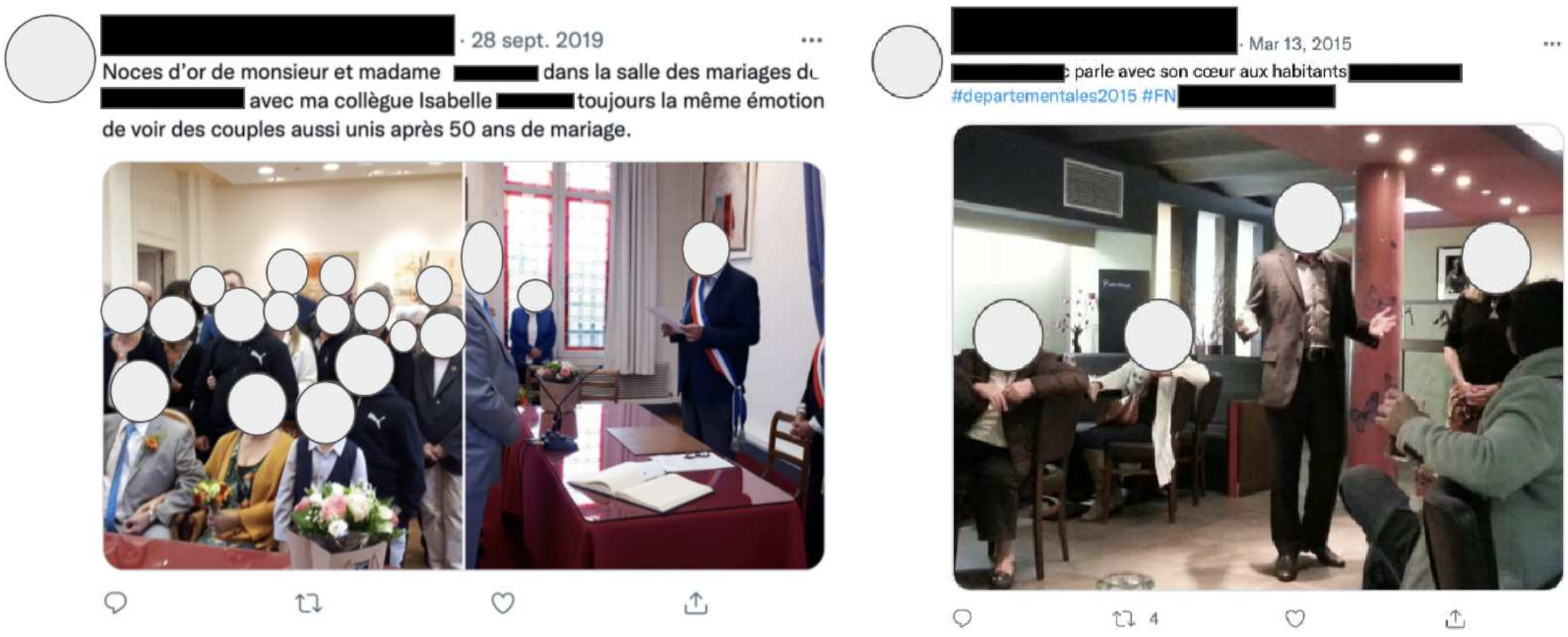

*Figure 4.7. Exemples de photos publiées par des élus*

Alors que les élus ont tendance à rédiger leurs propres tweets, souvent en concertation avec leur plume, plutôt que de retweeter ceux des autres utilisateurs, les adhérents vont plutôt être dans le retweet « sec » des comptes officiels des élus et de leur parti, notamment en période de campagne électorale pour inciter leurs *followers* à voter pour leur candidat favori. Ils se contentent ainsi de partager les discours de leur parti sans ajouter de message personnel. Autrement dit, ils se font le relais d'une parole institutionnelle tout en effaçant leur personne de l'énonciation. Les publications des adhérents sont donc assez prudentes, du moins par rapport aux attentes de leur parti, puisqu'ils se contentent de diffuser les discours de leur organisation sans y ajouter de touche personnelle.

Les très nombreuses publications des militants sont principalement constituées de retweets. À la différence des adhérents, ils ne se contentent pas de relayer uniquement les tweets provenant de comptes officiels d'institutions ou de personnalités politiques, mais sont aussi très enclins à partager des tweets émanant de personnes ordinaires. Cette tendance suggère qu'ils accordent une moins grande attention à l'autorité des sources dont ils relaient les discours que les membres d'un parti, et qu'ils souhaitent surtout mettre en avant des prises de parole individuelles. Non sans ambivalence, cette forte propension aux retweets dénote un comportement impulsif mais pas totalement imprudent de leur part. En effet, d'un côté, cela indique que les militants ne prennent pas le temps de formuler leurs propres messages, ni de lire attentivement toutes les informations qu'ils relaient. De l'autre, cependant, cela leur permet de ne pas être tenus pour responsables des propos rapportés.

L'énonciation des militants sur Twitter se caractérise aussi par l'adoption d'un registre beaucoup plus familier que celui des élus et des adhérents. Ils utilisent de façon abondante des émojis, des mèmes et des gifs. Par ailleurs, leurs publications ne portent pas que sur l'actualité politique, mais concernent aussi des aspects plus personnels de leur vie quotidienne. Par exemple, les profils des militants sont souvent ornés d'images mettant en scène des animaux de compagnie, que ce soit à travers la photo d'un chat lové sur un canapé ou celle d'un chien jouant dans un parc. Ils partagent aussi de nombreux panneaux de citations (Pasquier, 2018, p. 85-93) remplissant une fonction phatique (Berriche et Altay, 2020), sauf qu'à la différence des internautes ordinaires, ils ne les adressent pas aux membres de leur famille ou à leurs amis, mais plutôt à leur cercle militant ou aux personnalités politiques qu'ils soutiennent (cf. Figure 4.8).

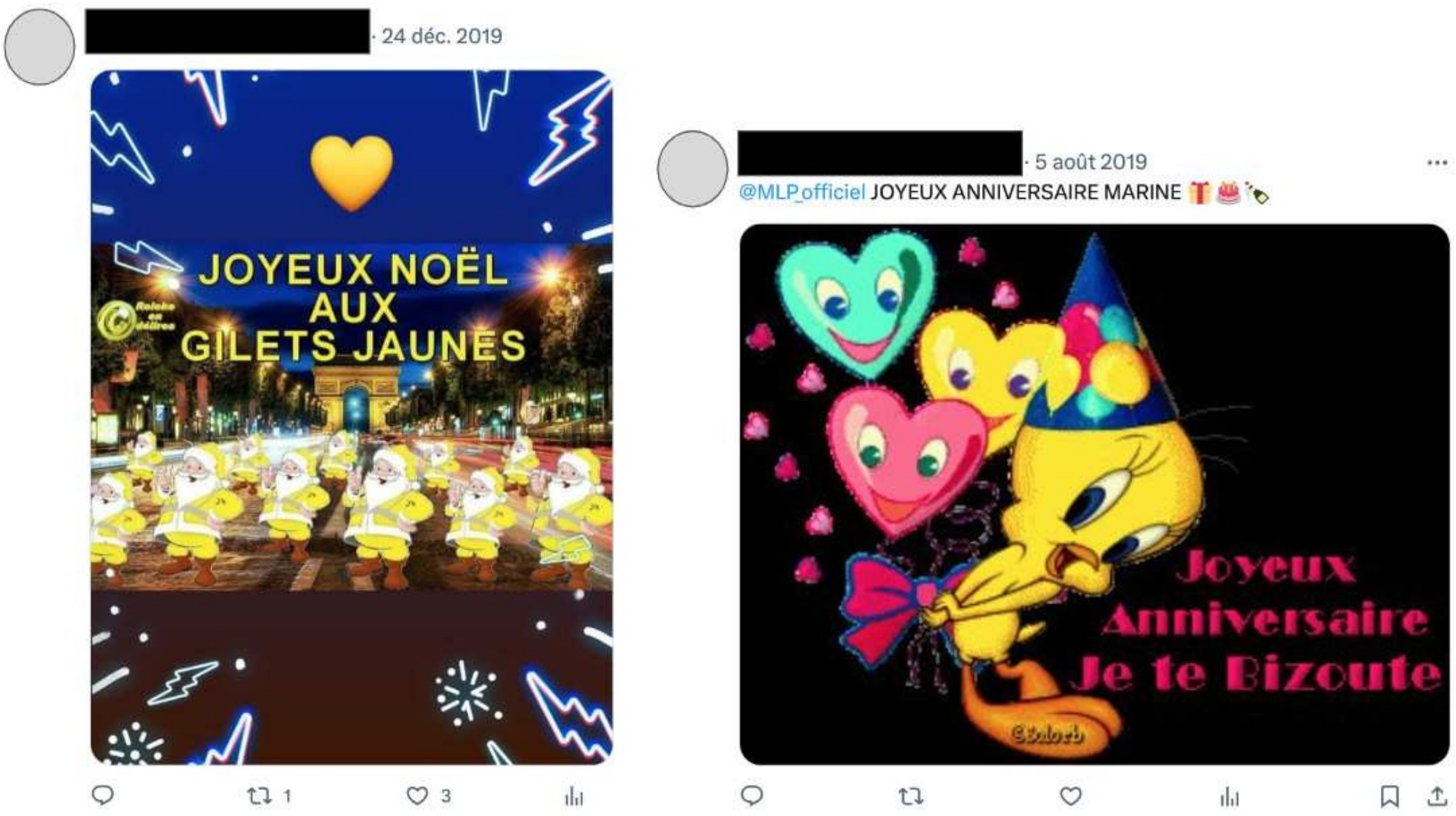

*Figure 4.8. Exemple de « panneaux de citation » publiés par des comptes de militants*

Tout comme les militants, les utilisateurs dépourvus d'attache partisane sont nettement plus actifs sur Twitter que ceux qui exercent des responsabilités politiques ou qui sont membres d'un parti. La plupart des tweets qu'ils relaient ne sont pas émis par des personnalités politiques, des institutions ou des médias, mais plutôt par des utilisateurs ordinaires qui s'expriment à la première personne du singulier. En fait, ils se distinguent surtout des militants, au niveau de leurs usages de Twitter, de par leur forte propension à relayer et à écrire des *replies*, c'est-à-dire des réponses rédigées en réaction à d'autres tweets. Cette pratique est très rare chez les autres relayeurs de *fake news* et dénote une posture critique de leur part. En effet, à la différence des *tweets* (expression de sa propre opinion) ou *retweets* (relais d'une autre opinion), les *replies* ou *quotes* peuvent avoir une position critique à l'égard d'un autre *tweet* (Roth et al., 2021 ; Hargittai et al., 2008). Il semblerait ainsi qu'ils utilisent le signe @ pour apostropher des membres du pouvoir et se faire entendre. Contrairement aux élus, adhérents et militants qui cherchent à donner de la visibilité à des énoncés de leur parti ou candidat favori, les utilisateurs sans attache partisane cherchent à être vus, ou plutôt entendus, par des comptes bénéficiant d'une forte visibilité. Par exemple, certains internautes interpellent fréquemment des comptes comme @BFMTV ou @JLMelenchon et dénoncent leurs « mensonges » en ayant recours à des gifs (cf. Figure 4.9) qui remplacent certaines expressions familières comme « c'est du pipeau » ou « mon œil ».

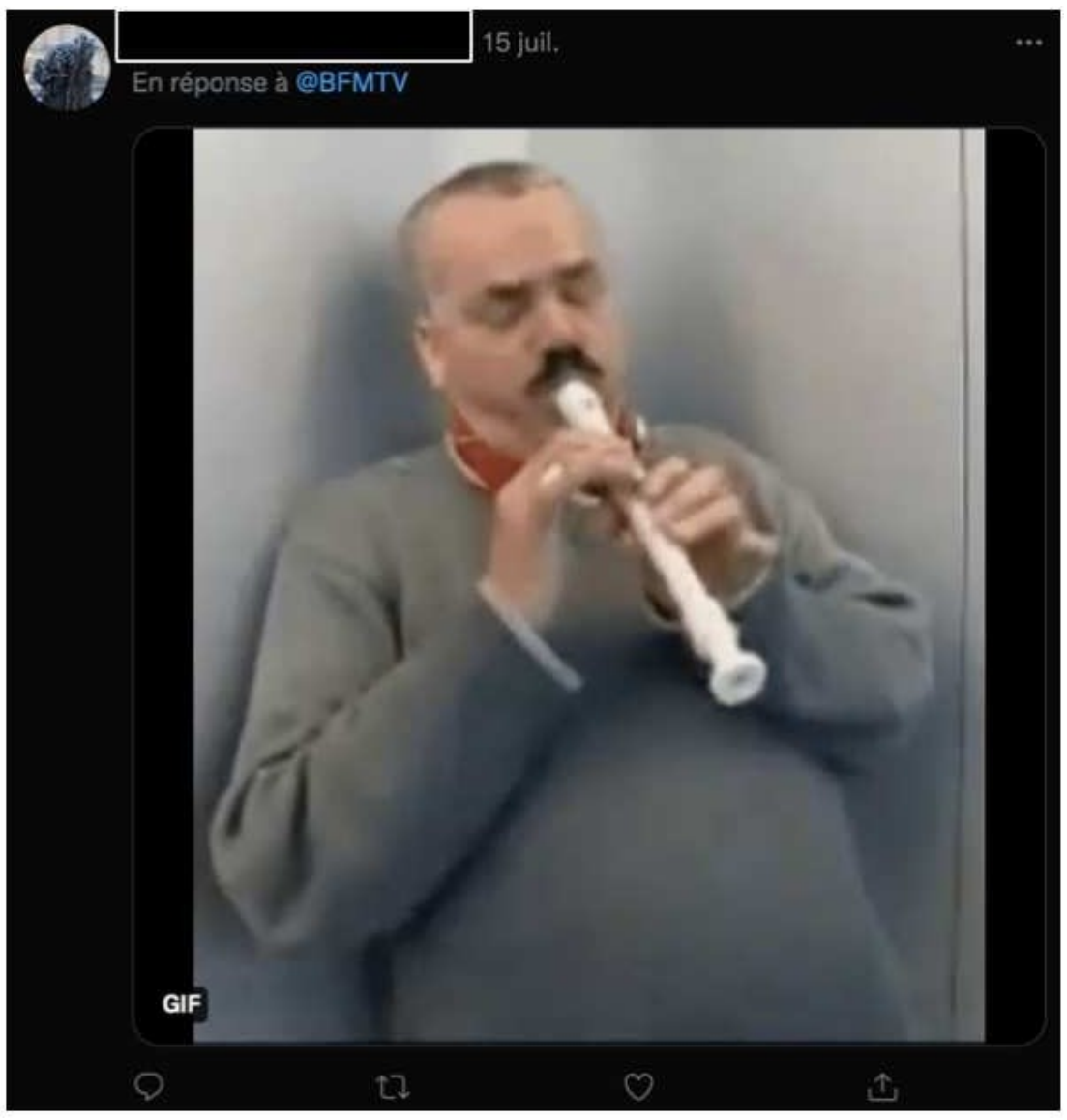

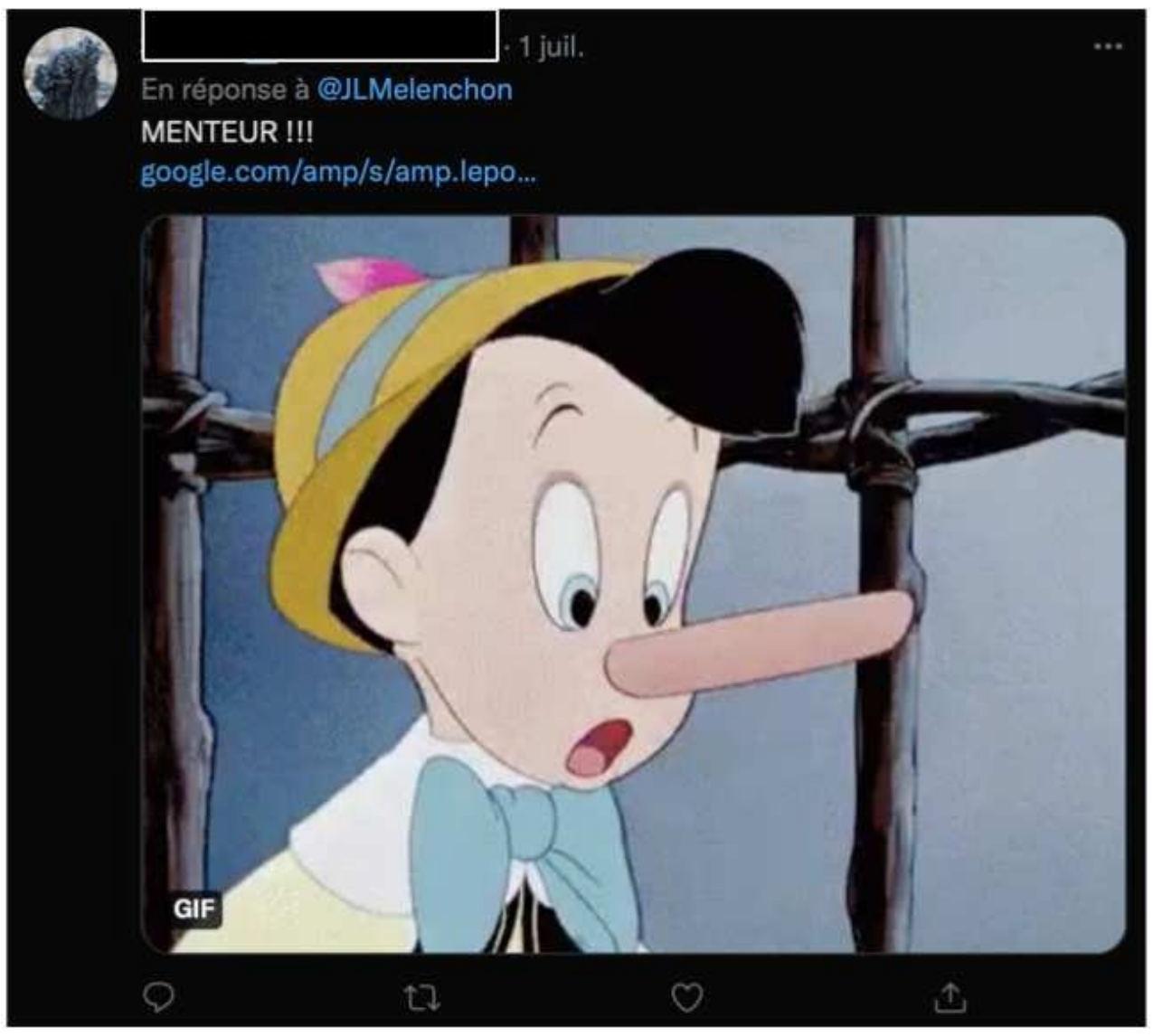

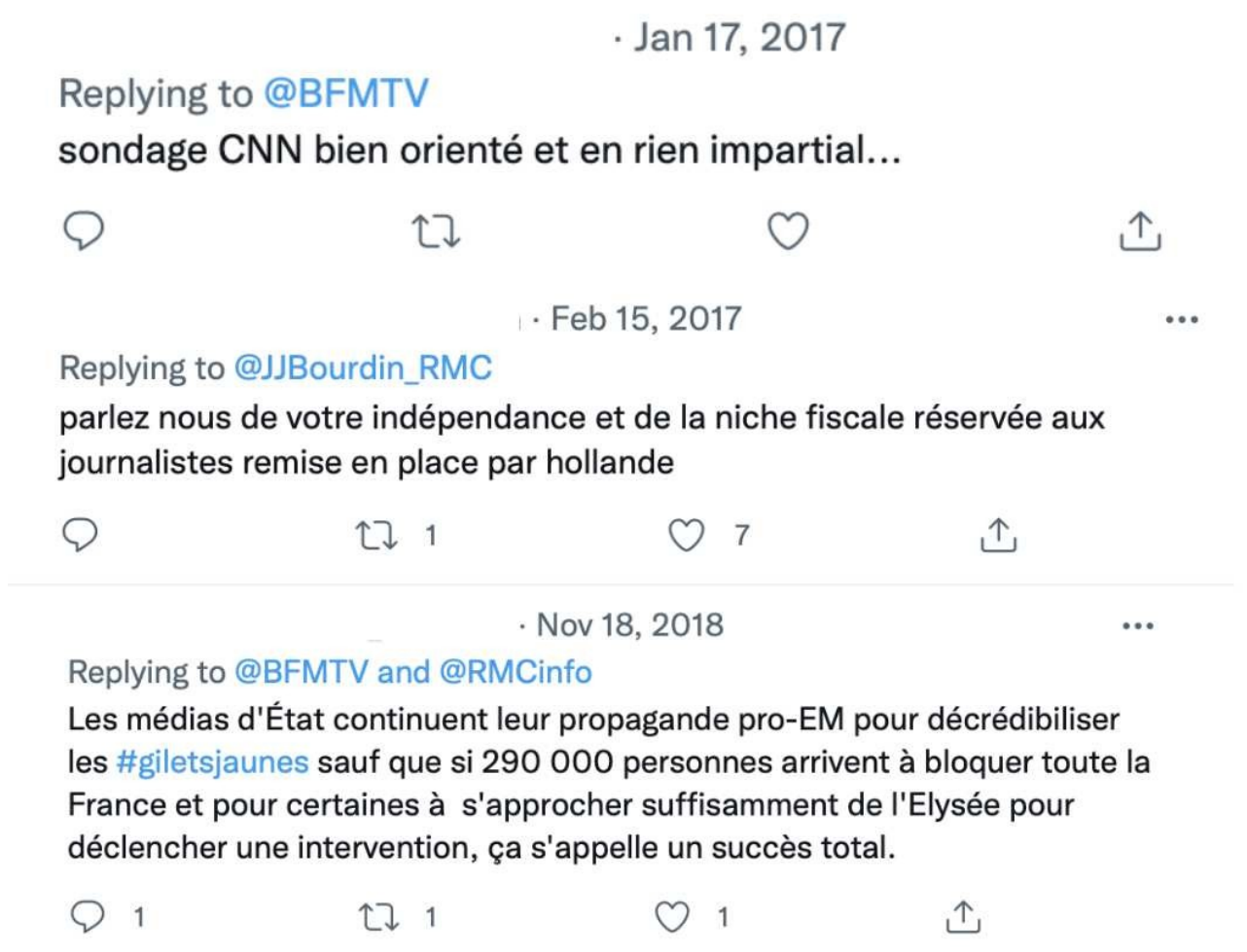

*Figure 4.9. Exemples de replies publiés par des utilisateurs indépendants*

Au cours d'un entretien, un enquêté a confirmé avoir fréquemment recours à l'option *reply* :

> *C'est rare que j'écrive un tweet. C'est très rare. En fait, je suis plutôt dans la réponse. [...]. Y a monsieur Castaner qui m'a fermé mon compte Twitter pendant 6 mois. J'ai fait des dizaines de tweets sur monsieur Castaner parce qu'en fait il était proche de la mafia et j'ai trouvé des photos sur Internet où on le voyait avec des anciens mafieux [...] J'ai publié les photos en m'adressant directement à monsieur Castaner. Je lui disais : « monsieur Castaner, comment expliquez-vous que… ? ». Et mon compte a été coupé pendant 6 mois. Donc l'État a quand même une mainmise sur les réseaux sociaux. Je n'ai jamais été insultant mais par contre c'était massif.* [165]

Contrairement aux comptes détenus par des relayeurs de *fake news* exerçant un rôle de représentant politique ou ayant adhéré à un parti, les comptes des relayeurs de *fake news* sans attache n'utilisent pas Twitter pour défendre un agenda politique. Leurs prises de parole sur le réseau social sont dominées par une logique de contestation. Si certains thèmes propres à l'extrême droite ou à l'extrême gauche sont fréquemment mis en avant, l'on peut surtout noter la présence de nombreuses publications exprimant un rejet des élites et une critique du système électoral (cf. Figure 4.10). Ils publient par exemple régulièrement des caricatures des membres du gouvernement ou des personnalités politiques.

[165] Homme, 56 ans, Bac +2, consultant, entretien réalisé le 19 novembre 2022.

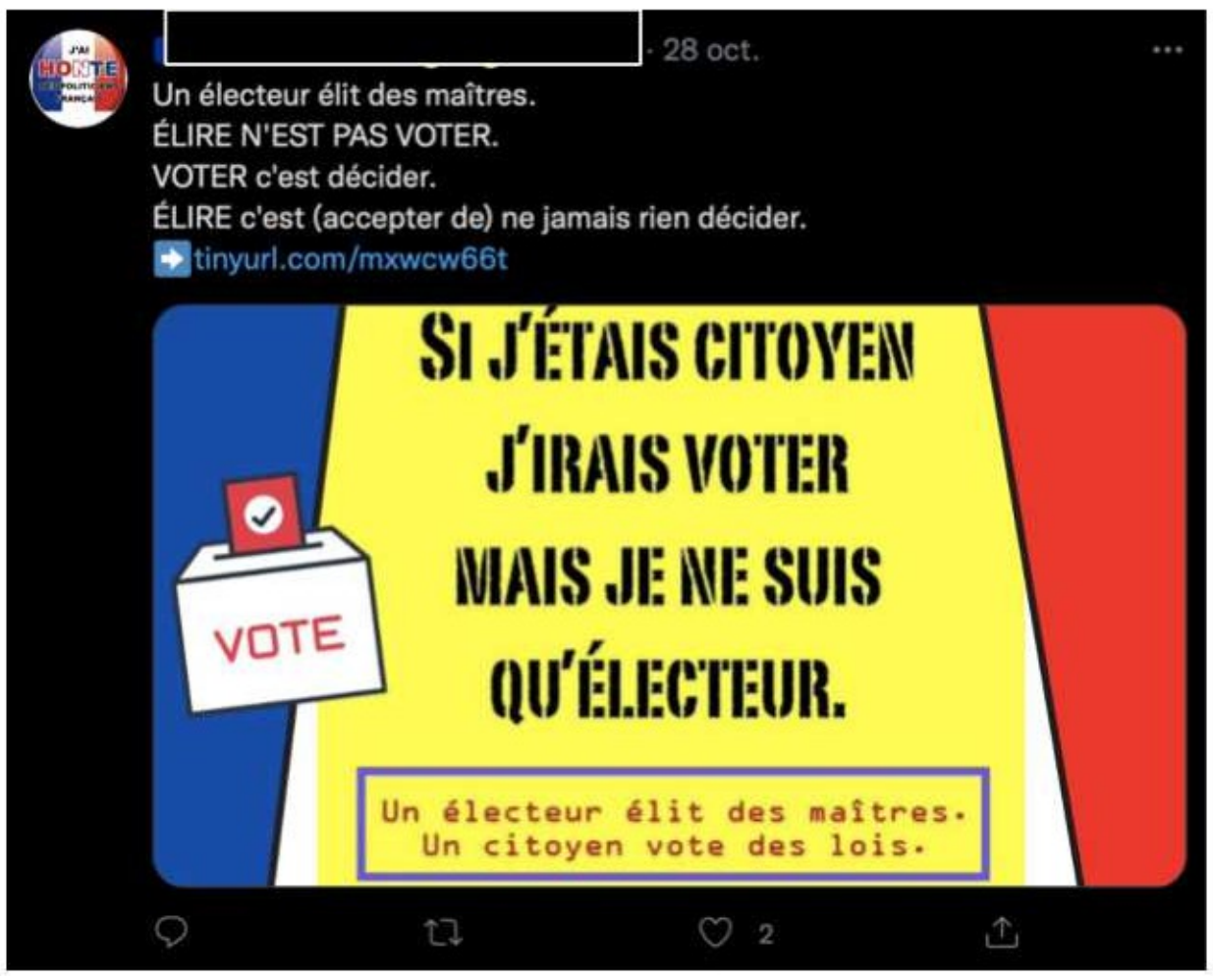

*Figure 4.10. Exemple de post publié par un utilisateur indépendant*

En résumé, plus les relayeurs de *fake news* sont intégrés à une organisation politique, plus ils bénéficient d'une visibilité importante sur Twitter et moins ils partagent de *fake news*. Par exemple, alors que les élus se trouvent régulièrement dans des situations de prise de parole en public hors ligne et sont exposés à une forte visibilité en ligne, ils partagent moins de *fake news*, ont une activité plus modérée et modalisent davantage leur énonciation sur Twitter. De leur côté, les membres d'un parti ou les personnes exerçant d'importantes responsabilités professionnelles sont régulièrement conduits à s'exprimer dans des cadres formels et ont souvent recours à des stratégies pour contrôler leur image et leur visibilité sur Twitter. À l'inverse, les utilisateurs sans attache politique ou inactifs sur le plan professionnel ont moins d'obligations à respecter et disposent de plus de temps pour interagir avec leurs pairs, que ce soit en ligne ou dans des conversations informelles ; ce qui favorise une plus grande activité sur Twitter et une énonciation plus relâchée (cf. Tableau 4.1.).

| | Élus n=12 | Adhérents n=23 | Militants n=50 | Indépendants n=36 | Peu politisés n=3 |
|---|---|---|---|---|---|
| **nombre moyen de *fake news*** | 5 | 7 | 10 | 15 | 1 |
| **recours à un pseudonyme** | 8 % | 48 % | 76 % | 92 % | 67 % |
| **nombre moyen d'abonnés** | 12 120 | 5 209 | 2 900 | 1 385 | 139 |
| **nombre moyen d'abonnements** | 4 521 | 4 266 | 2 259 | 1 612 | 566 |
| **nombre moyen de (re)tweets par jour** | 26 | 38 | 102 | 107 | 4 |

*Tableau 4.1. Caractéristiques des comptes Twitter selon leur degré d'affiliation à un parti politique*

## 4.4. La prudence énonciative : un outil d'interaction sociale

Cette section vise à revenir sur différentes pratiques observées au cours des analyses précédentes à travers lesquelles les individus cherchent à préserver leur réputation ou à maintenir leur intégration dans différents groupes sociaux. Ces pratiques, ainsi que les conditions dans lesquelles elles se produisent, sont décrites ci-dessous en reposant sur des distinctions conceptuelles de pragmatique linguistique, ainsi que sur les apports des travaux d'analyses conversationnelles.

Le premier type de pratique correspond à une forme d'évitement et d'auto-censure. Les utilisateurs choisissent de ne pas partager d'information, ni d'exprimer leur opinion que cela soit dans des situations en ligne ou hors ligne. Invisible ou silencieuse, cette pratique est impossible à capturer en reposant uniquement sur des données collectées sur les réseaux sociaux, et difficilement perceptible au cours d'observations *in situ*, dans la mesure où les utilisateurs ne produisent pas d'énoncé, ni de trace d'énonciation, à moins qu'ils n'aient recours à des stratégies rhétoriques comme la prétérition, ou ne produisent des gestes et des expressions témoignant d'une attitude de retrait. Dans le cadre de cette enquête, cette pratique a pu être relevée au cours de quelques entretiens lorsque des utilisateurs ont indiqué avoir déjà décidé de se taire dans une situation d'interactions hors ligne ou de ne rien publier sur les réseaux sociaux. On peut rappeler par exemple les témoignages de Viviane ou de Gauthier, déjà mentionnés plus haut :

> *Moi je faisais profil bas. De tout façon il y a un tel conformisme, on va dire de gauche, où toutes les évidences de gauche sont présentées avec la force de l'évidence incontestable qu'on va pas rentrer à batailler, à argumenter, c'est pas possible. Donc bon on dit : « oh oui » ou on dit rien.*[166]

> *À l'université, je ne le dis surtout pas ! [...] L'université c'est quand même majoritairement à gauche, je suis obligé de faire profil bas, même pas profil bas, de me faire tout petit. Je ne ramène jamais ma fraise. Moi à la fac, personne ne le sait, je suis un caméléon.*[167]

En écho aux travaux de Nina Eliasoph (1998) sur l'évitement des discussions politiques et de Elisabeth Noelle-Neumann sur les mécanismes de « spirale de silence » (1974), cette pratique est surtout mise en œuvre dans des espaces publics, lorsque les personnes sentent que leur opinion est minoritaire ou risque d'être mal perçue. Mais elle peut aussi se produire lors de discussions interpersonnelles informelles, où les individus cherchent à préserver leur face et celle de leurs partenaires (Goffman, 1973 ; 1974).

Le deuxième type de pratique vise à restreindre de façon stratégique la visibilité d'un énoncé. Les utilisateurs ont recours à des tactiques qui leur permettent de limiter leur audience à certains individus et d'en exclure d'autres. Cette restriction de visibilité repose sur un ensemble de stratégies destinées à contrôler l'accès à ses propres publications et à gérer les frontières de son audience. Plusieurs outils permettent d'opérer cette sélection, parmi lesquels on peut citer le blocage d'utilisateurs, le passage d'un compte en mode privé, ou encore l'utilisation de listes d'abonnés pour cibler certaines catégories de personnes tout en excluant d'autres. Les individus peuvent également choisir de se replier vers certains espaces de discussion au détriment d'autres, soit parce que ceux-ci sont plus confidentiels, soit parce que les règles de modération y sont moins strictes. Ces pratiques s'inscrivent dans une démarche de gestion calculée de la visibilité (Cardon, 2008), où les individus cherchent à protéger leur énonciation des regards indiscrets ou des personnes qu'ils jugent non légitimes à avoir accès à leurs propos pour éviter les sanctions symboliques ou sociales qu'ils pourraient entraîner.

---

[166] Femme, 70 ans, Bac +8, retraitée, ex-chercheuse, entretien réalisé le 15 mars 2022.
[167] Homme, 23 ans, Bac +4, étudiant, entretien réalisé le 5 décembre 2022.

Le troisième type de pratique se produit lorsqu'un locuteur choisit de supprimer un énoncé qu'il a publié antérieurement. Cette pratique relève d'une rétractation *a posteriori* : le locuteur revient sur son propre discours en l'effaçant du champ de la communication. Contrairement à une simple révision de son propos, le locuteur opère une tentative d'effacement total de toute trace énonciative dans l'espoir de faire disparaître à la fois l'énoncé et l'engagement qu'il contenait. Cette rétractation peut faire suite à une prise de conscience des répercussions qu'a eu ou que pourrait avoir un énoncé sur la réputation du locuteur. Par exemple, Jocelyne, Sylvia et Moïse ont indiqué avoir déjà supprimé un de leurs tweets après une vérification personnelle, une exposition à un *fact-check* ou des réactions négatives de la part des autres utilisateurs.

> *J'utilise peu Facebook mais encore un petit peu et Twitter beaucoup parce que l'information est rapide, claire… Enfin claire, en même temps il y a beaucoup de fake. Je me suis déjà faite avoir. J'essaie de voir et puis sinon ben je supprime. Mais bon ça m'arrive. Pas souvent hein. Mais faut être honnête ça m'est arrivé après vérification.*[168]

> *— Ça vous est déjà arrivé que quelqu'un qualifie une information que vous avez partagée de fake news ?*
> *— Sur Facebook ça m'est arrivé récemment. Ils renvoyaient vers un article qui montrait que ce qui avait été publié était faux. Alors euh je ne sais plus sur quoi c'était… Mais euh il semblerait qu'ils avaient raison donc finalement j'ai supprimé ma publication.*
> *— Ah oui d'accord. Peut-être un fact-check ?*
> *— Oui voilà c'est ça*
> *— Et ça vous a convaincu ?*
> *— Ça m'a convaincue. Les articles qui montrent qu'une info est fausse ne sont pas toujours convaincants pour moi mais là en l'occurrence il l'était. Ah attendez… Je me souviens maintenant de ce que c'était. C'était Elisabeth Borne, on lui avait prêté des propos qu'elle n'avait pas dit en fait. Je ne sais plus à quel sujet. Mais du coup j'ai pris ça pour argent*

[168] Femme, 67 ans, certificat d'étude, retraitée, entretien réalisé le 19 mai 2022.

*comptant alors que non c'était faux. Alors assez souvent j'arrive à faire la différence. Quand c'est un compte parodique je le vois. Mais là c'est vrai que je me suis fait avoir.*[169]

*— Je tweetais ce qui me passait par la tête, tout simplement. Des fois sans trop réfléchir. D'ailleurs j'ai pu regretter après.*
*— Ha oui, il y a des tweets que vous avez regretté d'avoir postés ?*
*— Ouais j'ai supprimé des tweets oui bien sûr. Certains de moi-même. D'autres que j'ai retirés parce qu'on m'a fait comprendre que ce n'était pas forcément compréhensible. Les réactions n'étaient pas celles que j'espérais ou négatives, donc j'ai supprimé.*
*— Vous auriez un exemple ?*
*— J'ai pas forcément d'exemple précis en tête mais je sais que j'avais fait un tweet sur la laïcité avec un dessin pas très clair pour les gens. Je pense qu'en fait ils pensaient que j'attaquais la laïcité alors que je la défendais.*[170]

Toutefois, il est important de noter que la suppression d'un énoncé dans l'espace numérique ne garantit pas une disparition totale. Les captures d'écran, les partages ou les discussions subséquentes peuvent continuer à faire circuler l'énoncé supprimé, créant ainsi un paradoxe d'effacement : bien que le locuteur tente d'annuler son discours, celui-ci peut continuer à exister dans des formes dérivées, échappant au contrôle initial.

Le quatrième type de pratique consiste à opérer un réajustement discursif ou une auto-correction. Un locuteur modifie son discours après avoir pris conscience que l'énoncé qu'il a partagé était erroné ou inapproprié. Il peut reconnaître publiquement une erreur et corriger son propos. Ce type de pratique permet de gérer les conséquences d'un énoncé erroné sans recourir à la suppression qui pourrait être perçue comme une tentative de dissimulation ou de fuite. En corrigeant publiquement son propos, l'individu témoigne de sa capacité à reconnaître ses erreurs et fait montre d'une ouverture d'esprit ainsi que d'une honnêteté intellectuelle. Cette pratique est cependant assez rare. Le témoignage de Christian par exemple montre que même lorsque la volonté de corriger est présente, des facteurs tels que la perte de mémoire concernant l'énoncé original ou le temps demandé par la correction peuvent freiner l'initiative de réajustement discursif.

[169] Femme, 58 ans, Bac +3, inactive, ex enseignante, entretien réalisé le 9 juin 2022.
[170] Homme, 45 ans, Bac +4, fonctionnaire, entretien réalisé le 18 mai 2022.

*Oui la semaine dernière on m'a pointé que ce que j'avais partagé était un fake. Puis, effectivement en regardant rapidement en dix minutes je me suis aperçu que c'était un fake. Mais là j'ai un trou, je ne sais plus sur quoi c'était. Alors c'est un cas concret intéressant parce que je me suis dit que j'allais fouiller plus et donc j'ai laissé en me disant que j'allais faire un correctif et finalement j'ai laissé tomber. Il m'avait tapé : « c'est un fake mon gars ».*[171]

Le cinquième type de pratique se manifeste par l'utilisation de modalisateurs tels que des adverbes de doute (« peut-être », « vraisemblablement »), permettant de tempérer un engagement fort vis-à-vis de l'information partagée. Cette précaution langagière permet au locuteur d'exprimer une prise de distance avec le point de vue contenu dans l'énoncé.

Le sixième type de pratique observé s'apparente à une forme de dédoublement ou de dissociation énonciative. L'utilisateur a par exemple recours à un pseudonyme ou à un deuxième compte pour dissocier son identité civile de l'énoncé partagé. Les distinctions conceptuelles apportées par Oswald Ducrot (1984) entre sujet parlant, locuteur et énonciateur offrent une grille d'analyse pertinente pour comprendre cette pratique. Lorsqu'un utilisateur utilise un pseudonyme pour publier un message sur un réseau social, il s'opère une dissociation entre le sujet parlant (la personne réelle qui écrit le tweet) et le locuteur (l'identité associée au pseudonyme). Dans ce cas, le sujet parlant est la personne réelle, physique, derrière l'écran, qui tape et publie le tweet. Cette personne est souvent invisible ou non identifiée directement si elle utilise un pseudonyme. Le locuteur est ici l'identité sous laquelle l'énoncé est publié, c'est-à-dire le compte Twitter pseudonymisé. C'est ce locuteur à qui est attribuée la responsabilité publique du tweet, même si cette personne est fictive ou anonyme. L'énonciateur, quant à lui, se réfère à la perspective ou au point de vue exprimé dans l'énoncé. Ce point de vue peut coïncider avec celui du locuteur lorsque l'utilisateur s'exprime de manière personnelle, ou bien être mis en scène de façon détachée, représentant une posture fictive ou une identité construite. Ces distinctions montrent que sur les réseaux sociaux, les rôles de sujet parlant, locuteur et énonciateur peuvent être dissociés, permettant ainsi aux utilisateurs de se distancier de leurs propres propos.

Enfin, le septième type de pratique, que l'on peut qualifier d'effacement énonciatif, se traduit par le fait que le locuteur gomme les traces de son implication dans l'énoncé. Il devient ainsi une sorte de ventriloque, laissant l'énoncé parler à sa place sans s'engager personnellement. Cela est souvent observé dans les retweets « secs » (i.e. partage de contenus sans ajout de commentaires) où le locuteur évite de s'impliquer directement. Par exemple, en retweetant un message polémique sans ajouter son avis personnel, l'utilisateur efface son rôle en ne s'associant pas explicitement à l'énoncé. L'usage courant de la formule « RT is not endorsement » illustre aussi ce phénomène, où l'acte de relayer est désolidarisé d'un soutien explicite, permettant ainsi au locuteur de minimiser sa responsabilité énonciative. De nouveau, les distinctions conceptuelles apportées par Oswald Ducrot (1984) entre sujet parlant, locuteur et énonciateur offrent une grille d'analyse pertinente pour comprendre cette pratique. Ici le sujet parlant (l'utilisateur qui émet le retweet) s'efface par rapport au locuteur premier (l'utilisateur à l'origine du tweet) et au point de vue de l'énonciateur exprimé dans l'énoncé.

[171] Homme, 55 ans, Bac +8, à son compte, ex-chercheur, entretien réalisé le 6 mars 2022.

À l'issue de ces analyses, nous proposons de désigner l'ensemble des pratiques décrites ci-dessus par la notion de « prudence énonciative ». Cette notion a déjà été mentionnée par Patrick Charaudeau (2010) dans son article « Une éthique du discours médiatique est-elle possible ? » afin de qualifier les discours savants par contraste avec les discours journalistiques. Dans cet article, Patrick Charaudeau indique que la prudence énonciative, se manifeste par des verbes et des adverbes de modalité (« il est probable que… », « on peut dire que… », « vraisemblablement »). Cependant, il ne fournit pas une définition détaillée de la prudence énonciative ni ne la situe dans un cadre théorique plus large. Cette thèse contribue à enrichir cette notion en l'illustrant par une série de pratiques ayant été observées au cours d'une enquête empirique et en la rattachant à différents concepts de pragmatique linguistique ou de sociologie interactionniste.

Comme l'ont montré les descriptions ci-dessus, la prudence énonciative ne se limite pas seulement à l'utilisation de modalisateurs, mais englobe également diverses stratégies telles que l'évitement, le dédoublement énonciatif ou les rétractations. À travers ces pratiques les utilisateurs intègrent dans leur énonciation une anticipation des potentielles réactions de leur audience réelle ou imaginée, et plus précisément des sanctions qu'ils sont susceptibles d'encourir. Cette observation suggère que la vigilance des utilisateurs des réseaux sociaux est moins dirigée vers la factualité intrinsèque d'un énoncé que vers sa nature contestable.

Autrement dit, l'anticipation des réactions des autres joue un rôle plus important que la seule qualité épistémique d'un contenu dans la mise en œuvre de cette pratique, suggérant que celle-ci constitue davantage un outil d'interaction sociale qu'une compétence cognitive.

## Conclusion du quatrième chapitre

Les analyses conduites tout au long de ce chapitre ont montré comment les pratiques informationnelles et numériques des utilisateurs qui relaient des *fake news* sur Twitter, ainsi que leurs stratégies de présentation de soi et leurs régimes d'énonciation, sont susceptibles de varier selon leur position dans l'espace social, les multiples rôles qu'ils doivent endosser dans leur vie quotidienne et les différentes situations d'interactions dans lesquelles ils sont amenés à recevoir ou à partager des informations. Autrement dit, alors que les analyses présentées dans le chapitre précédent ont fait ressortir des profils de relayeurs de *fake news* relativement homogène, ce chapitre met non seulement à jour différents profils de relayeurs de *fake news*, mais montre également comment ces derniers peuvent revêtir des facettes identitaires multiples et variées selon les espaces de visibilité, les cercles relationnels et les contextes de la vie quotidienne au sein desquels ils naviguent.

Prêter attention à ces variations tant inter qu'intra-individuelle a permis de mieux comprendre les mécanismes sociaux et situationnels qui favorisent ou inhibent la mise en circulation de *fake news* et la propension des utilisateurs à faire preuve d'une plus ou moins grande prudence énonciative.

Deux résultats principaux ressortent ainsi de ce chapitre. D'un côté, il permet de montrer que les réseaux sociaux n'ont pas fait disparaitre les régulations implicites qui émanent des structures sociales. Les pratiques des utilisateurs sur les réseaux sociaux ne sont pas totalement dissociées des attentes et des contraintes sociales qui encadrent leurs comportements hors ligne. Ainsi, l'intégration dans l'espace politique et social module l'exposition et le partage de *fake news* en faisant peser différentes contraintes sur les prises de paroles des individus. D'un autre côté, ce chapitre montre également comment les pratiques des relayeurs de *fake news* ne se réduisent pas simplement à leur position dans l'espace social, mais sont aussi construites par des dynamiques interactionnelles. Autrement dit, il existe des situations où les personnes font particulièrement attention à leur énonciation. D'autres au contraire, où la parole se libère. Ainsi, alors que les positions sociales des utilisateurs fournissent un cadre général qui structure leurs pratiques, ce sont les situations d'interactions concrètes qui les orientent vers des régimes d'énonciation spécifiques. Plus précisément, ce sont les potentielles réactions critiques ou sanctions des audiences imaginées qui façonnent les pratiques des utilisateurs et les poussent à faire preuve d'une plus ou moins grande prudence énonciative. Afin d'obtenir une compréhension plus fine des mécanismes qui contribuent à limiter le partage de *fake news*, l'enjeu du prochain chapitre sera justement de s'intéresser aux réactions critiques que peuvent exprimer des utilisateurs au cours d'un échange ou d'une discussion, notamment en ligne.

# Troisième partie. Entre critiques et silence

D'abord érigées en problème public par les discours médiatiques et institutionnels, puis en objet de recherche par les travaux académiques, les *fake news* sont également devenues un sujet de préoccupation majeur pour le grand public ces dernières années. 56 % des personnes interrogées par l'Institut Reuters ont déclaré se sentir très concernées par la problématique des *fake news* (Newman et al., 2021) et 50 % des Américains sondés par le Pew Research Center ont indiqué placer cet enjeu au-dessus d'autres problèmes sociaux comme le racisme ou le sexisme (Mitchell et al., 2019). Ces inquiétudes ont conduit de nombreuses personnes à modifier leur manière de s'informer, notamment à adopter des tactiques pour naviguer plus prudemment dans l'écosystème informationnel numérique. Par exemple, 78 % des répondants ont précisé se tourner régulièrement vers des rubriques de *fact-checking* et 32 % ont rapporté avoir déjà signalé un contenu sur les réseaux sociaux (*Ibid*). De plus, l'augmentation des recherches Google effectuées sur les *fake news* et des publications mentionnant le terme sur les réseaux sociaux (Li et Su, 2020) suggère que les débats sur les *fake news* ont largement infusé les espaces de discussion en ligne du grand public.

Limités à des données déclaratives, ces différents constats ne signifient pas que les individus sont nécessairement plus aptes que par le passé à discerner le vrai du faux. Ils invitent cependant à s'interroger sur la manière dont ceux-ci formulent le problème des *fake news*, au cours de leurs échanges ordinaires, et tentent d'y apporter des solutions. Après tout, alors que de nombreuses personnes se disent préoccupées par la circulation de *fake news* au quotidien, et utilisent régulièrement ce mot lors de leur navigation en ligne, pourquoi ne pas chercher à examiner leurs pratiques consistant à les dénoncer, plutôt que d'étudier uniquement – comme l'ont fait la majorité des études jusqu'à présent – leur propension à y adhérer ou à les partager ?

L'objectif de cette troisième partie est justement de partir à la rencontre des utilisateurs des réseaux sociaux qui interviennent dans le flux d'une conversation pour faire part d'un trouble ou exprimer une critique. Elle restitue les résultats d'une enquête menée sur Facebook qui, comme l'étude présentée dans la partie précédente, articule également des analyses quantitatives de traces numériques à des observations en ligne et des entretiens.

Le chapitre 5 introduit le concept de « point d'arrêt » pour qualifier des formes de lecture oppositionnelle qui se traduisent par l'expression publique de désaccords, de signalements, de critiques, de corrections ou de dénonciations au cours d'échanges en ligne. Après avoir présenté les perspectives théoriques mobilisées pour conceptualiser la notion de « point d'arrêt », le chapitre détaille l'approche méthodologique déployée pour identifier ces formes d'expression au sein de commentaires Facebook, ainsi que pour tenir compte de la multiplicité et de l'hétérogénéité des espaces de communication qui existent sur le réseau social. Les différentes méthodes mobilisées ont ensuite permis d'examiner comment la propension des utilisateurs à exprimer des points d'arrêt, ainsi que leurs régimes d'énonciation, varient selon les divers espaces de communication identifiés sur Facebook. Les résultats montrent que les pratiques consistant à exprimer des points d'arrêt sont plus fréquentes au sein d'espaces de conversation peu politisés et/ou bénéficiant d'une audience importante et/ou rassemblant des personnes avec des points de vue hétérogènes. Ils soulignent également comment, au-delà de ces variables situationnelles, des mécanismes sociaux, tels que le fait d'être membre d'un groupe de modération collective, sont aussi susceptibles d'encourager l'expression de points d'arrêt. En somme, les réseaux sociaux ne constituent donc pas un univers de pratiques homogènes totalement dépourvus de règles. Ils sont plutôt composés d'une pluralité d'espaces, régis par différents contrats de communication et normes d'interactions, favorisant ou entravant des mécanismes d'autorégulation conversationnelle.

Le chapitre 6 approfondit les résultats du chapitre précédent en analysant les profils des utilisateurs des réseaux sociaux qui expriment des points d'arrêt sur Facebook et les réactions que cela déclenche dans la suite des conversations. Tout comme le partage de *fake news*, l'expression de points d'arrêt est l'apanage d'une minorité d'utilisateurs, généralement très actifs sur les réseaux sociaux, plutôt jeunes, politisés et éduqués. Par ailleurs, les points d'arrêt exprimés dans les commentaires Facebook restent la plupart du temps sans réponse, quand ils ne déclenchent pas des réactions injurieuses et négatives. Ainsi, bien que les utilisateurs des réseaux sociaux soient en mesure de mobiliser des formes de distance critique au cours d'échanges en ligne, celles-ci permettent rarement l'émergence de véritables débats délibératifs, pas plus que l'expression d'un pluralisme agonistique, mais donnent plutôt lieu à des « dialogues de sourds » entre une minorité d'utilisateurs particulièrement actifs en ligne. Cela étant, la rareté des réponses suscitées par les points d'arrêt ne signifie pas que les utilisateurs qui y sont exposés y sont totalement indifférents. L'absence de trace numérique est peut-être le signe d'une attention silencieuse ou d'un évitement délibéré de leur part – une hypothèse qui invite les futures recherches à sortir des réseaux sociaux afin de partir à la rencontre de ces utilisateurs invisibles et silencieux.

# Chapitre 5. L’incursion de « points d’arrêt » dans les échanges en ligne

Le chapitre précédent a montré comment les utilisateurs des réseaux sociaux qui partagent des *fake news* sur Twitter sont susceptibles de faire preuve d’une plus ou moins grande prudence énonciative selon leur position dans l’espace social, les rôles qu’ils doivent endosser et les situations dans lesquelles ils sont amenés à recevoir ou partager des informations. En effet, que cela soit en ligne ou hors ligne, différentes contraintes sociales et attentes normatives pèsent sur les comportements des individus. En cas de non-respect de celles-ci, des sanctions et rappels à l'ordre sont mis en place par les autres partenaires de l’action, notamment à travers des interventions critiques. Rappelons-nous par exemple du témoignage de Jocelyne lorsque celle-ci a mentionné que l’un de ses fils avait institué une amende dans le groupe WhatsApp familial pour éviter l’émergence de tensions autour du sujet du vaccin contre le Covid-19.

Ce cinquième chapitre vise précisément à décrire comment ces interventions critiques se manifestent au cours d’échanges en ligne. Au-delà de rendre compte des divers régimes de critiques employés par les utilisateurs des réseaux sociaux, l’objectif est également d’expliquer quelles sont les conditions socio-techniques qui favorisent ou entravent ces formes d’expression sur les réseaux sociaux.

Pour répondre à ces questions, ce chapitre se structure en trois sections. La première effectue un travail de conceptualisation afin d’ancrer les pratiques des utilisateurs consistant à exprimer des critiques en ligne dans un cadre théorique compréhensif. En partant des acquis des travaux de sociologie de la réception, d’analyses conversationnelles et de sociologie pragmatique sur la lecture oppositionnelle, l’expression de désaccords et la dénonciation publique, elle propose de forger un concept adapté aux spécificités des pratiques et des espaces numériques : celui de point d’arrêt. L’analyse d’un corpus de 40 000 commentaires Facebook a ensuite permis d'illustrer comment cette pratique se manifeste concrètement au sein d'échanges en ligne et de dresser une typologie des points d’arrêt en déterminant les éléments mis en cause et les appuis mobilisés par les utilisateurs dans leurs critiques. Ce double travail de conceptualisation et de description a ensuite permis d’automatiser la détection de points d’arrêt au sein d’un corpus de plus de 400 000 commentaires Facebook grâce à des méthodes d’apprentissage supervisé. Une réplication a également pu être amorcée sur un corpus de près de 3,5 millions de commentaires Youtube.

La deuxième section vise à rendre compte de la diversité des espaces de communication présents sur un même réseau social – en l'occurrence sur Facebook. L’enjeu est surtout de décrire le fonctionnement et les caractéristiques principales des pages et des groupes publiquement accessibles sur la plateforme afin d’identifier les contrats de communication qui les sous-tendent et de questionner la façon dont ils peuvent être perçus par les utilisateurs du réseau social.

Enfin, la troisième section examine comment l’expression de points d’arrêt varie en fonction des différents espaces de communication identifiés sur Facebook et cherche à expliquer les mécanismes sociaux et situationnels favorisant ou entravant ces formes d’expression sur les réseaux sociaux.

## 5.1. Définir et détecter des points d'arrêt dans des conversations en ligne

Alors que la majorité des études contemporaines se sont attachées à expliquer pourquoi les individus adhèrent à des *fake news* ou les partagent, quelques rares enquêtes empiriques se sont intéressées à leurs capacités à les réfuter (Micallef et al., 2020), que cela soit en corrigeant publiquement des informations perçues comme erronées (*user correction* ; Bode et al., 2024) ou en contestant des discours jugés néfastes (*counter speech* ; Buerger, 2021). Dans la littérature anglo-saxonne, ces actions menées par les utilisateurs du web pour réguler l'écosystème informationnel numérique, en dehors des procédures de modération institutionnelle, ont été conceptualisées comme des formes d'interventions civiques en ligne (Porten-Cheé et al., 2020) et de contrôle social informel (Atchison, 2000 ; Watson et al., 2019).

Malgré leurs apports conceptuels et empiriques, la plupart de ces recherches reposent cependant sur des approches expérimentales ou des questionnaires ne permettant pas de décrire et de comprendre quand et comment ces pratiques se manifestent de façon spontanée dans des discussions en ligne. Ces études ont également tendance à adopter un prisme normatif, considérant d'emblée les critiques des utilisateurs comme de « bonnes actions » émises par des « citoyens rationnels ». Or, dans une conversation de tous les jours, il est tout à fait possible qu'une personne ait tort de dire à une autre personne qu'elle se trompe, que cela soit parce que sa correction est erronée ou parce que son interlocuteur la juge inappropriée.

L'enjeu de cette section est d'adopter une approche descriptive et compréhensive afin d'appréhender les critiques exprimées par les utilisateurs sans *a priori*, c'est-à-dire en considérant les utilisateurs des réseaux comme capables de formuler des critiques, mais sans préjuger de la pertinence ou de la validité de celles-ci. Autrement dit, l'objectif est de décrire et de comprendre comment les utilisateurs expriment des critiques en partant de leurs pratiques et non de ce qui est considéré comme « rationnel » d'un point de vue normatif.

### 5.1.1. Qu'est-ce qu'un « point d'arrêt » ?

Avant d'offrir un cadre conceptuel permettant d'appréhender les pratiques des utilisateurs des réseaux sociaux consistant à intervenir dans le flux d'une conversation pour exprimer une critique, revenons sur la manière dont la réception de l'information peut être modélisée à l'ère du numérique (voir Chapitre 2, section 2.1.).

À l'origine de la production d'un contenu se trouve un média qui, à partir des cadres culturels, sociaux, économiques et politiques dominants dans la société, encode un sens préféré dans un message afin d'orienter l'interprétation des récepteurs en favorisant certaines lectures par rapport à d'autres (Hall, 1973). Le média établit ainsi ce que Véron (1985) appelle un « contrat de lecture » : une sorte de pacte implicite entre l'émetteur et le récepteur, où ce dernier est invité à s'identifier à la figure du destinataire construite par le média. Par exemple, à travers leur titre, leur mise en page et leurs choix de sujet, les magazines comme *Marie-Claire* ou *Elle* inscrivent leurs contenus dans un cadre interprétatif qui s'adresse principalement à une audience féminine. Ce cadre structure l'expérience de lecture en fonction de cette identification anticipée, invitant les lecteurs à s'aligner sur les valeurs, les préoccupations et les goûts attribués à cette audience ciblée.

Avant l'essor des réseaux sociaux, les contenus médiatiques étaient traditionnellement reçus par les publics soit directement via les canaux de transmission d'informations du média (e.g. chaîne de télévision, journal papier, etc.), soit indirectement, lors de conversations interpersonnelles se déroulant en face à face. Dans ces conversations, le contenu pouvait être réencodé par le locuteur, apportant ainsi une nouvelle orientation à la lecture initialement proposée par le média.

Avec le numérique, les dynamiques de réception de l'information se sont complexifiées. Les contenus peuvent toujours être reçus directement via les canaux traditionnels du média, désormais également disponibles en ligne (e.g. site web, compte Twitter, page Facebook, etc.) ou au cours de conversations interpersonnelles privées (désormais aussi bien en face à face qu'en ligne). Toutefois, le principal changement provoqué par l'essor du web et des réseaux sociaux est de permettre à une pluralité de locuteurs de relayer publiquement des contenus médiatiques. Ces acteurs, parmi lesquels se trouvent aussi bien des médias, que des partis politiques, des entreprises, des personnalités publiques ou de simples quidams introduisent à leur tour leur propre perspective et fixent ainsi un « contrat de conversation » (Granier, 2011). Ce contrat de conversation ne remplace pas le contrat de lecture initial du média mais le double en ajoutant une nouvelle couche d'interprétation accessible à une audience potentiellement beaucoup plus large. Ce double encodage – par le média et par les utilisateurs – ouvre alors un espace où les sens peuvent être négociés, contestés et redéfinis de manière plus participative et moins prévisible que dans des contextes de réception d'information se déroulant hors ligne. Ainsi, à l'ère du numérique, la réception de l'information ne peut plus être envisagée uniquement à travers le prisme du contrat de lecture établi par le média. Il faut également tenir compte du contrat de conversation qui se forme entre les utilisateurs en ligne et qui peut soit renforcer, soit contester le sens privilégié initialement encodé. Afin de considérer ensemble ces deux types de contrat, il est utile de mobiliser la notion de « contrat de communication » (Charaudeau, 2017). Celle-ci désigne « un cadre minimal, nécessaire à l'intercompréhension, une base commune de reconnaissance, de cadrage du sens, de stabilisation d'une partie de la production/reconnaissance du sens, à partir de laquelle peut se jouer une multiplicité de variations et de créations de sens. » (Charaudeau, 2004, p. 120).

Dans ce cadre, imaginons que des récepteurs réagissent de façon critique à un contenu médiatique. Qu'est-ce que cela signifie exactement ? À quoi sont-ils opposés ? Au contrat de lecture initialement proposé par le contenu médiatique ou au contrat de conversation qui se développe dans les échanges en ligne ? Comment ces réactions critiques se manifestent-elles et comment les appréhender ?

Sur le web ou les réseaux sociaux, des réactions critiques de la part d'utilisateurs peuvent tout d'abord relever d'une forme de lecture oppositionnelle telle qu'étudiée par les travaux de sociologie de la réception (Hall, 1973). Dans ce cas, les utilisateurs rejettent le sens encodé par l'émetteur et lui opposent une interprétation alternative. Néanmoins, ce processus reste privé et interne, donc difficilement accessible aux chercheurs. Si les utilisateurs extériorisent leurs réactions critiques, celles-ci peuvent alors s'apparenter à l'expression de désaccords ou de dénonciations publiques, deux pratiques ayant été déjà bien examinées par les travaux d'analyses conversationnelles et de sociologie pragmatique. Les premiers ont montré comment l'expression d'une opinion contraire était généralement considérée comme un « acte menaçant pour la face », susceptible de mettre en danger la relation interpersonnelle ou l'issue de la conversation (Brown et Levinson 1978 ; 1987), et donc constituait généralement une réaction « non-préférée » au cours d'interactions verbales se déroulant en face à face (Sacks, 1987). Les seconds, quant à eux, ont mis en lumière les règles de grammaire que doivent respecter les prises de paroles des personnes pour être jugées recevables dans l'espace public (Boltanski et al., 1984 ; Cardon et al., 1995)

Malgré leurs apports théoriques importants, ces perspectives présentent des limites pour appréhender l'expression de critique en ligne. En effet, sur les réseaux sociaux, les interactions ne se déroulent pas en face à face et ce n'est pas parce que les espaces sont publiquement accessibles et visibles qu'ils constituent pour autant des sphères publiques (Papacharissi, 2002). Ces transformations de la morphologie de l'espace public soulèvent des questions quant à la nécessité pour les utilisateurs des réseaux sociaux de respecter les règles de la « grammaire publique » (Lemieux, 2000) ou de maintien de la « face » (Goffman, 1955). En effet, il est possible de faire l'hypothèse que les individus encourent moins de risques à exprimer des désaccords en ligne plutôt qu'au cours d'interactions verbales en face à face, et n'aient pas besoin d'avoir recours à des adoucisseurs pour désamorcer leur opposition, dans la mesure où « l'impossibilité de construire une relation de coprésence et le caractère différé des échanges font que sauver sa face et celle des autres devient une contrainte moins lourde dans l'interaction » (Beaudoin et Velkovska, 1999, p. 130). Par ailleurs, il est également possible qu'en ligne, il ne soit pas nécessaire de désingulariser son énonciation ou de monter en généralité pour exprimer des critiques jugées recevables dans l'espace public. Cependant, on peut également émettre l'hypothèse que la publication d'un commentaire critique est plus risquée en ligne qu'hors ligne car celui-ci devient potentiellement visible par tous pour une durée indéterminée.

Afin d'adopter un cadre conceptuel ajusté aux spécificités des situations de réception d'information en ligne, et à leur nouveau régime de visibilité, il peut être pertinent de mobiliser des travaux appliquant une approche de sociologie pragmatique à l'étude des usages numériques. Benjamin Loveluck (2016), par exemple, a étudié le vigilantisme numérique, qu'il définit comme des « pratiques privées et extrajudiciaires de maintien de l'ordre et d'exercice de la justice » (*Ibid*, p. 129) par lesquelles les individus cherchent à alerter les autorités ou l'opinion publique, ou à se faire justice à eux-mêmes en engageant des formes actives de surveillance, de répression ou de dissuasion. Benjamin Loveluck identifie quatre formes principales d'auto-justice en ligne : le signalement, l'enquête, la traque et la dénonciation organisée. Sans rentrer dans les détails de ces quatre pratiques, l'expression de critiques ou de désaccords face à un contenu médiatique sur les réseaux sociaux peut-être perçue comme une forme de signalement consistant à « dénoncer des conduites dans l'espace public considérées comme inciviles, voire "honteuses" en ayant recours au *shaming* » (*Ibid*, p. 134). Néanmoins, la pratique de signalement, telle que décrite par Benjamin Loveluck, suggère une directionnalité et une intentionnalité forte, avec un objectif de sanction clair en vue. Or, il est possible que l'expression d'une critique ou d'un désaccord au cours d'une conversation en ligne se manifeste par des dynamiques plus diffuses et variées, où l'utilisateur peut ne pas avoir une intention claire de déclencher un changement de comportement ou de mener une action collective, mais simplement de marquer une opposition, une désapprobation, un doute ou un questionnement. Il est également possible qu'une intervention critique sur les réseaux sociaux découle moins de l'exécution d'un plan prémédité que d'une ouverture à la contingence de la situation (Quéré, 2007). Ce caractère spontané et instantané de l'action fait penser au régime d'excitation exploratoire conceptualisé par Nicolas Auray (2017) qui se caractérise par une vigilance flottante et une attention divisée, favorisant à la fois la dispersion et l'attention et permettant aux individus d'être constamment en alerte, prêts à signaler un problème lorsqu'ils sont insatisfaits, ou à mener une enquête plus approfondie lorsqu'ils sont confrontés à une énigme.

Au terme de ce travail de conceptualisation, on peut appréhender l'expression de critique en ligne face à un contenu médiatique comme une lecture oppositionnelle qui se manifeste par l'expression publique d'un désaccord, d'un signalement, d'une correction ou d'une dénonciation. Cette réaction correspond à un instant précis dans le discours d'un individu sans pour autant définir l'ensemble de son discours ou de sa pensée. En effet, il est possible d'observer des infléchissements, des oscillations, voire des rétractations après ce moment de critique initiale. Pour capturer ce moment spécifique, tout en conservant une flexibilité qui permet d'inclure les nuances et les évolutions possibles, nous proposons de forger le concept de « point d'arrêt ». La formule de point d’arrêt invite à ne pas considérer cette action comme une fin en soi, c’est-à-dire comme un point final qui viendrait clore une discussion, mais plutôt comme une pause, comparable à un pouce pendant un jeu. Ce moment de suspension permet à l'utilisateur de marquer une opposition ou un désaccord, tout en laissant ouverte la possibilité d'une reprise de l’échange ou d'une évolution de son discours.

### 5.1.2. Annotation manuelle d’un corpus de commentaires Facebook

Une fois les premiers contours du concept de point d’arrêt esquissés, l'étape suivante a consisté à examiner sous quelles formes d’expression cette pratique se manifeste concrètement dans des échanges en ligne. L’objectif ici est d’affiner la définition du concept de point d’arrêt à partir de données empiriques afin qu’elle reflète avec justesse les pratiques observées sur les réseaux sociaux.

Pour ce faire, des annotations ont été réalisées sur un échantillon de 43 228 commentaires, représentant environ 10 % du corpus total. Cet échantillon a été constitué à partir d’une sélection aléatoire de 1 148 posts (pour plus d’information sur l’ensemble des données récoltées sur Facebook et les échantillons étudiés, voir chapitre 2, section 2.3). Le choix de tirer aléatoirement une liste de posts et d’analyser l’ensemble des commentaires associés, plutôt que des commentaires isolés les uns des autres, a permis de suivre des fils de discussion complets directement sur Facebook et d’interpréter le sens des commentaires dans leur contexte d'interaction.

***Pluralité des formes de points d’arrêt***

Au cours de ce travail d’annotation, diverses pratiques et formes d’expression correspondant à la définition du concept de point d’arrêt ont été identifiées au milieu d’un ensemble très disparate de commentaires. Ces points d’arrêt se déclinent en plusieurs types, chacun étant dirigé vers différents éléments que l’on pourrait qualifier de « perturbateurs » dans la mesure où ce sont précisément ces éléments qui semblent troubler les utilisateurs de Facebook ou leur donner l'impression de traverser une zone de turbulences.

*Critique de la crédibilité du contenu*

Un premier ensemble de points d'arrêt vise à contester la crédibilité d'un contenu, la plupart du temps en le qualifiant simplement de « fake » ou d' « intox » sans développer davantage la critique exprimée. Dans certains cas, les utilisateurs apportent des arguments précis ou détaillés. Ils pointent notamment des incohérences ou une contradiction entre la description de la réalité véhiculée par le contenu et leur propre représentation du réel, façonnée par leurs perceptions, leurs expériences et les autres informations qu'ils ont consultées. Par exemple, si un contenu évoque un événement survenu il y a plusieurs années, certains utilisateurs interviennent en indiquant la date où il s'est produit afin de montrer qu'il n'est plus d'actualité (e.g. « Article du 13 avril 2016 ! Les médias en ont parlé à l'époque… » ; « pourquoi relayer de vieilles informations ? N'y a t'il pas suffisamment de thèmes importants à vos yeux de nos jours ??? »). Dans la même veine, si un article prête des propos à une personnalité politique, sans rendre compte de l'ensemble du discours, ni du contexte dans lequel il a été prononcé, quelques utilisateurs apportent alors des nuances ou des explications (« Cessez de déformer les propos du président ! C'est de la désinformation ! » ; « Il faut écouter tout son discours et pas seulement sortir la phrase polémique, surtout pour ceux qui se sentent visés! »). Pour étayer leurs critiques, ces utilisateurs vont parfois repêcher d'anciens posts, s'appuyer sur des captures d'écran, citer des articles de *fact-checking* ou d'autres sources d'informations :

> *Toutes les études à ce sujets faites par des vrai scientifiques montrent que ce n'est pas lié. Ce genre d'article est dangereux car rapporte de fausses informations qui peuvent entrainer la mort, comme on a pu le constater avec la récente épidémie de rougeole qui s'est développée à cause de personnes refusant de vacciner : http:// www.europe1.fr/ sante/ epidemie-de-roug eole-en-nouvell e-aquitaine-un- mort-dans-la-vi enne-3573242*

*Critique du manque de pertinence et d'intérêt du contenu*

Une deuxième série de points d'arrêt cible toujours le contenu d'un post mais dénonce cette fois-ci son manque de pertinence ou d'intérêt pour le débat public, plutôt que son caractère erroné ou invraisemblable. Ici, les utilisateurs expriment leur désapprobation en qualifiant les contenus de « conneries », de « foutaises » ou de « bullshit », en utilisant des interjections comme « pff » ou « grr », ou encore des expressions comme « c'est des bêtises », « c'est n'importe quoi », « on s'en fout ». La majorité de ces interventions sont laconiques et lapidaires et traduisent un agacement face à ce qui est perçu comme une perte de temps ou une distraction inutile (e.g. « marre de perdre du temps à lire vos posts »). Les utilisateurs éprouvent probablement un sentiment de « fatigue informationnelle » (Mercier, 2023) ou de « saturation cognitive » et semblent ainsi chercher à recentrer la discussion sur des sujets qu'ils estiment plus dignes d'intérêt sans avoir à s'engager dans une argumentation détaillée.

*Critique de la source du contenu*

Un troisième groupe de points d'arrêt est orienté vers la source d'un contenu en dénonçant son manque de fiabilité, ses biais idéologiques, ses motivations économiques ou son intention parodique. Ici, la critique n'est donc pas dirigée contre le contenu d'une information en tant que tel mais contre la ligne éditoriale des médias ou la déontologie des journalistes à l'origine de sa production. Les utilisateurs vont alors qualifier le site de « complotiste », de « satirique », de « piège à clics » ou de « mensonger », ou encore avoir recours à des expressions comme « y-a-t'il encore une vraie presse d'information ? » ou « à quand un vrai travail journalistique ? ». Ces points d'arrêt s'appuient parfois sur des outils comme *Le Décodex*, qui évalue la crédibilité des sites d'information, ou font suite à une consultation de la section « à propos » d'un site web pour vérifier l'identité, les objectifs et la ligne éditoriale du média.

*Critique de la forme du contenu*

Un quatrième type de points d'arrêt porte spécifiquement sur la forme d'un contenu. En prêtant attention à différents détails, tels que la police, la ponctuation, la typographie, le logo d'un article ou sa composition visuelle, les utilisateurs font montre d'un certain « flair sémiologique » (Allard-Huver, 2018) qui les amène à questionner le réalisme et l'authenticité d'un contenu. Par exemple, s'ils repèrent sur une photo un doigt en trop ou des caractères issus d'alphabets différents, ils peuvent soupçonner l'image d'être un montage ou d'avoir été générée par une intelligence artificielle. Occasionnellement, les critiques des utilisateurs découlent d'une petite enquête, comme une recherche par image inversée permettant de vérifier l'origine d'une photo.

*Critique du locuteur*

Un cinquième groupe de points d'arrêt s'adresse directement au locuteur à l'origine de la publication d'un contenu sur une page ou un groupe Facebook. Dans ce cas, les utilisateurs ne critiquent pas directement le contenu lui-même, ni le média à l'origine de sa production, mais plutôt l'acte de le partager et de lui conférer de la visibilité sur les réseaux sociaux. Autrement dit, l'intervention des utilisateurs vise la responsabilité énonciative du locuteur, c'est-à-dire la décision et les intentions derrière le partage du contenu (e.g. « Et quoi? Vous voulez faire passer quel message ? Votre publication n'a aucun intérêt sur ce groupe… »). Ces points d'arrêt sont parfois (dits) accompagnés par des actions concrètes visant à sanctionner le locuteur, par exemple en signalant son profil, en se désabonnant de sa page ou en quittant le groupe : « Voilà trop c'est trop, à force de publier ce genre d'articles honteux et autres débilités, je me désabonne ! » ; « Et sinon vous pensez vous excusez et supprimer ça à un moment ??!!! Si vous vous prenez pour une page d'information faites le bien. »)

*Critique des autres commentateurs/récepteurs*

Enfin, un sixième ensemble de points d'arrêt concerne les autres commentateurs d'un post et, plus largement, son audience imaginée, c'est-à-dire les utilisateurs qui y sont exposés mais n'y réagissent pas publiquement. Le plus souvent, il s'agit de critiques adressées à la cantonade qui ne ciblent pas d'utilisateurs précis ou de commentaires particuliers. Dans ces points d'arrêt, les utilisateurs blâment la crédulité des autres, ou plutôt les croyances qu'ils leur prêtent, avec des formules telles que : « Dire qu'il y en a qui vont y croire… » ou « j'imagine même pas l'état du cerveau des personnes qui sont persuadés de la réalité de ce mensonge ». Cette posture critique est proche de l'effet « troisième personne », suggérant que les individus ont tendance à croire que les autres sont plus susceptibles d'être influencés par des contenus médiatiques qu'eux-mêmes. D'une certaine manière, ces points d'arrêt comportent une dimension démonstrative et ostentatoire : les utilisateurs qui les expriment ne cherchent pas tant à rectifier ce qu'ils perçoivent comme des erreurs qu'à afficher leurs compétences en matière de discernement et de critique et leur maîtrise des outils et des codes de la vérification en ligne.

Le travail d'annotation réalisé sur un échantillon de commentaires a ainsi permis de relever toute une gamme d'expressions et de pratiques pouvant être qualifiées de points d'arrêt. Certains sont étayés par des preuves ou accompagnés d'actions punitives, tandis que d'autres sont simplement énoncés sous la forme d'opinions ou de remarques sommaires dans des registres oscillant entre moquerie et indignation. L'expression de points d'arrêt semble ainsi se situer à l'interstice de différentes pratiques : de l'alerte à la dénonciation en passant par l'enquête ou la sanction. Elle peut aussi prendre la forme de pinaillage ostentatoire, de taquinerie complice ou de moquerie indulgente. Cette variété dans les modalités d'expression de points d'arrêt confirme la pertinence du choix de forger un concept flexible, tenant compte des spécificités des pratiques numériques, plutôt que de reprendre des notions initialement établies pour appréhender des pratiques se déroulant hors ligne.

Les pratiques expressives des utilisateurs ont également montré qu'ils dirigeaient leurs critiques vers différentes entités : le contenu, son format, sa source, son locuteur, ses commentateurs ou récepteurs. Inspirée par la pragmatique de l'énonciation, cette approche permet d'illustrer comment le grand public perçoit et définit le problème des *fake news*. Dans l'ensemble, les utilisateurs qui expriment des points d'arrêt paraissent avoir été sensibilisés aux discours des journalistes et des institutions. Ils ont recours à des termes similaires à ceux utilisés dans les discours publics et se sont appropriés certains outils de vérification et de régulation. Leur approche dépasse cependant la simple vérification des faits pour englober une critique plus large des mécanismes de diffusion et d'interprétation de l'information. Plutôt que de corriger simplement une erreur dans un énoncé, ils semblent surtout attachés à préserver la qualité de l'atmosphère de leurs espaces de communication (sans que celle-ci ne soit pour autant orientée vers un horizon normatif précis). Cela suggère qu'ils ne perçoivent pas les contenus médiatiques comme des entités isolées, mais comme des éléments inscrits dans un réseau complexe de production, de diffusion et de réception de l'information. Cette prise en compte du point de vue des utilisateurs à partir de leurs points d'arrêt renforce ainsi l'idée de parler de troubles de l'information et de la communication plutôt que de problèmes de *fake news*.

***En-deçà des points d'arrêt : d'autres formes de distanciation critiques et réflexives***

L'analyse du corpus a également fait ressortir des formes subtiles d'engagement critique qui, sans être tout à fait des points d'arrêt, font un pas de côté par rapport au message véhiculé dans l'énoncé et dénotent ainsi une posture réflexive de la part des utilisateurs des réseaux sociaux. En effet, de la même manière que notre rapport à un récit n'est pas limité à l'alternative binaire croire/ne pas croire et qu'il subsiste toujours un régime pluriel du croire (Veyne, 1983), notre rapport à une information ne se limite pas à être d'accord/ne pas être d'accord ; réfuter ou approuver un énoncé. On a pu dénombrer quatre types de réactions qui, sans être tout à fait des points d'arrêt témoignent d'une forme de distance critique de la part des utilisateurs.

*Alignement incrédule*

Le premier type de réaction peut être qualifié d'« alignement incrédule » dans la mesure où le commentaire produit est aligné avec le point de vue défendu par le locuteur ou l'énoncé partagé mais émet un doute quant à sa valeur de vérité. Dans la plupart des cas, ces commentaires sont introduits par l'expression « si c'est vrai », puis se poursuivent par des propos injurieux ou indignés, écrits dans un style emporté. Par exemple, en réaction au contenu ci-dessous (cf. Figure 5.1), quelques commentaires insultent Emmanuel Macron et appellent à sa destitution tout en laissant ouverte la possibilité que leur jugement soit infondé : « Si réellement il a dit cette horreur, il doit être juger pour trahison et incitation aux attentats, doit être destituer » ; « J'espère que cette info est fausse, sinon destitution immédiate et le faire juger lui et son gouvernement par le peuple ». Ces tournures permettent aux utilisateurs d'exprimer vivement leurs opinions tout en restant dans le cadre de la supposition, c'est-à-dire sans s'engager pleinement sur la véracité des faits. Certains utilisateurs vont aussi modaliser leur discours avec des expressions comme « bien sûr tout ça est à mettre au conditionnel » et font ainsi preuve d'une certaine prudence énonciative à l'instar des utilisateurs de Twitter étudiés dans le chapitre précédent. Quelques commentaires enfin vont pleinement reconnaître qu'ils perçoivent le contenu comme une *fake news* mais minimiser l'importance de son manque de véracité en ayant recours à des conjonctions ou locutions adversatives comme « mais » ou « quand même », rappelant les réflexions d'Octave Mannoni (1964) sur la croyance et le déni dans son texte intitulé « Je sais bien mais quand même » ou les travaux de linguistes appréhendant l'expression « quand même » comme un marqueur interactif situé dans une zone intermédiaire entre l'approbation et la réfutation (Moeschler et De Spengler, 1981).

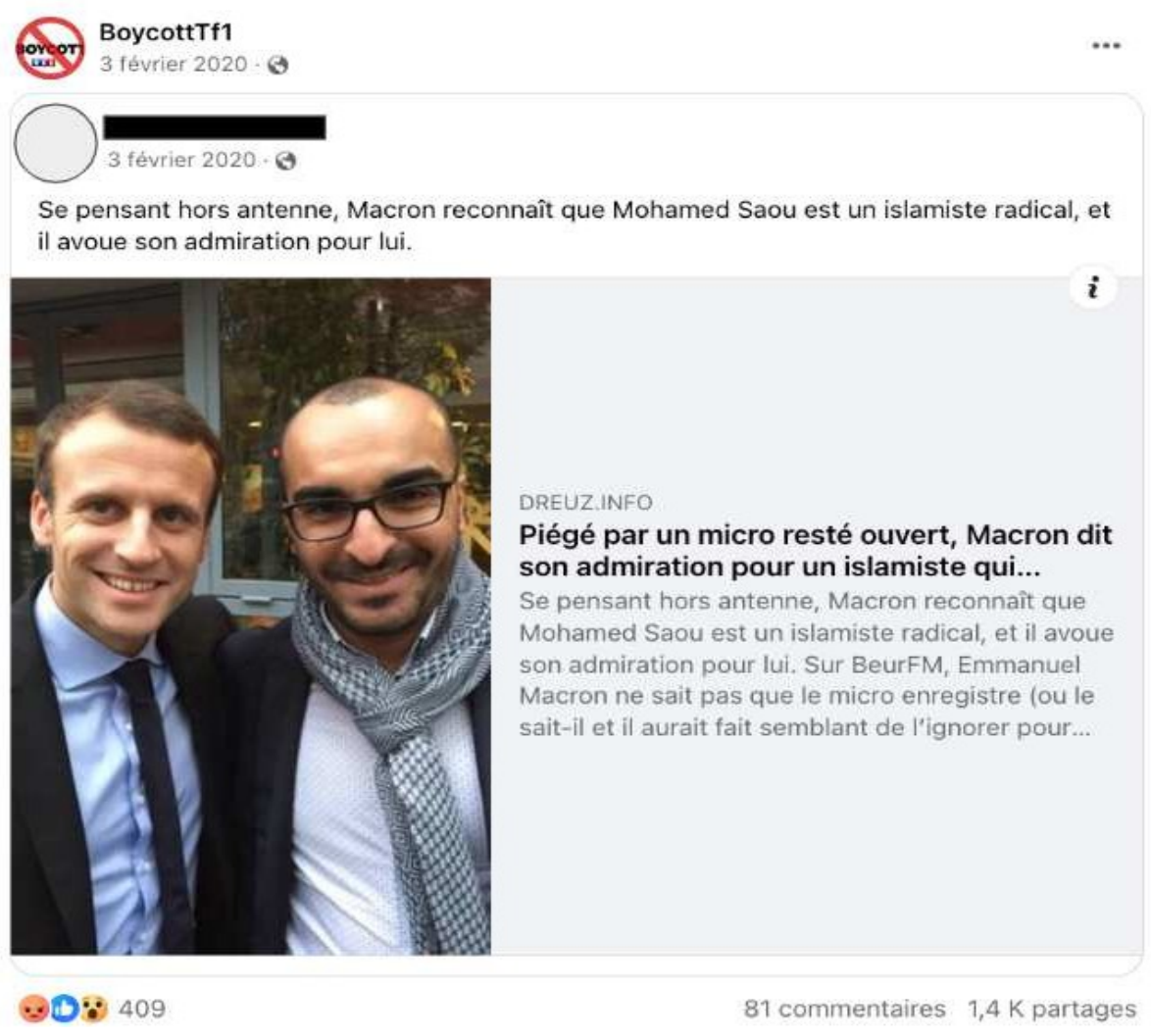

*Figure 5.1. Exemple de contenu ayant déclenché des réactions d'alignement incrédule*

*Suspension de l'incrédulité*

Le deuxième type de commentaire, souvent émis en réaction à des contenus parodiques, s'apparente à une forme de « suspension de l'incrédulité » semblable à celle que mettent en œuvre les récepteurs d'une œuvre de fiction. Les utilisateurs savent que le contenu publié n'est pas factuel mais satirique. Cependant, ils le jugent tout de même « vraisemblable », « probable », « pas loin de la vérité » ou « proche de la réalité ». Ils n'y croient pas vraiment mais s'amusent à y croire et à jouer avec l'idée qu'il pourrait être vrai, voire expriment le souhait qu'il le soit en rédigeant des commentaires tels que « Mince, j'aurais voulu y croire ! » ou « Par contre, ça aurait été tellement drôle ». De la même manière que le détour par la fiction a permis à de nombreux philosophes des Lumières de « [donner] au combat de la raison l'allure d'une fête » (Barthes, 1973, p. 83), le recours à la parodie permet de refléter la réalité avec un miroir déformant pour mieux la critiquer. Par exemple, le post ci-dessous fait référence de façon très exagérée aux discussions sur l'allongement de l'âge légal de départ à la retraite. Plusieurs commentaires confirment que les utilisateurs dressent un parallèle entre les contenus parodiques et les œuvres de fiction : « Plus qu'un article du Gorafi, un roman d'anticipation » ; « Le Gorafi un voyant incompris » ; « Satirique ou visionnaire ? ».

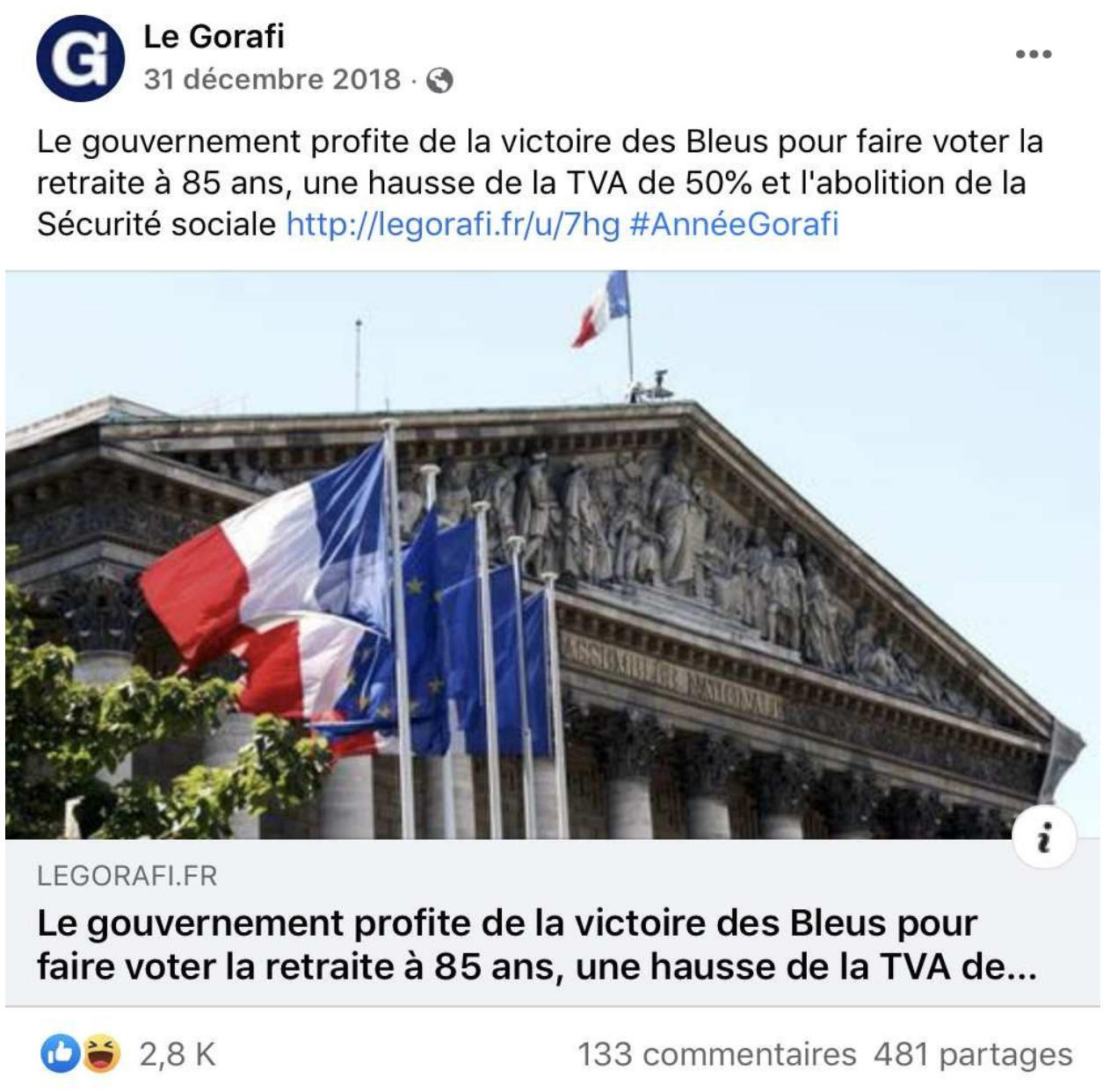

*Figure 5.2. Exemple de contenu ayant déclenché des réactions de suspension d'incrédulité*

### *Ironie et second degré*

Le troisième type de réaction se caractérise par un usage de l'ironie et du second degré pour exprimer différentes formes de désaccords (pour une analyse linguistique de cette pratique discursive, voir Baklouti, 2016). En conceptualisant l'ironie comme une reprise en écho ou une mention implicite d'un énoncé que le locuteur désapprouve, Dan Sperber et Deirdre Wilson (1981) ont avancé l'idée que l'essence de l'ironie ne serait pas de dire simplement le contraire de ce que l'on pense mais surtout de se distancier d'un point de vue énoncé par un autre locuteur. Sur Facebook, les utilisateurs qui ont recours à un registre d'énonciation ironique pour critiquer un contenu le font souvent en ajoutant une touche d'humour. Par exemple, les commentaires : « Mince, il est remort », « Michel Sardémarre », « Il meurt combien de fois par an lui ? » exprimés face au contenu parodique ci-dessous font écho de façon moqueuse à un ancien article de *Nordpresse* ayant déjà annoncé la mort de Michel Sardou et souligne ainsi de manière implicite le caractère absurde du contenu.

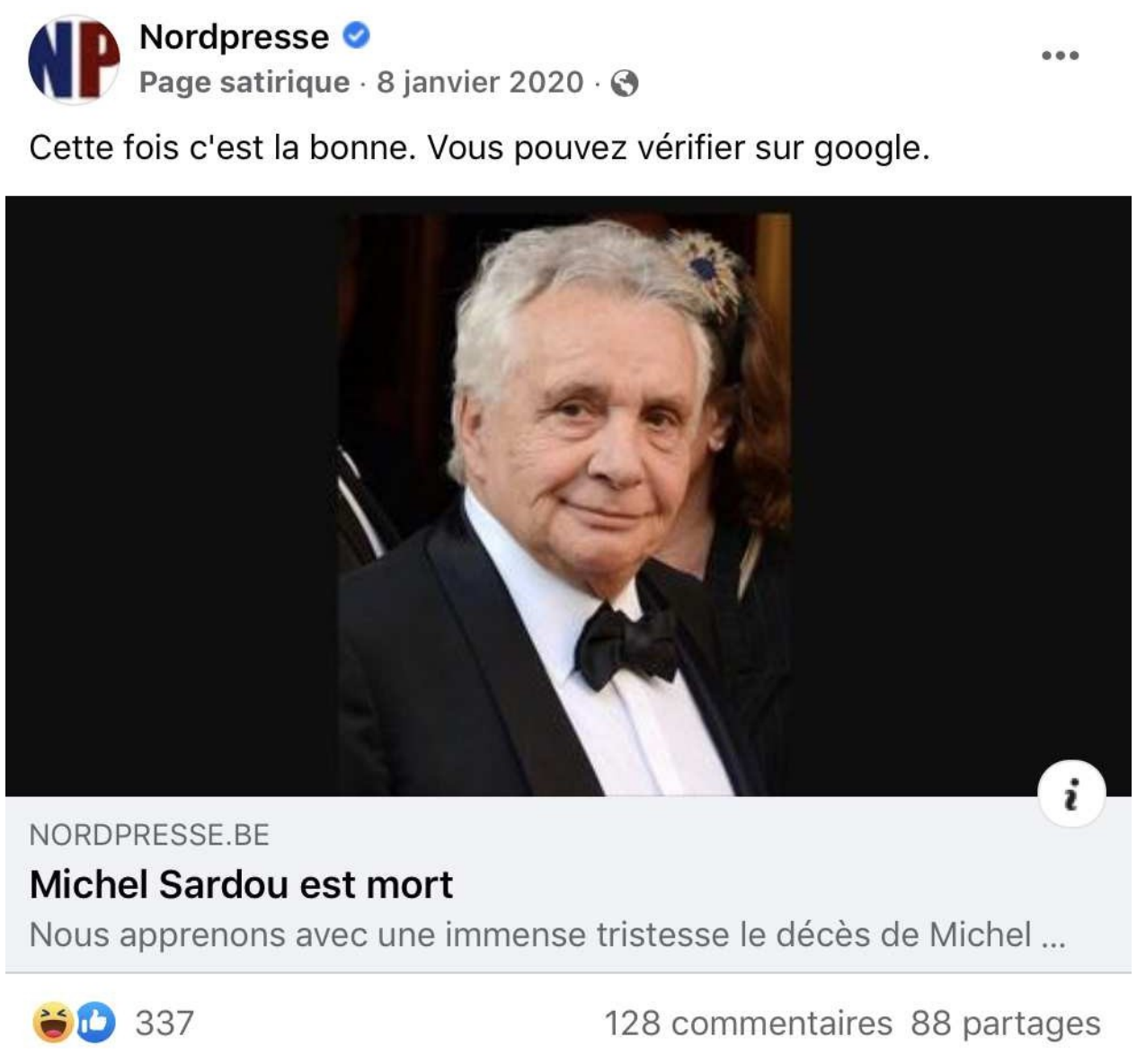

*Figure 5.3. Exemple de contenu ayant déclenché des réactions ironiques*

*Prise de conscience et auto-correction*

Enfin, le quatrième type de réaction dénote une forme de prise de conscience suivie d'une auto-correction de la part des commentateurs. Après avoir été momentanément induits en erreur par un post, les utilisateurs s'en aperçoivent et l'admettent publiquement en écrivant des commentaires tels que « ah zut...je me suis fait avoir » ou « Abuser putain j'avais pas vue. Ça craint ». Dans la majorité des cas, ces réactions sont émises face à des contenus parodiques, notamment provenant de la source *BMF news* pouvant facilement être confondue avec le média *BFM TV* (cf. Figure 5.4). Ces commentaires suggèrent que les utilisateurs ont dans un premier temps eu un aperçu rapide du contenu et de sa source, mais ont été suffisamment surpris ou dérouté pour y regarder par deux fois et tenter de mettre fin à leur trouble. Cette alternance entre distraction et attention, caractéristique d'un regard fureteur et curieux, est proche du régime d'excitation exploratoire conceptualisé par Nicolas Auray (2013).

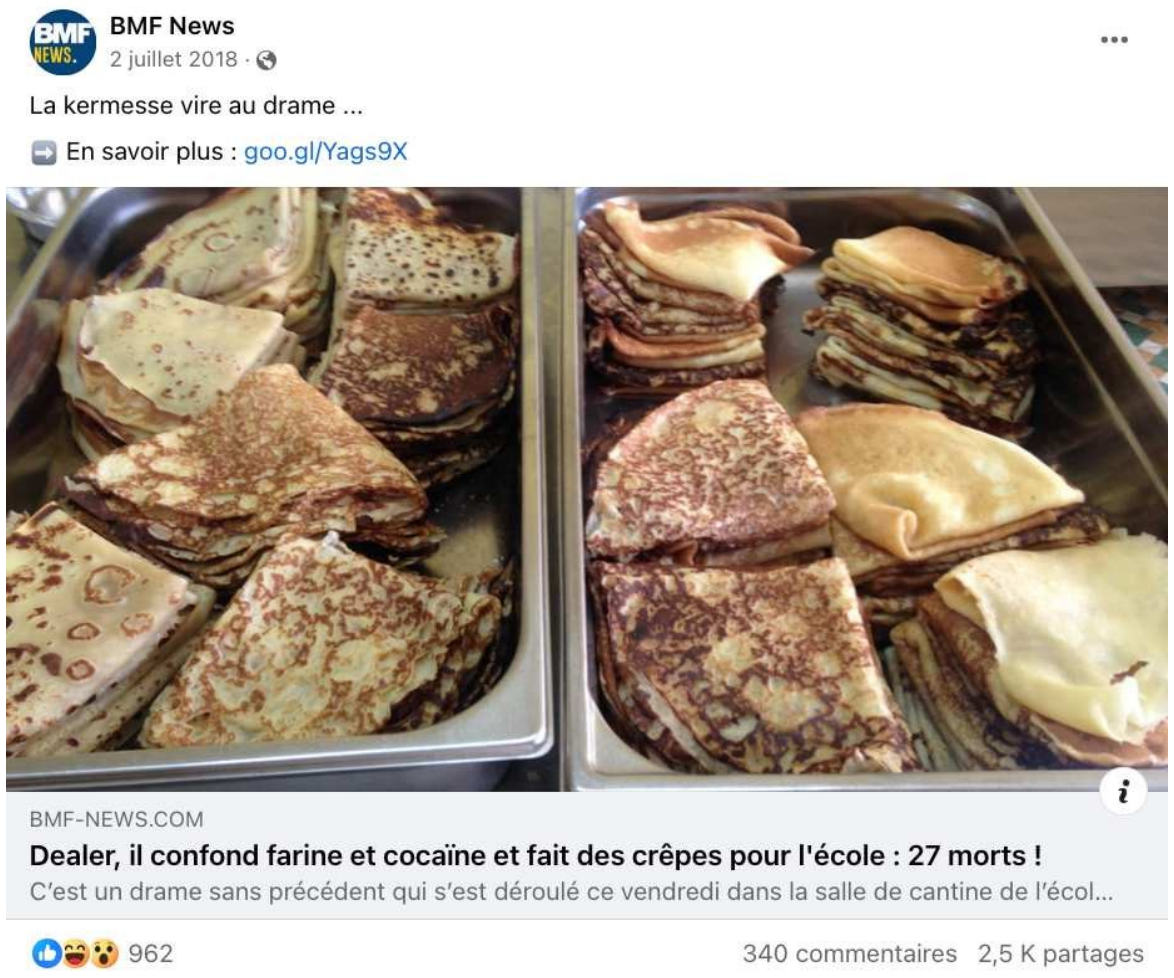

*Figure 5.4. Exemple de contenu ayant déclenché des réactions de prise de conscience et d'auto-correction*

### *Définition pratique des points d'arrêt*

Si les observations en ligne réalisées ont permis de tenir compte des multiples formes d'expression mobilisées par les utilisateurs de Facebook pour exprimer leur distance critique face à différents types de contenu, le travail d'annotation mené sur l'échantillon de 43 228 commentaires a été limité à une tâche binaire consistant à classer comme des « points d'arrêt » les commentaires exprimant une critique explicite – que celle-ci soit exprimée vers l'énoncé, le locuteur, la source, les autres commentateurs, etc. — et comme « autre » ceux n'en comportant pas. Pour le post ci-dessous, voici quelques exemples de commentaires classés comme « point d'arrêt » ou comme « autre ».

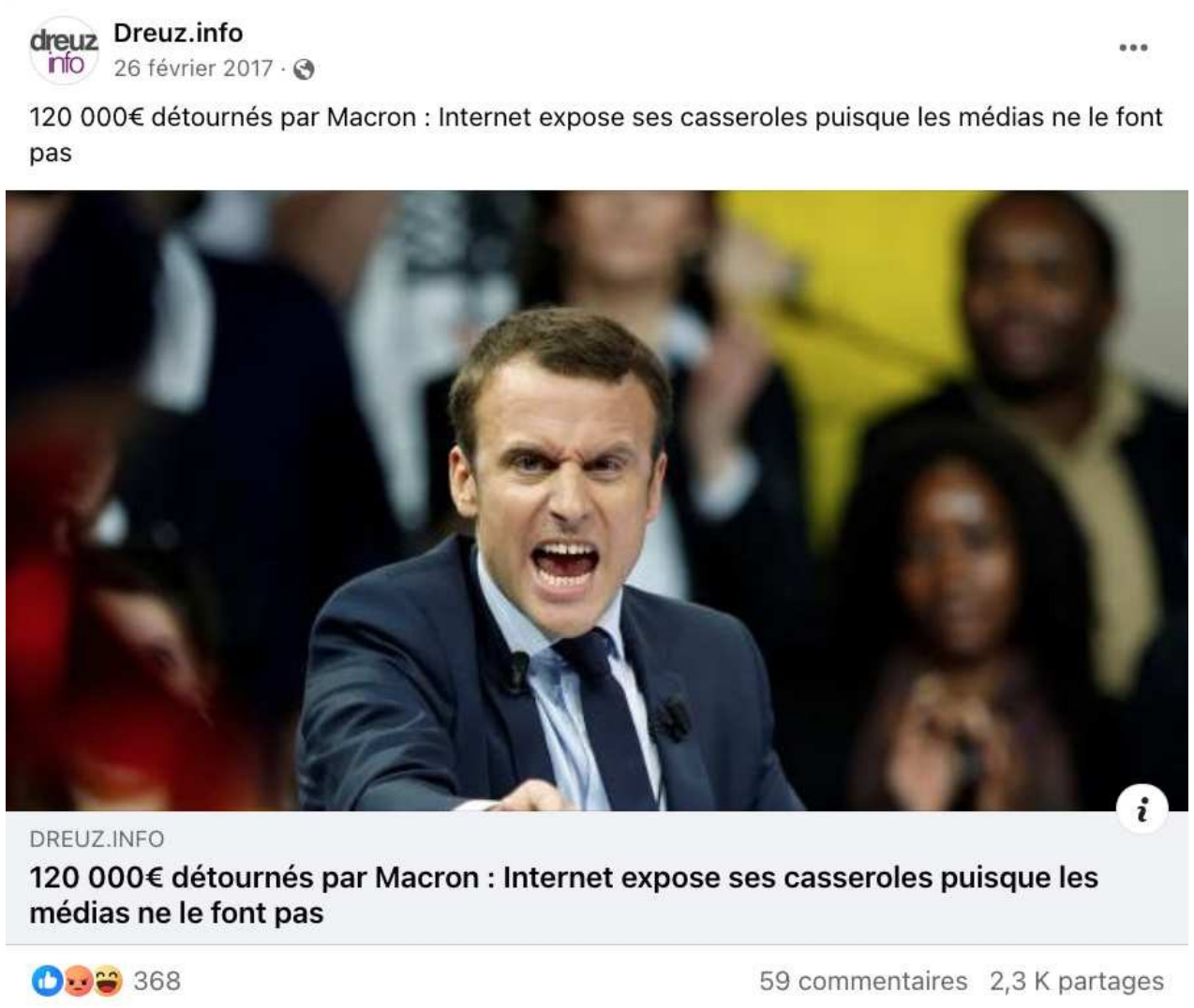

*Figure 5.5. Exemple d'un post Facebook dont les commentaires ont été annotés manuellement*

| Sélection de commentaires | Annotations |
|---|---|
| Voleurs de Politiciens! | autre |
| C'est faux vos casseroles restent chez vous et je vous supprime. | point d'arrêt |
| Il est jeune mais a déjà autant d'expérience que les vieux | autre |
| a vrai dire ! ils me font tous vomir !! | autre |
| INTOX | point d'arrêt |
| C'est n'importe quoi, t'as même pas lu l'article ils racontent que de la merde | point d'arrêt |
| Ce site diffuse régulièrement de fausses informations ou des articles trompeurs. Restez vigilant et cherchez d'autres sources plus fiables. Si possible, remontez à l'origine de l'information. | point d'arrêt |
| tu as vu ta tronche macron !!!!! | autre |
| un site complotiste | point d'arrêt |
| Fumier | autre |

*Tableau 5.1. Exemples de commentaires annotés comme « point d'arrêt » ou comme « autre »*

Le processus d’annotation a inclus des phases de lecture, de relecture et de discussions. Bien que ce travail ait demandé beaucoup de temps, il s'est avéré bénéfique pour la suite de l’enquête. Conduit en parallèle des observations menées sur les pages et groupes Facebook (*infra*), il a permis de s'imprégner pleinement des discussions qui se sont déroulées dans différents espaces de communication. Cela a également fourni une base solide pour recourir à des méthodes d’apprentissage supervisée comme cela sera détaillé dans la section suivante.

Au terme du travail d’annotation, 4 306 commentaires ont été identifiés comme des « points d’arrêt », soit près de 10 % de l’ensemble de l’échantillon.

### 5.1.3. Apprentissage supervisé

Afin d’étendre l’identification de point d’arrêt à l’ensemble du corpus de 441 149 commentaires, des méthodes d’apprentissage supervisé ont été mobilisées (pour plus d’informations sur l’utilisation de modèle de langage pré-entraîné en sciences sociales, voir Chapitre 2, section 2.2.3). Les annotations réalisées sur 43 228 commentaires ont permis d’entraîner un algorithme en utilisant le modèle de langage pré-entraîné BERT (Delvin et al., 2019) dans sa version francophone CamemBERT (Martin et al., 2019). Le score F1 obtenu est de 0,74 (cf. Figure 5.6), ce qui signifie que le modèle atteint un bon équilibre entre précision (la proportion de véritables points d’arrêt parmi ceux identifiés par le modèle) et rappel (la proportion de véritables points d’arrêt correctement identifiés parmi tous les points d’arrêt présents dans le corpus). Une relecture qualitative d’un échantillon d’une centaine de commentaires classés par le modèle a permis de confirmer la pertinence des prédictions (pour des exemples, voir Tableau 5.2). Pour minimiser le risque de faux positifs, seuls les commentaires classés comme des points d’arrêt avec une probabilité égale ou supérieure à 0,75 ont été retenus comme tels, tandis que les autres ont été recodés en « autre ».

| | precision | recall | f1-score | support |
|---|---|---|---|---|
| 0 | 0.97 | 0.98 | 0.97 | 3923 |
| 1 | 0.76 | 0.73 | 0.74 | 408 |
| accuracy | | | 0.95 | 4331 |
| macro avg | 0.86 | 0.85 | 0.86 | 4331 |
| weighted avg | 0.95 | 0.95 | 0.95 | 4331 |

*Figure 5.6. Évaluation des performances du modèle d'identification des points d'arrêt*

| Sélection de commentaires | Prédiction | Probabilité |
|---|---|---|
| C'est du mytho | stop | 0.991002 |
| Pas surpris de ça à l'époque les femmes étaient vu comme ca | stop | 0.684420 |
| je t'"ai déjà expliqué que tu diffuses une fausse info, c'est l'apprentissage de plusieurs langues avec un choix de langue, lis donc un peu les 1er commentaires, tu sais que le FN dis n'importe quoi | stop | 0.990790 |
| Je ne suis ni riche ni pauvre mais alors là en lisant ça j'ai eu un choc c'est de la discrimination honte à vous je ne sais pas qui vous eteS pour publier autant de connerie mais un conseil le matin regarder vous dans la glace et regarder au fond de vous cela ne doit pas être très jolie je vous plein et si cela as été écrit pour faire le buzz c'est encore pire je vous dit pas merci | stop | 0.987751 |
| Du grand n'importe quoi !!!!!! Le rédacteur de cet article est complètement ignorant sur le sujet ! Un tissu de mensonge, une vraie honte ! | stop | 0.992446 |
| Sauf que ce n'est pas le fait qu'elle aille ou non dans le sens de mon propos qui pose problème, mais le protocole et ses failles | stop | 0.775043 |
| Perso j'dis ça et j'dis rien mais la ressemblance entre la femme à droite, celle qui a les cheveux teintés en brun (en fait, il n'y en qu'une) et une certaine Brigitte M ....j'veux pas foutre la merde, hein! | no_stop | 0.969428 |
| J'exige que ça cesse | no_stop | 0.998840 |
| Elle est vraiment inconsciente | no_stop | 0.991131 |
| Ça m'étonnerait quand même qu'il soit libre après ce qu'il a fait ...en tout cas si ça s'avère vrai et bien là c'est une preuve que la justice ne font pas leur travail et qui incite les gens à faire du mal ! Je suis de Rouen alors je vais aller vérifier sa à la source direct.. d'ailleurs ils ont pas été assassiné sur place?? | no_stop | 0.936597 |
| Le pire c est que personne ne bouge pour défendre ce pauvre chien | no_stop | 0.998577 |
| Bien sûr que l'on pense au fric c'est le nerf de la guerre et il faut bien que l'on vive,au rythme ou il nous ponctionne il vas nous rester que les deux yeux pour pleurer, monsieur t'en mieux pour vous si vous en avez du fric | no_stop | 0.998852 |

*Tableau 5.2. Exemple de commentaires avec les labels prédits par le modèle et la probabilité associée*

Recourir à des méthodes d'apprentissage supervisé a permis de détecter 42 847 points d'arrêt au sein du corpus initial de 441 149 commentaires. Le taux moyen de points d'arrêt obtenu s'élève donc à 9,7 %.

Le modèle entraîné à partir de commentaires Facebook a également été utilisé pour détecter la présence de points d'arrêt au sein d'un corpus de 3 216 654 commentaires Youtube. Ces derniers représentent 10 % des commentaires analysés par des chercheurs du médialab dans le cadre d'une enquête dédiée à mesurer et cartographier l'empreinte antisémite et les discours de haine sur Youtube (de Dampierre et al., 2022 ; Tainturier et al., 2023). Cette fois-ci, 240 256 points d'arrêt ont été détectés, soit 7,42 %. Il est difficile de conclure à partir de ces seuls constats que les points d'arrêt sont moins fréquents sur Youtube que sur Facebook. L'objectif principal de cette réplication est surtout de montrer que le travail réalisé dans le cadre de notre enquête sur Facebook pourra alimenter de futures études, notamment sur d'autres espaces de discussion.

Bien que les points d'arrêt ne représentent qu'une petite proportion de l'ensemble des commentaires, ils ne sont pas négligeables et suggèrent une présence régulière de discours critiques dans les conversations en ligne. Des disparités importantes ont cependant été observées à l'échelle des posts et des pages ou groupes Facebook, où le nombre de points d'arrêt par post oscille entre 0 et 1 515, avec une moyenne de 1,42, une variance de 2 969 et un écart-type de 54,5. De plus, 49,8 % des posts contiennent au moins un point d'arrêt, tandis que 50,2 % des posts n'en contiennent aucun. Comment expliquer ces disparités ? Pourquoi le nombre de points d'arrêt varie-t-il autant d'un post à l'autre et d'une page ou d'un groupe à l'autre ?

Avant de répondre à cette question, il est nécessaire de présenter tout d'abord la méthodologie déployée pour caractériser les contenus ayant fait l'objet de signalements et les différents espaces de communication au sein desquels ils ont été partagés.

## 5.2. Un réseau social, des espaces de communication

De nombreuses études sur la circulation et la réception de l'information ont souligné l'importance d'adopter des approches multi-plateformes (Cardon et al., 2019 ; Cointet et al., 2021 ; Bode et Vraga, 2018 ; Rogers, 2023). Toutefois, peu de recherches ont pris en compte la diversité et l'hétérogénéité des espaces de communication qui existent au sein d'un même réseau social. Or, les pratiques des utilisateurs ne varient pas seulement d'une plateforme à l'autre, mais fluctuent également au sein d'une seule et même plateforme en fonction des différents espaces de communication qu'elle propose, de leur degré de visibilité et de la manière dont les utilisateurs s'approprient leurs fonctionnalités pour les adapter à leur sociabilité (Cardon, 2008 ; Balleys, 2014).

Après avoir présenté brièvement les principaux espaces de communication associés à différentes fonctionnalités de Facebook, ainsi que les usages variés qu'en font les utilisateurs, cette section se concentre sur les pages et groupes publiquement accessibles sur le réseau social. Malgré la diversité des contextes sociaux dans lesquels sont situés les utilisateurs exposés à leurs contenus, elle montre comment ces derniers ont la possibilité de s'appuyer sur des indices d'énonciation communs pour discerner le contrat de communication implicite à chaque page et groupe Facebook. La méthodologie mise en place pour identifier ces contrats de communication est alors documentée dans les deux dernières sous-sections.

### 5.2.1. Facebook et ses espèces d'espaces

Lancé par Mark Zuckerberg en février 2004, Facebook (initialement nommé *The Facebook*) était un site web conçu exclusivement pour les étudiants de l'Université de Harvard. Rapidement, la plateforme s'est ensuite étendue à d'autres universités américaines avant d'être ouverte au grand public en septembre 2006.

À ses débuts, Facebook était principalement utilisé par des adolescents et étudiants de premier cycle (Wilson et al., 2012). Plusieurs personnes interrogées dans le cadre de notre enquête ont d'ailleurs déclaré s'être inscrites sur la plateforme au collège ou au lycée. Elles considéraient alors le réseau social comme un prolongement de MSN ou de Skyblog et se rendaient dessus pour écrire des messages sur les murs de leurs amis ou partager des photos avec eux.

> *Pfiou… Alors quand ? Ça doit faire 12 ans je crois. J'ai 26 ans du coup ça devait être quand je devais avoir 14-15 ans. Du coup ça remonte un petit peu. Pourquoi ? Ben globalement parce que la plupart de mes amis étaient dessus et qu'il y avait pas mal d'informations qui transitaient par-là dans le groupe, notamment en termes de messagerie c'était pratique. Donc voilà au début c'était exclusivement pour un but social.*[172]

> *J'étais au lycée je crois. Voilà on a fait ça pour le fun. On ne savait pas ce que ça allait devenir. On sortait de MSN. Facebook, c'est arrivé du jour au lendemain. On a utilisé Facebook pour sa messagerie au début même si ce n'était pas comme aujourd'hui avec Messenger. Et en fait voilà c'était plutôt le réseau social dans sa définition la plus primaire avant l'évolution qu'on a pu constater aujourd'hui. C'était simplement pour échanger avec des pairs.*[173]

Au fil des années, ces utilisateurs sont devenus de jeunes adultes et leurs usages ont évolué avec l'apparition de nouvelles fonctionnalités sur la plateforme. L'introduction des pages publiques en 2007 a notamment marqué un tournant en permettant aux entreprises, associations et personnalités publiques de se connecter directement avec leurs audiences, transformant ainsi Facebook en une plateforme de communication professionnelle, marketing et politique. À partir de 2010, les groupes ont ensuite donné la possibilité aux utilisateurs de se rassembler autour de causes, de problématiques ou d'activités communes, qu'elles soient professionnelles, personnelles ou militantes, afin de faciliter l'organisation d'événements, de partage de contenus et de débats.[174] Un an plus tard, en 2011, c'est la fonctionnalité *Messenger* qui a été lancée pour permettre aux utilisateurs de communiquer entre eux de façon privée et instantanée. Enfin, en mars 2017, inspiré par le succès d'Instagram et Snapchat, Facebook a lancé la fonctionnalité *Stories* pour permettre aux

---

[172] Homme, 26 ans, Bac +6, interne en médecine, entretien réalisé le 18 janvier 2023.
[173] Homme, 34 ans, Bac +3, styliste-modéliste, entretien réalisé le 13 février 2023.
[174] Pour plus d'informations sur les différences entre les pages et les groupes Facebook, voir : https://www.facebook.com/notes/10160197330146729/

utilisateurs de partager des photos et des vidéos qui disparaissent après 24 heures.

À l'instar des forums étudiés par Valérie Beaudouin et Julia Velkovska (1999), Facebook constitue donc un « espace de communication multiforme » au sein duquel les utilisateurs ont la possibilité de publier des contenus sur plusieurs supports à la fois, tels que des pages ou groupes publics, des profils personnels, des *stories* ou messages privés. Coexistent ainsi différents espaces et situations d'interactions, plus ou moins publics et visibles, où règnent différentes ambiances, ce qui demande aux utilisateurs de faire preuve d'une certaine agilité, celle de basculer d'un régime d'énonciation à un autre sans provoquer de situation de *contexte collapse* (pour rappel, cette notion est discutée au chapitre 2, section 2.1.3) afin de naviguer sans accrocs entre ces multiples espaces.

Ces évolutions de Facebook ont été accompagnées de nouveaux usages. Aujourd'hui, bien que Facebook soit encore principalement utilisé pour se divertir et rester en contact avec des amis, une part significative des utilisateurs est régulièrement exposée à des contenus d'actualité politique et une minorité active participe au partage et à la discussion de contenus politiques. Par exemple, parmi les 68 % des Américains qui utilisent Facebook (Gottfried, 2024), 36 % s'y informent régulièrement, ce qui représente 54 % des utilisateurs de la plateforme (Shearer et Mitchell, 2021). Ainsi, Facebook n'est plus seulement un espace de divertissement et de conversation entre amis, mais est devenu, pour la plupart de ses utilisateurs, un espace hybride où se mêlent loisirs et informations :

> *Ça a forcément évolué parce que le réseau social a aussi évolué. Donc oui au début, c'était vraiment pour publier 2-3 trucs comme ça et puis après je me suis un peu piqué au jeu du débat sur les réseaux sociaux, surtout au moment où on pouvait commenter les articles, tout ça, dans des contextes soit d'élection, et puis après dans la lutte contre les fake news en santé et en science.*[175]

En parallèle de l'évolution de Facebook, il est important de souligner que le nombre d'usagers a aussi beaucoup augmenté. Aujourd'hui, le public de Facebook est beaucoup moins juvénile que par le passé. Les seniors ont rejoint la plateforme, tandis que les jeunes se sont tournés vers Instagram, TikTok et Snapchat pour échapper au regard de leurs aînés. Par ailleurs, contrairement à Twitter, Facebook est utilisé par un public plus large et populaire (Pasquier, 2018). Avec ce changement démographique, les usages de Facebook ont également évolué. Davantage de contenus concernant l'actualité politique sont partagés sur la plateforme par rapport à ses débuts dans différents registres oscillant de l'indignation injurieuse à la contestation argumentée.

[175] Homme, 44 ans, Bac +8, médecin, entretien réalisé le 2 août 2022.

En somme, le développement de Facebook s'est accompagné de deux dynamiques : d'une part, d'une démultiplication de ses espaces de communication ; de l'autre, d'une augmentation importante de son nombre d'utilisateurs. Plus que d'un élargissement des espaces de communication de Facebook et de ses publics, il est sans doute plus juste de parler d'une diversification. Ainsi, Facebook n'est plus seulement l'espace de divertissement et de conversation entre amis qu'il était à ses débuts, mais est plutôt devenu un espace hybride, où se mêlent loisirs et informations, utilisé par des individus de différentes générations.

### 5.2.2. Le cas des pages et des groupes publiquement accessibles sur Facebook

Malgré la diversité des espaces de communication disponibles sur Facebook, l'enquête menée dans le cadre de cette thèse se concentre uniquement sur des contenus et commentaires publiés sur des pages et des groupes publiquement accessibles. Elle ne permet donc pas de rendre compte des échanges qui se déroulent au sein d'espaces privés sur la plateforme.

Bien qu'il n'ait pas été possible d'observer directement les utilisateurs naviguer et interagir sur des pages et groupes Facebook publiquement accessibles, les entretiens réalisés ont permis de recueillir des informations sur les types de pages suivies par les enquêtés, sur la manière dont ils ont été exposés aux contenus, et sur leur perception de ces espaces de discussion. Les réponses montrent que les utilisateurs s'abonnent souvent à une variété de pages, allant des pages de divertissement aux journaux d'actualité, en passant par des groupes thématiques concernant des sujets variés comme la science ou les animaux. Par exemple, un enquêté décrit ses abonnements ainsi :

*Alors dans les pages de divertissement, j'ai pas mal de pages de memes, les neurchi, voilà en France. Je suis pas mal de pages de journaux différents, surtout français et j'ai pas mal aussi de journaux de littérature scientifique que je suis sur le site et ensuite c'est de groupes qui vont plus donner des… comment dire… ben chacun pour des intérêts variés là j'en ai un sous les yeux qui concerne les insectes.*[176]

Les utilisateurs peuvent être exposés de différentes manières à un post publié sur une page ou un groupe Facebook :
(1) via leur fil d'actualité – soit parce qu'ils sont abonnés à la page ou membres du groupe en question, soit parce qu'un de leur contact a réagi au post ;
(2) via une notification – si un de leur contact les a mentionné dans un commentaire ou leur a partagé le post en message privé ou si Facebook les avise de nouvelles activités sur la page ou le groupe ;
(3) via une recherche spécifique.

Une fois exposés à un contenu, les utilisateurs peuvent choisir de le consulter pour le lire attentivement, de passer à un autre post ou de réagir par un commentaire. Dans la plupart des cas, les utilisateurs semblent surtout commenter des posts auxquels ils ont été exposés à la suite d'une réaction d'un de leurs contacts :

*Non je ne la suis pas. En fait, je ne l'ai jamais suivie. Je ne sais pas comment je me suis retrouvé dessus. Je pense que j'ai dû passer de lien en lien pour me retrouver là ou quelqu'un a dû me citer une info qui venait de cette page et j'ai dû remonter le fil jusqu'à arriver à l'article où il parlait et du coup j'ai dû commenter. Je ne suis plus certain comment je suis tombé dessus honnêtement.*[177]

*Alors, de manière générale, je ne cherchais pas ce genre de page mais j'y tombais parce que j'avais des contacts qui… euh… En fait, j'ai découvert par exemple « La Gauche m'a tuer » parce que bah les contacts que j'avais… Je fais du tennis de table donc j'avais beaucoup de contacts qui sont des gens qui jouent en club de tennis de table et qui ne sont pas forcément… Ce sont des gens avec qui je suis en lien pour le sport mais pas forcément pour les idées et qui donc…*

[176] Homme, 26 ans, Bac +6, interne en médecine, entretien réalisé le 18 janvier 2023.
[177] Homme, 28 ans, Bac +5, développeur, entretien réalisé le 2 mars 2023.

*Bah postaient des choses où je me disais « ben c'est quoi ça ? » et où j'allais voir et je trouvais ça à la fois choquant mais intéressant de pouvoir discuter.*[178]

*En fait la manière donc je tombe sur ces liens-là, c'est que j'ai des amis intimes qui partagent n'importe quoi. Et qui partagent des discussions de comptoirs mais de ghettos, de quartiers. Et les discussions de comptoirs de quartier, c'est plein de stéréotype du style… Enfin on est dans la conspiration la plus extrême.*[179]

Tous les commentaires laissés sont potentiellement visibles pour l'ensemble des contacts du commentateur, ainsi que pour l'ensemble des abonnés ou membres de la page ou du groupe en question à condition qu'ils n'aient pas été supprimés. La manière dont les utilisateurs ont été exposés à un contenu peut façonner leur perception de l'audience qui y est également confrontée. Cette audience imaginée joue un rôle important dans leur décision de réagir ou non à un post, ainsi que dans leur façon de s'exprimer (Marwick et boyd, 2011). Par exemple, le fait d'être exposé à un contenu perçu comme incivil via leur fil d'actualité à la suite d'un commentaire d'un ami peut donner l'impression aux utilisateurs d'être dans une « sphère publique personnelle » (Gagrčin, 2022). En revanche, s'ils font une recherche pour retrouver un post publié sur un groupe, dont ils ne connaissent pas personnellement les membres, ils peuvent percevoir cet espace comme moins intime.

Par ailleurs, le contexte dans lequel les utilisateurs se trouvent hors ligne peut également influencer la manière dont ils vont appréhender une situation d'énonciation en ligne. Par exemple, si deux amis complices sont côte à côte derrière un écran de portable ou d'ordinateur, il est possible qu'ils réagissent à un post publié sur une page ou un groupe publiquement accessible pour s'en moquer ou rire ensemble. Cette proximité physique peut en effet renforcer leur sentiment de connivence et déteindre sur leur intervention en ligne.

Il ressort ainsi que les pages et les groupes publiquement accessibles sur Facebook constituent des espaces de communication très variés qu'il est impossible de classer sous une même catégorie. En effet, un même contenu publié sur une même page ou un même groupe sera reçu par une pluralité d'utilisateurs, tous situés dans des contextes sociaux différents et exposés à des audiences disparates. Cette diversité rend la réception des contenus extrêmement variable et difficile à capturer dans sa totalité. S'il est quasiment impossible de savoir comment les utilisateurs ont été exposés à un post – sont-ils tombés dessus par hasard ? ont-ils reçu une notification ? – il est en revanche possible de faire l'hypothèse qu'ils sont tous exposés à des indices communs leur permettant – pour peu qu'ils y prêtent attention – de cerner le contrat de communication sous-tendant les publications et discussions d'une page ou d'un groupe Facebook. Nous n'allons pas pouvoir saisir tous ces indices, ni étudier comment ceux-ci interagissent avec les contextes sociaux dans lesquels se trouvent les utilisateurs hors ligne. Nous proposons néanmoins d'identifier les différents contrats de communication susceptibles d'être perçus par les utilisateurs de Facebook à partir de différents indices potentiellement visibles par tous.

---

[178] Homme, 34 ans, Bac, accompagnant éducatif et social, entretien réalisé le 1er août 2022.
[179] Homme, 34 ans, Bac +3, styliste-modéliste, entretien réalisé le 13 février 2023.

### 5.2.3. La carte et les territoires : une analyse de réseau exploratoire

Afin de saisir les contrats de communication sous-jacents à différents groupes et pages Facebook, une analyse de réseau a dans un premier temps été conduite. L'idée était de relier ensemble des pages et des groupes ayant publié les mêmes URLs signalées. L'hypothèse était que des pages et des groupes partageant les mêmes contenus renfermeraient des contrats de communication assez proches. En effet, relayer des URLs provenant des mêmes sources et portant sur des thématiques identiques dénote une proximité en termes de centres d'intérêt et de valeurs — suggérant ainsi une orientation éditoriale similaire ou, pour le dire autrement, que les pages et les groupes diffusent la même « musique d'ambiance ».

À partir de notre base de données contenant 904 URLs signalées, partagées par 3 520 pages et groupes Facebook, un graphe bipartite a été créé puis projeté en graphe monopartite en utilisant les bibliothèques Python *NetworkX* et *Pelote*. L'algorithme de détection de communautés de Louvain a ensuite permis de faire ressortir plusieurs clusters (cf. Figure 5.7) et la bibliothèque *ipysigma* de visualiser le graphe de façon interactive (Plique, 2022).

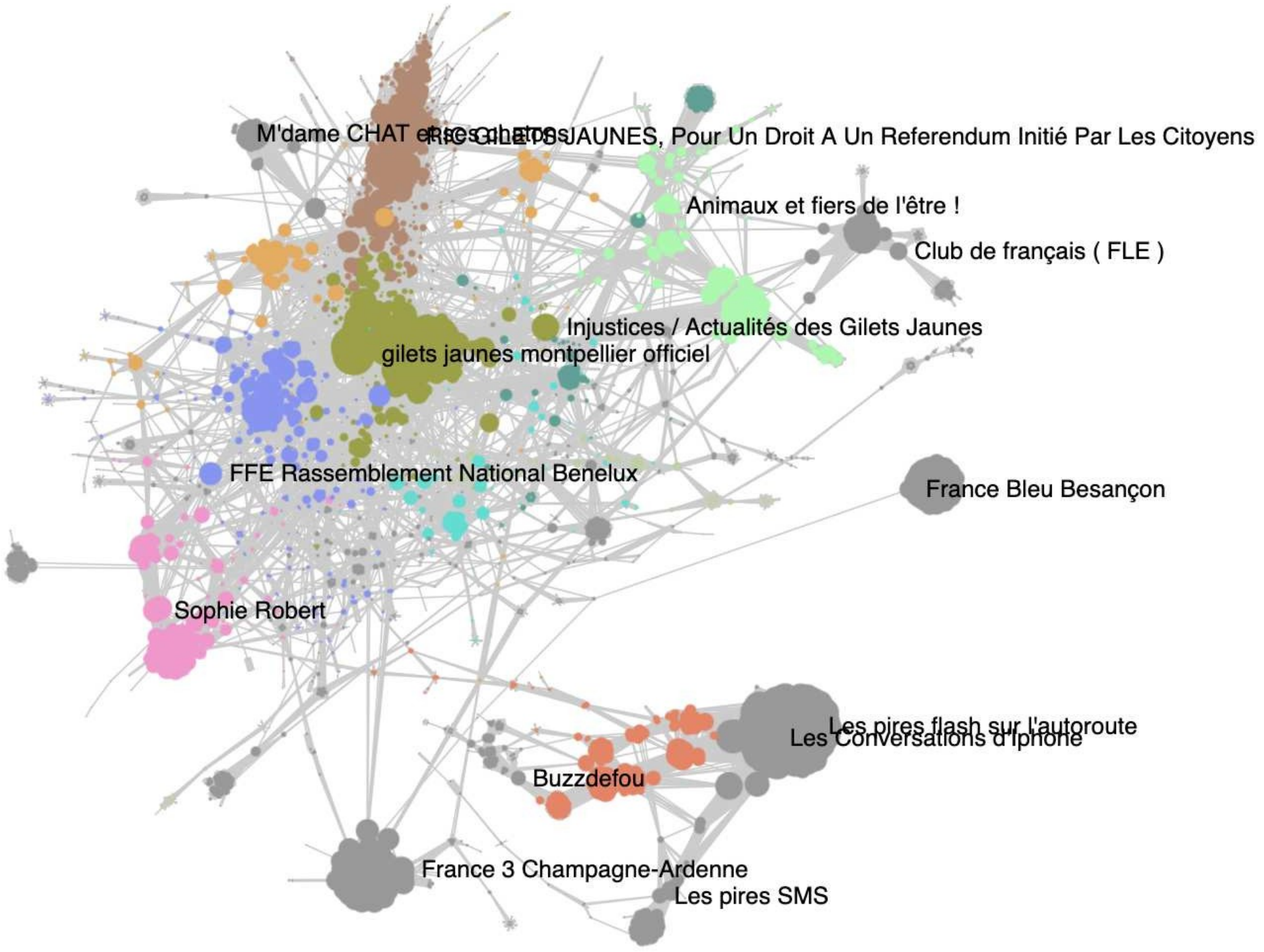

*Figure 5.7. Cartographie des pages et des groupes Facebook partageant des URLs signalées*

Sur le graphe ci-dessus, les clusters marron, jaune orangé et vert olive (en haut à gauche) sont composés de pages et de groupes associés au mouvement des Gilets Jaunes ; le cluster rose (en bas à gauche) à des pages et des groupes d'extrême droite ; et le cluster orange foncé (en bas à droite) à des pages et des groupes de divertissement. De leur côté, les clusters mauve, turquoise, vert clair, vert foncé et gris sont moins cohérents et plus difficiles à interpréter. Le cluster mauve semble regrouper des pages et des groupes dédiés à la politique mais associés à des sensibilités divergentes (aussi bien de gauche que de droite) ou sans affiliation partisane claire. De même pour le cluster turquoise, bien que celui-ci comporte davantage de pages et de groupes de gauche. Les clusters vert foncé et vert clair comportent quant à eux un ensemble de pages et de groupes dédiés aux animaux, à l'environnement, à la santé ou à des causes principalement défendues par des partis de gauche. Enfin, les divers clusters gris, situés en périphérie du graphe, rassemblent des pages et des groupes concernant des localités ou des sujets très précis (comme l'enseignement du français comme langue étrangère).

Afin d’obtenir des communautés davantage distinctes et cohérentes, il aurait sans doute été nécessaire de récolter toutes les URLs partagées sur une période donnée (par exemple de 6 mois ou 1 an) par chaque page et groupe du corpus, plutôt que de reposer seulement sur une liste d’URLs signalées. Cela aurait en effet permis de capturer un ensemble plus représentatif de leur activité de publications. Cependant, la collecte de telles données exigeait beaucoup de temps et aurait donné lieu à une base de données difficile à manipuler pour conduire une analyse de réseau.

En raison de ces limites, des annotations détaillées et des observations en ligne ont été réalisées pour décrire plus finement les différents groupes et pages Facebook du corpus et capturer avec plus de justesse les contrats de communication qui les sous-tendent.

### 5.2.4. Saisir les contrats de communication des pages et des groupes Facebook

Afin d’identifier avec le plus de finesse possible les contrats de communication sous-jacents à chaque espace de communication, des observations non participantes ont été conduites, avec l’aide de deux stagiaires, sur environ 200 pages et groupes Facebook, sélectionnés de façon aléatoire dans notre corpus. L'atmosphère et l’ambiance de chaque page et groupe ont été décrites sur un journal de terrain en faisant particulièrement attention aux détails suivants : (1) le nom de la page ou du groupe (et ses éventuels changements) ; (2) la description indiquée dans son onglet à propos ; (3) la présence ou l’absence d’une charte de modération ; (4) la taille de l’audience (i.e. le nombre d’abonnés ou de membres) ; (5) l’activité de la page ou du groupe (e.g. le nombre de publications par jour, semaine, mois ou année) ; (6) les types de sources et de contenus publiés ; (7) les réactions des utilisateurs (i.e. volume d’engagement, registre des commentaires, etc.). Les observations n’ont pas été réalisées au jour le jour mais en remontant les fils d’actualité des pages et des groupes. Quand les publications étaient très fréquentes, l’outil de recherche avancée de Facebook a été utilisé afin de saisir d’éventuelles évolutions au niveau des contenus publiés et des interactions entre les utilisateurs.

Suite à ces observations, des grilles d’annotation (cf. Annexes 5 à 7) ont été construites pour attribuer des catégories à chaque page et groupe Facebook du corpus, ainsi qu’aux contenus des URLs signalées afin d’affiner l’analyse de leurs sources et thématiques effectuée au chapitre 2 (cf. section 2.3.1). Un sous-ensemble commun d'URLs, de pages et de groupes Facebook a d'abord été codé de façon collective avec une collègue et deux stagiaires. Des discussions ont permis d'affiner les catégories et les critères d'annotation, puis de valider et stabiliser les grilles utilisées. Chaque page et groupe Facebook a ensuite été classé selon 17 modalités. La présence ou l'absence d'une charte de modération a également été notée. De leur côté, les sources des URLs et les thématiques des contenus ont été respectivement classées selon 7 et 12 catégories. Des seuils ont également été fixés pour qualifier la taille de leur audience comme (1) très petite ; (2) petite ; (3) moyenne ; (4) grande ; (5) très grande.

Grâce à ce travail d'annotation, des analyses quantitatives ont ainsi pu être réalisées (cf. Tableaux 5.3 à 5.5), permettant de confirmer ou d'affiner les constats issus des observations, notamment en mesurant la proportion de chaque catégorie de sources et de thèmes pour chaque catégorie de pages et de groupes. De leur côté, les entretiens ont permis de tenir compte des perceptions des utilisateurs par rapport aux espaces étudiés.

| Caractéristiques des pages et des groupes Facebook | | | | | | |
|---|---|---|---|---|---|---|
| Type de pages/groupes | nombre de pages/groupes | nombre de posts | nombre médian d'abonnés/de membres | nombre de commentaires | nombre moyen de commentaires par post | Pourcentage avec charte de modération |
| actualités diverses | 128 | 653 | 7 046 | 4 142 | 6 | 47,6 |
| vie locale | 375 | 1 656 | 68 039 | 15 634 | 9 | 55,4 |
| animaux | 191 | 623 | 20 158 | 16 386 | 26 | 37,0 |
| anti-système | 292 | 4 099 | 7 335 | 25 632 | 6 | 44,2 |
| centre | 51 | 144 | 3 596 | 671 | 5 | 39,6 |
| divertissement | 648 | 3 731 | 159 794 | 76 874 | 21 | 4,5 |
| droite | 109 | 551 | 6 477 | 6 804 | 12 | 27,8 |
| extrême droite | 239 | 2 534 | 5 586 | 23 478 | 9 | 17,7 |
| extrême gauche | 176 | 1 611 | 11 085 | 5 717 | 4 | 62,9 |
| gauche | 321 | 891 | 5 247 | 4 829 | 5 | 36,9 |
| gilets jaunes | 411 | 9 131 | 9 363 | 7 694 | 1 | 67,7 |
| humour | 89 | 795 | 128 721 | 105 916 | 133 | 10,7 |
| média alternatif | 43 | 222 | 56 788 | 4 143 | 19 | 10,8 |
| média mainstream | 68 | 560 | 1 873 207 | 82 093 | 147 | 15,0 |
| religion | 64 | 344 | 17 0296 | 1 227 | 4 | 41,0 |
| santé/ sciences/ bien-être | 258 | 2 503 | 12 6494 | 58 631 | 23 | 29,2 |
| éducation | 57 | 109 | 39 399 | 1 278 | 12 | 15,6 |

*Tableau 5.3. Statistiques descriptives des pages et des groupes Facebook de l'ensemble du corpus*

| Sources des URLs signalées | | | | | | | |
|---|---|---|---|---|---|---|---|
| Type de pages/groupes | contre-information | divertissement | Presse locale | *mainstream* media | parodique | partisan/ engagé | santé alternative |
| actualités diverses | 16,23 | 5,05 | 3,22 | 24,96 | 17,00 | 25,11 | 8,42 |
| actualités et vie locale | 3,26 | 3,99 | 12,20 | 65,34 | 5,13 | 5,50 | 4,59 |
| animaux | 1,12 | 26,65 | 6,26 | 12,84 | 17,17 | 19,58 | 16,37 |
| anti-système | 30,98 | 2,42 | 0,22 | 15,15 | 9,69 | 32,47 | 9,08 |
| centre | 4,86 | 0,69 | 2,78 | 53,47 | 6,94 | 27,78 | 3,47 |
| divertissement | 2,76 | 83,41 | 0,21 | 4,07 | 3,54 | 1,85 | 4,15 |
| droite | 33,03 | 1,09 | 1,63 | 19,60 | 10,53 | 31,03 | 3,09 |
| extrême droite | 62,35 | 1,34 | 0,20 | 10,06 | 7,22 | 15,94 | 2,88 |
| extrême gauche | 7,14 | 2,17 | 0,56 | 28,37 | 5,71 | 47,73 | 8,32 |
| gauche | 6,40 | 6,06 | 2,92 | 27,05 | 4,71 | 38,61 | 14,25 |
| gilets jaunes | 18,05 | 1,06 | 0,25 | 19,54 | 8,18 | 49,85 | 3,07 |
| humour | 6,42 | 27,30 | 0,13 | 7,55 | 39,62 | 10,69 | 8,30 |
| média alternatif | 48,20 | 0,00 | 0,00 | 19,82 | 5,86 | 16,22 | 9,91 |
| média mainstream | 0,00 | 0,18 | 1,79 | 96,25 | 0,00 | 1,79 | 0,00 |
| religion | 15,12 | 5,81 | 0,00 | 13,66 | 7,56 | 8,43 | 49,42 |
| santé/ sciences/ bien-être | 11,83 | 13,22 | 0,24 | 8,39 | 3,36 | 11,83 | 51,14 |
| éducation | 1,83 | 1,83 | 3,67 | 29,36 | 15,60 | 9,17 | 38,53 |

*Tableau 5.4. Pourcentage de chaque type de source des URLs signalées partagées par chaque type de page ou groupe*

| | Thématiques des URLs signalées | | | | | | | | | | | |
|---|---|---|---|---|---|---|---|---|---|---|---|---|
| Type de pages/groupes | animaux | inclusion sociale | crime/ violence | divertissement | environnement/ nature | faits divers | immigration | politique | religion | santé | société | économie |
| actualités diverses | 2,14 | 2,14 | 4,13 | 2,14 | 3,68 | 9,34 | 4,44 | 39,05 | 5,97 | 7,20 | 10,41 | 9,34 |
| vie locale | 4,95 | 1,03 | 3,68 | 43,60 | 4,53 | 10,81 | 1,09 | 13,59 | 0,48 | 3,20 | 8,76 | 4,29 |
| animaux | 82,83 | 0,32 | 0,32 | 1,12 | 2,25 | 3,37 | 0,00 | 0,96 | 0,16 | 5,62 | 2,73 | 0,32 |
| anti-système | 1,20 | 1,78 | 2,22 | 0,90 | 2,34 | 2,95 | 3,93 | 47,96 | 12,08 | 8,88 | 7,56 | 8,20 |
| centre | 0,00 | 2,78 | 0,69 | 1,39 | 2,08 | 0,69 | 2,78 | 65,97 | 9,72 | 2,08 | 6,94 | 4,86 |
| divertissement | 5,44 | 3,70 | 3,89 | 19,24 | 1,88 | 23,24 | 0,94 | 9,89 | 0,86 | 25,27 | 5,15 | 0,51 |
| droite | 0,18 | 2,54 | 3,27 | 1,63 | 0,54 | 5,63 | 14,88 | 36,66 | 18,15 | 0,36 | 10,71 | 5,44 |
| extrême droite | 0,16 | 1,89 | 1,42 | 0,79 | 0,95 | 3,67 | 15,04 | 26,72 | 40,57 | 2,29 | 3,08 | 3,43 |
| extrême gauche | 1,55 | 2,61 | 1,18 | 1,30 | 4,35 | 1,68 | 0,93 | 59,22 | 0,87 | 4,16 | 8,07 | 14,09 |
| gauche | 7,52 | 5,61 | 1,46 | 3,03 | 28,40 | 2,58 | 3,48 | 23,34 | 0,67 | 7,63 | 9,54 | 6,73 |
| gilets jaunes | 0,16 | 1,56 | 0,99 | 0,30 | 0,74 | 1,73 | 1,50 | 70,74 | 1,73 | 3,58 | 7,27 | 9,70 |
| humour | 1,89 | 1,64 | 3,27 | 8,55 | 1,13 | 11,82 | 2,26 | 40,50 | 2,01 | 13,71 | 7,80 | 5,41 |
| média alternatif | 4,05 | 0,90 | 4,95 | 3,60 | 4,05 | 3,15 | 7,66 | 44,59 | 4,50 | 6,31 | 8,11 | 8,11 |
| média mainstream | 3,57 | 4,64 | 5,18 | 8,57 | 7,32 | 8,21 | 2,86 | 25,36 | 1,43 | 12,86 | 13,39 | 6,61 |
| religion | 4,94 | 4,94 | 3,49 | 0,58 | 0,00 | 3,49 | 8,43 | 12,5 | 32,56 | 20,64 | 4,94 | 3,49 |
| santé/ sciences | 4,12 | 1,24 | 1,56 | 1,68 | 1,92 | 4,12 | 1,00 | 19,22 | 0,48 | 47,62 | 14,22 | 2,84 |
| éducation | 0,00 | 0,00 | 1,83 | 20,18 | 0,00 | 1,83 | 0,00 | 7,34 | 0,00 | 30,28 | 34,86 | 3,67 |

*Tableau 5.5. Pourcentage de chaque type de thématique des URLs signalées partagées par chaque type de page ou groupe*

Afin de visualiser plus facilement les relations entre toutes ces variables, une analyse factorielle de correspondance a été réalisée (cf. Figure 5.8). Cette analyse a permis de dégager deux axes principaux. Le premier axe, représenté horizontalement, semble caractériser le degré de politisation des espaces de communication. Plus on se déplace vers la droite, plus les contrats apparaissent politisés dans la mesure où l'on retrouve les contenus portant sur des sujets politiques ou des thématiques comme l'immigration et l'environnement, venant de médias partisans ou de contre-information. À l'inverse, les contenus situés dans la partie gauche du plan portent sur des sujets de divertissement ou traitent des faits divers.

Le second axe, représenté verticalement, est plus complexe à interpréter. Il semble permettre de saisir le degré de pluralisme d'un espace et la diversité de son audience dans la mesure où dans la partie supérieure du graphe se trouvent les pages et groupes d'humour, de divertissement, de médias *mainstream* et d'actualités diverses, tandis que dans la partie inférieure on peut voir les pages et groupes dédiés à la santé, à la religion ou associés à la droite et à l'extrême droite.

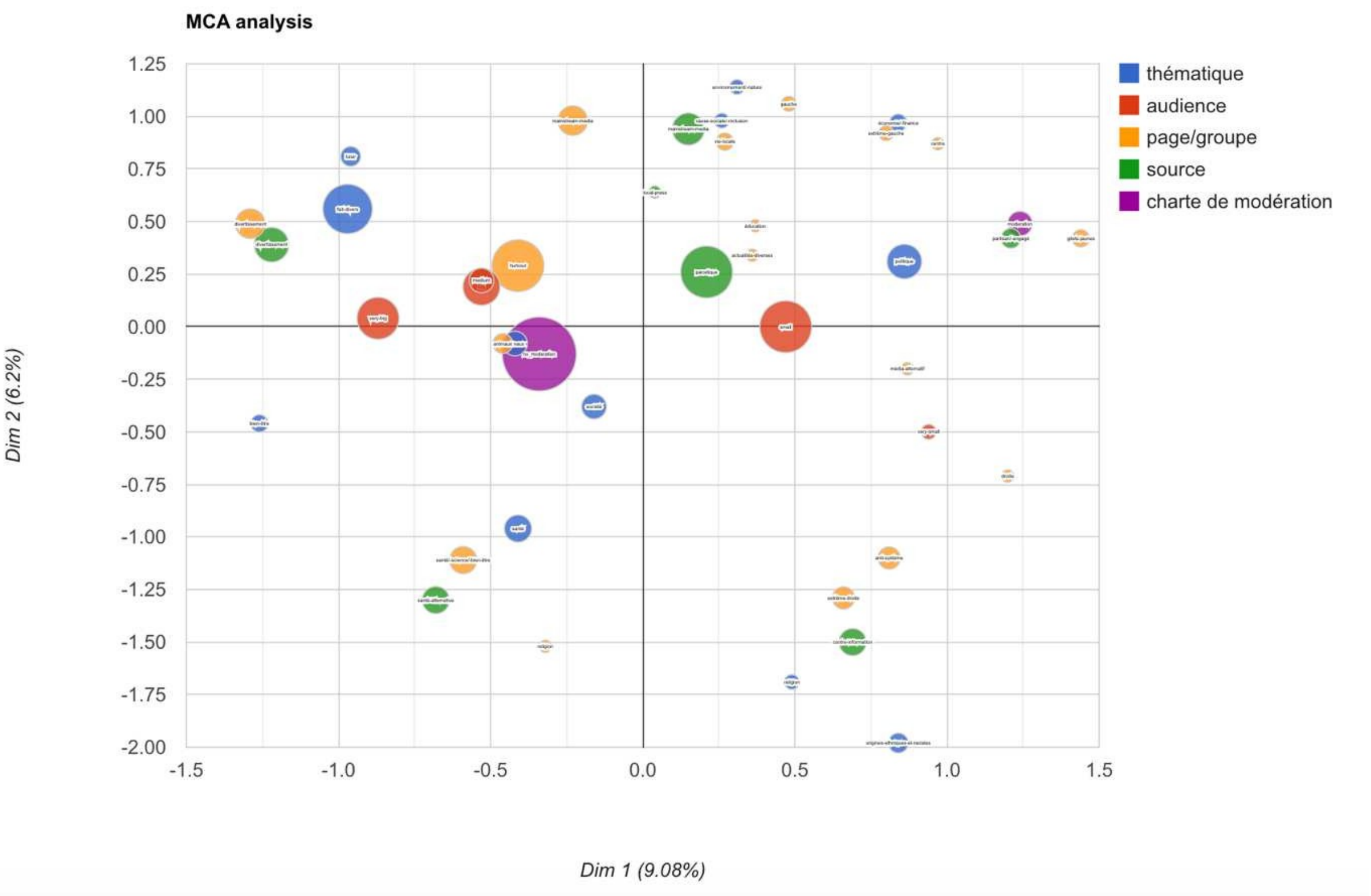

*Figure 5.8. Projection des variables sur un plan factoriel*

Afin de présenter brièvement ces différents espaces de communication, et les types de contenus qu'ils mettent en circulation, ces pages et groupes peuvent ainsi être regroupés en quatre types d'espaces : (1) politique ; (2) généraliste ; (3) spécialisé ; (4) récréatif.

Tout d'abord, une première série de pages et de groupes peut être qualifiée de « politique ». Dans l'ensemble, ces espaces de communication publient principalement des contenus dont les thématiques sont régulièrement discutées dans le débat public (e.g. élections, projets de loi, conjoncture économique, etc.). Des disparités importantes ont cependant été relevées entre plusieurs sous-groupes. Certains affichent une position politique précise et peuvent être situés entre l'extrême gauche et l'extrême droite de l'échiquier politique français. Leur nom ou leur onglet à propos indiquent qu'ils soutiennent une personnalité ou un parti spécifique (cf. Figure 5.9), et leurs publications sont très resserrées autour des thématiques portées par le programme de leur camp idéologique. D'autres ont des positions politiques plus élastiques ou flottantes. Par exemple, les pages et groupes associés au mouvement des gilets jaunes ou à des opinions anti-système, ou encore tenus par des médias alternatifs sont très critiques à l'égard des élites et des institutions, mais ne sont ni spécifiquement de gauche ou de droite (cf. Figure 5.10). Selon leur sensibilité politique, ces espaces vont publier des contenus différents. Sans surprise, les pages et groupes affiliés à des partis de gauche vont plutôt publier des contenus alignés avec leur idéologie, tandis que ceux affiliés à des partis de droite privilégient des thématiques qui leur sont propres. Les pages et groupes centristes, en revanche, tendent à publier des contenus plus *mainstream*, souvent en soutien au parti du gouvernement. Contrairement à l'enquête menée sur Twitter, il n'a pas été possible d'inférer la position idéologique des utilisateurs abonnés aux pages ou membres d'un groupe. Néanmoins, il est possible de faire l'hypothèse que ces utilisateurs sont davantage partisans que ceux abonnés à des pages ou membres de groupes non politiques. Autrement dit, plus une page ou un groupe est politiquement défini, plus il est probable que son réseau de discussion soit homogène.

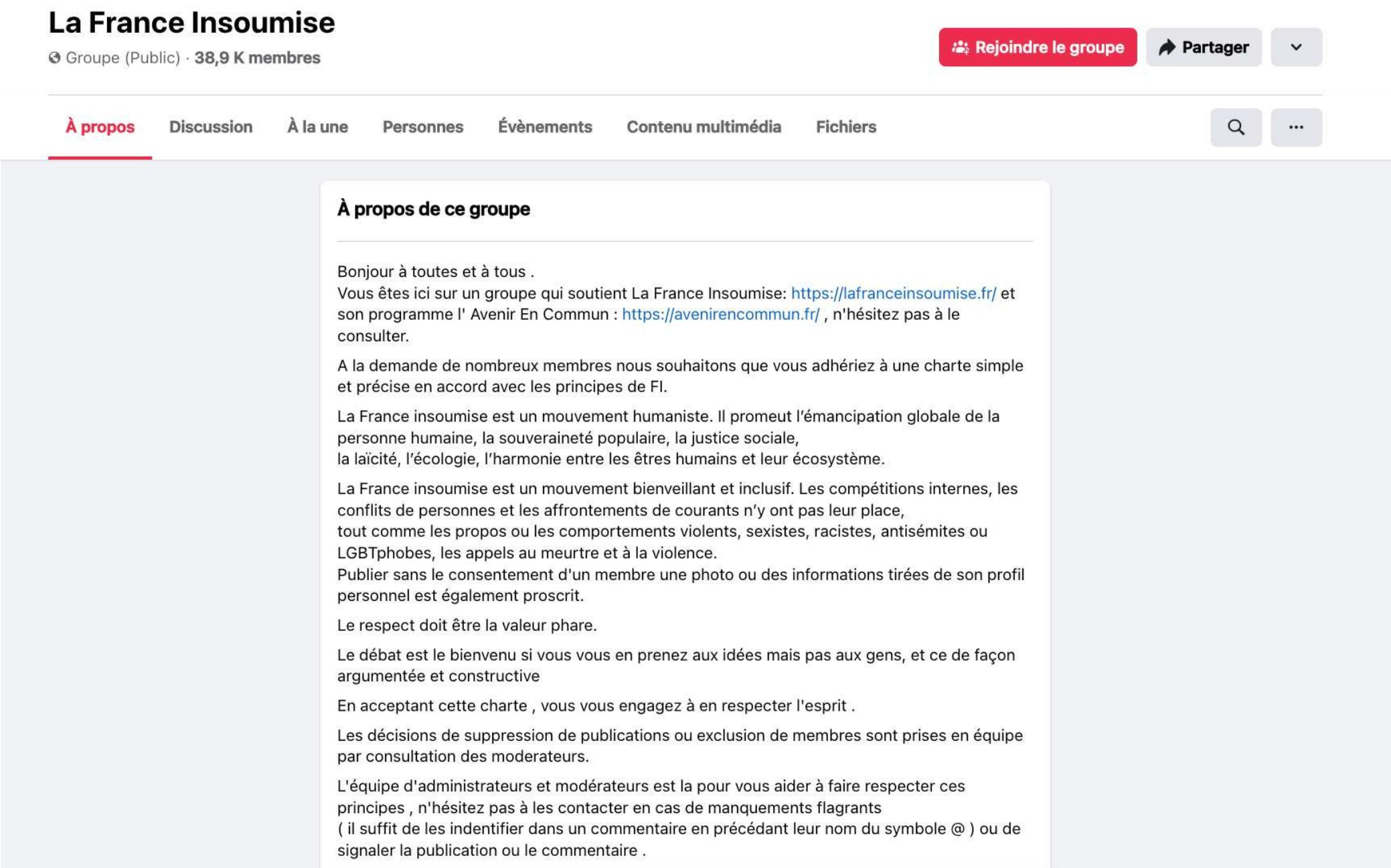

*Figure 5.9. Exemple d'onglet à propos d'un groupe de La France Insoumise*

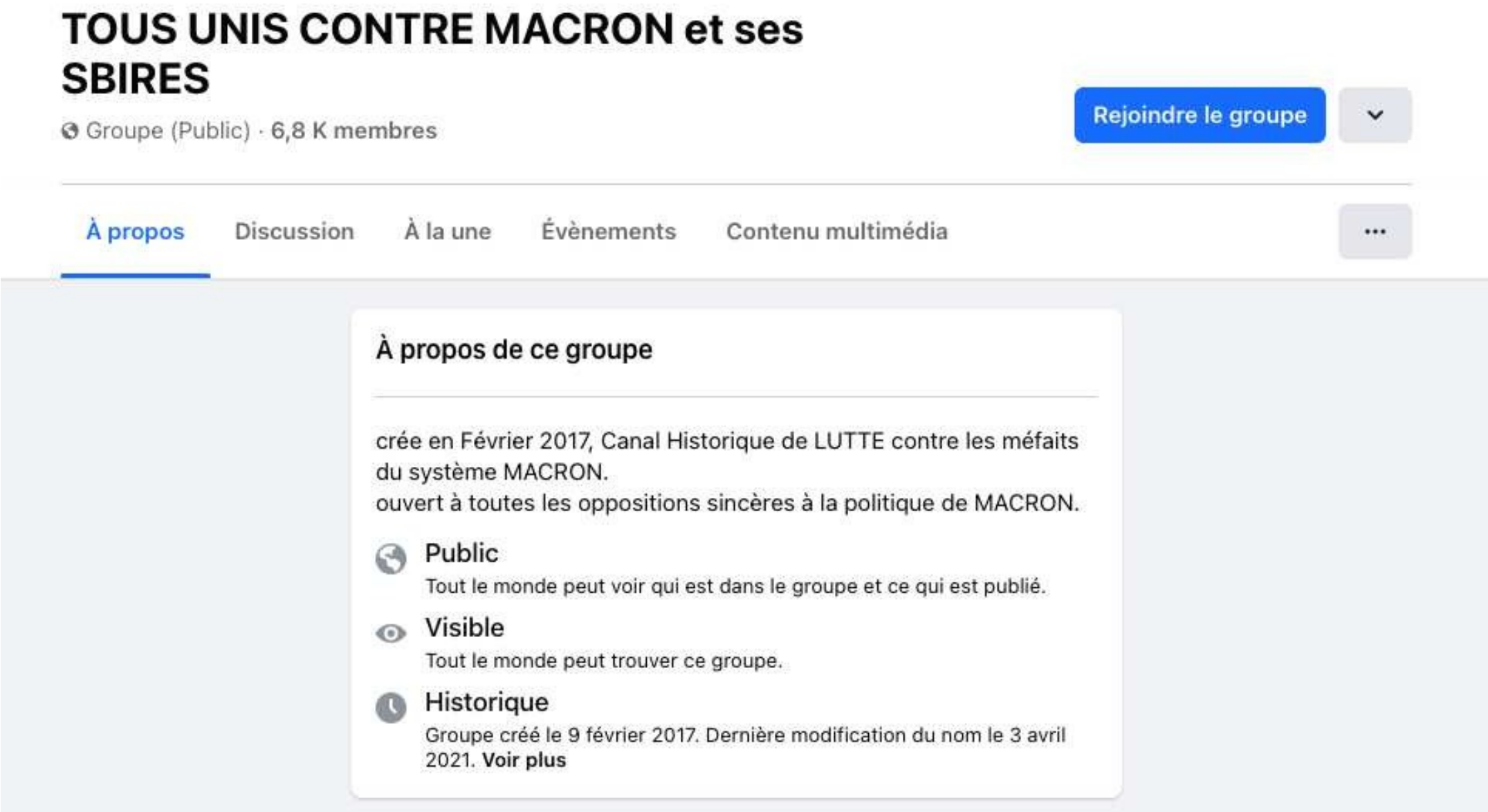

*Figure 5.10. Exemple d'onglet à propos d'un groupe anti-système*

Une deuxième série de pages et groupes peuvent être perçus comme « généralistes ». Ils visent à s'adresser au grand public ou à toucher des personnes avec des points de vue contradictoires. C'est le cas notamment des pages et groupes dédiés à débattre et discuter de l'actualité politique, tel que France : DÉBATS sur la Politique (cf. Figure 5.11). Ces espaces sont conçus pour favoriser l'échange et la confrontation des idées, et il est ainsi possible qu'ils attirent un public diversifié et parfois opposé dans ses opinions. Ces espaces généralistes concernent également les pages et groupes de médias *mainstream*, qui bien qu'ils aient pour la plupart une ligne éditoriale bien délimitée, ont un nombre médian d'abonnés beaucoup plus élevé que celui des autres groupes (i.e. 1 873 207 contre 158 753 en moyenne), ce qui leur permet de garantir un certain pluralisme des points de vue. Par ailleurs, ces espaces adoptent souvent une approche nuancée ou omnibus, cherchant à couvrir un large éventail de sujets et de perspectives.

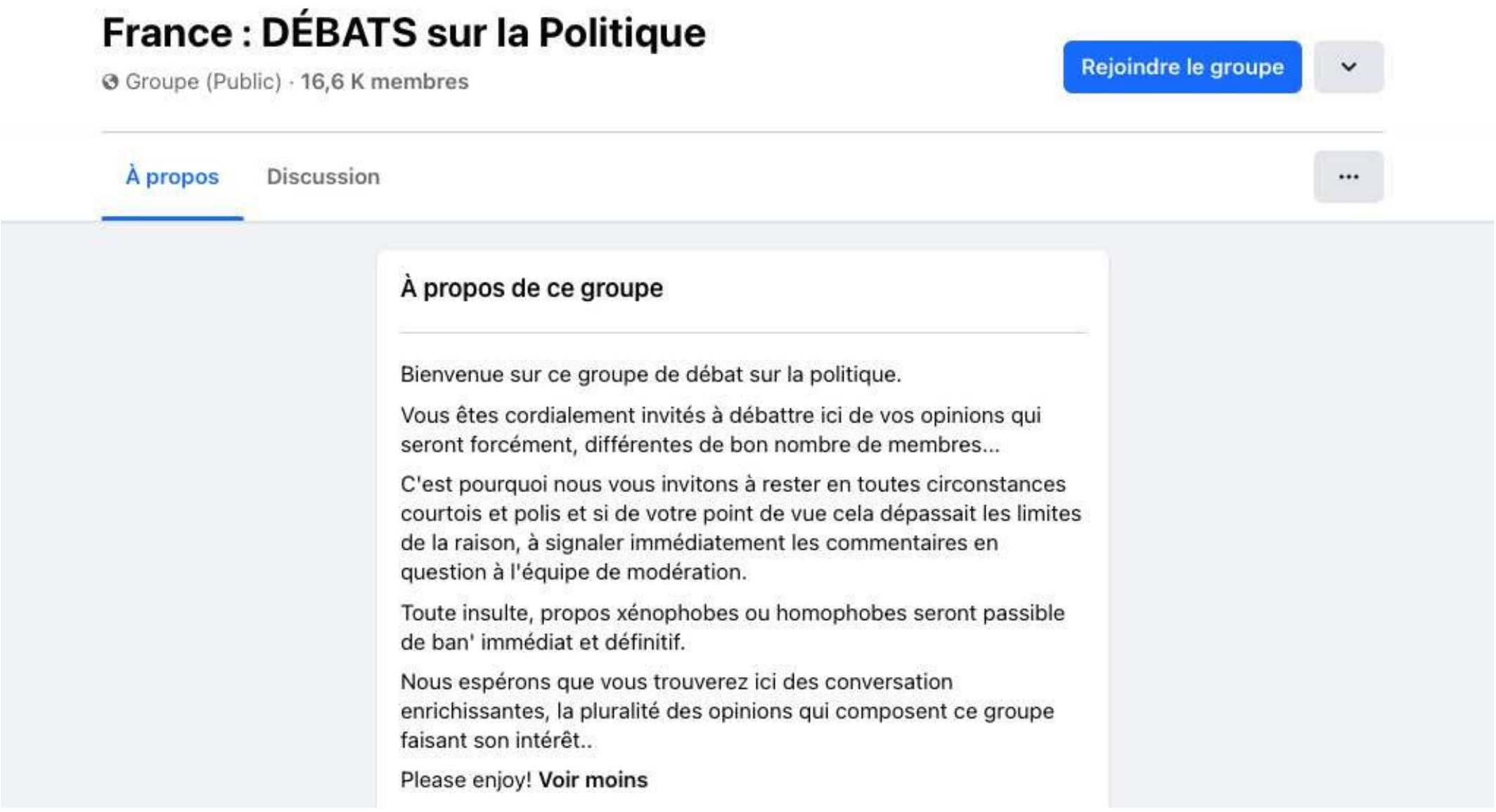

*Figure 5.11. Exemple de description d'un groupe classé comme « actualités diverses »*

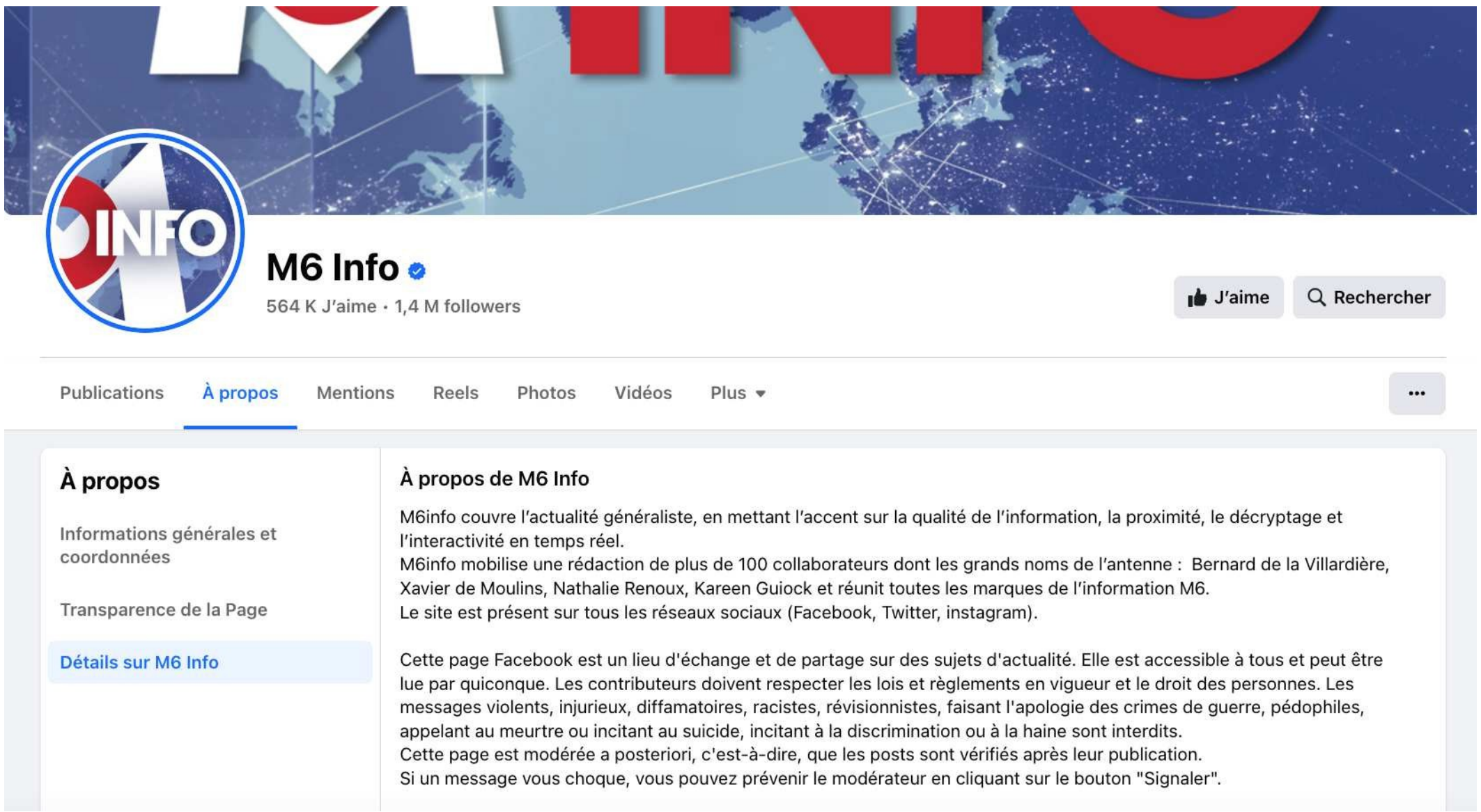

*Figure 5.12. Exemple de description d'une page de média mainstream*

Une troisième série de pages et de groupes peuvent être qualifiés de « spécialisés » dans la mesure où la plupart de leurs contenus sont centrés sur des sujets spécifiques (e.g. relatifs à la santé, aux animaux, à l'éducation, à la religion ou à une ville ou commune française). D'une façon générale, ces pages et ces groupes ne sont pas explicitement rattachés à un camp politique particulier et n'affichent pas d'orientation politique explicite. Le lien partagé entre les membres est fondé sur la géographie ou sur un centre d'intérêt commun, mais n'est pas politique *a priori*. Ces espaces n'en demeurent pas moins politisables dans la mesure où ils rassemblent des utilisateurs qui ont les mêmes centres d'intérêt ou habitent à proximité les uns des autres et peuvent ainsi être amenés à se regrouper autour de préoccupations ou de causes communes. Ces espaces se caractérisent donc probablement par une certaine homogénéité, si ce n'est un degré d'interconnaissance élevé dans le cas des pages et des groupes dédiés à la vie quotidienne au sein d'une localité française.

Enfin, un nombre important de pages et de groupes Facebook sont dédiés à l'humour ou au divertissement. Leurs noms comportent des termes comme « waouh », « buzz », « rire » et leurs descriptions incluent des messages tels que : « Si toi aussi tu as envie de rire, alors tu es sur la bonne page ! » ou encore : « Une page dans laquelle vous aurez la chance de dire souvent WOW! De l'humour, du talent, du choquant ». La plupart de ces pages et groupes attirent une audience importante, avec une médiane située entre 130 000 et 160 000 membres ou abonnés, et ont un rythme de publications soutenu, avec 10 à 15 posts par jour. Si l'atmosphère est globalement détendue et rieuse sur la plupart de ces espaces, des différences sont à noter entre ceux portés sur le divertissement et ceux consacrés à l'humour. Les premiers privilégient des contenus sensationnalistes, centrés sur les loisirs et la vie quotidienne, tandis que les seconds se concentrent davantage sur des contenus parodiques, souvent à caractère politique. En effet, 40,5 % des URLs signalées sur les pages et groupes humoristiques sont liées à la politique, contre seulement 9,9 % pour les espaces dédiés au divertissement. Ainsi, bien que ces deux types d'espace semblent partager une ambiance similaire, davantage propice à l'amusement qu'à la consultation d'informations d'actualité, certaines distinctions notables doivent être prises en compte pour étudier les réactions des utilisateurs face à leurs contenus.

Différents espaces de discussions potentiellement sous-tendus par différents contrats de communication ont ainsi été identifiés. Loin de former un tout homogène, les pages et les groupes publics de Facebook sont donc éclatés en différentes zones d'interlocution, plus ou moins publiques, politisées et hétérogènes. De façon intéressante, si l'on projette les catégories utilisées au cours du travail d'annotation sur le graphe présenté précédemment, une certaine cohérence ressort par rapport aux clusters identifiés par l'algorithme de Louvain (cf. Figure 5.13). Cette section illustre bien la complémentarité des méthodes numériques et qualitatives, notamment lorsque de nombreux allers-retours entre les deux sont réalisés. En l'occurrence, une analyse de réseau exploratoire a tout d'abord permis d'avoir une vue d'ensemble des relations de proximité entre les pages et les groupes Facebook du corpus. Ensuite, des observations ont permis d'affiner les interprétations, puis d'élaborer des grilles d'annotation afin de saisir leur atmosphère. Enfin, les annotations ont permis de réaliser quelques statistiques descriptives permettant de confirmer et d'objectiver les impressions issues des observations. En combinant ces différentes analyses, on a pu saisir les contrats de lecture et de conversation de différents espaces de communication.

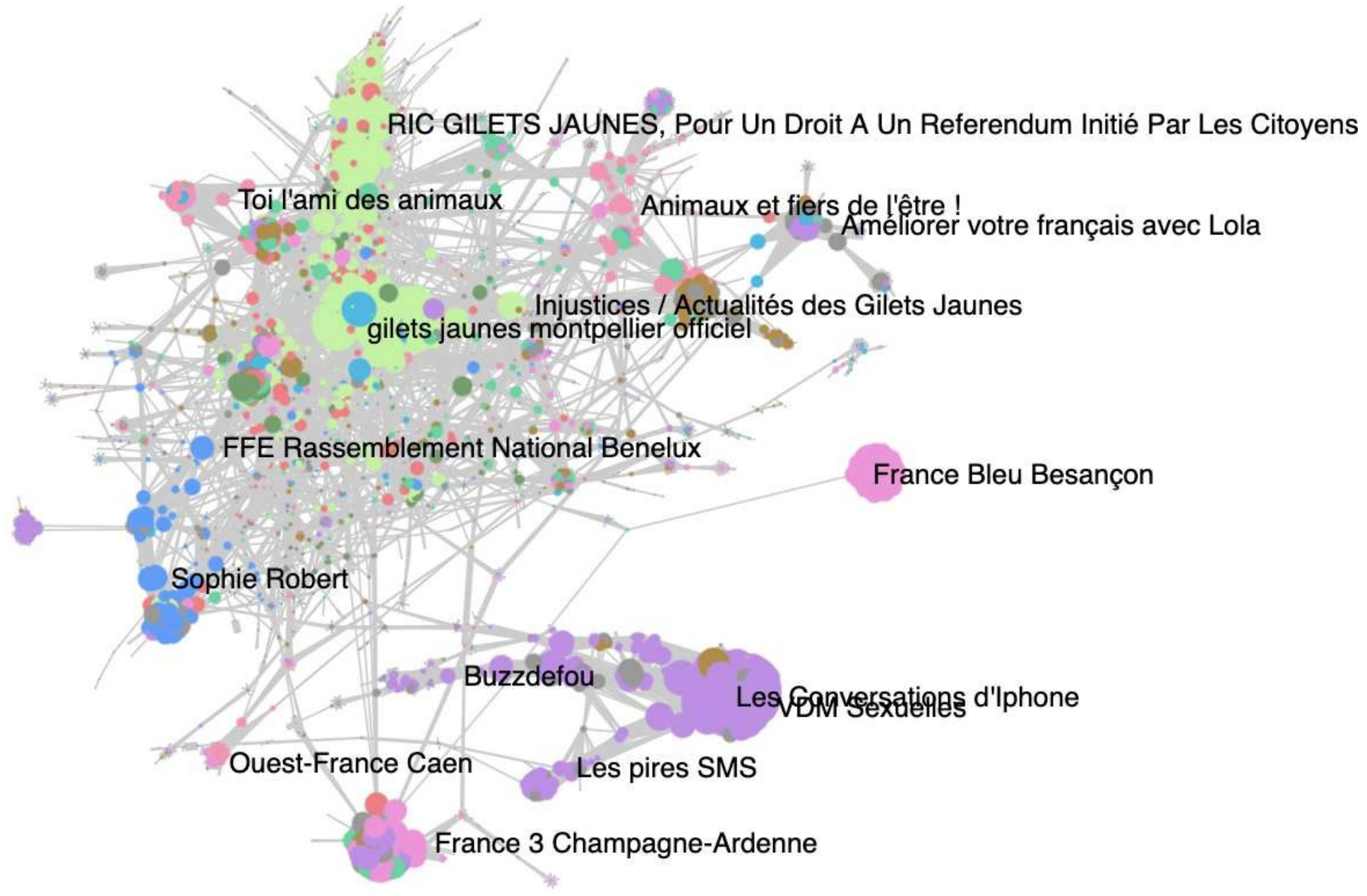

*Figure 5.13. Cartographie des pages et des groupes Facebook partageant des URLs signalées*

Dans une certaine mesure, ces différentes catégories d'espaces de communication peuvent être mises en perspective avec les divers types de collectifs auprès desquels Nina Eliasoph (1998) a mené une enquête ethnographique. En effet, le terrain de la chercheuse était composé de deux groupes de loisir, organisés autour d'un club de danse country western et d'une fraternité ; d'un ensemble de groupes de bénévoles, dont une association de lutte anti-drogue et une organisation de parents d'élèves ; et de deux groupes de militants, l'un contre des déchets toxiques et l'autre en faveur du désarmement. Contrairement à Nina Eliasoph, néanmoins, nous n'avons pas pu suivre les utilisateurs des réseaux sociaux dans différents contextes d'interactions.

À partir du fond de carte esquissé dans cette section, permettant de délimiter différents espaces de communication, la prochaine section vise à examiner si le taux d'expression de points d'arrêt et la façon dont ils sont énoncés varient selon les différents espaces de communication identifiés.

## 5.3. L’expression de points d’arrêt : une pratique située et socialisée

Le travail d’annotation réalisé a permis d’examiner comment l’expression de points d’arrêt varie selon cinq dimensions : (1) la source d’une URL ; (2) le thème d’un contenu ; (3) la taille de l’audience d’une page ou d’un groupe Facebook ; (4) son atmosphère ; (5) la présence ou l’absence d’une charte de modération. Ces différentes dimensions peuvent être appréhendées comme autant d’indices sur lesquels les utilisateurs des réseaux sociaux ont la possibilité de s’appuyer pour cerner les caractéristiques d’une situation d’énonciation en ligne.

Les statistiques descriptives (cf. Figure 5.14 ; 5.17 et 5.18) effectuées à partir de ces variables font ressortir cinq constats principaux : les utilisateurs de Facebook expriment davantage de points d’arrêt en réaction à des contenus qui (1) proviennent de sources bénéficiant d’une faible autorité dans l’espace médiatique français ; (2) portent sur des thématiques sensibles, fréquemment associées à des discours de haine dans le débat public ; sont publiés sur des pages et groupes Facebook qui (3) s’adressent à une audience importante et/ou diverse, (4) affichent un contrat de communication peu politisé et/ou pluraliste, mais (5) ne possèdent pas systématiquement de charte de modération.

Les observations et les entretiens ont permis d’affiner et d’approfondir ces constats : d’une part, en examinant les divers régimes d’énonciation et points d’appui mobilisés par les utilisateurs de Facebook pour exprimer des points d’arrêt ; de l’autre, en questionnant les mécanismes sociaux, au-delà des seuls indices d’énonciation mentionnés ci-dessus, susceptibles d'encourager ou de réfréner l’expression de points d’arrêt. Les différents résultats obtenus sont détaillés dans les sous-sections ci-dessous.

### 5.3.1. Attention à l’autorité des sources et à l’intérêt des sujets pour le débat public

Le premier résultat montre que les utilisateurs des réseaux sociaux ne sont pas indifférents à l’autorité des sources des contenus qui circulent sur Facebook, ni à l’intérêt de leur sujet et thématique pour le débat public. En effet, les URLs issues de sources de divertissement (14,34%), de contre-information (10,43%) ou de santé alternative (10,32%) déclenchent proportionnellement beaucoup plus de points d’arrêt que celles de médias *mainstream* (7,39%) ou de presse locale (3,33% ; cf. Figure 5.14). Ce classement est très proche de celui des *fact-checkers* vu au chapitre 2 (cf. Figure 2.1) et suggère qu’une partie des utilisateurs des réseaux sociaux adoptent des critères d’évaluation épistémique similaires à ceux des professionnels de l’information. D’ailleurs, alors que les sites *La Gauche m’a tuer* et *Santé + Mag* sont ceux qui ont publié le plus de contenus évalués comme des *fake news* par les *fact-checkers*, ils sont aussi parmi ceux qui ont suscité le plus de points d’arrêt sur Facebook. Cette concordance entre les jugements des *fact-checkers* et des utilisateurs exprimant des points d’arrêt peut s’expliquer par une intériorisation des normes de la profession journalistique par ces derniers. Celle-ci se remarque à travers les nombreux points d’arrêt faisant référence à des articles de *fact-checking* ou mobilisant des outils comme le *Décodex* du Monde ou le site *HoaxBuster* pour étayer leurs mises en doute. En simplifiant les processus de vérification des informations, ces ressources semblent avoir rendu la vérification de faits non seulement plus accessible, mais aussi plus routinière, et ainsi permis à un plus grand nombre de personnes de se familiariser avec les critères d'évaluation utilisés par les professionnels de l'information.

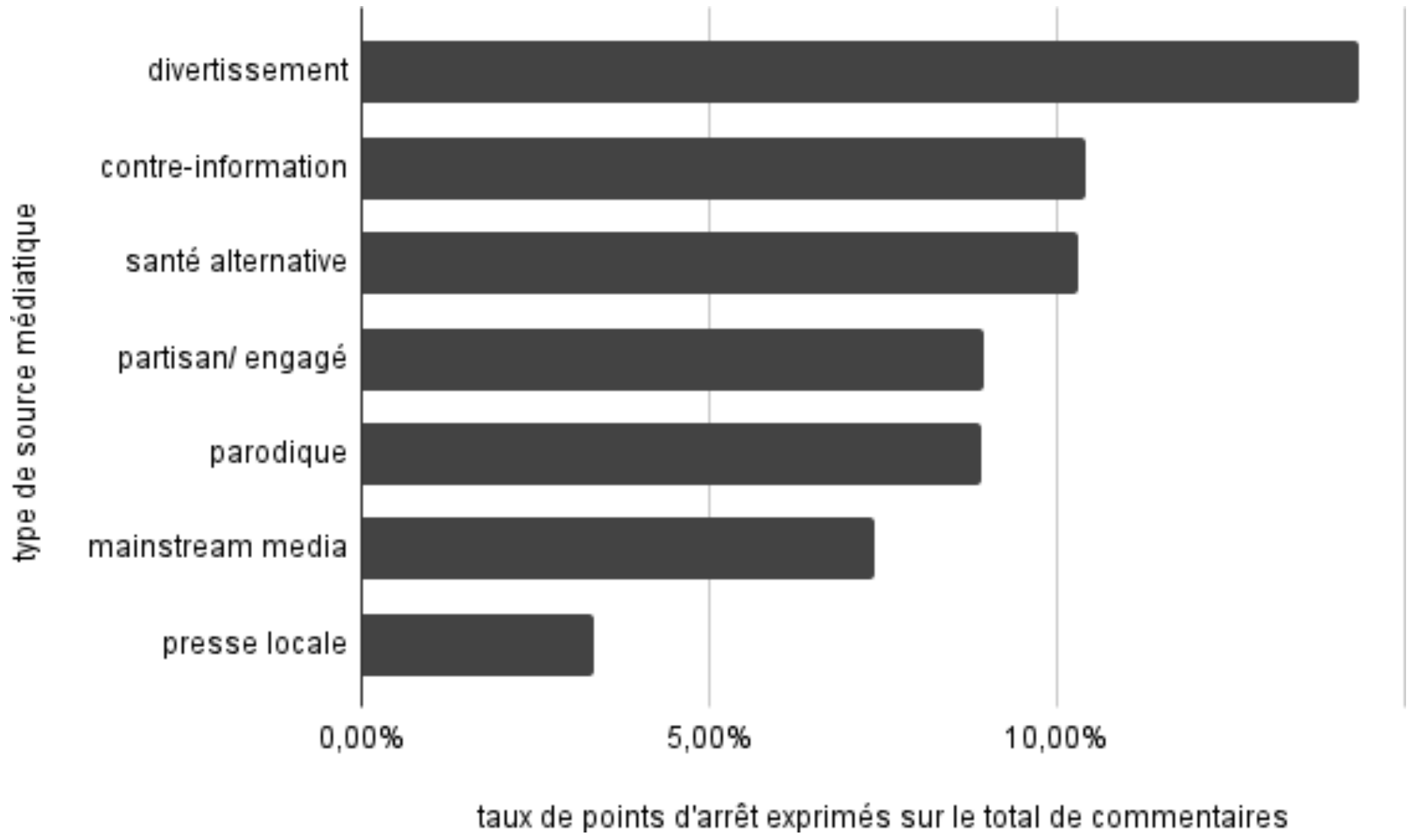

*Figure 5.14. Taux de points d'arrêt exprimés sur le total de commentaires reçus par chaque source médiatique*

Afin de comprendre dans quelles circonstances les utilisateurs des réseaux sociaux font particulièrement preuve de vigilance à l'égard des sources d'informations auxquelles ils sont exposés, il aurait été nécessaire de les suivre dans leurs usages numériques quotidiens en conduisant une enquête ethnographique. Loin d'être systématique, il est possible que l'attention portée par des individus à la fiabilité d'une source soit activée de façon plus ou moins importante par différents types d'épreuves – ne serait-ce qu'en les faisant sourciller ou plisser les yeux. Les entretiens réalisés n'ont pas permis de tenir compte de ces potentielles variations situationnelles. Néanmoins, ils ont fait ressortir des mécanismes plus structurels, soulignant que les individus ont plus tendance à faire preuve de circonspection à l'égard d'une source lorsque des pratiques de vérification et de sourçage leur ont été inculquées à travers leur éducation familiale ou scolaire. Un étudiant en sixième année de médecine[180] a par exemple expliqué avoir été sensibilisé par ses études à la notion de consensus scientifique et entraîné à la lecture critique d'articles scientifiques.

[180] Homme, 26 ans, Bac +6, interne en médecine, entretien réalisé le 18 janvier 2023.

*j'ai toujours vérifié s'il y avait plusieurs sources qui parlaient de la même information, vérifié le site d'où venait l'information et du coup d'autres informations qui venaient du même site, voire si elles étaient fiables, si elles avaient été contredites ailleurs, si typiquement c'était un site parodique. Et euh aller me renseigner, quand ça touchait à la science, je pouvais me référer au consensus scientifique, aller rechercher des études brutes et voir ce qu'elles en disaient surtout si elles citaient des études, j'avais l'avantage d'avoir été formé à lire des études scientifiques.*

Un autre enquêté[181], ayant passé le début de son enfance en Algérie, puis grandi dans un quartier populaire d'une ville de Province, a expliqué avoir été « sectorisé dans un des quartiers les plus riches de [s]a ville [et] été confronté à des camarades de classe qui avaient une éducation différente de la [s]ienne ». À plusieurs reprises, celui-ci a mentionné comment ce contexte socio-culturel l'avait rendu particulièrement attentif à vérifier la fiabilité des sources.

*J'ai toujours été attentif parce que j'ai grandi dans un environnement social, dans un pays… Je n'ai pas grandi en France. J'ai grandi dans un autre pays où la désinformation est quelque chose de banal, normal. Donc j'avais la chance depuis que j'étais jeune, mes parents m'ont appris que tout ce que j'écoute, tout ce que j'apprends, tout ce que j'entends, tout ce que je vois, n'est pas vraiment vrai et qu'il faut toujours vérifier.*

*Dans mon contexte, les études m'ont formaté pour ne pas croire tout ce que je lis partout. [...] On est la génération où quand on faisait une dissertation, si on ne mettait pas les sources, même si on avait raison, on avait zéro. [...] Dès lors qu'on a un certain niveau d'études, même si ce n'est pas énorme, mais dès lors qu'on a passé notre bac, qu'on l'ait eu ou pas, on ne peut pas ne pas confronter les sources. Ce n'est pas possible. Si on ne le fait pas, c'est qu'on est malhonnête intellectuellement.*

Si les utilisateurs qui expriment des points d'arrêt semblent surtout faire montre de vigilance à l'égard de sources qui ne bénéficient pas d'une forte autorité dans l'espace médiatique français, cela ne les empêche pas d'exprimer ponctuellement des critiques à l'égard des médias *mainstream* et de leurs journalistes. De façon intéressante, la base de données Condor de Facebook décrite au chapitre 2 a permis d'identifier des URLs ayant été évaluées comme « vraies » par des *fact-checkers* mais signalées comme « fausses » par des utilisateurs de Facebook, à l’instar du contenu ci-dessous publié par le média *20 Minutes* (cf. Figure 5.15). De nombreux utilisateurs sont intervenus pour contester cette publication, notamment en expliquant qu’Erdogan avait effectivement prononcé la phrase citée, mais que son sens avait été altéré car elle avait été « mal traduite », « mal interprétée », « déformée », « remixée » ou « sortie de son contexte ». Au total, 165 points d’arrêt ont été dénombrés sur les 1 400 commentaires déclenchés par le post (e.g. « 20 minutes je vous conseil fortement à relire et à relire sérieux vous traduisez très mal, et bien évidemment cela est fait pour salir la Turquie donc next !!! » ; « Quand on sort des propos de son contexte pour attirer la haine… » ; « C'est de la diabolisation contre Erdogan, l'interprétation est eronnée »). Ces commentaires montrent que les utilisateurs ne se contentent pas de vérifier la factualité brute des informations mais se préoccupent également de l’interprétation, du contexte, et de la manière dont les faits sont encadrés et présentés.

[181] Homme, 30 ans, Bac +5, entrepreneur, entretien réalisé le 26 janvier 2023.

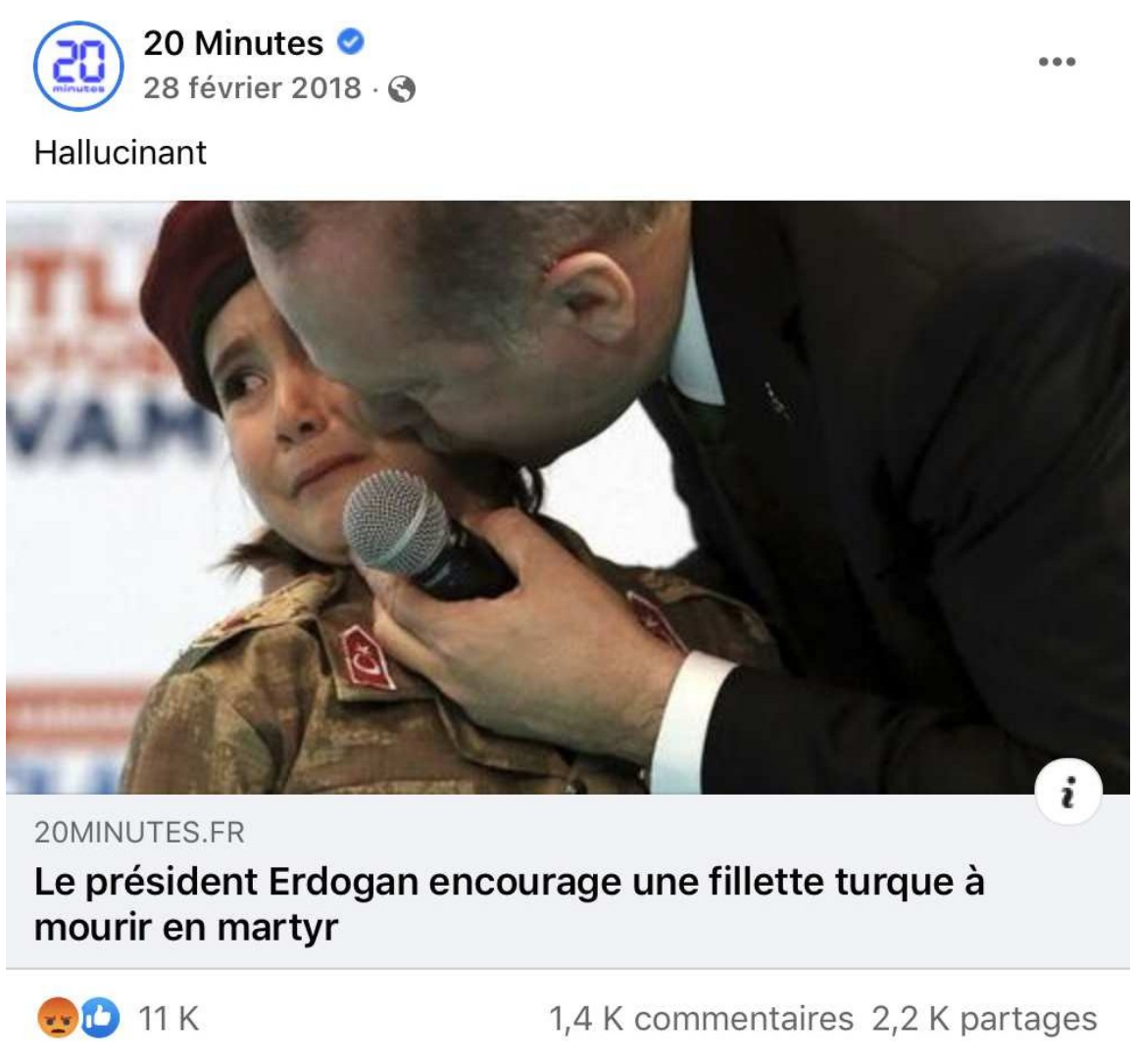

*Figure 5.15. Premier exemple d’un contenu évalué comme « vrai » par des fact-checker mais ayant fait l’objet de signalements de la part d’utilisateurs sur Facebook*

Dans une certaine mesure, la factualité d’un énoncé est parfois considérée comme un ersatz de la réalité par certains utilisateurs des réseaux sociaux. Ils vont alors non seulement critiquer les médias qui cherchent à véhiculer des représentations qui leur semblent déformées du réel — par exemple en instrumentalisant des faits avérés qui ne sont plus d’actualité — mais aussi chercher à compléter le travail des *fact-checkers* en se positionnant comme des « débunkers » comme l'explique cet enquêté, intervenu pour contester un post diffusée sur la page La gauche m'a tuer (cf. Figure 5.16).

> *Alors cet article là où je suis intervenu, on peut comparer mes commentaires à du débunkage. Il s'agissait ici d'un article sur un événement datant d'il y a plus de deux ans, où La gauche m'a tuer cherche à attiser la haine en sous-entendant que ça vient juste de se passer. Le pire est qu'il leur arrive régulièrement de reposter tous les X mois le même article comme ça pour pouvoir attiser plusieurs fois la haine sur le même événement. Mes interventions ont surtout consisté à modérer les gens qui s'horrifiaient de ce « nouvel acte de violence » et bien leur expliquer que c'est une vieille affaire.*[182]

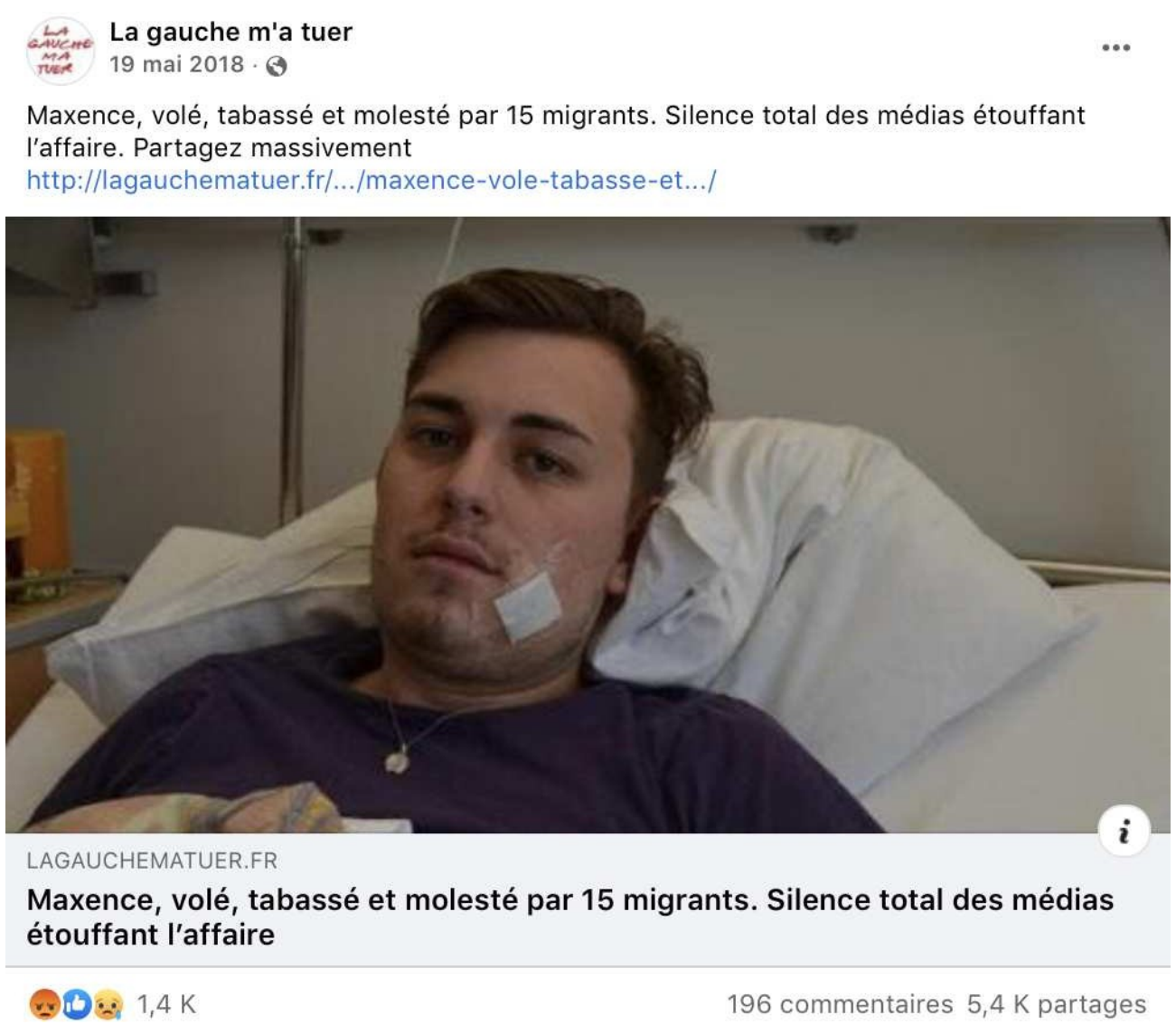

*Figure 5.16. Deuxième exemple d'un contenu évalué comme « vrai » par des fact-checker mais ayant fait l'objet de signalements de la part d'utilisateurs sur Facebook*

Ces exemples ne signifient pas que la majorité des *fact-checkers* sont indifférents aux dynamiques d'instrumentalisation et de mise à l'agenda. Au contraire, plusieurs d'entre eux les mentionnent explicitement dans leurs articles. Cependant, ils révèlent comment les termes *fake news* et *fact-checks* tendent à structurer le débat de manière manichéenne, en réduisant la question des *fake news* à un simple problème de factualité. Les points d'arrêt des utilisateurs permettent ainsi de souligner que la seule factualité des contenus ne suffit pas à garantir la qualité du débat public et que d'autres dimensions sont importantes à prendre en considération, telles que la pertinence contextuelle ou le cadrage interprétatif des contenus. En témoigne, d'ailleurs, la vigilance accrue des utilisateurs à l'égard de certains sujets (cf. Figure 5.17). Par exemple, les contenus portant sur des thématiques sensibles comme la religion (17,3%), l'inclusion sociale (14,8%) ou l'immigration (12,4%) suscitent plus de points d'arrêt que ceux concernant des sujets d'actualité liés à la politique (9,4%), l'économie (9,2%) ou l'environnement (8,9%), ou traitant de faits divers (8,5%) ou de sujets de divertissement (8,3%). Ces contenus peuvent être perçus comme davantage susceptibles de renforcer des préjugés ou des stéréotypes et d'exacerber les divisions sociales que ceux portant sur d'autres sujets dans la mesure où ils concernent souvent des valeurs et croyances personnelles ou des questions d'identité.

[182] Homme, 34 ans, Bac, accompagnant éducatif et social, entretien réalisé le 1er août 2022.

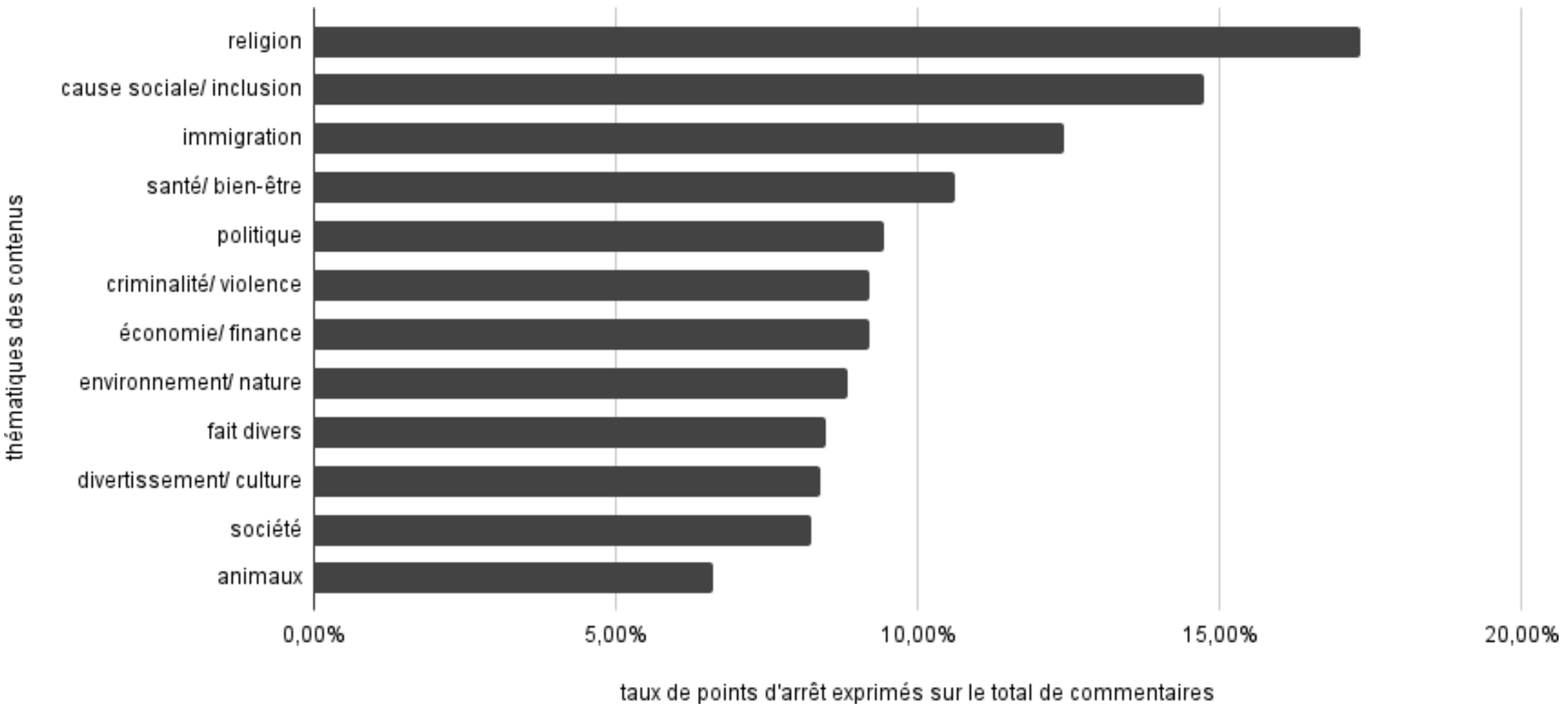

*Figure 5.17. Taux de points d'arrêt exprimés sur le total de commentaires reçus par chaque type de contenu selon sa thématique*

### 5.3.2. Des opportunités discursives différenciées selon le degré de pluralisme et de politisation des espaces de communication

Le deuxième résultat indique que plus les espaces de communication s'adressent à des audiences larges et/ou hétérogènes, plus ils sont susceptibles de favoriser l'expression de points d'arrêt. À l'inverse, plus ils sont centrés sur une idéologie ou un parti politique spécifique, plus les taux de points d'arrêt y sont bas. En effet, ce sont les pages et les groupes Facebook qualifiés d'actualités diverses qui ont suscité les taux de points d'arrêt les plus élevés (18,35%). Par ailleurs, les taux de points d'arrêt énoncés sur des pages et groupes associés au mouvement des gilets jaunes (16,0%) ou à des opinions anti-systèmes (12,2%) sont entre 1,5 et 2,3 plus élevés que ceux émis sur des pages de droite (8,2%) et d'extrême droite (7,0%). Ces divergences sont cependant moins marquées avec les pages et groupes de gauche et d'extrême gauche dont les taux de points d'arrêt s'élèvent respectivement à 11,6% et 11,8%. Il faut également souligner que peu de points d'arrêt ont été exprimés sur les pages et groupes de *média mainstream* (6,8%) et situés au centre de l'échiquier politique (7,8%). Enfin, les pages et groupes spécialisés ont globalement obtenu des taux de points d'arrêt légèrement en dessous de la moyenne, avec cependant un écart important entre les espaces dédiés à des sujets de santé (11,99%) et ceux visant à regrouper des personnes habitant dans une même ville ou commune (5,29%).

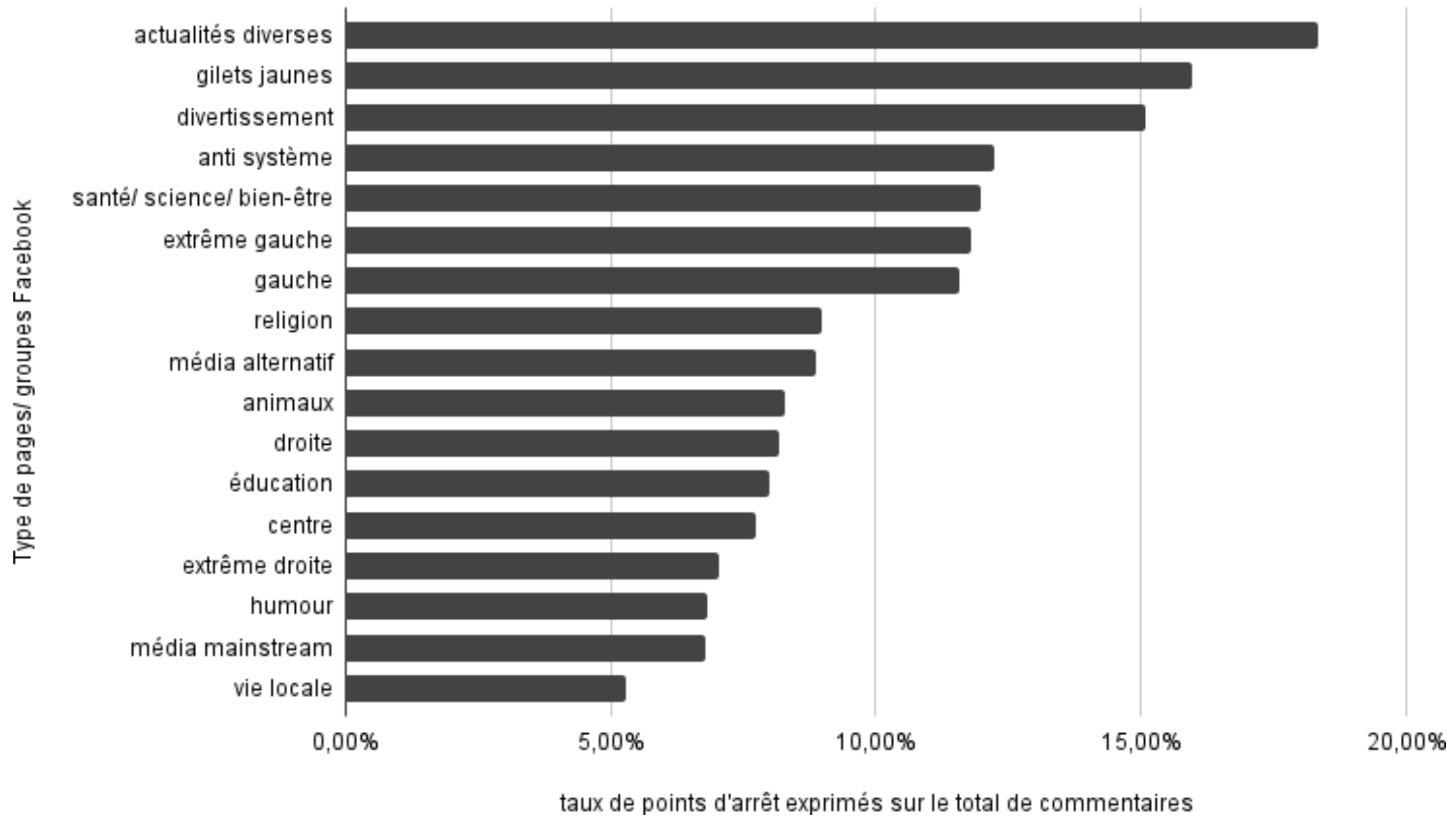

*Figure 5.18. Taux de points d'arrêt exprimés sur le total de commentaires reçus par chaque type de page/groupe Facebook*

Différentes analyses et études de cas permettent d'illustrer ces variations et d'expliquer pourquoi la propension des utilisateurs à exprimer des points d'arrêt varie selon le degré de pluralisme et de politisation des pages et des groupes Facebook.

Prenons, dans un premier temps, l'écart entre les pages et groupes d'actualités diverses et de médias *mainstream*. Alors que les deux types d'espace affichent globalement un contrat de lecture et de conversation pluraliste, le taux de points d'arrêt des premiers est trois fois plus important que celui des seconds. Plusieurs caractéristiques propres à chacun de ces espaces permettent d'expliquer cet écart. Tout d'abord, les pages et groupes d'actualités diverses sont spécifiquement orientés vers la confrontation d'opinions. Par exemple, il est indiqué dans l'onglet à propos du groupe France : DÉBATS sur la Politique : « Bienvenue sur ce groupe de débat sur la politique. Vous êtes cordialement invité ici à débattre de vos opinions qui seront forcément différentes de bon nombre de membres ». Comme lors de débats télévisés en période d'élections électorales, l'expression de désaccords semble ainsi constituer une forme de réaction recherchée ou attendue (Kerbrat-Orecchioni, 2016). En revanche, les pages Facebook de médias *mainstream* cherchent avant tout à être des « espace[s] pertinent[s] et agréable[s] de discussion sur l'actualité » (comme cela est indiqué dans l'onglet à propos de la page du *Monde*) ou des « lieu[x] d'échange et de partage sur des sujets d'actualité [...] accessible[s] à tous et [pouvant] être lu[s] par quiconque » (comme cela est indiqué dans l'onglet à propos de la page de *M6 Info*). Elles visent à garantir un pluralisme d'opinion mais semblent davantage encourager les commentaires « polis et courtois » que l'expression de désaccords. Par ailleurs, les pages et groupes des médias *mainstream* publient quasiment exclusivement (96,2%) des contenus issus de leurs propres médias – soit de sources bénéficiant d'une certaine autorité dans l'espace médiatique français – alors que les pages et groupes d'actualités diverses partagent une plus grande variété de contenus, provenant de sources moins établies, ce qui incite probablement les utilisateurs à questionner plus fréquemment ces informations. Enfin, bien que les pages et groupes des médias *mainstream* aient un nombre d'abonnés 27 fois plus grand que les pages et groupes d'actualités diverses, cela ne joue pas en leur faveur. Alors que leurs publications d'URLs signalées ont suscité en moyenne 147 commentaires, celles des pages et groupes d'actualités diverses en ont généré seulement 6 en moyenne. Dans cette masse de commentaires, les points d'arrêt ne peuvent donc qu'être dilués et peu visibles sur les pages et groupes des médias *mainstream*. Cela ne signifie pas pour autant que les discussions y sont de moins bonne qualité que sur les pages et groupes d'actualités diverses, mais suggère plutôt que la nature des interactions diffère. D'autres indicateurs que les taux de points d'arrêt, tels que la longueur des commentaires ou la propension à citer des sources, sont importants à prendre en compte pour examiner la nature et la teneur des échanges.

Voyons à présent comment et pourquoi la propension des utilisateurs de Facebook à exprimer des points d'arrêt varie selon que les espaces de discussion cherchent à rassembler des personnes de sensibilités politiques variées ou homogènes. Pour cela, les réactions émises face au contenu ci-dessous, issu du site *LesObservateurs.ch*, constituent un bon cas d'étude (cf. Figure 5.19). En effet, ce contenu a été partagé par 18 pages et groupes Facebook différents affichant soit un positionnement politique précis (notamment d'extrême droite), soit des opinions anti-système regroupant différentes sensibilités partisanes (notamment sur les groupes des gilets jaunes). Ce cas permet ainsi de comparer comment les réactions des utilisateurs varient d'un espace de communication à l'autre, indépendamment du contenu qui y est partagé.

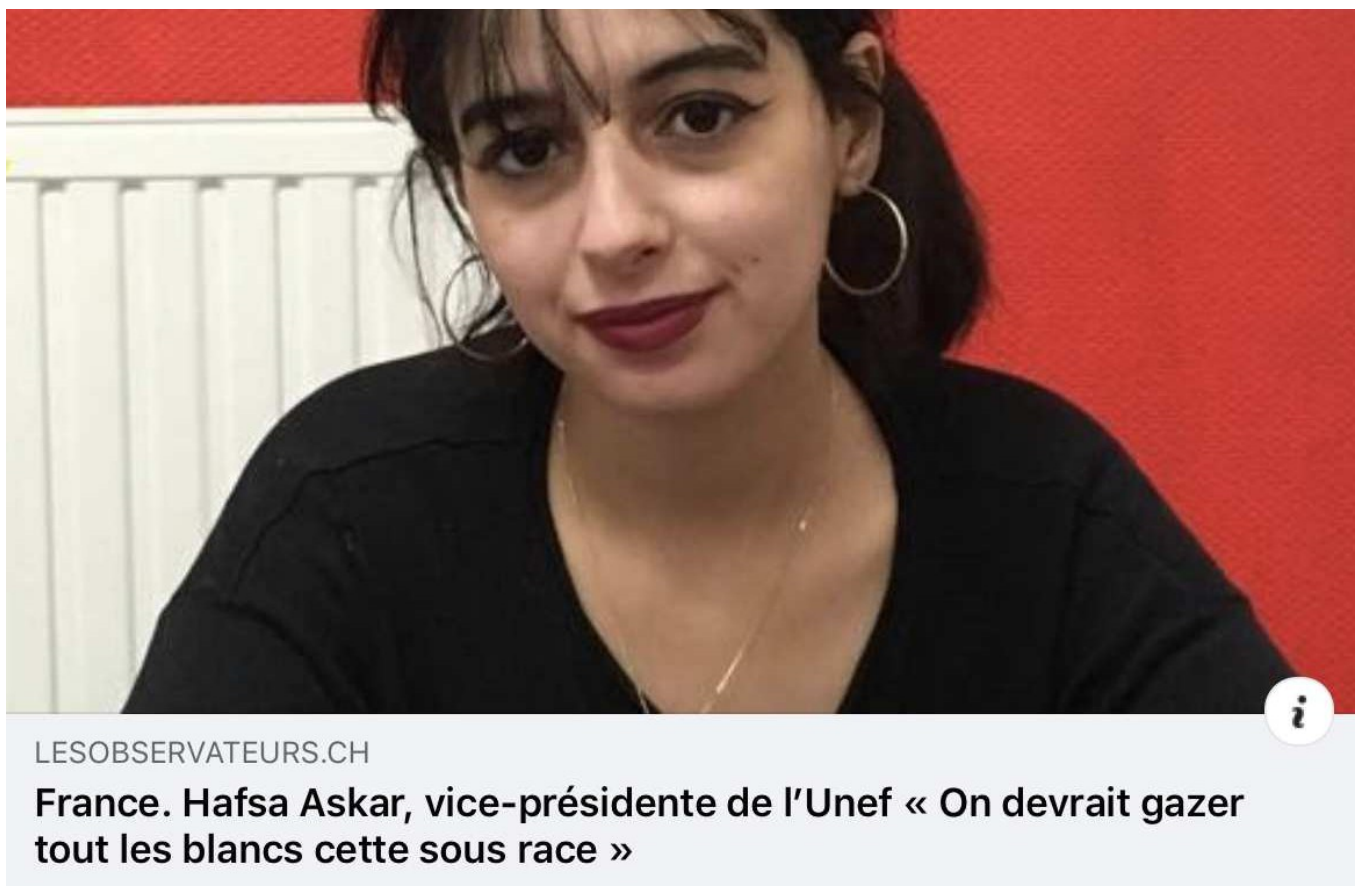

*Figure 5.19. Exemple de contenu partagé par des pages et groupes politisés s'adressant à des audiences plus ou moins hétérogènes*

C'est sur les pages et groupes « Gilets Jaunes le Mouvement » (27 000 abonnés) et « Révolte française » (22 000 abonnés) que les taux de points d'arrêt ont été les plus élevés. Par exemple, sur les 122 commentaires suscités par le post de la page « Gilets Jaunes le Mouvement », environ 30 points d'arrêt ont été relevés, soit 30% du total des commentaires. Sur ces espaces, les points d'arrêt comportent des marqueurs de colère (e.g. écriture en majuscule, présence d'émoji avec un visage rouge) et sont rédigés sur un ton réprobateur. La majorité de ces points d'arrêt s'adresse à l'auteur du post, ou plus largement au mouvement qu'il représente, et dénonce son intention idéologique, perçue comme manipulatoire, en l'accusant de « jouer le jeu des fachos Racistes Nationalistes [...] à deux jours des élections européennes ». Ces réactions montrent comment un contrôle mutuel peut être exercé au sein d'espaces de communication très politisés mais partiellement pluralistes. Plutôt que de correspondre à des sphères publiques habermassiennes, ceux-ci s'apparentent davantage à des arènes de luttes agonistiques au sein desquelles les désaccords sont énoncés de façon conflictuelle et polémique. Le chapitre 6 approfondira cette analyse en examinant les réponses reçues par les points d'arrêt au sein de ces espaces afin de questionner dans quelle mesure ils permettent réellement l'expression d'un pluralisme agonistique.

A l'inverse, sur des pages et groupes explicitement d'extrême droite, qui bénéficient d'une audience similaire, voire plus large, les taux de points d'arrêt sont nettement plus faibles. Par exemple, sur la page « Je vote Rassemblement national » (35 983 abonnés), seulement une dizaine de points d'arrêt ont été observés sur un total de 866 commentaires. Par ailleurs, ces points d'arrêt sont exprimés avec une moins grande assertivité que sur les espaces de communication des gilets jaunes ou anti-système. Les utilisateurs emploient des tournures interrogatives ou des adverbes de doute : « c est bien vrai ? » ; « des preuves ? » ; « fake ?? ». Quelques réactions s'apparentent également à des formes d'alignement incrédule : « On a du mal à croire ça ?? Il faut se méfier des réseaux sociaux ??? Si c'est vraiment ce que cette merde a dit !! Alors qu'elle aille se faire foutre !!! ». Cette plus grande réserve dans l'expression de critiques peut s'expliquer par des dynamiques de spirale du silence (Noelle-Neumann, 1974). Par crainte d'énoncer une opinion minoritaire, dans un environnement où règne une forte cohésion idéologique, les utilisateurs préfèrent sans doute ne pas prendre le risque de faire connaître leurs désaccords, ou du moins éviter de s'engager trop fermement dans une opposition en laissant ouverte la possibilité de se rétracter si nécessaire.

Il est également intéressant de noter que sur des pages soutenant des opinions d'extrême droite mais pas officiellement affiliées à un parti, comme « Johnny Front Des Patriotes Français » (48 395 abonnés), les points d'arrêt sont encore plus rares (seulement 1 sur 287 commentaires). Par ailleurs, pratiquement aucun point d'arrêt n'a été relevé sur les pages et groupes avec un faible nombre d'abonnés ou de membres. Les utilisateurs se montrent aussi plus désinvoltes et irrévérencieux dans leurs expressions, sans doute parce que ces espaces sont moins modérés et contrôlés et offrent plus de latitude pour des opinions radicales que ceux officiellement tenus par des responsables d'un parti. Une observation qui va dans le même sens que les résultats du chapitre précédent : de la même manière que les utilisateurs de Twitter sont plus susceptibles de faire preuve de prudence énonciative lorsqu'ils sont intégrés à des groupes partisans officiels, les utilisateurs de Facebook sont plus enclins à exprimer des points d'arrêt sur des pages et groupes politique officiels, probablement contrôlés par des cadres du parti, que sur des espaces militants mais non affiliés. Dans le même temps, tout comme le partage de *fake news* est associé à un degré de politisation individuel plus élevé, l'expression de point d'arrêt est limitée par un degré de politisation important des espaces de communication.

Si les pages et groupes Facebook qui visent à rassembler des personnes partageant les mêmes sensibilités politiques sont moins propices à l'expression de points d'arrêt que ceux affichant un contrat de communication pluraliste, une question se pose : pourquoi les pratiques consistant à exprimer des points d'arrêt sont plus élevées sur les pages et groupes de gauche que ceux de droite ? Deux hypothèses principales peuvent être avancées. La première est que les opposants aux opinions de gauche sont plus enclins à se mobiliser pour émettre des critiques sur les réseaux sociaux que ceux opposés à des opinions de droite. La seconde est que les divisions internes sont plus prononcées au sein de la gauche. Bien qu'il n'ait pas été possible de déterminer la position politique exacte de chaque membre ou abonné d'une page ou d'un groupe Facebook, des interventions émanant de personnes se réclamant de la gauche ont été relevées, critiquant leur propre camp, notamment sur des sujets liés à la vaccination. Par exemple, voici quelques points d'arrêt émis en réaction à cet énoncé : « Encore un bébé de deux mois qui décède 48H après avoir reçu 8 vaccins ».

> *Mais arrêtez de partager ces fausses informations bordel. On est sur le groupe d'un mouvement de gauche ou de fachos ?*

> *Au-delà du fait qu'il est inadmissible de faire de la pub à un site d'extrême droite, même "l'info" en elle même est pourrie. Elle a pour seul but de créer une défiance envers la vaccination. Si cet* *enfant existe et qu'il est malheureusement décédé, rien ne prouve qu'il y ait un lien avec le vaccin. Où est-il question dans "l'avenir en commun" de promouvoir des croyances obscurantistes contre les progrès de la médecine ? Il faudrait quand même que les gens qui administrent les pages LFI soient un peu plus conséquents.*

Ces points d'arrêt font écho au témoignage d'une militante LFI exposé au chapitre 4. Celle-ci avait en effet indiqué s'être fait reprocher par ses amis de gauche de « relayer la propagande d'extrême droite ».

Analysons, enfin, pourquoi les taux de points d'arrêt des pages et groupes qui publient des contenus centrés sur des thématiques spécifiques sont globalement en dessous de la moyenne, mais au-dessus de ceux des espaces de communication très politisés de droite et d'extrême droite, et pourquoi il existe un écart important entre les pages et groupes traitant de sujets de santé et ceux qui concernent la vie quotidienne au sein d'une localité. Il est difficile de comparer et d'interpréter de façon rigoureuse ces divergences entre différents espaces de communication, étant donné qu'ils ne publient pas du tout les mêmes types de contenus. Cela rend en effet impossible une analyse similaire à celle de l'étude de cas sur le contenu issu du site *LesObservateurs.ch*, où il a été possible d'examiner si des utilisateurs réagissent différemment à un même contenu selon l'espace de communication dans lequel il a été diffusé. Toutefois, les observations effectuées sur chaque type de page et de groupe spécialisé ont révélé des dynamiques conversationnelles distinctes. Par exemple, l'écart notable entre les pages traitant de sujets de santé et celles concernant la vie quotidienne au sein d'une localité peut être éclairé par la nature des discussions dans ces espaces. Les sujets de santé, surtout lorsqu'ils touchent à des questions controversées comme la vaccination ou les traitements alternatifs, tendent à polariser davantage les opinions et à inciter à des interventions critiques. Cependant, comme cela sera montré dans la section 5.3.4., ces interventions semblent moins découler du degré de pluralisme offert *per se* par les pages et groupes de santé que d'une coordination externe entre divers collectifs visant à défendre la science, la rationalité et l'esprit critique.

En ce qui les concerne, les discussions se déroulant au sein de pages et de groupes dédiés à une localité française sont souvent axées sur des préoccupations communes et quotidiennes, ce qui tend à favoriser la recherche de consensus entre les individus plutôt que l'expression de désaccords. De plus, dans ces espaces, l'expression de points d'arrêt peut être perçue comme une pratique sociale plus risquée en raison du fort niveau d'interconnaissance et de la proximité physique entre les abonnés d'une page ou les membres d'un groupe. Les utilisateurs peuvent en effet craindre que leurs commentaires en ligne aient des répercussions sur leurs relations dans la vie réelle, les incitant ainsi à éviter les confrontations publiques et à adopter un ton plus conciliant.

En définitive la principale conclusion qui ressort à l'issue de cette sous-section est que les espaces de communication en ligne offrent des opportunités d'expression de points d'arrêt différenciés selon leur degré de pluralisme et de politisation. Les pages et groupes d'actualités diverses se distinguent particulièrement par leur capacité à générer des discussions entre utilisateurs aux opinions variées et s'apparentent ainsi, dans une certaine mesure, à des « tiers espaces » (Wright, 2012a ; 2012b). Ce concept désigne des espaces conversationnels non politisés *a priori*, au sein desquels se déroulent des conversations informelles entre citoyens ordinaires sur les réseaux sociaux, portant sur l'actualité politique (Graham et al., 2016). Selon Scott Wright (2016), ces « tiers espaces » sont davantage susceptibles d'accueillir des conversations contradictoires, réunissant des internautes aux opinions diverses, échangeant de manière rationnelle ou apaisée, tout en constituant des vecteurs de politisation pour des individus peu intéressés par l'actualité politique. À l'inverse, les espaces plus politisés tendent à renforcer les divisions idéologiques et à limiter les échanges contradictoires, contribuant ainsi à la fragmentation du discours public en ligne (González-Bailón et al., 2023).

### 5.3.3. Entre humour et divertissement : exprimer des points d'arrêt sans casser l'ambiance

Le troisième résultat confirme que les espaces de communication *a priori* peu politisés, s'adressant à une audience large et diversifiée, comme ceux dédiés au divertissement, ont tendance à susciter davantage de points d'arrêt que les pages et groupes destinés à une audience très politisée et homogène. Cela étant, bien que les groupes et pages humoristiques attirent également un public important et ne soient pas non plus ouvertement politisés, leur taux de points d'arrêt moyen (6,8 %) est deux fois moins élevé que celui des espaces de divertissement (15,1 %). Cet écart s'explique par des variations dans les contrats de communication propres à ces deux types d'espaces, ainsi que par la capacité des utilisateurs à déceler et à comprendre les attentes implicites qui les sous-tendent.

Comme cela a été décrit dans la section 5.2.4., les pages et groupes de divertissement se présentent comme des espaces de discussions orientés vers des sujets variés, tels que la mode, le sport, le cinéma, la musique, les séries ou les jeux vidéo. Bien que leur objectif apparent soit d'informer les utilisateurs sur les dernières tendances, de leur révéler des scoops sur des célébrités ou de leur proposer des bons plans pour leur vie sociale et quotidienne, plusieurs indices montrent clairement que leur but principal est de générer un maximum d'engagement en partageant des contenus surprenants ou choquants. L'intention commerciale derrière ces contenus n'est toutefois pas explicitement mentionnée, pas plus que leur caractère parfois douteux ou fallacieux. Il est possible que les utilisateurs acceptent ce contrat de communication ambigu, mais seulement jusqu'à un certain point : tant que les contenus demeurent divertissants ou présentent une utilité pratique. Toutefois, lorsque ces pages et groupes s'écartent trop de leur intention affichée en publiant des contenus hors-sujet ou insensés, il est compréhensible que certains utilisateurs réagissent vivement en exprimant des points d'arrêt. Ceux-ci sont alors majoritairement dirigés contre le locuteur pour condamner sa responsabilité énonciative. Par exemple, alors que la page « Astuce de grand-mère » se présente comme « un site pour avoir des astuces d'antan », un utilisateur intervient pour dire : « la page "astuces de grand mere" n a rien à voir avec son nom! Cette page est malsaine et calomnieuse et irrespectueuse! Moi je la supprimé tellement que ces sujets sont immoraux ! ». Ce type de réaction illustre comment les utilisateurs, lorsqu'ils estiment que les contenus ne correspondent plus à l'intention affichée, n'hésitent pas à condamner ouvertement la page en question, marquant ainsi leur désapprobation et leur méfiance face à une intention perçue comme trompeuse.

*A contrario*, sur les pages et groupes d'humour, l'intention de partager des contenus invraisemblables à des fins humoristiques est explicitement indiquée. Dans ces espaces, les utilisateurs sont invités à découvrir le décalage entre ce qui est dit explicitement par l'énonciateur et ce qui est sous-entendu par le locuteur (Charaudeau, 2006 ; 2013 ; Fernandez et Vivero García, 2006). Ainsi, à moins de ne pas avoir saisi cette intention humoristique, les utilisateurs ne peuvent pas accuser le média ou l'auteur du post de chercher à les tromper ou de ne pas respecter leur contrat de communication. Dans l'ensemble, les utilisateurs semblent comprendre sans difficulté cette intention humoristique. En effet, sur les pages et les groupes d'humour, les contenus font essentiellement l'objet de bavardages familiers entre internautes. Plusieurs d'entre eux taguent un de leur contact qui souvent leur répond par un commentaire rédigé dans un registre familier. La majorité des commentaires sont écrits en argots Internet et utilisent de nombreux émojis et acronymes comme « Mdr » ou « Ptdr » en multipliant les « r » comme pour faire entendre des éclats de rire. Les fils de discussion s'ouvrent et se referment comme ils ont commencé : sur une remarque facétieuse. L'atmosphère rieuse des pages et des groupes d'humour n'empêche pas toutefois quelques utilisateurs d'exprimer des points d'arrêt, le plus souvent non pas par manque de compréhension de l'intention parodique, mais parce qu'ils estiment que le contenu déroge au contrat de communication humoristique comme en témoigne le commentaire suivant : « Ouille... Pas très convaincue du caractère comique de cette publication… ». Néanmoins, ces critiques représentent une très faible proportion par rapport à la flopée de commentaires blagueurs émis par les autres utilisateurs. Le nombre moyen de commentaires par post sur les pages et groupes d'humour est en effet de 133, alors qu'il n'est que de 21 sur les espaces de divertissement. Les points d'arrêt peuvent alors apparaître comme dissonants par rapport aux autres commentaires si bien que l'on peut se demander s'ils ne sont pas perçus comme des fausses notes par les autres utilisateurs – une question que nous explorerons dans le chapitre 6.

Une autre raison pour laquelle les taux de points d'arrêt sont plus élevés sur les pages et groupes de divertissement que sur ceux explicitement orientés vers l'humour est que ces réactions sont souvent attendues et même recherchées dans ces premiers espaces. Bien que les pages et groupes de divertissement n'aient pas pour objectif d'être signalés ou de perdre des abonnés, leur intention est clairement de susciter des réactions en provoquant ou en choquant le plus grand nombre d'utilisateurs. Comme mentionné précédemment, certains utilisateurs désapprouvent cette approche et sanctionnent les pages ou groupes en se désabonnant ou en les bloquant. Cependant, un grand nombre d'utilisateurs ne cherchent pas à contester leur contrat de communication, mais plutôt à s'en emparer pour afficher leurs compétences critiques et renforcer leurs liens de sociabilité. Les contenus relayés sont tellement gros, tellement ébouriffants, qu’il est facile de les critiquer et de les contester. Par exemple, face à la publication ci-dessous (cf. Figure 5.20), de nombreux internautes sont intervenus pour montrer qu’ils avaient l'œil : « photomontage lol , le ga a était rajouter sur la photo regarder bien mdrr » ; « Olalaaa mais vous avez pas encore remarquer que la photo et truquée on peut tout faire avaler au gens c'est fou ça» ; « montage bidon, dégueulasse ; tellement mal fait » ; « On fait quoi des jambes juste en dessous de l'homme ? Mdr ! » ; « photoshop mal fait »

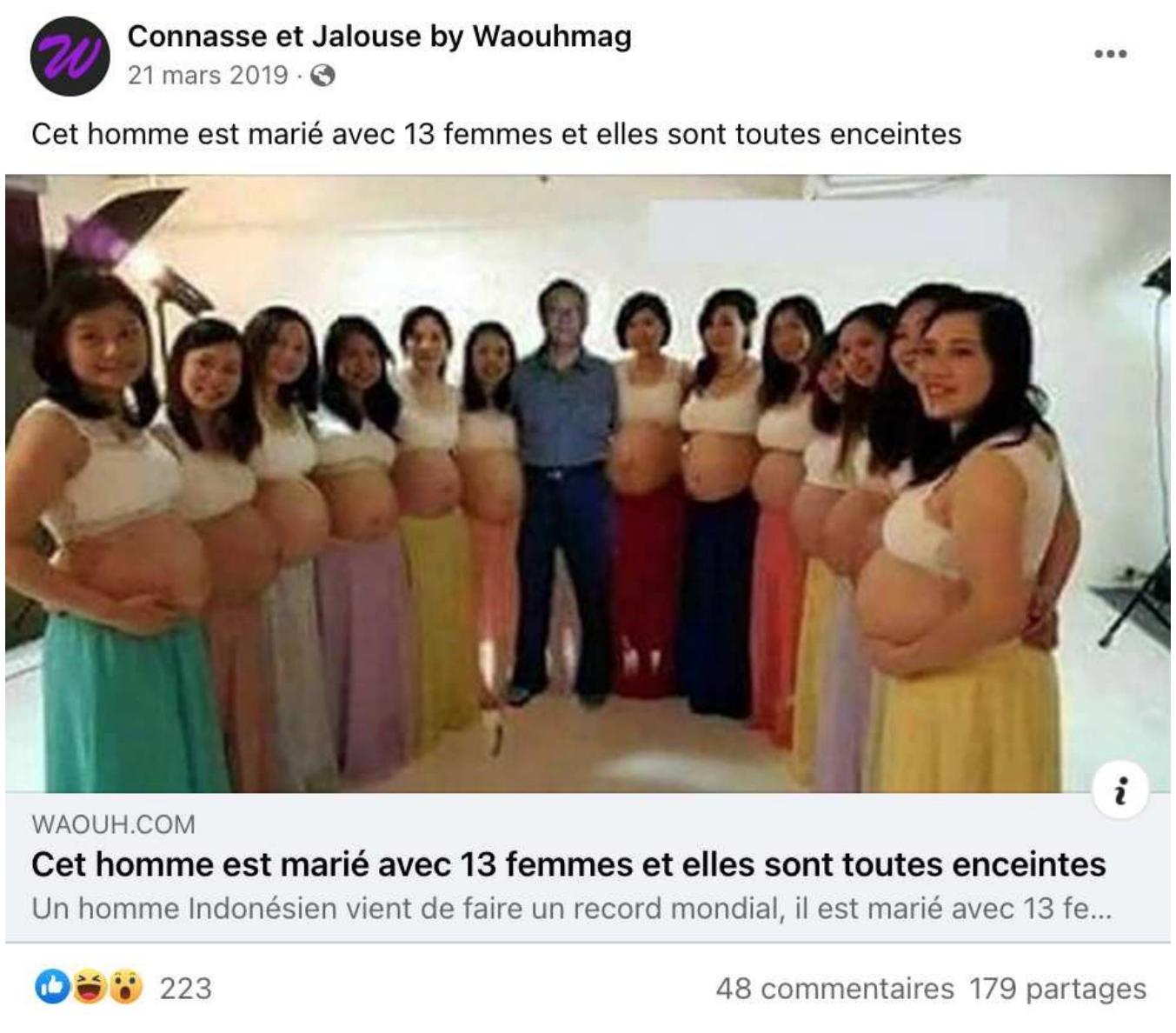

*Figure 5.20. Exemple de contenu partagé par plusieurs pages et groupes de divertissement*

Dans de nombreux cas, l'incrédulité commune des abonnés de la page permet de faire émerger des liens de complicité entre eux. Ensemble, ils se moquent du manque de crédibilité des contenus ; ensemble, ils s'auto-félicitent les uns les autres de ne pas être dupes. Par exemple, beaucoup de points d'arrêt comportent des tags accompagnés d'un nom d'utilisateurs et d'une remarque comme « t'as vu » ou « bro », « mec ». Enfin, plusieurs remarques ironiques et sarcastiques ont aussi été relevées sur ces espaces, témoignant une prise de distance critique de la part des utilisateurs sans pour autant marquer de points d'arrêt. En effet, ces interventions ne visent ni à rediriger la conversation ni à modifier l'atmosphère de la situation, mais plutôt à intensifier son caractère récréatif et divertissant. À travers leurs jeux de mots, les utilisateurs visent à divertir l'audience et à accroître leur popularité, comme en témoigne le nombre de réactions suscitées par ces critiques à un contenu annonçant la mort d'une célébrité déjà décédée dix ans plus tôt.

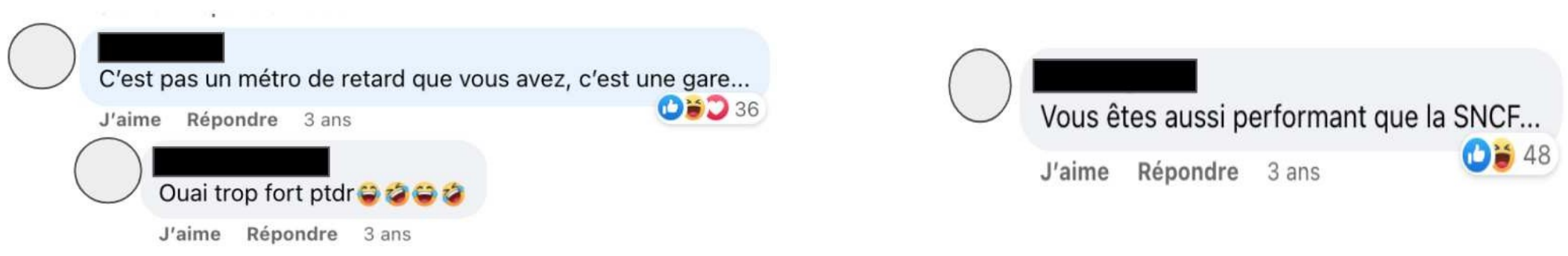

*Figure 5.21. Exemple de remarques ironiques formulées sur des pages/groupes de divertissement*

## 5.3.4. Entre autorégulation et modération collective

Le quatrième résultat suggère que la propension des utilisateurs à exprimer des points d'arrêt n'est pas significativement différente selon la présence ou l'absence d'une charte de modération sur une page ou un groupe Facebook, mais peut être encouragée par l'appartenance des utilisateurs à des groupes de modération collective.

Le taux de point d'arrêt des pages et des groupes qui possèdent une charte de modération est très légèrement inférieur à celui des espaces qui en sont dépourvus (i.e. 9,6% contre 9,7%). Par ailleurs, très peu de points d'arrêt exprimés par des utilisateurs détenant des badges d'administrateur ou de modérateur ont été relevés au cours de l'enquête ethnographique. Plutôt que de s'inscrire dans un cadre de régulation formelle, les points d'arrêt exprimés sur Facebook semblent donc davantage être le fruit de pratiques participatives et décentralisées, reflétant une forme d'autorégulation informelle qui dépasse la simple application de règles explicites. Ce constat ne permet néanmoins pas de conclure que la présence de charte de modération constitue une entrave à l'expression de points d'arrêt ni que leur absence constitue un levier. En fait, quand des utilisateurs s'abonnent à une page ou rejoignent un groupe public, il est probable qu'ils ne prêtent pas attention à la présence ou à l'absence d'une charte de modération ou qu'ils ne se souviennent pas s'il y en a une ou non au moment de poster un commentaire. En effet, la plupart du temps celle-ci se trouve nichée dans l'onglet à propos de la page ou du groupe, et il est peu probable que les utilisateurs la consultent avant de laisser un commentaire. Ainsi, l'expression de points d'arrêt est probablement davantage favorisée par d'autres facteurs situationnels : les indices vus ci-dessus jouent donc un rôle sans doute plus important que la présence ou l'absence d'une charte de modération. Par ailleurs, il est également possible que sur certains espaces de communication des modérateurs filtrent les commentaires et post problématiques avant que les utilisateurs n'aient eu le temps d'exprimer des points d'arrêt. Des recherches antérieures ont en effet démontré que la présence d'une charte de modération favorise non seulement une intervention plus fréquente des modérateurs, notamment à travers la suppression de commentaires, mais aussi une forme accrue d'auto-surveillance et d'auto-censure de la part des utilisateurs eux-mêmes (Gibson, 2019). À l'inverse, dans les groupes de divertissement, où l'objectif est de maximiser les interactions, la présence de modérateur est quasiment inexistante, ce qui laisse davantage d'espace aux utilisateurs pour prendre en charge la régulation des échanges et la gestion des désaccords.

Au-delà d'être une pratique située, variant d'un espace de communication à l'autre, selon la façon dont les utilisateurs des réseaux sociaux les perçoivent (et co-définissent les règles implicites de leurs échanges), l'expression de points d'arrêt dépend également de mécanismes sociaux et de logiques d'action collective, tels que le fait d'être intégré à des groupes de modération. Au cours d'un entretien, un enquêté, Julien, a par exemple indiqué avoir été invité à rejoindre le collectif #jesuislà après avoir posté un commentaire en ligne.

*Alors aujourd'hui, c'est moins le cas parce que j'ai rejoint une communauté. On me l'a proposé justement suite à des commentaires sur… Je ne sais plus quel article. Mais un jour où j'ai commenté un article comme ça quelqu'un est venu me voir en message privé pour me dire : "mais on a une communauté où l'idée c'est de se rassembler".*[183]

L'objectif de ce collectif est de lutter « contre la haine, la désinformation et [d']encourager le civisme en ligne pour faire d'Internet un endroit meilleur ».[184] Comme l'explique Julien, la stratégie de contre-discours proposée par le collectif consiste à exploiter l'algorithme de commentaires de Facebook pour amplifier leurs propres commentaires critiques tout en ensevelissant ceux qu'ils considèrent comme haineux afin de ne pas laisser les individus les plus extrêmes accaparer les espaces de visibilité en ligne :

*En fait le postulat de cette communauté…. L'idée de base de ce groupe c'est de se dire - c'est même pas de se dire c'est un fait qui est validé - c'est que les commentaires Facebook, il y a deux choses à savoir qui sont importantes : c'est de un que les commentaires qui sont les plus likés ou avec le plus de réactions ou de réponses, c'est ceux qui apparaissent en premier quand on ouvre les commentaires d'un post, et une étude scientifique a montré que l'avis des gens est influencé par les premiers commentaires sous une publication et que du coup il y a vraiment une clé dans l'opinion publique qui est faite par les premiers commentaires et du coup le but de cette communauté, c'est de prendre des articles plusieurs fois par jour - alors c'est des articles qui sont récents, qui ont moins d'une heure en général - et de se dire « ça part mal dans les commentaires pour cet article : il y a beaucoup de gens qui sont soit racistes, soit complotistes, enfin voilà ça part loin ». Donc on propose à la communauté du groupe d'aller commenter avec le hashtag #je suis là et d'aller liker les commentaires des autres de la communauté. Comme ça ça met des commentaires qui sont bienveillants, dans une démarche positive, ce n'est jamais des commentaires qui rentrent dans le conflit avec les autres, ça met ces commentaires en haut et donc qui vont potentiellement faire pencher l'opinion des personnes qui sont un peu dans l'hésitation et qui vont lire ça.* [185]

Plusieurs enquêtés, notamment un étudiant en médecine et un chirurgien, ont également indiqué avoir fréquenté des cercles zététiques. L'essor de ces mouvements décrit dans le premire chapitre (cf. section 1.2.4) pourrait d'ailleurs expliquer pourquoi les pages et groupes dédiés à la santé affichent un taux de points d'arrêt plus élevé que d'autres catégories de groupes et pages spécialisés. En effet, ces groupes se concentrent souvent sur des thématiques controversées telles que les vaccins, les OGM et la 5G, ce qui génère des discussions intenses et polarisées.

---

[183] Homme, 34 ans, Bac, accompagnant éducatif et social, entretien réalisé le 1er août 2022.
[184] Pour plus d'informations voir : https://jesuisla.net/
[185] Homme, 34 ans, Bac, accompagnant éducatif et social, entretien réalisé le 1er août 2022.

> *Alors c'est assez drôle parce qu'en fait j'ai été contacté par des gens qui se coltinent les anti-vaccins depuis des années. Puis bon ils avaient trouvé que j'étais capable d'argumenter, que j'avais un certain nombre de connaissances et ils m'ont dit "ben voilà, viens, on a un groupe". Et donc je suis rentré dans un groupe Facebook où je me suis retrouvée effectivement à échanger avec ces gens-là et c'est là que j'ai découvert les pages en fait.*[186]

Appartenir à des groupes de modération collective présente un double intérêt pour les utilisateurs des réseaux sociaux. D'une part, cela permet de surmonter les mécanismes de spirale du silence, en offrant un soutien qui encourage l'expression d'opinions divergentes. Par exemple, dans le cadre du collectif #JeSuisLà, ce hashtag est conçu pour agir comme une barrière de protection, incitant les membres à prendre la parole non seulement en ligne, mais aussi dans d'autres espaces de discussion, y compris hors ligne. Le sentiment d'appartenance à une communauté renforce la confiance des individus, leur permettant de se sentir moins isolés et de gagner le courage nécessaire pour exprimer des points de vue qui pourraient autrement être réprimés (Garland et al., 2020 ; Friess et al., 2021). D'autre part, cette dynamique communautaire facilite également le travail de vérification et de contre-argumentation. Les membres de ces groupes partagent des ressources telles que des fiches d'information, des liens et des statistiques qui renforcent leurs arguments.

> *On avait un espace collaboratif où on a déconstruit toute l'argumentation des anti-vaccins. [...] Voilà donc on était dans un espace collaboratif où on échangeait. Il y avait d'autres médecins, des biologistes, des pharmaciens. [...]. On s'est dit, voilà qu'il fallait qu'on collabore, qu'on essaie de trouver des gens, et au fil du temps sur les forums, on a réussi à attraper des gens qui manifestement avaient des compétences et qu'on a amené dans le groupe pour travailler.* [187]

---

[186] Homme, 44 ans, Bac +8, médecin, entretien réalisé le 2 août 2022.
[187] *Ibid.*

*En fait, on a fait des pages et des documents via les Vaxxeuses et d'autres collectifs qui permettent en gros d'avoir des trucs tout fait parce qu'il y a des arguments qui reviennent tout le temps. [...] Donc si tu veux, à chaque fois pour certains trucs on avait nos blogs, et des ressources documentaires qu'on pouvait ressortir pour faciliter un petit peu les choses parce qu'on ne peut pas expliquer 20 fois le formaldéhyde, 20 fois l'aluminium, etc. Donc c'est vrai qu'on avait des ressources qui étaient déjà définies.* [188]

## Conclusion du cinquième chapitre

En adoptant une perspective inverse à celle de la majorité des études quantitatives sur les *fake news*, c'est-à-dire en partant de la façon dont le problème des *fake news* est posé par les utilisateurs des réseaux sociaux plutôt que par les discours publics, ce chapitre a permis de rendre compte des pratiques des utilisateurs consistant à intervenir dans le flux d'une conversation en ligne pour exprimer des mises en doute, des critiques ou des désaccords. En prenant appui sur des travaux de sociologie de la réception, d'analyses conversationnelles et de sociologie pragmatique sur la lecture oppositionnelle, l'expression de désaccords et la dénonciation publique, une nouvelle notion a été introduite, celle de point d'arrêt, pour qualifier ces pratiques énonciatives.

Bien que les utilisateurs reprennent parfois les mêmes éléments de langage que ceux des discours publics pour dénoncer le problème des *fake news* et semblent avoir intériorisé certaines normes journalistiques, ils prennent également appui sur d'autres valeurs et mettent en lumière d'autres troubles lorsqu'ils expriment des points d'arrêt dans une conversation en ligne. Loin de réduire la problématique des *fake news* à la seule question de la factualité d'un énoncé, ils l'associent à différentes causes : les biais idéologiques des médias, leurs motivations économiques, le manque de responsabilité des pages et des groupes qui diffusent ces informations, ainsi que le manque d'esprit critique des autres utilisateurs.

---

[188] *Ibid.*

Au-delà de rendre compte des divers régimes de critiques employés par les utilisateurs des réseaux sociaux, ce chapitre a également permis d'examiner comment la propension des utilisateurs à exprimer des points d'arrêt se manifeste différemment selon les situations d'énonciation et les contrats de communication qui les sous-tendent. Au terme de ce chapitre, nous avons ainsi pu identifier les espaces qui favorisent l'expression de points d'arrêt et rendre compte des variations, non seulement dans la propension des utilisateurs à les exprimer, mais aussi dans la manière dont ils les énoncent. Ces constats invitent à considérer les compétences de distance critiques moins comme une disposition cognitive ancrée dans l'esprit des personnes que comme une capacité construite par les divers contextes d'énonciation dans lesquels elles évoluent.

Afin d'approfondir ces résultats, le dispositif d'enquête mis en place dans ce chapitre pourrait être étendu à d'autres plateformes. À titre d'exemple, nous avons mobilisé les données d'une enquête ayant mesuré et cartographié l'empreinte antisémite et les discours de haine sur Youtube (de Dampierre et al., 2022 ; Tainturier et al., 2023). À partir de ces données, nous avons pu détecter les points d'arrêt dans les commentaires YouTube et observer leur répartition au sein des différentes communautés de la cartographie des chaînes YouTube réalisée par l'étude (cf. Annexe 8). Une interprétation approfondie n'a pas pu être proposée car nous n'avons pas directement contribué à ce projet de recherche et n'avons pas pu effectuer un travail ethnographique similaire à celui réalisé sur Facebook permettant de saisir l'atmosphère de différents espaces et les dynamiques interactionnelles entre les utilisateurs. Cependant, cette analyse met en lumière le potentiel d'adapter et de reprendre les méthodes employées dans notre étude pour de futures recherches sur d'autres plateformes.

En définitive, ce chapitre montre que les réseaux sociaux ne forment pas un tout homogène. Ils se composent plutôt d'une pluralité d'espaces de communication qui, bien qu'ils permettent des formes d'énonciation plus souples et relâchées, n'empêchent pas l'émergence de formes d'autorégulation conversationnelle. Est-ce à dire pour autant que les points d'arrêt peuvent être considérés comme des indicateurs permettant de mesurer, sinon la qualité d'un débat, du moins son ouverture à une pluralité d'opinion ? Nous allons explorer cette question dans le prochain chapitre notamment en examinant comment les points d'arrêt sont perçus par les autres utilisateurs.

# Chapitre 6. Apports et limites des points d’arrêt

Le chapitre précédent a montré comment la propension des utilisateurs à exprimer des points d’arrêt varie selon plusieurs indices d’énonciation susceptibles de les aider à comprendre le contrat de communication qui sous-tend une situation d’interactions en ligne. Par exemple, les utilisateurs ont davantage tendance à exprimer des points d’arrêt face à des contenus publiés sur des pages et des groupes Facebook qui s’adressent à une audience importante et/ou hétérogène et/ou peu politisée, et ce d’autant plus lorsque les contenus proviennent de sources dotées d’une faible autorité dans l’espace médiatique français ou portent sur des thématiques sensibles, fréquemment associées à des discours de haine dans le débat public.

À première vue, ces pratiques énonciatives, consistant à marquer publiquement un désaccord ou une critique face à des contenus perçus comme problématiques, peuvent être interprétées comme des indicateurs d’une ouverture de certains espaces de communication à une pluralité d’opinions — un critère important pour garantir la qualité du débat public, que cela soit d’après une perspective délibérative ou agonistique de l’espace public. En effet, dans les modèles de la délibération d’inspiration habermassienne, l’ouverture à une pluralité d’opinions permet de soumettre tous les arguments à l’épreuve de la raison critique afin de dépasser les conflits et d’atteindre un consensus (Habermas, 1984 ; 1996). À l’inverse, dans les conceptions agonistiques de la sphère publique, inspirées des travaux de Chantal Mouffe (2000 ; 2013), l’ouverture à la pluralité des opinions n’a pas pour but de parvenir à un consensus, mais plutôt d’accepter l’existence de divergences irréconciliables tout en reconnaissant chaque adversaire comme légitime.

Pour que l’expression de point d’arrêt soit réellement le signe d’une ouverture des espaces de communication à une pluralité d'opinions, que cela soit d’après une perspective délibérative ou agonistique, il faudrait donc que plusieurs conditions soient remplies : (1) d’abord que l'expression de point d’arrêt soit accessible au plus grand nombre d’utilisateurs indépendamment de leurs caractéristiques socio-démographiques ; (2) ensuite, qu’elle encourage la participation des autres utilisateurs et favorise la poursuite des échanges ; (3) enfin, que ces échanges permettent soit (i) un rapprochement des points de vue des interlocuteurs (perspective délibérative), soit (ii) une reconnaissance réciproque des positions divergentes (perspective agonistique).

L'un des objectifs de ce chapitre est d'explorer si les points d'arrêt répondent à ces critères, afin de comprendre dans quelle mesure certains espaces de communication en ligne peuvent être considérés comme des sphères publiques, et quelles sont les conditions socio-techniques favorisant leur émergence.

Cependant, restreindre l’analyse à ces seules perspectives reviendrait à projeter sur les utilisateurs des réseaux sociaux une intention univoque : celle de préserver ou de restaurer un espace de débat public, qu’il soit orienté vers des échanges délibératifs ou des confrontations agonistiques. Or, une telle projection fait abstraction de la diversité des usages concrets des réseaux sociaux. Il est donc nécessaire d'élargir le cadre d’analyse pour prendre en compte la possibilité que les points d’arrêt formulés sur les réseaux sociaux ne visent pas uniquement à rétablir un débat rationnel ou à exprimer un dissensus légitime. Comme cela a été suggéré dans le chapitre précédent, ces interventions critiques peuvent parfois chercher à préserver des formes d’interactions plus légères, à renforcer des liens de sociabilité ou à restaurer des espaces de badinage et de distraction.

Dans cette optique, plutôt que de se limiter à une évaluation normative des critères de délibération et d’agonisme, ce chapitre propose également de s’intéresser à la manière dont les points d’arrêt sont effectivement perçus et jugés par les autres utilisateurs des réseaux sociaux. Comment ces derniers réagissent-ils face à ces interventions critiques ? Les trouvent-ils justifiées et légitimes ? Ou, au contraire, les considèrent-ils comme des intrusions indésirables dans les échanges ? Loin d’imposer une grille d’interprétation extérieure, il s’agit ici de laisser les utilisateurs eux-mêmes définir ce qu’ils considèrent comme pertinent ou impertinent, approprié ou non, en fonction des situations d’interaction.

Cette démarche pragmatique vise ainsi à identifier les conditions concrètes de « félicité » des points d’arrêt, c’est-à-dire les circonstances dans lesquelles ces interventions sont jugées adéquates par ceux qui y sont confrontés. En analysant ces micro-situations de jugement, il devient possible de mieux comprendre les logiques d'action des utilisateurs et les dynamiques de légitimation ou de contestation qui se jouent dans les échanges en ligne.

Après avoir examiné quelles sont les caractéristiques des utilisateurs qui expriment des points d'arrêt sur les réseaux sociaux, ce chapitre se penche sur les réactions que déclenchent ces interventions chez les autres utilisateurs. Une réflexion sur les conditions socio-techniques qui favorisent ou, au contraire, entravent l'émergence de sphères publiques délibératives ou agonistiques est enfin proposée, ainsi que sur celles qui permettent aux utilisateurs de préserver, de restaurer ou de créer des espaces communs d'intercompréhension répondant à leurs propres attentes normatives.

## 6.1. Minorité expressive, majorité silencieuse

Malgré leurs divergences, les conceptions de l'espace public inspirées des travaux de Jürgen Habermas et de Chantal Mouffe partagent un horizon commun : celui d'encourager la participation du plus grand nombre au débat public. Que celui-ci soit orienté vers la recherche de consensus par l'échange d'arguments rationnels ou vers la reconnaissance de conflits par la confrontation respectueuse d'opinions incompatibles, il doit garantir une égalité discursive entre tous les utilisateurs, ou du moins leur offrir des opportunités de participation équitables.

L'enjeu de cette section est ainsi d'explorer si l'expression de points d'arrêt est une pratique accessible à tous les utilisateurs des réseaux sociaux, indépendamment de leurs caractéristiques socio-démographiques, ou si elle est inégalement répartie au sein de la population. Autrement dit, il s'agit de se demander si des facteurs comme l'âge, le genre, le niveau d'éducation, ou encore l'appartenance sociale influencent la capacité des internautes à exprimer publiquement des désaccords ou des critiques.

Cette section vise également à étudier si les points d'arrêt favorisent la poursuite des discussions sur les réseaux sociaux ou si au contraire ils tendent à rester sans réponse. En d'autres termes, la question est de savoir si les points d'arrêt agissent comme des leviers à la participation en ligne ou s'ils fonctionnent plutôt comme des mécanismes de clôture, limitant la diversité des opinions exprimées.

### 6.1.1. Reproduction des inégalités associées à la participation en ligne

Parmi les 294 988 commentateurs recensés dans le corpus étudié, seulement 7,1 % ont exprimé au moins un point d'arrêt, et, parmi cette minorité, plus de 83,7 % n'ont produit qu'un seul point d'arrêt, tandis qu'à peine 0,02 % en ont formulé plus de dix. Ces chiffres indiquent que l'expression de points d'arrêt est l'apanage d'une petite fraction de commentateurs sur Facebook et invitent à s'interroger sur les caractéristiques qui distinguent ces utilisateurs des autres. Si des données sur le nombre de vues par publication avaient été disponibles, il aurait été intéressant de calculer la proportion d'utilisateurs ayant commenté chaque post par rapport à son audience totale. Cela aurait permis d'évaluer la part d'utilisateurs qui, bien qu'exposés à des contenus sur Facebook, choisissent de ne pas y réagir.

Ces résultats s'inscrivent dans la lignée des travaux relatifs à la participation en ligne. Sur les forums, blogs et réseaux sociaux, seule une toute petite minorité d'utilisateurs est responsable de la majorité des contenus et des commentaires produits (Sun et al., 2014 ; Van Mierlo, 2014). Ce phénomène est souvent décrit par la « règle des 90-9-1 », selon laquelle 90 % des utilisateurs se contentent de consommer des contenus sans interagir avec ; 9 % les modifient ou commentent ; et seulement 1 % sont des créateurs actifs de ces contenus. Par exemple, une étude de Gibson (2019) indique que les 1 % des commentateurs les plus prolifiques génèrent à eux seuls plus d'un tiers des commentaires (34,5 %), tandis que les 10 % les plus actifs en produisent les deux tiers (66,8 %).

Cette minorité d'utilisateurs actifs se distingue souvent par des spécificités socio-démographiques. En général, ce sont les individus les plus diplômés qui participent le plus en ligne, que ce soit en produisant des contenus originaux, en prenant part à des discussions sur des forums, ou en réagissant à des articles de presse (Schradie, 2011). Cela s'explique en partie par le fait que les utilisateurs dotés d'un capital scolaire important maîtrisent davantage les compétences techniques requises pour naviguer aisément sur les plateformes numériques et créer des contenus écrits ou multimédias (Hargittai et al., 2014). Par exemple, les travaux sur la sociologie des contributeurs de Wikipédia montrent que ces derniers sont majoritairement des hommes disposant d'un niveau d'éducation élevé et d'un certain niveau de compétences en informatique (Julien, 2012). Par contraste, les individus avec un faible capital scolaire tendent à adopter des usages plus modestes et pragmatiques d'Internet et évitent de passer par des formats textuels complexes. Une enquête conduite par Dominique Pasquier (2018) suggère que ces utilisateurs considèrent avant tout Internet comme un outil pour resserrer les liens familiaux et accéder à des informations pratiques. Ils préfèrent ainsi partager directement des contenus tout faits plutôt que d'en produire eux-mêmes.

Par ailleurs, plusieurs études ont mis en lumière la persistance des inégalités de genre dans les interactions en ligne et la création de contenus. [189] Comme lors d'échanges en face à face (Zimmerman et West, 1975), les femmes tendent à intervenir moins souvent, ont moins de chances de voir leurs sujets de discussion poursuivis, et sont plus fréquemment interrompues que les hommes sur les forums de discussion en ligne (Sutton, 1994 ; Karpowitz et al., 2012).

Dans le cas de notre enquête, les entretiens réalisés auprès d'utilisateurs exprimant des points d'arrêt ont fait ressortir des profils très homogènes (cf. Tableau 2.2), avec un échantillon d'enquêtés composé exclusivement d'hommes, relativement jeunes, âgés de 20 à 47 ans (34 ans en moyenne), très diplômés et/ou politisés, ainsi que très actifs sur les réseaux sociaux et très intéressés par l'actualité. Par exemple, parmi les 16 personnes interrogées, sept possèdent un diplôme de niveau Bac +5, tandis que deux autres détiennent un doctorat. Dans l'ensemble les pratiques informationnelles et numériques des utilisateurs interrogés ne sont pas du tout représentatives de celles de la majorité de la population. Ils consacrent un temps important à s'informer chaque jour et ont installé plusieurs applications sur leur smartphone pour suivre les actualités des médias traditionnels ; ce qui indique que leur exposition à des informations d'actualité n'est pas accidentelle mais activement recherchée. Par ailleurs, comme mentionné dans le chapitre précédent, leur parcours académique, leurs professions (médecin, enseignant, etc.) ou les fonctions qu'ils exercent au sein d'une municipalité ou d'une organisation (maire, président d'association, etc.) les conduisent à se référer régulièrement au « consensus scientifique » ou à « l'intérêt général ». Ils sont ainsi très attentifs à la fiabilité des sources et des contenus qu'ils consultent et choisissent de lire des revues qui offrent une pluralité de points de vue comme *Courrier international* ou des analyses critiques comme *Arrêt sur Image*. Ils regardent également des émissions de débats avec des spécialistes comme *C'est dans l'air* ou écoutent de longs podcasts d'histoire ou de vulgarisation scientifique. Un enquêté a aussi indiqué avoir recours à des outils d'intelligence artificielle comme *Perplexity* pour approfondir ses recherches. [190]

[189] Cette problématique est actuellement traitée par Emma Gauthier dans le cadre de la thèse qu'elle effectue au LISIS.

Sur les 16 personnes interrogées, les trois quarts ont déclaré être très intéressées par la politique et ont affiché un positionnement politique très précis. Un participant, par exemple, a déclaré qu'il se décrirait comme « néo-rocardien ». [191] La plupart ont indiqué se situer à gauche de l'échiquier politique, et quatre se revendiquent plutôt du centre-droit. Parmi ces individus, quasiment aucun n'est membre d'un parti politique. Seulement un enquêté a été maire de sa ville pendant six ans. La majorité autrement se considère comme des militants ou des citoyens engagés, à la fois en ligne et hors ligne.

Le profil de Julien illustre bien le caractère très politisé et engagé de certains des utilisateurs qui expriment des points d'arrêt. Très actif sur Facebook, il publie environ un post par jour sur divers sujets d'actualité politique (notamment liés à l'écologie ou au féminisme) et critique souvent les mesures prises par le gouvernement ainsi que les discours émis par des personnalités d'extrême droite. Son profil est très ouvert et permet de voir qu'il soutient

---

[190] Homme, 28 ans, Bac +5, développeur, entretien réalisé le 2 mars 2023.
[191] Homme, 47 ans, Bac +5, ingénieur, entretien réalisé le 12 août 2024.

*Europe Écologie Les Verts* et *La France Insoumise* et qu'il est impliqué dans des organisations citoyennes comme *The Global Warming Proofing*[192] visant à sensibiliser la population aux enjeux du réchauffement climatique en menant des opérations de « désintoxication » des réseaux sociaux cherchant à lutter contre la désinformation climatique en publiant régulièrement des *fact-checking* et des infographies pédagogiques.

D'une manière générale, les points d'arrêt exprimés par Julien sont étroitement liés à ses positions politiques et découlent d'une logique agonistique. Son but est de contrer le cadrage idéologique que cherchent à imposer des médias d'extrême droite et de défendre des causes soutenues par la gauche. D'autres interventions de Julien ont ainsi concerné des sujets féministes ou liés aux violences sexistes et sexuelles. Par exemple, il s'est mobilisé pour inviter ses contacts Facebook à signaler une page anti-IVG, et il a indiqué être intervenu récemment sur un post Facebook pour s'opposer à des commentaires niant le caractère abusif du *stealthing* (i.e. le fait de retirer un préservatif pendant un rapport sexuel sans le consentement de son/sa partenaire). Par ailleurs, l'intérêt de Julien pour différents sujets d'actualité politique et son implication dans la vie locale semblent plus largement le pousser à faire montre d'une vigilance citoyenne au quotidien, voire d'une forme de vigilantisme numérique (Loveluck, 2016). Il intervient dans plusieurs groupes de sa commune pour signaler le non-respect de certaines règles (cf. Figure 6.1). Par exemple, il n'hésite pas à envoyer un mail au service client d'une entreprise, à inciter d'autres habitants à coller des panneaux stop pub sur leur boîte aux lettres et à proposer des actions collectives pour dénoncer le non-respect de cette loi. Ses pratiques de signalement ne se résument donc pas au fait d'écrire un simple commentaire sur Facebook et dépassent la seule question de l'évaluation de la qualité épistémique d'une information. Enfin, quelques indices laissent penser que Julien est une personne qui aime bien chercher la petite bête, même lorsqu'il échange avec des contacts qui partagent son point de vue. Par exemple, après avoir émis un commentaire critique en réaction à une publication d'une association au sein de laquelle il est investi, il reconnaît « [avoir] envie d'être chiant » et fait preuve d'auto-dérision en se qualifiant de « gros relou qui ne lâche jamais l'affaire et qui a toujours raison ». Au cours d'entretiens, quelques enquêtés ont également indiqué se considérer comme têtus et obstinés, ou se sont décrits, avec une pointe d’humour dans la voix, comme maniaques et obsessionnels — des réponses qui suggèrent que l'expression des points d'arrêt est non seulement liée à un niveau élevé de politisation, mais également à certains traits de personnalité spécifiques.

[192] Afin de préserver l'anonymat de l'enquête, le nom de l'organisation et la description de son activité ont été modifiés.

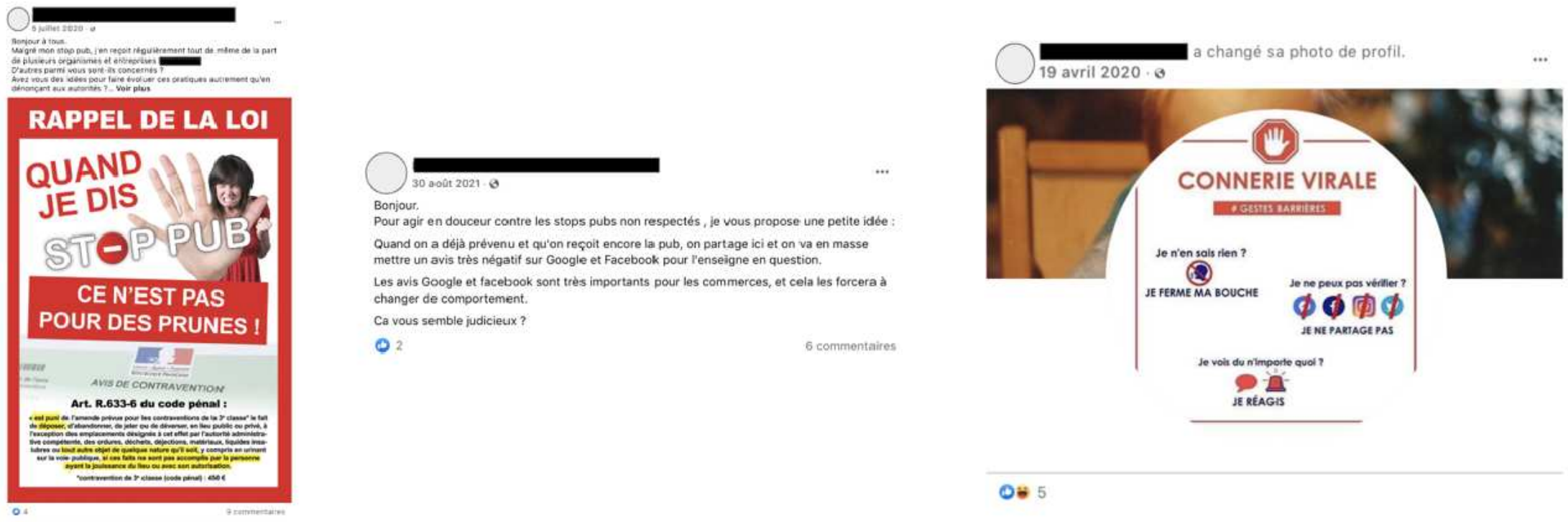

*Figure 6.1. Captures d’écran issues du compte de Julien*

Parmi les utilisateurs avec lesquels des entretiens ont été réalisés se trouvent également quelques profils moins politisés et/ou moins dotés en capital scolaire. Damien[193], par exemple, n’a pas poursuivi d’études supérieures après avoir passé son bac, et travaille aujourd’hui comme superviseur dans un aéroport. Il ne se sent proche d’aucun parti politique et a indiqué avoir voté pour Emmanuel Macron en 2017 et Marine Le Pen en 2022 sans réelle conviction « pour avoir du changement [...] et sanctionner Macron ». Depuis 2013, il administre une page Facebook suivie par 145 000 comptes. Celle-ci vise à communiquer des informations pratiques à propos d’une route nationale française (e.g. sur la météo, le trafic, l’ouverture de restaurants, etc.). Comme il l’explique lui-même, le fait de gérer cette page le pousse régulièrement à vérifier scrupuleusement les informations qu'il partage avec ses abonnés, et lui donne l’opportunité d’être ponctuellement en contact avec des journalistes.

193 Homme, 36 ans, Bac, superviseur des bagages dans un aéroport, entretien réalisé le 16 février 2023.

*Ben… C'est que… Comme moi j'ai une page où je publie beaucoup d'informations, je fais attention aux sources oui.*

*Du coup, avec ma page, ben j'ai dans mon répertoire des journalistes de TF1, BFM [...]. Ils ont déjà pris des photos que j'avais publiées sur ma page et ils m'ont demandé les autorisations. Donc, j'ai des contacts avec eux quand ils me demandent l'autorisation de prendre des photos.*

L'expérience de Damien dans l'administration d'une page Facebook semble également l'avoir sensibilisé aux enjeux de modération de contenus en ligne, et lui avoir permis d'acquérir sur le tas une forme de responsabilité éditoriale. Par exemple, il indique veiller à ne pas laisser de commentaires susceptibles de causer du tort aux personnes impliquées dans un accident de la route ou à leur entourage. Ses décisions de suppression ne sont donc pas prises selon sa sensibilité personnelle mais dans l'intérêt de sa communauté.

*On est trois à gérer la page. Je suis le créateur. Et en fait pareil quand il y a des commentaires qui passent à côté du filtre à injure et que ça nous plaît pas – pas parce qu'on aime pas le commentaire mais parce que ça peut être… Genre s'il y a un accident et qu'il y a une personne blessée gravement dedans, si la personne juge le conducteur et qu'on sait que le conducteur est entre la vie et la mort – ce qu'on a eu il y a deux-trois semaines là – on va modérer pour que la famille ne voit pas les commentaires.*

Ainsi, au-delà de l'aspect technique de la gestion d'une page, Damien a progressivement intériorisé des règles de modération qu'il applique de manière pragmatique, en fonction des situations et des interactions spécifiques auxquelles il fait face. Ses décisions ne sont pas dictées par une théorie abstraite de la modération, mais par les ajustements qu'il opère au fil de son expérience, en tenant compte des besoins de sa communauté et des dynamiques relationnelles.

Au fil du temps, les pratiques numériques de Damien paraissent également avoir augmenté son intérêt pour l'information d'actualité et contribué à son éveil politique. Contrairement à ses débuts en tant qu'électeur, où il ne s'intéressait pas aux programmes des candidats, il se renseigne désormais davantage en consultant des sources qu'ils jugent fiables et neutres comme *Le Parisien*. Cette évolution montre comment une pratique quotidienne, ancrée dans des interactions concrètes, peut influencer l'intérêt pour l'information et les compétences d'un individu en matière de discernement des sources et de modération de contenus, illustrant l'apprentissage progressif de ce que l'on pourrait appeler une « grammaire publique » de la prise de parole numérique.

> *Je pense qu'on peut plus débattre maintenant qu'avant avec les réseaux sociaux parce qu'on a plus d'informations sur… Enfin je vous dis ça, quand j'ai eu 18 ans, quand j'ai voté pour la première fois, je ne lisais pas forcément le programme de chaque candidat, alors que maintenant en ligne, ben si !* Le Parisien *ou quoi ce soit ils mettent quelques infos sur le programme de chaque candidat là ça permet vraiment de se faire un peu plus une opinion sur le programme des personnes politiques.*

Ainsi, bien que l'expression de points d'arrêt en ligne n'échappe pas à la reproduction d'inégalités classiques, en étant dominée par une minorité d'utilisateurs au profil très homogènes, les réseaux sociaux semblent tout de même avoir facilité l'accès à certaines pratiques de vérification et de modération pour des publics plus diversifiés. Cette légère ouverture a été perceptible à travers les observations réalisées sur des comptes d'utilisateurs, qui ont révélé une plus grande diversité de profils par rapport aux entretiens. Bien que les comptes soient la plupart du temps fermés, les utilisateurs indiquent leur vrai prénom et parfois leur profession ou les études qu'ils ont suivies. Le genre des utilisateurs a pu être déterminé facilement à partir des noms d'utilisateurs. Environ 2 000 noms d'utilisateurs des comptes à l'origine d'un point d'arrêt dans le corpus de commentaires annotés manuellement ont été classés selon trois modalités : (1) homme, (2) femme, (3) indéterminable (prénoms mixtes, noms écrits dans un alphabet non latin, pseudonymes, comptes gérés par un collectif, une association ou une organisation). Un classifieur a ensuite été utilisé pour élargir les annotations à l'ensemble des utilisateurs à l'origine d'un point d'arrêt. Cette approche a permis de déterminer que 50,1 % des utilisateurs à l'origine d'un point d'arrêt sont des femmes, 48,3 % des hommes, et 1,6 % appartiennent à des comptes dont le genre n'a pas pu être déterminé.[194] Cette répartition est assez proche de celle de la population générale sur Facebook en France, où 52,5 % des utilisateurs sont des femmes et 47,5 % des hommes.[195]

Alors que l'expression de points d'arrêt semble être équitablement répartie entre les hommes et les femmes, il est surprenant de n'avoir obtenu de réponse positive que de la part d'hommes à nos demandes d'entretien. Ce décalage fait écho avec l'enquête de Jennifer Stromer-Galley (2003). Celle-ci a également envoyé des messages privés à des personnes participant à des forums. Sur les 69 réponses positives qu'elle a reçues, 62 provenaient d'utilisateurs masculins. Il est possible que les femmes soient plus méfiantes à l'égard de messages reçus par des inconnus sur les réseaux sociaux et que les hommes soient plus enclins à y répondre. La méfiance des femmes à l'égard des messages privés provenant d'inconnus sur les réseaux sociaux peut être liée à une perception accrue des risques en ligne, tels que le harcèlement ou les abus, ce qui les rend plus prudentes quant aux interactions non sollicitées (Herring et al., 2002 ; Jane, 2018).

La mixité observée dans l'expression des points d'arrêt en ligne peut surprendre au regard des travaux sur la participation en ligne – d'autant plus que cette pratique est probablement plus risquée pour les femmes étant donné qu'elles sont souvent plus exposées au cyber harcèlement et aux discours de haine en ligne (Citron, 2016). Par ailleurs, de nombreuses études conduites sur les profils des individus qui font de la vulgarisation scientifique sur Youtube ou gravitent autour des mouvements rationalistes et des cercles zététiques ont montré l'importante prédominance des hommes dans ces milieux (Dauphin, 2022 ; Debove et al., 2021 ; Martins Velho et al., 2020 ; Morcillo et al., 2019 ; Amarasekara et Grant, 2019). On aurait ainsi pu s'attendre à ce que l'expression de points d'arrêt en ligne soit davantage le fait d'utilisateurs masculins dans la mesure où cette pratique n'est pas déconnectée des logiques de debunking et de pensée critique, traditionnellement dominées par des hommes.

[194] Pour ces trois modalités, le score F1 était respectivement de 0,96 ; 0,96 et 0,79.

[195] Statista. (2023). *Répartition des utilisateurs de Facebook parmi les internautes en France de 2016 à 2023, selon le sexe*. https://fr.statista.com/statistiques/491487/taux-penetration-facebook-france/

Toutefois, malgré la mixité globale observée dans l'expression de points d'arrêt, il est important de noter que la propension des hommes et des femmes à intervenir dans le flux d'une conversation varie selon les espaces de communication et les types de contenu abordés (cf. Figures 6.2 à 6.4). Par exemple, les hommes ont plus tendance à intervenir au sein de pages ou de groupes politisés, notamment en réaction à des contenus portant sur des sujets d'intérêt général en lien avec l'actualité politiques ou provenant de médias partisans. À l'inverse, environ 70 % des utilisateurs qui expriment des points d'arrêt sur des pages/groupes liés à la santé, aux animaux ou à l'éducation sont des femmes. Par ailleurs, celles-ci sont davantage susceptibles que les hommes d'intervenir en réactions à des contenus venant de médias de divertissement et de santé alternative ou portant sur des sujets comme l'immigration et l'inclusion sociale. Ces constats font écho aux observations recueillies auprès d'un enquêté occupant des responsabilités au sein de l'association #JeSuislà. [196] Celui-ci, ayant accès aux statistiques de la page du collectif en France, a indiqué que les femmes représentaient environ deux tiers des membres actifs, soulignant ainsi une implication féminine significative dans les activités du collectif. L'enquêté a également noté que les membres masculins du mouvement étaient souvent moins impliqués. Ce phénomène contraste avec la prédominance masculine observée dans d'autres milieux comme les cercles zététiques ou la vulgarisation scientifique sur YouTube. Ces différences d'implications entre les hommes et les femmes dans le collectif #JeSuisLà pourrait refléter une approche plus axée sur la préservation d'espaces de discussion respectueux, bienveillants et inclusifs, dans une logique d'entraide et de modération collective, plutôt que sur la factualité des contenus dans une dynamique de confrontation compétitive.

[196] Homme, 38 ans, Bac +5, consultant, entretien réalisé le 21 mars 2023.

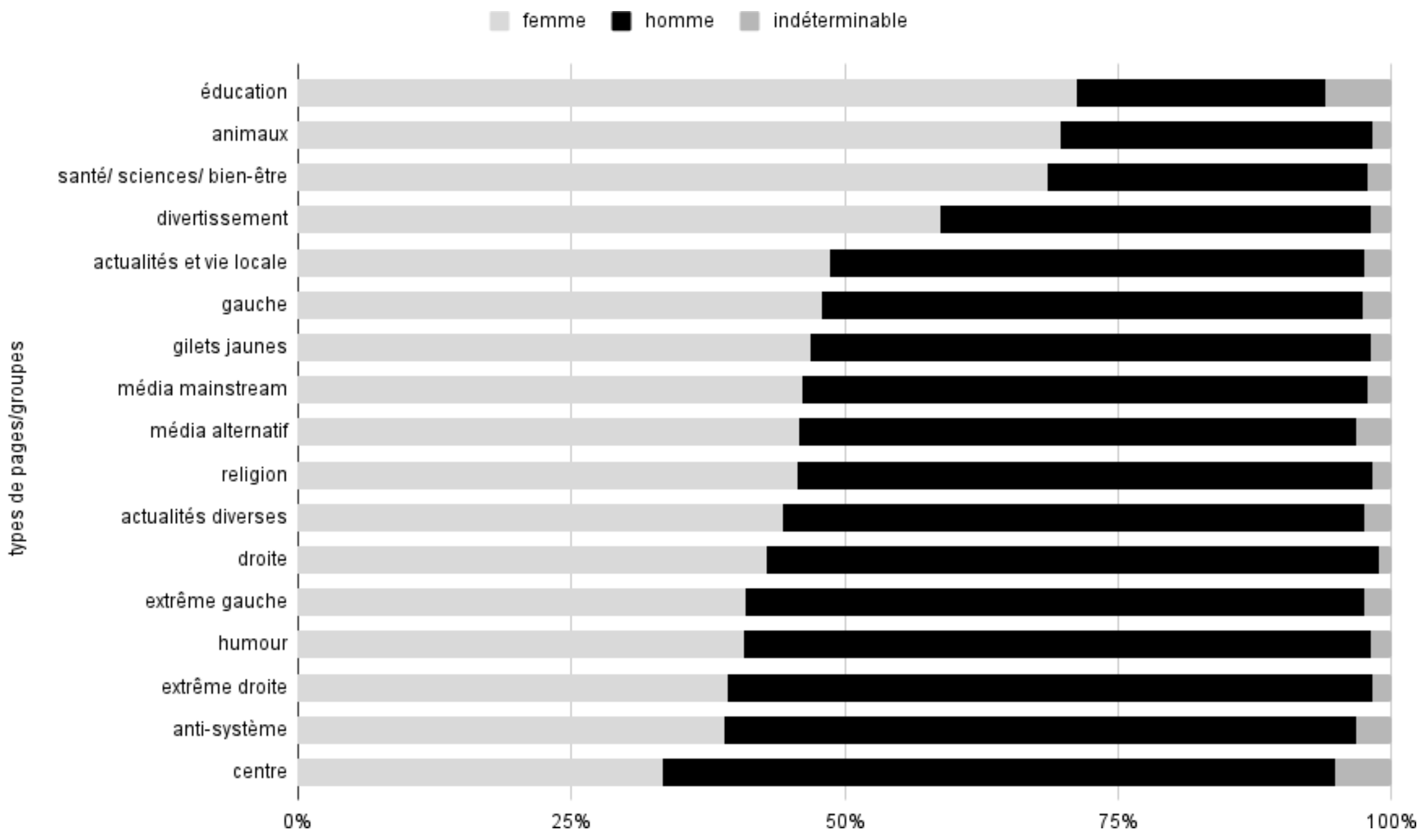

*Figure 6.2. Proportions de points d'arrêt exprimés par des hommes ou des femmes selon les types de pages et de groupes Facebook*

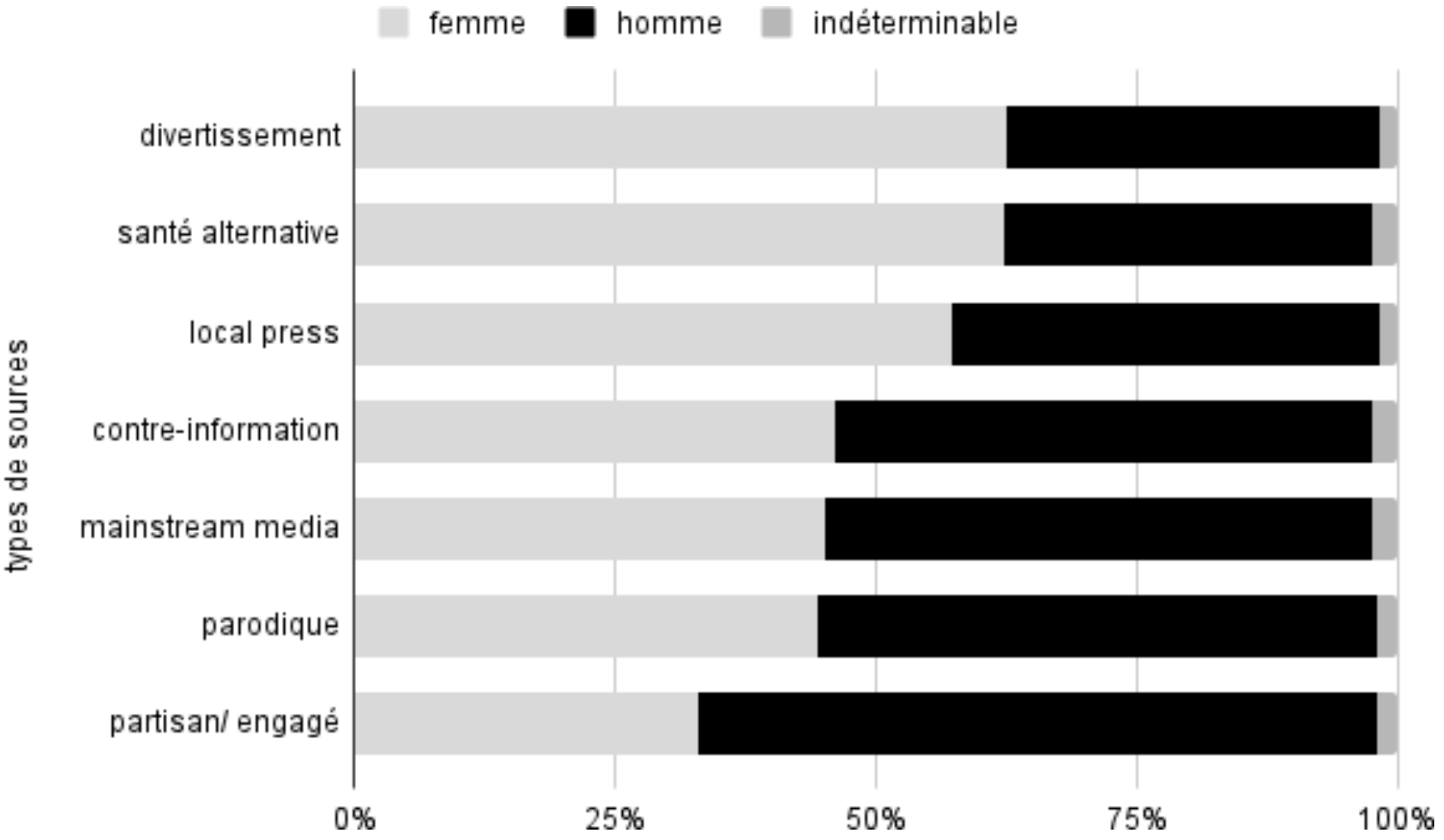

*Figure 6.3. Proportions de points d'arrêt exprimés par des hommes ou des femmes selon les sources des contenus*

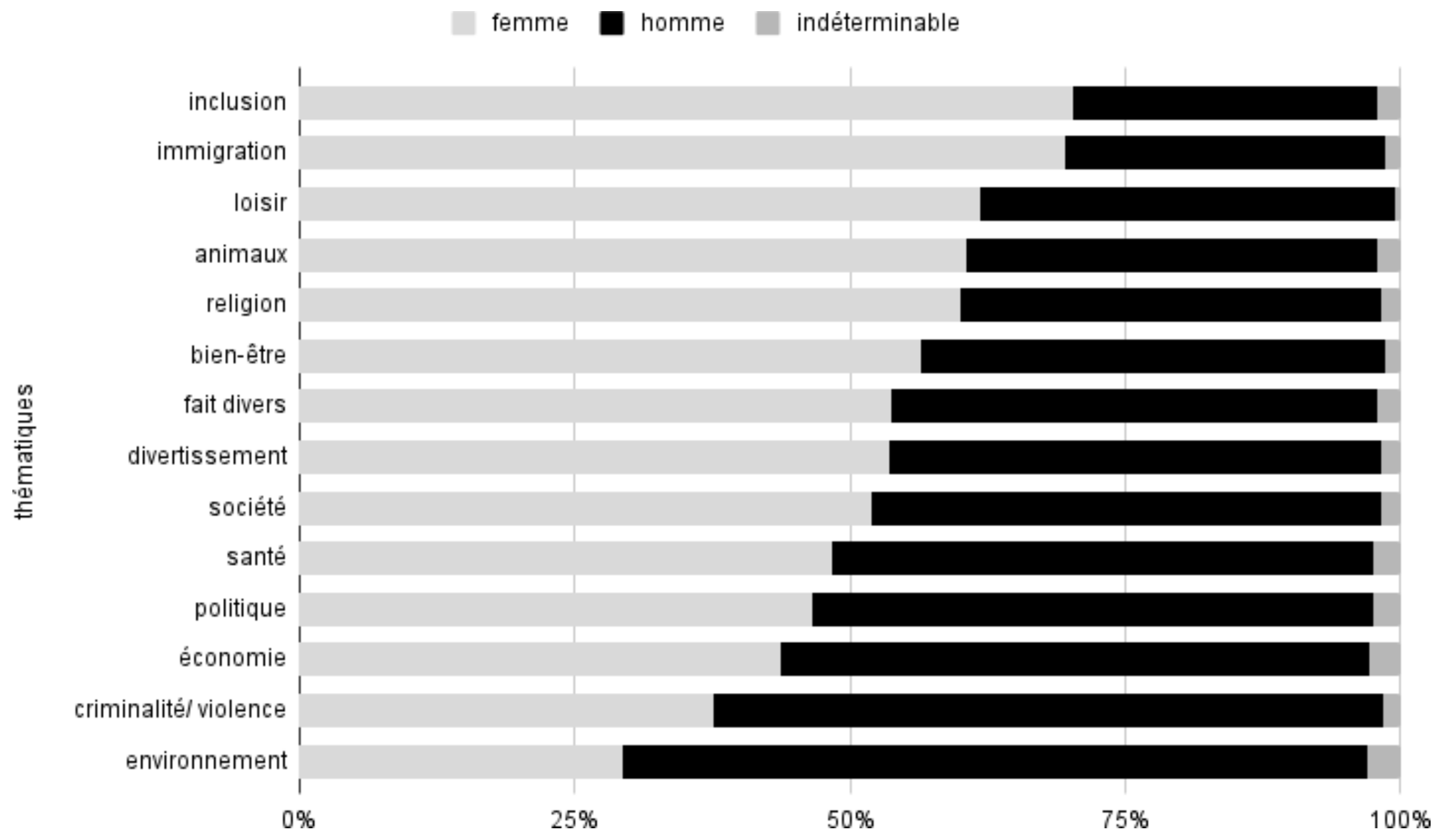

*Figure 6.4. Proportions de points d'arrêt exprimés par des hommes ou des femmes selon les thématiques des contenus*

Dans la même veine, que les entretiens, les observations sur les profils des utilisateurs font ressortir que l'expression des points d'arrêt est souvent le fruit d'individus assez politisés, bien qu'il n'ait été possible de déterminer la position politique des comptes que dans un cas sur cinq. Ces pratiques toutefois sont loin d'être l'apanage exclusif d'utilisateurs militants. Elles proviennent également d'internautes plus ordinaires partageant majoritairement sur leur compte des contenus éloignés des thématiques politiques, tels que des sujets liés aux animaux, au sport (comme le football ou le basket), ou encore à la musique. Bien que ces utilisateurs n'aient pas toujours le profil d'un citoyen politiquement engagé, et qu'ils ne soient pas forcément très diplômés, cela ne les empêche pas d'être en mesure de faire preuve d'un certain sens de la formule leur permettant d'intervenir de manière percutante dans les discussions en ligne. Par exemple, un jeune homme d'une vingtaine d'années, manifestement amateur de rap, est intervenu plusieurs fois de manière concise et rythmée avec des formules telles que « bien vu, bien dit, bien trouvé » ou encore « la meilleure manière d'y répondre c'est avec le silence ». Ce type de commentaire révèle une capacité à s'approprier les codes de communication propres à l'espace numérique.

Il apparaît donc que, même si les points d'arrêt sont principalement le fait d'une minorité d'utilisateurs, leur expression s'étend à une diversité de profils, y compris des individus moins politisés ou moins diplômés. Il convient de noter que cette minorité diffère de celle des relayeurs de *fake news*, notamment en termes d'âge, avec une prévalence plus marquée des jeunes adultes parmi ceux qui expriment des points d'arrêt, alors que Facebook est de plus en plus utilisé par des générations plus âgées.

Bien que ces observations reposent sur des indices encore limités et nécessitent des recherches supplémentaires pour être confirmées, elles sont en phase avec les résultats d'études antérieures basées sur des données déclaratives ou des questionnaires, ainsi qu'avec une large part de la littérature existante.

### 6.1.2. Rareté des réponses et des réactions

Bien que l'expression de points d'arrêt soit le fait d'une infime minorité d'utilisateurs sur Facebook, il est possible que ces « super-participants » jouent un rôle discursif positif en encourageant la participation de nouveaux utilisateurs ou en facilitant les débats (Albrecht, 2006 ; Graham et Wright, 2014 ; Kies, 2010). Aussi, est-il pertinent de s'interroger sur les effets que peut avoir l'expression de points d'arrêt sur les dynamiques de participation en ligne. Favorisent-ils le déroulement de discussions ? Ou bien, au contraire, engendrent-ils des mécanismes d'*exit* ou de silenciation ? (Hirschman, 1970).

Pour apporter des éléments de réponses à ces questions, deux indicateurs ont été calculés à partir des commentaires de premier niveau du corpus (i.e. en excluant les réponses) : (1) la proportion de points d'arrêt ayant reçu des réponses et des réactions ; (2) la moyenne de réponses et de réactions déclenchées par chaque point d'arrêt. Ces indicateurs ont ensuite été comparés avec ceux obtenus pour les autres types de commentaires.

Les résultats indiquent que la majorité des points d'arrêt exprimés sous un post Facebook reste sans réponse et ne déclenche pas de réactions (i.e. de *like* ou d'émoji). En effet, parmi les 32 400 points d'arrêt de premier niveau du corpus, seulement 5 379 ont reçu au moins une réponse, soit 16,6 % ; et 14 451 ont suscité au moins une réaction, soit 44,6 %. Ces proportions sont plus faibles que celles des autres commentaires du corpus, puisque 22 % et 53,7 % d'entre eux ont respectivement déclenché au moins une réponse ou une réaction.

Ce constat d'un faible taux de réponse est partagé par les utilisateurs interrogés. Plusieurs d'entre eux ont indiqué ne recevoir que très rarement des réactions ou des réponses après avoir laissé un commentaire critique sur Facebook. Certains ont également précisé qu'ils ne cherchaient pas à déclencher de discussions ou de débats en commentant un post, et que même si cela arrivait il ne répondrait pas forcément.

> *Je n'attends aucune interaction. Je n'attends rien [...] et même s'il y en a ce n'est pas dit que je réponde à la réaction qu'il y a eu quoi. Je m'en fous un peu en fait. J'ai dit ce que j'avais à dire. Bon ben soit une discussion s'entame mais pourquoi pas on peut discuter. Ça m'est déjà arrivé [...] mais c'est plutôt rare, franchement c'est plutôt rare. En général, j'écris ce que j'ai à dire, tout le monde s'en fout et c'est peut-être bien mieux comme ça et puis voilà quoi. Regardez le truc là que j'ai écrit, ouais y a des gens qui ont mis des like, des coeur, tout ce que vous voulez, personne n'a réagi de façon écrite, ça prouve bien que tout le monde s'en fout.* [197]

Par ailleurs, le nombre moyen de réponses et de réactions reçues par chaque point d'arrêt est très faible, respectivement de 0,6 et de 2,5. Ce nombre est néanmoins légèrement plus élevé que celui des autres commentaires, ce qui suggère que, lorsqu'ils suscitent des réponses, les points d'arrêt génèrent des discussions plus longues que les autres types de commentaires.

Enfin, les caractéristiques socio-démographiques des utilisateurs qui répondent et réagissent aux points d'arrêt n'ont pas pu être analysées, mais en prenant appui sur les travaux sur la participation en ligne cités dans la section précédente, on peut faire l'hypothèse que les utilisateurs qui répondent à des points d'arrêt partagent des caractéristiques socio-démographiques similaires à ceux qui en expriment ou qui rédigent des commentaires sur les réseaux sociaux.

[197] Homme, 41 ans, Bac, manager, entretien réalisé le 8 novembre 2022.

En résumé, ces différents résultats suggèrent que les points d'arrêt exprimés sur les réseaux sociaux ne favorisent pas forcément la participation en ligne des autres utilisateurs. En effet, ces interventions critiques sont globalement moins susceptibles de susciter des réponses et des réactions que les autres types de commentaires. Il ressort cependant que les points d'arrêt qui reçoivent des réponses génèrent des discussions plus longues et déclenchent des réactions plus nombreuses que les autres types de commentaires. Il est important toutefois de ne pas interpréter trop hâtivement une absence de réponse comme un signe négatif ; tout comme un nombre élevé de réponses n'est pas forcément positif.

Le fait de ne pas réagir à un point d'arrêt ne reflète pas forcément un manque d'intérêt, un défaut d'attention ou une appréhension à entrer dans une discussion, mais peut au contraire suggérer une forme d'*épochè* de la part des utilisateurs, c'est-à-dire une suspension temporaire de leur jugement leur permettant de prendre le temps de réfléchir plutôt que d'exprimer de façon immédiate leur point de vue. D'autre part, la longueur d'un échange ne préjuge en rien de sa qualité. Au contraire, comme l'énonce la loi de Godwin, plus une discussion se prolonge, plus elle peut facilement dériver vers des débats stériles ou polarisés, où l'échange d'arguments et d'opinions contradictoires laisse place à des attaques personnelles ou à une escalade des tensions. Un nombre élevé de réponses à un point d'arrêt n'est donc pas synonyme d'un débat constructif ou pluraliste.

Avant de tirer des conclusions sur le faible nombre de réponses obtenues par les points d'arrêt, il est donc crucial d'examiner plus en détail la nature de ces réponses. La section suivante propose justement de se consacrer à cette analyse afin de mieux comprendre les dynamiques conversationnelles qui entourent l'expression des points d'arrêt.

## 6.2. Entre dialogues de sourds et écoute silencieuse

La plupart des enquêtes portant sur la manière dont les corrections apportées par des utilisateurs sur les réseaux sociaux sont perçues par les autres reposent sur des méthodes expérimentales ou des questionnaires, et s'inscrivent dans un cadre d'analyse orienté vers l'idéal d'un espace public délibératif. Autrement dit, ces travaux tendent à évaluer l'efficacité des corrections en fonction de leur capacité à convaincre les autres utilisateurs. Une correction est ainsi considérée comme réussie si elle parvient à modifier l'opinion de ceux qui y sont exposés (Vraga et Bode, 2017 ; 2018 ; Margolin et al., 2018 ; Bode et al., 2020 ; Pasquetto et al, 2022 ; Badrinathan et Chauchard, 2024), et comme inefficace si elle échoue à entraîner un changement d'avis ou provoque des réactions négatives (Masullo et Kim, 2021 ; Mosleh et al., 2021b ; Heiss et al., 2023).

Ce type d'approche, bien qu'utile pour mesurer l'impact direct des corrections, reste néanmoins focalisé sur une vision normative de l'espace public en ligne. Il présente les utilisateurs qui corrigent les autres sur les réseaux sociaux comme des « citoyens rationnels », comme si leurs corrections étaient nécessairement « justes » et devraient convaincre les autres utilisateurs, et comme si les réactions de ces derniers se limitaient à un simple accord ou désaccord. Or, présupposer que l'objectif principal des interactions est de parvenir à un consensus rationnel peut occulter l'intérêt d'autres formes de réponse telles que la résistance, l'humour, le rejet, l'ironie ou l'indifférence. De plus, comme cela a été mentionné dans le chapitre précédent, les méthodes expérimentales et questionnaires mobilisés par ces enquêtes ne permettent pas de reproduire de manière authentique les critiques exprimées par les utilisateurs sur les réseaux sociaux, ni de capturer les réactions naturelles et spontanées qu'elles déclenchent.

Plutôt que d'évaluer les effets des points d'arrêt avec un cadre d'analyse uniquement orienté vers l'idéal d'un espace public délibératif (Stromer-Galley, 2007 ; Greffet et Wojcik, 2008 ; Monnoyer-Smith, Wojcik, 2012 ; Wright, 2016), l'enjeu de cette section est d'examiner attentivement et de décrire finement les réactions des utilisateurs face à des points d'arrêt sans partir de catégories déjà constituées ni d'horizon normatif particulier. L'idée est de s'en remettre aux autres utilisateurs pour juger du caractère approprié ou inapproprié d'un point d'arrêt.

### 6.2.1. Pluralité des types de réponses et de réactions

Quels types de réponses et de réactions déclenchent les points d'arrêt sur les réseaux sociaux ? Sont-elles plutôt favorables ou défavorables aux points d'arrêt ? Ceux-ci vont-ils déclencher des échanges interactifs entre plusieurs utilisateurs ? Si oui, ces échanges s'apparentent-ils alors à des débats, des discussions ou des disputes ?

Afin d'apporter des éléments de réponses à ces questions, les analyses de cette section proposent de se concentrer sur les points d'arrêt de premier niveau qui ont suscité des réponses. Parmi les 5 379 points d'arrêt de premier niveau identifiés dans notre corpus, ceux-ci ont généré en moyenne 3,6 réponses chacun et ont suscité 11 réactions (*likes, angry,* etc.). Cela représente un total de 19 249 commentaires de second niveau en réponse aux points d'arrêt. Ce volume de commentaires étant beaucoup trop important pour conduire des analyses fines et rigoureuses, nous avons décidé de limiter notre analyse à l'échantillon de commentaires déjà annotés manuellement dans le chapitre précédent (cf. section 5.1.2). Sur les 4 306 points d'arrêt identifiés, 2 521 constituent des commentaires de premier niveau. Parmi eux, 442, répartis au sein de 189 fils de discussion, ont reçu des réponses, soit 17,42%. Au total, ceux-ci ont généré 2 055 réponses.

Les 2 055 réponses émises en réaction à des points d'arrêt déjà annotées ont été relues et analysées. Bien qu'aucune grille d'annotation formelle n'ait été utilisée, quatre types principaux de réponses ont pu être dégagés : (1) les réponses positives ; (2) les réponses négatives ; (3) les réponses mobiles ; (4) les réponses obliques.

#### *Entre approbation, remerciement, consolidation et surenchère*

Un premier ensemble de réponses correspond à des réactions positives par rapport aux points d'arrêt exprimés en commentaires. Dans la majorité des cas, il s'agit de simples marques d'approbation où les utilisateurs se contentent d'exprimer leur accord avec des remarques telles que « bien dit », « 100 % d'accord » ou « c'est bien vrai ». Ces réponses se limitent souvent à des formules brèves et immédiates, traduisant un soutien direct à la critique formulée. Dans certains cas, plus rares, il arrive que des utilisateurs apportent des arguments ou des preuves supplémentaires pour consolider la critique formulée par un point d'arrêt. Dans d'autres cas, enfin, les réponses relèvent plutôt de la surenchère ou de la complicité moqueuse. Par exemple, des commentaires comme « beaucoup sont débiles et croient tout ce qui se dit aux infos ... pas de cerveau .... pour se poser des questions ... !!!! » manifestent un effet de troisième personne où les utilisateurs se moquent collectivement de la crédulité présumée de certains groupes sans chercher véritablement à entrer en dialogue avec des opposants. Ces interactions témoignent d'un désir de renforcer les liens avec des pairs partageant les mêmes opinions, plutôt que de favoriser le débat et révèlent que l'expression de points d'arrêt, loin d'initier un véritable échange, sert souvent à créer une forme de cohésion au sein d'un groupe d'opinions similaires, en excluant ou ridiculisant ceux perçus comme des opposants. Certains utilisateurs réagissent probablement en voyant ces commentaires apparaître sur leur fil d'actualité, ce qui déclenche des moqueries partagées et une validation sociale des points de vue exprimés. Ainsi, plus qu'à entrer en dialogue avec leurs opposants, ceux qui expriment des points d'arrêt cherchent sans doute l'approbation de leurs pairs.

***Entre désapprobation, injure, contre-argumentation et moquerie***

Un deuxième ensemble de réponses s'apparente à des réactions négatives par rapport aux points d'arrêt exprimés en commentaires. Dans de nombreux cas, les utilisateurs expriment leur désaccord avec des formules telles que « vous vous trompez » ou « c'est faux ». Parfois, ils accompagnent leur désapprobation de nouveaux arguments ou preuves pour contre-argumenter la critique initiale. Toutefois, ces échanges sont surtout souvent marqués par une forte dose de tension, et un certain nombre de commentaires se traduisent par des insultes ou des propos agressifs. La plupart des enquêtés interrogés ont d'ailleurs indiqué avoir reçu plutôt des oppositions véhémentes que des soutiens ou des remerciements :

> *Donc dès qu'on contre-argumente avec quelqu'un, on sait qu'on va se faire grosso modo agressé, voire menacé de morts en message privé. C'est très très courant. Ou bien si on met un* *truc favorable sur la vaccination, on sait qu'il y a des gens qui vont répondre pour nous dire qu'on est tantôt des trolls de l'industrie pharmaceutique, qu'on a des conflits d'intérêt, etc, etc.*[198]

*Par défaut je pense que les gens qui trouvent un argument intéressant, ils ne le relèvent pas trop, c'est-à-dire qu'il y en a de temps en temps qui likent ou qui disent merci pour les explications mais la plupart du temps on a plutôt des réactions contre.*[199]

*Malheureusement ça ne marche pas. Ces personnes ça ne marche pas. J'ai encore un exemple récent. Alors là ça ne parle pas de racisme, ça parle du covid. Euh d'une contact que j'ai qui est une femme qui est complètement dans la sphère de manipulation par le covid, enfin voilà anti-vax et tout, et quand il y a eu les manifestations contre le pass vaccinal, elle avait juste mis une photo du genre « eh on nous dit qu'il y a personne pour la manif à Paris » et en fait la photo elle était prise du haut des champs élysées et on voyait une foule colossale. Et en fait j'ai juste fait une recherche inversée et en trois seconde je suis tombée sur le fait que cette photo ben en fait elle venait d'un article de 2018 où la France avait gagné la coupe du monde de Foot et je lui ai juste renvoyé pour lui dire « ben en fait la photo elle n'est pas de la manif » mais elle s'est énervée. Quand je fais ça, on est sur des gens qui s'énervent. C'était la même chose avec des contacts du club. C'est des gens qui nous détestent quand on leur met la preuve sous le nez qu'ils se sont trompés sur un truc.*[200]

Enfin, les points d'arrêt peuvent être également l'objet d'un autre type de désapprobation visant non pas à contester leurs objections épistémiques mais leur caractère inapproprié par rapport au contexte d'interactions. On peut alors dire que l'expression d'un point d'arrêt est considérée comme une « faute de grammaire ». Le plus souvent, ce type de réprobation est émis sur un ton oscillant entre une ironie moqueuse et une indulgence amusée. Les utilisateurs reprochent à l'internaute à l'origine d'un point d'arrêt de manquer d'humour ou de second degré. Parfois, cela est fait sur le ton de l'humour, ce qui permet de réorienter la discussion vers un régime de légèreté et de familiarité sans rentrer dans le conflit. Ce type de réprobation taquine a principalement été observée sur les pages d'humour et de divertissement, en réaction à des contenus parodiques ou de sources sensationnalistes. Par exemple, ci-dessous (cf. Figure 6.5), l'auteur du post répond de façon ironique à l'utilisateur lui ayant adressé un point d’arrêt. Alors que ce dernier lui demande de vérifier les sources des informations qu’ils partagent, l’auteur du post lui répond « Source moutarde lolll ». Le jeu de mot qu’il fait entre les mots « source » et « sauce » lui permet de faire écho à la fois au contenu initial qu’il a partagé sur McDonald’s et au point d’arrêt émis par un utilisateur et ainsi d’indiquer de façon subtile qu’il est conscient du manque de fiabilité de la source mais que sa publication vise davantage à divertir qu’à informer. Ce type de réponse sert à repositionner la conversation sur un mode plus léger, tout en marquant la distance entre l’intention de l’auteur et la critique de l’utilisateurs à l’origine d’un point d’arrêt.

[198] Homme, 44 ans, Bac +8, médecin, entretien réalisé le 2 août 2022.
[199] Homme, 24 ans, Bac +5, étudiant, entretien réalisé le 29 août 2023.
[200] Homme, 34 ans, Bac, accompagnant éducatif et social, entretien réalisé le 1er août 2022.

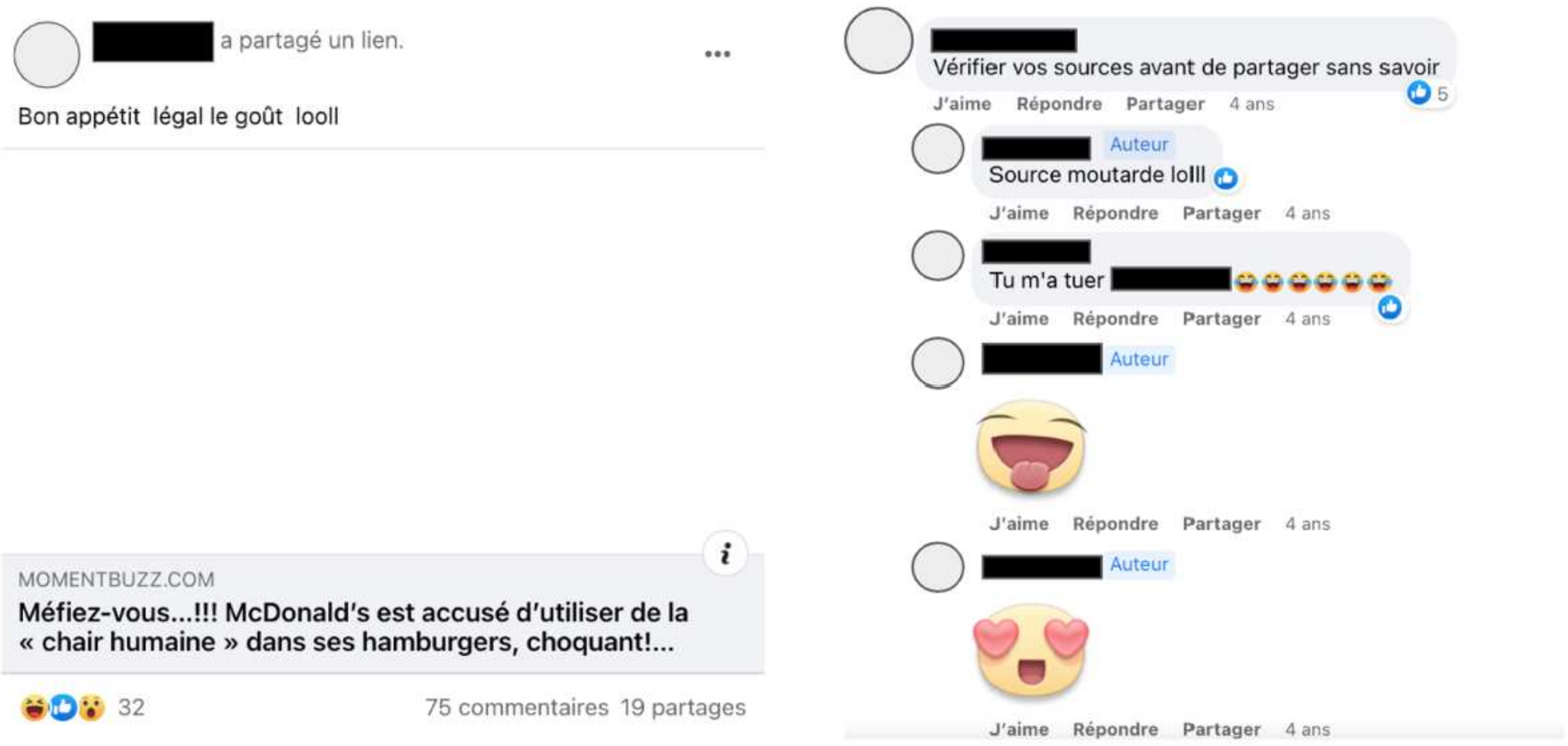

*Figure 6.5. Capture d’écran d’un échange sur une page de divertissement*

### ***Entre question, précision, concession et nuance***

Un troisième ensemble de réponses renferme des réactions que l’on peut qualifier de mobiles dans la mesure où les utilisateurs qui les émettent font un « pas vers l’autre », par exemple en posant une question, ou indiquent qu’ils sont prêts à faire évoluer leur opinion, par exemple en ayant recours à des marqueurs de concession. Il est particulièrement important de relever quand ces réponses sont émises par l’auteur du post ou du point d’arrêt. En effet, cela permet de comparer le contenu de la réponse avec la position formulée au départ. Cela donne ainsi des indications sur le degré de délibération ou d'agonisme dans l'échange. Au fil des analyses, il a été très rare de constater qu’un auteur reconnaisse son erreur et remercie pour la correction apportée. En revanche, on a observé davantage de réponses où l'auteur nuance ses propos initiaux ou concède un bon point à un contre-argument, montrant une certaine évolution de son opinion. Le témoignage de Thibaud, membre du collectif #JeSuisLà confirme que ce second cas est plus fréquent, bien que, dans les faits, il reste peu courant dans les commentaires observés.

> *Souvent la personne ne va pas dire « bon ben je change d'avis », mais peut tempérer son avis et dire « ah oui c'est vrai que bon ça je vous le concède » ou alors on va avoir une discussion un peu animée et la personne va dire « bon bah merci pour la discussion » ou alors c'est moi qui vais le dire et c'est vrai que là déjà on a fait un pas vers l'autre.*[201]

***Digressions conversationnelles***

Enfin, une quatrième série de réponses peuvent être qualifiées d'« obliques » dans le sens où elles s'éloignent du point d'arrêt initial pour faire bifurquer l'échange vers une forme de conversation « privée-publique ». Ces réponses se distinguent par leur ton léger, souvent humoristique, et se manifestent à travers des blagues, des *private jokes* ou des échanges empreints d'affection entre des participants qui semblent se connaître. Les utilisateurs utilisent fréquemment des gifs, des emojis ou des références culturelles partagées pour réorienter la discussion vers une interaction plus détendue et personnelle. Ces réponses obliques participent ainsi à la création d'une connivence entre certains membres, en transformant un débat sérieux en un moment de complicité sociale, tout en atténuant les tensions initiales soulevées par le point d'arrêt.

Ce travail de description a permis d'approximer la prévalence de différents types de réactions. Aucune quantification rigoureuse n'a cependant pu être réalisée car plusieurs commentaires correspondent à plusieurs types et il est difficile de rassembler dans une même catégorie des commentaires très disparates les uns des autres. En fait, dans un premier temps, il est particulièrement utile et intéressant de suivre la dynamique des échanges en prêtant attention au rôle et au statut des intervenants : est-ce l'auteur initial du post qui réagit ?

[201] Homme, 38 ans, Bac +5, consultant, entretien réalisé le 21 mars 2023.

D'autres utilisateurs ? L'auteur du point d'arrêt intervient-il à nouveau ? Suivre ces interactions permet d'observer si des ajustements progressifs ont lieu dans les échanges et surtout comment les utilisateurs parviennent (ou pas) à créer des espaces communs d'intercompréhension.

Par exemple, sur un groupe dédié à la vie locale d'une commune située dans le département de l'Isère, en région Auvergne-Rhône-Alpes, le contenu parodique ci-dessous a été publié par Rebecca une habituée du groupe qui en est aussi modératrice (cf. Figure 6.6). Un premier point d'arrêt dirigé vers la source est tout d'abord exprimé avec un émoji tirant la langue pour indiquer un ton complice : « Sciences Info est un site de fake news :p ». Puis, d'autres commentaires, émis sur un ton plus sérieux et parfois moralisateur, s'ensuivent : « Faut pas croire tout ce que vous voyez sur le net » ou encore « Avez-vous lu l'article avant de le publier ? Sûrement pas, car vous auriez vite compris qu'il s'agit d'un canular ». Ces remarques sérieuses émanent d'internautes qui, bien qu'ils comprennent le caractère humoristique du post et en rient entre eux (en témoignant par des réactions sous forme d'émojis ou de gifs), s'inquiètent de la crédulité des autres membres du groupe. Rebecca, l'utilisatrice à l'origine de la publication dans le groupe intervient alors pour expliciter son intention : « L'article est drôle, je partage pour qu'on décompresse avant la semaine qui redémarre ! ». Son intervention reçoit le soutien de nombreux membres du groupe qui semblent bien la connaître, dans la mesure où Rebecca leur répond « vous êtes les seules à me connaitre <3 », illustrant ainsi comment la proximité affective – même virtuelle – permet de mieux interpréter les intentions des autres et d'assouplir le langage. À l'inverse, pour les liens faibles, cette compréhension mutuelle se révèle plus difficile comme l'admet l'une des utilisatrices ayant formulé initialement d'un point d'arrêt : « Désolé, mais je ne vous connais pas et je ne connais pas forcément votre humour donc vous pouvez comprendre que ça m'a surpris ce genre de publication que certaines personnes peuvent prendre au 1er degré en voyant le titre ». Rebecca *like* son commentaire et lui répond : « et bien enchantée madame :) oui je vous comprends effectivement. C de l'humour et j'aime le côté décalé de l'article dans ce monde actuel où la parano réussit à enterrer le bon sens! ».

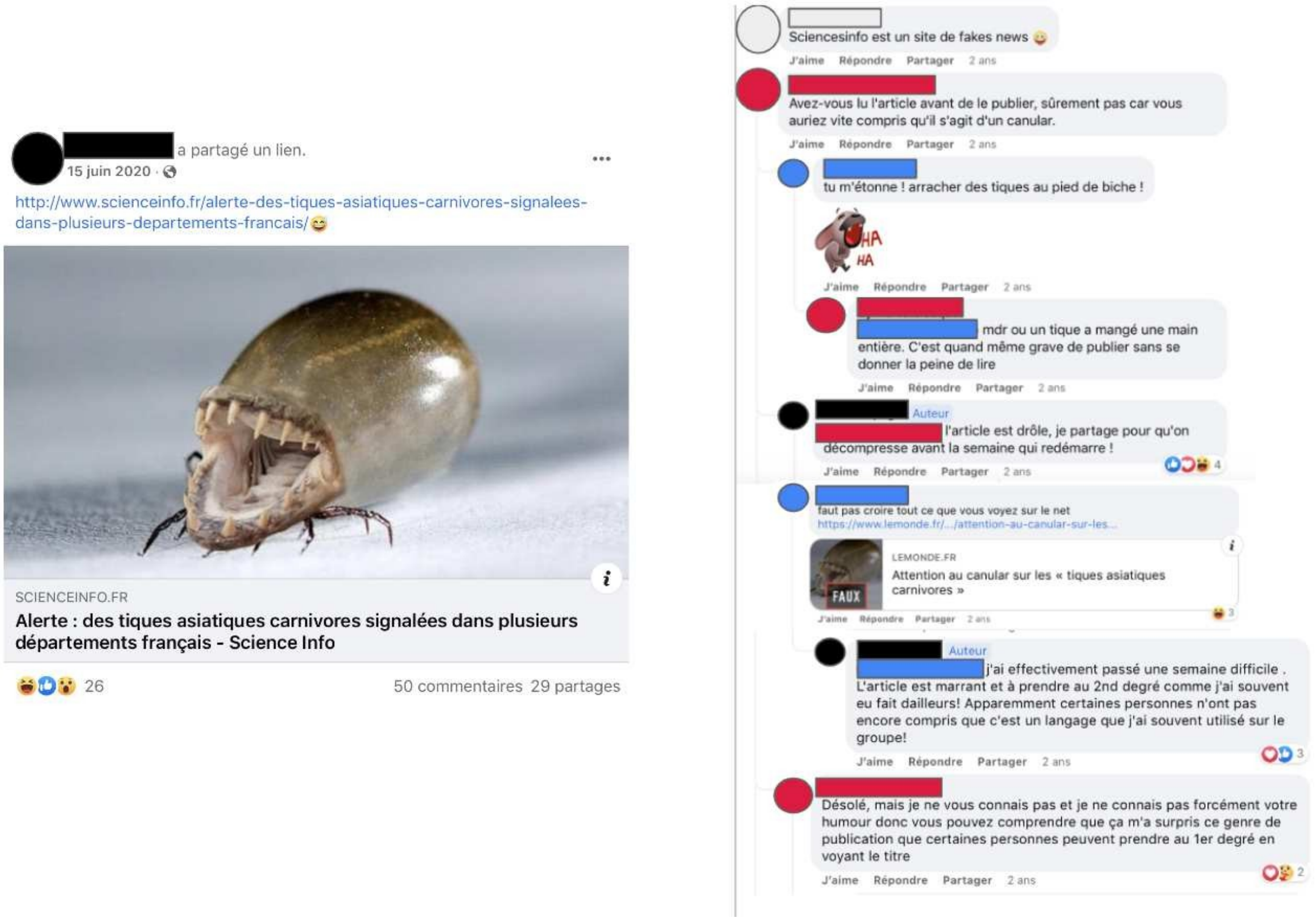

*Figure 6.6. Capture d'écran d'un échange sur une page d'une commune française*

À travers ce cas, on voit ainsi comment les utilisateurs des réseaux sociaux peuvent parvenir à créer des espaces communs d'intercompréhension malgré des divergences initiales sur la façon d'appréhender la situation d'interactions.

Les différentes analyses effectuées dans cette section ont aussi permis de dégager quelques indices sur lesquels un chercheur peut s'appuyer pour examiner si un échange tend plutôt vers la délibération, la préservation de l'ordre des interactions ou la contestation agonistique. Par exemple, le fait qu'un utilisateur à l'origine d'une publication réponde à un point d'arrêt suggère l'ouverture d'un débat. De même lorsque l'utilisateur à l'origine d'un point d'arrêt intervient à nouveau dans l'échange. Le nombre de personnes différentes qui répondent au point d'arrêt permet aussi d'évaluer le degré de pluralisme de la discussion.

### 6.2.2. Entre sentiment d'à-quoi-bonisme et entêtement : mécanisme de silenciation et envenimement des échanges

Face aux nombreuses réactions négatives qu'ils reçoivent sur les réseaux sociaux, les utilisateurs qui expriment des points d'arrêt ont souvent fait part d'un sentiment de fatigue et de lassitude, teinté d'une certaine irritation. Nombreux sont ceux qui, après avoir été la cible d'insultes, d'agressions verbales ou de menaces, ont indiqué avoir fini par se retirer des échanges ou alors par adopter un ton plus emporté.

Un enquêté, par exemple, a mentionné avoir été exclu de plusieurs groupes Facebook défendant l'idée que la Terre est plate. [202] À chaque fois qu'il « formulait une critique ou un argument rationnel », il se faisait « bannir tout de suite » a-t-il expliqué. En réaction à ces exclusions répétées, il a indiqué avoir « appris à [se] taire et à rester discret ». Afin de pouvoir réintégrer certains de ces groupes sans se faire immédiatement « éjecté », il a décidé de créer plusieurs comptes sous pseudonymes. Une fois de retour dans ces groupes, cependant, son objectif n'était plus d'intervenir dans les discussions pour apporter des contre-arguments mais de mener des observations afin de comprendre les motivations profondes des autres membres. Ce comportement témoigne d'un processus de silenciation, où des utilisateurs sont réduits au silence, non par conviction, mais par peur de représailles ou par lassitude face à des échanges qu'ils perçoivent comme stériles et agressifs. Ce retrait n'est pas un désengagement total, mais une stratégie de protection, permettant de rester en position d'observateur critique tout en évitant les confrontations directes.

En opposition à cette attitude de retrait, certains utilisateurs continuent d'intervenir dans des groupes de discussion malgré l'inutilité perçue de l'effort argumentatif. Cette persistance s'accompagne souvent d'un durcissement du ton, comme l'expliquent ces deux enquêtés :

> *Mais avec le temps, je dois admettre que ma capacité à débattre sereinement s'est érodée, et que la rhétorique chargée d'invective oscillant entre le passif-agressif et l'agressif-agressif – que*

[202] Homme, 47 ans, Bac +5, ingénieur, entretien réalisé le 12 août 2024.

*les spécialistes nomment « Syndrome du conducteur parisien » – a remplacé le discours équilibré de logos, pathos, ethos.* [203]

*Au début, j'y suis allé pas en colère. J'étais plutôt dans une optique pédagogue en disant « oui le tétanos n'est pas une maladie immunisante mais le vaccin est efficace » par exemple ou des choses comme ça. Un certain nombre de choses très basiques quoi : par exemple, si on vaccine les enfants contre l'hépatite b c'est parce que les enfants font beaucoup plus de formes chroniques et donc sont beaucoup plus à même de faire des cancers à terme. Enfin voilà, j'étais vraiment dans une optique plutôt pédagogue et bon là j'ai pris, là les coups ont plu. Quand on rentre là dedans, on se rend compte qu'il y a un niveau d'agressivité qui est quand même assez stupéfiant.* [204]

En réalité, plutôt que de faire ressortir une opposition binaire entre ceux qui choisissent de persister (*voice*) et ceux qui préfèrent se mettre en retrait (*exit*), les entretiens ont plutôt mis en lumière chez de nombreux enquêtés une ambivalence subtile entre le désir de ne plus s'engager dans des débats vains et le besoin de continuer à corriger des informations erronées. Par exemple, au cours d'un entretien, Bastien a déclaré : « Plus ça va et moins j'ai envie de perdre mon temps à critiquer ce genre de choses ». Quelques minutes plus tard, cependant, lorsqu'il prend conscience du nombre de commentaires qu'il a laissés sous un post contre lequel il a exprimé un point d'arrêt, il s'étonne :

*Ha bon ? J'en ai mis plusieurs ? Ah oui effectivement, j'ai commenté des… Je devais être énervé ce jour-là. C'est des choses que je ne fais absolument plus parce que je n'ai absolument plus de temps à perdre dans ce genre de choses et puis surtout c'est de la polémique qui n'apporte rien. Je répondais à des gens, à chaque personne, mais typiquement ce genre de chose, j'essaie de ne plus le faire… Enfin si je l'ai encore fait la semaine dernière mais on va dire que je ne rentre pas… Je le fais de temps en temps mais voilà j'essaie de pas perdre de temps à faire ça parce que la personne n'a généralement pas l'intention de changer d'avis et moi non plus donc ça n'a pas trop d'intérêt.* [205]

---

[203] Homme, 32 ans, Bac +8, enseignant, entretien réalisé le 1er août 2022.
[204] Homme, 44 ans, Bac +8, médecin, entretien réalisé le 2 août 2022.
[205] Homme, 45 ans, Bac +4, ingénieur du son, entretien réalisé le 16 janvier 2023.

La réponse de Bastien est intéressante car elle illustre une contradiction interne entre ce qu'il pense devoir faire (se retirer des débats) et ce qu'il continue néanmoins à faire (réagir et commenter). Tout se passe comme si sa volonté de cesser de participer à des échanges en ligne se heurtait à chaque fois à une impulsion irrépressible de contre-argumentation. Son affirmation « je ne le ferais plus » suivie de l'aveu « Enfin si je l'ai encore fait la semaine dernière » témoigne d'une rationalisation postérieure : il se justifie après coup, conscient que son comportement contredit ses intentions.

Cet entêtement à rester engagé dans une discussion jugée stérile peut être illustré par le mème *Someone is WRONG on the Internet*, créé par Randall Munroe dans sa bande dessinée xkcd. Celui-ci montre un personnage incapable de se déconnecter d'une discussion en ligne, malgré l'heure tardive, parce qu'il est convaincu qu'une personne a tort (cf. Figure 6.7).

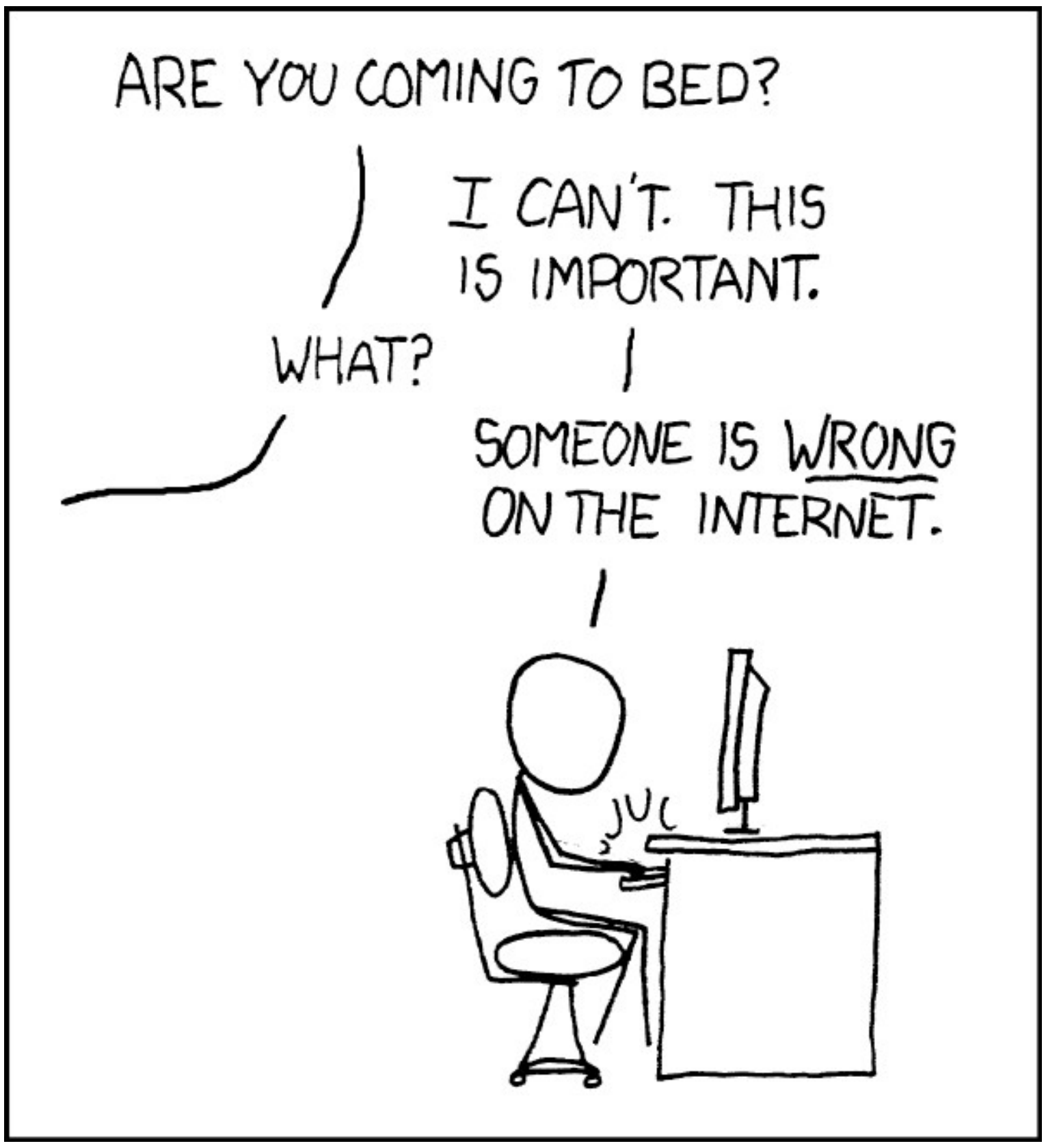

*Figure 6.7. Mème tiré de la bande dessinée xkcd de Randall Munroe, publié sous licence Creative Commons BY-NC 2.5. Source originale : https://xkcd.com/386/*

Cette attitude ambivalente, consistant à rester engagé dans un débat sans fin tout en étant conscient d'avance qu'il sera infructueux soulève une question paradoxale : pourquoi certains utilisateurs des réseaux sociaux persistent-ils à contre-argumenter alors qu'ils savent d'avance qu'ils ne vont pas réussir à convaincre leur interlocuteur ?

Ce mécanisme relève de ce que Marc Angenot (2008) décrit dans son ouvrage *Dialogues de sourds* comme un échange où les participants ne s'efforcent pas réellement de convaincre leurs opposants directs, mais plutôt de justifier et de valider leur position auprès d'un auditoire universel (Perelman et Olbrechts-Tyteca, 1988). L'argumentation vise alors à obtenir l'approbation d'un tiers spectral, un arbitre qui serait capable de reconnaître la légitimité, la rationalité, la morale et la justice du discours du locuteur contre celui de son adversaire. Cette analyse permet d'expliquer pourquoi, malgré la stérilité apparente des débats, les participants continuent à argumenter : ils ne cherchent plus à convaincre leur adversaire, mais à gagner l'adhésion d'un public plus large et invisible. Ainsi, les interactions en ligne se rapprochent davantage de monologues croisés que de véritables débats, chaque participant s'adressant à une audience invisible, ce qui peut contribuer à l'escalade des tensions.

### 6.2.3. S'adresser aux gens qui doutent ou n'interviennent pas dans les discussions

Seule une minorité d'utilisateurs publie régulièrement des commentaires en ligne, et parmi eux, seulement une petite proportion exprime des points d'arrêt ou y réagit. Toutefois, il ne faut pas oublier qu'à côté de ces utilisateurs actifs se trouvent également de nombreux *lurkers,* c'est-à-dire des personnes qui lisent les échanges qui se déroulent en ligne mais n'y contribuent pas de manière visible (Falgas, 2017). Or, c'est précisément à ces utilisateurs invisibles et silencieux que plusieurs enquêtés ont indiqué s'adresser lorsqu'ils interviennent publiquement sur les réseaux sociaux pour exprimer des critiques ou des désaccords face à des contenus qu'ils jugent problématiques :

*Mon idée n'est pas de convaincre les gens directement avec qui je parle parce que je sais que je parle avec les gens les plus extrêmes souvent dans… Je n'ai pas envie de dire dans les idées parce que ce n'est pas toujours le cas mais les plus extrêmes dans la méthode de pensée, dans la façon de s'exprimer. Du coup, je trouvais ça intéressant d'avoir des échanges avec ces personnes pour les autres personnes qui peuvent lire. Je sais qu'il y a aussi potentiellement pas mal d'autres personnes qui lisent.* [206]

Contrairement aux utilisateurs les plus actifs et expressifs, les individus silencieux sur les réseaux sociaux sont souvent moins politisés et ont des opinions moins tranchées (Scheufele et Nisbet, 2003 ; Bail et al, 2018). Ainsi, lorsqu'ils sont exposés à des contenus contestés par des points d'arrêt, il est possible qu'ils s'interrogent sans pour autant exprimer ouvertement leurs doutes. *A contrario*, les individus qui formulent des commentaires de soutien à des informations contestées sont déjà très probablement convaincus par l'opinion qu'elles véhiculent et donc moins susceptibles d'être amenés à se poser des questions face à un point d'arrêt.

*Il y a des personnes qui sont hyper militantes et là on ne peut pas trop argumenter. Mais celles qui doutent… Je trouve que ce qui est intéressant c'est de trouver celui qui doute mais à raison, c'est-à-dire raisonnablement. Qui doute pour des raisons légitimes, qui a besoin de comprendre choses, qui veut pas qu'on lui impose. Ça, ça m'intéresse plus que de me confronter aux militants… La discussion avec les militants ne m'intéresse pas dans le sens où je ne cherche pas à les faire changer d'avis.* [207]

Cette idée de s'adresser à la « majorité silencieuse » fait écho aux actions de la communauté #jesuislà. Plutôt que de tenter de convaincre ceux qui publient des fausses informations ou des discours de haine, les membres de #jesuislà visent à atténuer l'impact de ces contenus sur les utilisateurs indécis. Ils espèrent que leurs interventions influenceront les lecteurs silencieux ou les encourageront à intervenir eux-mêmes. Une étude sur ce mouvement en Suède a montré que cette stratégie permet de favoriser l'émergence de nouveaux contre-discours (Buerger, 2021), en brisant les mécanismes de spirale du silence et d'ignorance pluraliste, tout en réduisant la polarisation perçue. Par exemple, une recherche sur plus de 7 500 commentaires Facebook a testé l'efficacité du contre-discours pour répondre à des représentations négatives des Roms en Slovaquie entre 2016 et 2017 (Miškolci et al., 2018). Les résultats ont montré que le contre-discours n'a pas particulièrement influencé le comportement des utilisateurs ayant posté des commentaires anti-Roms, mais a néanmoins entraîné une augmentation du nombre de commentaires pro-Roms dans le fil de discussion.

[206] Homme, 33 ans, Bac+5, *data scientist*, entretien réalisé le 9 mars 2023.
[207] Homme, 44 ans, Bac +8, médecin, entretien réalisé le 2 août 2022.

Les résultats de ces études confirment l'hypothèse émise dans la section 6.1.2, à savoir que l'absence de réaction ou la présence de commentaires négatifs ne doivent pas être interprétées comme un échec ou un signe d'inefficacité du point d'arrêt. Elles invitent à redéfinir l'objectif normatif du point d'arrêt : plutôt que de convaincre directement les utilisateurs qui partagent des contenus considérés comme des *fake news* ou adhèrent à ces contenus, il s'agit de s'adresser aux utilisateurs invisibles et silencieux. Cela pose une question centrale pour les futures recherches : comment évaluer l'efficacité du contre-discours ? Il devient crucial d'inventer des dispositifs d'enquête permettant d'observer les réactions des *lurkers*, et non seulement celles des participants visibles.

## Conclusion du sixième chapitre

Ce dernier chapitre a permis de dégager deux constats principaux. D'une part, seul un faible pourcentage d'utilisateurs exprime des « points d'arrêt ». D'autre part, ces points d'arrêt ne sont pas nécessairement synonymes d'une ouverture des espaces de communication à une pluralité d'opinions. En effet, ils traduisent souvent des répliques isolées ou des expressions de désaccords, sans pour autant structurer un échange dynamique et argumenté. Cependant, l'absence de réponse à ces points d'arrêt ne doit pas être interprétée comme une simple faiblesse du débat en ligne. Au contraire, cette absence de réciprocité ouvre des questions fondamentales pour l'étude des publics numériques : quels sens les utilisateurs attribuent-ils à leurs interventions solitaires, et comment perçoivent-ils le silence qui leur répond ? Ce constat suggère la nécessité d'enquêtes auprès des utilisateurs silencieux, souvent désignés comme une majorité passive. Comprendre leurs motivations – qu'il s'agisse de l'indifférence, de l'accord tacite, du désengagement stratégique, ou de la crainte de harcèlement en ligne – pourrait enrichir la compréhension des logiques interactionnelles à l'œuvre dans les espaces de communication numérique.

Ainsi, l'analyse de ces dynamiques silencieuses, souvent invisibles dans les études quantitatives de traces numériques, invite à envisager des pistes méthodologiques nouvelles pour appréhender l'ensemble des acteurs des débats en ligne – actifs comme silencieux. Ce silence pourrait révéler des formes latentes d'adhésion, de résistance, ou de désengagement, ouvrant la voie à une compréhension plus nuancée des « publics » numériques, en interaction constante avec l'écosystème informationnel.

# Conclusion générale

Les analyses produites à travers chacun de ces six chapitres permettent d'achever cette thèse par une conclusion générale. Après avoir synthétisé les principaux résultats issus des deux enquêtes réalisées sur Twitter et Facebook, des réponses peuvent être apportées pour expliquer les deux paradoxes qui ont guidé la rédaction de cette thèse. Ces résultats permettent de formuler deux arguments plus généraux contribuant à enrichir les conceptualisations des notions d'espace public et de rationalité. Des pistes pour de futures recherches sont ensuite proposées pour dépasser les limites de notre dispositif méthodologique. Enfin des recommandations pour le débat public sont suggérés à partir des bases conceptuelles et empiriques apportés par la thèse, ainsi que celles d'autres travaux de recherche en sciences sociales, pour reformuler le problème des *fake news* et réorienter l'action publique.

## Synthèse des résultats empiriques

Les deux enquêtes conduites dans cette thèse ont permis de faire ressortir trois résultats principaux.

Le premier résultat montre que le partage de *fake news* est davantage lié à des caractéristiques socio-politiques spécifiques, notamment à un fort degré de politisation et à une défiance importante à l'égard des élites et des institutions, qu'à des processus cognitifs reflétant un manque de raisonnement analytique. Les utilisateurs qui relaient des *fake news* sur Twitter sont en effet fortement concentrés aux extrêmes de l'échiquier politique, et très critiques à l'encontre du gouvernement et des partis politiques traditionnels. Leur langage n'est pas moins réflexif mais ils ont des pratiques militantes importantes et une activité de partages d'informations très élevée sur les réseaux sociaux qui les distinguent très nettement des utilisateurs ordinaires qui passent peu de temps à consulter des informations d'actualité et sont plutôt des *lurkers*. Les utilisateurs qui partagent des *fake news* sur Twitter sont également plutôt âgés et très éduqués. Ces constats soulignent que les utilisateurs des réseaux sociaux ne sont ni prémunis du partage de *fake news* par leurs capacités de raisonnement analytique, ni par leur intérêt pour l'information d'actualité ou leur niveau d'éducation. Ils mettent ainsi en évidence les limites de l'idéal du citoyen bien informé et rationnel (Schudson, 1998).

Le deuxième résultat indique que les utilisateurs des réseaux sociaux exposés à des *fake news* sont en mesure de déployer des compétences de distance critique. Cette capacité varie selon leur position dans l'espace social, ainsi que les contraintes énonciatives et les contrats de communication qui sous-tendent les situations d'interactions dans lesquelles ils sont amenés à recevoir ou à partager des informations. Deux compétences critiques ont été mises en évidence. Tout d'abord, une capacité à faire preuve de prudence énonciative. Celle-ci se traduit par une série de pratiques à travers lesquelles les individus intègrent dans leur énonciation une anticipation des attentes et réactions de leur public potentiel. Ensuite, une capacité à intervenir dans le flux d'un échange en ligne pour exprimer des points d'arrêt, c'est-à-dire des désaccords, corrections ou dénonciations à travers lesquels les utilisateurs font part d'un trouble et cherchent à rétablir l'ordre des interactions. Loin d'être des constantes dépendant uniquement de facteurs cognitifs, ces compétences critiques sont plutôt des pratiques situées et socialisées à travers lesquelles les publics semblent moins soucieux de la factualité intrinsèque d'un énoncé que de leur réputation et de leur intégration dans des groupes sociaux, ainsi que de la préservation de l'ordre de l'interaction.

Le troisième résultat suggère que ce n'est pas parce que la majorité des utilisateurs ne partagent pas de *fake news* sur les réseaux sociaux et que certains sont capables, dans certaines circonstances, de faire preuve de distance critique que des troubles de l'information et de la communication ne sévissent pas dans l'écosystème informationnel contemporain. En effet, bien que les utilisateurs qui partagent des *fake news* ne représentent qu'une petite minorité d'internautes, leurs discours sont potentiellement surreprésentés dans l'espace public numérique en raison de leur hyperactivité en ligne. Cet accaparement des espaces de visibilité numérique par une minorité d'internautes peut sur le long-terme contribuer à façonner les thèmes inscrits à l'agenda du débat public et ainsi influencer l'opinion publique. Par ailleurs, les pratiques de prudence énonciative mises en œuvre par certains utilisateurs visent à préserver leur réputation ou leur intégration dans des groupes sociaux mais ne garantit pas la qualité des énoncés qu'ils relaient. Certains utilisateurs peuvent avoir recours à des stratégies de prudence énonciative pour mettre en circulation des énoncés contestables sans risquer d'être sanctionnés. Enfin, les interventions dans les échanges en ligne qui ont lieu sous la forme de points d'arrêt sont rares et l'apanage d'une minorité d'utilisateurs. Loin de favoriser l'émergence de véritables débats délibératifs, ou l'expression d'un pluralisme agonistique, elles donnent plutôt lieu à des dialogues de sourds entre une minorité d'utilisateurs particulièrement actifs en ligne.

## Explications des paradoxes

Ces différentes analyses permettent d'expliquer les deux paradoxes soulevés dans l'introduction de cette thèse après avoir confronté les représentations issues des discours publics sur les *fake news* aux constats empiriques des études académiques :

(1) Pourquoi la majorité des enquêtes empiriques montre que les *fake news* ne représentent qu'une petite proportion du total d'informations consultées et partagées par les utilisateurs des réseaux sociaux alors que ces derniers ne sont ni soumis à un contrôle éditorial, ni à des règles de déontologie journalistique ?

À cette question la thèse apporte une réponse claire. Les réseaux sociaux ne constituent pas un univers de pratiques homogènes totalement dépourvu de règles. Ils sont plutôt composés d'une pluralité d'espaces, régis par différents contrats de communication et normes d'interactions, favorisant ou entravant des mécanismes d'autorégulation conversationnelle. Loin de manquer de discernement, les utilisateurs des réseaux sociaux sont dotés de capacités critiques qui leur permettent d'identifier ces contrats de communication et normes d'interactions afin d'ajuster leurs pratiques énonciatives en conséquence. Ainsi, si les utilisateurs des réseaux sociaux consomment et partagent peu de *fake news*, bien que les plateformes numériques soient dépourvues de *gatekeepers* pour leur faire barrage, c'est parce qu'il existe des mécanismes sociaux, notamment liés au fait d'appartenir à des groupes sociaux, qui leur permettent de faire attention et d'auto-réguler leurs espaces de conversations, par exemple en faisant preuve de prudence énonciative ou en exprimant des points d'arrêt.

(2) Comment comprendre la montée de la polarisation politique alors que les utilisateurs ne semblent pas si réceptifs aux *fake news* ?

Ce constat paradoxal peut s'expliquer à la fois par des mécanismes d'*agenda-setting* et de spirale du silence. Bien que la proportion d'utilisateurs partageant des *fake news* soit marginale, leur hyperactivité sur les réseaux sociaux leur permet d'occuper une place importante dans les conversations en ligne. Cette surexposition donne une visibilité disproportionnée à des opinions polarisantes, créant ainsi l'illusion que ces opinions dominent le débat public. Par ailleurs, les modes d'expression de cette minorité peuvent exacerber les clivages et contribuer à invisibiliser et à marginaliser certaines voix, notamment celles des plus modérés. Enfin, la minorité d'utilisateurs à l'origine de points d'arrêt ne favorise pas toujours l'émergence de véritables débats délibératifs, ou l'expression d'un pluralisme agonistique, mais peut donner lieu à des dialogues de sourds entre une minorité d'utilisateurs particulièrement actifs en ligne. Il apparaît ainsi réducteur de présenter les *fake news* comme la cause unique de la montée de la polarisation politique et les capacités de discernement des individus comme la panacée pour lutter contre cet enjeu. En réalité, les *fake news* sont moins la source de divisions sociales et politiques que leur expression (Kreiss, 2019). Elles constituent davantage un symptôme de dysfonctionnements structurels plus profonds qui affectent l'écosystème informationnel dans son ensemble et menace le bon déroulement du débat public. Le dispositif d'enquête de cette thèse n'a pas été conçu pour identifier les causes de ces dysfonctionnements, mais d'autres travaux suggèrent qu'ils résultent de transformations structurelles du paysage médiatique (Benkler et al., 2018 ; Cagé et al., 2022). Ainsi, traiter la polarisation uniquement comme un problème de qualité de l'information, c'est s'attaquer aux symptômes tout en négligeant les causes profondes des troubles qui affectent l'espace public. La qualité du débat public ne dépend pas uniquement du nombre de *fake news* qui circulent sur les réseaux sociaux, ni des seules capacités des individus à les identifier, mais repose sur la possibilité de créer des espaces communs d'intercompréhension permettant l'expression d'une pluralité de points de vue et l'exercice d'un contrôle mutuel entre les différents interlocuteurs d'un échange.

## Principaux arguments de la thèse

En articulant étroitement une analyse compréhensive des pratiques énonciatives des utilisateurs des réseaux sociaux face aux *fake news* à une description des espaces de communication dans lesquels elles se déploient, cette thèse apporte un autre regard sur les capacités critiques des publics numériques et sur l'architecture de l'écosystème informationnel contemporain. À rebours des discours académiques et médiatiques présentant les utilisateurs des réseaux sociaux comme des individus crédules, isolés dans des bulles de filtres en raison de leurs biais cognitifs, et les réseaux sociaux comme des marchés de l'information dérégulés, elle montre que l'écosystème informationnel numérique doit plutôt être appréhendé comme une constellation d'arènes de discussion, sous-tendues par différents contrats de communication et normes d'interactions que les utilisateurs sont en mesure d'identifier afin d'ajuster leurs pratiques énonciatives en conséquence.

### Sans *gatekeeper* mais pas sans filtre : maintien de contraintes collectives et nouvelles formes de contrôle dans l'écosystème informationnel numérique

L'approche pragmatique adoptée dans cette thèse a permis de mettre en lumière la diversité et les oscillations des régimes d'énonciation déployés par les utilisateurs des réseaux sociaux face aux *fake news.* A partir de cette diversité et de ces oscillations, nous avons pu identifier la façon dont les utilisateurs des réseaux sociaux perçoivent les situations dans lesquelles ils évoluent et les règles de grammaire qu'ils cherchent à respecter. Cette diversité et ces oscillations de pratiques invitent à repenser la façon dont sont appréhendés l'espace public et l'écosystème informationnel contemporain.

Si les réseaux sociaux facilitent la diffusion d'énoncés échappant au contrôle éditorial et aux normes déontologiques du journalisme, ils ne constituent pas pour autant des espaces de communication dépourvus de règles ou de hiérarchie. Ils restent au contraire traversés par des régulations implicites qui émanent des structures sociales. Comme l'a démontré le chapitre 4 de cette thèse, les pratiques des utilisateurs sur les réseaux sociaux sont profondément liées aux attentes et contraintes sociales qui encadrent leur comportement hors ligne. Conscients que leurs prises de parole en ligne peuvent avoir des répercussions dans leur vie sociale, professionnelle ou familiale, ces utilisateurs ajustent leur discours en fonction des attentes des différentes audiences qui pourraient y être exposées. Certains adoptent même des stratégies de dissociation entre leurs énoncés et leur identité personnelle afin de se protéger. Autrement dit, les attentes des membres de leur réseau, le risque de sanctions, l'importance de la réputation, et la crainte de représailles dans d'autres sphères sociales agissent comme autant de filtres qui modulent la manière dont les individus partagent ou consomment l'information en ligne. À travers leurs expériences et leur vie quotidienne, ils ont intériorisé ces règles et contraintes collectives, qui continuent de structurer leurs interactions numériques, même en l'absence de *gatekeeper* traditionnel.

Par ailleurs, si l'on s'affranchit des conceptions normatives traditionnelles de l'espace public, notamment celles qui cherchent à établir des principes universels et rigides de ce qui constitue une délibération rationnelle ou légitime, et qu'on observe à la place de façon compréhensive comment les utilisateurs définissent ce qu'ils considèrent comme pertinent ou impertinent, approprié ou non, en fonction des situations d'interaction, on peut découvrir comment ils tentent et parviennent à créer des espaces communs d'intercompréhension, qui n'ont pas pour toujours pour horizon la délibération mais la préservation de l'ordre des interactions.

Ainsi, plutôt que de concevoir l'écosystème informationnel numérique comme un marché cognitif monolithique ou une agora idéale de discussion rationnelle, il semble plus pertinent de l'appréhender comme une constellation d'arènes de discussion, chacune régie par ses propres règles, filtres et dynamiques de contrôle.

## Repenser les notions d’esprit critique et de rationalité

Les pratiques informationnelles et conversationnelles des utilisateurs des réseaux sociaux, analysées dans le cadre de cette thèse, invitent à repenser les définitions des notions d’esprit critique et de rationalité. La mise au jour de deux pratiques en particulier — la prudence énonciative et l'expression de points d’arrêt — montre l’intérêt de ne pas conceptualiser la rationalité ou l’esprit critique comme une série de dispositions cognitives se mesurant uniquement par des tests de logique au cours d’expériences en laboratoire, mais comme des pratiques situées et socialisées, susceptibles d’être observées dans des situations d'interactions ordinaires pour peu que l’on adopte une approche descriptive et compréhensive, affranchie des attentes normatives fixées par les discours publics.

Pour comprendre ce (re)positionnement théorique, il peut être utile de mettre brièvement en discussion — et en tension — plusieurs conceptions de la rationalité et du raisonnement proposées par différentes approches de sciences humaines et sociales.

Objet de questionnement pluridisciplinaire depuis l’Antiquité, le concept de rationalité a fait couler l’encre de nombreux philosophes, économistes, sociologues et psychologues au fil des siècles. De Platon aux philosophes des Lumières en passant par Descartes, de nombreux penseurs ont tout d’abord érigé la Raison comme un guide suprême pour accéder à la Vérité en imposant l’idée que l’acquisition de connaissances objectives et universelles ne pouvaient découler que de raisonnements logiques ou de réflexions pondérées – pas d’expériences sensibles et encore moins des émotions ou des passions des Hommes. Influencés par cette longue tradition philosophique, les économistes classiques puis néoclassiques ont alors cherché à expliquer les comportements et prises de décisions des individus en partant d’une hypothèse : celle de la rationalité des acteurs. Ce modèle, dit de *l’homo economicus* estime que les individus agissent de façon à optimise leur bien-être en cherchant à maximiser les bénéfices de leurs actions tout en minimisant leurs coûts. En pratique, cependant, les comportements des individus suivent rarement les prédictions faites par la théorie des choix rationnels. Ne disposant pas d’informations complètes, les individus ne peuvent pas prendre de décisions optimales. Selon la théorie de la rationalité limitée, élaborée par Herbert Simon, les individus vont plutôt prendre une décision quand celle-ci leur paraîtra suffisamment satisfaisante (*satisficing*).

Pour comprendre les limites de la rationalité humaine, de nouvelles méthodes de recherche ont décidé de ne pas s'arrêter aux comportements observables des individus (behaviorisme) mais de s'intéresser aux processus mentaux qui régissent leurs actions et prises de décisions. Se sont ainsi développées de nouvelles sous-disciplines, enrichies d'un cadre explicatif intégrant l'étude des mécanismes cognitifs, telles que l'économie comportementale et la psychologie expérimentale (Kahneman et Tversky, 1979). En observant les erreurs que les individus commettent lors de tâches de logique ou face à des illusions d'optique, les chercheurs ont commencé à formaliser les théories à processus duel, arguant que, plutôt que de raisonner par inférence bayésienne ou déduction, les individus s'appuient fréquemment sur leurs intuitions et leurs émotions pour faire des choix ou prendre des décisions. Avec quelques nuances, ces perspectives théoriques ont en commun de distinguer deux modes de pensée : d'une part, le Système 1, intuitif, rapide et émotionnel ; de l'autre le Système 2, lent, logique et rationnel. À la gourmandise impulsive du Système 1 pour tout ce qui est simple et efficace, s'oppose la paresse du Système 2 pour tout ce qui demande effort et réflexion. C'est donc la capacité à laisser le système 2 prendre le relais du système 1 qui distingue les bons des mauvais raisonneurs (Kahneman, 2011 ; Evans et Stanovich, 2013 ; Evans, 2008).

Si les théories à processus duel rejettent le modèle du choix rationnel dans sa validation empirique, elles ont toujours le même horizon normatif. En effet, alors que l'approche des « heuristiques et biais » est souvent présentée comme un amendement empirique aux conceptions standards de la rationalité (notamment en économie), et ébranle la valeur descriptive des théories formelles de la rationalité, elle semble en accord avec sa légitimité prescriptive (Bergeron et al., 2018). En effet, bien que les travaux d'économie comportementale et de psychologie cognitive aient mis en évidence les limites du modèle de l'*homo œconomicus* pour décrire les comportements effectivement adoptés par les individus, ces approches cherchent tout de même de façon plus ou moins explicite à corriger les biais et les erreurs de raisonnement des individus afin de réorienter leurs actions vers la situation optimale à laquelle parviendrait un *homo œconomicus* parfaitement rationnel. Ces approches se distinguent ainsi de la rationalité limitée décrite par Herbert Simon car selon ce dernier, les individus ne cherchent pas à maximiser leur bien-être par un calcul rationnel, mais plutôt à atteindre une solution satisfaisante à travers un processus de jugement.

Si les approches qui expliquent les erreurs des individus par leur défaut de raisonnement analytique sont dominantes dans le champ de la psychologie cognitive et des études contemporaines sur les *fake news* elles ne font pas moins l'objet de débats théoriques et méthodologiques fondamentaux.

Sur le plan théorique, tout d'abord, les approches classiques du raisonnement issues des théories à processus duel sont concurrencées par des approches alternatives, reposant notamment sur des perspectives pragmatiques, interactionnistes ou écologiques (Gigerenzer, 2002 ; Mercier et Sperber, 2011 ; 2017). Les divergences entre ces approches résultent d'une façon différente de conceptualiser la fonction du raisonnement. Alors que les théories à processus duels partent du principe que la fonction du raisonnement est uniquement épistémique, les approches pragmatiques, interactionnistes ou écologiques de la rationalité proposent de faire un pas de côté par rapport à l'idée que le raisonnement sert seulement à accéder à des vérités supérieures. Au lieu d'évaluer le niveau de raisonnement de chaque individu, à partir des attentes normatives fixées par les discours publics, ou d'axiomes incertains, ces approches proposent en premier lieu de s'interroger sur le rôle du raisonnement. Pourquoi les humains sont-ils dotés de capacités de raisonnement ? À quoi leur sert donc cette compétence dans la vie quotidienne ? D'une façon générale, l'hypothèse défendue par les approches pragmatiques, interactionnistes ou écologiques est que la rationalité est contextuelle, adaptée aux situations sociales et environnementales. Elles rejettent la vision d'une rationalité universelle et calculatrice, pour valoriser au contraire l'idée d'une rationalité située, ancrée dans des pratiques, des interactions et des environnements spécifiques. Par exemple, la théorie argumentative du raisonnement proposée par Hugo Mercier et Dan Sperber avance que la fonction première du raisonnement n'est pas d'aboutir à des réponses correctes mais avant tout de produire et d'évaluer des raisons. Autrement dit, le raisonnement permet aux individus de justifie leurs croyances auprès des autres tout en passant au crible leurs arguments. Cette approche invite à mettre en œuvre une perspective communicationnelle dans laquelle la vérification des informations est une activité fondamentalement interactionnelle. Elle soutient ainsi que la fonction du raisonnement est avant tout argumentative et qu'il s'agit d'une compétence sociale. Ces approches soulignent aussi que les individus utilisent des heuristiques et des ajustements dynamiques pour naviguer dans des situations complexes, sans chercher à maximiser systématiquement une logique formelle ou abstraite. Autrement dit, certaines heuristiques pourraient s'apparenter plutôt à des inférences pertinentes, car adaptées au contexte dans lequel se trouvent les individus, plutôt que comme des erreurs logiques systématiques de leur part (Hertwig et Gigerenzer, 1999).

Sur le plan méthodologique ensuite, les approches pragmatiques, interactionnistes ou écologiques de la rationalité et du raisonnement ont apporté des critiques importantes aux dispositifs expérimentaux qui évaluent *in silico* les capacités de raisonnement des individus en leur demandant de résoudre des problèmes logico-mathématiques En effet, en utilisant des designs expérimentaux alternatifs, plusieurs travaux ne parviennent pas à répliquer certains résultats issus d'études classiques de psychologie du raisonnement. L'un des exemples le plus célèbre pour illustrer ces problèmes de réplication est très sûrement celui de la tâche de sélection de Wason (1966), l'une des tâches les plus utilisées en psychologie pour démontrer le biais de confirmation des individus. Par exemple, une interprétation pragmatique de la tâche de Wason, proposée par Vittorio, Kemmelmeier, Van der Henst et Sperber (2001) dans un article intitulé « Raisonneurs incompétents ou virtuoses pragmatiques ? » permet de mettre au jour les oscillations des taux de succès selon la manière dont l'énoncé a été cadré. Pour les chercheurs, cette variabilité des performances s'explique par des considérations pragmatiques en termes de maximisation de pertinence. Ces résultats montrent non pas que les gens sont de mauvais raisonneurs, mais plutôt qu'ils sont des virtuoses pragmatiques, capables d'ajuster leur interprétation de la tâche à différentes situations. Dans une veine similaire, à la place de la formulation abstraite de la tâche de Wason, Leda Cosmides et John Tooby (1992) ont reformulé le problème dans une version sociale, c'est-à-dire qu'ils ont transformé l'énoncé initial en une question plus susceptible d'être rencontrée dans la vie quotidienne mais impliquant les mêmes procédures de décision.

Malgré des divergences épistémologiques fondamentales, les approches de psychologie qui reposent sur une perspective pragmatique, interactionniste et écologique peuvent être mises en discussion avec des approches de sociologie pragmatique (Thévenot, 2006, p. 197-201 ; Vayre, 2022). Ces deux approches partagent une vision contextualisée et située de l'action humaine et renforcent la pertinence de penser les comportements humains non comme des réponses à des normes abstraites, mais comme des ajustements contextuels, adaptés à des situations vécues

Cette contextualisation est centrale dans les travaux de Julien Talpin et Mathieu Berger. Par exemple, Julien Talpin (2006, 2010) montre que les compétences civiques des individus se construisent à travers l'engagement participatif et l'expérience de la délibération. Mathieu Berger, lui, met en lumière la manière dont les individus « répondent en citoyens » dans des situations de participation profane, un aspect qui renforce l'idée que les compétences civiques et les comportements rationnels ne sont pas des qualités fixes, mais bien des compétences situées, adaptées aux contraintes du moment. Ces résultats conduisent à dépasser la conception classique de la rationalité, souvent limitée aux espaces institutionnels et procéduraux où l'usage public de la raison est orienté vers la délibération et la recherche de consensus Une approche élargie de la rationalité invite à observer, dans des espaces plus informels, un processus public de construction de la raison et de compréhension mutuelle. Ici, la discussion n'est pas nécessairement tournée vers une prise de décision ou une action politique directe, mais se manifeste dans des échanges quotidiens dont la finalité principale est la préservation du lien social.

## Limites de la thèse et pistes pour de futures recherches

La principale limite de la thèse réside dans la composition des corpus de *fake news* et des échantillons d'enquêtés étudiés. Cette limite résulte des deux définitions opérationnelles du terme *fake news* retenues pour délimiter des corpus de *fake news*, ainsi que du dispositif d'enquête mis en place pour identifier des utilisateurs des réseaux sociaux ayant été exposés à ces contenus.

En prenant appui sur un corpus de contenus classés comme des *fake news* par les principales rubriques de *fact-checking* françaises, la première enquête a permis d'identifier l'ensemble des utilisateurs ayant partagé au moins une *fake news* sur la Twittosphère française et d'analyser leurs pratiques informationnelles et conversationnelles. Cette approche a permis de confronter la façon dont les utilisateurs des réseaux sociaux sont dépeints dans le débat public, à partir des catégories d'analyse issues des discours journalistiques, à des données empiriques. Cette approche a également permis d'inscrire notre thèse dans la continuité de la majorité des études scientifiques et ainsi de confirmer, réfuter ou approfondir leurs résultats afin de prendre position dans les débats académiques qui entourent les *fake news*.

En prenant appui sur un corpus de contenus signalés comme des *fake news* par des utilisateurs de Facebook, la seconde étude a permis d'identifier l'ensemble des utilisateurs ayant réagi par un commentaire à au moins un de ces contenus sur Facebook et d'analyser leurs régimes d'énonciation. Cette approche nous a donné l'opportunité de dépasser la seule question de la factualité de l'information et d'étudier plus largement les réactions des utilisateurs à des énoncés dont la qualité épistémique est incertaine afin d'examiner comment ils formulent le problème des *fake news* et tentent de le résoudre au cours de situations qu'ils sont susceptibles de percevoir comme problématiques.

Au-delà d'apporter des contributions empiriques importantes dans le champ des études sur les *fake news*, ces deux enquêtes ont permis d'apporter des contributions conceptuelles et méthodologiques aux travaux de sociologie de la réception, de sociologie pragmatique et de sociologie des usages du numérique. Articuler des méthodes quantitatives et qualitatives a permis d'interpréter de façon fine et précise les traces numériques émises par des utilisateurs en réaction à des *fake news* sur les réseaux sociaux, et ainsi de cerner la diversité des régimes d'énonciation et plus encore de mettre au jour leurs compétences de distance critique. Cette approche hybride, fondée sur l'analyse des traces numériques, constitue une nouveauté dans les études sur la réception des *fake news*.

Malgré leurs apports, ces deux enquêtes présentent cependant une limite majeure : elles se concentrent sur les utilisateurs actifs qui ont réagi à des *fake news* sur les réseaux par la production de traces numériques (en l'occurrence de tweets ou commentaires Facebook), sans prendre en compte les utilisateurs exposés à ces contenus mais qui n'y ont pas réagi publiquement, ou ceux qui n'y ont pas été exposés du tout. Les deux études de cette thèse portent donc sur une frange spécifique d'utilisateurs des réseaux sociaux et non sur un échantillon représentatif. Reformulée dans le contexte numérique, cette citation de Dominique Mehl (2004, p. 146) s'applique parfaitement à notre échantillon : « Public en réaction, il n'est pas représentatif des utilisateurs pris dans leur ensemble. Public critique, il n'est pas le miroir exact de l'audience globale des réseaux sociaux. Public actif, il n'est pas le reflet fidèle de l'utilisateur moyen ». Dans une certaine mesure, et pour reprendre les mots de Sabine Chalvon-Demersay (2003), on pourrait dire que notre recherche se concentre sur des « publics particulièrement concernés ».

De plus, notre approche se fonde sur une définition relativement restreinte des *fake news*, ce qui laisse de côté des phénomènes émergents tels que les *deepfakes* ou d'autres stratégies de désinformation. Pour les recherches futures, il serait bénéfique d'adopter une perspective plus large, en s'intéressant aux pratiques des publics dans des contextes naturels et variés. Plutôt que de partir des définitions prédéfinies des *fake news* et des traces numériques laissées par les utilisateurs, nous suggérons d'adopter une approche inversée : identifier un échantillon représentatif d'internautes, puis observer leurs pratiques *in situ*, via des entretiens et la collecte éventuelle de leurs données. Cette méthode permettrait de dépasser les limites actuelles de l'analyse basée sur les seules traces numériques, et d'explorer les comportements des utilisateurs invisibles et silencieux, souvent absents des études actuelles.

Ainsi, il est crucial d'adopter une posture plus globale, en sortant des cadres restrictifs des réseaux sociaux et en partant des pratiques réelles des utilisateurs. Ce changement de perspective permettrait d'élargir notre compréhension des processus de réception et de production d'information à l'ère numérique, en tenant compte des publics passifs et non visibles.

## Implications et recommandations pour le débat public

Les apports de cette thèse ne sont pas uniquement destinés au monde académique mais visent également à nourrir le débat public. Comme cela a été montré dans le premier chapitre, les discours publics sur les *fake news* ont orienté l'action publique vers des mesures de régulation ou d'éducation à l'esprit critique en véhiculant des représentations alarmistes de l'écosystème informationnel numérique et des utilisateurs des réseaux sociaux. Les résultats de la thèse invitent à certaines bifurcations et inflexions tant au niveau des discours à porter que des actions à mener.

### Un problème public à reformuler

#### ***Des fake news aux troubles de l'information et de la communication***

Depuis l'apparition du terme *fake news* dans le débat public en 2016, de nombreux acteurs ont recommandé de l'éviter. Aux États-Unis, des distinctions telles que *misinformation*, *malinformation* et *disinformation* ont été proposées par des chercheurs comme Claire Wardle et Hossein Derakhshan (2017). En novembre 2018, après le rejet d'une loi « relative à la lutte contre les fausses informations », le Parlement a finalement adopté une loi « relative à la lutte contre la manipulation de l'information en période électorale ». Ces tentatives de redéfinition montrent l'importance du choix des mots utilisés dans la formulation des problèmes publics et indiquent une volonté d'élargir la problématique des *fake news* au-delà de la simple factualité des énoncés, en prenant en compte l'intention de leurs producteurs et diffuseurs. À l'issu de cette thèse, nous proposons d'aller un cran plus loin en invitant les acteurs du débat public à parler plutôt de problème d'écologie informationnelle et de troubles de l'information et de la communication. Cette nouvelle terminologie permet de dépasser la focalisation étroite sur la seule factualité des énoncés et d'englober des informations biaisées ou trompeuses – mais pas forcément fausses – qui sont souvent amplifiées par les médias traditionnels eux-mêmes afin d'aborder plus largement les cadres idéologiques et les usages médiatiques qui influencent la façon dont les informations sont perçus selon les espaces de communication où elles sont mises en circulation (Watts et al., 2021 ; Krämer, 2021).

### *Éviter les discours techno-déterministes et reconnaître l'existence de causes plus structurelles*

Ces dernières années, la polarisation politique a souvent été attribuée à l'influence des réseaux sociaux, accusés de favoriser la diffusion de fausses informations et d'exacerber les divisions au sein de la société. Cependant, une analyse plus approfondie révèle que cette explication techno-déterministe est réductrice et néglige des causes structurelles plus profondes. La polarisation actuelle et le partage de *fake news* reflètent une crise de la représentation, où une partie croissante de la population ne se sent plus représentée par les institutions politiques traditionnelles. Cette déconnexion alimente la défiance envers les élites et les médias, perçus comme déconnectés des réalités quotidiennes des citoyens. Les réseaux sociaux, bien que souvent pointés du doigt, agissent davantage comme des amplificateurs de tendances préexistantes que comme des causes premières de la polarisation. Ils offrent des espaces de communication où les opinions peuvent être exprimées et partagées rapidement, mais ils ne créent pas *ex nihilo* les divisions. Éviter les discours techno-déterministes permet ainsi de ne pas négliger les facteurs socio-politiques à l'origine de la polarisation et de mieux penser à des solutions ciblant directement ces causes structurelles.

### *Ne pas sous-estimer les capacités critiques des publics*

Dans les discours publics, les utilisateurs des réseaux sociaux sont souvent représentés par des synecdoques. Leurs réactions face aux *fake news* sont réduites à des millions de *likes*, de partages ou de commentaires. Ces chiffres alimentent une présomption de crédulité généralisée et masquent la diversité de leurs pratiques. Des milliers de vues sur Youtube sont rapidement assimilées à un refus de se faire vacciner ou de porter un masque. Comme l'ont montré les enquêtes conduites dans le cadre de cette thèse, pourtant, il est particulièrement difficile d'interpréter de façon juste les traces numériques des utilisateurs. Celles-ci cachent parfois des stratégies de prudence énonciative ou de distance critique.

Sans aller jusqu'à voir naïvement derrière chaque *like* un citoyen éclairé, il peut être utile de ne pas partir systématiquement d'une présomption d'irrationalité. En effet, cela peut conduire les pouvoirs publics à prendre des mesures inadaptées. Par exemple, les gouvernements et institutions définissent souvent leurs stratégies de communication en se représentant le public comme irrationnel, peu capable de se faire un jugement en situation d'incertitude informationnelle et prompt à des mobilisations paniques (Ward, 2018). Or, l'expérience montre que cette représentation paternaliste est erronée et conduit souvent à une communication contre-productive. Par exemple, les leçons des expériences gouvernementales de la gestion de la crise du COVID montrent qu'il est important que les gouvernements ne craignent pas la panique des populations et leur adressent des messages clairs, ne dissimulent pas les incertitudes et rendent compte de la façon la plus juste possible l'état des savoirs scientifiques (Petersen, 2021). Le risque autrement est d'accroitre la défiance des publics à l'encontre des médias, des élites et des institutions.

Reformuler le problème des *fake news* en termes de troubles de l'informations et de la communication en tenant compte de ses causes structurelles et des capacités critiques des utilisateurs permet ainsi de repenser les approches visant à préserver la qualité du débat public.

## Une action publique à réorienter

La manière dont le problème des *fake news* a été abordé dans le débat public a conduit à privilégier des approches centrées sur la factualité des énoncés et les biais cognitifs des individus. Par exemple, des initiatives telles que le *fact-checking*, la modération des contenus ou encore l'éducation à l'esprit critique s'inscrivent dans cette logique. Cette section propose d'élargir ces perspectives en formulant des suggestions qui visent à (1) se décentrer de la seule factualité des énoncés pour adopter une approche écologique qui tient compte de l'écosystème informationnel dans sa globalité, et à (2) déplacer l'accent mis sur les biais cognitifs individuels, notamment des jeunes issus des milieux populaires, pour s'intéresser aux pratiques énonciatives des utilisateurs en situation d'interactions.

## ***Se décentrer de la factualité des énoncés et adopter une approche écologique des troubles de l'information et de la communication***

### *Du fact-checking à l'indépendance des médias*

Le *fact-checking* a une place importante dans la lutte contre la désinformation, mais il ne doit pas constituer l'unique réponse. En effet, les approches centrées sur la vérification des faits, comme le *fact-checking*, se concentrent sur des énoncés unitaires, en les classant comme vrais ou faux, mais elles omettent souvent de questionner le cadrage et la mise en visibilité de ces informations. Or, celles-ci circulent dans un écosystème complexe où des acteurs variés, des algorithmes aux choix éditoriaux des médias, influencent quels énoncés sont amplifiés ou marginalisés. Il est donc nécessaire de développer des approches plus systémiques. Cela inclut des efforts visant à garantir l'indépendance des médias, à éviter les pressions économiques ou politiques qui compromettent leur rôle de contre-pouvoir.

### *Responsabiliser les discours des élites et des institutions*

Alors que le partage de *fake news* est davantage lié à une défiance envers les élites et les institutions qu'à un manque de raisonnement analytique de la part du grand public, il est prioritaire de responsabiliser les discours des personnalités et des institutions publiques, plutôt que de chercher uniquement à corriger les erreurs des individus. C'est ce que défend souvent Rasmus Kleis Nielsen, chercheur associé au *Reuters Institute for the Study of Journalism.*[208] En effet, en raison de leur grande visibilité en ligne, les figures publiques jouent un rôle crucial dans la manière dont les informations sont perçues et leurs déclarations peuvent grandement influencer la confiance des publics dans le système médiatique.

Au lieu de se concentrer uniquement sur les utilisateurs ordinaires des réseaux sociaux, une approche plus efficace consisterait ainsi à modifier les incitations et les pratiques des élites politiques et autres personnalités publiques. Une étude menée par Nyhan et Reifler (2015) illustre ce point. Avant les élections de 2012, un groupe aléatoire de législateurs de neuf États américains a reçu des messages leur rappelant les coûts politiques liés à la diffusion de déclarations fausses. Les résultats ont montré que les législateurs ayant reçu ces messages étaient moins susceptibles de voir la véracité de leurs déclarations remise en question publiquement, ce qui suggère que le fait d'être sensibilisé aux conséquences de la diffusion de fausses informations peut encourager des discours plus responsables et réduire la propagation de fausses informations. Cela pourrait également contribuer à restaurer une partie de la confiance du public dans les institutions et les élites.

---

[208] Nielsen, R. (2024, January 2). Forget technology — politicians pose the gravest misinformation threat. *Financial Times*. https://www.ft.com/content/5da52770-b474-4547-8d1b-9c46a3c3bac9

***Repenser les dispositifs d'éducation aux médias, à l'esprit critique et au numérique et élargir leurs publics***

De nombreux dispositifs d'éducation aux médias ciblent spécifiquement les jeunes, issus de milieux populaires, et sont dédiés à leur faire prendre conscience de leur biais cognitifs. Ces efforts ne sont ni inutiles ni superflus. Il ne s'agit pas ici de remettre en cause leur pertinence, mais plutôt de souligner l'intérêt d'autres approches qui mettent davantage l'accent sur les pratiques concrètes des individus que sur leurs seuls mécanismes de pensée, et visent à s'adresser à des publics plus diversifiés.

*Encourager l'expression publique en ligne et l'écoute*

La majorité des dispositifs mis en œuvre pour lutter contre les *fake news* visent à conduire les utilisateurs à réfléchir par deux fois avant d'agir. Par exemple, de nombreuses plateformes utilisent le principe de la « friction » comme levier comportemental : elles multiplient les clics requis pour partager un article (en demandant, par exemple, si l'utilisateur souhaite le lire ou en lui demandant d'évaluer la véracité du titre). Si cette approche peut dissuader la diffusion de contenus erronés, elle risque également de freiner la circulation d'informations fiables, d'autant que seuls quelques utilisateurs partagent activement, tandis que la majorité tend à s'autocensurer.

Une critique récurrente est que ces dispositifs visent surtout à faire taire les diffuseurs de fausses informations, alors que le véritable enjeu consiste à encourager l'expression de ceux qui restent silencieux, en particulier les individus marginalisés. Autrement dit, l'objectif n'est pas uniquement de limiter la diffusion de la désinformation, mais aussi de promouvoir une prise de parole plus large et diversifiée en ligne. Il ne s'agit pas seulement d'apprendre à identifier les fausses informations, mais aussi de développer une véritable « grammaire publique numérique » : un langage commun permettant d'articuler les idées et les préoccupations au sein d'espaces de communication en ligne.

Des initiatives telles que La Zone d'Expression Prioritaire (ZEP) et « Je suis là » illustrent cette démarche. La ZEP, lancée en 2015 par Emmanuel Vaillant, Edouard Zambeaux et Thibault Renaudin, est un média associatif qui accompagne l'expression des jeunes à travers des ateliers d'écriture et de création médiatique. De son côté, l'association « Je suis là » incarne une approche visant à redonner la parole aux personnes habituellement mises en retrait des espaces de débat public.

Parallèlement, il est crucial de travailler à l'éducation à l'écoute pour ceux qui, en raison de leur position sociale ou culturelle, sont souvent « sourds » aux voix divergentes. Sur le plan de l'éducation aux médias, il est envisageable de mettre en place des dispositifs visant à cultiver l'écoute active chez certains groupes (notamment les élites et les générations plus âgées) tout en facilitant l'expression pour d'autres (comme les jeunes et les publics sous-représentés). Par exemple, dans les salles de classe, la participation des élèves est souvent évaluée uniquement en fonction de leur prise de parole orale. Valoriser également la participation écrite et l'écoute active pourrait enrichir ce processus et favoriser une communication plus inclusive.

En définitive, dans un débat démocratique, le droit d'expression doit être accompagné du devoir d'écoute. En promouvant ces valeurs, nous pourrions non seulement enrichir le discours public, mais aussi contribuer à construire une société plus inclusive et engagée.

*Des publics négligés*

Alors que certains dispositifs d'éducation aux médias sont pleinement intégrés au programme scolaire, ils s'adresse principalement aux jeunes, notamment ceux qui sont issus des milieux populaires mais tendent à ignorer certains publics, notamment les personnes plus âgées et les élites. Cette focalisation omet des pans entiers de la population qui jouent également un rôle clé dans les dynamiques informationnelles. Il est ainsi important de promouvoir des dispositifs qui favorisent le dialogue intergénérationnel. Par exemple, des programmes tels que la « Parentalité numérique » du CLEMI visent à sensibiliser les parents aux enjeux de l'éducation aux médias, favorisant ainsi le dialogue intergénérationnel.

En reformulant le problème des *fake news* en termes de troubles de l'information et de la communication, cette thèse n'entend pas dire que les valeurs de vérité et d'objectivité n'ont pas d'importance dans le débat public. Nos résultats contribuent à voir les choses sous un autre angle. Nous espérons que nous avons produit de la connaissance mais aussi servir intérêt public. Finalement, les défis posés par la circulation des fausses informations ne doivent pas se résumer à une simple question de manque d'esprit critique. Il est important de comprendre que ces phénomènes révèlent une crise plus profonde de la démocratie et des formes de représentation sociale. Plutôt que de se concentrer uniquement sur les individus et leurs biais, il est temps d'adopter une approche plus systémique, visant à renforcer les institutions démocratiques et à revitaliser l'espace public. En d'autres termes, l'éducation aux médias ne devrait pas seulement viser à corriger les comportements individuels, mais aussi à rétablir des canaux de communication plus inclusifs et à créer des espaces où les citoyens, quelle que soit leur position sociale, puissent s'exprimer, se faire entendre, et participer activement au débat public.

# Bibliographie

# Annexes

## Annexe 1. Revue de littérature non exhaustive d'études mesurant l'audience des *fake news* et/ou leur place dans le total de contenus consommés par les individus.

| Étude | Échantillon | Données | Pays | Période | Corpus | Définition des *fake news* | Audience/ Engagement | Place dans la consommation globale de contenus |
|---|---|---|---|---|---|---|---|---|
| Allcott et Gentzkow, 2017 | 1 200 | Navigation web<br><br>Questionnaire | États-Unis | Novembre 2016 | 65 sites de *fake news*<br><br>665 médias *mainstream* | *Snopes*, *PolitiFact*, *BuzzFeed* | Sites de *fake news* : 159M impressions<br><br>Médias *mainstream* : 3Md impressions | N/A |
| Nelson et Taneja, 2018 | 1 000 | Navigation web | États-Unis | Janvier 2016 - Janvier 2017 | 30 sites de *fake news*<br><br>24 médias *mainstream* | *OpenSources* | Sites de *fake news* : 675 000 visiteurs uniques<br><br>Médias *mainstream* : 28M visiteurs uniques | Sites de *fake news* : 4,5 min/ mois<br><br>Médias *mainstream* : 9 min/ mois |
| Fletcher et al., 2018 | N/A | Navigation web<br><br>Engagement Facebook | France | Janvier - Octobre 2017 | 38 sites de *fake news*<br><br>5 médias *mainstream* | *Les Décodeurs* | Sites de *fake news* : 1 % (*reach* moyen individuel)<br><br>Médias *mainstream* : 12,6 % à 22,3 %<br><br>Pages facebook de sites de *fake news* : 0,1 à 113,7 M d'interactions<br><br>Pages facebook de médias *mainstream* : 13,4 à 56,6 M d'interactions | Sites de *fake news* : 0,2 à 9,4 M de minutes par mois.<br><br>Médias *mainstream* : 47,3 à 178 M minutes par mois |
| | N/A | // | Italie | // | 21 sites de *fake news*<br><br>5 médias *mainstream* | *BUTAC*<br>*Bufale*<br>*Bufalopedia* | Sites de *fake news* : 1 % (*reach* moyen individuel)<br><br>Médias *mainstream* : 6,2 % à 50,9 %<br><br>Pages facebook de sites de *fake news* : 0,0 à 7,2 M d'interactions<br><br>Pages facebook de médias *mainstream* : 1,6 à 55,4 M d'interactions | Sites de *fake news* : 0,1 à 7,5 M de minutes par mois<br><br>Médias *mainstream* : 13,9 à 443,5 M de minutes par mois |
| Gaumont et al. 2018 | 2,4 M | Engagement Twitter | France | Août 2016 - Mai 2017 | 179 URLs de *fake news*<br><br>60 M de tweets | *Les Décodeurs* | N/A | 0,081 % de tous les tweets partagés |
| Guess et al., 2019 | 1 191 | Partage Facebook | États-Unis | Avril - Novembre 2016 | 21 sites de *fake news* | *Buzzfeed News* | 8,5 % ont partagé un site de *fake news* | N/A |
| Grinberg et al., 2019 | 16 442 | Engagement Twitter | États-Unis | Août - Décembre 2016 | 300 sites de *fake news*<br><br>475 médias *mainstream* | *Buzzfeed News*<br>*Politifact*<br>*FactCheck.org*<br>*Snopes.com* | 1 % ont été exposés et 0,1 % ont partagé 80 % des *fake news* | 1,18 % du total d'exposition à des informations politiques. |
| Allen et al., 2020 | TV panel : 150 000<br><br>Web panel : 60-90 000 | TV<br><br>Navigation web | États-Unis | Janvier 2016 - Décembre 2018 | 98 sites de *fake news*<br><br>800 médias *mainstream*<br><br>400 programmes télé<br><br>2 000 sites non liés à l'actualité | Grinberg et al., 2019<br>*NewsGuard*<br>*Buzzfeed* | Seulement 1,97 % ont consommé plus de contenus issus de site de *fake news* que de média *mainstream* | moins d'une minute par jour<br><br>1 % de la consommation d'informations d'actualité<br><br>0,15 % du total de consommation médiatique |
| Guess et al. 2020 | 2 525 | Navigation web | États-Unis | Octobre - Novembre 2016 | 300 sites de *fake news*<br><br>500 sites d'actualité | Grinberg et al., 2019 | 44,3 % ont consulté un site de *fake news* | 64,2 secondes en moyenne<br><br>5,9 % du total d'informations consommées |
| Hopp et al., 2020 | 543 | Engagement Twitter | États-Unis | Mars - Juin 2017 | 106 sites de *fake news* | *About.com, Aloisius, CBS* | 5 % (partage Twitter) | N/A |
| | | Engagement Facebook | | | 1 914 médias *mainstream* | *News, The Daily Dot, Fake News Watch, Fake News Checker, Melissa Zimdars, NPR, Snopes Field Guide, The New Republic, et U.S. News, World Report* | 28,9 % (partage Facebook) | |
| Osmundsen et al., 2021 | 2 337 | Engagement Twitter | États-Unis | Décembre 2018 - Janvier 2019 | 42 sites de *fake news*<br><br>260 médias *mainstream* | Guess et al., 2019 | 11 % mais 1 % est responsable du partage de 75 % des *fake news* | 3 % de tous les tweets partagés<br><br>4 % du total d'informations d'actualité partagées |
| Cordonnier et Brest 2021 | 2 372 | Navigation web | France | Septembre - Octobre 2020 | 2 295 sites web<br><br>651 chaînes YouTube | Algorithme de la société *Storyzy* | 39 % | 0,16 % du temps total de connexion à Internet et 5 % du temps total d'information en ligne |
| Guess et al., 2021 | N/A | Vues Facebook | États-Unis | Janvier-Décembre 2018 | 966 591 URLs | *NewsGuard* | 11 % (exposition)<br><br>15 % (partage) | N/A |
| Altay et al., 2022 | 300 000 | Navigation web<br><br>Engagement Facebook | États-Unis | 2017-2021 | 128 sites de *fake news*<br><br>2 322 médias *mainstream* | *NewsGuard* | 3,6 % (web)<br><br>18,52 % (Facebook) | N/A |
| | 55 500 | // | Royaume-Uni | // | 6 sites de *fake news*<br><br>334 médias *mainstream* | // | 0,10 % (web)<br><br>2,44 % (Facebook) | N/A |
| | 23 000 | // | France | // | 51 sites de *fake news*<br><br>231 médias *mainstream* | // | 4,43 % (web)<br><br>23,00 % (Facebook) | N/A |
| | 27 000 | // | Allemagne | // | 28 sites de *fake news*<br><br>220 médias *mainstream* | // | 0,97 % (web)<br><br>18,52 % (Facebook) | N/A |
| Moore et al., 2023 | 1 151 | Navigation web | États-Unis | 2 Octobre - 9 Novembre 2020 | 1 796 sites de *fake news*<br><br>5 471 sites d'actualité | *NewsGuard* | 26,2 % | 38,6 secondes en moyenne |
| Zhou et al., 2023 | 140 000 | Navigation web | États-Unis | Janvier - Décembre 2018 | 504 sites de *fake news*<br><br>707 médias *mainstream* | *Buzzfeed*<br>*Politifact*<br>*FactCheck.org*<br>*Snopes.com*<br>*Allcott and Gentzkow, 2017,*<br>*Guess et al., 2020,*<br>*NewsGuard* | 21 % mais 1 % est responsable d'avoir visité 65,3 % de toutes les pages web des sites de *fake news* | N/A |
| Allen et al., 2024 | 233M | Vue Facebook | États-Unis | Janvier - Mars 2021 | 13 206 URLs sur les vaccins | *Facebook third-party fact-checkers*<br>Lasser et al., 2023<br>*Crowdsourcing*<br>MLP | 8,7 millions de vues | 0,3 % des 2,7 milliards d'URLs vues<br><br>5,1% de vues pour les contenus issus de site jugés peu fiables |
| Baribi-Bartov et al., 2024 | 2 707 relayeurs de *fake news* prolifiques | Engagement Twitter | États-Unis | Août-Novembre 2020 | 5 534 URLs fiables et non fiables | Grinberg et al., 2019<br>*NewsGuard* | 0,3 % ont partagé 80 % des *fake news*<br><br>Leurs tweets atteignent 5,2 % des comptes d'électeurs américains enregistrés sur Twitter | 7 % de tous les contenus politiques partagés sur Twitter |

## Annexe 2. Revue de littérature non exhaustive d'études mesurant les effets de différents types de contenu informationnel sur les croyances, attitudes et comportements des individus.

| Étude | Échantillon | Pays | Période | Traitement/ Variable instrumentale | Effet mesuré | Résultats |
|---|---|---|---|---|---|---|
| DellaVigna et Kaplan, 2007 | données de vote pour 9 256 villes | États-Unis | Octobre 1996 - Novembre 2000 | introduction de *Fox News* sur le marché de la télévision par câble | • participation électorale<br>• choix électoral | • augmentation de la part des voix républicaines aux élections présidentielles de 0,4 à 0,7 points.<br>• impact global sur le vote à l'échelle nationale estimé entre 0,15 et 0,2 point.<br>• 3 à 28 % de téléspectateurs non républicains ont été convaincus de voter républicain. |
| Bond et al., 2012 | 61M | États-Unis | 2 Novembre 2010 | exposition à un message de mobilisation politique encourageant à voter avec ou sans incitations sociales | • vote auto-déclaré<br>• recherche d'informations sur les élections<br>• comportement électoral | • augmentation de 2,08 % des déclarations de vote pour le groupe ayant reçu un message avec incitations sociales par rapport à celui sans incitation sociale.<br>• augmentation de 0,26 % de la recherche d'informations pour le groupe ayant reçu un message avec incitations sociales.<br>• augmentation de 0,39 % de la participation électorale pour le groupe ayant reçu un message avec incitations sociales mais pas de différence entre le groupe sans incitations sociales et le groupe contrôle.<br>• effets indirects de 0,099 % sur les déclarations de vote ; 0,011 % sur la recherche d'informations ; 0,224 % sur le vote réel pour les amis proches des utilisateurs ayant reçu un message avec incitations sociales. |
| Hopkins et Ladd, 2014 | 22 592 | États-Unis | 2000 | introduction de *Fox News* sur le marché de la télévision par câble | • intention de vote | • pas d'effet sur l'ensemble de la population.<br>• effet significatif de 2,6 points sur les intentions de vote des Républicains et des indépendants. |
| Broockman et Green, 2014 | N1 : 2 984<br><br>N2 : 3 557 | États-Unis | 6-8 Octobre 2012<br><br>29 Octobre - 4 Novembre 2012 | exposition (répétée à des publicités politiques sur Facebook | • N1 : mémorisation des pubs ; reconnaissance des noms des candidats ; évaluations des candidats<br>• N2 : évaluation du candidat ; évaluation de l'opposant ; | • N1 : aucun effet significatif sur la reconnaissance du candidat, l'évaluation du candidat ou le soutien électoral<br>• N2 : aucun effet significatif sur la reconnaissance du candidat, l'évaluation du candidat ou le soutien électoral mais effet de 5,3 points sur la mémorisation |
| | | | | | mémorisation des enjeux de campagne mis en avant dans la pub ; mémorisation de la publicité ; fréquence d'utilisation de Facebook au cours de la semaine passé | de la publicité |
| Kalla et Broockman, 2017 | Méta-analyse de 49 (quasi) expériences de terrain | US | 2004-2017 | campagnes et publicités politiques (par courrier, appels téléphoniques, porte-à-porte ou en ligne ou via la TV) | • choix électoraux | • aucun effet significatif sauf pour 2 études<br>• estimation optimiste : effet de persuasion auprès d'1 électeur sur 175<br>• estimation la plus vraisemblable selon les auteurs : effet de persuasion auprès d'1 électeur sur 800.<br>• lorsque les campagnes contactent les électeurs longtemps avant le jour de l'élection et mesurent les effets immédiatement, les campagnes semblent souvent persuader les électeurs. Cependant, cette persuasion initiale disparaît avant le jour de l'élection et les mêmes traitements cessent généralement de fonctionner à l'approche du jour de l'élection.<br>• les campagnes peuvent avoir des effets persuasifs significatifs dans les campagnes primaires et les référendums, lorsque les indices partisans ne sont pas présents. |
| | 9 nouvelles expériences | US | 2015-2016 | porte-à-porte | | • les efforts de persuasion par porte-à-porte semblent avoir peu d'effet sur le choix de vote général dans les élections lorsque menés près de la date de l'élection, mais peuvent avoir un impact plus marqué lorsqu'ils sont réalisés plus tôt ou dans des contextes sans indices partisans.<br>• en moyenne, le contact personnel - comme le porte-à-porte ou les appels téléphoniques - mené dans les deux mois précédant une élection générale n'a pas d'effet substantiel sur le choix de vote. |
| Martin et Yurukoglu, 2017 | N/A | États-Unis | 1998-2008 | position d'une chaîne dans la liste des chaînes de télévision par câble | • exposition<br>• opinion politique<br>• choix électoral | • regarder *Fox News* 2,5 minutes supplémentaires par semaine a augmenté la part des voix du candidat républicain à la présidentielle de 0,3 point<br>• regarder *MSNBC* 2,5 minutes supplémentaires par semaine n'a pas d'effet significatif<br>• si *Fox News* n'avait pas été présente avant l'élection de 2000, la part des votes pour le candidat républicain aurait été réduite de 0,46 point. |
| Spenkuch et Toniatti, 2018 | N/A | États-Unis | 2004-2012 | exposition à des publicités politiques | • taux de participation électorale<br>• part des voix des candidats. | • presque aucun impact sur la participation électorale globale.<br>• mais effet positif et significatif sur les parts de voix des candidats, e.g. voir environ 22 publicités de plus pour un candidat plutôt qu'un autre augmente la différence partisane des parts de voix d'environ 0,5 point. |
| Bail et al., 2018 | 1 652 | États-Unis | Octobre 2017 | suivre un bot retweetant des messages de comptes Twitter d'un bord politique opposé | • attitude politique | • renforcement des attitudes politiques préalables.<br>• augmentation significative pour les Républicains (entre 0,11 et 0,59 point) mais pas pour les Démocrates. |
| Lévy, 2021 | 17 635 | États-Unis | Février-Mars 2018 | exposition à des contenus d'un bord politique opposé | • attitude politique | • effet sur l'orientation des sites d'information consultés par les individus.<br>• diminution des attitudes négatives envers le parti politique opposé.<br>• pas d'effet sur les opinions politiques<br>• l'algorithme de Facebook est moins susceptible de fournir aux individus des publications provenant de médias aux opinions contraires, même s'ils y sont abonnés. |
| Ash et al., 2022 | 661 000 | États-Unis | 2000-2020 | exposition à Fox News | • taux de participation électorale<br>• attitude politique | • une diminution d'un écart-type de la position de *Fox News* (i.e. ~ 29 positions, ce qui induit ~ 7 minutes de visionnage supplémentaire par semaine) a augmenté les parts de voix républicaines d'au moins 0,5 point de pourcentage lors des élections présidentielles, sénatoriales, de la Chambre des représentants et des gouverneurs.<br>• les effets de *Fox News* ont augmenté régulièrement entre 2004 et 2016 avant de se stabiliser. |
| Broockman, et Kalla, 2022 | 763 | États-Unis | Août-Octobre 2020 | remplacement du visionnage de Fox News par le visionnage de CNN parmi les téléspectateurs de Fox News | • croyances factuelles<br>• attitudes politiques<br>• perceptions de l'importance d'enjeux d'actualité<br>• opinion politique | • effet de 0,11 à 0,23 écarts-types sur la mémorisation de nouvelles négatives sur Donald Trump et des nouvelles positives sur Joe Biden, et vice versa<br>• connaissance accrue (effet de 0,07 à 0,21 écarts-types) de faits défavorables aux Républicains et favorables aux Démocrates.<br>• augmentation du soutien pour le vote par correspondance (VBM) : 0,24 écarts-types<br>• réduction de l'accord avec l'idée que le VBM entraînera une fraude généralisée : 0,18 écarts-types |
| | | | | | | • réduction moyenne des évaluations de Trump de 0,10 écart-type.<br>• diminution de 2,9 points dans la note de température des sentiments envers Trump<br>• augmentation des attitudes défavorables envers Fox News : 0,09 écarts-types.<br>• désaccord accru avec l'affirmation "Si Trump faisait quelque chose de mal, Fox News en parlerait" : 0,20 écarts-types.<br>• aucun effet significatif et cohérent n'a été trouvé sur les indices mesurant les attitudes générales envers les médias, la confiance en CNN, la confiance en FoxNews et les émotions à court terme.<br>• légère augmentation de la consommation de CNN après la période incitative.<br>• certains effets à long terme sur les attitudes et croyances, mais non significatifs après ajustement pour comparaisons multiples. |
| Nyhan et al., 2023 | Facebook : 231M<br><br>Questionnaire : 23 377 | États-Unis | Juin- Septembre 2020<br><br>traitement : Septembre - Décembre 2020 | diminution de l'exposition à des contenus venant de sources d'un camp politique aligné | • exposition<br>• attitude politique | • diminution de l'exposition à des contenus venant de sources pro-attitudinales (de 53,7 % à 36,2 %) et à des contenus incivils (de 3,15 % à 2,81 %), comportant des insultes (de 0,034 % à 0,030 %) et à des *fake news* (de 0,76 % à 0,55 %).<br>• augmentation légère de l'exposition à des contenus venant de sources variées (de 20,7 % à 27,9 %).<br>• diminution de l'engagement total avec des contenus pro-attitudinales de 0,24 à 0,12 écart-type<br>• augmentation de l'engagement passif et actif avec des sources transversales de 0,11 et 0,04 écart-type<br>• pas d'effet sur les attitudes politiques (e.g. polarisation affective, extrémisme idéologique et attitudes envers les candidats et le système électoral)<br>• pas de différence selon l'idéologie politique, la sophistication politique, la littératie numérique ou l'exposition préalable à des contenus similaires. |
| Guess et al., 2023a | Facebook : 2 391<br><br>Instagram : 21 373 | États-Unis | Août-Décembre 2020 | fil d'actualité classé par ordre chronologique inversé au lieu du classement algorithmique par défaut | • exposition<br>• attitude politique | • diminution de l'activité et du temps passé sur la plateforme<br>• augmentation de l'exposition à des contenus politiques et peu fiables<br>• diminution de l'exposition à des contenus qualifié d'incivil ou contenant des termes injurieux |
| | | | | | | • augmentation de l'exposition à des contenus provenant d'amis modérés et de sources avec des audiences idéologiquement mixtes<br>• pas d'effet sur les attitudes politiques |
| Guess et al., 2023b | Facebook : 3 781<br><br>Questionnaire : 23 402 | États-Unis | Septembre-Décembre 2020 | suppression de l'exposition à des contenus partagés (et non directement publiés) par d'autres utilisateurs | • exposition<br>• engagement<br>• attitude politique | • diminution de l'exposition à des contenus politiques, y compris de ceux provenant de sources non fiables<br>• diminution du nombre total de clics et de réactions<br>• diminution du nombre de clics sur les actualités partisanes<br>• diminution des connaissances sur l'actualité<br>• pas d'effet sur les attitudes politiques |
| González-Bailón et al., 2023 | 208M | États-Unis | 1er Septembre - 1er Février 2021 | filtre exercé par l'algorithme de Facebook | • exposition<br>• engagement | • indice de ségrégation de 0,35 (au niveau des noms de domaine) et de 0,45-0,5 (au niveau des URLs)<br>• les audiences conservatrices sont plus isolées dans leur consommation de nouvelles politiques que les audiences libérales<br>• la plupart des sources de désinformation sont préférées par les audiences conservatrices, avec une homogénéité plus marquée dans les Pages et les Groupes qui partagent de la désinformation.<br>• la polarisation idéologique augmente à mesure qu'on descend dans l'entonnoir d'engagement, indiquant que les algorithmes et l'amplification sociale contribuent à une ségrégation accrue entre les audiences libérales et conservatrices.<br>• le score de ségrégation sur Facebook est trois fois plus élevé que celui observé sur des sites web pour le même ensemble d'individus, soulignant l'impact unique de la plateforme sur la ségrégation des audiences. |
| Ash et al., 2023 | 1 480 | États-Unis | 1er Février - 30 Avril 2020 | couverture médiatique de la chaîne Fox News (FNC) durant les premiers mois de la pandémie de COVID-19 | • comportements de distanciation sociale<br>• habitudes d'achat de produits liés au COVID-19<br>• opinions et attitudes par rapport au Covid-19 | • une diminution d'un écart-type dans la position de la chaîne FNC (indiquant une audience plus élevée) a augmenté le temps quotidien passé à l'extérieur de 3,5 minutes, la distance quotidienne parcourue de 0,3 km et a diminué les dépenses hebdomadaires pour les produits liés au COVID-19 de 106 $.<br>• les spectateurs de FNC :<br>- ont été plus lents à adopter des changements de comportement en réponse au COVID-19 par rapport aux spectateurs d'autres chaînes d'information ; |
| | | | | | | - ont été plus susceptibles de croire en l'efficacité de l'hydroxychloroquine comme traitement du COVID-19, soutenue par le président Trump ;<br>- ont privilégié les considérations économiques par rapport aux préoccupations de santé publique pendant la pandémie. |
| Allen et al., 2024 | Facebook : 233M<br><br>N1 : 8 603<br><br>N2 :  10 122 | États-Unis | Traces numériques :<br>Janvier - Mars 2021<br><br>Expérience 1 :<br>17 Mars - 22 Mai 2022.<br><br>Expérience 2 :<br>14 Juillet - 3 août 2022 | exposition à 13 206 URLS portant sur les vaccins | • intention vaccinale | • L'exposition à une seule information erronée sur les vaccins a diminué les intentions de vaccination de 1,5 point en moyenne.<br>• En tenant compte du nombre de vues des contenus, ceux sceptiques à l'égard des vaccins non fact-checkés ont réduit les taux de vaccination de −2,28 points comparé à −0,05 point pour les *fake news*<br>• L'impact des *fake news* était 50 fois moins important que celui de contenus sceptiques à l'égard des vaccins mais non qualifiés de *fake news* par des *fact-checkers* |

# Annexe 3. Rôle de l’hyperactivité des comptes dans le partage de *fake news*

Étant donné que les comptes de l’échantillon de relayeurs des *fake news* publient beaucoup plus de tweets que ceux du groupe de contrôle, nous avons cherché à vérifier si leur hyperactivité expliquait à elle seule le fait qu’ils aient partagé au moins une *fake news.* Pour ce faire, nous avons analysé l'activité sur deux ans des 3 661 690 comptes ayant partagé au moins un lien provenant d'une liste de 420 médias français. La distribution cumulative de l'activité montre que les utilisateurs qui partagent des *fake news* sont effectivement beaucoup plus actifs que les autres.

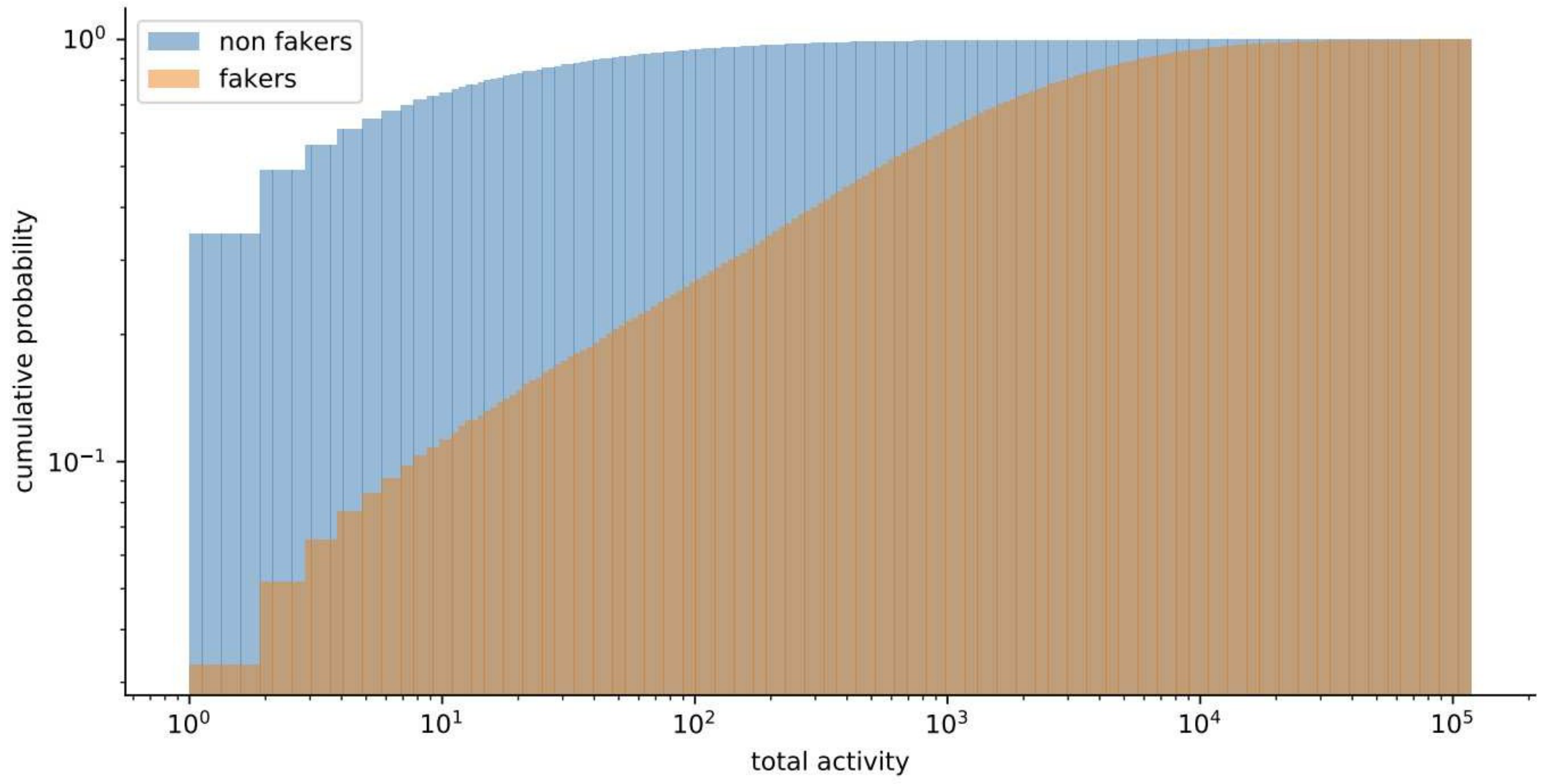

Nous avons ensuite développé un modèle pour estimer combien de *fake new*s chaque compte partagerait, en tenant compte de (1) la distribution du nombre de tweets produits par ces 3,7 millions de comptes sur deux ans, et (2) la probabilité de partager une *fake news* (calculée comme le ratio entre le nombre total de tweets contenant des *fake news* et l'activité totale de ces comptes). Ce modèle permet de déterminer si le partage de *fake news* est simplement dû à l'hyperactivité des utilisateurs, ou si cela relève d’un comportement spécifique concentré chez certains individus.

Enfin, nous avons comparé la distribution observée du nombre de *fake news* partagées par 31 760 utilisateurs à la distribution attendue, en répartissant les 87 364 *fake news* parmi les 184 millions de tweets produits par ces 3,7 millions d'utilisateurs, tout en tenant compte de leur activité réelle. La simulation a montré que le nombre d'utilisateurs partageant des *fake news* est plus élevé que prévu, bien que ces utilisateurs partagent en moyenne moins de *fake news* par compte. Les deux distributions sont proches, mais le test statistique ($\chi^2$ = 14 663, valeur de p = 0,0) révèle une différence significative.

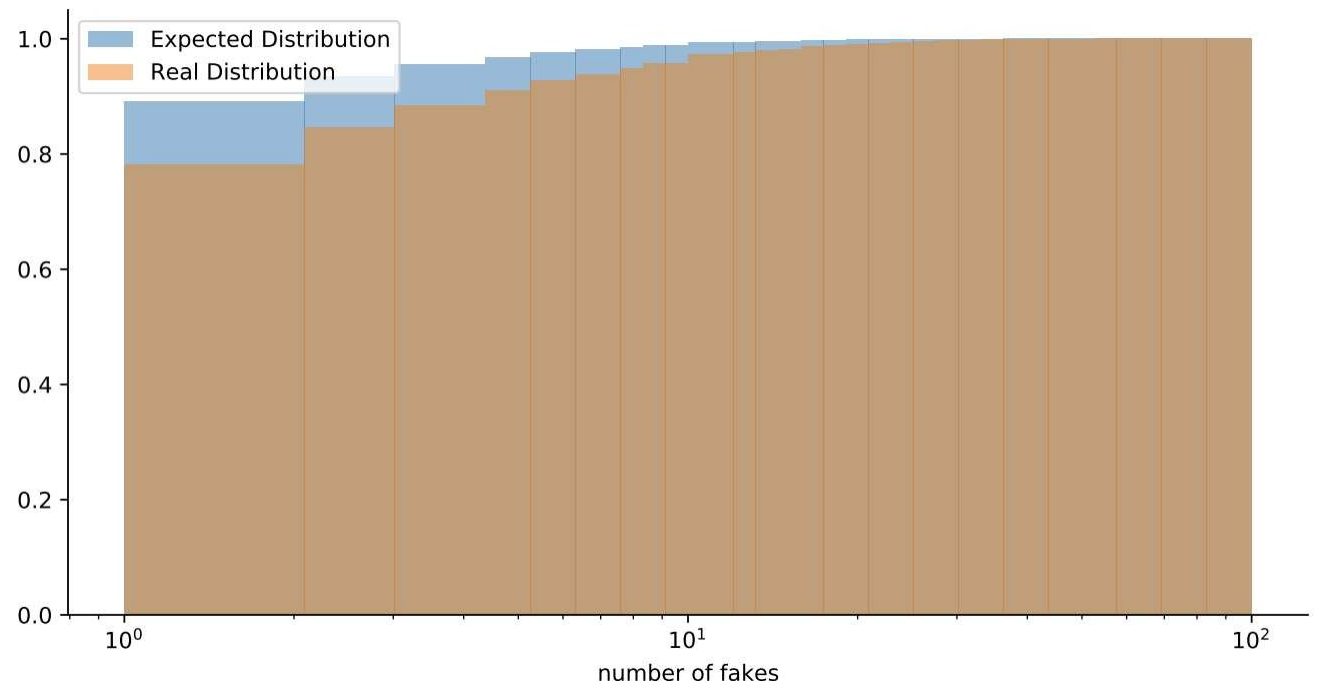

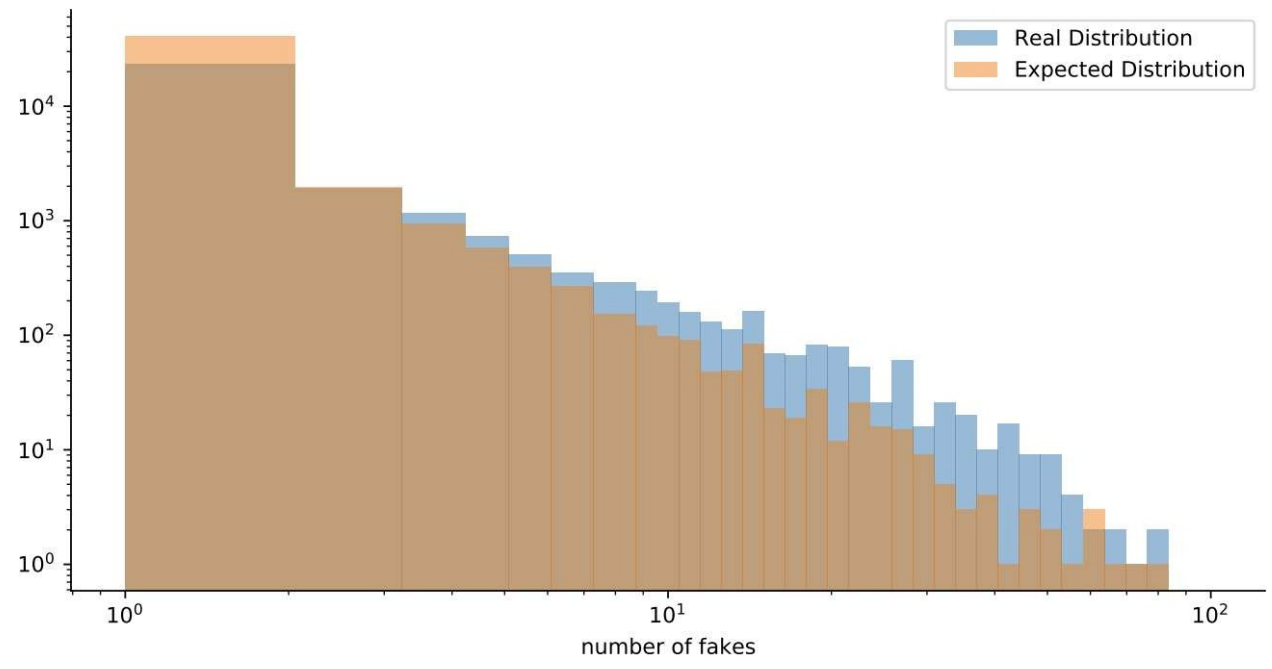

Il ressort donc que les utilisateurs qui partagent des *fake news* sont beaucoup plus actifs que ceux qui n'en partagent pas, mais cette hyperactivité ne suffit pas à elle seule à expliquer ce comportement.

## Annexe 4. Résultats du Botometer

En raison de la très forte activité des relayeurs de *fake news* sur Twitter, nous avons vérifié que les comptes de notre échantillon n'étaient pas des bots en utilisant le Botometer, un outil de détection de bots bien établi dans la littérature académique qui utilise diverses mesures pour déterminer la probabilité qu'un compte Twitter soit géré par un bot plutôt que par un être humain (Davis et al., 2016 ; Varol et al., 2017 ; Bessi & Ferrara, 2016 ; Wojcik et al., 2018 ; Keller & Klinger, 2019). Ces mesures incluent l'activité de publication du compte, le contenu des tweets et des caractéristiques du réseau telles que le rapport entre les abonnés et les amis. Le Botometer attribue un score entre 0 et 1, les scores plus élevés indiquant une probabilité plus élevée que le compte soit un bot. Les chercheurs définissent généralement un seuil manuel variant de 0,25 (Zhang et al., 2019) à 0,76 (Keller and Klinger, 2019) pour classer les comptes de manière binaire. Bien qu'il n'y ait pas de consensus sur le meilleur seuil à choisir dans la littérature scientifique, la plupart des articles en sélectionnent un oscillant entre 0,4 et 0,5 (Wojcik et al., 2018). Pour valider les performances du Botometer, nous avons examiné manuellement tous les profils d'utilisateurs avec un score supérieur à 0,4 (n=272). Parmi ces utilisateurs, seuls 11 ont été soupçonnés d'être des bots. Dans la mesure où le Botometer peut être imprécis dans la détection des bots et entraîner des faux positifs (Rauchfleisch & Kaiser, 2020 ; He et al., 2021), nous avons décidé de ne pas exclure ces utilisateurs de notre échantillon.

## Annexe 5. Grille d'annotations des noms de domaine des URLs signalées

| Catégorie | Caractéristiques | Exemples |
|---|---|---|
| **Divertissement** | Cette catégorie inclut les médias dont le principal objectif est de divertir le public. Le contenu est généralement centré sur le spectacle, la culture populaire, la musique, les films, la télévision, les célébrités, ou d'autres formes de loisir. Bien que certains articles puissent aborder l'actualité, le ton est souvent léger et le but est d'attirer l'attention ou d'amuser. | momentbuzz.com ; waouh.com ; topibuzz.com |
| **Contre-information** | Ces sources se concentrent sur des récits alternatifs à ceux proposés par les médias traditionnels. Elles sont souvent critiques des institutions établies et des discours dominants, et cherchent à remettre en question ou à « révéler » des informations censurées ou ignorées par les médias *mainstream*. | lesmoutonsrebelles.com ; dreuz.info ; lagauchematuer.fr |
| **Santé alternative** | Médias qui promeuvent des approches non conventionnelles de la santé, souvent en opposition aux pratiques médicales traditionnelles ou scientifiques. Ils peuvent couvrir des sujets tels que les médecines douces, les traitements naturels, ou les thérapies alternatives. Ils tendent à valoriser la méfiance vis-à-vis des autorités médicales et à favoriser des approches dites « naturelles » | sain-et-naturel.com ; sante-nutrition.org ; professeur-joyeux.com |
| **Partisan/ Engagé** | Ces médias affichent une orientation politique ou idéologique claire et soutiennent activement une cause, un parti, ou un mouvement social. Leur couverture de l'actualité est souvent orientée pour défendre ou promouvoir une vision spécifique, et ils peuvent produire des contenus militants ou activistes. | atlantico.fr ; valeursactuelles.com ; marianne.net |
| **Parodique** | Ce sont des sites ou des médias qui créent des contenus humoristiques ou satiriques. Bien que leurs articles prennent souvent la forme de l'actualité, ils ne visent pas à informer, mais à tourner en dérision les événements ou personnalités publiques. Les contenus sont souvent fictifs ou exagérés de manière absurde pour provoquer le rire. | bmf-news.com ; legorafi.fr ; scienceinfo.fr |
| **Média *mainstream*** | Ce sont des médias grand public, largement diffusés et reconnus pour leur couverture de l'actualité nationale ou internationale. Ils suivent généralement des pratiques journalistiques établies et visent une large audience. | ouest-france.fr ; lefigaro.fr ; 20minutes.fr |
| **Presse Locale** | Médias qui se concentrent sur l'actualité et les événements à une échelle géographique réduite, souvent une ville, une région ou un département. Ils couvrent les sujets d'intérêt local, allant des informations politiques locales aux événements communautaires. Leur lectorat est généralement composé de résidents locaux. | ledauphine.com ; nicematin.com ; leprogres.fr |

# Annexes 6. Grille d'annotations des thématiques des URLs signalées

| Catégorie | Sujets possibles | Exemples |
|---|---|---|
| **Religion** | débats sur la laïcité, conflits religieux, scandales au sein des institutions religieuses, célébrations et fêtes religieuses, spiritualité, montée des fondamentalismes | « Chambéry : La mairie fait avancer le feu d'artifice du 14 juillet au 24 juin pour fêter la fin du ramadan » |
| **Cause sociale/ inclusion** | débats sur les discriminations raciales ou sociales, discussions sur les droits des minorités, campagnes pour l'égalité femme-homme | « Paris : des manifestants de la Marche des fiertés dégradent la statue de Jeanne d'Arc » |
| **Immigration** | débats sur les politiques d'immigration, conditions d'accueil des réfugiés, intégration des migrants | « Norvège : expulsion des musulmans délinquants, 31% de baisse des crimes violents en un an ! \| Résistance Républicaine » |
| **Santé/ science/ bien-être** | Témoignages sur les effets secondaires de traitements médicaux, controverses sur les nouvelles découvertes scientifiques, débats autour de la vaccination et de la santé publique, conseils sur le bien-être, nouvelles sur des découvertes médicales ou innovations technologiques, discussions sur les théories scientifiques alternatives. | « Les médecins qui ont découvert les enzymes du cancer dans les vaccins ont tous été assassinés » |
| **Politique** | Élections, politiques publiques, vie des partis politiques, discours des élus, affaires gouvernementales, réformes législatives, tensions internationales, campagnes électorales, enjeux politiques locaux et internationaux | « Un plan secret avait été mis en place si Marine Le Pen accédait au pouvoir... Voilà ce qui se serait passé ! » |
| **Criminalité/ violence** | Reportages sur des affaires criminelles, témoignages de victimes de violences, documentaires sur des enquêtes policières, des procès ou des condamnations, histoires de gangs ou de criminalité organisée, histoires de récidives ou d'affaires non résolues. | « FLASH - À Calais, trois personnes en garde à vue pour viols avec torture » |
| **Économie/ finance** | Chômage, inflation, croissance économique, crises financières, scandale financier | « L'État français va injecter 10,5 milliards d'euros dans six grandes banques » |
| **Environnement/ nature** | Reportages sur des catastrophes naturelles, débats sur le changement climatique, nouvelles sur des initiatives de préservation environnementale, discussions sur l'impact des industries sur l'environnement, | « Climat : ne rien faire coutera bien plus cher que la transition écologique ! » |
| **Faits divers** | Incidents inhabituels, accidents de la vie quotidienne, phénomènes inexpliqués, histoires insolites ou choquantes, disparitions, | « Une jeune baby-sitter a défendu un garçon de 7 ans à main nue contre un renard agressif » |
| | découvertes macabres, drames familiaux, crimes atypiques. | |
| **Divertissement/ culture** | Célébrités, cinéma, musique, mode, jeux vidéo, sports | « Ballon d'Or : et de cinq pour Cristiano Ronaldo ! » |
| **Société** | Débats sur l'évolution des normes sociales et des modes de vie, discussions sur les rapports entre générations, l'éducation, la famille, points de vue sur les nouvelles technologies et leur impact sur la vie quotidienne. | « Au Danemark, les cours d'empathie sont obligatoires dans les écoles depuis 1993 » |
| **Animaux** | Droits des animaux, protection des espèces en danger, maltraitance animale, adoption d'animaux, élevage industriel, bien-être animal domestique. | « Marineland d'Antibes : polémique autour de l'état de santé d'un orque de 20 ans » |

## Annexes 7. Grille d'annotations des pages et des groupes Facebook ayant partagé des URLs signalées

| Catégorie | Caractéristiques | Exemples |
|---|---|---|
| **actualités diverses** | Informations générales couvrant un large éventail de sujets tels que la politique, l'économie, la culture, les événements mondiaux et locaux | France : DÉBATS sur la Politique ; Actualité du jour ; #DEBAT |
| **animaux** | Contenu centré sur les animaux, incluant des informations sur les soins, les droits des animaux, les histoires émouvantes, les vidéos et photos amusantes, ainsi que des conseils pour les propriétaires d'animaux de compagnie. | J'aime les chats ; Aider les animaux tous ensemble ; Les Droits des Animaux |
| **anti-système** | Contenu critiquant les institutions établies, y compris le gouvernement, les grandes entreprises, et les médias traditionnels. Souvent lié à des théories du complot et à une méfiance envers les autorités. | Pour la démission d'Emmanuel Macron ; Ce que vous Cachent les Médias ; Contre L'Elite Mondiale qui nous Manipule |
| **centre** | Contenu politique modéré, prônant des solutions équilibrées et consensuelles. Souvent critique à l'égard des extrêmes politiques et cherchant à représenter une perspective raisonnable et pragmatique. | Citoyens En Marche ! ; Ensemble avec Emmanuel Macron ; Sympathisants et militants de EM (En Marche) |
| **divertissement** | Contenu destiné à divertir, incluant des vidéos humoristiques, des mèmes, des nouvelles sur les célébrités, des critiques de films et de séries, ainsi que des jeux en ligne et des quiz. | Oh my mag ; 100% Insolites et Virales ; Top Astuces |
| **droite** | Contenu politique aligné avec les idées et les valeurs de la droite, incluant des discussions sur l'économie de marché, la sécurité, les traditions et souvent une approche conservatrice des politiques sociales. | Tous à Droite ; Laurent Wauquiez : « Pour moi, le modèle, c'est Nicolas Sarkozy » ; Les Républicains Bourg |
| **éducation** | Contenu axé sur l'apprentissage et l'enseignement, incluant des ressources pédagogiques, des conseils pour les étudiants et les enseignants, des discussions sur les politiques éducatives, et des innovations dans le domaine de l'éducation. | SOS Éducation ; Tu sais que tu es Educateur de Jeunes Enfants quand... ; Les profs réunis |
| **extrême droite** | Contenu politique radical aligné avec les idées de l'extrême droite, incluant souvent des discours nationalistes, xénophobes, et un fort accent sur la sécurité et l'ordre. | Sympathisants Debout la France ; Florian Philippot ; OBJECTIF 2022 EN BLEU MARINE |
| **extrême gauche** | Contenu politique radical aligné avec les idées de l'extrême gauche, prônant des changements sociaux et économiques radicaux, l'anti-capitalisme, et souvent une critique sévère des inégalités et des injustices sociales. | INSOUMIS ; MAA - Mouvement Anarchiste Anticapitaliste ; JLM LA VOIX DU PEUPLE INSOUMIS |
| **gauche** | Contenu politique aligné avec les idées et les valeurs de la gauche, incluant des discussions sur la justice sociale, l'égalité économique, les droits des minorités, et des politiques progressistes. | Peuple de gauche ; #GROUPE LES TULIPES ; BENOIT HAMON |
| **gilets jaunes** | Contenu lié au mouvement des Gilets Jaunes en France, incluant des revendications sociales et économiques, des témoignages de manifestants, et des critiques des politiques gouvernementales perçues comme injustes. | Carte des rassemblements ; GILET JAUNE NON CENSURE ; Gilets Jaunes le Mouvement |
| **humour** | Contenu humoristique destiné à faire rire, incluant des blagues, des mèmes, des vidéos drôles, et des anecdotes amusantes. Souvent léger et conçu pour divertir un large public. | SecretNews ; Le Gorafi ; Nordpresse |
| **média alternatif** | Contenu produit par des sources médiatiques non traditionnelles, souvent avec une perspective critique envers les médias *mainstream*. | Dreuz.info ; WikiStrike Officiel ; La Tribune des Pirates |
| **média *mainstream*** | Contenu produit par les grandes entreprises médiatiques traditionnelles, couvrant un large éventail de sujets d'actualité avec une approche journalistique professionnelle et un large auditoire. | Ouest France ; Le Monde ; Le Figaro |
| **religion** | Contenu lié aux pratiques, croyances, et événements religieux. Peut inclure des discussions théologiques, des conseils spirituels, des informations sur les fêtes religieuses, et des témoignages de foi. | Entre frères et soeurs avançons dans la Religion ; Chrétiens de France venez rejoindre le groupe ; Le Monde Juif |
| **santé/ sciences/ bien-être** | Contenu axé sur la santé physique et mentale, les avancées scientifiques, les conseils de bien-être, et les pratiques de vie saine. Peut inclure des articles sur la nutrition, l'exercice, les traitements médicaux, et la recherche scientifique. | Alimentation Danger ; Allo Docteurs - Le Mag de la Santé ; Astuces minceur |
| **vie locale** | Contenu centré sur les événements, les nouvelles, et les discussions pertinentes à une communauté ou une région spécifique. Inclut souvent des informations sur les services locaux, les événements communautaires, et les préoccupations locales. | Tu es de Rians si… ; Entre Montpellierains ; Tu es d'Aix en Pce si … |

# Annexes 8. Taux de points d’arrêt selon les territoires Youtube

| Community_label | No_Stop | Stop | Taux de point d'arrêt |
|---|---|---|---|
| **Science et Histoire** | 130932 | 16576 | 12,66% |
| **Armée et Stratégie** | 24208 | 2148 | 8,87% |
| **Médias Centraux** | 1423756 | 115500 | 8,11% |
| **Contenus à clic** | 148633 | 11783 | 7,93% |
| **Droite Masculiniste** | 344724 | 26443 | 7,67% |
| **Tech** | 11423 | 841 | 7,36% |
| **Médias de Gauche** | 279826 | 19652 | 7,02% |
| **Identitaires** | 9133 | 616 | 6,74% |
| **Culture et Pop culture** | 370836 | 21352 | 5,76% |
| **Droite Confusionniste** | 448946 | 24368 | 5,43% |
| **Magazines Féminins** | 24237 | 977 | 4,03% |